%% file: main.tex
\documentclass[11pt,oneside,phd]{style/dms}

\usepackage[utf8]{inputenc} %
\usepackage[T1]{fontenc}    %

\usepackage{lmodern}

\anglais

\def\sloppy{%
  \tolerance 500%
  \emergencystretch 3em%
  \hfuzz .5pt
  \vfuzz\hfuzz}
\usepackage{graphicx,amssymb,icomma}

\usepackage[labelfont=bf, width=\linewidth]{caption}
\usepackage{subcaption}

\usepackage[onehalfspacing]{setspace}

\usepackage{imakeidx}  %
\usepackage{hyperref}  %
\usepackage{hypcap}   %
\usepackage{bookmark} %
\usepackage{glossaries}

\usepackage[american]{babel}
\usepackage{csquotes}

\usepackage[dvipsnames]{xcolor}
\usepackage{microtype}
\usepackage{graphicx}
\usepackage{multirow}
\usepackage{dsfont}
\usepackage{algorithm}
\usepackage[noEnd=true, indLines=true]{algpseudocodex}
\algrenewcommand\algorithmicrequire{\textbf{Inputs:}}
\algrenewcommand\algorithmicensure{\textbf{Outputs:}}

\usepackage{mathtools} %
\usepackage{booktabs} %
\usepackage{tikz} %

\usepackage[
    style=authoryear,
    natbib=true,
    backend=biber,
    hyperref=true,
    backref=true,
    sorting=nyt,
    maxnames=1000,
    maxcitenames=2,
    uniquename=false,
    uniquelist=false,
    dashed=false,
]{biblatex}
\bibliography{references/references,references/gflownet,references/gflownet_tmp}
\DeclareNameAlias{sortname}{given-family}
\DeclareFieldFormat[article,inbook,incollection,inproceedings,patent,thesis,unpublished]{citetitle}{#1\isdot}
\DeclareFieldFormat[article,inbook,incollection,inproceedings,patent,thesis,unpublished]{title}{#1\isdot}
\DefineBibliographyStrings{english}{
  backrefpage={cited on p.},
  backrefpages={cited on pp.}
}
\renewbibmacro{in:}{}

\renewbibmacro*{cite}{%
  \printtext[bibhyperref]{%
    \iffieldundef{shorthand}
      {\ifthenelse{\ifnameundef{labelname}\OR\iffieldundef{labelyear}}
         {\usebibmacro{cite:label}%
          \setunit{\printdelim{nonameyeardelim}}}
         {\printnames{labelname}%
          \setunit{\printdelim{nameyeardelim}}}%
       \usebibmacro{cite:labeldate+extradate}}
      {\usebibmacro{cite:shorthand}}}}

\renewbibmacro*{citeyear}{%
  \printtext[bibhyperref]{%
    \iffieldundef{shorthand}
      {\iffieldundef{labelyear}
         {\usebibmacro{cite:label}}
         {\usebibmacro{cite:labeldate+extradate}}}
      {\usebibmacro{cite:shorthand}}}}

\renewbibmacro*{textcite}{%
  \ifnameundef{labelname}
    {\iffieldundef{shorthand}
       {\printtext[bibhyperref]{%
          \usebibmacro{cite:label}}%
        \setunit{%
          \global\booltrue{cbx:parens}%
          \printdelim{nonameyeardelim}\bibopenparen}%
        \ifnumequal{\value{citecount}}{1}
          {\usebibmacro{prenote}}
          {}%
        \printtext[bibhyperref]{\usebibmacro{cite:labeldate+extradate}}}
       {\printtext[bibhyperref]{\usebibmacro{cite:shorthand}}}}
    {\printtext[bibhyperref]{\printnames{labelname}}%
     \setunit{%
       \global\booltrue{cbx:parens}%
       \printdelim{nameyeardelim}\bibopenparen}%
     \ifnumequal{\value{citecount}}{1}
       {\usebibmacro{prenote}}
       {}%
     \usebibmacro{citeyear}}}

\renewbibmacro*{cite:shorthand}{%
  \printfield{shorthand}}

\renewbibmacro*{cite:label}{%
  \iffieldundef{label}
    {\printfield[citetitle]{labeltitle}}
    {\printfield{label}}}

\renewbibmacro*{cite:labeldate+extradate}{%
  \printlabeldateextra}

\input{math_commands.tex}

\usepackage{hyperref}
\usepackage{url}
\usepackage{mathtools}
\usepackage{amsthm}
\usepackage{lipsum}
\usepackage{thmtools}
\usepackage{thm-restate}
\usepackage{adjustbox}
\usepackage{enumitem}
\usepackage{parskip}
\setlist[itemize]{parsep=0pt}
\setlist[enumerate]{parsep=0.5\parsep}

\usepackage[mode=buildnew]{standalone}
\usepackage{tikz}
\usepackage{pgfplots}
\pgfplotsset{compat=1.18}
\usetikzlibrary{backgrounds, scopes}
\usetikzlibrary{shapes.geometric}

\usepackage[capitalize,noabbrev]{cleveref}

\indexsetup{firstpagestyle=empty}
\makeindex

\definecolor{mydarkblue}{rgb}{0,0.08,0.45}
\hypersetup{ %
    pdfdisplaydoctitle,
    pdftitle={Generative Flow Networks: Theory and Applications to Structure Learning},
    pdfauthor={Tristan Deleu},
    pdfcreator={Tristan Deleu},
    pdfkeywords={Generative flow networks, Bayesian inference, Structure learning, Bayesian networks, Reinforcement learning, Variational inference},
    pdfborder=0 0 0,
    pdfpagemode=UseNone,
    colorlinks=true,
    linkcolor=mydarkblue,
    citecolor=mydarkblue,
    filecolor=mydarkblue,
    urlcolor=mydarkblue,
}

\usepackage[most]{tcolorbox}
\tcbuselibrary{breakable}

\theoremstyle{plain}
\newtheorem{theorem}{Theorem}[section]
\tcolorboxenvironment{theorem}{
  breakable=true,
  oversize,
  enhanced jigsaw,
  sharp corners,
  boxrule=0pt,
  colback=Red!5!white,
  colframe=Red!5!white,
  borderline west={4pt}{0pt}{Red!75!black}
}

\newtheorem{proposition}[theorem]{Proposition}
\tcolorboxenvironment{proposition}{
  breakable=true,
  oversize,
  enhanced jigsaw,
  sharp corners,
  boxrule=0pt,
  colback=RedOrange!5!white,
  colframe=RedOrange!5!white,
  borderline west={4pt}{0pt}{RedOrange!75!black},
}

\newtheorem*{proposition611}{Proposition 6.1.1}
\tcolorboxenvironment{proposition611}{
  breakable=true,
  oversize,
  enhanced jigsaw,
  sharp corners,
  boxrule=0pt,
  colback=RedOrange!5!white,
  colframe=RedOrange!5!white,
  borderline west={4pt}{0pt}{RedOrange!75!black},
}

\newtheorem{lemma}[theorem]{Lemma}
\tcolorboxenvironment{lemma}{
  breakable=true,
  oversize,
  enhanced jigsaw,
  sharp corners,
  boxrule=0pt,
  colback=NavyBlue!5!white,
  colframe=NavyBlue!5!white,
  borderline west={4pt}{0pt}{NavyBlue!75!black}
}

\newtheorem{corollary}[theorem]{Corollary}
\tcolorboxenvironment{corollary}{
  breakable=true,
  oversize,
  enhanced jigsaw,
  sharp corners,
  boxrule=0pt,
  colback=YellowOrange!5!white,
  colframe=YellowOrange!5!white,
  borderline west={4pt}{0pt}{YellowOrange!75!black}
}

\theoremstyle{definition}
\newtheorem{definition}[theorem]{Definition}
\tcolorboxenvironment{definition}{
  breakable=true,
  oversize,
  enhanced jigsaw,
  sharp corners,
  boxrule=0pt,
  colback=Gray!5!white,
  colframe=Gray!5!white,
  borderline west={4pt}{0pt}{Gray!75!black},
  parbox=false
}

\newtheorem{example}[theorem]{Example}
\tcolorboxenvironment{example}{
  breakable=true,
  oversize,
  enhanced jigsaw,
  sharp corners,
  boxrule=0pt,
  colback=ForestGreen!5!white,
  colframe=ForestGreen!5!white,
  borderline west={4pt}{0pt}{ForestGreen!75!black}
}

\theoremstyle{remark}

\tcolorboxenvironment{proof}{
  breakable=true,
  oversize,
  enhanced jigsaw,
  sharp corners,
  interior hidden,
  frame hidden,
  boxrule=0pt,
  borderline west={4pt}{0pt}{gray!10},
  parbox=false,
  before upper=\vspace*{-2em},
}

\crefformat{chapter}{#2Chapter~#1#3}
\crefmultiformat{chapter}{#2Chapters~#1#3}%
{ \&~#2#1#3}{, #2#1#3}{, \&~#2#1#3}
\crefformat{section}{#2Section~#1#3}
\crefmultiformat{section}{#2Sections~#1#3}%
{ \&~#2#1#3}{, #2#1#3}{, \&~#2#1#3}
\crefformat{appendix}{#2Appendix~#1#3}

\crefformat{equation}{(#2#1#3)}
\crefmultiformat{equation}{#2Equations~#1#3}%
{ \&~#2#1#3}{, #2#1#3}{, \&~#2#1#3}

\crefformat{table}{#2Table~#1#3}
\crefformat{figure}{#2Figure~#1#3}
\crefmultiformat{figure}{#2Figures~#1#3}%
{ \&~#2#1#3}{, #2#1#3}{, \&~#2#1#3}

\crefformat{theorem}{#2Theorem~#1#3}
\crefmultiformat{theorem}{#2Theorems~#1#3}%
{ \&~#2#1#3}{, #2#1#3}{, and~(#2#1#3)}

\crefformat{lemma}{#2Lemma~#1#3}
\crefformat{proposition}{#2Proposition~#1#3}
\crefmultiformat{proposition}{#2Propositions~#1#3}%
{ \&~#2#1#3}{, #2#1#3}{, and~(#2#1#3)}

\crefformat{algorithm}{#2Algorithm~#1#3}
\crefmultiformat{algorithm}{#2Algorithms~#1#3}%
{ \&~#2#1#3}{, #2#1#3}{, and~(#2#1#3)}

\crefformat{corollary}{#2Corollary~#1#3}
\crefformat{definition}{#2Definition~#1#3}
\crefformat{remark}{#2Remark~#1#3}
\crefformat{example}{#2Example~#1#3}

\crefformat{assumption}{#2Assumption~#1#3}
\crefmultiformat{assumption}{#2Assumptions~#1#3}%
{ \&~#2#1#3}{, #2#1#3}{, and~(#2#1#3)}

\DeclareCiteCommand{\citeyear}
    {}
    {\bibhyperref{\printfield{year}}\bibhyperref{\printfield{extrayear}}}
    {\multicitedelim}
    {}

\DeclareCiteCommand{\citeyearpar}
    {}
    {\mkbibparens{\bibhyperref{\printfield{year}}\bibhyperref{\printfield{extrayear}}}}
    {\multicitedelim}
    {}

\usepackage{pifont}
\newcommand{\cmark}{\ding{51}}%
\newcommand{\xmark}{\ding{55}}%
\newcommand{\ccomp}{\textcolor{Green}{\cmark}}
\newcommand{\xcomp}{\textcolor{Red}{\xmark}}
\newcommand{\qcomp}{\tikz[baseline=-0.75ex]\fill[Gray] (0,0) circle (.5ex);}

\newcommand{\ie}{\textit{i.e.},~}
\newcommand{\eg}{\textit{e.g.},~}
\newcommand{\terminal}{\bot}
\newcommand{\stopaction}{\mathrm{stop}}
\newcommand{\parents}{\mathrm{Pa}}
\newcommand{\children}{\mathrm{Ch}}
\newcommand{\markov}{\mathrm{Markov}}
\renewcommand{\KL}{\mathrm{KL}}

\newcommand\independent{\protect\mathpalette{\protect\independenT}{\perp}}
\def\independenT#1#2{\mathrel{\rlap{$#1#2$}\mkern2mu{#1#2}}}

\newcommand{\soft}{\mathrm{soft}}
\newcommand{\maxent}{\mathrm{MaxEnt}}
\newcommand{\fldb}{\mathrm{FL}\textrm{-}\mathrm{DB}}
\newcommand{\mdb}{\mathrm{M}\textrm{-}\mathrm{DB}}
\newcommand{\pisql}{\pi\textrm{-}\mathrm{SQL}}

\input{style/widebar}

\numberwithin{equation}{section}
\numberwithin{table}{chapter}
\numberwithin{figure}{chapter}

\makeglossaries
\input{prologue/glossaries}

\begin{document}
\frontmatter

\version{1}

\title{Generative Flow Networks:\\Theory and Applications to Structure Learning}
\author{Tristan Deleu}
\copyrightyear{2024}

\department{D\'{e}partement d'Informatique et de Recherche Op\'{e}rationnelle}

\date{August 31, 2024} %

\sujet{Informatique}

\president{Dhanya Sridhar}  %

\directeur{Yoshua Bengio}

\membrejury{Simon Lacoste-Julien}  %

\examinateur{Kevin P. Murphy}   %

\repdoyen{William J. McCausland} %

\maketitle

\maketitle

\francais
\input{prologue/abstract-fr}

\anglais
\input{prologue/abstract-en}

\anglais

\cleardoublepage
\pdfbookmark[chapter]{\contentsname}{toc}  %
\setlength{\parskip}{0pt}
\tableofcontents
\setlength{\parskip}{.5\baselineskip plus 2pt}
\addtocontents{toc}{\protect\setcounter{tocdepth}{-1}}  %
\cleardoublepage
\listoftables
\cleardoublepage
\listoffigures

\addtocontents{toc}{\protect\setcounter{tocdepth}{3}}  %

\chapter*{Glossary}
\renewcommand{\glossarysection}[2][]{}
\glsdisablehyper
\printglossary

\input{prologue/acknowledgements}

\NoChapterPageNumber
\cleardoublepage

\input{prologue/prologue}

\mainmatter
\addtocontents{toc}{\protect\setcounter{tocdepth}{0}}  %
\input{chapters/00_Introduction}
\addtocontents{toc}{\protect\setcounter{tocdepth}{3}}

\input{chapters/01_Background}

\part{Generative Flow Networks}
\label{part:gflownet}

\input{chapters/02_Probabilistic_Inference_Strctured}
\input{chapters/03_Flow_Networks}
\input{chapters/04_Generative_Flow_Networks}
\input{chapters/05_GFlowNets_MaxEnt_RL}
\input{chapters/06_GFlowNets_General_State_Spaces}

\addtocontents{toc}{\protect\newpage}  %
\part{Bayesian Structure Learning}
\label{part:bayesian-structure-learning}

\input{chapters/07_DAG_GFlowNet}
\input{chapters/08_JSP_GFN}

\addtocontents{toc}{\protect\setcounter{tocdepth}{0}}  %
\input{chapters/09_Conclusion}

\addtocontents{toc}{\protect\setcounter{tocdepth}{3}}

\printindex

\def\bibname{References}
\printbibliography

\appendix

\part*{Appendices}
\label{part:appendices}

\addtocontents{toc}{\protect\setcounter{tocdepth}{1}}
\input{appendix/01_Generative_Flow_Networks}
\input{appendix/02_Bayesian_Structure_Learning}

\end{document}

%% file: math_commands.tex
\usepackage{amsmath,amsfonts,bm}

\def\eqref#1{equation~\ref{#1}}

\def\1{\bm{1}}

\def\vzero{{\bm{0}}}

\def\vb{{\bm{b}}}

\def\vg{{\bm{g}}}

\def\vp{{\bm{p}}}

\def\vu{{\bm{u}}}
\def\vv{{\bm{v}}}
\def\vw{{\bm{w}}}
\def\vx{{\bm{x}}}
\def\vy{{\bm{y}}}
\def\vz{{\bm{z}}}

\def\mA{{\bm{A}}}

\def\mF{{\bm{F}}}

\def\mI{{\bm{I}}}

\def\mM{{\bm{M}}}

\def\mP{{\bm{P}}}

\def\mR{{\bm{R}}}

\def\mX{{\bm{X}}}

\DeclareMathAlphabet{\mathsfit}{\encodingdefault}{\sfdefault}{m}{sl}
\SetMathAlphabet{\mathsfit}{bold}{\encodingdefault}{\sfdefault}{bx}{n}

\def\gA{{\mathcal{A}}}
\def\gB{{\mathcal{B}}}
\def\gC{{\mathcal{C}}}
\def\gD{{\mathcal{D}}}
\def\gE{{\mathcal{E}}}
\def\gF{{\mathcal{F}}}
\def\gG{{\mathcal{G}}}
\def\gH{{\mathcal{H}}}
\def\gI{{\mathcal{I}}}
\def\gJ{{\mathcal{J}}}

\def\gL{{\mathcal{L}}}
\def\gM{{\mathcal{M}}}
\def\gN{{\mathcal{N}}}

\def\gQ{{\mathcal{Q}}}

\def\gS{{\mathcal{S}}}
\def\gT{{\mathcal{T}}}
\def\gU{{\mathcal{U}}}

\def\gX{{\mathcal{X}}}
\def\gY{{\mathcal{Y}}}

\def\sN{{\mathbb{N}}}

\def\sP{{\mathbb{P}}}

\def\sR{{\mathbb{R}}}

\newcommand{\E}{\mathbb{E}}

\newcommand{\KL}{D_{\mathrm{KL}}}

\DeclareMathOperator*{\argmax}{arg\,max}
\DeclareMathOperator*{\argmin}{arg\,min}

\DeclareMathOperator{\Tr}{Tr}

%% file: prologue/glossaries.tex
\newglossaryentry{structuredag}
{
  name={$G$},
  description={Directed acyclic graph structure of a Bayesian network}
}

\newglossaryentry{structureparams}{
  name={\ensuremath{\theta}},
  description={Parameters of the conditional probability distributions of a Bayesian network}
}

\newglossaryentry{numnodes}{
  name={$d$},
  description={Number of random variables in a Bayesian network}
}

\newglossaryentry{randomvars}{
  name={$X_{1}, \ldots, X_{d}$},
  description={Random variables in a Bayesian network}
}

\newglossaryentry{conditionalindependence}{
  name={$X \independent Y \mid Z$},
  description={Random variable $X$ is independent of $Y$ conditioned on $Z$}
}

\newglossaryentry{dag}{
  name={DAG},
  description={Directed acyclic graph}
}

\newglossaryentry{cpdag}{
  name={CPDAG},
  description={Completed partially directed acyclic graph}
}

\newglossaryentry{mec}{
  name={MEC},
  plural={Markov equivalence classes},
  description={Markov equivalence class}
}

\newglossaryentry{dataset}{
  name={\ensuremath{\gD}},
  description={Dataset of observations}
}

\newglossaryentry{numsamples}{
  name={\ensuremath{N}},
  description={Number of samples in the dataset $\gD$}
}

\newglossaryentry{em}{
  name={EM},
  description={Expectation maximization}
}

\newglossaryentry{elbo}{
  name={ELBO},
  description={Evidence lower-bound}
}

\newglossaryentry{iid}{
  name={iid.},
  description={Independent and identically distributed}
}

\newglossaryentry{ebm}{
  name={EBM},
  description={Energy-based model}
}

\newglossaryentry{mcmc}{
  name={MCMC},
  description={Markov chain Monte Carlo}
}

\newglossaryentry{samplespace}{
  name={$\gX$},
  description={Sample space of the target distribution / Set of terminating states}
}

\newglossaryentry{pointeddag}{
  name={$\gG = (\protect\widebar{\gS}, \gA)$},
  description={Pointed DAG with state space $\protect\widebar{\gS}$ and $\gA$ the set of edges}
}

\newglossaryentry{statespace}{
  name={$\gS$},
  description={State space}
}

\newglossaryentry{augmentedstatespace}{
  name={\ensuremath{\protect\widebar{\gS}}},
  description={Augmented state space $\protect\widebar{\gS} = \gS \cup \{\terminal\}$}
}

\newglossaryentry{initialstate}{
  name={$s_{0}$},
  description={Initial state}
}

\newglossaryentry{terminalstate}{
  name={$\terminal$},
  description={Terminal state}
}

\newglossaryentry{parents}{
  name={$\parents_{\gG}(s)$},
  description={Parents of a state $s$ in $\gG$}
}

\newglossaryentry{children}{
  name={$\children_{\gG}(s)$},
  description={Children of a state $s$ in $\gG$}
}

\newglossaryentry{partialtrajectories}{
  name={$s_{m} \rightsquigarrow s_{n}$},
  description={Set of (partial) trajectories from $s_{m}$ to $s_{n}$}
}

\newglossaryentry{completetrajectories}{
  name={$\gT$},
  description={Set of complete trajectories}
}

\newglossaryentry{completetrajectory}{
  name={$\tau = (s_{0}, s_{1}, \ldots, s_{T}, \terminal)$},
  description={A complete trajectory (with often the convention $s_{T+1} = \terminal$)}
}

\newglossaryentry{concatenatetrajectories}{
  name={$\tau \oplus \tau'$},
  description={Concatenation of the (partial) trajectory $\tau$ with the trajectory $\tau'$}
}

\newglossaryentry{simplexchildren}{
  name={$\Delta(\children_{\gG})$},
  description={Space of probability distributions over the children of some $s\in\gG$}
}

\newglossaryentry{forwardtransitionprobability}{
  name={\ensuremath{P_{F}}},
  description={Forward transition probability}
}

\newglossaryentry{terminatingstateprobability}{
  name={$P_{F}^{\top}$},
  description={Terminating state probability distribution (associated with $P_{F}$)}
}

\newglossaryentry{mdp}{
  name={MDP},
  description={Markov decision process}
}

\newglossaryentry{mdpmath}{
  name={$\gM = (\gG, r)$},
  description={Markov decision process with structure $\gG$ and reward function $r$}
}

\newglossaryentry{maxentrl}{
  name={MaxEnt RL},
  description={Maximum entropy reinforcement learning}
}

\newglossaryentry{rl}{
  name={RL},
  description={Reinforcement learning}
}

\newglossaryentry{sql}{
  name={SQL},
  description={Soft Q-Learning}
}

\newglossaryentry{numtrajectories}{
  name={\ensuremath{n(x)}},
  description={Number of complete trajectories leading to the terminating state $x$}
}

\newglossaryentry{flow}{
  name={\ensuremath{F^{\star}}},
  description={Valid trajectory flow}
}

\newglossaryentry{flownetwork}{
  name={\ensuremath{\protect\widebar{\gF}(\gG)}},
  description={Set of all trajectory flows over $\gG$}
}

\newglossaryentry{totalflow}{
  name={$Z^{\star}$},
  description={Total flow of a valid trajectory flow}
}

\newglossaryentry{backwardtransitionprobability}{
  name={\ensuremath{P_{B}}},
  description={Backward transition probability}
}

\newglossaryentry{simplexparents}{
  name={$\Delta(\parents_{\gG})$},
  description={Space of probability distributions over the parents of some $s\in\gG$}
}

\newglossaryentry{backwardinduceddistribution}{
  name={$P_{B}(\tau\mid x)$},
  description={Distribution over trajectories leading to a terminating state $x$ induced by $P_{B}$}
}

\newglossaryentry{lhs}{
  name={LHS},
  description={Left hand side}
}

\newglossaryentry{rhs}{
  name={RHS},
  description={Right hand side}
}

\newglossaryentry{markovianflownetwork}{
  name={\ensuremath{\protect\widebar{\gF}_{\markov}(\gG)}},
  description={Set of all Markovian flows over $\gG$}
}

\newglossaryentry{fm}{
  name={FM},
  description={Flow matching condition / loss}
}

\newglossaryentry{db}{
  name={DB},
  description={Detailed balance condition / loss}
}

\newglossaryentry{tb}{
  name={TB},
  description={Trajectory balance condition / loss}
}

\newglossaryentry{subtb}{
  name={SubTB},
  description={Sub-trajectory balance condition / loss}
}

\newglossaryentry{reward}{
  name={$R$},
  description={Reward function of the GFlowNet}
}

\newglossaryentry{gflownet}{
  name={GFlowNet},
  description={Generative flow network}
}

\newglossaryentry{gfn}{
  name={GFN},
  description={Generative flow network}
}

\newglossaryentry{residual}{
  name={$\Delta(\cdot; \phi)$},
  description={Algorithm-specific residual for a flow matching loss}
}

\newglossaryentry{phi}{
  name={$\phi$},
  description={Parameters of the GFlowNet}
}

\newglossaryentry{pib}{
  name={$\pi_{b}$},
  description={Behavior policy for off-policy algorithms}
}

\newglossaryentry{hvi}{
  name={HVI},
  description={Hierarchical variational inference}
}

\newglossaryentry{ppo}{
  name={PPO},
  description={Proximal policy optimization}
}

\newglossaryentry{energy}{
  name={$\gE(x)$},
  description={Energy function}
}

\newglossaryentry{pcl}{
  name={PCL},
  description={Path Consistency Learning}
}

\newglossaryentry{pisql}{
  name={$\pi$-SQL},
  description={Policy-parametrized Soft Q-Learning}
}

\newglossaryentry{fldb}{
  name={FL-DB},
  description={Forward-looking detailed balance condition / loss}
}

\newglossaryentry{rlhf}{
  name={RLHF},
  description={Reinforcement learning from human feedback}
}

\newglossaryentry{returntime}{
  name={$\sigma_{B}$},
  description={Return time (random time) to a set $B$}
}

\newglossaryentry{hittingtime}{
  name={$\eta_{B}$},
  description={Hitting time (random time) to a set $B$}
}

\newglossaryentry{forwardreferencekernel}{
  name={$\kappa_{F}$},
  description={Forward reference kernel of the measurable pointed graph}
}

\newglossaryentry{backwardreferencekernel}{
  name={$\kappa_{B}$},
  description={Backward reference kernel of the measurable pointed graph}
}

\newglossaryentry{measurablepointedgraph}{
  name={$\gG = ((\protect\widebar{\gS}, \Sigma), \kappa_{F}, \kappa_{B}, \nu)$},
  description={Measurable pointed graph}
}

\newglossaryentry{productsigmaalgebra}{
  name={\ensuremath{\Sigma_{S} \otimes \Sigma_{T}}},
  description={Product $\sigma$-algebra between $\Sigma_{S}$ and $\Sigma_{T}$}
}

\newglossaryentry{productkernel}{
  name={$\kappa_{1} \otimes \kappa_{2}$},
  description={Product kernel between two transition kernels $\kappa_{1}$ and $\kappa_{2}$}
}

\newglossaryentry{productkernelrec}{
  name={$P_{F}^{\otimes T}$},
  description={Recursive product kernel ($T$ times product of $P_{F}$)}
}

\newglossaryentry{compositionkernel}{
  name={$\kappa_{1} \cdot \kappa_{2}$},
  description={Composition kernel between two transition kernels $\kappa_{1}$ and $\kappa_{2}$}
}

\newglossaryentry{compositionkernelrec}{
  name={$\kappa^{T}$},
  description={Recursive composition kernel ($T$ times product of $\kappa$)}
}

\newglossaryentry{dqn}{
  name={DQN},
  description={Deep Q-Networks}
}

\newglossaryentry{jsd}{
  name={JSD},
  description={Jensen Shannon divergence}
}

\newglossaryentry{rmse}{
  name={RMSE},
  description={Root mean-square error}
}

\newglossaryentry{eshd}{
  name={$\E$-SHD},
  description={Expected structural hamming distance}
}

\newglossaryentry{shd}{
  name={SHD},
  description={Structural hamming distance}
}

\newglossaryentry{auroc}{
  name={AUROC},
  description={Area under the receiver operating characteristic curve}
}

\newglossaryentry{bgescore}{
  name={BGe score},
  description={Bayesian Gaussian equivalent score}
}

\newglossaryentry{bdescore}{
  name={BDe score},
  description={Bayesian Dirichlet equivalent score}
}

%% file: prologue/abstract-fr.tex
\chapter*{Résumé}
\label{chap:abstract-fr}
\vspace*{-4em}
Découvrir la structure d'un modèle causal seulement à partir de donnée souffre de problèmes d'identifiabilité. En général, plusieurs modèles équivalents peuvent tout aussi bien expliquer la donnée observée, même s'ils impliquent des conclusions causales complètement différentes. Ainsi, choisir un de ces éléments de manière arbitraire pourrait donner lieu à des décisions dangereuses si le modèle n'est pas aligné avec la manière dont le monde fonctionne réellement. Il est donc impératif de maintenir une notion d'incertitude épistémique sur les différents candidats pour limiter les risques posés par ces modèles non alignés, surtout lorsqu'il y a peu de donnée.

En prenant une perspective bayésienne, cette incertitude peut être représentée par une distribution postérieure sur les modèles, conditionnée sur les observations. Mais comme c'est le cas pour beaucoup de problèmes en inférence bayésienne, la postérieure est typiquement impossible à calculer à cause du grand nombre de structures possibles, représentées par des graphes dirigés acycliques (DAGs). Des approximations sont donc nécessaires. Même s'il y a eu d'énormes avancées en modélisation générative ces dernières années, menées par la puissante combinaison de l'inférence variationelle et de l'apprentissage profond, la plupart de ces modèles sont particulièrement adaptés à des espaces continus. Par conséquent, cela les rend inapplicables pour des problèmes avec des objets discrets comme des graphes dirigés, avec des contraintes complexes d'acyclicité.

Dans la première partie de cette thèse, nous introduisons les \emph{réseaux à flots génératifs} (GFlowNets), une nouvelle classe de modèles probabilistes specialement créés pour representer des distributions sur des objets discrets et compositionnels comme des graphes. Les GFlowNets traitent la génération d'un échantillon comme un problème de décisions séquentielles, en le construisant morceau par morceau. Ces modèles décrivent des distributions définies à une constante de normalisation près en imposant la conservation de certains flots à travers un réseau. Nous mettrons l'accent sur les liens qui existent avec divers domaines de l'apprentissage statistique, comme l'inférence variationelle et l'apprentissage par renforcement, et nous discuterons d'extensions à des espaces généraux.

Ensuite dans la deuxième partie de cette thèse, nous montrerons comment les GFlowNets sont capables d'approcher la distribution postérieure sur les structures de DAG des réseaux bayésiens, en fonction d'observations. Mais au delà de la structure seule, nous montrerons que les paramètres des distributions conditionelles peuvent également être intégrés dans l'approximations de la postérieure représentée par le GFlowNet, ce qui nous permet une plus grande flexibilité dans la manière dont les réseaux bayésiens sont définis.

\vspace*{1em}
\textbf{Mots-clés:} Réseaux à flots génératifs, Inférence bayésienne, Apprentissage de structure, Réseaux bayésiens, Apprentissage par renforcement, Inférence variationelle

%% file: prologue/abstract-en.tex
\chapter*{Abstract}
\label{chap:abstract-en}
Discovering the structure of a causal model purely from data is plagued with problems of identifiability. In general, without any assumptions about data generation, multiple equivalent models may explain observations equally well even if they could entail widely different causal conclusions. As a consequence, choosing an arbitrary element among them could result in unsafe decisions if it is not aligned with how the world truly works. It is therefore imperative to maintain a notion of epistemic uncertainty about our possible candidates to mitigate the risks posed by these misaligned models, especially when the data is limited.

Taking a Bayesian perspective, this uncertainty can be captured through the posterior distribution over models given data. As is the case with many problems in Bayesian inference though, the posterior is typically intractable due to the vast number of possible structures, represented as directed acyclic graphs (DAGs). Hence, approximations are necessary. Although there have been significant advances in generative modeling over the past decade, spearheaded by the powerful combination of amortized variational inference and deep learning, most of these models focus on continuous spaces, making them unsuitable for problems involving discrete objects like directed graphs, with highly complex acyclicity constraints.

In the first part of this thesis, we introduce \emph{generative flow networks} (GFlowNet), a novel class of probabilistic models specifically designed for distributions over discrete and compositional objects such as graphs. GFlowNets treat generation as a sequential decision making problem, constructing samples piece by piece. These models describe distributions defined up to a normalization constant by enforcing the conservation of certain flows through a network. We will highlight how they are rooted in various domains of machine learning and statistics, including variational inference and reinforcement learning, and discuss extensions to general spaces.

Then in the second part of this thesis, we demonstrate how GFlowNets can approximate the posterior distribution over the DAG structures of Bayesian networks given data. Beyond structure alone, we show that the parameters of the conditional distributions can also be integrated in the posterior approximated by the GFlowNet, allowing for flexible representations of Bayesian networks.

\vspace*{1.5em}
\textbf{Keywords:} Generative flow networks, Bayesian inference, Structure learning, Bayesian networks, Reinforcement learning, Variational inference

%% file: prologue/acknowledgements.tex
\chapter*{Acknowledgements / Remerciements}

First and foremost, I would like to thank my advisor Yoshua Bengio for giving me this unique opportunity and his invaluable support throughout my Ph.D. I feel extremely fortunate to have had the chance to work with Yoshua. Thank you so much for trusting me, for giving me the freedom to grow as a researcher, and for always being there. Un grand merci pour tout Yoshua.

I am also grateful for all the professors at Université de Montréal \& Mila I had the opportunity to work with and from whom I learned so much, Doina Precup, Guillaume Lajoie, Laurent Charlin, Pierre-Luc Bacon, Simon Lacoste-Julien, and Stefan Bauer. I want to extend these thanks to Dhanya Sridhar and Kevin Murphy, who have also kindly accepted to be part of my Ph.D.~committee. I want to give a special thanks to all the brilliant researchers I have had the chance to collaborate with over the years, Ant\'onio, Chris, Cristian, David, Edward, Giancarlo, Jithendaraa, Katie, Kolya, Leo, Mansi, Mizu, Moksh, Padideh, Quentin, Salem, and Sébastien. You have shaped the way I have become a researcher, and it is not an understatement to say that this Ph.D.~thesis would not be what it is without you.

But as the life of a researcher is not limited to their work, I would like to thank Akram, Alex, Clara, Joey, Mandana, Maxime, Mélisande, Rim, Samuel, Simon, 
and Victor, and many others at Mila for making my days brighter in Montreal. I want to also thank Julie, who has been so incredibly helpful and so kind throughout all these years.

J'aimerais également remercier, dans la langue de Molière cette fois, tous les soutiens que j'ai reçu de l'autre côté de l'Atlantique. A commencer par ma famille, et en particulier à ma mère Catherine qui a toujours été d'un immense soutien inconditionnel et à qui je dois tellement, mais également à Véronique, François, Tiffany, Nathalie, et Olivier. Je voudrais également remercier mes amis qui ont rendu l'aventure d'autant plus joyeuse, François-Xavier, Jules, Laure, Manon, Marion, Mathilde, Pauline, Pierre-Yves, Romain, Sébastien, Tristan, et tant d'autres. Un grand merci à vous tous.

Finally, I would like to thank my life partner Padideh for her love and her constant support throughout this journey. You have been a shining beacon, and I can only hope that one day I can give back for your own adventure.

%% file: prologue/prologue.tex
\chapter*{Prologue}
\label{chap:prologue}

A Ph.D.~really is a journey, and mine was no exception. What I am about to present in this thesis only represents a part of what I have had the chance to work on over the course of my Ph.D. This choice was motivated by the ambition of presenting my work as a coherent whole that can be read as a standalone piece (including novel results that were never published before). Although the vast majority of the content in this monograph has been re-worked to improve the clarity and consistency of this manuscript, most of the content is based on the following peer-reviewed publications \& preprints (listed in chronological order, ${}^{*}$\,indicates equal contribution):

\begin{itemize}[itemsep=1ex]
    \item \textbf{Tristan Deleu}, Ant\'{o}nio G\'{o}is, Chris Emezue, Mansi Rankawat, Simon Lacoste-Julien, Stefan Bauer, Yoshua Bengio (2022). \emph{Bayesian Structure Learning with Generative Flow Networks}. Conference on Uncertainty in Artificial Intelligence (UAI). (\cref{chap:dag-gflownet}) \notecite{deleu2022daggflownet}
    \item Mizu Nishikawa-Toomey, \textbf{Tristan Deleu}, Jithendaraa Subramanian, Yoshua Bengio, Laurent Charlin (2023). \emph{Bayesian learning of Causal Structure and Mechanisms with GFlowNets and Variational Bayes}. Graphs and More Complex Structures for Learning and Reasoning Workshop (AAAI). (\cref{chap:jsp-gfn}) \notecite{nishikawa2023vbg}
    \item Moksh Jain, \textbf{Tristan Deleu}, Jason Hartford, Cheng-Hao Liu, Alex Hernandez-Garcia, Yoshua Bengio (2023). \emph{GFlowNets for AI-Driven Scientific Discovery}. Digital Discovery, Royal Society of Chemistry. (\cref{chap:probabilistic-inference-structured-objects}) \notecite{jain2023gfnscientific}
    \item Nikolay Malkin$^{*}$, Salem Lahlou$^{*}$, \textbf{Tristan Deleu}$^{*}$, Xu Ji, Edward Hu, Katie Everett, Dinghuai Zhang, Yoshua Bengio (2023). \emph{GFlowNets and Variational Inference}. International Conference on Learning Representations (ICLR). (\cref{chap:generative-flow-networks}) \notecite{malkin2023gfnhvi}
    \item Yoshua Bengio$^{*}$, Salem Lahlou$^{*}$, \textbf{Tristan Deleu}$^{*}$, Edward Hu, Mo Tiwari, Emmanuel Bengio (2023). \emph{GFlowNet Foundations}. Journal of Machine Learning Research (JMLR). (\cref{chap:flow-networks,chap:generative-flow-networks}) \notecite{bengio2023gflownetfoundations}
    \item \textbf{Tristan Deleu}, Yoshua Bengio (2023). \emph{Generative Flow Networks: a Markov Chain Perspective}. (\cref{chap:gflownets-general-state-spaces}) \notecite{deleu2023gfnmarkovchain}
    \item Salem Lahlou, \textbf{Tristan Deleu}, Pablo Lemos, Dinghuai Zhang, Alexandra Volokhova, Alex Hern\'{a}ndez-Garc\'{i}a, L\'{e}na N\'{e}hale Ezzine, Yoshua Bengio, Nikolay Malkin (2023). \emph{A Theory of Continuous Generative Flow Networks}. International Conference on Machine Learning (ICML). (\cref{chap:gflownets-general-state-spaces}) \notecite{lahlou2023continuousgfn}
    \item \textbf{Tristan Deleu}, Mizu Nishikawa-Toomey, Jithendaraa Subramanian, Nikolay Malkin, Laurent Charlin, Yoshua Bengio (2023). \emph{Joint Bayesian Inference of Graphical Structure and Parameters with a Single Generative Flow Network}. Advances in Neural Information Processing Systems (NeurIPS). (\cref{chap:jsp-gfn}) \notecite{deleu2023jspgfn}
    \item \textbf{Tristan Deleu}, Padideh Nouri, Nikolay Malkin, Doina Precup, Yoshua Bengio (2024). \emph{Discrete Probabilistic Inference as Control in Multi-path Environments}. Conference on Uncertainty in Artificial Intelligence (UAI). (\cref{chap:probabilistic-inference-structured-objects,chap:gflownet-maxent-rl}) \notecite{deleu2024gfnmaxentrl}
\end{itemize}

But these difficult choices meant that some of my work had to be left out of this manuscript. The following publications will \emph{not} be covered in this thesis, listed thematically.

\textbf{Meta-learning}
\begin{itemize}[itemsep=1ex]
    \item \textbf{Tristan Deleu}, David Kanaa, Leo Feng, Giancarlo Kerg, Yoshua Bengio, Guillaume Lajoie, Pierre-Luc Bacon (2022). \emph{Continuous-Time Meta-Learning with Forward Mode Differentiation}. International Conference on Learning Representations (ICLR). \textbf{Spotlight} \notecite{deleu2022comln}
    \item Ramnath Kumar, \textbf{Tristan Deleu}, Yoshua Bengio (2023). \emph{The Effect of Diversity in Meta-Learning}. AAAI Conference on Artificial Intelligence. \textbf{Oral} \notecite{kumar2023diversitymetalearning}
\end{itemize}

\textbf{Structure learning}
\begin{itemize}[itemsep=1ex]
    \item Yoshua Bengio, \textbf{Tristan Deleu}, Nasim Rahaman, Nan Rosemary Ke, S\'{e}bastien Lachapelle, Olexa Bilaniuk, Anirudh Goyal, Christopher Pal (2020). \emph{A Meta-Transfer Objective for Learning to Disentangle Causal Mechanisms}. International Conference on Learning Representations (ICLR) \notecite{bengio2019metatransfer}
    \item S\'{e}bastien Lachapelle, Philippe Brouillard, \textbf{Tristan Deleu}, Simon Lacoste-Julien (2020). \emph{Gradient-Based Neural DAG Learning}. International Conference on Learning Representations (ICLR) \notecite{lachapelle2020grandag}
\end{itemize}

\textbf{Causal representation learning}
\begin{itemize}[itemsep=1ex]
    \item S\'{e}bastien Lachapelle${}^{*}$, \textbf{Tristan Deleu}${}^{*}$, Divyat Mahajan, Ioannis Mitliagkas, Yoshua Bengio, Simon Lacoste-Julien, Quentin Bertrand (2023). \emph{Synergies Between Disentanglement and Sparsity: a Multi-Task Learning Perspective}. International Conference on Machine Learning (ICML) \notecite{lachapelle2023synergiesdisentanglementsparsity}
\end{itemize}

%% file: chapters/00_Introduction.tex
\chapter*{Introduction}
\label{chap:introduction}
\addcontentsline{toc}{chapter}{Introduction}

In 1976, renowned British statistician George E. P. Box wrote that \textit{``all models are wrong, but some are useful''}, to comment on the incremental and interactive nature of scientific discovery (this exact statement appearing only three years later; \citealp{box1979robustnessscientificmodel}). He further went on to say that

\begin{minipage}{\textwidth}
\vspace*{1ex}
\begin{center}
\begin{minipage}{0.8\textwidth}
        \textit{``In applying mathematics to subjects such as physics or statistics we make tentative assumptions about the real world which we know are false but which we believe may be useful nonetheless.''}
    \begin{flushright}
    ---\;George E. P. Box \citeyearpar{box1976sciencestats}
    \end{flushright}
\end{minipage}
\end{center}
\vspace*{1ex}
\end{minipage}

This is the essence of what a \emph{model} is: a convenient simplification of the intricacies observed in the real world \citep{cox1990rolemodelsstats}. Even though these models may be merely statistical to explain what they observe, scientists using these models strive for a representation as faithful as possible, even going towards a \emph{causal} understanding of the world to explain how it responds to active interactions \citep{pearl2009causality}. But if all models are wrong, how can we find the ones that are useful? This question is of utmost importance since a bad model has the potential to be dangerous just as much as a good one may be useful (if not more). This is true now more than ever as artificial intelligence is progressing at such a fast pace and the safety of our models becomes critical \citep{bengio2023aicatastrophicrisk}.

Finding a useful model is all the more difficult that it is typically based on \emph{evidence}. In many cases, especially when the evidence is limited (but not only), multiple concurrent models may explain what we observe equally well and committing to a single candidate is prone to catastrophic failures beyond our observations \citep{hodges1987uncertainty,bengio2024cautiousscientistai}. When faced with risky choices, a rational decision-maker tends to act as if it was maximizing some utility under \emph{uncertainty} \citep{ma2022advancesbayesianml}. To account for this epistemic uncertainty, it appears natural to take a \emph{Bayesian approach}, which Box himself was a fervent advocate of \citep{box1973bayesianinference}, where the uncertainty is quantified by the \emph{posterior distribution} over models given some observations.  Adopting a Bayesian perspective to quantify the \emph{structural uncertainty} of the models can be traced back to the research agenda proposed by \citet{draper1987researchagendamodeluncertainty}, and was addressed in part throughout the 1990s \citep{madigan1994occam,chatfield1995modeluncertainty,draper1995assessmentmodeluncertainty}.

This now raises two questions: (1) how can we represent a statistical (or even causal) model of a system of interest, and (2) how can we represent a posterior distribution over those models? For the former, we choose the well-established framework of \emph{Bayesian networks} \citep{pearl1988probabilisticreasoning} for its flexibility and potential extensions to causality. For the latter though, defining a distribution over the full characterization of a Bayesian network (\ie its parameters \emph{and} its structure) proves to be particularly challenging. The main reason being that the structure of a Bayesian network is defined as a \emph{directed acyclic graph}, thus the posterior must be a distribution over a vast number of discrete objects (directed graphs) with a complex acyclicity constraint. While we could leverage generic tools for Bayesian inference such as \emph{Markov chain Monte Carlo} methods \citep{gelfand1990mcmc,madigan1995structuremcmc}, are there more appropriate approaches based on the combined strengths of modern generative models and deep learning \citep{lecun2015deeplearning,goodfellow2016deeplearning}?

In this thesis, we introduce \emph{generative flow networks} (GFlowNets; \citealp{bengio2021gflownet,bengio2023gflownetfoundations}) as a general framework to sample from distributions over discrete quantities that exhibit some form of \emph{compositional} aspect, such as directed acyclic graphs. Generative flow networks are \emph{discrete}, \emph{sequential}, and \emph{amortized} generative models in the following sense:

\begin{itemize}[itemsep=1ex]
    \item \emph{Discrete}: the objects generated by the GFlowNets are discrete, and will typically have a \emph{compositional} structure as described in \cref{chap:probabilistic-inference-structured-objects} (see \cref{sec:compositional-objects} in particular for examples). This is an under-developed area in modern generative modeling, which has been primarily focused on continuous problems \citep{kingma2013vae,rezende2014dlgm} or continuous relaxations of discrete ones \citep{maddison2017concretedistribution,jang2017gumbelsoftmax}. We will see in \cref{chap:gflownets-general-state-spaces} how GFlowNets can be extended beyond discrete state spaces, including to continuous cases;
    \item \emph{Sequential}: GFlowNets treat generation as a \emph{sequential decision making} problem, where an object is constructed one piece at a time. They are aligned with the recent trend of considering generative models as a sequential process \citep{song2019sorebasedmodels,brown2020gpt3,lipman2023flowmatching,tong2024conditionalflowmatching} as opposed to a static one \citep{goodfellow2014gan}. In the case of GFlowNets, this is inspired by the framework of \emph{reinforcement learning} \citep{sutton2018introrl}, and we will see in \cref{chap:gflownet-maxent-rl} that they are in fact closely related to entropy-regularized reinforcement learning \citep{haarnoja2017sql};
    \item \emph{Amortized}: GFlowNets amortize the cost of sampling by training a policy that can eventually be used to generate new objects. This policy will typically be parametrized by a neural network, leveraging their generalization capabilities in different contexts with some shared structure, and are trained by minimizing some objectives we will introduce in \cref{sec:flow-matching-losses}. We will also see in \cref{sec:gflownets-variational-inference} that GFlowNets are deeply rooted in the literature of \emph{amortized variational inference} \citep{kingma2013vae}.
\end{itemize}

Although GFlowNets are generative models in the ``classical'' sense of the term (under the generative/discriminative dichotomy in probabilistic modeling; \citealp{ng2001discriminativevsgenerative}), they are notably different from the more ``modern'' expectation of what generative models are: instead of being a model that learns a generative process from a dataset of examples (\eg image generation), GFlowNets rely on a predefined \emph{reward function} that gives a notion of preference for some objects over others (\ie objects with a higher reward will be sampled more frequently than those with a lower reward). This is a setting frequently encountered in Bayesian inference, but also more generally in scientific discovery \citep{noe2019boltzmanngenerators}.
\newpage

\section*{Outline}
\label{sec:intro-outline}
The first part of this thesis will be dedicated to the theory of generative flow networks in its generality. In \cref{chap:probabilistic-inference-structured-objects} we introduce the problem of sampling from an \emph{energy-based model}, which is a distribution defined up to an intractable normalization, that will serve as a through line during the first part. We will argue via multiple examples how sampling from a distribution over discrete and compositional objects can be seen as a sequential decision making problem, which in some cases falls into the framework of \emph{maximum-entropy reinforcement learning} (MaxEnt RL).

In the general case though, this approach based on MaxEnt RL is biased and does not sample from the target distribution as expected. This motivates the introduction of GFlowNets as a general solution that does not suffer from the same bias. We first introduce in \cref{chap:flow-networks} some properties of \emph{flow networks}, which serve as the foundations for GFlowNets, as well as various conditions to characterize them. Armed with this, we then present in \cref{chap:generative-flow-networks} how to use flow networks in conjunction with appropriate \emph{boundary conditions} for generative modeling. We show how the conservation laws characterizing the flow networks can be turned into practical learning objectives, and show novel convergence guarantees of the approximate distribution towards the target, making GFlowNets rooted in the field of \emph{variational inference}.

Then in \cref{chap:gflownet-maxent-rl} we circle back and show that GFlowNets and MaxEnt RL are actually ``two faces of the same coin'', provided that the reward function is properly corrected. We show that many of the objectives introduced in \cref{chap:generative-flow-networks} find natural counterparts in the MaxEnt RL literature. Finally in \cref{chap:gflownets-general-state-spaces}, through an excursion to \emph{Markov chains}, we show how GFlowNets can be extended to general state spaces, including continuous ones, despite being originally created specifically for discrete settings.

The second part of this thesis will then focus specifically on the problem of \emph{Bayesian structure learning}, that is to model the posterior distribution over Bayesian networks, using GFlowNets. In \cref{chap:dag-gflownet} we first present how GFlowNets can be used to sample from a distribution over directed acyclic graphs, by constructing a directed graph one edge at a time while efficiently guaranteeing acyclicity at every stage. This allows us to not only model the \emph{marginal} posterior distribution over the structure of a Bayesian network alone, but also serves as the basis for modeling the \emph{joint} posterior over the full characterization of a Bayesian network. In \cref{chap:jsp-gfn}, we finally introduce two separate methods using GFlowNets to model the \emph{joint} posterior distribution over the structure \emph{and} parameters of a Bayesian network, including one with a single GFlowNet leveraging the extension to general state spaces introduced in \cref{chap:gflownets-general-state-spaces}.

%% file: chapters/01_Background.tex
\chapter{Background}
\label{chap:background}
In this chapter, we provide a brief overview of the necessary concepts of probabilistic modeling useful throughout this thesis. Other domains will also be introduced in further chapters, such as notions of graph theory in \cref{sec:elements-graph-theory} and reinforcement learning in \cref{sec:maximum-entropy-rl}.  %

\section{Bayesian networks}
\label{sec:bayesian-networks}
Probabilistic graphical models \citep{koller2009pgm,murphy2023pml2book} offer a way to represent probability distributions of complex systems in a compact and modular fashion. Although there exists a variety of ways to encode modularity, such as factor graphs \citep{kschischang2001loopybp}, we will focus primarily on \emph{Bayesian network} in this thesis \citep{pearl1988probabilisticreasoning}, as they constitute the foundations for eventually representing causal knowledge, as we will see further in \cref{sec:causality}.

Before going further, we need to first recall the notion of \emph{conditional independence}, which plays a critical role in Bayesian networks. Two random variables $X$ and $Y$ are said to be independent given a third random variable $Z$, and denoted by \gls{conditionalindependence}, if and only if
\begin{equation}
    P(X, Y\mid Z) = P(X\mid Z)P(Y\mid Z).
    \label{eq:conditional-independence}
\end{equation}
This definition can be naturally extended to $Z \equiv \emptyset$, in which case $X$ and $Y$ are called \emph{marginally independent}, and to cases where $X$, $Y$, or $Z$ are sets of random variables. This characterization of conditional independence in \cref{eq:conditional-independence} can be equivalently written as $P(X\mid Y, Z) = P(X\mid Z)$. In other words, if we observe the value of $Z$, then knowing something about $Y$ does not provide any additional information about $X$.

\subsection{Representation of conditional independencies}
\label{sec:representation-conditional-independencies}
\begin{figure}[t]
    \centering
    \begin{adjustbox}{center}
        \includegraphics[width=480pt]{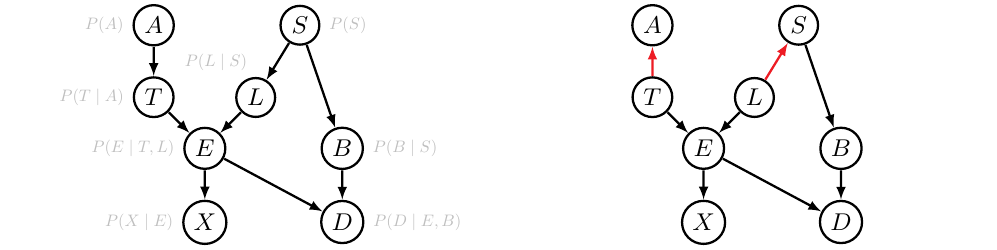}%
    \end{adjustbox}
    \begin{adjustbox}{center}%
        \begin{subfigure}[b]{240pt}%
            \caption{}%
            \label{fig:bayesian-network}%
        \end{subfigure}%
        \begin{subfigure}[b]{240pt}%
            \caption{}%
            \label{fig:bayesian-network-mec}%
        \end{subfigure}%
    \end{adjustbox}
    \caption[Bayesian network]{Factorization of the joint distribution $P(A, B, D, E, L, S, T, X)$ as a Bayesian network \citep{lauritzen1988asia}. (a) Illustration of the structure of a Bayesian network, with the corresponding conditional distributions. (b) An example of a Markov equivalent Bayesian network representing the same conditional independencies, where the edges $A \rightarrow T$ \& $S\rightarrow L$ have been reversed.}
    \label{fig:bayesian-network-and-mec}
\end{figure}

Abiding to the principle of modularity, a \emph{Bayesian network}\index{Bayesian network} is a compact representation of a joint distribution between \gls{numnodes} random variables \gls{randomvars}, describing some system of interest. Given a directed acyclic graph (\gls{dag})\index{Directed acyclic graph} \gls{structuredag}, where each node corresponds to one of those random variables, the joint distribution factorizes according to the graph $G$ in the following way
\begin{equation}
    P_{\theta}(X_{1}, \ldots, X_{d}) = \prod_{i=1}^{d}P\big(X_{i}\mid \parents_{G}(X_{i}); \theta_{i}\big),
    \label{eq:bayesian-network}
\end{equation}
where $\parents_{G}(X_{i})$ represents the parent random variables of $X_{i}$ in $G$, and $\gls{structureparams} = (\theta_{1}, \ldots, \theta_{d})$ are the parameters of the conditional distributions. The structure of the Bayesian network is typically fixed based on expert knowledge, although we will see in \cref{sec:structure-learning}, and more generally in the second part of this thesis, that it is possible to learn it from data. \cref{fig:bayesian-network} shows an example of a Bayesian network.

\begin{example}[Linear-Gaussian models]
    \label{ex:linear-gaussian-model}\index{Linear-Gaussian model}
    One class of Bayesian networks we will study later in this thesis is called \emph{linear-Gaussian models}. Given a DAG $G$, the conditional probabilities appearing in \cref{eq:bayesian-network} are Normal distributions, whose mean is a linear combination of the parent variables. We can write such a model as a set of \emph{structural equations} of the form
    \begin{align}
        X_{i} &:= \sum_{j=1}^{d}\mathds{1}\big(X_{j}\in\parents_{G}(X_{i})\big)\theta_{ji}X_{j} + \varepsilon_{i} && \textrm{where} & \varepsilon_{i} &\sim \gN(0, \sigma_{i}^{2}).
        \label{eq:linear-gaussian-model}
    \end{align}
    We use the convention ``$\theta_{ji}$'' to align with the convention for adjacency matrices, effectively interpreting the parameters $\theta$ as a weighted adjacency matrix, with zero entries when there is no edge between two nodes. We can show that the joint distribution in that case is a multivariate Normal distribution with $0$ mean, and whose covariance matrix depends on $\theta$ and $\sigma_{i}^{2}$'s. 
\end{example}

The factorization of the joint distribution $P_{\theta}(X_{1}, \ldots, X_{d})$ according to the graph reveals a number of conditional independencies between sets of random variables. This is encoded in $G$ via the notion of \emph{d-separation} \citep{pearl1988probabilisticreasoning}, which is a purely graphical concept.
\begin{definition}[d-separation]
    \label{def:d-separation}
    An undirected path (\ie a sequence of edges in $G$, ignoring their orientations) is \emph{d-separated} by a set of nodes $X_{C}$ if it contains at least one of the following sub-structures:
    \begin{enumerate}
        \item a \emph{chain} of the form $A \rightarrow C \rightarrow B$, with $C \in X_{C}$;
        \item a \emph{fork} of the form $A \leftarrow C \rightarrow B$, with $C \in X_{C}$;
        \item a \emph{v-structure} of the form $A \rightarrow C \leftarrow B$, with neither $C$ nor any of its descendents in $X_{C}$.
    \end{enumerate}
    Two sets of nodes $X_{A}$ and $X_{B}$ are d-separated by $X_{C}$ if any undirected path between nodes of $X_{A}$ and $X_{B}$ are d-separated by $X_{C}$.
\end{definition}
Although this may seem like a highly complex criterion to verify, there exist algorithmic solutions to easily uncover d-separation \citep{shachter1998bayesball}. When applying it to the structure of the Bayesian network, we can read conditional independencies (or lack thereof) in the joint distribution it represents. For example, one can verify that $A \not\independent B \mid X$ in the joint distribution described by the Bayesian network in \cref{fig:bayesian-network}, even though $A \independent B$. Being able to read conditional independencies from the graph is called the \emph{global Markov property}.
\begin{proposition}[Global Markov property]
    \label{prop:global-markov-property}
    Let $P$ be a distribution that factorizes according to a Bayesian network with structure $G$. Let $X_{A}$, $X_{B}$ and $X_{C}$ be three sets of random variables (nodes in $G$). If $X_{A}$ and $X_{B}$ are d-separated by $X_{C}$ in $G$, then $X_{A} \independent X_{B}\mid X_{C}$.
\end{proposition}
Recall that this is a purely graphical (hence non-parametric) criterion: it only guarantees that if there exists d-separation, then there must be conditional independence in $P_{\theta}$ \emph{regardless of the parameters $\theta$}. However, there could be additional conditional independencies that are not reflected in the structure $G$ but exist because of the specific parametrization $\theta$. Finally, while it is tempting to interpret the direction of the edges of the graph as being ``causal'' influences of one variable on another, it is important to note that Bayesian networks only encode statistical relationships (conditional independencies), and not causal ones. Nevertheless, they serve as the foundations for causal models, which we will introduce in \cref{sec:causality}.

\subsection{Markov equivalence}
\label{sec:markov-equivalence}\index{Markov equivalence class}
Since the DAG structure of a Bayesian network encodes the conditional independencies one can find in the joint distribution it represents, a question one could ask is whether this encoding is unique. The answer is unfortunately \emph{no}: there exist multiple DAG structures that represent the same set of conditional independencies. Two DAGs $G_{1}$ and $G_{2}$ encoding the same set of conditional independencies are called \emph{Markov equivalent}, and this equivalence relation creates a partition of the space of DAGs into \glspl{mec}. \cref{fig:bayesian-network-and-mec} depicts two DAGs that are Markov equivalent, despite having different structures. The following theorem gives a graphical criterion to check if two graphs are Markov equivalent.
\begin{theorem}[\citealp{verma1990mec}]
    \label{thm:markov-equivalence}
    Two Bayesian networks with DAG structures $G_{1}$ and $G_{2}$ are Markov equivalent if and only if they have the same skeleton and the same v-structures.
\end{theorem}
In this context, the skeleton of a directed graph corresponds to the same (undirected) graph with no orientation of its edges. Markov equivalence classes can be represented compactly as a \emph{completed partially directed acyclic graph} (CPDAG), made of the skeleton and v-structures shared across all the members of the equivalence class. The existence of Markov equivalence reinforces our observation in the previous section that the directions of the arrows do not have a causal interpretation.

\subsection{Sampling \& probabilistic inference}
\label{sec:sampling-probabilistic-inference}
Bayesian networks are \emph{generative models}, in the sense that they model the joint distribution of an entire system. In particular, this means that if the structure and the parameters of the model are known, it is possible to generate virtual data from it. The acyclic nature of the graph suggests a very simple algorithm to sample from a Bayesian network called \emph{ancestral sampling}, where the values of each random variable are sampled sequentially following a topological ordering of the DAG, based on the conditional probability distributions composing the factorization in \cref{eq:bayesian-network}.

But other than generating data, Bayesian networks are mainly used as a tool to answer questions about the system it is modeling. This is called \emph{probabilistic inference}. Probabilistic inference refers very generally to the problem of estimating quantities of the form $\E_{P_{\theta}}\big[f(X_{1}, \ldots, X_{d})\big]$, where $f$ is some function. In the context of Bayesian networks though, it will often refer to estimating conditional probabilities of the form $P_{\theta}(Z\mid X)$, where $\{X, Y, Z\}$ form a partition of $X_{1}, \ldots, X_{d}$ (\ie we marginalize over the variables in $Y$). That's why inference is related to answering questions about a system: we want to know (probabilistically) something about an unknown quantity $Z$, given that we observed some other variables $X$, ignoring (marginalizing out) all other variables $Y$. The unobserved $Z$ are called \emph{latent variables}. Although it is possible to leverage the conditional independencies encoded by the Bayesian network to simplify $P_{\theta}(Z\mid X)$ in some special cases, in general computing this type of conditional distributions is NP-hard \citep{dagum1993inferencenphard}. Therefore, we often resort to approximations such as \emph{variational inference} as presented further.

\subsection{Parameters learning}
\label{sec:parameters-learning}
We have assumed thus far that the structure $G$ of the Bayesian network and the parameters $\theta$ of its conditional distributions were known and fixed. However, it is possible to learn either of them from a dataset of observations $\gls{dataset} = \{\vx^{(n)}\}_{n=1}^{\gls{numsamples}}$. While we will see in \cref{sec:structure-learning} how to learn $G$ from data, a very common task is to learn the parameters $\theta$ of a Bayesian network with a known structure.

If we observe the whole system, meaning that each datapoint $\vx^{(n)} \in \sR^{d}$ in $\gD$ is a complete assignment of all the random variables, then we can learn the parameters by maximizing the (log-)likelihood of the data available:
\begin{equation}
    \widehat{\theta}_{\mathrm{MLE}} = \argmax_{\theta} \log P_{\theta}(\gD) = \argmax_{\theta} \sum_{n=1}^{N}\log P_{\theta}(\vx^{(n)}).
    \label{eq:maximum-likelihood-estimation}
\end{equation}
Thanks to the decomposition of the joint distribution in a Bayesian network \cref{eq:bayesian-network}, this maximization can be carried out for each individual parameter $\theta_{i}$ in a completely independent way. However, if we only observe part of the system (\ie $\vx^{(n)}$ is only a partial assignment, and some variables remain latent), then learning $\theta$ necessitates a step of probabilistic inference to ``complete'' the data. This is the idea behind the \emph{(variational) expectation maximization} algorithm, which we will introduce in \cref{sec:estimation-parameters}.

\section{Variational inference}
\label{sec:variational-inference}\index{Variational inference}
We consider a general inference problem of the form $P(\vz\mid \vx)$, where $\vz$ are latent variables and we observe $\vx$. For example, this may be derived from a joint distribution $P(\vz, \vx)$ described as a Bayesian network, with a known structure and parameterization. With some notable exceptions, for example when the Bayesian network has a near-tree structure \citep{zhang1996variableelimination}, computing this conditional distribution is typically intractable. While we could take a \emph{sample-based} approach to inference if we had access to samples of $P(\vz\mid \vx)$ (we will come back to it in \cref{sec:existing-approaches-sampling-ebm}), in this section we will focus on getting an approximation of this target distribution. We consider that the generative model $P(\vz, \vx)$ is known and tractable (\ie if we had full information, then we could compute this joint probability).

\subsection{Inference as optimization}
\label{sec:inference-as-optimization}
\emph{Variational inference} is the general principle of treating inference as an optimization problem. The goal is to approximate the intractable distribution $P(\vz\mid \vx)$ with a tractable (and often simpler) distribution $Q_{\phi}(\vz)$ coming from a family of distributions $\gQ = \{Q_{\phi}\mid \phi\in\Phi\}$ parametrized by $\phi$. For example, the variational family $\gQ$ may be the set of Normal distributions parametrized by their natural parameters $\phi = (\mu, \sigma^{2})$. This approximation can be quantified with some notion of discrepancy $D$ between the two distributions, which we seek to minimize
\begin{equation}
    Q_{\phi^{\star}} = \argmin_{Q_{\phi}\in\gQ} D\big(Q_{\phi}(\vz), P(\vz\mid \vx)\big).
    \label{eq:variational-inference-problem}
\end{equation}
The distribution $Q_{\phi}(\vz)$ approximates the conditional $P(\vz\mid \vx)$ \emph{for a specific $\vx$}, and therefore its dependency on $\vx$ is implicit here; however, we will see in the next section how we can \emph{amortize} inference and consider a distribution $Q_{\phi}(\vz\mid \vx)$ that depends explicitly on $\vx$. To measure the difference between two distributions $P$ and $Q$, there exists a broad class of divergences called \emph{$f$-divergences}\index{f-divergence@$f$-divergence} based on non-negative convex functions $f$ such that $f(1) = 0$, defined by
\begin{equation}
    D_{f}\big(P(\vz\mid \vx)\,\|\,Q(\vz)\big) = \int_{\vz}Q(\vz)f\left(\frac{P(\vz\mid \vx)}{Q(\vz)}\right)d\vz.
    \label{eq:f-divergence}
\end{equation}
A popular example of divergence is the \emph{Kullback-Leibler divergence}\index{KL divergence}, also known as the KL divergence, which corresponds to the $f$-divergence associated with the convex function $f(x) = -\log x$:
\begin{equation}
    \KL\big(Q(\vz)\,\|\,P(\vz\mid \vx)\big) = \int_{\vz}Q(\vz)\log \frac{Q(\vz)}{P(\vz\mid \vx)}d\vz.
    \label{eq:kl-divergence}
\end{equation}
A general property of $f$-divergences is that it achieves its minimum at 0 iff.~both distributions are equal, justifying our choice to quantify the accuracy of the approximation. Minimizing the discrepancy between $Q_{\phi}(\vz)$ and $P(\vz\mid \vx)$ may seem daunting, since $P(\vz\mid \vx)$ is intractable and yet necessary in the definition of a $f$-divergence like the KL divergence. Fortunately, we can rewrite the KL divergence to explicitly separate out the intractable evidence $P(\vx)$:
\begin{equation}
    \KL\big(Q_{\phi}(\vz)\,\|\,P(\vz\mid \vx)\big) = \int_{\vz}Q_{\phi}(\vz)\log \frac{Q_{\phi}(\vz)}{P(\vz\mid \vx)} = \underbrace{\int_{\vz}Q_{\phi}(\vz)\log \frac{Q_{\phi}(\vz)}{P(\vz, \vx)}}_{\triangleq\,-\gJ(\phi)} + \log P(\vx).
    \label{eq:derivation-elbo-from-kl}
\end{equation}
The quantity $\gJ(\phi)$ is called the \emph{evidence lower-bound} (\gls{elbo}), and gets its name from the fact that it is a lower bound of the (log-)evidence $\log P(\vx) \geq \gJ(\phi)$ (since the KL divergence is non-negative). It is clear that minimizing the KL divergence is exactly equivalent to maximizing the ELBO, which is more commonly written as
\begin{equation}
    \gJ(\phi) = \E_{Q_{\phi}}\big[\log P(\vx\mid \vz)\big] - \KL\big(Q_{\phi}(\vz)\,\|\,P(\vz)\big).
\end{equation}
To find the distribution $Q_{\phi}(\vz)$ that best approximates $P(\vz\mid \vx)$ (in the sense of the KL divergence), it is therefore standard to obtain the variational parameters $\phi^{\star}$ that maximize $\gJ(\phi)$ using gradient-based methods.

\subsection{Amortized inference}
\label{sec:amortized-inference}\index{Variational inference!Amortization}
This variational approach considers that we perform inference for a single observation $\vx$. If we wanted to do inference for a novel observation $\vx'$, then we would have to go through the same exercise and find a completely separate variational distribution $Q'_{\phi}(\vz)$ that minimizes $\KL\big(Q'_{\phi}(\vz)\,\|\,P(\vz\mid \vx')\big)$, and therefore solve a completely new optimization problem. This quickly becomes cumbersome, as each observation requires its own variational distribution.

Instead of treating each as being separate optimization problems, we could make use of some common information shared between observations to simplify inference. We can make the variational distribution $Q_{\phi}(\vz\mid \vx)$ depend explicitly on the observation $\vx$, and learn a single set of variational parameters $\phi$ that would work for any observation $\vx$. This process of sharing parameters across observations is called \emph{amortization} \citep{gershman2014amortizedinference,amos2023amortizedoptimization}. $Q_{\phi}(\vz\mid \vx)$ is known as an \emph{inference model}, in contrast with the generative model $P(\vz, \vx)$. For example, the inference model can be defined as a Normal distribution $Q_{\phi}(\vz\mid \vx) = \gN(\vz\mid \mu_{\phi}(\vx), \sigma^{2}_{\phi}(\vx)\big)$, where this time the parameters of the Normal distribution are functions of the observation $\vx$ parametrized by $\phi$. %

\subsection{Estimation of the parameters}
\label{sec:estimation-parameters}
So far, we have assumed that the generative model $P_{\theta}(\vz, \vx)$ was known (\eg a Bayesian network with fixed structure and parametrization). In practice, we may want to also learn the parameters $\theta$ of this model along with the variational parameters of the inference model $Q_{\phi}(\vz\mid \vx)$, based on a dataset of observations $\gD = \{\vx^{(1)}, \ldots, \vx^{(N)}\}$. To estimate the parameters of a model in the presence of latent variables via maximum likelihood of $P_{\theta}(\gD)$, it is standard to use the \emph{expectation maximization algorithm} (\gls{em}; \citealp{dempster1977em}). When this is coupled with a variational approximation of $P(\vz\mid \vx)$, this is known as the \emph{variational EM algorithm} \citep{neal1998variationalem}. The ELBO in that case becomes
\begin{equation}
    \gJ(\theta, \phi) = \sum_{n=1}^{N}\E_{Q_{\phi}(\vz\mid \vx^{(n)})}\big[\log P_{\theta}(\vx^{(n)}\mid \vz)\big] - \KL\big(Q_{\phi}(\vz\mid \vx^{(n)})\,\|\,P_{\theta}(\vz)\big)
    \label{eq:elbo}
\end{equation}
where we made the dependency on both parameters $\theta$ of the generative model and $\phi$ of the inference model explicit. It is common practice to reweight $\gJ(\theta, \phi)$ by the number of observations in order to normalize the value of the objective.

Following closely the formulation of the EM algorithm\index{Variational inference!Variational EM}, the ELBO may be maximized using coordinate ascend, alternating between (1) finding the variational parameters $\phi$ that lead to a faithful approximation of $P_{\theta}(\vz\mid \vx)$ (\ie maximizing $\gJ(\theta, \phi)$ wrt.~$\phi$, E step), and (2) finding the parameters of the generative model $\theta$ that best explain the observations completed with $Q_{\phi}(\vz\mid \vx)$ (\ie maximizing $\gJ(\theta, \phi)$ wrt.~$\theta$, M step). However recently, powered by the advances in deep learning and automatic differentiation, it has been common practice to jointly maximize $\gJ(\theta, \phi)$ using gradient-based methods \citep{kingma2013vae,rezende2014dlgm}.

\paragraph{Estimation of the gradients of the ELBO} Optimization with gradient-based methods necessitates access to the gradients of $\gJ(\theta, \phi)$. In particular, $\nabla_{\phi}\gJ(\theta, \phi)$ requires computing gradients of quantities of the form $\E_{Q_{\phi}}\big[f(\vz)\big]$, with appropriate functions $f$ (\eg $f(\vz) = \log P_{\theta}(\vx^{(n)}\mid \vz)$ in the first term of \cref{eq:elbo}). In order to account for the dependency of the distribution $Q_{\phi}$, over which the expectation is taken, on the parameters $\phi$ when computing the gradient $\nabla_{\phi}\E_{Q_{\phi}}\big[f(\vz)\big]$, we can push the parameters $\phi$ inside the expectation via a change of variables \citep{rubinstein1992sensitivityanalysis,schulman2015gradientstochastic}. To make things more concrete, suppose that the variational family is of the form $Q_{\phi}(\vz\mid \vx) = \gN\big(\vz\mid \mu_{\phi}(\vx), \sigma^{2}_{\phi}(\vx)\big)$. We can rewrite an expectation wrt.~$Q_{\phi}$ as
\begin{equation}
    \E_{Q_{\phi}}\big[f(\vz)\big] = \E_{\varepsilon\sim \gN(0, 1)}\big[f(\mu_{\phi}(\vx) + \sigma_{\phi}(\vx)\varepsilon)\big].
\end{equation}
With this, we can then compute the gradient of this expectation wrt.~$\phi$ by simply differentiating inside the expectation as the distribution $\gN(0, 1)$ no longer depends on $\phi$
\begin{equation}
    \nabla_{\phi}\E_{Q_{\phi}}\big[f(\vz)\big] = \E_{\varepsilon\sim \gN(0, 1)}\big[\nabla_{\phi} f(\mu_{\phi}(\vx) + \sigma_{\phi}(\vx)\varepsilon)\big].
    \label{eq:reparametrization-trick}
\end{equation}\index{Reparametrization trick}
This is known as the \emph{pathwise derivative} \citep{mohamed2020mcestimationml}, or the \emph{reparametrization trick} \citep{kingma2013vae}. This leads to a natural estimation of the gradient of the ELBO, based on a Monte-Carlo estimate of \cref{eq:reparametrization-trick}. Estimating the gradient $\nabla_{\theta}\gJ(\theta, \phi)$ on the other hand is typically easier since there is no need for reparametrization.

This is an effective way to estimate the gradients of the ELBO when the latent variables are continuous. But when $\vz$ is discrete, variational inference becomes more challenging as there is in general no direct counterpart of the reparametrization trick in the discrete case. If the number of possible values the latent variables $\vz$ may take is small enough, the expectations appearing in \cref{eq:elbo} can be computed analytically. However, enumeration of the values taken by $\vz$ is rarely feasible, and the presence of discrete latent variables often necessitates continuous relaxations of the problem \citep{jang2017gumbelsoftmax,maddison2017concretedistribution}, the use of score-function estimators that generally suffer from high variance \citep{williams1992reinforce,kleijnen1996scorefunction,rubinstein1996scorefunction}, or biased estimations of the gradients via the straight-through estimator \citep{bengio2013sraightthroughestimator}.

\subsection{Wake-sleep algorithm}
\label{sec:wake-sleep-algorithm}\index{Variational inference!Wake-sleep algorithm}\index{Wake-sleep algorithm|see {Variational inference}}
The difficulty of maximizing the ELBO in cases where the latent variables are discrete comes from the estimation of $\nabla_{\phi}\gJ(\theta, \phi)$ (for the E step in a variational EM setting). This is due to the fact that the expectations are taken wrt.~the inference model $Q_{\phi}$ in the KL divergence. The KL divergence being asymmetric, choosing to minimize $\KL\big(Q_{\phi}(\vz\mid \vx)\,\|\,P_{\theta}(\vz\mid \vx)\big)$ in this order (also called the \emph{reverse KL}) may seem arbitrary since any measure quantifying the difference between these two distribution would do. While this is considered standard in variational inference, we could use any other $f$-divergence with another convex function $f$.

Similar to variational EM, the \emph{wake-sleep algorithm} \citep{hinton1995wakesleep,bornschein2015reweightedwakesleep} minimizes the difference between the conditional distribution induced by the generative model $P_{\theta}(\vz, \vx)$ and the inference model $Q_{\phi}(\vz\mid \vx)$ by alternating between two phases:
\begin{enumerate}
    \item A \emph{wake phase}, where the reverse KL divergence $\KL\big(Q_{\phi}(\vz\mid \vx)\,\|\,P_{\theta}(\vz\mid \vx)\big)$ is being minimized wrt.~the parameters of the generative model $\theta$. This is exactly equivalent to the M step in variational EM (\ie maximizing the ELBO wrt.~$\theta$);
    \item A \emph{sleep phase}, where this time the \emph{forward KL} divergence is being minimized (as opposed to the reverse KL in the E step of variational EM) wrt.~the parameters of the inference model $\phi$. The forward KL divergence is the $f$-divergence with $f(x) = x\log x$, corresponding to $\KL\big(P_{\theta}(\vz\mid \vx)\,\|\,Q_{\phi}(\vz\mid\vx)\big)$. Following the same argument as in \cref{sec:inference-as-optimization}, this corresponds to maximizing the log-likleihood objective
    \begin{equation}
        \gJ_{\mathrm{sleep}}(\phi) = \E_{P_{\theta}(\vz, \vx)}\big[\log Q_{\phi}(\vz\mid \vx)\big].
    \end{equation}
\end{enumerate}
Unlike using the reverse KL divergence for the update of both $\theta$ and $\phi$ in variational EM, this does not require the inference model to be reparametrizable anymore, and therefore is readily applicable to cases where the latent variables are discrete. However, unlike the EM algorithm, wake-sleep has no guarantee to converge to a maximum of the likelihood \citep{neal1997wakesleepconvergence}.

\section{Bayesian statistics}
\label{sec:bayesian-statistics}\index{Bayesian inference}
When we introduced Bayesian networks, we assumed that they were parametrized by some $\theta$ that one could maybe learn from data. These learned parameters may sometimes be unreliable, especially when the amount of data is relatively small and the risk of overfitting to that dataset is high. In those cases, it is beneficial to take a \emph{Bayesian approach} and to treat the parameters as random quantities themselves. The problem of Bayesian inference is therefore to characterize the \emph{posterior} distribution $P(\theta\mid \gD)$, defined by \emph{Bayes' rule}\index{Bayesian inference!Bayes' rule} as
\begin{equation}
    P(\theta\mid \gD) = \frac{P(\gD\mid \theta)P(\theta)}{P(\gD)},
    \label{eq:bayes-rule}
\end{equation}
where $P(\theta)$ is called a \emph{prior} over the parameters $\theta$, $P(\gD\mid \theta)$ the \emph{likelihood} of the dataset, and $P(\gD)$ the \emph{marginal likelihood} (or evidence), which is obtained by marginalizing over all possible $\theta$
\begin{equation}
    P(\gD) = \int_{\theta}P(\gD\mid \theta)P(\theta)d\theta.
    \label{eq:marginal-likelihood}
\end{equation}
In general, computing the marginal likelihood above is intractable, making the computation of the posterior distribution itself intractable as well. A notable (but rare) exception is when the likelihood and the prior distribution are ``conjugate'' of one another, in which case the posterior $P(\theta\mid \gD)$ remains in the same family of distributions as the prior. While we will briefly mention a variational approach to approximate the posterior in \cref{sec:variational-bayes}, approximating distributions known up to a normalization constant, as is the case for $P(\theta\mid \gD)$, will be central in the first part of this thesis.

\subsection{Bayesian model averaging}
\label{sec:bayesian-model-averaging}
More generally than modeling uncertainty over the parameters of the same model, the Bayesian approach applies to complete hypotheses $h$ as well, where the posterior distribution $p(h\mid \gD)$ is over models of data. For example, a hypothesis might be the full specification of the parameters along with the structure of a Bayesian network; this will be studied extensively in the second part of this thesis. We will assume that hypotheses $h\in\gH$ belong to a space which may be discrete (\eg the DAG $G$ of a Bayesian network), continuous (\eg the parameters $\theta$ of a Bayesian network with fixed structure), or a mixture of both (\eg both the structure $G$ and the parameters $\theta$ are unknown).

\paragraph{Bayesian model selection}\index{Bayesian inference!Bayesian model selection} Suppose that we want to choose between two hypotheses $h_{1}$ and $h_{2}$ which one best fits the data available. If we know the posterior distribution, one way we can do that is to choose the hypothesis with the highest posterior probability. Although we saw that this distribution is typically intractable, we can fortunately compare the ratio of probabilities and cancel out the intractable marginal likelihood
\begin{equation}
    \frac{P(h_{1}\mid \gD)}{P(h_{2}\mid \gD)} = \frac{P(\gD \mid h_{1})}{P(\gD\mid h_{2})}\frac{P(h_{1})}{P(h_{2})}.
\end{equation}
The ratio of likelihood terms is also called the \emph{Bayes factor} \citep{kass1995bayesfactor}, and serves as a measure to quantify the strength of the evidence for either hypotheses in case of a uniform prior \citep{jeffreys1948theoryprobability}. More generally, we can look over the whole family of hypotheses and pick the one with highest posterior (log-)probability
\begin{equation}
    \widehat{h} = \argmax_{h\in\gH}\log P(h\mid \gD) = \argmax_{h\in\gH} \log P(\gD\mid h) + \log P(h).
\end{equation}
When the hypothesis $h \equiv G$ is the structure of a Bayesian network, and $P(\gD \mid G)$ then corresponds to the marginal likelihood (where the parameters $\theta$ have been marginalized out), this is also called \emph{empirical Bayes} \citep{carlin2000empiricalbayes}.

\paragraph{Bayesian model averaging}\index{Bayesian inference!Bayesian model averaging} Instead of selecting a single hypothesis with maximum posterior probability though, we could average out predictions based on \emph{all} the hypotheses, weighted by the posterior. We can do this by computing the \emph{posterior predictive} distribution of a new datapoint $x_{\mathrm{new}}$
\begin{equation}
    P(x_{\mathrm{new}}\mid \gD) = \int_{h}P(x_{\mathrm{new}}\mid h)P(h\mid \gD) dh,
    \label{eq:posterior-predictive}
\end{equation}
provided that $x_{\mathrm{new}} \independent \gD\mid h$; we will come back to this condition (and importantly, to failure cases) in \cref{sec:criticism-evaluation-metrics}. This holds for example when the hypothesis $h \equiv \theta$ are the parameters of a Bayesian network with fixed structure. %

\paragraph{Bayesian model combination} While averaging the predictions from multiple models is reminiscent of ensembling in frequentist statistics \citep{breiman1996bagging}, the Bayesian approach remains fundamentally different. Bayesian model averaging relies on the implicit assumption that the data was generated from exactly one ``true'' model \citep{reichelt2024beyondbma,bernardo2009bayesiantheory}. The soft weights $P(h\mid \gD)$ only reflect
a statistical inability to distinguish it based on limited data \citep{minka2000bmanotcombination}. Therefore, in the limit of large \emph{independent and identically distributed (\gls{iid})} data, the posterior will typically concentrate around a single model \emph{if this true model is in the family $\gH$} \citep{clydec2013bmaopen}, effectively ignoring incoherent hypotheses and reducing \cref{eq:posterior-predictive} to the prediction of a single model.

But most importantly, albeit theoretically grounded, predictions made with Bayesian model averaging tend to underperform when compared to frequentist ensembles \citep{domingos2000bmaoverfitting}. It has been conjectured that this is a consequence of the hypothesis class not being expressive enough to capture all the subtleties of the real world; the ``true'' model, assuming it exists, is not in $\gH$. To address this issue while still adhering to the Bayesian principle, multiple models may be \emph{combined} together \citep{monteith2011bayesianmodelcombination,kim2012bayesianclassifiercombination}.

\subsection{Variational Bayes}
\label{sec:variational-bayes}\index{Variational inference!Variational Bayes}\index{Variational Bayes|see {Variational inference}}
The variational approach presented in \cref{sec:estimation-parameters} allows us to find the best parameters $\theta$ of the generative model with some latent variables $\vz$. What if the latent variables of interest are the model parameters $\theta$ themselves though, as it is the case in Bayesian inference? For example if the parameters can be naturally broken down into $\theta = (\theta_{1}, \theta_{2})$, we can choose a variational family of the form $Q_{\phi}(\theta) = Q_{\phi_{1}}(\theta_{1})Q_{\phi_{2}}(\theta_{2})$ (mean-field approximation), ignoring the dependencies between $\theta_{1}$ \& $\theta_{2}$ in the posterior approximation.

Using a similar argument as in \cref{sec:variational-inference}, it can be shown that the KL divergence between the approximation $Q_{\phi}(\theta)$ and the (intractable) posterior $P(\theta\mid \gD)$ can be decomposed as
\begin{equation}
    \KL\big(Q_{\phi}(\theta)\,\|\,P(\theta\mid \gD)\big) = -\gJ(\phi_{1}, \phi_{2}) + \log P(\gD),
    \label{eq:kl-variational-bayes}
\end{equation}
where $\gJ(\phi_{1}, \phi_{2})$ is an evidence lower-bound (ELBO) that can be written as
\begin{align}
    \gJ(\phi_{1}, \phi_{2}) &= -\KL\big(Q_{\phi_{1}}(\theta_{1})\,\|\,\widetilde{Q}_{1}(\theta_{1})\big) + C_{2}(\phi_{2})\\
    \textrm{where}\quad \widetilde{Q}_{1}(\theta_{1}) &\propto \exp\left(\int_{\theta_{2}}Q_{\phi_{2}}(\theta_{2})\log P(\gD\mid \theta_{1}, \theta_{2})d\theta_{2}\right),\label{eq:elbo-vb-anchor-distribution}
\end{align}
and $C_{2}(\phi_{2})$ is only a function of $\phi_{2}$ (independent of $\phi_{1}$). In a completely symmetric fashion, it is also easy to show that the ELBO can be written as $\gJ(\phi_{1}, \phi_{2}) = -\KL\big(Q_{\phi_{2}}(\theta_{2})\,\|\,\widetilde{Q}_{2}(\theta_{2})\big) + C_{1}(\phi_{1})$, where $\widetilde{Q}_{2}(\theta_{2})$ is defined similarly to \cref{eq:elbo-vb-anchor-distribution}, but replacing $Q_{\phi_{2}}$ by $Q_{\phi_{1}}$, and integrating wrt.~$\theta_{1}$.

We can find the best approximation of the posterior $P(\theta\mid \gD)$ (in the sense of the KL divergence) by maximizing the ELBO $\gJ(\phi_{1}, \phi_{2})$ in \cref{eq:kl-variational-bayes}. Loosely inspired by variational EM in \cref{sec:estimation-parameters}, we can maximize it using coordinate ascent, by repeating (1) minimizing $\KL\big(Q_{\phi_{1}}(\theta_{1})\,\|\,\widetilde{Q}_{1}(\theta_{1})\big)$ wrt.~$\phi_{1}$ given a fixed $Q_{\phi_{2}}$, and (2) minimizing $\KL\big(Q_{\phi_{2}}(\theta_{2})\,\|\,\widetilde{Q}_{2}(\theta_{2})\big)$ wrt.~$\phi_{2}$ given a fixed $Q_{\phi_{1}}$. This algorithm is called \emph{Variational Bayes} \citep{beal2003variationalbayes}. In some cases, one of the minimization steps above can be performed analytically.

\section{Structure learning}
\label{sec:structure-learning}\index{Structure learning}
We saw in \cref{sec:parameters-learning,sec:estimation-parameters} how to learn the parameters $\theta$ based on a dataset of (possibly partial) observations $\gD$, provided that the structure $G$ of the Bayesian network is known and fixed. In this section, we will go one step further and consider the ``inverse problem'' of learning the structure $G$ of the Bayesian network from a dataset $\gD$. This is an important problem in scientific discovery, since uncovering (potentially novel) statistical dependencies in a system is the main objective, and is at least as important as using the model itself for inference.

\subsection{Faithfulness \& identifiability}
\label{sec:faithfulness-identifiability}

To learn the structure of a Bayesian network, it is necessary to have a guarantee that any conditional independencies that can be found from data are going to be reflected in the graph that is eventually found. Otherwise, this inverse problem is ill-posed no matter how much data is available to accurately detect conditional independence. This is called the ``faithfulness'' of the data-generating distribution.

\begin{definition}[Faithfulness]
    \label{def:faithfulness}
    A distribution $P$ is \emph{faithful} to a DAG $G$ if any conditional independence in $p$ is reflected in the d-separations in $G$. Let $A$, $B$ and $C$ be three sets of nodes in $G$, with their respective sets of random variables $X_{A}$, $X_{B}$ and $X_{C}$. If $X_{A} \independent X_{B}\mid X_{C}$, then $A$ and $B$ are d-separated by $C$ in $G$.
\end{definition}

Faithfulness can be seen informally as the ``reverse'' of the global Markov property in \cref{prop:global-markov-property}. But even though the global Markov property is satisfied for any distribution represented by a Bayesian network, the notion of faithfulness is not automatic.

\begin{figure}[t]
    \centering
    \begin{adjustbox}{center}
        \includegraphics[width=480pt]{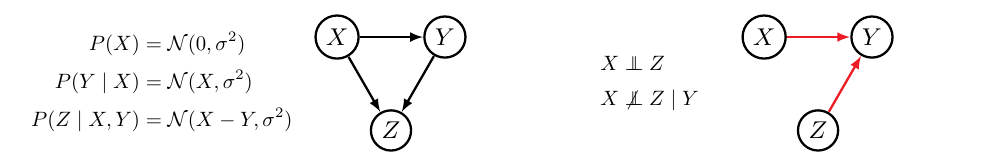}%
    \end{adjustbox}
    \begin{adjustbox}{center}%
        \begin{subfigure}[b]{240pt}%
            \caption{}%
            \label{fig:violation-faithfulness-bayesnet}%
        \end{subfigure}%
        \begin{subfigure}[b]{240pt}%
            \caption{}%
            \label{fig:violation-faithfulness-structlearn}%
        \end{subfigure}%
    \end{adjustbox}
    \caption[Example of a distribution violating the faithfulness assumption]{Example of a distribution violating the faithfulness assumption. (a) Example of a Bayesian network with its conditional distributions in the decomposition of $P(X, Y, Z)$. (b) The only conditional independence statements one can extract from $P(X, Y, Z)$ are $X\independent Z$ \& $X\not\independent Z\mid Y$. In a fully non-parametric way, this DAG is the only one representing exactly both statements.}
    \label{fig:violation-faithfulness}
\end{figure}

\begin{example}[Violation of faithfulness]
    \label{ex:violation-faithfulness}
    Consider the joint distribution $P(X, Y, Z)$ described by the Bayesian network in \cref{fig:violation-faithfulness-bayesnet}. The structure of the Bayesian network follows the ``natural'' description of the conditional distributions, with $Y$ depending on $X$, and $Z$ depending on both $X$ \& $Y$. The graph $G$ being complete, it does not encode any (non-trivial) conditional independencies a priori. However, one can show that the joint distribution $P(X, Y, Z)$ is a multivariate Normal distribution, whose only conditional independencies are $X \independent Z$ and $X \not\independent Z \mid Y$. Therefore, the distribution $p$ is not faithful to $G$, and any attempt to learn the structure of the Bayesian network from $P$ would result at best in the DAG in \cref{fig:violation-faithfulness-structlearn}.
\end{example}

Thankfully, faithfulness is guaranteed for \emph{almost} all distributions $P$ that factorize according to some DAG, in the sense that the set of distributions for which faithfulness is violated is of measure zero \citep[][Theorem 3.5]{koller2009pgm}; in the example above, the conditional distributions were carefully crafted to be unfaithful \citep[][Theorem 3.2]{spirtes2000causationpredictionsearch}. Therefore throughout this thesis, we will always assume faithfulness without explicitly mentioning it again.

Going back to the problem of finding the structure of a Bayesian network, the following theorem shows that if we had complete access to the distribution $P$, then we could recover the DAG \emph{up to Markov equivalence}, under faithfulness.

\begin{theorem}[Structure identifiability]
    \label{thm:structure-identifiability}
    If a distribution $P$ is faithful to $G^{\star}$, then the Markov equivalence class of $G^{\star}$ is identifiable from $P$.
\end{theorem}

Shy of any assumption, the caveat of recovering the DAG only up to Markov equivalence has to be expected, since Markov equivalent structures encode the same conditional independencies. With additional assumptions on the data generating process, stronger forms of identifiability are possible, sometimes even of a single DAG the distribution is faithful to. For example, if $P$ is a non-linear additive noise model, where the structural equation model for each variable can be written as
\begin{align}
    X_{i} &:= f_{i}\big(\parents_{G^{\star}}(X_{i})\big) + \varepsilon_{i} && \textrm{with} & \varepsilon_{i} &\sim \gN(0, \sigma_{i}^{2}),
\end{align}
and where the functions $f_{i}$ are non-linear and depend on the parent variables of $X_{i}$, then the DAG $G^{\star}$ is identifiable from $P$ \citep{peters2014causaldiscoveryanm}. Another classic example of identifiability of a single DAG is in the case where $P$ is a linear-Gaussian model with equal variance (\ie $\sigma_{i}^{2} = \sigma^{2}$ in \cref{eq:linear-gaussian-model}; \citealp{peters2014identifiabilitysamevariance}). We can also further identify the structure more precisely if we have access to experimental data \citep{hauser2012interventionalmec}; see \cref{sec:interventions} for an informal example.

\subsection{Score-based methods}
\label{sec:score-based-methods}
If we knew $P$ completely, then this would lead to a simple strategy for structure learning (again, up to Markov equivalence): find the conditional independencies encoded in $P$, and construct a structure that follows them (in the sense of d-separation). However in practice, we never have access to $P$ itself, but only to a finite dataset of observations $\gD$. \emph{Constraint-based} methods approach structure learning in a similar way, by testing conditional independencies from data using frequentist independence tests \citep{spirtes2000causationpredictionsearch,pearl2009causality}.

Of greater interest in this thesis, \emph{score-based} methods on the other hand treat the problem of learning the structure of the Bayesian network as an optimization problem. For some score function ``$\mathrm{score}(G; \gD)$'' that quantifies how well $G$ fits the dataset, the problem becomes to find a DAG that maximizes this score:
\begin{equation}
    \widehat{G} = \argmax_{G\in\mathrm{DAG}} \mathrm{score}(G; \gD).
    \label{eq:score-based-structure-learning}
\end{equation}

\paragraph{Score functions} The most natural example of score function is the \emph{likelihood score}, which corresponds to the log-likelihood of the maximum likelihood model
\begin{equation}
    \mathrm{score}_{L}(G; \gD) \triangleq \max_{\theta}\log P(\gD\mid G, \theta) = \log P(\gD\mid G, \widehat{\theta}_{\mathrm{MLE}}).
\end{equation}
This score has the notable drawback of inevitably favoring dense structures (similar effect as standard overfitting in machine learning), meaning that using this score always leads to a trivial structure where all the nodes are connected together \citep{koller2009pgm}. To mitigate this issue, it is possible to regularize this score by the number of free parameters in $G$, in order to penalize dense structures. The \emph{Bayesian information criterion (BIC) score} \citep{schwarz1978bicscore} is one example. Another example of regularized score that we will use extensively in \cref{chap:dag-gflownet} is the so-called \emph{Bayesian score}, which treats structure learning as a Bayesian model selection problem\index{Bayesian inference!Bayesian model selection} (introduced in \cref{sec:bayesian-model-averaging}) over the entire space of DAGs, with the score being
\begin{equation}
    \mathrm{score}_{B}(G; \gD) \triangleq \log P(\gD \mid G) + \log P(G).
    \label{eq:bayesian-score}
\end{equation}
Although computing the marginal likelihood $P(\gD\mid G)$ is in general intractable, we will see in \cref{app:bayesian-scores} that this score can be computed analytically for some classes of Bayesian networks (\eg linear-Gaussian models, or discrete Bayesian networks with Dirichlet prior).

\paragraph{Optimization of the score} Albeit conceptually simple, the optimization problem in \cref{eq:score-based-structure-learning} is extremely challenging in practice for two reasons:
\begin{enumerate}
    \item The space of DAGs is discrete, meaning that we cannot use any of the conventional tools from continuous optimization, such as gradient-based methods;
    \item The space of DAGs is combinatorially large. The number of DAGs over $d$ nodes one can build is of the order $2^{\Theta(d^{2})}$ \citep{robinson1973countingdags,rodionov1992numberofdags}. To give some perspective, there are about $10^{18}$ DAGs over $d=10$ nodes, and about $10^{72}$ DAGs over $d=20$ nodes.
\end{enumerate}
This means that this combinatorial optimization problem is NP-hard. Worse, even if we restrict the search space to only DAGs having at most $p$ parents (with $p \geq 2$), then this problem remains NP-hard \citep{chickering1996nphard}. Therefore it is necessary to fall back on heuristic approaches in order to find a DAG maximizing the score \citep{cooper1992structurelearning}, such as evolutionary algorithms \citep{wong2004evolutionarystructurelearning}, or greedy approaches \citep{chickering2002ges,tsamardinos2006mmhc}.

\subsection{Continuous relaxations}
\label{sec:structure-learning-continuous-relaxations}
Another approach that has gained popularity recently is to relax the combinatorial optimization problem \cref{eq:score-based-structure-learning} into a continuous and constrained optimization problem. This stems from the observation of \citet{zheng2018notears} that a graph $G$ with adjacency matrix $A$ is a DAG if and only if
\begin{equation}
    \Tr\big(\exp(A)\big) = d,
    \label{eq:dags-no-tears-condition}
\end{equation}
where $d$ is the number of nodes in $G$, and $\exp(A)$ is the matrix exponential. The function $\Tr\big(\exp(A)\big)$ effectively counts the number of cycles in the graph, reweighted by their length, and it being equal to $d$ means that the only ``cycles'' that exist in a DAG are those of length $0$. Other characterizations of DAGs based on operations on the adjacency matrix also exist \citep{yu2019daggnn,yu2021nocurl,bello2022dagma}. If the score function can be expressed as a function of the adjacency matrix, which is the case for the scores mentioned in the previous section, we can treat \cref{eq:dags-no-tears-condition} as a constraint, and search over the space of all relaxed adjacency matrices
\begin{align}
    \max_{A\in [0, 1]^{d\times d}}\ & \mathrm{score}(A; \gD)\\[-0.5em]
    \mathrm{s.t.}\ & \Tr\big(\exp(A)\big) = d.
\end{align}
This constrained optimization problem can then be solved, this time with gradient-based methods, provided that the score is differentiable wrt.~$A$, for example using the \emph{augmented Lagrangian} algorithm \citep{bertsekas2016nonlinearprog}. While \citet{zheng2018notears} applied this exclusively to the discovery of linear-Gaussian models, we extended this idea to non-linear models \citep{lachapelle2020grandag}.

\subsection{Identifiability \& Bayesian perspective}
\label{sec:identifiability-bayesian-perspective}
The result of \cref{thm:structure-identifiability} only guarantees the identifiability of the Markov equivalence class of $G^{\star}$ if we know $P$ itself. However in practice in structure learning, we don't have access to the whole distribution, but only to a finite dataset $\gD$ of samples from that distribution; identifiability really is an asymptotic notion, in the limit of infinite data. The risk in practice is two-fold: (1) we may not have enough data in $\gD$ to fully identify the Markov equivalence class (MEC) of $G^{\star}$, and (2) even if we identify it, a structure learning algorithm may choose an arbitrary element of the equivalence class, since there is non-identifiability \emph{within} the MEC.

To get a more comprehensive view, we could instead take a Bayesian approach as we described in \cref{sec:bayesian-statistics}, and get a ``soft'' notion of identification via the posterior distribution \citep{cole2020parameterredundancyidentifiability}. In cases where the model is not identifiable, this gets reflected in the posterior distribution $P(G\mid \gD)$, which then has support over the whole class of non-identifiability (\eg the MEC in the limit of infinite data), provided that the prior puts non-zero probability on these models. But the prior can provide additional information to break symmetry in case of non-identifiability \citep{lindley1972bayesianstatsreview}; \eg it has become increasingly popular to consider sparsity inducing penalties in order to prove identifiability \citep{lachapelle2024mechanismsparsity}. If the model is identifiable (\eg to a single DAG, see examples in \cref{sec:faithfulness-identifiability}), the posterior distribution then is consistent and converges to a Dirac distribution, under some regularity conditions (\eg via the Bernstein-von Mises theorem; \citealp{ma2022advancesbayesianml}).

And this is without counting on the main advantage of Bayesian inference: to account for uncertainty over models, especially in cases where the dataset is finite. All these advantages motivate our Bayesian approach for structure learning, which we will detail in the second part of this thesis. However, this will come at the cost of characterizing an intractable posterior distribution, which we will have to approximate.

\section{Causality}
\label{sec:causality}
In \cref{sec:bayesian-networks}, we emphasized that the directed edges of a Bayesian network do not encode any notion of direct causal relationship. These models were developed only to describe statistical relationships of a system that we \emph{passively} observe. A \emph{causal model}, on the other hand, extends the notion of Bayesian network where this time the edges of the DAG correspond to direct causal relationships. Although functionally similar, a causal model allows us to not only passively observe a system, but also reason about \emph{actively} experimenting on it, just like how a scientist would in order to verify a hypothesis. This is done via the notion of \emph{interventions}. Causal models also allow us to reason about fictional situations called ``counterfactuals'' \citep{pearl2009causality}, which we will not detail.

\subsection{Interventions}
\label{sec:interventions}
Let's start with an illustrative example to highlight the fundamental differences between observing and acting on a system. Suppose that we have two random variables of interest: a variable $X$ encoding the weather (\eg as a categorical variable), and $Y$ encoding the measurement of a barometer (continuous measure of the atmospheric pressure). From historical data, passively observing the weather and recording the corresponding value from the barometer every day, it is clear that there exists a dependence between these two variables (high pressure correlates with good weather), and in particular that only the DAGs $X \rightarrow Y$ or $Y \rightarrow X$ may encode the system. However based on our discussion on Markov equivalence in \cref{sec:markov-equivalence}, these two DAGs encode the same conditional independencies (by \cref{thm:markov-equivalence}), and therefore may equally model the system.

If all we could do was passively observe $X$ \& $Y$, then this is the only conclusion we can draw (\cref{thm:structure-identifiability}). But suppose that we actively change the measurement of the barometer (\eg by moving the needle towards higher values). This (unfortunately) has no effect on the weather. The operation of actively changing the value of $Y$ is called performing an \emph{intervention} on $Y$, and it induces a distribution on $(X, Y)$ that has the property of leaving the marginal over $X$ unchanged (\ie intervening on $Y$ does not affect our observational distribution over $X$). Therefore we can confidently conclude that the causal model that best describes the system is $X \rightarrow Y$ (\ie the weather \emph{causes} the measurement on the barometer).

\begin{figure}[t]
    \centering
    \begin{adjustbox}{center}
        \includegraphics[width=480pt]{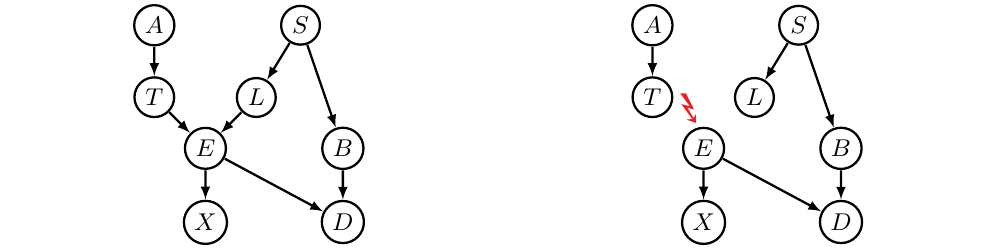}%
    \end{adjustbox}
    \begin{adjustbox}{center}%
        \begin{subfigure}[b]{240pt}%
            \caption{$P(A, B, D, E, L, S, T, X)$}%
            \label{fig:causal-graph-before}%
        \end{subfigure}%
        \begin{subfigure}[b]{240pt}%
            \caption{$P(A, B, D, L, S, T, X\mid \mathrm{do}(E=e))$}%
            \label{fig:causal-graph-after}%
        \end{subfigure}%
    \end{adjustbox}
    \caption[The effect of an intervention on the causal graph]{The effect of an intervention on the causal graph. (a) Example of a causal graph representing a family of distributions, including the joint (observational) distribution $P(A, B, D, E, L, S, T, X)$. (b) The mutilated graph corresponding to intervening on the random variable $E$ (and setting it to the value $e$); the incoming edges of $E$ have been removed.}
    \label{fig:mutilated-graph}
\end{figure}

We saw that intervening on the system induces a new joint distribution over the variables of interest. We will now formalize this idea more generally for a causal model with a DAG structure $G$. For a set of indices $\gI$, if the pair $\{X_{\gI}, X_{-\gI}\}$ forms a partition of $X_{1}, \ldots, X_{d}$, then the interventional distribution of acting on the variables $X_{\gI}$ by setting them to $x_{\gI}$ is given by
\begin{equation}
    P_{\theta}\big(X_{-\gI}\mid \mathrm{do}(X_{\gI} = x_{\gI})\big) = \mathds{1}(X_{\gI} = x_{\gI})\prod_{i\notin\gI}P\big(X_{i}\mid \parents_{G}(X_{i});\theta_{i}\big).
    \label{eq:interventional-distribution}
\end{equation}
We use the notation ``$\mathrm{do}(X_{\gI} = x_{\gI})$'' to distinguish it from standard conditioning, and to make the active nature of an intervention more explicit (as in \emph{doing} an action). From this equation, it is clear what is the effect of an intervention on the distribution: it sets the values of the variables $X_{\gI}$, disregarding the values of their parents in $G$. From a graphical point of view, this can be seen as encoding this distribution with a \emph{mutilated DAG}, where the incoming edges to $X_{\gI}$ have been removed; see \cref{fig:mutilated-graph}. While we saw that a Bayesian network encodes a joint distribution over $X_{1}, \ldots, X_{d}$, a causal model effectively represents a whole \emph{family} of distributions, one for each possible intervention on some random variables; this notably includes the ``observational'' distribution where no variable has been intervened on, matching the joint distribution represented in a Bayesian network. Data collected by intervening on the system will be called \emph{interventional data}, as opposed to \emph{observational data} (\eg the datasets of observations $\gD$ we considered so far in previous sections).

Going back to the example of the weather and barometer, and using our new notations, we concluded that for some value of the pressure $y$, $P_{\theta}(X\mid \mathrm{do}(Y = y)) = P_{\theta}(X)$; intervening on the barometer has no effect on the weather. This is to be contrasted with the conditional distribution $P_{\theta}(X\mid Y = y) = P_{\theta}(X, Y = y) / P_{\theta}(Y = y)$, which only encodes statistical relationships; if $y$ is large, then there is a higher probability of having good weather, and is generally distinct from $P_{\theta}(X)$.

\subsection{Causal identifiability}
\label{sec:causal-identifiability}
In \cref{sec:faithfulness-identifiability}, we focused on the identifiability of the structure of a Bayesian network (or its Markov equivalence class), in the context of structure learning. But identifiability comes in widely different forms, especially when it comes to causality. For example in the context of \emph{causal inference}, identifiability could also mean the ability to express an interventional query involving ``$\mathrm{do}$'' operators with quantities that don't (\ie conditional distributions), which can then be estimated from data \citep{bareinboim2022pearlshierarchy}. More recently, there has also been a growing interest for identifiability in the context of \emph{causal representation learning} \citep{scholkopf2021causalrepresentationlearning,lachapelle2023synergiesdisentanglementsparsity}.

Going back to the question of structure identifiability, which is more aligned with the setting studied in this thesis, many of the considerations we had for Bayesian networks transfer directly to causal models as well. In particular, most of the tools for \emph{causal structure learning} (also known as \emph{causal discovery}) are derived from those we presented earlier. However, the presence of interventions may help identification of the causal structure (more precisely than at the level of a Markov equivalence class; \citealp{hauser2012interventionalmec}). This is notably useful when the dataset we use for structure learning also includes \emph{interventional data}. %

%% file: chapters/02_Probabilistic_Inference_Strctured.tex
\chapter{Probabilistic Inference over Structured Objects}
\label{chap:probabilistic-inference-structured-objects}

\begin{minipage}{\textwidth}
    \itshape
    This chapter contains material from the following papers:
    \begin{itemize}[noitemsep, topsep=1ex, itemsep=1ex, leftmargin=3em]
        \item Moksh Jain, \textbf{Tristan Deleu}, Jason Hartford, Cheng-Hao Liu, Alex Hernandez-Garcia, Yoshua Bengio (2023). \emph{GFlowNets for AI-Driven Scientific Discovery}. Digital Discovery, Royal Society of Chemistry. \notecite{jain2023gfnscientific}
        \item \textbf{Tristan Deleu}, Padideh Nouri, Nikolay Malkin, Doina Precup, Yoshua Bengio (2024). \emph{Discrete Probabilistic Inference as Control in Multi-path Environments}. Conference on Uncertainty in Artificial Intelligence (UAI). \notecite{deleu2024gfnmaxentrl}
    \end{itemize}
    \vspace*{5em}
\end{minipage}

In this thesis, we will put an emphasis on sampling from distributions defined over very structured objects, with applications to structure learning from a Bayesian perspective in the second part. We will see via multiple examples in this chapter how we can treat sampling from such distributions as a \emph{sequential decision making} problem, which can eventually be solved as a control problem, albeit in limited cases. The objective of the first part of this thesis will then be to extend these ideas so that they can be applied more broadly.

\section{Energy-based models}
\label{sec:energy-based-models}
Throughout the first part of this thesis, we are interested in the broad question of sampling from an \emph{energy-based model}\index{Energy-based model} (\gls{ebm}; \citealp{lecun2006ebms}) over a \emph{discrete} and \emph{structured} sample space \gls{samplespace}. The ``structured'' aspect of $\gX$ will be further expanded in \cref{sec:compositional-objects}. Given an energy function \gls{energy}, which we assume to be known and fixed (\ie this is an oracle that can be queried for any $x\in\gX$), our objective is to sample from the following \emph{Gibbs distribution}\index{Gibbs distribution}:
\begin{equation}
    P^{\star}(x) = \frac{\exp(-\gE(x))}{Z}.
    \label{eq:gibbs-distribution}
\end{equation}
This energy function encodes some notion of \emph{preference} for certain objects, which can come in different forms (\eg the compatibility between an image and a label, or between a sentence in English and its candidate translation in French). It is often convenient to use an energy since it can be an arbitrary function that does not necessarily need to be normalized. All the complexity related to the normalization is deferred to the \emph{partition function} $Z = \sum_{x'\in\gX}\exp(-\gE(x'))$, which is often intractable when the space $\gX$ is combinatorially large, as is the case for the majority of applications we consider here.

\subsection{Why sampling from an EBM?}
\label{sec:sampling-ebm}
Although the problem of sampling from \cref{eq:gibbs-distribution} may seem somewhat restricted at first glance, it is in fact a fundamental component for inference and learning in energy-based models.

\paragraph{Sampling-based inference} Probabilistic inference as a general principle can be cast as computing quantities of the form $\E_{P^{\star}}[f(x)]$, with an appropriately chosen function $f$, and where the expectation is taken wrt. the Gibbs distribution. For example, if we want to find the probability $P^{\star}(x \in A)$ of an object having some property $A$, then we would use $f(x) = \mathds{1}(x \in A)$ in our general form. Even if in certain specific cases exact inference may be possible \citep{wainwright2008pgm}, typically these quantities can only be approximated since the expectation requires summing over all the elements of the (combinatorially large) space $\gX$. If we have a process to obtain samples from \cref{eq:gibbs-distribution}, then we can obtain a \emph{Monte Carlo} estimate of this expectation: if $\{x^{(1)}, \ldots, x^{(N)}\}$ are iid. samples from $P^{\star}$, then
\begin{equation}
    \E_{P^{\star}}\big[f(x)\big] \approx \frac{1}{N}\sum_{n=1}^{N}f(x^{(n)}).
    \label{eq:monte-carlo-estimate}
\end{equation}
This estimate of the query of interest is unbiased, and consistent by the law of large numbers. This is an alternative to using variational inference as described in \cref{sec:variational-inference}.

\paragraph{Learning the energy function} In this paragraph only, we will assume that the energy function $\gE_{\theta}(x)$ is parametrized by some unknown $\theta$, that we would like to learn based on a dataset of observations $\gD = \{x^{(1)}, \ldots, x^{(N)}\}$. Learning the parameters $\theta$ can be done by maximizing the (average) log-likelihood of the data under the (now parametric) Gibbs distribution $P^{\star}_{\theta}(x)$
\begin{equation}
    \gL(\theta) \triangleq \E_{P_{\gD}}\big[\log P^{\star}_{\theta}(x)\big] = -\frac{1}{N}\sum_{n=1}^{N}\gE_{\theta}(x^{(n)}) - \log Z_{\theta},
    \label{eq:maximum-likelihood-ebm}
\end{equation}
where $P_{\gD}$ is the empirical distribution over the dataset $\gD$. This can be maximized using gradient-based methods, as long as one can easily compute the gradient of $\gL(\theta)$ wrt.~the parameters $\theta$. It can be shown \citep{hinton2002contrastivedivergence} that this gradient has a particularly simple form
\begin{equation}
    \nabla_{\theta}\gL(\theta) = -\E_{P_{\gD}}\big[\nabla_{\theta}\gE_{\theta}(x)\big] + \E_{P_{\theta}^{\star}}\big[\nabla_{\theta}\gE_{\theta}(x)\big].
    \label{eq:gradient-mle-ebm}
\end{equation}
While the first term can be easily computed by averaging over $\gD$, provided one can compute $\nabla_{\theta}\gE_{\theta}(x)$, the second term is an expectation under the Gibbs distribution itself, and we saw in \cref{eq:monte-carlo-estimate} that we can obtain a Monte Carlo estimate of it if we had a way to sample from $P^{\star}_{\theta}$.

But beyond the sole interest in statistics, sampling plays a crucial role in scientific discovery in general, for example in the generation of diverse proteins with specific structures and properties \citep{watson2023rfdiffusion,huguet2024foldflow2}. Aligned with scientific discovery, sampling also plays an important role in \emph{active learning}, where the most informative and promising elements are sampled for later evaluation. Finally, we cannot talk about sampling without mentioning the massive impact that \emph{generative AI} has nowadays \citep{brown2020gpt3,betker2023dalle3}, with its seemingly endless potential but also its societal risks \citep{oecdai2024risksai}.

\subsection{Sampling from an EBM with MCMC}
\label{sec:existing-approaches-sampling-ebm}
One of the most versatile tool for sampling from an EBM (or distributions known up to a normalization constant at large) is a technique called \emph{Markov chain Monte Carlo}\index{Markov chain Monte Carlo} (\gls{mcmc}; \citealp{gelfand1990mcmc}). The idea is to construct a Markov chain on the sample space $\gX$ whose stationary distribution is the Gibbs distribution in \cref{eq:gibbs-distribution}, meaning that after running the chain for long enough the samples will eventually be distributed approximately as $P^{\star}$.

A popular strategy to build such a Markov chain is the \emph{Metropolis-Hastings} algorithm \citep{metropolis1953mcmc,hastings1970mcmc}, where moves from the current state $x\in\gX$ to a new state $x'\in\gX$ are proposed by a known (and easy to sample from) distribution $Q(x'\mid x)$. This \emph{proposal distribution} is typically a distribution over local ``mutations'' of $x$, \eg adding or removing a single edge in a graph; see also \cref{sec:structure-mcmc} for an example where $P^{\star}$ is a distribution over graphs. To ensure that the stationary distribution of this chain is indeed $P^{\star}$, the new state $x'$ is only accepted ($x^{(t+1)} = x'$) randomly with probability $\min\!\big(1, A(x^{(t)}, x')\big)$, where
\begin{equation}
    A(x^{(t)}, x') \triangleq \frac{P^{\star}(x')Q(x^{(t)}\mid x')}{P^{\star}(x^{(t)})Q(x'\mid x^{(t)})} = \frac{Q(x^{(t)}\mid x')}{Q(x'\mid x^{(t)})}\exp\!\big(\gE(x^{(t)}) - \gE(x')\big),
    \label{eq:metropolis-hastings-acceptance}
\end{equation}
and otherwise, the state remains unchanged ($x^{(t+1)} = x^{(t)}$). We can observe that the acceptance probability $A$ only depends on the difference in energies between $x^{(t)}$ and $x'$, and not on the intractable partition function $Z$. For large enough $t \gg 1$, $x^{(t)}$ sampled using this process is approximately distributed as $P^{\star}$. The pseudo-code for the Metropolis-Hastings algorithm is available in \cref{alg:metropolis-hastings-algorithm}.

\begin{algorithm}[t]
    \caption{Metropolis-Hastings algorithm}
    \label{alg:metropolis-hastings-algorithm}
    \begin{algorithmic}[1]
        \Require An initial state $x^{(0)} \in \gX$, the proposal distribution $Q(x'\mid x)$.
        \For{$t \geq 0$}
            \State Sample a proposed move: $x' \sim Q(x'\mid x^{(t)})$
            \State Compute the the acceptance probability $A\!\big(x^{(t)}, x'\big)$ \cref{eq:metropolis-hastings-acceptance}
            \State Sample $u \sim U(0, 1)$
            \State Update the state:
            \Statex \qquad$\displaystyle x^{(t+1)} = \left\{\begin{array}{lll}
                x' & \textrm{if $u \leq A\!\big(x^{(t)}, x'\big)$} & \textrm{(accept)}\\
                x^{(t)} & \textrm{otherwise} & \textrm{(reject)}
            \end{array}\right.$
        \EndFor
    \end{algorithmic}
\end{algorithm}

It is important to note that the quality of the samples obtained with MCMC highly depends on how close the Markov chain is to stationarity. The time it takes to reach this state, called the \emph{mixing time}, depends on both the target distribution \cref{eq:gibbs-distribution} (and by extension, the energy function $\gE(x)$) and on the proposal distribution. For example, MCMC is known to converge slowly when the target distribution has multiple modes separated by large regions of low probability and $Q(x'\mid x)$ only proposes local moves around $x$ \citep{roberts2008understandingmcmc}. To maximize the chances of getting approximate samples of $P^{\star}$, it is common practice to discard early samples for a ``burn-in'' period. Although there is a vast literature on analysing and improving the convergence of MCMC methods \citep{swendsen1986replicaexchange,robert2018acceleratingmcmc}, an exhaustive treatment is outside the scope of this thesis. Besides mixing, MCMC also suffers from \emph{autocorrelation}: due to its sequential nature, samples may be highly correlated, making the estimation of quantities in \cref{eq:monte-carlo-estimate} less efficient.

It cannot be understated how influential this algorithm has been (and MCMC in general). It has been considered one of the top 10 algorithms of the 20th century by the IEEE magazine \emph{Computing in Science \& Engineering} \citep{beichl2000metropolistop10}, and has been the cornerstone of major scientific discoveries, including in high-energy physics and lattice field theory \citep{lautrup1985mcmcheighenergyphysics}, the estimation of the effects of monetary policies in macroeconomy (which won the Nobel prize in economics in 2005; \citealp{sims2012mcnobeleconomics}), or the modeling of complex chemical systems at different scales (\citealp{minary2010conformationalopt}; which once again won the Nobel prize in chemistry in 2013; \citealp{levitt2014mcnobelchemistry}).

\paragraph{Using gradient information} In the standard version of the Metropolis-Hastings algorithm presented above, the proposal distribution $Q(x'\mid x)$ is not aware of the energy ``surface'' around the current state $x$ (the difference in energies only appearing in \cref{eq:metropolis-hastings-acceptance}), meaning that the moves are non-adaptive. On the contrary, we could imagine proposing larger moves when $\gE(x)$ has larger variations. Suppose in this paragraph only that $\gX \subseteq \sR^{d}$ is a continuous sample space, and that the energy function is differentiable; this is a setting which has been, by and large, the most studied in the EBM literature. We could use the \emph{Langevin dynamics} in order to propose new moves, with
\begin{equation}
    Q(x'\mid x) \propto \exp\left(-\frac{1}{4\beta}\|x' - x + \beta \nabla_{x}\gE(x)\|^{2}\right),
    \label{eq:proposal-langevin-dynamics}
\end{equation}
where $\beta > 0$ is a fixed step-size. Using this proposal distribution in \cref{alg:metropolis-hastings-algorithm} is called the \emph{Metropolis-adjusted Langevin algorithm} (MALA; \citealp{roberts1996mala}). Langevin Monte Carlo has recently gained popularity thanks to its applications to score-based models \citep{song2019sorebasedmodels}, which are closely related to EBMs. However, while there has been some works adapting these ideas to discrete spaces \citep{grathwohl2021gwg,zhang2022dmala}, in general these methods cannot be applied to our setting where $\gX$ is discrete and structured since they rely on the score function $\nabla_{x}\gE(x)$, which is undefined in our case a priori.

\subsection{Compositional objects}
\label{sec:compositional-objects}
\begin{figure}[t]
    \centering
    \begin{adjustbox}{center}
    \includegraphics[width=480pt]{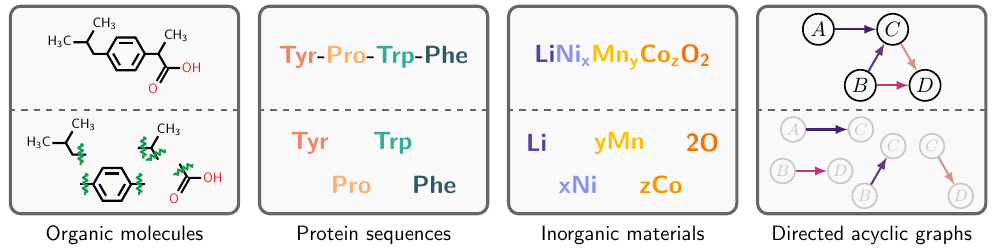}
    \end{adjustbox}
    \caption[Examples of compositional objects]{Examples of compositional objects. An organic molecule is a collection of molecular fragments assembled together. A protein sequence is a sequence of amino acids. An inorganic material is composed multiple elements (often organized as a crystal; \citealp{milaai4science2023crystalgfn}). A directed acyclic graph is a collection of edges, combined together so that they don't form any cycle.}
    \label{fig:compositional-objects}
\end{figure}

We focus our attention to cases where the sample space $\gX$ of the EBM is discrete and structured. The structure of this space will often come from some \emph{compositional} aspect of the objects to be generated, meaning that the elements $x\in\gX$ can be decomposed into pieces, and those pieces can then be reassembled to create new objects. We provide multiple examples of compositional objects and the pieces they are made of in \cref{fig:compositional-objects}. Of major interest for us in this thesis, we can for example view a directed acyclic graph (DAG) encoding the structure of a Bayesian network as a collection of edges connecting a fixed set of nodes. We keep this notion of compositionality rather vague on purpose to avoid being restrictive, and we may also consider somewhat loose notions of composition. For example in the context of natural language, a sentence may be viewed as being ``compositional'' in the sense that it is a sequence of words.

If we were to use MCMC methods in order to sample from a distribution over compositional objects, then the local ``mutations'' necessary to move in the sample space may correspond to replacing one piece with another, or adding / removing a piece from the object, \emph{provided that these moves lead to another valid object} in $\gX$. This is quite restrictive, and may lead to significant complexity to validate if a move is legal or not. For example, removing a fragment from a molecule may or may not be physically plausible. Applying MCMC in the context of DAGs will be detailed in \cref{sec:structure-mcmc}.

\section{Marginal sampling as sequential decision making}
\label{sec:marginal-sampling-sequential-decision-making}
Instead of sampling an object directly from $P^{\star}$, possibly by moving in $\gX$ with MCMC, we could leverage the compositional structure of each object $x\in\gX$ and construct it \emph{from scratch} by adding pieces together sequentially. In this section, we will show with multiple examples how we can sample from a complex distribution over compositional objects through a sequence of (simpler) decisions. We will also introduce notions of graph theory and probabilities which will be central in the rest of this thesis.

\subsection{The Galton board}
\label{sec:galton-board}
\begin{figure}[t]
    \centering
    \begin{adjustbox}{center}
        \includegraphics[width=480pt]{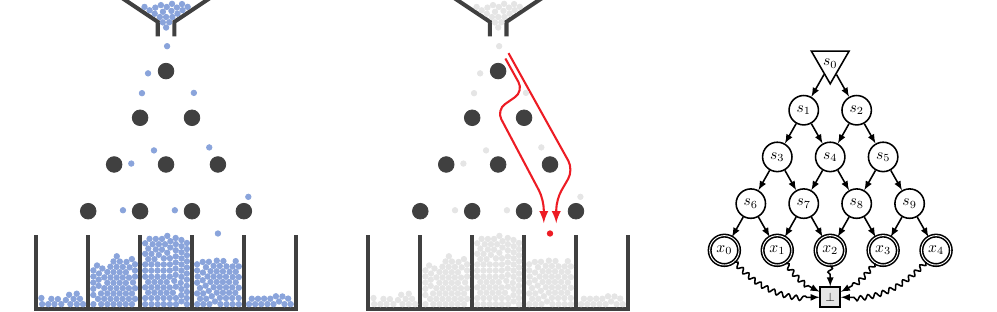}%
    \end{adjustbox}
    \begin{adjustbox}{center}%
        \begin{subfigure}[b]{160pt}%
            \caption{}%
            \label{fig:galton-board-1}%
        \end{subfigure}%
        \begin{subfigure}[b]{160pt}%
            \caption{}%
            \label{fig:galton-board-2}%
        \end{subfigure}%
        \begin{subfigure}[b]{160pt}%
            \caption{}%
            \label{fig:galton-board-3}%
        \end{subfigure}%
    \end{adjustbox}
    \caption[The Galton board]{The Galton board. (a) Visualization of the Galton board, where blue beads fall down the apparatus into one of the five bins. (b) A bead falling into a specific bin may take multiple paths, bouncing off a different sequence of pins. (c) Representation of the Galton board as a pointed DAG $\gG$. The intermediate states $s_{i}$ represent the different pins organized in quicunx, and the terminating states $x_{k}$ represent the bins.}
    \label{fig:galton-board}
\end{figure}

In his book on \emph{Natural Inheritance} in 1889, Francis Galton proposed a mechanical device to illustrate what he called ``the curve of frequency''. This device, called the \emph{Galton board} and illustrated in \cref{fig:galton-board-1}, is composed of a funnel at the top guiding beads into a series of pins disposed in a quincunx pattern. As the beads fall down due to gravity, they will bounce off each pin either to the left or to the right, going down each row to finally end up in one of the bins at the bottom of the apparatus. \citet{galton1889naturalinheritance} observed that as the numbers of beads and bins increase, the shape outlined by the beads at the bottom of the board resembles a ``\emph{Normal curve of frequency}''; this is an illustration of what we now know as the \emph{central limit theorem}.

Beyond this asymptotic curiosity, and the unsettling conclusions Galton drew from it, this model is simple enough that we can analyze its behavior with a fixed number of $n+1$ bins denoted by $\{0, \ldots, n\}$. Assuming that a certain bead has probability $p$ of going left and $1 - p$ of going right each time it bounces off a pin (in reality, $p = 0.5$), it is well known that the bins in which the bead will fall is a random variable $B$ that is distributed as a \emph{binomial distribution}, where
\begin{equation}
    P(B = k) = \binom{n}{k}p^{k}(1-p)^{n-k}.
    \label{eq:binomial-distribution}
\end{equation}
While the binomial distribution could be written in terms of an energy function and its sample space $\gX = \{0, \ldots, n\}$ is discrete, this example deviates from our setting highlighted in \cref{sec:compositional-objects} in that $\gX$ has no apparent structure. Nevertheless, this serves as an intuitive illustration of a general principle that we will use extensively in the rest of this thesis:

\begin{minipage}{\textwidth}
    \centering
    \begin{minipage}{0.9\textwidth}
        \itshape
        We have a process to get samples from a more complex distribution as a sequence of multiple simple decisions.
    \end{minipage}
\end{minipage}

In our example, the distribution we sample from is arguably complex, since a bead falls in a certain bin with distribution $\mathrm{Binomial}(n, p)$, the sequence of decisions corresponding to going left at each pin with probability $p$ (or $\mathrm{Bernoulli}(p)$).

A second general principle is that we are interested in the \emph{marginal distribution} over which bins the beads fall into, rather than the distribution over the exact paths the beads may take (\ie how they get to a certain bin).

\begin{minipage}{\textwidth}
    \centering
    \begin{minipage}{0.9\textwidth}
        \itshape
        We are only interested in the outcomes of the sequential process, not in the precise sequences of decisions themselves.
    \end{minipage}
\end{minipage}

As an example when $p = 0.5$, beads tend to fall in the middle bins with higher probability since there are many different paths they could follow to reach those bins (see also \cref{fig:galton-board-2}); this is reflected by the binomial coefficient in \cref{eq:binomial-distribution} counting the number of paths to bin $k$. Playing around with the distributions of each individual decision leads to various marginals at the end of this process; our objective will be eventually to carefully tune each of these distributions so that this marginal matches a complex distribution of interest.

\subsection{Elements of graph theory}
\label{sec:elements-graph-theory}
To formalize the example of the Galton board, and in preparation for more complex scenarios, we need to introduce a few notions of graph theory. We represent each decision point (\eg each pin) as a state $s$ in a state space $\gS$, connected together as a \emph{pointed directed acyclic graph} to represent the outcomes of each decision (see also \cref{fig:galton-board-3}).

\begin{definition}[Pointed DAG]
    \label{def:pointed-dag}\index{Directed acyclic graph!Pointed DAG}
    Let $\gls{augmentedstatespace} = \gS \cup \{\terminal\}$ be a state space \gls{statespace} augmented with a terminal state $\terminal\notin\gS$, and let $\gA \subseteq \gS\times\widebar{\gS}$ be a set of edges connecting these states. A (connected) directed acyclic graph (DAG) \gls{pointeddag} is called a \emph{pointed DAG} if there exists a unique \emph{initial state} $s_{0}\in\gS$ such that \gls{initialstate} is a root of the DAG (\ie it has no parent), and \gls{terminalstate} is the unique state with no child.
\end{definition}

Unlike most works on generative flow networks \citep{bengio2023gflownetfoundations}, where the terminal state is denoted by ``$s_{f}$'', throughout this thesis we will use ``$\terminal$'' to represent the terminal state, as in \citet{lahlou2023continuousgfn}. We use a symbol distinct from the notation ``$s$'' (used for the states of $\gS$) to emphasize that the terminal state is \emph{not} an element of the state space (albeit an element of the \emph{augmented} space $\widebar{\gS}$). We will always assume that all the states of $\gG$ are accessible from $s_{0}$, and that the terminal state $\terminal$ is also accessible from all the states of $\gG$. We denote by \gls{parents} and \gls{children} respectively the parents and the children of a state $s\in\widebar{\gS}$ in $\gG$.

The bins of the Galton board are also represented as states in this graph (\ie $\gX \subseteq \gS$); in this example, those states only have a single decision leading to the terminal state, signaling the end of the sequential process. The states where termination is possible are called \emph{terminating states}.

\begin{definition}[Terminating states]
    \label{def:terminating-states}\index{Terminating state}
    Let $\gG = (\widebar{\gS}, \gA)$ be a pointed DAG. The set of \emph{terminating states} $\gX \subseteq \gS$ of $\gG$ is defined as the parents of the terminal state in $\gG$: $\gX \triangleq \parents_{\gG}(\terminal)$.
\end{definition}

Based on our example in the previous section, we can see why the choice of $\gX$, the sample space of the distribution of interest, to denote the set of terminating states is not coincidental. While in that example all the terminating states have $\terminal$ as their unique child (\ie we are forced to terminate once the bead reaches one of the bins), it is important to note that in general reaching a terminating state does not necessarily force termination, only taking the transition $x \rightarrow \terminal$ to the terminal state does. In other words, it is completely possible that a pointed DAG contains edges from a terminating state to some non-terminal $s \in \gS$; see \cref{fig:gflownet-blogpost} further for an example. The sequences of decisions taken to reach a certain terminating state (\eg the pins a certain bead bounced off of) are encoded through \emph{trajectories} in $\gG$.

\begin{definition}[Complete trajectories]
    \label{def:complete-trajectories}
    Let $\gG = (\widebar{\gS}, \gA)$ be a pointed DAG, with $s_{0}$ its initial state. The trajectory \gls{completetrajectory} is called a \emph{complete trajectory} if for all $t \geq 0$, $s_{t} \rightarrow s_{t+1} \in \gA$ (with the convention $s_{T+1} = \terminal$). The set of all the complete trajectories over $\gG$ is denoted by \gls{completetrajectories}.
\end{definition}
It is clear that for a complete trajectory $\tau = (s_{0}, s_{1}, \ldots, s_{T}, \terminal)$, $s_{T}\in \gX$ is a terminating state. In some cases, it will also be convenient to talk about \emph{partial trajectories}, which are defined similarly except that the start and end points of those trajectories are not necessarily the initial state $s_{0}$ or the terminal state $\terminal$ respectively. We will use ``\gls{partialtrajectories}'' to denote the set of partial trajectories from $s_{m} \in \gS$ to $s_{n}\in\widebar{\gS}$ ($s_{n}$ may be the terminal state); in particular, it is clear that $\gT \equiv s_{0}\rightsquigarrow \terminal$. If $\tau$ and $\tau'$ are two partial trajectories such that the end point of $\tau$ matches the start point of $\tau'$, we will denote by \gls{concatenatetrajectories} the concatenation of both trajectories. We will also use the abuse of notation ``$s \rightarrow s'\in \gG$'' when the context is clear to denote $s \rightarrow s' \in \gA$, where $\gA$ is the set of edges of $\gG$. %

\paragraph{Notations} We will often visualize a pointed DAG $\gG$ as a graphical structure like the one shown in \cref{fig:galton-board-3}. We will use standard circles \raisebox{-1pt}{\tikz{\node[thick, minimum size=0.9em, inner sep=0pt, circle, draw=black] {};}} to denote the states in $\gS$, and double circles \raisebox{-2pt}{\tikz{\node[thick, minimum size=0.9em, inner sep=0pt, circle, draw=black, double] {};}} to highlight the terminating states in $\gX$. We identify the initial state $s_{0}$ with a triangular node \raisebox{-1pt}{\tikz{\node[thick, minimum size=0.8em, inner sep=0pt, isosceles triangle, isosceles triangle apex angle=60, draw=black] {};}}, and the terminal state as \raisebox{-1pt}{\tikz{\node[thick, inner sep=2pt, draw=black, fill=gray!20, scale=0.9] {$\bot$};}}. We use \raisebox{0pt}{\tikz{\draw[decorate, decoration={snake=coil, post length=5pt, segment length=5pt, amplitude=2pt}, -latex, thick] (0, 0) -- ++(0:1.8em);}} to denote the terminating edges leading to $\terminal$. Note that when the context is clear (\eg in \cref{fig:molecule-sequential-generation}), we may duplicate the node \raisebox{-1pt}{\tikz{\node[thick, inner sep=2pt, draw=black, fill=gray!20, scale=0.9] {$\bot$};}} for visual clarity, even though it must be understood as being \emph{one single state $\terminal$}. We will give a summary of these notations in the annotated \cref{fig:gflownet-blogpost-pointed-dag}. In the second part of this thesis, we will relax these notations (\eg the terminal state will be implicit, as in \cref{fig:dag-gflownet}).

\subsection{Forward transition probabilities}
\label{sec:forward-transition-probabilities}
In addition to the structure of the state space encoding the outcomes of each decision made sequentially, we also need to specify \emph{how} these decisions are made. For example in the Galton board, each bead had probability $p$ of falling to the left if it bounced off a pin. We can formalize this with a probability distribution to transition from any state to the next states (\eg transitioning with a $\mathrm{Bernoulli}(p)$).

\begin{definition}[Consistent forward transition probabilities]
    \label{def:consistent-forward-transition-probability}\index{Transition probability!Forward transition probabilities|textbf}
    Let $\gG = (\widebar{\gS}, \gA)$ be a pointed DAG. A function $\gls{forwardtransitionprobability}: \gS \rightarrow \Delta(\children_{\gG})$ is called a \emph{forward transition probability} distribution consistent with $\gG$ if for any $s\in\gS$, $P_{F}(\cdot\mid s)$ is a properly defined distribution over $\children_{\gG}(s)$, \ie for all $s'\in\children_{\gG}(s)$, $P_{F}(s'\mid s) \geq 0$ and
    \begin{equation}
        \sum_{s'\in\children_{\gG}(s)}P_{F}(s'\mid s) = 1.
        \label{eq:consistent-forward-transition-probability}
    \end{equation}
\end{definition}
In the definition above, for some state $s\in\gS$, we use the notation ``\gls{simplexchildren}'' to denote the space of probability distributions over the children of $s$ in $\gG$. In the context of reinforcement learning, a forward transition probability is also called a \emph{policy} \citep{puterman1994mdp}, provided the underlying environment is deterministic (\ie there is a one-to-one mapping between actions and next states). Although we will use ``policies'' in \cref{sec:probabilistic-inference-control-problem}, the term ``forward transition probability'' will be prevalent in \cref{chap:flow-networks,chap:generative-flow-networks}. We can also naturally extend the definition of $P_{F}$ for some partial trajectory $\tau = (s_{m}, s_{m+1}, \ldots, s_{n})$ as
\begin{equation}
    P_{F}(\tau) = \prod_{t=m}^{n-1}P_{F}(s_{t+1}\mid s_{t}).
    \label{eq:forward-transition-probability-trajectory}
\end{equation}
Consistent forward transition probabilities have the interesting property that they also induce a distribution over partial trajectories starting at any state $s$ and ending at $\terminal$.
\begin{lemma}[Distribution over suffixes]
    \label{lem:PF-distribution-suffix}
    Let $\gG = (\widebar{\gS}, \gA)$ be a pointed DAG, and let $P_{F}: \gS \rightarrow \Delta(\children_{\gG})$ be a forward transition probability consistent with $\gG$. For any state $s\in\gS$, $P_{F}$ induces a distribution over partial trajectories $\tau$ starting at $s$ and ending at the terminal state $\terminal$, \ie $P_{F}(\tau) \geq 0$ and
    \begin{equation}
        \sum_{\tau: s\rightsquigarrow \terminal}P_{F}(\tau) = 1.
        \label{eq:PF-distribution-suffix}
    \end{equation}
\end{lemma}

\begin{proof}
    Since $P_{F}$ is a forward transition probability, we clearly have $P_{F}(\tau) \geq 0$. We can prove \cref{eq:PF-distribution-suffix} by strong induction on the maximum length of the partial trajectories. For any state $s\in \gS$, let $d_{s}$ denote the maximum length of a partial trajectory from $s$ to the terminal state $\terminal$.
    
    \emph{Base case:} Since $\gG$ is a pointed DAG, there exists at least a state $s\in\gS$ such that $d_{s} = 1$. In that case, $\terminal$ is the only child of $s$ (otherwise there would be a longer trajectory from $s$ to $\terminal$, making $d_{s} > 1$), and there exists a single partial trajectory of the form $s\rightsquigarrow \terminal$, which is the transition $s \rightarrow \terminal$. Therefore
    \begin{equation}
        \sum_{\tau: s\rightsquigarrow \terminal}P_{F}(\tau) = P_{F}(\terminal \mid s) = 1.
    \end{equation}

    \emph{Induction step:} Suppose that \cref{eq:PF-distribution-suffix} is valid for all states $s'$ such that $d_{s'} \leq d$ for some $d > 0$. Since $\gG$ is a pointed DAG, there exists a state $s\in\gS$ such that $d_{s} = d+1$. We can decompose the set of all partial trajectories from $s$ into a disjoint union of partial trajectories going through each child $s' \in \children_{\gG}(s)$ as
    \begin{equation}
        \{\tau: s\rightsquigarrow \terminal\} = \bigcup_{s'\in\children_{\gG}(s)}\{(s\rightarrow s') \oplus \tau'\mid \tau': s'\rightsquigarrow \terminal\}.
    \end{equation}
    Since all the partial trajectories of the form $s' \rightsquigarrow \terminal$ have a maximum depth $d_{s'} \leq d$, we can apply the induction step to them. Using the decomposition above, we get
    \begin{equation}
        \sum_{\tau: s\rightsquigarrow \terminal}P_{F}(\tau) = \sum_{s'\in\children_{\gG}(s)}P_{F}(s'\mid s)\sum_{\tau': s'\rightsquigarrow \terminal}P_{F}(\tau') = \sum_{s'\in\children_{\gG}(s)}P_{F}(s'\mid s) = 1.
    \end{equation}
    This proves that \cref{eq:PF-distribution-suffix} is also valid for a state $s$ with maximum trajectory length $d+1$.
\end{proof}

A direct consequence of this lemma is that $P_{F}$ induces in particular a probability distribution over complete trajectories (\ie starting at $s_{0}$). However as we noted in \cref{sec:galton-board}, we are interested in a distribution over terminating states (\ie a distribution over the bins in which the beads fall) derived from this distribution over complete trajectories, and not $P_{F}(\tau)$ itself. This is called the \emph{terminating state probability} distribution.

\begin{definition}[Terminating state probability]
    \label{def:terminating-state-probability}\index{Terminating state probability|textbf}
    Let $\gG = (\widebar{\gS}, \gA)$ be a pointed DAG, and $\gX \subseteq \gS$ be the set of terminating states in $\gG$. Let $P_{F}: \gS \rightarrow \Delta(\children_{\gG})$ be an arbitrary forward transition probability consistent with $\gG$. The \emph{terminating state probability} distribution \gls{terminatingstateprobability} (associated with $P_{F}$) is defined for all $x\in\gX$ as
    \begin{equation}
        P_{F}^{\top}(x) \triangleq \sum_{\tau: x\rightarrow\terminal\in\tau}P_{F}(\tau) = \sum_{\tau: s_{0}\rightsquigarrow x}\prod_{t=0}^{T_{\tau}}P_{F}(s_{t+1}\mid s_{t}),
        \label{eq:terminating-state-probability}
    \end{equation}
    with the conventions $s_{T_{\tau}+1} = \terminal$, and necessarily $s_{T_{\tau}} = x$ for all complete trajectories $\tau$ such that $x\rightarrow \terminal \in \tau$.
\end{definition}
Putting it in words, the terminating state probability distribution is the marginal distribution of $P_{F}(\tau)$ over trajectories terminating at $x$ (\ie $x\rightarrow \terminal \in \tau$). It is important to note that this is different from the marginal over trajectories simply going through $x$, since these are not necessarily guaranteed to directly take the transition $x\rightarrow\terminal$. For example in \cref{fig:gflownet-blogpost} further, although the trajectory $(s_{0}, s_{2}, x_{4}, x_{6}, \terminal)$ passes through the terminating state $x_{4}$, it would not contribute to $P_{F}^{\top}(x_{4})$. The following proposition guarantees that for any forward transition probability $P_{F}$, its corresponding terminating state probability is a properly defined distribution over the set of terminating states $\gX$.
\begin{proposition}
    \label{prop:terminating-state-proper-distribution}
    The terminating state probability distribution $P_{F}^{\top}$ is a well-defined probability distribution over the terminating states, in the sense that for all $x\in\gX$, $P_{F}^{\top}(x) \geq 0$, and $\sum_{x\in\gX}P_{F}^{\top}(x) = 1$.
\end{proposition}

\begin{proof}
    Since $P_{F}$ is a forward transition probability, it is clear that $P_{F}^{\top}(x) \geq 0$ for any $x\in\gX$. The normalization is a consequence of \cref{lem:PF-distribution-suffix} applied to the initial state $s_{0}$:
    \begin{equation}
        \sum_{x\in\gX}P_{F}^{\top}(x) = \sum_{x\in\gX}\sum_{\tau: x\rightarrow\terminal\in\tau}P_{F}(\tau) = \sum_{\tau: s_{0}\rightsquigarrow \terminal}P_{F}(\tau) = 1.
    \end{equation}
\end{proof}
Interpreting $P_{F}^{\top}$ as a marginal distribution suggests a natural procedure for sampling from it, detailed in \cref{alg:sampling-terminating-state-probability}: (1) sample a complete trajectory following the forward transition probabilities $P_{F}$, and (2) return only the terminating state reached at the end of the trajectory, right before transitioning to $\terminal$. This is the same procedure as sampling from $P_{F}^{\top} = \mathrm{Binomial}(n, p)$ associated with $P_{F}(\cdot \mid s) = \mathrm{Bernoulli}(p)$ by letting a bead fall down the Galton board, and returning the bin in which it eventually fell into. Since $\gG$ is acyclic, this procedure is guaranteed to terminate in time $O(|\gS|)$ (worst case). However in many interesting cases, the ``depth'' of all the terminating states from the initial state $s_{0}$ will be significantly smaller than the total number of states, as in \cref{fig:galton-board}, but even more so when the state space $\gS$ has some compositional aspect which is reflected in the structure of $\gG$.

\begin{algorithm}[t]
    \caption{Sampling from the terminating state probability distribution $P_{F}^{\top}$ -- \cref{def:terminating-state-probability}}
    \label{alg:sampling-terminating-state-probability}
    \begin{algorithmic}[1]
        \Require A pointed DAG $\gG = (\widebar{\gS}, \gA)$, a forward transition probability $P_{F}: \gS \rightarrow \Delta(\children_{\gG})$.
        \Ensure A sample $x \sim P_{F}^{\top}$ of the terminating state probability distribution associated with $P_{F}$.
        \State Initialize the state: $s_{0}$
        \Repeat
            \State Sample the next state: $s_{t+1} \sim P_{F}(\cdot\mid s_{t})$
        \Until{$s_{T+1} = \terminal$}
        \State \Return $x \equiv s_{T}$
    \end{algorithmic}
\end{algorithm}

\subsection{Generation of small organic molecules}
\label{sec:generation-small-organic-molecules}
Even though the example of the Galton board in \cref{sec:galton-board} is interesting to give intuitions about how a sequential process can induce a terminating state probability distribution, it is limited in that $\gX$ is not a structured sample space. In this section, we consider a more realistic scenario where we would like to sample from a distribution over small organic molecules (\ie molecules containing C-H or C-C bonds). Molecules have a natural compositional structure as a combination of multiple fragments together, as illustrated in \cref{fig:compositional-objects}. Since the number of possible molecules is extremely large \citep{fink2005moleculenumber}, defining a distribution by assigning a probability for each individual molecule is practically impossible, due to the normalization constraint. It would be far simpler to just \emph{ignore} the normalization, and assign a certain value for each molecule. In that case, the energy function $\gE(x)$ in \cref{eq:gibbs-distribution} may correspond to the binding affinity with a target protein for example, in order to generate molecules that are more likely to bind to this protein (\eg \citet{bengio2021gflownet} used the binding affinity to the soluble epoxide hydrolase (sEH) protein, a well-studied protein playing a role in respiratory and heart disease \citep{imig2009seh}). %

\begin{figure}[t]
    \centering
    \begin{adjustbox}{center}
        \includegraphics[width=480pt]{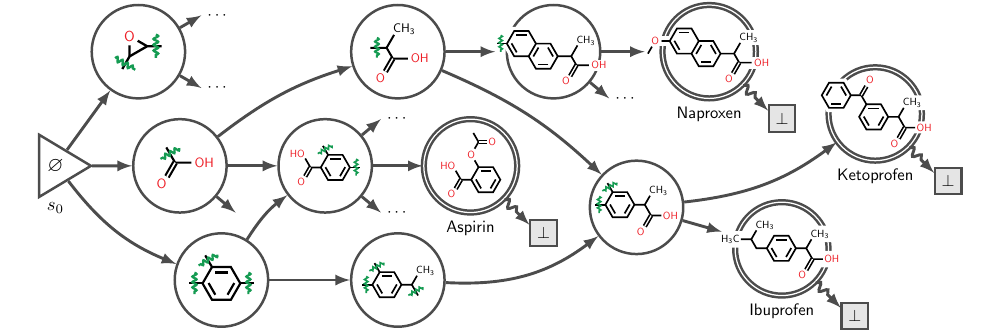}
    \end{adjustbox}
    \caption[Pointed DAG for the generation of small organic molecules]{Pointed DAG for the generation of small organic molecules. At each step of generation, a new molecular fragment chosen from a fixed vocabulary of fragments is added at some end of the partial molecule. The process eventually leads to fully formed molecules (\eg aspirin), where termination is possible. Following a forward transition probability distribution $P_{F}$ defined on this graph yields a terminating state probability distribution $P_{F}^{\top}$ over fully formed molecules.}
    \label{fig:molecule-sequential-generation}
\end{figure}

Instead of using MCMC methods for sampling from this distribution \citep{seff2019gibbsmolecules,xie2021mars}, we could construct a molecule \emph{from scratch} by adding one fragment at a time \citep{you2018moleculegeneration,bengio2021gflownet}, as shown in \cref{fig:molecule-sequential-generation}. Starting from the empty state, we choose a fragment from some vocabulary to be added to a (possibly partial) molecule at each step, until we reach the terminal state. This example highlights a major difference with MCMC: the sequence of decisions leading to the generation of a full molecule may go through partial molecules that are not physically plausible, whereas MCMC methods would only allow moves from a full molecule to another full molecule (\ie moves proposed by $Q(x'\mid x)$ in \cref{alg:metropolis-hastings-algorithm} are restricted to valid elements of $\gX$). Using the terminology introduced in previous sections, the set of terminating states $\gX$ corresponds to full molecules, and $\gS\backslash \gX$ is the set of partial molecules.

The advantage of having compositional objects, as noted in \cref{sec:compositional-objects}, becomes apparent here: having substructures shared between different molecules means that we have many states (decision points) along the complete trajectories to any full molecule, and therefore the complexity of the target Gibbs distribution can be broken down into a sequence of much simpler distributions ($P_{F}(\cdot \mid s)$). This also reduces significantly the number of steps required to sample a full molecule using \cref{alg:sampling-terminating-state-probability}. However, there is a trade-off between the length of the complete trajectories and the complexity of the transition probability at each state: while we could imagine having a naive pointed DAG $\gG$ where $s_{0}$ is directly connected to all the elements of $\gX$ (\eg all the full molecules), this would make $P_{F}(\cdot \mid s_{0})$ extremely complex (in fact, we would be back to finding $P^{\star}$ directly). We will see in \cref{sec:probabilistic-inference-control-problem} and subsequent chapters how to find a forward transition probability whose corresponding terminating state distribution $P_{F}^{\top}$ matches \cref{eq:gibbs-distribution} for a given energy function $\gE(x)$.

\subsection{Autoregressive sampling from a factor graph}
\label{sec:sampling-factor-graphs}
\begin{figure}[t]
    \begin{adjustbox}{center}
        \includegraphics[width=480pt]{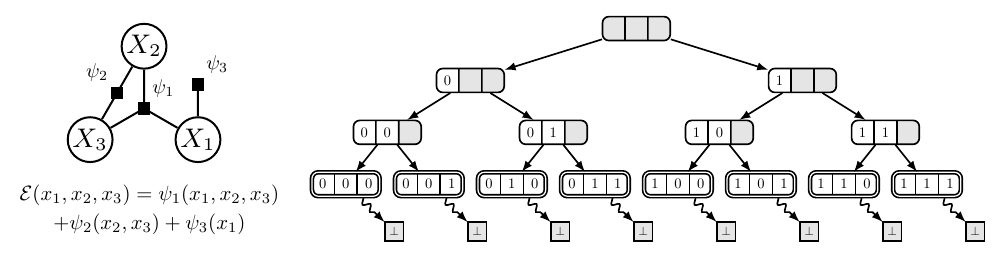}
    \end{adjustbox}
    \caption[Autoregressive sampling from a discrete factor graph]{Autoregressive sampling from a discrete factor graph, viewed from the perspective of sequential generation in a pointed DAG. (Left) An example of a factor graph over 3 binary random variables $X_{1}$, $X_{2}$, and $X_{3}$, with 3 factors $\psi_{1}$, $\psi_{2}$, and $\psi_{3}$. The energy function associated with the distribution represented by this factor graph is the sum of its factors. (Right) Representation of the sequential generation of a full assignment of the variables $(x_{1}, x_{2}, x_{3})$ as a pointed DAG $\gG$. $\terminal$ is displayed as separate nodes for clarity only, and should be understood as a single state.}
    \label{fig:treesample-intro}
\end{figure}

As a final example, this time anchored in the field of probabilistic modeling, we consider the problem of inference in a factor graph as introduced by \citet{buesing2020treesample}. Suppose that we have $d$ discrete random variables $(X_{1}, \ldots, X_{d})$, each $X_{i}$ taking values in $\{1, \ldots, K\}$, meaning that the sample space $\gX = \{1, \ldots, K\}^{d}$ is exponentially large. We assume that the joint distribution of these random variables is given by a \emph{factor graph}\index{Factor graph}, which can be expressed as a Gibbs distribution like in \cref{eq:gibbs-distribution}, with the energy function
\begin{equation}
    \gE(x_{1}, \ldots, x_{d}) = \sum_{m=1}^{M}\psi_{m}(x_{[m]}).
    \label{eq:energy-factor-graph}
\end{equation}
Each factor $\psi_{m}$ is a known function that only depends on a subset of random variables (the values of which are denoted by ``$x_{[m]}$''). A factor graph can be represented as a bipartite graph whose nodes are the random variables $\{X_{1}, \ldots, X_{d}\}$ on the one hand, and the factors $\{\psi_{1}, \ldots, \psi_{M}\}$ on the other hand; see \cref{fig:treesample-intro} for an example of a factor graph with $d=3$ variables and $M=3$ factors connecting them.

Inference in factor graphs is a very well studied topic, whether it is exact \citep{dechter1999bucketelimination} or approximate \citep{murphy1999loopybp,kschischang2001loopybp,desa2015gibbsfactorgraph}. As an alternative, and similar to our previous examples, \citet{buesing2020treesample} treated sampling from this factor graph as a sequential decision making problem, where the value of each variable is assigned one at a time following a fixed order. \cref{fig:treesample-intro} shows the pointed DAG $\gG$ for a factor graph with binary variables ($K = 2$), where we first assign the value of $X_{1}$, then $X_{2}$, and finally $X_{3}$. Therefore to obtain a full sample from $P^{\star}(x)$, this process requires a sequence of exactly $d$ steps. Interestingly, and this time unlike our previous examples, $\gG$ has a tree structure (with the exception of the terminal state $\terminal$, whose parents are always all the terminating states in $\gX$ by \cref{def:terminating-states}), which is typically the case for autoregressive sequence generation tasks \citep{bachman2015datagensequential,weber2015virl,angermueller2020dynappo,jain2022gfnbioseq,feng2022metarlbo,rafailov2024fromrtoqstar}.

\section{Probabilistic inference as a control problem}
\label{sec:probabilistic-inference-control-problem}
Continuing with our example in \cref{sec:sampling-factor-graphs}, if the forward transition probability $P_{F}$ was known, following this distribution would yield samples from the terminating state distribution $P_{F}^{\top}$. However, we don't want to sample from any distribution, but very specifically from the Gibbs distribution with the energy function in \cref{eq:energy-factor-graph}. How can we find $P_{F}$ so that $P_{F}^{\top}$ matches this distribution of interest, for a given energy function $\gE(x)$? In this section, we will show that we can treat this as a control problem within the framework of entropy-regularized reinforcement learning \citep{ziebart2010maxent,fox2016tamingnoiserl}. We will use ``policies'' and denote them by ``$\pi$'' (instead of ``forward transition probabilities'' and ``$P_{F}$'') to be more aligned with the naming conventions in reinforcement learning.

\subsection{Maximum entropy reinforcement learning}
\label{sec:maximum-entropy-rl}
We recall some notions of (maximum entropy) reinforcement learning \citep{sutton2018introrl}, with notations and concepts adapted to our setting. We consider a finite-horizon Markov Decision Process (\gls{mdp}) \gls{mdpmath}, where $\gG=(\widebar{\gS}, \gA)$ is a pointed DAG specifying the structure of the MDP, and $r$ is a specific reward function defined below. $\gS$ plays the role of the state space and $\gA$ the action space of the MDP, both being discrete and finite. The MDP is deterministic, in the sense that we are guaranteed to transition to a certain state $s'$ from a state $s\in\gS$ after following an action $a = s\rightarrow s' \in \gA$. Since $\gG$ is a pointed DAG, all the (complete) trajectories in this MDP start at the initial state $s_{0}$, and are guaranteed to eventually end at the terminal state $\terminal$ (finite-horizon). We also assume that we are in an undiscounted regime, meaning that the discount factor $\gamma = 1$.

Since the MDP is deterministic, we can identify the action leading to a new state $s'$ from $s$ with the transition $s \rightarrow s' \in \gG$ itself. As such, we will write all quantities involving state-action pairs $(s, a)$ where $a \in \gA$ (\eg the rewards, or the state-action values) directly in terms of $(s, s')$ instead. The reward function $r(s, s')$ of this MDP is defined such that the sum of rewards along a complete trajectory (also called the \emph{return}) only depends on the energy of the terminating state it reaches: for a trajectory $\tau = (s_{0}, s_{1}, \ldots, s_{T}, \terminal)$, we have
\begin{equation}
    \sum_{t=0}^{T}r(s_{t}, s_{t+1}) = -\gE(s_{T}),
    \label{eq:reward-soft-mdp}
\end{equation}
with the convention $s_{T+1} = \terminal$. In particular, this covers the case of a sparse reward that is received only at the end of the trajectory (\ie $r(s_{T}, \terminal) = -\gE(s_{T})$, and zero everywhere else). In reinforcement learning, the objective is in general to find an \emph{optimal policy} $\pi_{\mathrm{RL}}^{\star}(s'\mid s)$ that maximizes the expected return
\begin{equation}
    \pi^{\star}_{\mathrm{RL}} = \argmax_{\pi} \E_{\tau}\Bigg[\sum_{t=0}^{T} r(s_{t}, s_{t+1})\Bigg],
    \label{eq:reinforcement-learning-objective}
\end{equation}
where the expectation is taken wrt. the distribution $\pi(\tau)$ over complete trajectories induced by $\pi$ (see \cref{sec:forward-transition-probabilities}). With our particular choice of reward function in \cref{eq:reward-soft-mdp}, this maximization would correspond to finding a policy (that might be deterministic) that would lead to a state $x\in\gX$ with the lowest energy here, or equivalently finding a mode of the Gibbs distribution $P^{\star}$. This differs from our initial goal of \emph{sampling} from the whole distribution $P^{\star}$, and not only getting the objects with highest probability.

To account for possibly suboptimal objects, we need to regularize the objective in \cref{eq:reinforcement-learning-objective}. In maximum entropy reinforcement learning\index{Reinforcement learning!Maximum entropy reinforcement learning} (\gls{maxentrl}; \citealp{ziebart2010maxent,fox2016tamingnoiserl}), our objective is instead to search for a \emph{stochastic} policy that also maximizes the expected return, and the entropy $\gH(\pi(\cdot \mid s)) = -\sum_{s'\in\children_{\gG}(s)}\pi(s'\mid s)\log \pi(s'\mid s)$ of the policy $\pi$ in state $s$, in order to encourage diversity of the actions taken:
\begin{equation}
    \pi^{\star}_{\maxent} = \argmax_{\pi}\E_{\tau}\Bigg[\sum_{t=0}^{T}r(s_{t}, s_{t+1}) + \gH(\pi(\cdot \mid s_{t}))\Bigg],
    \label{eq:maxent-rl-objective}
\end{equation}
where the expectation is again taken wrt. $\pi(\tau)$. In general, there may be a temperature parameter $\alpha > 0$ that controls the importance of the entropy regularization relative to the original objective, which we assume to be equal to $1$ here for simplicity; we will come back to the impact of the temperature parameter in \cref{chap:gflownet-maxent-rl}. This type of entropy regularization falls into the broader domain of \emph{regularized MDPs} \citep{geist2019regularizedmdps}, and we will see another example of regularization based on the relative entropy in \cref{sec:relative-entropy-regularization-maxent-rl}. MaxEnt RL has been proven to be particularly beneficial for improving exploration \citep{haarnoja2017sql} and for robust control under model misspecification \citep{eysenbach2022maxentrlrobust}.

\citet{haarnoja2017sql} showed that the policy maximizing the objective in \cref{eq:maxent-rl-objective} can be written explicitly as $\pi^{\star}_{\maxent}(s'\mid s) = \exp\big(Q^{\star}_{\soft}(s, s') - V^{\star}_{\soft}(s)\big)$, where $Q^{\star}_{\soft}$ is the soft state-action value function and $V^{\star}_{\soft}$ the soft state value function which satisfy the following \emph{soft Bellman optimality equations} (adapted to our setting, \ie undiscounted and deterministic MDP):
\begin{align}
    Q^{\star}_{\soft}(s, s') &= r(s, s') + V^{\star}_{\soft}(s')\label{eq:soft-bellman-optimality-equations-intro}\\
    V^{\star}_{\soft}(s) &= \log \sum_{s''\in\children_{\gG}(s)}\exp\big(Q^{\star}_{\soft}(s, s'')\big).\label{eq:soft-bellman-optimality-equations-V-intro}
\end{align}
These equations differ from the standard Bellman optimality equations \citep{sutton2018introrl} in that the the greedy operation appearing in the optimal policy $\pi^{\star}_{\mathrm{RL}}$ is replaced by a softmax operation in $\pi^{\star}_{\maxent}$, and the ``$\max$'' operation in the optimal state value function $V^{\star}$ is replaced by the log-sum-exp in \cref{eq:soft-bellman-optimality-equations-V-intro}.

\subsection{Sampling terminating states from the soft MDP}
\label{sec:sampling-terminating-states-soft-mdp}
From the literature on \emph{control as inference} \citep{ziebart2008maxentirl,levine2018controlasinference}, it can be shown that the policy maximizing the MaxEnt RL objective in \cref{eq:maxent-rl-objective} induces a distribution over dynamically consistent trajectories that depends on the sum of rewards obtained along the trajectory.

\begin{proposition}
    \label{prop:maxent-rl-distribution-propto-return}
    Let $\gM$ be a deterministic \& undiscounted finite-horizon MDP. The policy $\pi^{\star}_{\maxent}$ maximizing \cref{eq:maxent-rl-objective} induces a distribution over complete trajectories $\tau = (s_{0}, s_{1}, \ldots, s_{T}, \terminal)$ such that
    \begin{equation}
    \pi^{\star}_{\maxent}(\tau) = \prod_{t=0}^{T}\pi^{\star}_{\maxent}(s_{t+1}\mid s_{t})\propto \exp\Bigg(\sum_{t=0}^{T}r(s_{t}, s_{t+1})\Bigg),
    \label{eq:maxent-rl-distribution-propto-return}
\end{equation}
with the convention $s_{T+1} = \terminal$.
\end{proposition}

\begin{proof}
    Using the soft Bellman optimality equation \cref{eq:soft-bellman-optimality-equations-intro} in $\pi_{\maxent}^{\star}$, we have
    \begin{equation}
        \pi^{\star}_{\maxent}(s'\mid s) = \exp\big(Q^{\star}_{\soft}(s, s') - V^{\star}_{\soft}(s)\big) = \exp\big(r(s, s') + V^{\star}_{\soft}(s') - V^{\star}_{\soft}(s)\big).
    \end{equation}
    By telescoping the state value functions, we can conclude that
    \begin{align}
        \pi^{\star}_{\maxent}(\tau) = \prod_{t=0}^{T}\pi^{\star}_{\maxent}(s_{t+1}\mid s_{t}) &= \prod_{t=0}^{T}\exp\big(r(s_{t}, s_{t+1}) + V^{\star}_{\soft}(s_{t+1}) - V^{\star}_{\soft}(s_{t})\big)\\
        &= \exp\Bigg(\sum_{t=0}^{T}\big(r(s_{t}, s_{t+1}) + V^{\star}_{\soft}(s_{t+1}) - V^{\star}_{\soft}(s_{t})\big)\Bigg)\\
        &= \exp\Bigg(\sum_{t=0}^{T}r(s_{t}, s_{t+1}) + V^{\star}_{\soft}(\terminal) - V^{\star}_{\soft}(s_{0})\Bigg)\\
        &= \frac{1}{\exp\big(V^{\star}_{\soft}(s_{0})\big)}\exp\Bigg(\sum_{t=0}^{T}r(s_{t}, s_{t+1})\Bigg)
    \end{align}
    since $V^{\star}_{\soft}(\terminal) = 0$ (the terminal state has no child). $V^{\star}_{\soft}(s_{0})$ is a constant independent of $\tau$.
\end{proof}

With our choice of reward function in \cref{eq:reward-soft-mdp}, where the sum of rewards only depends on the energy of the terminating state, the distribution in \cref{prop:maxent-rl-distribution-propto-return} corresponds \emph{almost} exactly to $P^{\star}$ in \cref{eq:gibbs-distribution}. However, the key difference between \cref{eq:maxent-rl-distribution-propto-return} and our goal of sampling from the Gibbs distribution is that $\pi^{\star}_{\maxent}(\tau)$ is a distribution over complete trajectories, while $P^{\star}$ is over $\gX$ only. In other words, we are interested in the terminating state distribution $\pi^{\star\top}_{\maxent}(x)$, a distribution over the \emph{outcomes} of the sequential process, and not a distribution over \emph{how} we got there.

Fortunately when there is a unique complete trajectory $s_{0} \rightsquigarrow x$ for any $x\in\gX$, then the terminating state distribution matches exactly $\pi^{\star}_{\maxent}(\tau)$ (\ie there is only a single term in the sum appearing in \cref{def:terminating-state-probability}). For instance, this is the case in our factor graph inference example of \cref{sec:sampling-factor-graphs} (see \cref{fig:treesample-intro}), and this is true more generally whenever $\gG$ has a tree structure. Therefore in those cases, if we know the policy $\pi^{\star}_{\maxent}$ maximizing \cref{eq:maxent-rl-objective}, then we can sample from $P^{\star}$ by simply applying \cref{alg:sampling-terminating-state-probability} (\ie sampling a complete trajectory using this policy, and returning the state right before reaching $\terminal$). Furthermore, unlike MCMC which produces autocorrelated samples, we can easily obtain independent samples from $P^{\star}$ by running this procedure multiple times.

\begin{algorithm}[t]
    \caption{Soft Q-Learning algorithm}
    \label{alg:soft-q-learning}
    \begin{algorithmic}[1]
        \Require A MDP $\gM = (\gG, r)$, a step size $\beta > 0$, a behavior policy $\pi_{b}(\cdot \mid s)$
        \Ensure A stochastic policy $\pi^{\star}_{\maxent}$ maximizing \cref{eq:maxent-rl-objective}
        \State Initialize $Q(s, s')$ for all $s \rightarrow s' \in \gG$, such that $Q(x, \terminal) = 0$ for all $x \in \gX$.
        \Loop
            \State Start a trajectory at the initial state $s_{0}$
            \Repeat
                \State Sample the next state: $s_{t+1} \sim \mathrm{Categorical}\big(\pi_{b}(\cdot\mid s_{t})\big)$%
                \State $Q(s_{t}, s_{t+1}) \leftarrow Q(s_{t}, s_{t+1}) + \beta \big[r(s_{t}, s_{t+1}) + \mathrm{LogSumExp}_{s'}Q(s_{t+1}, s') - Q(s_{t}, s_{t+1})\big]$ %
            \Until{$s_{t+1} = \terminal$}
        \EndLoop
        \State \Return $\pi^{\star}_{\maxent}(s'\mid s) \propto \exp\big(Q(s, s')\big)$
    \end{algorithmic}
\end{algorithm}

\paragraph{Finding the optimal policy} To find $\pi^{\star}_{\maxent}$, we can then use any algorithm from the MaxEnt RL literature \citep{nachum2017pcl,haarnoja2018sac}. Probably the simplest is the \emph{\glsdesc{sql}} algorithm\index{Soft Q-Learning} \citep{haarnoja2017sql}, which is the direct counterpart of the well-known Q-Learning algorithm in standard reinforcement learning \citep{watkins1989qlearning}, whose objective is to find a state-action value function $Q(s, s')$ that satisfies the soft Bellman optimality equations; see \cref{alg:soft-q-learning} for reference. Soft Q-Learning is an \emph{off-policy} algorithm, meaning that we can use any arbitrary behavior policy \gls{pib} in order to transition to the next state $s_{t+1}$ on line 5 (as opposed to being \emph{on-policy}, where we would be forced to use the policy derived from the current $Q(s, s')$). In their application to inference in factor graphs described in \cref{sec:sampling-factor-graphs}, \citet{buesing2020treesample} leveraged the tree structure of $\gG$ and used an algorithm based on \emph{Monte Carlo tree search} (MCTS; \citealp{coulom2006mcts}) in order to find $\pi^{\star}_{\maxent}$.  %

\subsection{Multi-path environments \& biased sampling}
\label{sec:multi-path-environment-biased-sampling}

\begin{figure}[t]
    \begin{adjustbox}{center}
        \includegraphics[width=480pt]{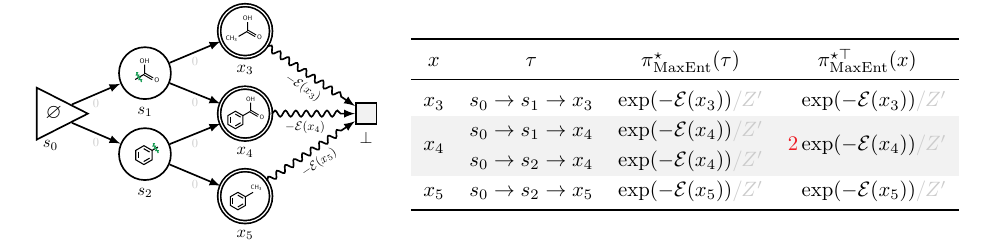}
    \end{adjustbox}
    \caption[Biased sampling with MaxEnt RL]{Illustration of the bias of the terminating state distribution associated with $\pi^{\star}_{\maxent}$ on a soft MDP with a DAG structure. The labels on each transition of the MDP corresponds to the reward function, satisfying \cref{eq:reward-soft-mdp} (\ie sparse reward setting). The terminating state distribution $\pi^{\star\top}_{\maxent}(x)$ is computed by marginalizing $\pi^{\star}_{\maxent}(\tau)$ over trajectories leading to $x$ (\eg two trajectories $s_{0} \rightarrow s_{1} \rightarrow x_{4}$ and $s_{0} \rightarrow s_{2} \rightarrow x_{4}$ to $x_{4}$). $\pi^{\star}_{\maxent}(\tau)$ is computed based on \cref{prop:maxent-rl-distribution-propto-return}. The terminating state distribution $\pi^{\star\top}_{\maxent}(x)$ should be contrasted with the Gibbs distribution $P^{\star}(x) \propto \exp(-\gE(x))$ in \cref{eq:gibbs-distribution}. The normalization constant is $Z' = \exp(-\gE(x_{3})) + 2\exp(-\gE(x_{4})) + \exp(-\gE(x_{5}))$. This is a simplified version of the molecule generation task described in \cref{sec:generation-small-organic-molecules}.}
    \label{fig:toy-molecule-maxentrl}
\end{figure}

We saw that when $\gG$ has a tree structure, meaning that there is a unique complete trajectory going to any terminating state $x\in\gX$, simply following $\pi_{\maxent}^{\star}$ yields samples from the Gibbs distribution if the reward is defined as \cref{eq:reward-soft-mdp}. But when there are multiple ways of constructing a terminating state, as in \cref{sec:galton-board,sec:generation-small-organic-molecules}, this does not hold any longer. It is important to recall that while the distribution over \emph{how} the objects are generated (\ie the complete trajectories) is proportional to $\exp(-\gE(x))$ by \cref{prop:maxent-rl-distribution-propto-return}, we are interested in the (marginal) distribution over the \emph{outcomes} (\ie the corresponding terminating state distribution). In fact, it is easy to show (based on \cref{def:terminating-state-probability} and \cref{prop:maxent-rl-distribution-propto-return}) that in general the terminating state distribution associated with $\pi_{\maxent}^{\star}$ is given by
\begin{equation}
    \pi_{\maxent}^{\star\top}(x) = \sum_{\tau:x\rightarrow \terminal \in \tau} \pi^{\star}_{\maxent}(\tau) \propto n(x)\exp\left(\sum_{t=0}^{T}r(s_{t}, s_{t+1})\right),
    \label{eq:terminating-state-distribution-optimal-maxentrl-policy}
\end{equation}
where $\gls{numtrajectories} = |\{\tau \in \gT\mid x\rightarrow \terminal \in \tau\}|$ is the number of complete trajectories that terminate at $x$, and where $s_{T} = x$ and $s_{T+1} = \terminal$ \citep{bengio2021gflownet}. With the reward function $r(s, s')$ defined as \cref{eq:reward-soft-mdp}, this means that $\pi^{\star\top}_{\maxent}(x) \propto n(x)\exp(-\gE(x))$, which does \emph{not} correspond to our target distribution \cref{eq:gibbs-distribution}. As an illustrative example, we consider a simplified version of the small molecule generation example of \cref{sec:generation-small-organic-molecules} in \cref{fig:toy-molecule-maxentrl}, where the sample space only contains 3 molecules. If we follow the optimal policy $\pi_{\maxent}^{\star}$, the molecule $x_{4}$ will be artificially over-represented since it has 2 complete trajectories leading to it.

Our objective in the first part of this thesis is to build a framework that does not suffer from this bias anymore. Our goal will still be to find a policy whose terminating state distribution matches the Gibbs distribution of interest, but the approach will be \emph{a priori} quite different from the control perspective we adopted in this section. This policy (or forward transition probability distribution $P_{F}$) will be derived from a \emph{flow} in a flow network, as we will see in the following chapters, but we will eventually tie it back to MaxEnt RL in \cref{chap:gflownet-maxent-rl}.

%% file: chapters/03_Flow_Networks.tex
\chapter{Flow Networks}
\label{chap:flow-networks}

\begin{minipage}{\textwidth}
    \itshape
    This chapter contains material from the following paper:
    \begin{itemize}[noitemsep, topsep=1ex, itemsep=1ex, leftmargin=3em]
        \item Yoshua Bengio$^{*}$, Salem Lahlou$^{*}$, \textbf{Tristan Deleu}$^{*}$, Edward Hu, Mo Tiwari, Emmanuel Bengio (2023). \emph{GFlowNet Foundations}. Journal of Machine Learning Research (JMLR). \notecite{bengio2023gflownetfoundations}
    \end{itemize}
    \vspace*{5em}
\end{minipage}

In the previous chapter, we saw that when the objects have some compositional structure, we can naturally sample from the Gibbs distribution in \cref{eq:gibbs-distribution} as a sequential decision making process through a pointed DAG $\gG$ whose terminating states $\gX$ is the sample space of $P^{\star}$. The main challenge was to find a forward transition probability $P_{F}$ such that its corresponding terminating state distribution matched the target Gibbs distribution. We saw in \cref{sec:probabilistic-inference-control-problem} that naively applying (maximum entropy) reinforcement learning in order to find this $P_{F}$ could lead to a distribution which is biased when $\gG$ is a general DAG, with multiple ways of constructing the objects of interest. To avoid this bias, we will see in the next chapter how to find such a $P_{F}$ in a general case for any structure $\gG$. This solution will be based on a standard tool of graph theory called \emph{flow networks}, of which we introduce the foundations in this chapter.

\section{Flow networks \& Markovian flows}
\label{sec:flow-networks-markovian-flows}
What made the maximum entropy reinforcement learning approach fail in \cref{sec:multi-path-environment-biased-sampling} was the possibility of having multiple paths leading to the same object in the pointed DAG $\gG$, where the probabilities of each path were added together. This was effectively treating different paths going to the same object as being completely ``independent'', even though they could have some components in common (and they even share the same endpoint). To alleviate this issue, we need a structure that will take into account these dependencies between paths. This will come in the form of a \emph{flow network}, which we will introduce in this section, along with its Markovian counterpart which will play a central role in practice when we will use them for generative modeling in \cref{chap:generative-flow-networks}.

\subsection{Flow networks}
\label{sec:flow-networks}
In \cref{sec:elements-graph-theory}, we introduced the notion of pointed DAG $\gG$ to encode the structure of a state space $\gS$. We augment this DAG with a function $F^{\star}$ over complete trajectories of $\gG$ called a \emph{flow}.
\begin{definition}[Trajectory flow]
    \label{def:trajectory-flow}\index{Flow!Trajectory flow}\index{Flow network}
    Let $\gG$ be a pointed DAG, and $\gT$ be the set of complete trajectories over $\gG$. A non-negative function over complete trajectories $\gls{flow}: \gT \rightarrow \sR^{+}$ is called a \emph{trajectory flow}. The set of all trajectory flows is denoted by $\gls{flownetwork} \triangleq \gF(\gT, \sR^{+})$ (\ie the set of all non-negative functions over $\gT$). The pair $(\gG, F^{\star})$ is called a \emph{flow network}.
\end{definition}
Although a trajectory flow $F^{\star}$ is defined as a function over complete trajectories, it can also naturally induce a measure over the measurable space of complete trajectories $(\gT, \Sigma)$, where the $\sigma$-algebra $\Sigma = 2^{\gT}$ is the power set of $\gT$. We make an abuse of notation here and denote by $F^{\star}$ both the trajectory flow (a function), and its corresponding measure, where for every subset of complete trajectories $A \subseteq \gT$, we have
\begin{equation}
    F^{\star}(A) = \sum_{\tau \in A}F^{\star}(\tau).
    \label{eq:induced-trajectory-flow-measure}
\end{equation}
In the special case where the subset $A$ is a singleton trajectory $\{\tau\}$, we will simply write its measure as $F^{\star}(\tau)$ to identify it with the trajectory flow function. By another abuse of notation, we will use $F^{\star}$ to also denote flows going through specific states, transitions, or partial trajectories (see \cref{sec:markovian-flows}). For example, the \emph{state flow} $F^{\star}(s)$ corresponds to the amount of flow going through the state $s\in\gS$, and is defined by
\begin{equation}
    F^{\star}(s) \triangleq F^{\star}(\{\tau \in \gT\mid s \in \tau\}) = \sum_{\tau: s\in\tau}F^{\star}(\tau).
    \label{eq:state-flow}\index{Flow!State flow}
\end{equation}
We will always assume that the trajectory flow $F^{\star}$ is non-identically zero, and that for any state $s\in\gS$, there exists at least a trajectory going through $s$ with positive flow, to guarantee that $F^{\star}(s) > 0$. Similarly, the \emph{edge flow} $F^{\star}(s\rightarrow s')$ corresponds to the amount of flow going through the transition $s\rightarrow s'\in \gG$:
\begin{equation}
    F^{\star}(s\rightarrow s') \triangleq F^{\star}(\{\tau\in \gT\mid s\rightarrow s' \in \tau\}) = \sum_{\tau: s\rightarrow s' \in \tau} F^{\star}(\tau).
    \label{eq:edge-flow}\index{Flow!Edge flow}
\end{equation}
These state and edge flows will play a crucial role throughout this section, and in particular the edge flows through terminating transitions of the form $F^{\star}(x\rightarrow \terminal)$ for $x\in\gX$, as we will see in \cref{chap:generative-flow-networks}. For any flow network, these flows are related via a ``flow conservation'' rule, which we will expand on in \cref{sec:flow-matching-conditions}.

\begin{proposition}
    \label{prop:identities-state-edge-flows}
    Given a flow network $(\gG, F^{\star})$, the state and edge flows satisfy
    \begin{align}
        \forall s \in \gS,\ \qquad F^{\star}(s) &= \sum_{s'\in\children_{\gG}(s)}F^{\star}(s\rightarrow s')\\
        \forall s' \neq s_{0},\qquad F^{\star}(s') &= \sum_{s\in\parents_{\gG}(s')}F^{\star}(s\rightarrow s')\label{eq:identity-state-edge-flows-parents}
    \end{align}
\end{proposition}
\begin{proof}
    For any $s\in\gS$, the set of complete trajectories going through $s$ is the (disjoint) union of the sets of trajectories going through $s \rightarrow s'$, for all $s'\in\children_{\gG}(s)$
    \begin{equation}
        \{\tau \in \gT\mid s\in\tau\} = \bigcup_{s'\in\children_{\gG}(s)}\{\tau\in\gT\mid s\rightarrow s' \in \tau\}.
    \end{equation}
    In other words, any complete trajectory $\tau$ going through $s$ has to also go through the transition $s \rightarrow s'$, for one of the children $s'\in\children_{\gG}(s)$. Therefore, it follows by definition of the state and edge flows that
    \begin{equation}
        F^{\star}(s) = \sum_{\tau: s\in\tau}F^{\star}(\tau) = \sum_{s'\in\children_{\gG}(s)}\sum_{\tau: s\rightarrow s'\in\tau}F^{\star}(\tau) = \sum_{s'\in\children_{\gG}(s)}F^{\star}(s\rightarrow s').
    \end{equation}
    Similarly, for $s'\neq s_{0}$, the set of complete trajectories going through $s'$ is the (disjoint) union of the sets of trajectories going through $s \rightarrow s'$ for all $s \in \parents_{\gG}(s')$
    \begin{equation}
        \{\tau\in\gT\mid s'\in\tau\} = \bigcup_{s\in\parents_{\gG}(s')}\{\tau\in\gT\mid s \rightarrow s'\in\tau\}.
    \end{equation}
    Again, any complete trajectory $\tau$ going through $s'$ has necessarily gone through the transition $s \rightarrow s'$, for one of the parents $s \in \parents_{\gG}(s')$ if $s' \neq s_{0}$ is not the initial state. Therefore
    \begin{equation}
        F^{\star}(s') = \sum_{\tau: s'\in\tau}F^{\star}(\tau) = \sum_{s \in \parents_{\gG}(s')}\sum_{\tau: s\rightarrow s'\in\tau}F^{\star}(\tau) = \sum_{s\in\parents_{\gG}(s')}F^{\star}(s\rightarrow s').
    \end{equation}
\end{proof}
The condition ``$s'\neq s_{0}$'' in \cref{eq:identity-state-edge-flows-parents} highlights the particular role played by the initial state: it acts as a single source for all the flow going through the flow network. Similarly, the terminal state $\terminal$ is a single sink state where all the flow ends at. Using the measure perspective of $F^{\star}$, the total amount of flow going through the flow network then corresponds to the flow over the whole space of complete trajectories.

\begin{definition}[Total flow]
    \label{def:total-flow}\index{Flow!Total flow}
    Let $(\gG, F^{\star})$ be a flow network. The \emph{total flow} \gls{totalflow} of the flow network is the sum of the flows of all the complete trajectories (equivalently, the measure of the whole set $\gT$)
    \begin{equation}
        Z^{\star} \triangleq F^{\star}(\gT) = \sum_{\tau\in\gT}F^{\star}(\tau).
        \label{eq:total-flow}
    \end{equation}
\end{definition}
Since $F^{\star}$ is non-identically zero, the total flow $Z^{\star} > 0$ itself is clearly guaranteed to be positive. The following proposition makes the intuition above clear that all the flow through the flow network originates from the initial state $s_{0}$, linking the total flow $Z^{\star}$ with the flow going through (going out of) the initial state $s_{0}$.
\begin{proposition}[Initial flow]
    \label{prop:initial-flow-total-flow}
    Let $(\gG, F^{\star})$ be a flow network, and let $Z^{*}$ be its total flow. The state flow at the initial state is equal to the total flow: $F^{\star}(s_{0}) = Z^{*}$.
\end{proposition}
\begin{proof}
    All the complete trajectories go through the initial state. Therefore
    \begin{equation}
        F^{\star}(s_{0}) = \sum_{\tau: s_{0}\in\tau}F^{\star}(\tau) = \sum_{\tau\in\gT}F^{\star}(\tau) = Z^{\star}.
    \end{equation}
\end{proof}
We will consistently use the ``${}^{\star}$'' notation to denote any quantity related to a valid trajectory flow (\eg the total flow, or the corresponding transition probabilities in \cref{def:transition-probabilities-from-flow}). In further sections, and especially in \cref{chap:generative-flow-networks}, the functions without the star notation will denote any function that may not necessarily correspond to a proper trajectory flow, and therefore may not satisfy the properties above such as the identities in \cref{prop:identities-state-edge-flows} and the link between the total flow and the state flow at the initial state in \cref{prop:initial-flow-total-flow}.

\begin{figure}[t]
    \centering
    \begin{adjustbox}{center}
        \includegraphics[width=520pt]{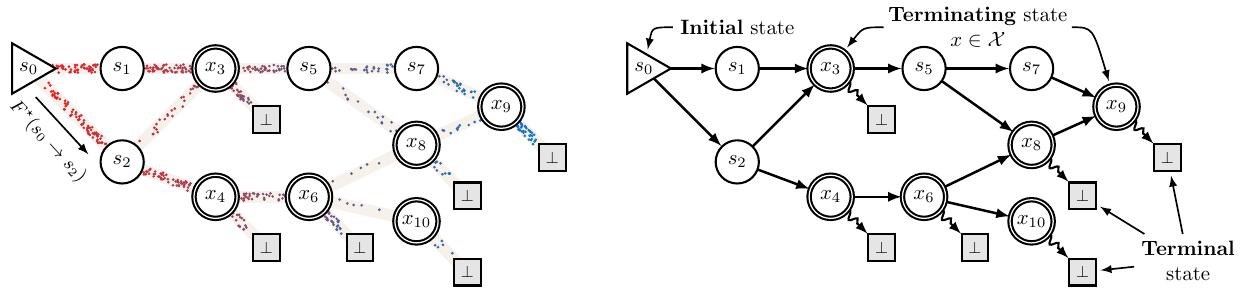}%
    \end{adjustbox}
    \begin{adjustbox}{center}%
        \begin{subfigure}[b]{260pt}%
            \caption{}%
            \label{fig:gflownet-blogpost-flow}%
        \end{subfigure}%
        \begin{subfigure}[b]{260pt}%
            \caption{}%
            \label{fig:gflownet-blogpost-pointed-dag}%
        \end{subfigure}%
    \end{adjustbox}
    \caption[Illustration of a flow network]{Illustration of a flow network. (a) A flow of particles traverses the network of pipes, from $s_{0}$ to the terminal state $\terminal$ (illustrated by the gradual change in color). The edge flow $F^{\star}(s\rightarrow s')$ quantifies the ``amount of particles'' going through $s \rightarrow s'$. (b) Representation of the underlying pointed DAG $\gG$, with the nomenclature introduced in \cref{sec:elements-graph-theory}.}
    \label{fig:gflownet-blogpost}
\end{figure}

A more concrete interpretation of a flow network is as a ``fluid'' going through conduits, as illustrated in \cref{fig:gflownet-blogpost-flow}. A flow can be viewed as a stream of particles flowing through these conduits from the initial state $s_{0}$ (the source of all the fluid, \eg a faucet) to the terminal state $\terminal$ (a sink). The edge flow $F^{\star}(s \rightarrow s')$ is the total amount of flow going through a certain conduit. In general though, since flows are defined at a global level over complete trajectories, these particles have the rather unnatural property (from a physical perspective) that they can remember their history (\ie which sequence of conduits they took). We will see in \cref{sec:markovian-flows} a family of trajectory flows where particles will no longer have memory of their past, making this physical intuition more realistic.

\subsection{Backward transition probabilities}
\label{sec:backward-transition-probabilities}
In \cref{sec:forward-transition-probabilities}, we defined the notion of forward transition probability distribution $P_{F}$ as being a distribution over the children of a particular state $s$. In what follows, it will be necessary to also consider distributions over the \emph{parents} of a state. We call these \emph{backward transition probabilities}.

\begin{definition}[Consistent backward transition probabilities]
    \label{def:consistent-backward-transition-probability}\index{Transition probability!Backward transition probabilities|textbf}
    Let $\gG = (\widebar{\gS}, \gA)$ be a pointed DAG. A function $\gls{backwardtransitionprobability}: \gS \rightarrow \Delta(\parents_{\gG})$ is called a \emph{backward transition probability} distribution consistent with $\gG$ if for any $s \neq s_{0}$, $P_{B}(\cdot \mid s)$ is a properly defined distribution over $\parents_{\gG}(s)$, \ie for all $s'\in\parents_{\gG}(s)$, $P_{B}(s'\mid s) \geq 0$ and
    \begin{equation}
        \sum_{s'\in\parents_{\gG}(s)}P_{B}(s'\mid s) = 1.
        \label{eq:consistent-backward-transition-probability}
    \end{equation}
\end{definition}
We use $\gS$ as the domain of $P_{B}$ in our definition for simplicity, even though the backward transition probability is undefined for the initial state $s_{0}$, since it has no parent (by definition of a pointed DAG in \cref{def:pointed-dag}). Moreover, while it would be technically possible to define $P_{B}$ at the terminal state $\terminal$, as a distribution over $\gX = \parents_{\gG}(\terminal)$, this is intentionally left out from the definition here since $\gX$ may be much larger than any other $\parents_{\gG}(s)$ and defining such a ``$P_{B}(\cdot \mid \terminal)$'' would eventually be as hard as computing the Gibbs distribution in \cref{eq:gibbs-distribution}, which we want to avoid. Similar to \cref{def:consistent-forward-transition-probability}, for any state $s\neq s_{0}$, we use the notation ``\gls{simplexparents}'' to denote the space of probability distributions over the parents of $s$ in $\gG$.

Again similar to how we extended the notion of forward transition probabilities to (partial) trajectories, we can naturally do the same for $P_{B}$ as well. For some partial trajectory $\tau = (s_{m}, s_{m+1}, \ldots, s_{n})$,
\begin{equation}
    P_{B}(\tau) = \prod_{t=m}^{n-1}P_{B}(s_{t}\mid s_{t+1}).
    \label{eq:PB-partial-trajectory}
\end{equation}
We saw in \cref{lem:PF-distribution-suffix} that a forward transition probability distribution consistent with $\gG$ induced a distribution over the partial trajectories starting at any arbitrary state. The following lemma shows similarly that a backward transition probability distribution would also induce a distribution over partial trajectories, but this time ending at any state $s \in \gS$.
\begin{lemma}[Distribution over prefixes]
    \label{lem:PB-distribution-prefix}
    Let $\gG = (\widebar{\gS}, \gA)$ be a pointed DAG, and let $P_{B}: \gS \rightarrow \Delta(\parents_{\gG})$ be a backward transition probability consistent with $\gG$. For any state $s\in\gS$ such that $s\neq s_{0}$, $P_{B}$ induces a distribution over partial trajectories $\tau$ starting at the initial state $s_{0}$ and ending at $s$, \ie $P_{B}(\tau) \geq 0$ and
    \begin{equation}
        \sum_{\tau: s_{0}\rightsquigarrow s}P_{B}(\tau) = 1.
        \label{eq:PB-distribution-prefix}
    \end{equation}
\end{lemma}

\begin{proof}
    The proof is similar to the one of \cref{lem:PF-distribution-suffix}, and we prove \cref{eq:PB-distribution-prefix} by strong induction on the maximum length of the partial trajectories. For any state $s\in\gS$, let $d_{s}$ denote the maximum length of a partial trajectory from the initial state $s_{0}$ to $s$.

    \emph{Base case:} There exists at least a state $s\in\children_{\gG}(s_{0})$ such that $d_{s} = 1$. In that case, $s_{0}$ is the only parent of $s$, and there exists a single partial trajectory of the form $s_{0} \rightsquigarrow s$, which is the transition $s_{0} \rightarrow s$. In that case
    \begin{equation}
        \sum_{\tau: s_{0}\rightsquigarrow s}P_{B}(\tau) = P_{B}(s_{0}\mid s) = 1.
    \end{equation}

    \emph{Induction step:} Suppose that \cref{eq:PB-distribution-prefix} is valid for all states $s'$ such that $d_{s'} \leq d$ for some $d > 0$. Since $\gG$ is a pointed DAG, there exists a state $s\in\gS$ such that $d_{s} = d+1$. We can decompose the set of all partial trajectories to $s$ into a disjoint union of partial trajectories going through each parent $s' \in \parents_{\gG}(s)$ as
    \begin{equation}
        \{\tau: s_{0}\rightsquigarrow s\} = \bigcup_{s'\in\parents_{\gG}(s)}\{\tau' \oplus (s'\rightarrow s)\mid \tau': s_{0}\rightsquigarrow s'\}.
    \end{equation}
    Since all the partial trajectories of the form $s_{0}\rightsquigarrow s'$ have a maximum depth $d_{s'} \leq d$, we can apply the induction step to them. Using the decomposition above, we get
    \begin{equation}
        \sum_{\tau: s_{0}\rightsquigarrow s}P_{B}(\tau) = \sum_{s'\in\parents_{\gG}(s)}P_{B}(s'\mid s)\sum_{\tau': s_{0}\rightsquigarrow s'}P_{B}(\tau') = \sum_{s'\in\parents_{\gG}(s)}P_{B}(s'\mid s) = 1.
    \end{equation}
    This proves that \cref{eq:PB-distribution-prefix} is also valid for a state $s$ with maximum trajectory length $d+1$.
\end{proof}
A consequence of this lemma is that a backward transition probability distribution induces a distribution over the set of partial trajectories $s_{0}\rightsquigarrow x$ leading to some terminating state $x\in\gX$. This distribution over partial trajectories is denoted \gls{backwardinduceddistribution} \citep{malkin2022trajectorybalance}, and it will be useful in order to estimate the terminating state probability $P_{F}^{\top}(x)$ of a specific state, using importance sampling \citep{zhang2022ebgfn}; see \cref{sec:estimation-terminating-state-probability} for details.

\subsection{Probability distribution induced by a flow}
\label{sec:probability-distribution-induced-flow}
To bridge the gap between flow networks, which may seem a bit \emph{ad hoc} thus far, and transition probabilities, we can derive forward and backward transition probability distributions from a valid flow $F^{\star}$ by simply normalizing it with appropriate quantities. 
\begin{definition}[Transition probabilities]
    \label{def:transition-probabilities-from-flow}\index{Transition probability!Flow}
    Let $(\gG, F^{\star})$ be a flow network. For any transition $s \rightarrow s' \in \gG$ such that $s'\neq \terminal$ and any transition $s' \rightarrow s''\in\gG$, the \emph{forward transition probability} distribution $P^{\star}_{F}$ and the \emph{backward transition probability} distribution $P^{\star}_{B}$ (associated with $F^{\star}$) are defined by
    \begin{align}
        P^{\star}_{F}(s''\mid s') &\triangleq \frac{F^{\star}(s'\rightarrow s'')}{F^{\star}(s')} & P^{\star}_{B}(s\mid s') &\triangleq \frac{F^{\star}(s\rightarrow s')}{F^{\star}(s')}.
    \end{align}
\end{definition}
It is clear from \cref{prop:identities-state-edge-flows} that $P_{F}^{\star}$ is a properly defined forward transition probability consistent with $\gG$ (\cref{def:consistent-forward-transition-probability}), and $P_{B}^{\star}$ is a backward transition probability consistent with $\gG$ (\cref{def:consistent-backward-transition-probability}). When flow networks are interpreted as being a fluid going through pipes, the forward transition probability $P^{\star}_{F}$ has a natural interpretation as ``following the flow'', in the sense that at each state $s'$, we have a higher chance of going to a state $s''$ as the flow $F^{\star}(s'\rightarrow s'')$ is larger.

Although in general computing the terminating state distribution $P_{F}^{\top}(x)$ associated with an arbitrary $P_{F}$ might be challenging due to the summation over all the trajectories $s_{0} \rightsquigarrow x$ (\cref{def:terminating-state-probability}), the following proposition shows that the terminating state distribution associated with $P_{F}^{\star}$ derived from a flow $F^{\star}$ has a notably simple form that only depends on the flow at the terminating edge.
\begin{proposition}
    \label{prop:terminating-state-probability-from-flow}\index{Terminating state probability}
    Let $(\gG, F^{\star})$ be a flow network, with $P_{F}^{\star}$ the forward transition probability associated with the trajectory flow $F^{\star}$, and let $Z^{\star}$ be the total flow of $F^{\star}$. The corresponding terminating state probability distribution satisfies for all $x\in\gX$
    \begin{equation}
        P_{F}^{\star\top}(x) = \frac{F^{\star}(x \rightarrow \terminal)}{Z^{\star}}.
        \label{eq:terminating-state-probability-from-flow}
    \end{equation}
\end{proposition}

\begin{proof}
    By definition of the terminating state probability of $x\in\gX$ (\cref{def:terminating-state-probability})
    {\allowdisplaybreaks%
    \begin{align}
        P_{F}^{\star\top}(x) &= \sum_{\tau:s_{0}\rightsquigarrow x}\prod_{t=0}^{T_{\tau}}P_{F}^{\star}(s_{t+1}\mid s_{t}) = \sum_{\tau: s_{0}\rightsquigarrow x}\prod_{t=0}^{T_{\tau}}\frac{F^{\star}(s_{t}\rightarrow s_{t+1})}{F^{\star}(s_{t})}\label{eq:proof-terminating-state-probability-from-flow-eq1}\\
        &= \frac{F^{\star}(x\rightarrow \terminal)}{F^{\star}(s_{0})}\sum_{\tau:s_{0}\rightsquigarrow x}\prod_{t=1}^{T_{\tau}}\frac{F^{\star}(s_{t-1}\rightarrow s_{t})}{F^{\star}(s_{t})}\\
        &= \frac{F^{\star}(x\rightarrow \terminal)}{Z^{\star}}\sum_{\tau: s_{0}\rightsquigarrow x}\prod_{t=1}^{T_{\tau}}P_{B}^{\star}(s_{t-1}\mid s_{t})\label{eq:proof-terminating-state-probability-from-flow-eq2}\\
        &= \frac{F^{\star}(x\rightarrow \terminal)}{Z^{\star}}\sum_{\tau: s_{0}\rightsquigarrow x}P_{B}^{\star}(\tau) = \frac{F^{\star}(x\rightarrow \terminal)}{Z^{\star}},\label{eq:proof-terminating-state-probability-from-flow-eq3}
    \end{align}}%
    where we used the conventions $s_{T_{\tau}} = x$ and $s_{T_{\tau}+1} = \terminal$, the definition of $P_{F}^{\star}$ in terms of the edge and state flows of $F^{\star}$ in \cref{eq:proof-terminating-state-probability-from-flow-eq1}, the definition of $P_{B}^{\star}$ and \cref{prop:initial-flow-total-flow} in \cref{eq:proof-terminating-state-probability-from-flow-eq2}, and \cref{lem:PB-distribution-prefix} in \cref{eq:proof-terminating-state-probability-from-flow-eq3} applied to $x$ since $P_{B}^{\star}$ is a backward transition probability consistent with $\gG$.
\end{proof}
The proposition above is very important. It shows that \emph{if we know the flow $F^{\star}$}, then the terminating state distribution associated with $P_{F}^{\star}$ (which is derived from the flow) is a distribution known up to a normalization constant, which is reminiscent of a Gibbs distribution in \cref{eq:gibbs-distribution}. Therefore we can reduce the problem of sampling from the Gibbs distribution to finding a valid flow such that $F^{\star}(x\rightarrow \terminal) = \exp(-\gE(x))$; all we would have to do to sample from the Gibbs distribution would be to follow $P_{F}^{\star}$ and return the terminating state. How to find such a flow that satisfies these ``boundary conditions'' will be the subject of \cref{chap:generative-flow-networks}. We will often say that $P_{F}^{\star\top}$ is the distribution (over terminating states) induced by the flow $F^{\star}$.

Another consequence of this proposition is that since $P_{F}^{\star\top}$ is a properly defined distribution over terminating states, in addition to being equal to the flow at the initial state (\cref{prop:initial-flow-total-flow}), the total flow $Z^{\star} = \sum_{x\in\gX}F^{\star}(x\rightarrow \terminal)$ is equal to the sum of all the flow going through terminating edges. Just like $s_{0}$ is the unique source of all the flow in the flow network, $\terminal$ is the unique sink for all the flow.

\subsection{Markovian flows}
\label{sec:markovian-flows}
\begin{figure}[t]
    \begin{adjustbox}{center}
        \includegraphics[width=480pt]{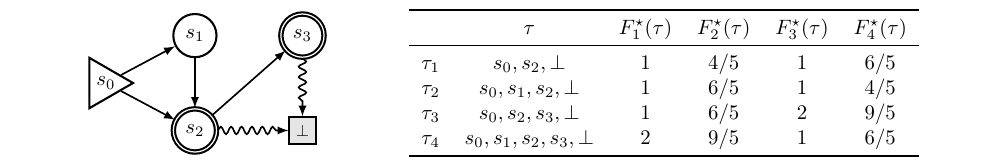}
    \end{adjustbox}
    \caption[Examples of Markovian, non-Markovian, and equivalent flows.]{Examples of Markovian, non-Markovian, and equivalent flows. The flows $F^{\star}_{2}$ \& $F^{\star}_{4}$ are Markovian, whereas $F^{\star}_{1}$ \& $F^{\star}_{3}$ are not. The first two flows are equivalent $F^{\star}_{1} \sim F^{\star}_{2}$. Similarly, $F^{\star}_{3} \sim F^{\star}_{4}$, but they are not equivalent to $F^{\star}_{1}$ and $F^{\star}_{2}$. All these trajectory flows have the same edge flow at the terminating edges ($F^{\star}(s_{2} \rightarrow \terminal) = 2$ \& $F^{\star}(s_{3} \rightarrow \terminal) = 3$).}
    \label{fig:markovian-non-markovian-flows}
\end{figure}

Defining a trajectory flow requires the specification of $|\gT|$ non-negative values, one for every complete trajectory, which can be exponentially large in the number of edges in $\gG$ in general. In this section, we introduce a special class of trajectory flows satisfying some \emph{Markov property}, which have the remarkable characteristic that they can be defined with much fewer values thanks to a factorization along $\gG$. Similar to the state and edge flows in \cref{sec:flow-networks}, we first need to define the flow going through a prefix (partial) trajectory of the form $\tau = (s_{0}, s_{1}, \ldots, s)$, ending at some state $s\in\gS$ (not necessarily the terminal state), by
\begin{equation}
    F^{\star}(\tau \preceq \cdot) \triangleq F^{\star}(\{\tau' \in \gT\mid \tau \preceq \tau'\}) = \sum_{\tau': \tau\preceq \tau'}F^{\star}(\tau'),
    \label{eq:prefix-trajectory-flow}
\end{equation}
where $\tau \preceq \tau'$ means that $\tau$ is a prefix of $\tau'$. This partial trajectory $\tau$ can be interpreted as the full history before reaching $s$. With this perspective, we will use $F^{\star}(\tau\preceq \cdot)$, alongside the state and edge flows, to define the Markovian nature of a trajectory flow $F^{\star}$.
\begin{definition}[Markovian flow]
    \label{def:markovian-flow}\index{Flow!Markovian flow}\index{Markovian flow|see {Flow}}
    Let $(\gG, F^{\star})$ be a flow network, with $\gG = (\widebar{\gS}, \gA)$. $F^{\star}$ is called a \emph{Markovian flow} if for any state $s\in\gS$, any partial trajectory $\tau = (s_{0}, s_{1}, \ldots, s)$ leading to $s$, and any transition $s \rightarrow s' \in \gG$
    \begin{equation}
        F^{\star}(s)F^{\star}(\tau' \preceq \cdot) = F^{\star}(s\rightarrow s')F^{\star}(\tau \preceq \cdot),
        \label{eq:markovian-flow-condition}
    \end{equation}
    where $\tau' = \tau \oplus (s \rightarrow s')$ is the (partial) trajectory where the transition $s\rightarrow s'$ has been appended to $\tau$. The set of Markovian flows is denoted by $\gls{markovianflownetwork} \subsetneq \widebar{\gF}(\gG)$. $(\gG, F^{\star})$ is called a \emph{Markovian flow network}.
\end{definition}
Unlike in \citep{bengio2023gflownetfoundations}, our definition of a Markovian flow does not involve probabilities over trajectories explicitly, and only relies on properties of the flow itself. However, the standard notion of a Markov process, where the probability of transitioning only depends on the current state and not the whole history, can be recovered as a consequence of \cref{def:markovian-flow}. For a flow $F^{\star}$, we can define a probability distribution over the space of complete trajectories $(\gT, \Sigma)$ (where $\Sigma = 2^{\gT}$) by simply normalizing the measure $F^{\star}$ by its total flow $Z^{\star}$. In other words, for any set $A \in \Sigma$, we have
\begin{equation}
    P^{\star}(A) \triangleq \frac{F^{\star}(A)}{Z^{\star}} = \frac{1}{Z^{\star}}\sum_{\tau\in A}F^{\star}(\tau).
\end{equation}
If we assume that $F^{\star}$ is Markovian, then by definition of conditional probability, the probability of transitioning to $s'$ conditioned on the whole history prior to $s$, given by the partial trajectory $\tau = (s_{0}, s_{1}, \ldots, s)$, is
\begin{equation}
    P^{\star}(s\rightarrow s'\mid \tau) = \frac{P^{\star}(\tau'\preceq \cdot)}{P^{\star}(\tau\preceq \cdot)} = \frac{F^{\star}(\tau'\preceq \cdot)}{F^{\star}(\tau\preceq \cdot)} = \frac{F^{\star}(s\rightarrow s')}{F^{\star}(s)} = P_{F}^{\star}(s'\mid s),
\end{equation}
where the forward transition probability of $F^{\star}$ only depends on the current state $s$. Note that the Markov property does not hold for all trajectory flows, and there exist non-Markovian flows. In \cref{fig:markovian-non-markovian-flows} we show multiple examples of Markovian and non-Markovian flows. For example, to see that $F_{1}^{\star}$ is not Markovian, we can observe that for the partial trajectory $\tau = (s_{0}, s_{1}, s_{2})$,
\begin{align}
    F_{1}^{\star}(s_{2}) &= \sum_{i=1}^{4}F_{1}^{\star}(\tau_{i}) = 5 &&& F_{1}^{\star}(s_{2}\rightarrow s_{3}) &= F_{1}^{\star}(\tau_{3}) + F_{1}^{\star}(\tau_{4}) = 3\\
    F_{1}^{\star}(\tau_{4} \prec \cdot) &= F_{1}^{\star}(\tau_{4}) = 2 &&& F_{1}^{\star}(\tau \prec \cdot) &= F_{1}^{\star}(\tau_{2}) + F_{1}^{\star}(\tau_{4}) = 3,
\end{align}
and therefore $F_{1}^{\star}(s_{2})F_{1}^{\star}(\tau_{4} \prec \cdot) \neq F_{1}^{\star}(s_{2}\rightarrow s_{3})F_{1}^{\star}(\tau \prec \cdot)$. On the other hand, using a similar reasoning, we have that $F_{2}^{\star}(s_{2})F_{2}^{\star}(\tau_{4}\prec \cdot) = F_{2}^{\star}(s_{2}\rightarrow s_{3})F_{2}^{\star}(\tau \prec \cdot) = 9$ (and similarly for all other states and partial trajectories), making $F_{2}^{\star}$ Markovian.

Although Markovian flows will play a key role in what follows, showing that a flow is Markovian via \cref{def:markovian-flow} can be tedious. The following proposition offers an alternative characterization of a Markovian flow, via a decomposition into its total flow and forward transition probabilities.

\begin{proposition}[Characterization of Markovian flows]
    \label{prop:characterization-markovian-flow}
    Let $(\gG, F^{\star})$ be a flow network, with $\gG = (\widebar{\gS}, \gA)$. $F^{\star}$ is a Markovian flow if and only if there exists a constant $C > 0$ and a forward transition probability $P_{F}: \gS \rightarrow \Delta(\children_{\gG})$ consistent with $\gG$ such that for any complete trajectory $\tau = (s_{0}, s_{1}, \ldots, s_{T}, \terminal) \in \gT$,
    \begin{equation}
        F^{\star}(\tau) = C\prod_{t=0}^{T}P_{F}(s_{t+1}\mid s_{t}),
        \label{eq:characterization-markovian-flow}
    \end{equation}
    where we used the convention $s_{T+1} = \terminal$. Under these conditions, $C$ corresponds to the total flow of $F^{\star}$, and $P_{F}$ is the forward transition probability associated with $F^{\star}$.
\end{proposition}

\begin{proof}
    $\Rightarrow$: Let $\tau_{t} = (s_{0}, s_{1}, \ldots, s_{t})$ denote the (partial) prefix trajectory of $\tau$, with the convention that $\tau_{T+1} = \tau$. If $F^{\star}$ is a Markovian flow, and since $\tau_{t+1} = \tau_{t} \oplus (s_{t} \rightarrow s_{t+1})$, then by definition we have that
    \begin{equation}
        F^{\star}(\tau_{t+1}\preceq \cdot) = \frac{F^{\star}(s_{t}\rightarrow s_{t+1})}{F^{\star}(s_{t})}F^{\star}(\tau_{t}\preceq \cdot) = P_{F}^{\star}(s_{t+1}\mid s_{t})F^{\star}(\tau_{t}\preceq\cdot).
        \label{eq:proof-:characterization-markovian-flow-eq1}
    \end{equation}
    An immediate consequence of \cref{eq:proof-:characterization-markovian-flow-eq1} is that
    \begin{equation}
        F^{\star}(\tau \preceq \cdot) = F^{\star}(\tau_{T+1}\preceq \cdot) = F^{\star}(\tau_{0} \preceq \cdot) \prod_{t=0}^{T}P^{\star}_{F}(s_{t+1}\mid s_{t}).
        \label{eq:proof-:characterization-markovian-flow-eq2}
    \end{equation}
    Since $\tau$ is a complete trajectory, it is clear from \cref{eq:prefix-trajectory-flow} that $F^{\star}(\tau\preceq \cdot) = F^{\star}(\tau)$, $\tau$ itself being the only complete trajectory whose prefix is $\tau$. Furthermore, since $\tau_{0} = (s_{0})$ only consists of the initial state, we also have that $F^{\star}(\tau_{0}\preceq \cdot) = F^{\star}(s_{0}) = Z^{\star}$, where the last equality is a consequence of \cref{prop:initial-flow-total-flow}. Using this in \cref{eq:proof-:characterization-markovian-flow-eq2}, we get the expected decomposition of the Markovian flow $F^{\star}$ in terms of its forward transition probabilities $P^{\star}_{F}$ and its total flow $Z^{\star}$
    \begin{equation}
        F^{\star}(\tau) = Z^{\star}\prod_{t=0}^{T}P^{\star}_{F}(s_{t+1}\mid s_{t}).
    \end{equation}

    $\Leftarrow$: To show the sufficiency of \cref{eq:characterization-markovian-flow}, we will prove that (1) the trajectory flow $F^{\star}$ is Markovian, and (2) that its total flow and associated forward transition probability are $C$ and $P_{F}$ respectively.
    \begin{enumerate}[leftmargin=*]
        \item Let $s\in\gS$ be a state, $\tau = (s_{0}, s_{1}, \ldots, s)$ a partial trajectory leading to $s$, and a transition $s \rightarrow s' \in \gG$. We denote by $\tau' = \tau \oplus (s \rightarrow s')$ the (partial) trajectory corresponding to $s \rightarrow s'$ being appended to $\tau$. Any complete trajectory $\tau''$ having $\tau$ as a prefix can be decomposed into the partial trajectory $\tau$, followed by a (partial) trajectory $\overrightarrow{\tau}$ from $s$ to the terminal state $\terminal$. By definition of $F^{\star}(\tau \preceq \cdot)$,
        {\allowdisplaybreaks%
        \begin{align}
            F^{\star}(\tau \preceq \cdot) &= \sum_{\tau'': \tau \preceq \tau''}F^{\star}(\tau'') = C\sum_{\tau'': \tau\preceq \tau''}\prod_{t=0}^{T_{\tau''}}P_{F}(s_{t+1}\mid s_{t})\\
            &= C\prod_{t=0}^{T_{\tau}-1}P_{F}(s_{t+1}\mid s_{t})\sum_{\overrightarrow{\tau}: s\rightsquigarrow \terminal}\prod_{t=T_{\tau}}^{T_{\overrightarrow{\tau}}-1}P_{F}(s_{t+1}\mid s_{t})\\
            &= CP_{F}(\tau)\sum_{\overrightarrow{\tau}: s\rightsquigarrow \terminal}P_{F}(\overrightarrow{\tau})\\
            &= CP_{F}(\tau)\label{eq:proof-characterization-markovian-flow-eq1}
        \end{align}}%
        where we used \cref{lem:PF-distribution-suffix} in \cref{eq:proof-characterization-markovian-flow-eq1}. Similarly, it is easy to show that
        \begin{equation}
            F^{\star}(\tau'\preceq \cdot) = CP_{F}(\tau') = F^{\star}(\tau\preceq \cdot)P_{F}(s'\mid s).
            \label{eq:proof-characterization-markovian-flow-eq2}
        \end{equation}
        Any complete trajectory $\tau''$ going through the transition $s\rightarrow s'$ can also be decomposed as (1) a prefix (partial) trajectory $\overleftarrow{\tau}$ from $s_{0}$ to $s$, (2) the transition $s \rightarrow s'$, and (3) a suffix (partial) trajectory $\overrightarrow{\tau}$ from $s'$ to the terminal state $\terminal$. By definition of the edge flow
        {\allowdisplaybreaks%
        \begin{align}
            F^{\star}(s\rightarrow s') &= \sum_{\tau'': s\rightarrow s'\in\tau''}F^{\star}(\tau'') = C\sum_{\tau'': s\rightarrow s'\in\tau''}\prod_{t=0}^{T_{\tau''}}P_{F}(s_{t+1}\mid s_{t})\\
            &= C\Bigg(\sum_{\overleftarrow{\tau}: s_{0}\rightsquigarrow s}\prod_{t=0}^{T_{\overleftarrow{\tau}}-1}P_{F}(s_{t+1}\mid s_{t})\Bigg)P_{F}(s'\mid s)\Bigg(\sum_{\overrightarrow{\tau}: s'\rightsquigarrow \terminal}\prod_{t=T_{s'}}^{T_{\overrightarrow{\tau}}-1}P_{F}(s_{t+1}\mid s_{t})\Bigg)\\
            &= C\Bigg(\sum_{\overleftarrow{\tau}: s_{0}\rightsquigarrow s}P_{F}(\overleftarrow{\tau})\Bigg)P_{F}(s'\mid s)\Bigg(\sum_{\overrightarrow{\tau}: s'\rightsquigarrow \terminal}P_{F}(\overrightarrow{\tau})\Bigg)\\
            &= C\Bigg(\sum_{\overleftarrow{\tau}: s_{0}\rightsquigarrow s}P_{F}(\overleftarrow{\tau})\Bigg)P_{F}(s'\mid s)\label{eq:proof-characterization-markovian-flow-eq3}
        \end{align}}%
        where we again used \cref{lem:PF-distribution-suffix} in \cref{eq:proof-characterization-markovian-flow-eq3}. By \cref{prop:identities-state-edge-flows}, and using the fact that $P_{F}$ is a forward transition probability consistent with $\gG$, we have
        \begin{equation}
            F^{\star}(s) = \sum_{s''\in\children_{\gG}(s)}F^{\star}(s\rightarrow s'') = C\Bigg(\sum_{\overleftarrow{\tau}: s_{0}\rightsquigarrow s}P_{F}(\overleftarrow{\tau})\Bigg)\Bigg(\sum_{s''\in\children_{\gG}(s)}P_{F}(s''\mid s)\Bigg) = C\sum_{\overleftarrow{\tau}: s_{0}\rightsquigarrow s}P_{F}(\overleftarrow{\tau})
            \label{eq:proof-characterization-markovian-flow-eq4}
        \end{equation}
        Combining \cref{eq:proof-characterization-markovian-flow-eq3} and \cref{eq:proof-characterization-markovian-flow-eq4}, this shows in particular that
        \begin{equation}
            F^{\star}(s\rightarrow s') = F^{\star}(s)P_{F}(s'\mid s).
            \label{eq:proof-characterization-markovian-flow-eq5}
        \end{equation}
        And again combining \cref{eq:proof-characterization-markovian-flow-eq2} and \cref{eq:proof-characterization-markovian-flow-eq5}, we have $F^{\star}(s)F^{\star}(\tau'\preceq \cdot) = F^{\star}(s\rightarrow s')F^{\star}(\tau\preceq \cdot)$. Since this is true for any transition $s\rightarrow s'$ and any partial trajectory $\tau$ leading to $s$, by \cref{def:markovian-flow}, this shows that $F^{\star}$ is a Markovian flow.

        \item We have shown already in \cref{eq:proof-characterization-markovian-flow-eq5} that $P_{F}$ is the forward transition probability associated with $F^{\star}$. To show that $C$ is the total flow of $F^{\star}$, we can apply \cref{lem:PF-distribution-suffix}
        \begin{equation}
            Z^{\star} = \sum_{\tau\in\gT}F^{\star}(\tau) = C\sum_{\tau: s_{0}\rightsquigarrow \terminal}\prod_{t=0}^{T_{\tau}-1}P_{F}(s_{t+1}\mid s_{t}) = C\sum_{\tau: s_{0}\rightsquigarrow \terminal} P_{F}(\tau) = C.
        \end{equation}
    \end{enumerate}
\end{proof}
While there exist other characterizations of Markovian flows, for example based on a backward transition probability $P_{B}$ \citep{bengio2023gflownetfoundations}, we only include this result based on $P_{F}$ instead since this is sufficient here, in particular for proving the statements related to flow matching conditions in \cref{sec:flow-matching-conditions}.

\paragraph{Flow networks in graph theory} Markovian flows are closer to the standard notion of ``flow networks'' in graph theory, where flows are specified only along the edges of a graph, and not more globally over complete trajectories; see also \cref{sec:flow-matching-condition} for a characterization of a Markovian flow in terms of its edge flows. The key difference is that standard flow networks typically have a notion of ``capacity'' associated with each edge of the graph, which is the maximal amount of flow that can go through it. In the case of Markovian flows though, there is no such capacity since the flows may be arbitrarily large (equivalently, each edge has infinite capacity), and their values will only be later grounded using boundary conditions (see \cref{chap:generative-flow-networks}). Applications of flow networks in operational research include traffic regulations \citep{coclite2005trafficregulation} and identification of decease outbreaks \citep{pinto2012deceaseoutbreak}.

\subsection{Equivalence relation between trajectory flows}
\label{sec:equivalence-relation-flows}
For a fixed pointed DAG, there might be different flows that essentially behave the same for what matters most to us eventually: using the flow as a sampling mechanism, as described in \cref{sec:probability-distribution-induced-flow}. To formalize this, we define a notion of \emph{equivalent flows} if their flows through any edge are equal.

\begin{definition}[Equivalent flows]
    \label{def:equivalent-flows}
    Let $\gG$ be a pointed DAG, and $F_{1}^{\star}, F_{2}^{\star} \in \widebar{\gF}(\gG)$ two trajectory flows. $F_{1}^{\star}$ and $F_{2}^{\star}$ are said to be \emph{equivalent}, denoted $F_{1}^{\star} \sim F_{2}^{\star}$, if they coincide on the edge flows:
    \begin{equation}
        F_{1}^{\star} \sim F_{2}^{\star} \qquad \Leftrightarrow \qquad \forall s \rightarrow s' \in \gG,\quad F_{1}^{\star}(s \rightarrow s') = F_{2}^{\star}(s \rightarrow s').
        \label{eq:equivalent-flows}
    \end{equation}
\end{definition}
We can show that this is a properly defined equivalence relation (\ie reflexive, symmetric, and transitive). For example in \cref{fig:markovian-non-markovian-flows}, we can show that $F^{\star}_{1} \sim F^{\star}_{2}$ (recall that $F^{\star}_{2}$ is Markovian while $F^{\star}_{1}$ isn't); moreover $F^{\star}_{3} \sim F^{\star}_{4}$, but they are not equivalent to $F^{\star}_{1}$ and $F^{\star}_{2}$. It is clear that two equivalent flows will necessarily induce the same transition probabilities (in the sense of \cref{def:transition-probabilities-from-flow}). Under this equivalence relation, Markovian flows play a remarkable role in that if two Markovian flows are equivalent, then they are equal.
\begin{proposition}
    \label{prop:equivalent-markovian-flows}
    Two Markovian flows are equivalent if and only if they are equal.
\end{proposition}
\begin{proof}
    If two Markovian flows are equal, then they are obviously equivalent. Conversely, for a pointed DAG $\gG$, let $F_{1}^{\star}, F_{2}^{\star} \in \widebar{\gF}_{\markov}(\gG)$ be two Markovian flows such that $F_{1}^{\star} \sim F_{2}^{\star}$. Let $Z_{1}^{\star}$ and $Z_{2}^{\star}$ be their respective total flows. By \cref{prop:identities-state-edge-flows} and equivalence of the Markovian flows, we have
    \begin{equation}
        Z_{1}^{\star} = F_{1}^{\star}(s_{0}) = \sum_{s'\in\children_{\gG}(s_{0})}F_{1}^{\star}(s_{0}\rightarrow s') = \sum_{s'\in\children_{\gG}(s_{0})}F_{2}^{\star}(s_{0}\rightarrow s') = F_{2}^{\star}(s_{0}) = Z_{2}^{\star}.
    \end{equation}
    By \cref{prop:characterization-markovian-flow}, we can decompose both Markovian flows in terms of their corresponding forward transition probabilities, denoted $P_{F}^{1\star}$ and $P_{F}^{2\star}$ respectively, and their total flows $Z_{1}^{\star} = Z_{2}^{\star}$. By definition of these forward transition probabilities in term of the edge flows, and using the equivalence of the flows, we have for any complete trajectory $\tau = (s_{0}, s_{1}, \ldots, s_{T}, \terminal) \in \gT$
    \begin{align}
        F_{1}^{\star}(\tau) &= Z_{1}^{\star}\prod_{t=0}^{T}P_{F}^{1\star}(s_{t+1}\mid s_{t}) = Z_{1}^{\star}\prod_{t=0}^{T}\frac{F_{1}^{\star}(s_{t}\rightarrow s_{t+1})}{\sum_{s''\in\children_{\gG}(s_{t})}F_{1}^{\star}(s_{t}\rightarrow s'')}\\
        &= Z_{2}^{\star}\prod_{t=0}^{T}\frac{F_{2}^{\star}(s_{t}\rightarrow s_{t+1})}{\sum_{s''\in\children_{\gG}(s_{t})}F_{2}^{\star}(s_{t}\rightarrow s'')} = Z_{2}^{\star}\prod_{t=0}^{T}P_{F}^{2\star}(s_{t+1}\mid s_{t}) = F_{2}^{\star}(\tau).
    \end{align}
    This shows that these two Markovian flows are equal: $F_{1}^{\star} = F_{2}^{\star}$.
\end{proof}
The proposition above shows in particular that a Markovian flow is uniquely defined by its edge flows, which will be expanded on in \cref{thm:flow-matching-condition}. But since they represent a priori a smaller family of flows, it is natural to wonder what we are losing when working with Markovian flows as opposed to (general) trajectory flows. It turns out that from the point of view of equivalent flows, we are not losing much in terms of generality, in the sense that for any trajectory flow, there exists an equivalent Markovian flow.
\begin{proposition}
    \label{prop:equivalent-markovian-flow-from-trajectory-flow}
    Let $(\gG, F^{\star})$ be a flow network. There exists a unique Markovian flow $F'^{\star} \in \widebar{\gF}_{\markov}(\gG)$ such that $F'^{\star} \sim F^{\star}$.
\end{proposition}

\begin{proof}
    This proposition can be proven by applying \cref{thm:flow-matching-condition} stated further in the next section. If $F^{\star}$ is a trajectory flow, then by \cref{prop:identities-state-edge-flows} we have that the edge flows derived from $F^{\star}$ satisfy the flow matching condition \cref{eq:flow-matching-condition} for all $s'\in\gS$ such that $s'\neq s_{0}$. Therefore, based on \cref{eq:flow-matching-unique-markovian-flow}, there exists a unique Markovian flow $F'^{\star}\in\widebar{\gF}_{\markov}(\gG)$ defined, for any complete trajectory $\tau = (s_{0}, s_{1}, \ldots, s_{T}, \terminal) \in \gT$, by
    \begin{equation}
        F'^{\star}(\tau) = F^{\star}(s_{0}\rightarrow s_{1})\prod_{t=1}^{T}\frac{F^{\star}(s_{t}\rightarrow s_{t+1})}{\sum_{s''\in\children_{\gG}(s_{t})}F^{\star}(s_{t}\rightarrow s'')},
    \end{equation}
    with the convention $s_{T+1} = \terminal$, such that the edge flows of $F'^{\star}$ match the edge flows of $F^{\star}$ (the edge flows of $F^{\star}$ being the function over the edges of $\gG$ that satisfies the flow matching condition in the application of \cref{thm:flow-matching-condition}). By definition, this means that $F'^{\star}\sim F^{\star}$.
\end{proof}
This shows that in each equivalence class stands out a particular flow which is Markovian, a property that no other member of the equivalence class has. In other words, the space of Markovian flows is equivalent to the quotient space of trajectory flows under the relation in \cref{def:equivalent-flows}: $\widebar{\gF}_{\markov}(\gG) \equiv \widebar{\gF}(\gG)\, /\! \sim$. Based on this observation, we will work exclusively with Markovian flows throughout this thesis from now on.

\section{Flow matching conditions}
\label{sec:flow-matching-conditions}
\begin{figure}[t]
    \centering
    \begin{adjustbox}{center}
        \includegraphics[width=480pt]{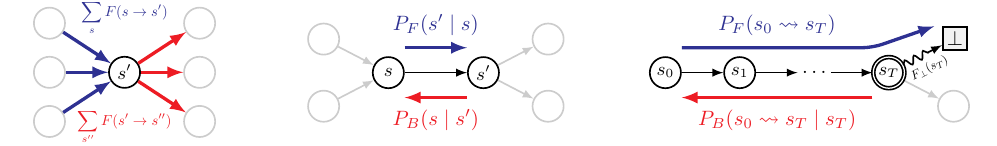}%
    \end{adjustbox}
    \begin{adjustbox}{center}%
        \begin{subfigure}[b]{120pt}%
            \caption{Flow matching}%
            \label{fig:flow-matching-condition}%
        \end{subfigure}%
        \begin{subfigure}[b]{180pt}%
            \caption{Detailed balance}%
            \label{fig:detailed-balance-condition}%
        \end{subfigure}%
        \begin{subfigure}[b]{180pt}%
            \caption{Trajectory balance}%
            \label{fig:trajectory-balance-condition}%
        \end{subfigure}%
    \end{adjustbox}\vspace*{1ex}
    \begin{adjustbox}{center}
        \includegraphics[width=480pt]{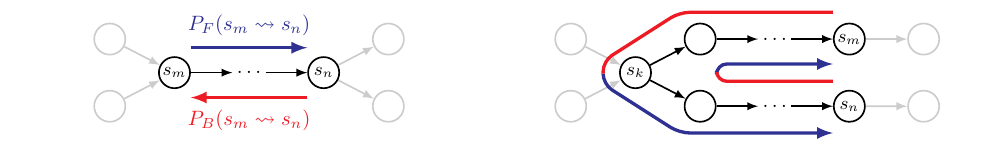}%
    \end{adjustbox}
    \begin{adjustbox}{center}%
        \begin{subfigure}[b]{240pt}%
            \caption{Sub-trajectory balance}%
            \label{fig:sub-trajectory-balance-condition}%
        \end{subfigure}%
        \begin{subfigure}[b]{240pt}%
            \caption{Generalized sub-trajectory balance}%
            \label{fig:generalized-sub-trajectory-balance-condition}%
        \end{subfigure}%
    \end{adjustbox}
    \caption[Illustration of the flow matching conditions]{Illustration of the flow matching conditions. (a) The flow matching condition operates at the level of states $s'$ (\cref{sec:flow-matching-condition}) (b) The detailed balance condition operates at the level of transitions (\cref{sec:detailed-balance-condition}). (c) The trajectory balance condition operates at the level of complete trajectories (\cref{sec:trajectory-balance-condition}). (d-e) The sub-trajectory balance condition operates at the level of partial trajectories (directed, and ``forward then backward''; \cref{sec:sub-trajectory-balance-condition}).}
    \label{fig:flow-matching-conditions}
\end{figure}

In the previous section we showed multiple properties of flow networks, \emph{provided that we know that $F^{\star}$ is a valid (Markovian) flow}. However, apart from the characterization \cref{prop:characterization-markovian-flow} that serves more as a way to verify that a flow is Markovian, we have not given any details on \emph{how} we can construct a valid flow over a given pointed DAG $\gG$. We will show in this section that this can be done by finding quantities that satisfy some \emph{conservation laws} at a certain level of granularity. These are generally called \emph{flow matching conditions}, and are illustrated in \cref{fig:flow-matching-conditions}.

\subsection{Flow matching condition}
\label{sec:flow-matching-condition}
Going back to the intuitive interpretation of a flow network as some fluid going through pipes (\cref{sec:flow-networks}), physics tells us that at any branching point, the total amount of fluid going into it has to be equal to the total amount of fluid going out of it. In other words, apart from the initial and terminal states which play a special role, there is no ``new fluid'' being created, nor is there any fluid ``leaking'' out of the pipes. This intuitive observation can be formalized as the following \emph{flow matching condition} \citep{bengio2021gflownet}, which characterizes the condition a function $F$ must satisfy to be a valid edge flow. Note that here we are using the notation ``$F$'' (a priori an arbitrary function) and not ``$F^{\star}$'', which we used to denote a valid flow in the previous section.
\begin{theorem}[Flow matching condition]
    \label{thm:flow-matching-condition}\index{Flow matching!Condition|textbf}\glsadd{fm}
    Let $\gG = (\widebar{\gS}, \gA)$ be a pointed DAG. A function $F: \gA \rightarrow \sR_{+}$ defines the edge flow of a unique Markovian flow $F^{\star}$ if and only if it satisfies the following \emph{flow matching condition} for all $s' \in \gS$ such that $s' \neq s_{0}$:
    \begin{equation}
        \sum_{s\in\parents_{\gG}(s')}F(s \rightarrow s') = \sum_{s''\in\children_{\gG}(s')}F(s'\rightarrow s'').
        \label{eq:flow-matching-condition}
    \end{equation}
    Moreover, under these conditions, the unique Markovian flow $F^{\star}$ whose edge flows match the function $F$ is defined, for any complete trajectory $\tau = (s_{0}, s_{1}, \ldots, s_{T}, \terminal) \in \gT$, by
    \begin{equation}
        F^{\star}(\tau) = F(s_{0} \rightarrow s_{1})\prod_{t=1}^{T}\frac{F(s_{t} \rightarrow s_{t+1})}{\sum_{s'' \in \children_{\gG}(s_{t})}F(s_{t} \rightarrow s'')},
        \label{eq:flow-matching-unique-markovian-flow}
    \end{equation}
    with the convention $s_{T+1} = \terminal$.
\end{theorem}
\begin{figure}[t]
    \centering
    \begin{adjustbox}{center}
    \includegraphics[width=480pt]{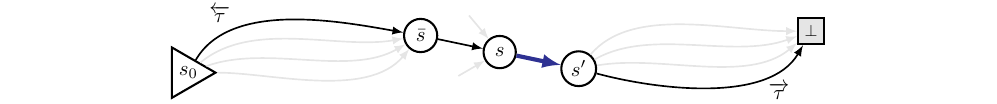}
    \end{adjustbox}
    \caption[Decomposition of a complete trajectory going through the transition $s\rightarrow s'$]{Decomposition of a complete trajectory going through the transition $s\rightarrow s'$ into a prefix $\overleftarrow{\tau}$ to a parent $\bar{s}$, a transition $\bar{s} \rightarrow s$, the transition $s\rightarrow s'$, and a suffix $\overrightarrow{\tau}$.}
    \label{fig:illustration-trajectory-proof}
\end{figure}
\begin{proof}
    $\Rightarrow$: If $F^{\star}$ is a Markovian flow, then its edge flow satisfies the flow matching condition \cref{eq:flow-matching-condition} for all $s'\in\gS$ such that $s'\neq s_{0}$ by \cref{prop:identities-state-edge-flows}. This proves the necessity of the flow matching condition for the edge flows of a Markovian flow.

    $\Leftarrow$: To prove sufficiency, we will prove that under the flow matching condition (1) the edge flows of the trajectory flow $F^{\star}$ match the function $F$, (2) the trajectory flow $F^{\star}$ is Markovian, and finally (3) the Markovian flow $F^{\star}$ is unique. In what follows, we will assume that the function $F$ satisfies \cref{eq:flow-matching-condition} for all $s'\neq s_{0}$.
    \begin{enumerate}[leftmargin=*]
        \item Let $s \rightarrow s' \in\gG$ be an arbitrary transition; we will assume for now that $s \neq s_{0}$. Any complete trajectory $\tau$ going through $s \rightarrow s'$ can be decomposed into (1) a prefix (partial) trajectory $\overleftarrow{\tau}$ from $s_{0}$ to some parent $\bar{s}\in\parents_{\gG}(s)$, (2) a transition $\bar{s} \rightarrow s$, (3) the transition $s \rightarrow s'$, and (4) a suffix (partial) trajectory $\overrightarrow{\tau}$ from $s'$ to the terminal state $\terminal$; see \cref{fig:illustration-trajectory-proof} for an illustration. By definition of the edge flow \cref{eq:edge-flow}, we have
        \begin{align}
            F^{\star}(s \rightarrow s') &= \sum_{\tau:s\rightarrow s'\in \tau}F^{\star}(\tau)\\
            &= \Bigg(\sum_{\bar{s}\in\parents_{\gG}(s)}\sum_{\overleftarrow{\tau}: s_{0}\rightsquigarrow \bar{s}}F(s_{0}\rightarrow s_{1})\prod_{t=1}^{T_{\overleftarrow{\tau}}}\frac{F(s_{t}\rightarrow s_{t+1})}{\sum_{s''\in\children_{\gG}(s_{t})}F(s_{t}\rightarrow s'')}\Bigg)\nonumber\\
            &\qquad \qquad \times \frac{F(s \rightarrow s')}{\sum_{s''\in\children_{\gG}(s)}F(s\rightarrow s'')}\Bigg(\sum_{\overrightarrow{\tau}: s'\rightsquigarrow \terminal}\prod_{t=T_{s'}}^{T_{\overrightarrow{\tau}}-1}\frac{F(s_{t} \rightarrow s_{t+1})}{\sum_{s''\in\children_{\gG}(s_{t})}F(s_{t}\rightarrow s'')}\Bigg)\label{eq:proof-flow-matching-condition-eq1}
        \end{align}
        where we used the convention $s_{T_{\overleftarrow{\tau}}+1} = s$ in the first term of the decomposition above. We introduce the notation $P_{F}$ to denote the distribution defined using the function $F$ by
        \begin{equation}
            P_{F}(s_{t+1}\mid s_{t}) \triangleq \frac{F(s_{t}\rightarrow s_{t+1})}{\sum_{s''\in\children_{\gG}(s_{t})}F(s_{t}\rightarrow s'')}.\label{eq:proof-flow-matching-condition-eq6}
        \end{equation}
        Note that while it is true that $P_{F}$ is the forward transition probability associated with $F^{\star}$, we will only use it as a new notation in this proof. $P_{F}$ is clearly a forward transition probability consistent with $\gG$. Using \cref{lem:PF-distribution-suffix}, the last term of \cref{eq:proof-flow-matching-condition-eq1} is equal to $1$:
        \begin{equation}
            \sum_{\overrightarrow{\tau}: s'\rightsquigarrow \terminal}\prod_{t=T_{s'}}^{T_{\overrightarrow{\tau}}-1}\frac{F(s_{t}\rightarrow s_{t+1})}{\sum_{s''\in\children_{\gG}(s_{t})}F(s_{t}\rightarrow s'')} = \sum_{\overrightarrow{\tau}: s'\rightsquigarrow \terminal}\prod_{t=T_{s'}}^{T_{\overrightarrow{\tau}}-1}P_{F}(s_{t+1}\mid s_{t}) = \sum_{\overrightarrow{\tau}:s'\rightsquigarrow \terminal}P_{F}(\overrightarrow{\tau}) = 1
            \label{eq:proof-flow-matching-condition-eq2}
        \end{equation}
        Similarly, we also introduce the notation $P_{B}$ to denote the backward transition probability defined using the function $F$ by
        \begin{equation}
            P_{B}(s_{t}\mid s_{t+1}) \triangleq \frac{F(s_{t}\rightarrow s_{t+1})}{\sum_{s''\in\parents_{\gG}(s_{t+1})}F(s''\rightarrow s_{t+1})}.
            \label{eq:proof-flow-matching-condition-eq5}
        \end{equation}
        Again, even though $P_{B}$ is the backward transition probability associated with $F^{\star}$, we only use it as a new notation here. $P_{B}$ is a backward transition probability consistent with $\gG$. We can rewrite the first term of \cref{eq:proof-flow-matching-condition-eq1} (for a fixed $\bar{s} \in \parents_{\gG}(s)$) in terms of $P_{B}$:
        {\allowdisplaybreaks
        \begin{align}
            \sum_{\overleftarrow{\tau}: s_{0}\rightsquigarrow \bar{s}}F(s_{0}\rightarrow s_{1})&\prod_{t=1}^{T_{\overleftarrow{\tau}}}\frac{F(s_{t}\rightarrow s_{t+1})}{\sum_{s''\in\children_{\gG}(s_{t})}F(s_{t}\rightarrow s'')}\\
            &= \sum_{\overleftarrow{\tau}:s_{0}\rightsquigarrow \bar{s}} F(s_{0} \rightarrow s_{1})\prod_{t=1}^{T_{\overleftarrow{\tau}}}\frac{F(s_{t}\rightarrow s_{t+1})}{\sum_{s''\in\parents_{\gG}(s_{t})}F(s''\rightarrow s_{t})}\label{eq:proof-flow-matching-condition-eq3}\\
            &= F(\bar{s}\rightarrow s)\sum_{\overleftarrow{\tau}: s_{0}\rightsquigarrow \bar{s}}\prod_{t=1}^{T_{\overleftarrow{\tau}}}\frac{F(s_{t-1}\rightarrow s_{t})}{\sum_{s''\in\parents_{\gG}(s_{t})}F(s''\rightarrow s_{t})}\\
            &= F(\bar{s}\rightarrow s)\sum_{\overleftarrow{\tau}: s_{0}\rightsquigarrow \bar{s}}\prod_{t=1}^{T_{\overleftarrow{\tau}}}P_{B}(s_{t-1}\mid s_{t})\\
            &= F(\bar{s}\rightarrow s)\sum_{\overleftarrow{\tau}: s_{0}\rightsquigarrow \bar{s}}P_{B}(\overleftarrow{\tau}) = F(\bar{s} \rightarrow s)\label{eq:proof-flow-matching-condition-eq4}
        \end{align}}%
        where we used the flow matching condition applied to $s_{t} \neq s_{0}$ in \cref{eq:proof-flow-matching-condition-eq3}, and \cref{lem:PB-distribution-prefix} in \cref{eq:proof-flow-matching-condition-eq4}. Using both \cref{eq:proof-flow-matching-condition-eq2} and \cref{eq:proof-flow-matching-condition-eq4} in the expression of $F^{\star}(s\rightarrow s')$ in \cref{eq:proof-flow-matching-condition-eq1}, we get
        \begin{equation}
            F^{\star}(s\rightarrow s') = \Bigg(\sum_{\bar{s}\in\parents_{\gG}(s)}F(\bar{s} \rightarrow s)\Bigg)\frac{F(s \rightarrow s')}{\sum_{s''\in\children_{\gG}(s)}F(s\rightarrow s'')} = F(s\rightarrow s'),
        \end{equation}
        where we again used the flow matching condition, this time applied to $s \neq s_{0}$. Note that this derivation is also valid if $s' = \terminal$, since in that case the third term in the decomposition \cref{eq:proof-flow-matching-condition-eq1} vanishes (being equal to \cref{eq:proof-flow-matching-condition-eq2}).

        If $s = s_{0}$, then we can directly write the edge flow $F^{\star}(s_{0} \rightarrow s')$ as
        {\allowdisplaybreaks
        \begin{align}
            F^{\star}(s_{0}\rightarrow s') &= \sum_{\tau: s_{0}\rightarrow s' \in \tau}F^{\star}(\tau)\\
            &= F(s_{0} \rightarrow s')\sum_{\tau':s'\rightsquigarrow \terminal}\prod_{t=1}^{T_{\tau'}-1}\frac{F(s_{t}\rightarrow s_{t+1})}{\sum_{s''\in\children_{\gG}(s_{t})}F(s_{t}\rightarrow s'')}\\
            &= F(s_{0}\rightarrow s')\sum_{\tau':s'\rightsquigarrow \terminal}\prod_{t=1}^{T_{\tau'}-1}P_{F}(s_{t+1}\mid s_{t})\\
            &= F(s_{0}\rightarrow s')\sum_{\tau': s'\rightsquigarrow \terminal}P_{F}(\tau') = F(s_{0}\rightarrow s'),\label{eq:proof-flow-matching-condition-eq7}
        \end{align}}%
        where we used \cref{lem:PF-distribution-suffix} to conclude in \cref{eq:proof-flow-matching-condition-eq7}. This proves that for any transition $s \rightarrow s' \in \gG$, the edge flow of $F^{\star}$ matches the function $F$: $F^{\star}(s\rightarrow s') = F(s\rightarrow s')$.

        \item Using our notation \cref{eq:proof-flow-matching-condition-eq6}, the trajectory flow $F^{\star}$ can be rewritten as
        \begin{align}
            F^{\star}(\tau) &= F(s_{0}\rightarrow s_{1})\prod_{t=1}^{T}\frac{F(s_{t}\rightarrow s_{t+1})}{\sum_{s''\in\children_{\gG}(s_{t})}F(s_{t}\rightarrow s'')}\\
            &= \Bigg(\sum_{s''\in\children_{\gG}(s_{0})}F(s_{0}\rightarrow s'')\Bigg)\prod_{t=0}^{T}P_{F}(s_{t+1}\mid s_{t})
        \end{align}
        Since the sum of flows going out of $s_{0}$ is a constant (independent of $\tau$), and $P_{F}$ is a forward transition probability consistent with $\gG$, we can conclude by \cref{prop:characterization-markovian-flow} that $F^{\star}$ is a Markovian flow.

        \item Suppose that there exists another Markovian flow $F'^{\star}$ such that the function $F$ matches its edge flow. By definition of equivalent flows, and since we already know that $F^{\star}$ also has its edge flow defined by the function $F$, we have $F'^{\star} \sim F^{\star}$. By \cref{prop:equivalent-markovian-flows}, this means that $F'^{\star} = F^{\star}$, showing the unicity of the Markovian flow $F^{\star}$.
    \end{enumerate}
    \vspace*{-1em}
\end{proof}
The flow matching condition is a condition that applies locally at the level of states $s'\neq s_{0}$. It is reminiscent of the \emph{global balance} condition in statistical physics \citep{mallick2009statphy}, which finds applications in irreversible Markov chain Monte Carlo algorithms \citep{suwa2010mcmcwithoutdetailedbalance,hukushima2016irreversiblemcmc}. But more generally, the connection with fluid dynamics allows us to make a number of analogies with various domains of physics; we will give an example from electrical engineering below. In fluid dynamics, a fluid with density $\rho$ and velocity $\vu$ ($\vu$ being a vector field) satisfy the following \emph{continuity equation}
\begin{equation}
    \frac{\partial \rho}{\partial t} + \mathrm{div}(\rho \vu) = 0,
    \label{eq:continuity-equation}
\end{equation}
where ``$\mathrm{div}$'' is the divergence operator. This equation can be interpreted as a conservation of mass in an infinitesimal volume. Note that the use of the continuity equation in machine learning has gained popularity recently with \emph{flow matching} generative models \citep{lipman2023flowmatching,liu2023rectifiedflow,albergo2023stochasticinterpolants,tong2024conditionalflowmatching}. Although they are loosely related to the ``flow matching'' condition presented here (as we will see below), it is important to distinguish these two concepts even if they share the same name.

In the case where the fluid is incompressible, meaning that its density is constant over time, this continuity equation becomes $\mathrm{div}(\vu) = 0$. The divergence being a local measure of how much fluid goes out of an infinitesimal volume, having it equal to zero means that the amount of flow going into that volume (with negative flux) and out of it (with positive flux) cancel each other out. If we interpret the function $F(s\rightarrow s')$ as being a discrete version of the vector field $\vu$, then this corresponds to the flow matching condition \cref{eq:flow-matching-condition}.

\begin{figure}[t]
    \centering
    \begin{adjustbox}{center}
        \includegraphics[width=480pt]{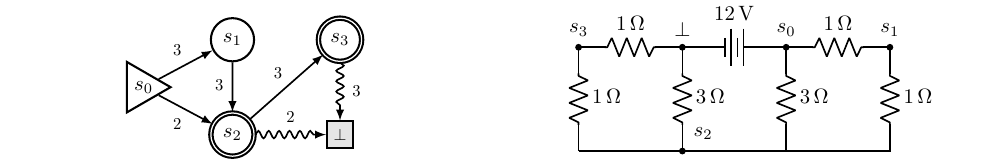}%
    \end{adjustbox}
    \begin{adjustbox}{center}%
        \begin{subfigure}[b]{240pt}%
            \caption{Flow network}%
            \label{fig:kirchhoff-law-flow-network}%
        \end{subfigure}%
        \begin{subfigure}[b]{240pt}%
            \caption{Electrical network}%
            \label{fig:kirchhoff-law-electrical}%
        \end{subfigure}%
    \end{adjustbox}
    \caption[Analogy between the flow matching condition and Kirchhoff's current law]{Analogy between the flow matching condition and Kirchhoff's current law. (a) A Markovian flow network with the edge flow defined at each transition. (b) An equivalent electrical circuit with a power source and 6 resistors. The values of the resistances are chosen in such a way that the current at each branch is equal to the edge flow in the corresponding flow network in (a). This electrical network must satisfy Kirchhoff's current law.}
    \label{fig:kirchhoff-law}
\end{figure}

\paragraph{Kirchhoff's current law} The flow matching condition is analogous to the first Kirchhoff's law in electrical networks, also known as \emph{Kirchhoff's current law}. This is a conservation law stating that at any junction point in an electrical network, the sum of currents flowing into that point is equal to the sum of current flowing out of it. In \cref{fig:kirchhoff-law} we illustrate this correspondence between the flow matching condition on a flow network and Kirchhoff's law on an equivalent electrical network. With this interpretation, the edge flow in \cref{fig:kirchhoff-law-flow-network} is exactly the current between the two corresponding nodes in \cref{fig:kirchhoff-law-electrical}. One difference between these two points of view is that the electrical circuit is a \emph{closed loop}, whereas the flow network goes from $s_{0}$ to the terminal state $\terminal$; we will come back to this closed loop perspective of flow networks in \cref{sec:cyclic-nature-flow-networks}.

\subsection{Detailed balance condition}
\label{sec:detailed-balance-condition}
Taking inspiration from the \emph{detailed balance} equations from the literature on Markov chains \citep{grimmett2020probability} characterizing the stationary distribution of a reversible Markov chain, we can introduce another characterization of a Markovian flow, this time working at the level of transitions instead of states as in the previous section. Instead of parametrizing a flow with its edge flows directly, we will use a combination of its state flows and transition probabilities. We call this the \emph{detailed balance condition} \citep{bengio2023gflownetfoundations}.

\begin{theorem}[Detailed balance condition]
    \label{thm:detailed-balance-condition}\index{Detailed balance!Condition|textbf}\glsadd{db}
    Let $\gG = (\widebar{\gS}, \gA)$ be a pointed DAG. A function $F: \gS \rightarrow \sR_{+}$ defines the state flow, $P_{F}: \gS \rightarrow \Delta(\children_{\gG})$ the forward transition probabilities, and $P_{B}: \gS \rightarrow \Delta(\parents_{\gG})$ the backward transition probabilities of a unique Markovian flow $F^{\star}$ if and only if they satisfy the following \emph{detailed balance condition} for all transitions $s\rightarrow s' \in \gG$ such that $s' \neq \terminal$:
    \begin{equation}
        F(s)P_{F}(s'\mid s) = F(s')P_{B}(s\mid s').
        \label{eq:detailed-balance-condition}
    \end{equation}
    Moreover, under these conditions, the unique Markovian flow $F^{\star}$ whose state flow, forward transition probabilities and backward transitions probabilities match $F$, $P_{F}$, and $P_{B}$ respectively is defined, for any complete trajectory $\tau = (s_{0}, s_{1}, \ldots, s_{T}, \terminal) \in \gT$, by
    \begin{equation}
        F^{\star}(\tau) = F(s_{0})\prod_{t=0}^{T}P_{F}(s_{t+1}\mid s_{t}),
        \label{eq:detailed-balance-unique-markovian-flow}
    \end{equation}
    with the convention $s_{T+1} = \terminal$.
\end{theorem}

\begin{proof}
    $\Rightarrow$: If $F^{\star}$ is a Markovian flow, let $P^{\star}_{F}$ and $P^{\star}_{B}$ denote its forward transition probabilities and backward transition probabilities respectively. By \cref{def:transition-probabilities-from-flow}, we can express both $P^{\star}_{F}$ and $P^{\star}_{B}$ in terms of the state and edge flows of $F^{\star}$: for any transition $s \rightarrow s' \in \gG$ such that $s' \neq \terminal$, we have
    \begin{align}
        P^{\star}_{F}(s'\mid s) &= \frac{F^{\star}(s\rightarrow s')}{F^{\star}(s)} & P^{\star}_{B}(s\mid s') &= \frac{F^{\star}(s\rightarrow s')}{F^{\star}(s')}.
    \end{align}
    With two ways of writing the edge flow $F^{\star}(s\rightarrow s')$, this shows that the detailed balance condition \cref{eq:detailed-balance-condition} is satisfied by the state flow, the forward transition probabilities, and the backward transition probabilities of $F^{\star}$. This proves the necessity of the detailed balance condition.

    $\Leftarrow$: Let $F$, $P_{F}$, and $P_{B}$ be functions satisfying the detailed balance condition for all transitions $s\rightarrow s'\in \gG$ such that $s'\neq \terminal$, and let $F^{\star}$ be the trajectory flow defined in \cref{eq:detailed-balance-unique-markovian-flow}. Since $F(s_{0})$ is a constant and $P_{F}$ is a forward transition probability consistent with $\gG$, the flow $F^{\star}$ is Markovian by \cref{prop:characterization-markovian-flow}. Moreover, this also shows that $P_{F}$ matches the forward transition probabilities of $F^{\star}$ (again, by \cref{prop:characterization-markovian-flow}). To show the sufficiency of the detailed balance condition, we then need to prove that (1) $F$ matches the state flow of $F^{\star}$, (2) $P_{B}$ matches its backward transition probabilities, and that (3) the Markovian flow $F^{\star}$ is unique.

    \begin{enumerate}[leftmargin=*]
        \item Let $s\in\gS$ be an arbitrary state. From the proof of \cref{prop:characterization-markovian-flow}, and in particular \cref{eq:proof-characterization-markovian-flow-eq4}, we know that the state flow $F^{\star}(s)$ can be written as
        {\allowdisplaybreaks%
        \begin{align}
            F^{\star}(s) &= F(s_{0})\sum_{\tau: s_{0}\rightsquigarrow s}P_{F}(\tau) = F(s_{0})\sum_{\tau: s_{0}\rightsquigarrow s}\prod_{t=0}^{T_{\tau}-1}P_{F}(s_{t+1}\mid s_{t})\\
            &= F(s_{0})\sum_{\tau: s_{0}\rightsquigarrow s}\prod_{t=0}^{T_{\tau}-1}\frac{F(s_{t+1})}{F(s_{t})}P_{B}(s_{t}\mid s_{t+1})\label{eq:proof-detailed-balance-condition-eq1}\\
            &= F(s)\sum_{\tau: s_{0}\rightsquigarrow s}\prod_{t=1}^{T_{\tau}}P_{B}(s_{t-1}\mid s_{t})\\
            &= F(s)\sum_{\tau: s_{0}\rightsquigarrow s}P_{B}(\tau\mid s) = F(s)\label{eq:proof-detailed-balance-condition-eq2}
        \end{align}}%
        where we used the convention $s_{T_{\tau}} = s$, the detailed balance condition in \cref{eq:proof-detailed-balance-condition-eq1} (since $s_{t+1} \neq \terminal$), and \cref{lem:PB-distribution-prefix} in \cref{eq:proof-detailed-balance-condition-eq2}. We implicitly assumed that $F(s) > 0$ for all states, which is a reasonable assumption since we guaranteed that the state flows $F^{\star}(s) > 0$ are positive. This shows that the function $F$ matches the state flow of $F^{\star}$.

        \item Let $s\rightarrow s' \in \gG$ be an arbitrary transition such that $s'\neq \terminal$. Again using the proof of \cref{prop:characterization-markovian-flow}, in particular \cref{eq:proof-characterization-markovian-flow-eq5}, the detailed balance condition applied to $s\rightarrow s'$, and using \cref{eq:proof-detailed-balance-condition-eq2} above to move from $F^{\star}(s)$ to $F(s)$ and back, we can write the edge flow of $F^{\star}$ as
        \begin{equation}
            F^{\star}(s\rightarrow s') = F^{\star}(s)P_{F}(s'\mid s) = F(s)P_{F}(s'\mid s) = F(s')P_{B}(s\mid s') = F^{\star}(s')P_{B}(s\mid s').
        \end{equation}
        By definition, this shows that $P_{B}$ matches the backward transition probabilities of $F^{\star}$:
        \begin{equation}
            P^{\star}_{B}(s\mid s') = \frac{F^{\star}(s\rightarrow s')}{F^{\star}(s')} = P_{B}(s\mid s').
        \end{equation}

        \item Suppose that there exists another Markovian flow $F'^{\star}$ such that $F$ matches its state flow, $P_{F}$ matches its forward transition probabilities, and $P_{B}$ matches its backward transition probabilities. We denote by $P'^{\star}_{F}$ the forward transition probabilities associated with $F'^{\star}$. By definition, we then have for any transition $s\rightarrow s' \in \gG$ (where $s'$ may be equal to the terminal state $\terminal$)
        \begin{equation}
            F'^{\star}(s\rightarrow s') = F'^{\star}(s)P'^{\star}_{F}(s'\mid s) = F(s)P_{F}(s'\mid s) = F^{\star}(s)P^{\star}_{F}(s'\mid s) = F^{\star}(s\rightarrow s'),
        \end{equation}
        where we used the fact that $F^{\star}$ and $F'^{\star}$ have both their matching state flows and forward transition probabilities equal to $F$ and $P_{F}$ respectively. Since this is valid for all transitions $s\rightarrow s'\in\gG$, by definition of equivalent flows we have $F'^{\star} \sim F^{\star}$. By \cref{prop:equivalent-markovian-flows}, this means that the two flows are equal, showing the unicity of $F^{\star}$.
    \end{enumerate}
    \vspace*{-1em}
\end{proof}
In their standard form, the detailed balance equations in the Markov chain literature assume that the transition kernel is reversible, meaning that the same kernel appears on both sides of the equation. With our notations here, this would mean having $P_{F}(s'\mid s)$ on the \gls{lhs} of \cref{eq:detailed-balance-condition}, and $P_{F}(s\mid s')$ on the \gls{rhs}. However, the latter is undefined in our case since we necessarily have $s \notin \children_{\gG}(s')$ ($\gG$ is a pointed DAG). This detailed balance condition is closer to the concept of \emph{skew detailed balance} equations used to define irreversible Markov chain Monte Carlo algorithms \citep{turitsyn2011skeweddb,hukushima2013irreversiblemcmc}, which use two separate kernels. We will come back to the interpretation of the state flow $F(s)$ as the stationary distribution of $P_{F}$ in \cref{sec:gflownet-discrete-spaces}.

Since the detailed balance condition operates at the level of individual transitions, it may seem more efficient at first glance than the flow matching condition which requires summing over all incoming and outgoing transitions. However, note that since $P_{F}(\cdot \mid s)$ and $P_{B}(\cdot \mid s)$ are consistent transition probabilities, they implicitly require summing over the children and the parents of $s$ respectively due to the normalization. The detailed balance condition offers an alternative characterization of a Markovian flow, and no condition is a priori better than another. Finally for certain structures of $\gG$ which we will study in \cref{sec:gflownet-over-dags}, it is possible to adapt the detailed balance condition to not depend on a state flow function $F(s)$ anymore.

\subsection{Trajectory balance condition}
\label{sec:trajectory-balance-condition}
While the detailed balance condition works at the level of individual transitions, at the other end of the spectrum the \emph{trajectory balance condition} \citep{malkin2022trajectorybalance} is a conservation law that operates at the level of complete trajectories in $\gT$.
\begin{theorem}[Trajectory balance condition]
    \label{thm:trajectory-balance-condition}\index{Trajectory balance!Condition|textbf}\glsadd{tb}
    Let $\gG = (\widebar{\gS}, \gA)$ be a pointed DAG, and $\gX \subseteq \gS$ the set of terminating states in $\gG$. A scalar $Z > 0$ defines the total flow, and functions $F_{\terminal}: \gX \rightarrow \sR_{+}$ the edge flow of the terminating transitions, $P_{F}: \gS \rightarrow \Delta(\children_{\gG})$ the forward transition probabilities, and $P_{B}: \gS \rightarrow \Delta(\parents_{\gG})$ the backward transition probabilities of a unique Markovian flow $F^{\star}$ if and only if they satisfy the following \emph{trajectory balance condition} for all complete trajectories $\tau=(s_{0}, s_{1}, \ldots, s_{T}, \terminal) \in \gT$:
    \begin{equation}
        Z\prod_{t=0}^{T}P_{F}(s_{t+1}\mid s_{t}) = F_{\terminal}(s_{T})\prod_{t=1}^{T}P_{B}(s_{t-1}\mid s_{t}),
        \label{eq:trajectory-balance-condition}
    \end{equation}
    where we used the convention $s_{T+1} = \terminal$, and $s_{T} \in\gX$ is a terminating state. Moreover, under these conditions, the unique Markovian flow $F^{\star}$ whose total flow, edge flow of the terminating transitions, forward transition probabilities and backward transition probabilities match $Z$, $F_{\terminal}$, $P_{F}$, and $P_{B}$ respectively is defined by
    \begin{equation}
        F^{\star}(\tau) = Z\prod_{t=0}^{T}P_{F}(s_{t+1}\mid s_{t}).
        \label{eq:trajectory-balance-unique-markovian-flow}
    \end{equation}
\end{theorem}

\begin{proof}
    $\Rightarrow$: If $F^{\star}$ is a Markovian flow, let $P^{\star}_{F}$, $P^{\star}_{B}$, and $Z^{\star}$ denote its forward transition probabilities, its backward transition probabilities, and its total flow respectively. By \cref{prop:initial-flow-total-flow}, we know that $Z^{\star} = F^{\star}(s_{0})$. By definition of the forward transition probabilities and the backward transition probabilities in terms of the edge flows, we have for any complete trajectory $\tau = (s_{0}, s_{1}, \ldots, s_{T}, \terminal)$
    \begin{align}
        Z^{\star}\prod_{t=0}^{T}P^{\star}_{F}(s_{t+1}\mid s_{t}) &= F^{\star}(s_{0})\prod_{t=0}^{T}\frac{F^{\star}(s_{t}\rightarrow s_{t+1})}{F^{\star}(s_{t})} = \frac{\prod_{t=0}^{T}F^{\star}(s_{t}\rightarrow s_{t+1})}{\prod_{t=1}^{T}F^{\star}(s_{t})}\\
        F^{\star}(s_{T}\rightarrow \terminal)\prod_{t=1}^{T}P^{\star}_{B}(s_{t-1}\mid s_{t}) &= F^{\star}(s_{T}\rightarrow \terminal)\prod_{t=1}^{T}\frac{F^{\star}(s_{t-1}\rightarrow s_{t})}{F^{\star}(s_{t})} = \frac{\prod_{t=0}^{T}F^{\star}(s_{t}\rightarrow s_{t+1})}{\prod_{t=1}^{T}F^{\star}(s_{t})}
    \end{align}
    with the convention $s_{T+1} = \terminal$. This shows that the trajectory balance condition \cref{eq:trajectory-balance-condition} is satisfied by the total flow, the edge flow of the terminating transitions, the forward transition probabilities, and the backward transition probabilities of $F^{\star}$. This proves the necessity of the trajectory balance condition.

    $\Leftarrow$: Let $Z$ be a scalar and $F_{\terminal}$, $P_{F}$, and $P_{B}$ be functions satisfying the trajectory balance condition for all complete trajectories $\tau = (s_{0}, s_{1}, \ldots, s_{T}, \terminal) \in \gT$. By \cref{prop:characterization-markovian-flow}, the flow $F^{\star}$ defined in \cref{eq:trajectory-balance-unique-markovian-flow} is Markovian. Moreover, this also shows that $Z$ is the total flow and $P_{F}$ matches the forward transition probabilities of $F^{\star}$ (again, by \cref{prop:characterization-markovian-flow}). To show the sufficiency of the trajectory balance condition, we then need to prove that (1) $F_{\terminal}$ matches the edge flow of the terminating transitions of $F^{\star}$, (2) $P_{B}$ matches its backward transition probabilities, and that (3) the Markovian flow $F^{\star}$ is unique.
    \begin{enumerate}[leftmargin=*]
        \item Let $x\in\gX$ be an arbitrary terminating state. By definition of the flow $F^{\star}$, we can write the edge flow of the terminating transition $F^{\star}(x\rightarrow \terminal)$ as %
        {\allowdisplaybreaks%
        \begin{align}
            F^{\star}(x \rightarrow \terminal) &= \sum_{\tau: x\rightarrow \terminal \in \tau}F^{\star}(\tau) = Z\sum_{\tau: s_{0}\rightsquigarrow x}\prod_{t=0}^{T_{\tau}}P_{F}(s_{t+1}\mid s_{t})\\
            &= F_{\terminal}(x)\sum_{\tau: s_{0}\rightsquigarrow x}\prod_{t=1}^{T_{\tau}}P_{B}(s_{t-1}\mid s_{t})\label{eq:proof-trajectory-balance-condition-eq1}\\
            &= F_{\terminal}(x),\label{eq:proof-trajectory-balance-condition-eq2}
        \end{align}}%
        where we used the conventions $s_{T_{\tau}} = x$, and $s_{T_{\tau}+1}=\terminal$, the trajectory balance condition in \cref{eq:proof-trajectory-balance-condition-eq1}, and \cref{lem:PB-distribution-prefix} applied to $x$ in \cref{eq:proof-trajectory-balance-condition-eq2}. This shows that the function $F_{\terminal}$ matches the edge flow of the terminating transitions of $F^{\star}$.

        \item Let $s\rightarrow s'\in\gG$ be an arbitrary transition such that $s'\neq \terminal$. Any complete trajectory going through $s\rightarrow s'$ can be decomposed into (1) a prefix (partial) trajectory $\overleftarrow{\tau}$ from $s_{0}$ to $s$, (2) the transition $s\rightarrow s'$, (3) a suffix (partial) trajectory $\overrightarrow{\tau}$ from $s'$ to some terminating state $x\in\gX$, and finally (4) a transition $x\rightarrow \terminal$. We can write the edge flow $F^{\star}(s\rightarrow s')$ as
        {\allowdisplaybreaks%
        \begin{align}
            F^{\star}(s\rightarrow s') &= \sum_{\tau:s\rightarrow s'\in\tau}F^{\star}(\tau) = Z\sum_{\tau:s\rightarrow s'\in\tau}\prod_{t=0}^{T_{\tau}}P_{F}(s_{t+1}\mid s_{t})\\
            &= \sum_{\tau: s\rightarrow s'\in\tau}F_{\terminal}(s_{T_{\tau}})\prod_{t=1}^{T_{\tau}}P_{B}(s_{t-1}\mid s_{t})\label{eq:proof-trajectory-balance-condition-eq3}\\
            &= \sum_{x\in\gX}F_{\terminal}(x)\Bigg(\sum_{\overleftarrow{\tau}:s_{0}\rightsquigarrow s}P_{B}(\overleftarrow{\tau})\Bigg)P_{B}(s\mid s')\Bigg(\sum_{\overrightarrow{\tau}: s'\rightsquigarrow x}P_{B}(\overrightarrow{\tau})\Bigg)\label{eq:proof-trajectory-balance-condition-eq4}\\
            &= P_{B}(s\mid s')\sum_{x\in\gX}F_{\terminal}(x)\Bigg(\sum_{\overrightarrow{\tau}: s'\rightsquigarrow x}P_{B}(\overrightarrow{\tau})\Bigg),\label{eq:proof-trajectory-balance-condition-eq5}
        \end{align}}%
        where we used the convention $s_{T_{\tau}+1}=\terminal$, the trajectory balance condition in \cref{eq:proof-trajectory-balance-condition-eq3}, the decomposition of complete trajectories going through $s\rightarrow s'$ in \cref{eq:proof-trajectory-balance-condition-eq4}, and \cref{lem:PB-distribution-prefix} in \cref{eq:proof-trajectory-balance-condition-eq5}. Using \cref{prop:identities-state-edge-flows}, we can also write the state flow $F^{\star}(s')$ as
        \begin{equation}
            F^{\star}(s') = \sum_{s''\in\parents_{\gG}(s')}F^{\star}(s''\rightarrow s') = \sum_{x\in\gX}F_{\terminal}(x)\Bigg(\sum_{\overrightarrow{\tau}: s'\rightsquigarrow x}P_{B}(\overrightarrow{\tau})\Bigg),\label{eq:proof-trajectory-balance-condition-eq6}
        \end{equation}
        where we used the fact that $P_{B}$ is a backward transition probability consistent with $\gG$. Combining \cref{eq:proof-trajectory-balance-condition-eq5} and \cref{eq:proof-trajectory-balance-condition-eq6}, we have that $F^{\star}(s\rightarrow s') = P_{B}(s\mid s')F^{\star}(s')$. By definition,
        \begin{equation}
            P_{B}^{\star}(s\mid s') = \frac{F^{\star}(s\rightarrow s')}{F^{\star}(s')} = P_{B}(s\mid s'),
        \end{equation}
        showing that $P_{B}$ matches the backward transition probabilities of $F^{\star}$.

        \item Suppose that there exists another Markovian flow $F'^{\star}$ such that $Z$ matches its total flow, $F_{\terminal}$ matches its edge flow at terminating transitions, $P_{F}$ matches its forward transition probabilities, and $P_{B}$ matches its backward transition probabilities. We denote by $P'^{\star}_{F}$ the forward transition probabilities associated with $F'^{\star}$, and $Z'^{\star}$ its total flow. We first prove that $F'^{\star}(s\rightarrow s') = F^{\star}(s\rightarrow s')$ for all transitions $s\rightarrow s'\in\gG$ by strong induction on the depth of the state $s$ in $\gG$. We will use $d_{s}$ to denote the maximum length of a trajectory from the initial state $s_{0}$ to $s$.

        \emph{Base case:} If $d_{s} = 0$, then it means that $s = s_{0}$ is the initial state. In that case, using \cref{prop:initial-flow-total-flow} and the definition of the forward transition probabilities associated with $F'^{\star}$ and $F^{\star}$
        \begin{align}
            F'^{\star}(s_{0}\rightarrow s') &= F'^{*}(s_{0})P'^{\star}_{F}(s'\mid s_{0}) = Z'^{\star}P'^{\star}_{F}(s'\mid s_{0}) = ZP_{F}(s'\mid s_{0})\\
            &= Z^{\star}P_{F}^{\star}(s'\mid s_{0}) = F^{\star}(s_{0})P^{\star}_{F}(s'\mid s_{0}) = F^{\star}(s_{0}\rightarrow s'),
        \end{align}
        since $Z$ and $P_{F}$ match the total flow and forward transition probabilities of both Markovian flows.

        \emph{Induction step:} If we have $F'^{\star}(s\rightarrow s') = F^{\star}(s\rightarrow s')$ for all transitions such that $d_{s} \leq d$, for some $d > 0$, then let $s \rightarrow s'$ be a transition such that $d_{s} = d+1$. Using \cref{prop:identities-state-edge-flows} (since $s \neq s_{0}$), we then have
        \begin{align}
            F'^{\star}(s\rightarrow s') &= P'^{\star}_{F}(s'\mid s)F'^{\star}(s) = P'^{\star}_{F}(s'\mid s)\sum_{\bar{s}\in\parents_{\gG}(s)}F'^{\star}(\bar{s}\rightarrow s)\\
            &= P_{F}(s'\mid s)\sum_{\bar{s}\in\parents_{\gG}(s)}F^{\star}(\bar{s}\rightarrow s) = P^{\star}_{F}(s'\mid s)F^{\star}(s) = F^{\star}(s\rightarrow s'),
        \end{align}
        where we used the induction step on $F'^{\star}(\bar{s}\rightarrow s) = F^{\star}(\bar{s}\rightarrow s)$ since $d_{\bar{s}} \leq d$, and the fact that $P_{F}$ matches the forward transition probabilities of both Markovian flows.

        This proves that for any transition $s\rightarrow s'\in\gG$, $F'^{\star}(s\rightarrow s') = F^{\star}(s\rightarrow s')$, meaning that $F'^{\star} \sim F^{\star}$. By \cref{prop:equivalent-markovian-flows}, we can conclude that the two Markovian flows are equal, showing the unicity of $F^{\star}$.
    \end{enumerate}
    \vspace*{-1em}
\end{proof}
Similar to how the conditions introduced in the previous sections have analogues in the (irreversible) Markov chain literature, the trajectory balance condition is evocative of the \emph{Kolmogorov criterion} \citep{kelly1979reversibilitystochasticnetworks}. Moreover, we can rewrite \cref{eq:trajectory-balance-condition} for a complete trajectory $\tau = (s_{0}, s_{1}, \ldots, s_{T}, \terminal)$ as
\begin{align}
    P_{F}(\tau) = \frac{F_{\terminal}(s_{T})}{Z}P_{B}(\tau \mid s_{T}),
    \label{eq:trajectory-balance-condition-compact}
\end{align}
where we used the notations $P_{F}(\tau)$ and $P_{B}(\tau \mid s_{T})$ we introduced in \cref{sec:forward-transition-probabilities,sec:backward-transition-probabilities}. If we interpret the factor $F_{\terminal}(s_{T}) / Z$ as a distribution over $\gX$ (which is true when we have a trajectory flow, see \cref{prop:terminating-state-probability-from-flow}), then the RHS of this equation can be seen as a distribution over all the complete trajectories in $\gT$ (recall that $P_{B}(\tau\mid s_{T})$ is a distribution over the trajectories terminating at $s_{T}$, by \cref{lem:PB-distribution-prefix}). Similarly, the LHS of \cref{eq:trajectory-balance-condition-compact} is also a distribution over all the complete trajectories (\cref{lem:PF-distribution-suffix}). Therefore, another way to interpret the trajectory balance condition is as an equality between two distributions over complete trajectories, one as a forward process and the other as a backward process.

\subsection{Sub-trajectory balance condition}
\label{sec:sub-trajectory-balance-condition}
As an alternative in between the fully local detailed balance condition at the level of transitions and the more global trajectory balance condition at the level of complete trajectories, \citet{malkin2022trajectorybalance} also introduced a similar criterion operating at the level of partial trajectories in $\gG$. This is called the \emph{sub-trajectory balance condition} \citep{madan2022subtb}, and uses a similar parametrization as the detailed balance condition in terms of state flow and transition probabilities.
\begin{theorem}[Sub-trajectory balance condition]
    \label{thm:sub-trajectory-balance-condition}\index{Sub-trajectory balance!Condition|textbf}\glsadd{subtb}
    If $(\gG, F^{\star})$ be a Markovian flow with $\gG = (\widebar{\gS}, \gA)$, then its state flow $F^{\star}: \gS \rightarrow \sR_{+}$, its forward transition probabilities $P^{\star}_{F}: \gS\rightarrow \Delta(\children_{\gG})$, and its backward transition probabilities $P^{\star}_{B}: \gS \rightarrow \Delta(\parents_{\gG})$ satisfy the following \emph{sub-trajectory balance condition} for all partial trajectories $\tau = (s_{m}, s_{m+1}, \ldots, s_{n})$ (for some $n > m$) such that $s_{n}\neq \terminal$:
    \begin{equation}
        F^{\star}(s_{m})\prod_{t=m}^{n-1}P^{\star}_{F}(s_{t+1}\mid s_{t}) = F^{\star}(s_{n})\prod_{t=m}^{n-1}P^{\star}_{B}(s_{t}\mid s_{t+1}).
        \label{eq:sub-trajectory-balance-condition}
    \end{equation}
\end{theorem}

\begin{figure}[t]
    \begin{adjustbox}{center}
        \includegraphics[width=480pt]{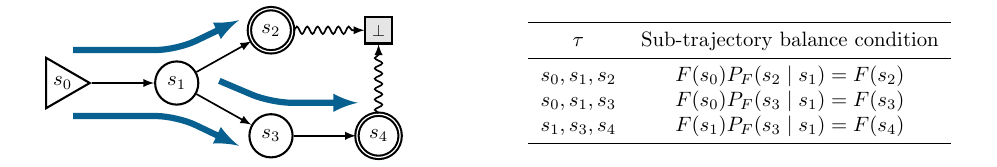}
    \end{adjustbox}
    \caption[Counter-example of the sufficiency of the sub-trajectory balance condition.]{Counter-example of the sufficiency of the sub-trajectory balance condition. We consider all the partial trajectories of length 2 in the pointed DAG (left), and write the corresponding sub-trajectory balance condition for each trajectory (right; simplified). This system of equations admits an infinite amount of solutions, among which many do not correspond to the state flow and transition probabilities of a valid flow (\eg \cref{eq:subtb-counter-example-solution}).}
    \label{fig:subtb-condition-counter-example}
\end{figure}

\begin{proof}
    By \cref{thm:detailed-balance-condition}, since $F^{\star}$ is a Markovian flow, we know that for any transition $s_{t}\rightarrow s_{t+1}\in\gG$, $F^{\star}$, $P^{\star}_{F}$, and $P^{\star}_{B}$ satisfy the detailed balance condition
    \begin{equation}
        F^{\star}(s_{t})P^{\star}_{F}(s_{t+1}\mid s_{t}) = F^{\star}(s_{t+1})P^{\star}_{B}(s_{t}\mid s_{t+1}).
    \end{equation}
    Applying the equality above for all the transitions in $\tau = (s_{m}, s_{m+1}, \ldots, s_{n})$ and taking their product yields the expected result (factoring out the common term $F^{\star}(s_{m+1})\times \ldots \times F^{\star}(s_{n-1})$).
\end{proof}
Unlike the results in the previous sections though, the sub-trajectory balance condition is only a necessary condition for a Markovian flow, but it is not sufficient. If the sub-trajectory balance condition is satisfied for all partial trajectories \emph{of fixed length}, then this may not define a valid Markovian flow. To see this, consider the counter-example in \cref{fig:subtb-condition-counter-example}. If we look at all the partial trajectories of length 2, we can write down the sub-trajectory balance condition \cref{eq:sub-trajectory-balance-condition} for each of them while simplifying the known terms (\eg $P_{F}(s_{1}\mid s_{0}) = 1$ because $s_{1}$ is the unique child of $s_{0}$, or $P_{B}(s_{3}\mid s_{4}) = 1$ because $s_{3}$ is the unique parent of $s_{4}$); for example for $\tau = (s_{0}, s_{1}, s_{2})$:
\begin{equation}
    F(s_{0})\underbrace{P_{F}(s_{1}\mid s_{0})}_{=\,1}P_{F}(s_{2}\mid s_{1}) = F(s_{2})\underbrace{P_{B}(s_{1}\mid s_{2})}_{=\,1}\underbrace{P_{B}(s_{0}\mid s_{1})}_{=\,1}
\end{equation}
The (simplified) equations for the sub-trajectory balance condition of all the partial trajectories of length 2 are given in \cref{fig:subtb-condition-counter-example} (right). This gives us a system of 3 equations with 7 unknowns and 1 constraint (for $P_{F}(\cdot \mid s_{1})$). This system is underdetermined and admits an infinite number of solutions; in particular, the following is a solution of this system:
\begin{align}
    P_{F}(s_{2}\mid s_{1}) &= P_{F}(s_{3}\mid s_{1}) = 1/2 & F(s_{0}) &= 4 & F(s_{1}) &= 2\label{eq:subtb-counter-example-solution}\\
    F(s_{2}) &= 2 & F(s_{3}) &= 2 & F(s_{4}) &= 1.
\end{align}
This solution is only one of many we could find. But if we assumed that $F$ defines the state flow and $P_{F}$ the forward transition probabilities of a valid Markovian flow, then the total flow would necessarily be $Z = F(s_{0}) = 4$ (by \cref{prop:initial-flow-total-flow}). However, this contradicts the conservation of the total flow throughout the network, since the total amount of flow going into $\terminal$ (which is supposed to be equal to the total flow) is $F(s_{2}) + F(s_{4}) = 3 \neq F(s_{0})$. Therefore, $F$ and $P_{F}$ cannot define the state flow and forward transition probabilities of a valid (Markovian) flow.

The sub-trajectory balance condition is not sufficient to define a valid flow: satisfying \cref{eq:sub-trajectory-balance-condition} for all partial trajectories of fixed length does \emph{not} yield a valid Markovian flow. This may appear as a surprise since the detailed balance condition of \cref{thm:detailed-balance-condition} is effectively a special case of the sub-trajectory balance condition applied to all partial trajectories of fixed length $1$ (\ie transitions), but it \emph{is} sufficient. Intuitively, what \cref{fig:subtb-condition-counter-example} shows is that the sub-trajectory balance condition may ``jump over'' a certain state. The value of $F(s_{1})$ is never grounded by $F(s_{0})$: only the third equation in the system involves $F(s_{1})$, which is independent of $F(s_{0})$, and the other partial trajectories of length $2$ starting at $s_{0}$ all ``jump over'' $s_{1}$. This is not a situation that we would encounter with detailed balance, as there would be an equation for the transition $s_{0} \rightarrow s_{1}$.

However, if the sub-trajectory balance condition is satisfied for all partial trajectories of \emph{any length} $\leq L$ (and not of fixed length anymore), then it becomes a sufficient condition for a Markovian flow. This is somewhat trivial because if it is satisfied for any length $\leq L$, it is in particular true for partial trajectories of length $1$ (\ie transitions), in which case it reduces to the detailed balance condition of \cref{thm:detailed-balance-condition}. Looser conditions are also possible depending on the structure of $\gG$; we will see an example in \cref{sec:jsp-gfn-sub-trajectory-balance-condition} where the sub-trajectory balance condition for some (but not all) partial trajectories of fixed length still defines a valid flow.

We can observe that the sub-trajectory balance condition unifies the detailed balance condition of \cref{thm:detailed-balance-condition} (applying the sub-trajectory balance condition to partial trajectories of length 1, \ie transitions), as well as the trajectory balance condition of \cref{thm:trajectory-balance-condition} (applying the condition to complete trajectories). That's why even if this is not a sufficient characterization of a Markovian flow, it will sometimes be useful to prove statements based on the sub-trajectory balance condition, such as in \cref{sec:equivalence-pcl-subtb}, and infer results for the detailed balance and trajectory balance conditions as byproducts. \citet{malkin2022trajectorybalance} further generalized this sub-trajectory balance condition to undirected paths that go ``backward then forward'' between two states $s_{m}$ and $s_{n}$:
\begin{equation}
    F^{\star}(s_{m})\prod_{t=k}^{m-1}P^{\star}_{B}(s_{t}\mid s_{t+1})\prod_{t=k}^{n-1}P^{\star}_{F}(s_{t+1}\mid s_{t}) = F^{\star}(s_{n})\prod_{t=k}^{n-1}P^{\star}_{B}(s_{t}\mid s_{t+1})\prod_{t=k}^{m-1}P^{\star}_{F}(s_{t+1}\mid s_{t}),
    \label{eq:generalized-sub-trajectory-balance-condition}
\end{equation}
where $s_{k}$ is a common ancestor of both $s_{m}$ and $s_{n}$; see \cref{fig:generalized-sub-trajectory-balance-condition} for an illustration of this generalized condition. This general form has been frequently used on undirected trajectories between 2 terminating states, similar to MCMC \citep{zhang2022ebgfn,deleu2023jspgfn,kim2024lsgfn}.  %

\section{Conditional flow networks}
\label{sec:conditional-flow-networks}

In \cref{sec:flow-networks}, we defined a flow network as a pointed DAG, augmented with some function $F^{\star}$ over the set of complete trajectories $\gT$. We can extend this notion by \emph{conditioning} each of its components on some additional information $y \in \gY$. In general, this conditioning variable can represent any information, either external to the network structure (but still influencing the flows), or internal (\eg $y$ can be a property of complete trajectories over another flow network, such as passing through a particular state). We will define a \emph{conditional flow network} as a collection of flow networks, indexed by $y\in\gY$.

\begin{definition}[Conditional flows]
    \label{def:conditional-flows}\index{Flow!Conditional flow}\index{Conditional flow|see {Flow}}
    Let $\gY$ be a set of conditioning values. We consider a family of pointed DAGs $\gG_{y} = (\widebar{\gS}_{y}, \gA_{y})$ indexed by $y\in\gY$, with their corresponding initial states denoted by $(s_{0}\mid y)\in\gS_{y}$. A \emph{conditional flow} is a family of trajectory flows $F^{\star}_{y} \in \widebar{\gF}(\gG_{y})$ defined on each $\gG_{y}$. The family $\{(\gG_{y}, F^{\star}_{y})\}_{y\in\gY}$ is called a \emph{conditional flow network}.
\end{definition}

We will assume that all the pointed DAGs have a common terminal state $\terminal$ that belongs to each $\widebar{\gS}_{y}$ (but similar to unconditional flow networks, is not an element of $\gS_{y}$). Since conditional flow networks are defined using the same components as their unconditional counterparts, they inherit all the properties from \cref{sec:flow-networks-markovian-flows}. In particular, we can directly extend any of the flow matching conditions in \cref{sec:flow-matching-conditions} by conditioning each component on $y\in\gY$. In the following sections, we will elaborate two important examples: flow networks conditioned on external information that modifies the flows $F^{\star}_{y}$, and state-conditional flow networks that depend on previously visited states.

\subsection{Parametric flow networks}
\label{sec:parametric-flow-networks}
While \cref{def:conditional-flows} is fully general and allows us to change the structure of the pointed DAG $\gG_{y}$ (along with its initial state) with $y\in\gY$, in many interesting cases we will have a common structure for the pointed DAG $\gG$, and only the flows are \emph{parametrized} by some external variable $y$.
\begin{definition}[Parametric flow network]
    \label{def:parametric-flow-network}\index{Flow network!Parametric}
    Let $\gG$ be a pointed DAG, and $\gY$ a set of conditioning values. A family $\{(\gG, F^{\star}_{y})\}_{y\in\gY}$ is called a \emph{parametric flow network} if for all $y\in\gY$, $F^{\star}_{y}\in\widebar{\gF}(\gG)$ is a trajectory flow.
\end{definition}
In this context, the conditioning values in $\gY$ will be called ``parameters''; however, it is important to note that these do not correspond to the parameters of a function approximating the flow (\eg parameters of a neural network, see \cref{chap:generative-flow-networks}), because for any parameter $y$, $F^{\star}_{y}$ is a proper trajectory flow and not an approximation. We will see in \cref{sec:vbg-variational-bayes-approach} an application of parametric flow networks.

\subsection{State-conditional flow networks}
\label{sec:state-conditional-flow-networks}
\begin{figure}[t]
    \centering
    \begin{adjustbox}{center}
        \includegraphics[width=480pt]{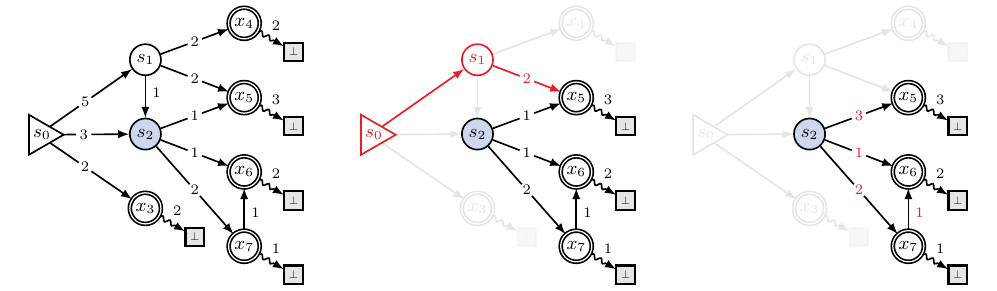}%
    \end{adjustbox}
    \begin{adjustbox}{center}%
        \begin{subfigure}[b]{160pt}%
            \caption{}%
            \label{fig:state-conditional-flow-1}%
        \end{subfigure}%
        \begin{subfigure}[b]{160pt}%
            \caption{}%
            \label{fig:state-conditional-flow-2}%
        \end{subfigure}%
        \begin{subfigure}[b]{160pt}%
            \caption{}%
            \label{fig:state-conditional-flow-3}%
        \end{subfigure}%
    \end{adjustbox}
    \caption[Example of a state-conditional flow network]{Example of a state-conditional flow network. (a) The original Markovian flow $F^{\star}$, with the edge flows indicated at each transition. (b) The subgraph of states reachable from $s_{2}$; there is a flow through $(s_{0}, s_{1}, s_{5}, \terminal)$ that contributes to $F^{\star}(s_{5}\rightarrow \terminal)$, but is not accounted for in $F^{\star}(s_{2})$. This shows that $F^{\star}(s_{2})$ does not marginalize the flows out of its descendants into $\terminal$. (c) The state-conditional flow $F_{s_{2}}^{\star}$, which differs from the original flow $F^{\star}$ on the subgraph, but satisfies the desired marginalization property. Note that the flows at the terminating edges are left unchanged.}
    \label{fig:state-conditional-flow}
\end{figure}

For an (unconditional) flow network $(\gG, F^{\star})$, where $\gG = (\widebar{\gS}, \gA)$, we saw in \cref{prop:initial-flow-total-flow} that the state flow at the initial state $s_{0}$ corresponds to the total amount of flow going through $\gG$. By \cref{prop:terminating-state-probability-from-flow}, this means that
\begin{equation}
    F^{\star}(s_{0}) = Z^{\star} = \sum_{x\in\gX}F^{\star}(x\rightarrow \terminal).
    \label{eq:conditional-flow-marginalization-simple}
\end{equation}
This equation is a simple form of ``marginalization'', where we sum over all the terminating states accessible from $s_{0}$ (all of them). All of the flow coming out of the initial state $s_{0}$ eventually ends up at one of the terminating states accessible from $s_{0}$. A natural question then is: does this marginalization also hold for $F^{\star}(s)$ the flow through any intermediate state $s\in\gS$? More precisely, we would like to know if $F^{\star}(s)$ corresponds to summing the edge flows at the terminating transitions, for all the terminating states that are accessible from $s$:
\begin{equation}
    F^{\star}(s) \overset{?}{=} \sum_{x: s\leq x} F^{\star}(x\rightarrow \terminal).
\end{equation}
Unfortunately, in general, the answer to this question is \emph{no} as illustrated in \cref{fig:state-conditional-flow}. In this example, we have $F^{\star}(s_{2}) = 4$, whereas the sum of edge flows at the terminating transitions accessible from $s_{2}$ is $6$ (the terminating states accessible from $s_{2}$ being $\{x_{5}, x_{6}, x_{7}\}$). This discrepancy is caused by the flow through $(s_{0}, s_{1}, x_{5}, \terminal)$ that contributes to $F^{\star}(x_{5} \rightarrow \terminal)$, but is not accounted for in $F^{\star}(s_{2})$ since this trajectory does not go through $s_{2}$. To address this, we can introduce a family of flow networks called a \emph{state-conditional flow network} indexed by the states of $\gG$, where each conditional flow satisfies \cref{eq:conditional-flow-marginalization-simple} for different subgraphs.

\begin{definition}[State-conditional flow network]
    \label{def:state-conditional-flow-network}\index{Flow network!State-conditional}
    Let $(\gG, F^{\star})$ be a flow network, where $\gG = (\widebar{\gS}, \gA)$, and let $\gX$ be the set of terminating states of $\gG$. For each state $s\in\gS$, let $\gG_{s}$ be the subgraph of $\gG$, rooted in $s$, containing all the states $s'$ accessible from $s$ in $\gG$ ($s' \geq s$). A family $\{(\gG_{s}, F^{\star}_{s})\}_{s\in\gS}$ is called a \emph{state-conditional flow network} if the terminating edge flows of $F^{\star}_{s}\in\widebar{\gF}(\gG_{s})$ at the terminating transitions match those of $F^{\star}$: for all $x\in\gX$,
    \begin{equation}
        F_{s}^{\star}(x\rightarrow \terminal) = F^{\star}(x\rightarrow \terminal).
        \label{eq:state-conditional-flow-network}
    \end{equation}
\end{definition}
Unlike parametric flow networks though, the structure of the pointed DAG in this conditional flow network depends on the anchor state $s$, as well as its initial states $(s_{0}\mid s) = s$. In particular, it is clear that $F^{\star}_{s_{0}}(\tau) = F^{\star}(\tau)$ always satisfies the conditions above. Since the definition of a state-conditional flow network depends on an original flow network, we must ensure that this definition is indeed valid, \ie that such a state-conditional flow network satisfying \cref{eq:state-conditional-flow-network} exists.

\begin{proposition}[Existence of state-conditional flows]
    \label{prop:existence-state-conditional-flow-network}
    For any flow network $(\gG, F^{\star})$, there exists a state-conditional flow network satisfying \cref{def:state-conditional-flow-network}.
\end{proposition}

\begin{proof}
    Let $s\in\gS$ be a fixed state. Since the structure of the pointed DAG $\gG_{s}$ rooted at $s$ is clearly well-defined, we just need to show that there exists a trajectory flow $F^{\star}_{s}\in\widebar{\gF}(\gG_{s})$ that satisfies \cref{eq:state-conditional-flow-network}. Let $x\in\gX$ be a terminating state accessible from $s \leq x$, and let $\gT_{s}$ be the set of complete trajectories in $\gG_{s}$; it is clear that $\gT_{s} \equiv s\rightsquigarrow \terminal$ corresponds to the partial trajectories in $\gG$, starting at $s$ and ending at $\terminal$. Finally, let $A_{x\mid s} = \{\tau\in\gT_{s}\mid x\rightarrow \terminal\in\tau\}$ be the set of complete trajectories (in $\gG_{s}$) terminating in $x$. The condition in \cref{eq:state-conditional-flow-network} then reads
    \begin{equation}
        F^{\star}(x\rightarrow \terminal) = F^{\star}_{s}(x\rightarrow \terminal) = F^{\star}_{s}(A_{x\mid s}) = \sum_{\tau\in A_{x\mid s}}F^{\star}_{s}(\tau).
        \label{eq:proof-existence-state-conditional-flow-eq1}
    \end{equation}
    Note that in \cref{eq:proof-existence-state-conditional-flow-eq1}, $F^{\star}(x\rightarrow \terminal)$ is a given quantity because the flow $F^{\star}$ is known. Since $\gT_{s} = \bigcup_{x:s\leq x}A_{x\mid s}$ forms a partition of all the complete trajectories $\gT_{s}$, \cref{eq:proof-existence-state-conditional-flow-eq1} is a system of linear equations, with as many equations as there are terminating states accessible from $s$, and whose unknowns are $F^{\star}_{s}(\tau)$ for all $\tau\in\gT_{s}$, where each equation involves separate sets of unknowns. Therefore, there exists at least a solution $F^{\star}_{s}(\tau)$ of this linear system.

    We can construct such a solution explicitly in the following way. For some $\tau\in\gT_{s}$, we can start by selecting the complete trajectories $\bar{\tau}\in\gT$ in the original pointed DAG that contain $\tau$: $C_{\tau} \triangleq \{\bar{\tau}\in\gT\mid \tau \subseteq \bar{\tau}\}$. The key difference between the pointed DAG $\gG$ and the subgraph $\gG_{s}$ though is that $\gG$ may contain trajectories that terminate in some $x\geq s$, but do not pass though $s$, and those are therefore not covered by the trajectories of $\gG_{s}$ (see \cref{fig:state-conditional-flow-2} for an illustration). Let $U_{x\mid s}$ be the set of complete trajectories of $\gG$ handling this case
    \begin{equation}
        U_{x\mid s} \triangleq \{\bar{\tau}\in\gT\mid x\in\bar{\tau}, s\notin\bar{\tau}\}.
    \end{equation}
    For all $\tau\in\gT_{s}$ such that $\tau$ terminates in some $x\geq s$, we can therefore construct the trajectory flow $F^{\star}_{s}(\tau)$ as
    \begin{equation}
        F^{\star}_{s}(\tau) \triangleq F^{\star}(C_{\tau}) + \frac{1}{|A_{x\mid s}|}F^{\star}(U_{x\mid s}),
        \label{eq:proof-existence-state-conditional-flow-eq2}
    \end{equation}
    where all the quantities are well defined, since the sets $C_{\tau}$ and $U_{x\mid s}$ are subsets of $\gT$, and therefore have valid flows $F^{\star}$. It is easy to show that $F^{\star}_{s}(\tau)$ defined in \cref{eq:proof-existence-state-conditional-flow-eq2} is a solution of \cref{eq:proof-existence-state-conditional-flow-eq1}.
\end{proof}
The construction above shows that the state-conditional flow network associated with some flow network $(\gG, F^{\star})$ is not unique, since we could have redistributed the flow $F^{\star}(U_{x\mid s})$ going to $x$ but not through $s$ in any way among the trajectories in $A_{x\mid s}$. Going back to the motivation of interpreting state flows as marginalizing the edge flows at the terminating transitions, we can now leverage the marginalization property at the initial state of each $(\gG_{s}, F^{\star}_{s})$.

\begin{proposition}[Marginalization]
    \label{prop:state-conditional-initial-flow-marginalization}
    Let $(\gG, F^{\star})$ be a flow network, with $\gG = (\widebar{\gS}, \gA)$, and let $\gX$ be the set of terminating states of $\gG$. Let $\{(\gG_{s}, F^{\star}_{s})\}_{s\in\gS}$ be a corresponding state-conditional flow network. The state flow at the initial state satisfies the following marginalization property
    \begin{equation}
        F_{s}^{\star}(s_{0}\mid s) = F_{s}^{\star}(s) = \sum_{x: s\leq x}F^{\star}(x\rightarrow \terminal).
        \label{eq:state-conditional-initial-flow-marginalization}
    \end{equation}
\end{proposition}

\begin{proof}
    This is a direct consequence of \cref{prop:initial-flow-total-flow} and \cref{prop:terminating-state-probability-from-flow},
    \begin{equation}
        F^{\star}_{s}(s_{0}\mid s) = \sum_{\tau\in\gT_{s}}F^{\star}_{s}(\tau) = \sum_{x: s \leq x}F^{\star}_{s}(x\rightarrow \terminal) = \sum_{x: s\leq x}F^{\star}(x\rightarrow \terminal),
    \end{equation}
    where we used \cref{eq:state-conditional-flow-network}, and $\gT_{s}$ is the set of complete trajectories in $\gG_{s}$.
\end{proof}

This suggests a way to estimate the marginal \cref{eq:state-conditional-initial-flow-marginalization} over the terminating states accessible from $s$: for each $s\in\gS$, find a flow $F^{\star}_{s}$ using one of the flow matching conditions in \cref{sec:flow-matching-conditions}, under the constraint \cref{eq:state-conditional-flow-network}. For example, if we were to apply the \emph{flow matching condition} of \cref{thm:flow-matching-condition}, this amounts to finding a function $F: \gS \times \gA \rightarrow \sR_{+}$ that satisfies for all $\bar{s}\in\gS$ (anchor state) and for all $s'>\bar{s}$:
\begin{equation}
    \sum_{s\in\parents_{\gG}(s')}F(\bar{s}, s\rightarrow s') = \sum_{s''\in\children_{\gG}(s')}F(\bar{s}, s'\rightarrow s''),
    \label{eq:flow-matching-condition-state-conditional}
\end{equation}
under the constraint $F(\bar{s}, x\rightarrow \terminal) = F^{\star}(x\rightarrow \terminal)$ for any terminating state $x\geq \bar{s}$ accessible from $\bar{s}$. This kind of \emph{boundary condition} on the edge flow of the terminating transitions will play a central role in the next chapter, where flow networks will be used to model distributions over $\gX$. %

Finally, flow networks may also benefit from other marginalization properties, depending on the structure of the pointed DAG itself. For example, if we consider the pointed DAG of \cref{fig:treesample-intro} (inference over discrete factor graphs), then stopping the generation after one step effectively corresponds to marginalizing over the values of $X_{2}$ \& $X_{3}$. We will see another example of this in \cref{sec:jsp-gfn-parametrization-forward-trasition-probabilities}. This is all the more powerful when the variables of a joint distribution represented by the flow network can be assigned in different orders \citep[][Section 5.3]{bengio2023gflownetfoundations}.

%% file: chapters/04_Generative_Flow_Networks.tex
\chapter{Generative Flow Networks}
\label{chap:generative-flow-networks}

\begin{minipage}{\textwidth}
    \itshape
    This chapter contains material from the following papers:
    \begin{itemize}[noitemsep, topsep=1ex, itemsep=1ex, leftmargin=3em]
        \item Yoshua Bengio$^{*}$, Salem Lahlou$^{*}$, \textbf{Tristan Deleu}$^{*}$, Edward Hu, Mo Tiwari, Emmanuel Bengio (2023). \emph{GFlowNet Foundations}. Journal of Machine Learning Research (JMLR). \notecite{bengio2023gflownetfoundations}
        \item Nikolay Malkin$^{*}$, Salem Lahlou$^{*}$, \textbf{Tristan Deleu}$^{*}$, Xu Ji, Edward Hu, Katie Everett, Dinghuai Zhang, Yoshua Bengio (2023). \emph{GFlowNets and Variational Inference}. International Conference on Learning Representations (ICLR). \notecite{malkin2023gfnhvi}
    \end{itemize}
    \vspace*{5em}
\end{minipage}

In the previous chapter, we saw how flow networks were defined and some of their properties. In particular, we saw in \cref{sec:probability-distribution-induced-flow} that a flow network $(\gG, F^{\star})$ induces a distribution over terminating states by following its corresponding forward transition probability distribution $P_{F}^{\star}$. Going back to our original problem introduced in \cref{chap:probabilistic-inference-structured-objects}, we are more concerned with the ``inverse problem'': if we know the distribution (not the flow) up to an intractable normalization constant, can we sample from it by following the forward transition probability of a certain flow network? In this chapter, we will see how we can find a flow $F^{\star}$ whose corresponding terminating state distribution $P_{F}^{\star\top}$ matches the Gibbs distribution in \cref{eq:gibbs-distribution}. Unlike \cref{sec:probabilistic-inference-control-problem} though, where we viewed this problem as a control problem over a very specific MDP, this strategy will be valid for any arbitrary pointed DAG $\gG$ and not limited to tree structures.

\section{Discrete probabilistic modeling with flow networks}
\label{sec:probabilistic-modeling-flow-networks}
Recall that we are interested in the broad problem of sampling from a target probability distribution $P^{\star}$ defined over a discrete (and often compositional) sample space $\gX$, known only up to a normalization constant. This distribution may be defined based on some \emph{reward function} $R: \gX \rightarrow \sR_{+}$, such that for any $x\in\gX$
\begin{equation}
    P^{\star}(x) = \frac{R(x)}{Z},
    \label{eq:target-distribution-gflownet-reward}
\end{equation}
where $Z = \sum_{x\in\gX}R(x)$ is again the partition function. This is a general formulation for distributions defined up to some normalization constant; \eg in the case of a Gibbs distribution \cref{eq:gibbs-distribution}, we have $R(x) = \exp(-\gE(x))$. We use the term ``reward'' to denote the function $R$ by analogy with reinforcement learning \citep{bengio2021gflownet}. Intuitively, the reward function gives a notion of ``preference'' for some elements over others: objects with higher rewards are sampled more frequently than those with lower reward. We will come back to the connections existing between flow networks and reinforcement learning in the next chapter.

The idea behind using flow networks to represent the distribution in \cref{eq:target-distribution-gflownet-reward} is to construct a flow network $(\gG, F^{\star})$ such that (1) the sample space $\gX$ of $P^{\star}$ is the set of terminating states of $\gG$, and (2) the terminating state distribution $P_{F}^{\star\top}$ associated with the flow $F^{\star}$ matches $P^{\star}$. This would provide a natural way to sample from $P^{\star}$, despite the unknown normalization constant $Z$, via \cref{alg:sampling-terminating-state-probability}: start from the initial state $s_{0}$, sample consecutive states using $P_{F}^{\star}$ until we reach the terminal state $\terminal$, and return the last state visited right before terminating. In \cref{prop:terminating-state-probability-from-flow}, we saw that the terminating state distribution associated with a flow is proportional to its edge flows at the terminating transitions $F^{\star}(x\rightarrow \terminal)$. Thus in order to match \cref{eq:target-distribution-gflownet-reward}, this motivates some additional constraints the flow $F^{\star}$ must satisfy, beyond the flow matching conditions alone: a function $F: \gA \rightarrow \sR_{+}$ is said to satisfy the \emph{boundary conditions}\index{Boundary conditions} with the reward function \gls{reward} if for any $x\in\gX$, we have
    \begin{equation}
        F(x\rightarrow \terminal) = R(x).
        \label{eq:boundary-conditions}
    \end{equation}

In the context of differential equations, this type of boundary conditions is evocative of \emph{Neumann boundary conditions} \citep{cheng2005heritageboundary}, where the normal derivative of the solution of a partial differential equation must match a predefined function at the boundary of its domain; here, $F(x\rightarrow \terminal)$ plays the role of the normal derivative at the boundary $\gX$ of the state space (``going out'' of $\gX$). Using a flow network as a generative model is called a \emph{generative flow network}\index{Generative flow network|textbf}\index{GFlowNet|see {Generative flow network}} (\gls{gflownet}, or sometimes \gls{gfn}; \citealp{bengio2021gflownet,bengio2023gflownetfoundations}).

\begin{figure}[t]
    \centering
    \begin{adjustbox}{center}
    \includegraphics[width=520pt]{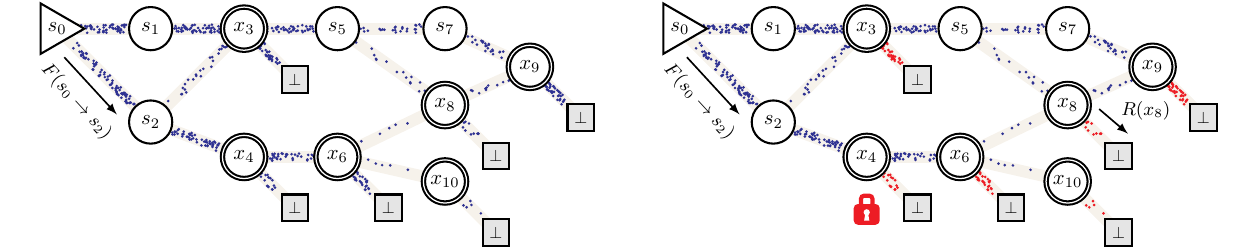}
    \end{adjustbox}
    \begin{adjustbox}{center}%
        \begin{subfigure}[b]{260pt}%
            \caption{Flow network}%
            \label{fig:gflownet-flow-comparison-1}%
        \end{subfigure}%
        \begin{subfigure}[b]{260pt}%
            \caption{Generative flow network}%
            \label{fig:gflownet-flow-comparison-2}%
        \end{subfigure}%
    \end{adjustbox}
    \caption[Difference between a flow network and a generative flow network]{Difference between a flow network and a generative flow network. (a) A flow network has a flow which is completely unrestricted, which only needs to satisfy some flow matching conditions. (b) A generative flow network (GFlowNet) is a flow network satisfying some boundary conditions \cref{eq:boundary-conditions}, where the flow at terminating edges must equal the reward at the terminating state.}
    \label{fig:gflownet-flow-comparison}
\end{figure}

\subsection{Flow matching condition under boundary constraints}
\label{sec:flow-matching-condition-boundary-constraint}
In \cref{sec:flow-matching-conditions}, we presented multiple ways to characterize \emph{an arbitrary} Markovian flow $F^{\star}$ by following some conservation law. Based on our discussion above, the only ingredient necessary to eventually sample from \cref{eq:target-distribution-gflownet-reward} is to find not just any flow, but a flow that also satisfies the boundary condition \cref{eq:boundary-conditions}. The following theorem combines together the flow matching condition of \cref{thm:flow-matching-condition} with the boundary condition in order to guarantee that the corresponding terminating state distribution is proportional to the reward function. This is largely considered as \emph{the} fundamental theorem of generative flow networks \citep{bengio2021gflownet}.%
\begin{theorem}[Flow matching condition]
    \label{thm:flow-matching-proportional-reward}\index{Flow matching!Condition}
    Let $\gG = (\widebar{\gS}, \gA)$ be a pointed DAG, where $\gX\subseteq \gS$ is the set of terminating states of $\gG$, and let $R: \gX \rightarrow \sR_{+}$ be a reward function defined over the terminating states. If a function $F: \gA \rightarrow \sR_{+}$ satisfies the \emph{flow matching condition} for all $s'\in\gS$ such that $s'\neq s_{0}$
    \begin{equation}
        \sum_{s\in\parents_{\gG}(s')}F(s\rightarrow s') = \sum_{s''\in\children_{\gG}(s')}F(s'\rightarrow s''),
        \label{eq:flow-matching-proportional-reward}
    \end{equation}
    and satisfies the boundary condition $F(x\rightarrow \terminal) = R(x)$ for all $x\in\gX$, then the terminating state probability distribution associated with the forward transition probability $P_{F}$ defined, for $s\rightarrow s'\in\gG$, by
    \begin{equation}
        P_{F}(s'\mid s) = \frac{F(s\rightarrow s')}{\sum_{s''\in\children_{\gG}(s)}F(s\rightarrow s'')}
        \label{eq:forward-transition-probability-flow-matching}
    \end{equation}
    is proportional to the reward: $P_{F}^{\top}(x) \propto R(x)$.
\end{theorem}

\begin{proof}
    This is a direct consequence of \cref{thm:flow-matching-condition} and \cref{prop:terminating-state-probability-from-flow}. Since $F$ satisfies the flow matching condition, then by \cref{thm:flow-matching-condition} there exists a unique Markovian flow $F^{\star}$ whose edge flow matches $F$. Moreover, this also shows that $P_{F}$ defined in \cref{eq:forward-transition-probability-flow-matching} matches the forward transition probability of $F^{\star}$ (by definition of $P_{F}^{\star}$). By \cref{prop:terminating-state-probability-from-flow}, the terminating state probability distribution associated with $P_{F}^{\star} = P_{F}$ is proportional to $F^{\star}(x\rightarrow \terminal) = F(x\rightarrow \terminal)$. Using the boundary condition $F(x\rightarrow \terminal) = R(x)$, we can conclude that $P_{F}^{\top}(x) \propto R(x)$.
\end{proof}

As an alternative way to write the flow matching condition together with the boundary conditions, \citet{bengio2021gflownet} in their original formulation incorporated the reward directly into \cref{eq:flow-matching-proportional-reward} as
\begin{equation}
    \sum_{s\in\parents_{\gG}(s')}F(s\rightarrow s') - \sum_{s''\in\underline{\children}_{\gG}(s')}F(s'\rightarrow s'') = R(s'),
    \label{eq:flow-matching-boundary-condition}
\end{equation}
using the convention $R(s') = 0$ for any intermediate state $s'\in\gS\backslash\gX$, and where $\underline{\children}_{\gG}(s') = \children_{\gG}(s') \backslash \{\terminal\}$ are all the children of $s' \neq s_{0}$ except the terminal state. Putting \cref{eq:flow-matching-boundary-condition} in words, this means that the total amount of flow going into any state $s'$ has to be equal to the total amount of flow going out, plus some residual $R(s')$.

\paragraph{Solution of a linear system} When the reward function $R$ is known, one way to interpret the constraint \cref{eq:flow-matching-boundary-condition} that the (non-negative) function $F$ must satisfy is as a system of $p \triangleq |\gS| - 1$ linear equations (one for each state $s'\neq s_{0}$). This system has the form $\mA F = \vb$, where $F$ can be viewed as a vector of $|\gA|$ unknowns (one for each edge of $\gG$), $\mA$ is a sparse matrix with entries in $\{-1, 0, +1\}$, and $\vb \in\sR^{p}$ is a vector that depends on $R$, with exactly $|\gX|$ non-zero entries. Treating $F$ as a vector in $\sR^{|\gA|}$ is called a \emph{tabular representation} of the flow, once again by analogy with the reinforcement learning literature \citep{sutton2018introrl}. In addition to this linear constraint, the vector $F \geq 0$ is also required to be non-negative. Without any assumption on the structure of the problem though (the sparsity pattern of $\mA$), there is no guarantee that this underdetermined system admits a solution at all, let alone a non-negative one. Fortunately, since $\gG$ is a pointed DAG, the following proposition shows the existence of a solution to this problem, regardless of the reward function $R$ (provided it is non-negative).

\begin{proposition}[Existence of a flow]
    \label{prop:existence-flow-flow-matching}
    Let $\gG = (\widebar{\gS}, \gA)$ be a pointed DAG, where $\gX \subseteq \gS$ is the set of terminating states of $\gG$, and let $R: \gX \rightarrow \sR_{+}$ be a reward function defined over the terminating states. There exists a function $F: \gA \rightarrow \sR_{+}$ satisfying the flow matching condition and boundary conditions in \cref{thm:flow-matching-proportional-reward} (or, alternatively \cref{eq:flow-matching-boundary-condition}).
\end{proposition}

\begin{proof}
    We can prove the existence of a flow by constructing it recursively. Let $\sigma$ be a topological ordering of the states in $\gG^{\top}$, the transpose of $\gG$ (\ie the pointed DAG constructed from $\gG$ by reversing all of its edges: $s' \rightarrow s \in \gG^{\top}$ if and only if $s \rightarrow s' \in \gG$); such a topological ordering exists thanks to the acyclicity of $\gG$. We first initialize the values of the flows at the terminating transitions of the form $x \rightarrow \terminal$ to satisfy the boundary conditions: $F(x \rightarrow \terminal) = R(x)$. Then the values of the flows at the other edges are determined by following the ordering of states $s'\in\sigma$: for any transition $s \rightarrow s'$, we define the flow $F(s\rightarrow s')$ as
    \begin{equation}
        F(s\rightarrow s') = \frac{1}{|\parents_{\gG}(s')|}\sum_{s''\in\children_{\gG}(s')}F(s'\rightarrow s'').
        \label{eq:proof-existence-flow-flow-matching}
    \end{equation}
    This is well defined because all the flows of the form $F(s'\rightarrow s'')$ have already been determined, since any child $s'' \in \children_{\gG}(s')$ appears before $s'$ in the ordering $\sigma$. It is clear that the function $F$ defined recursively by \cref{eq:proof-existence-flow-flow-matching} also satisfies the flow matching condition (in addition to the boundary conditions), and is non-negative.
\end{proof}

While a solution of \cref{eq:flow-matching-boundary-condition} exists, it is important to note that it is not unique, and the construction in \cref{eq:proof-existence-flow-flow-matching} was arbitrary (assigning the total amount of outgoing flow at $s'$ equally among its incoming edges). Nevertheless, any of these solutions will still induce the same terminating state distribution $P_{F}^{\top}(x) \propto R(x)$, for $P_{F}$ the corresponding forward transition probability defined in \cref{eq:forward-transition-probability-flow-matching}.

\paragraph{Relation to max-flow problems} We saw in \cref{sec:markovian-flows} that Markovian flows have a natural connection with (standard) flow networks in graph theory, where each edge has infinite capacity: $F(s\rightarrow s') \leq +\infty$. When introducing the boundary conditions of \cref{eq:boundary-conditions}, this creates new capacity constraints, where $R(x)$ now determines the capacity of the flow through the terminating transitions of the form $x \rightarrow \terminal$: $F(x\rightarrow \terminal) \leq R(x)$. Using this perspective, we can see that finding a flow satisfying \cref{eq:flow-matching-boundary-condition} is equivalent to finding a maximum flow (in a graph theoretical sense), and that any solution flow necessarily saturates the capacity for all the terminating transitions (\ie it satisfies the boundary conditions). To obtain a flow satisfying \cref{thm:flow-matching-proportional-reward}, we can therefore use any algorithm for finding a max-flow, such as the \emph{Ford-Fulkerson algorithm} \citep{ford1956fordfulkerson}, or the \emph{push-relabel algorithm} \citep{goldberg1988pushrelabel} that can find a solution in time $O(|\gS|^{3})$. Alternatively, there exists a dynamic programming algorithm capable of finding a solution in $O(|\gA|)$ \citep{lahlou2024gfnthesis}. In many cases though, the state space will be so large that these approaches become impractical. We will see in \cref{sec:gflownet-learning-flow-network} how this flow can be found approximately by \emph{learning} it with gradient methods (via amortization): from visiting a subset of possible transitions and terminating states, one can generalize more generally to other states, thanks to the generalization capabilities of neural networks.

\subsection{Flows over cyclic graphs}
\label{sec:flows-over-cyclic-graphs}

Generative flow networks adopt a purely \emph{constructive} approach, where pieces are only allowed to be added to ensure the acyclicity of the state space, and are never removed. Otherwise, if removal steps were also allowed, this would have the risk of introducing cycles in $\gG$, where some piece may be removed and added back to reach the same state. To handle both the addition and removal of pieces, similar to MCMC \citep{madigan1995structuremcmc}, a natural solution is to augment the states with a ``timestamp'' $t$ that increases at each step of the generation. Concretely, this means that the state space of this ``augmented'' generative flow network is $\gS \times \sN$, with transitions $(s, t) \rightarrow (s', t+1)$ for all $t \in \sN$ if and only if $s \rightarrow s'$ is a transition in the original state graph $\gG$ (not necessarily acyclic). Although this ensures that this new flow network is acyclic, this is now breaking the assumption that the state space should be finite; we will see in \cref{sec:gflownet-discrete-spaces} that under some conditions GFlowNets can also be extended to infinite state spaces.

Another way of accommodating for cycles in the state graph is to find a \emph{minimal flow} that satisfies the conditions of \cref{thm:flow-matching-proportional-reward}.
\begin{proposition}
    \label{prop:cyclic-flows-minimum-flow}
    Let $\gG = (\widebar{\gS}, \gA)$ be an arbitrary directed graph, where $\terminal \notin \gS$ is a terminal state that has no child in $\gG$, $\gX = \parents_{\gG}(\terminal)$ be the set of terminating states of $\gG$, and let $R: \gX \rightarrow \sR_{+}$ be a reward function defined over the terminating states. Let $F: \gA\rightarrow \sR_{+}$ be a function over the edges of $\gG$ solution of the following constrained optimization problem:
    \begin{equation}
        \begin{aligned}
            \min_{F \geq 0}\ & \sum_{s\rightarrow s'\in\gG}F(s\rightarrow s')\\
            \mathrm{s.t.}\ & \sum_{s\in\parents_{\gG}(s')}F(s\rightarrow s') - \sum_{s''\in\underline{\children}_{\gG}(s')}F(s'\rightarrow s'') = R(s') \qquad & \forall s'\neq s_{0}
        \end{aligned}
        \label{eq:cyclic-flows-constrained-optimization-problem}
    \end{equation}
    using the convention $R(s') = 0$ for any non-terminating state $s'\in\gS\backslash\gX$. Then the terminating state probability distribution associated with the forward transition probability $P_{F}$ defined in \cref{eq:forward-transition-probability-flow-matching} is proportional to the reward: $P_{F}^{\top}(x) \propto R(x)$.
\end{proposition}

\begin{proof}
    Although we can't directly apply \cref{prop:existence-flow-flow-matching} since $\gG$ is not necessarily acyclic anymore, by adopting the ``max-flow'' perspective mentioned in \cref{sec:flow-matching-condition-boundary-constraint}, the problem \cref{eq:cyclic-flows-constrained-optimization-problem} admits a solution (the constraint set is not empty), possibly infinite. We will show that if $F$ is a solution of the problem, then it has no ``positive cycle'', \ie a cycle where the flow is positive on each of its edges.

    Suppose that $F$ is a solution of \cref{eq:cyclic-flows-constrained-optimization-problem}, and that there exists a cycle $\gamma = (s_{1}, s_{2}, \ldots, s_{T}, s_{1})$ such that $F(s_{t}\rightarrow s_{t+1}) > 0$ (with the convention $s_{T+1} = s_{1}$). We can build a new function $F'$ from $F$ as
    {\allowdisplaybreaks%
    \begin{align}
        F'(s_{t} \rightarrow s_{t+1}) &= F(s_{t} \rightarrow s_{t+1}) - \varepsilon && \textrm{if $s_{t} \rightarrow s_{t+1} \in \gamma$}\\
        F'(s\rightarrow s') &= F(s\rightarrow s') && \textrm{otherwise},
    \end{align}}%
    where $\varepsilon > 0$ is chosen such that $F'$ stays non-negative. It is easy to see that $F'$ satisfies the flow matching condition (constraint of \cref{eq:cyclic-flows-constrained-optimization-problem}): for any state along $\gamma$, we removed the same amount $\varepsilon$ from its incoming and outgoing flow $F$ that was already satisfying the flow matching condition. However, the sum of flows of $F'$ is strictly smaller than the one of $F$
    \begin{equation}
        \sum_{s\rightarrow s'\in\gG}F'(s\rightarrow s') = \sum_{s\rightarrow s'\in\gG}F(s\rightarrow s') - T\varepsilon < \sum_{s\rightarrow s'\in\gG}F(s\rightarrow s'),
    \end{equation}
    which contradicts the fact that $F$ was a minimizer of \cref{eq:cyclic-flows-constrained-optimization-problem}. Since a solution $F$ of this problem has no positive cycle, we can construct a pointed DAG $\gG' = (\widebar{\gS}, \gA')$ from the directed graph $\gG$ by only extracting the edges that have positive flow: $s\rightarrow s'\in\gA'$ iff $F(s\rightarrow s') > 0$. In particular, each terminating transition of the form $x\rightarrow \terminal$ is kept in $\gG'$, provided its corresponding reward is positive (this is not limiting, since otherwise it wouldn't contribute to the terminating state probability anyway). Therefore, we have a function $F$ that satisfies the flow matching condition, as well as the boundary condition for the pointed DAG $\gG'$, since it satisfies the constraints in \cref{eq:cyclic-flows-constrained-optimization-problem}. By \cref{thm:flow-matching-proportional-reward}, this shows that the terminating state probability distribution associated with $P_{F}$ defined by \cref{eq:forward-transition-probability-flow-matching} is proportional to the reward: $P_{F}^{\top}(x) \propto R(x)$.
\end{proof}

These considerations are only necessary when we have little control over the way the state graph $\gG$ is defined (\eg black-box simulations). However in many practical cases, we are free to choose how the state space is structured (similar to how one can choose a transition kernel in MCMC), and therefore it is recommended to incorporate the acyclicity of $\gG$ in our design choices directly to avoid any additional complexity. All the GFlowNets introduced in this thesis will be guaranteed to have a DAG structure.

\subsection{Alternative conditions}
\label{sec:alternative-conditions}
The flow matching condition of \cref{thm:flow-matching-condition} is only one of many characterizations of a Markovian flow, as we saw throughout \cref{sec:flow-matching-conditions}. We can obtain similar results, based on any of the flow matching conditions we presented in the previous chapter. For example, we can combine the detailed balance condition of \cref{thm:detailed-balance-condition} with the boundary condition. Note that since the detailed balance condition does not depend directly on the edge flow function $F(s\rightarrow s')$ but on a state flow function $F(s)$ and transition probabilities, the boundary condition needs to be slightly adapted to only depend on $F(x)$ and $P_{F}(\terminal\mid x)$ (instead of $F(x\rightarrow \terminal)$ directly).
\begin{theorem}[Detailed balance condition]
    \label{thm:detailed-balance-proportional-reward}\index{Detailed balance!Condition}
    Let $\gG = (\widebar{\gS}, \gA)$ be a pointed DAG, where $\gX\subseteq \gS$ is the set of terminating states of $\gG$, and let $R: \gX \rightarrow \sR_{+}$ be a reward function defined over the terminating states. If the functions $F: \gS\rightarrow \sR_{+}$, $P_{F}: \gS \rightarrow \Delta(\children_{\gG})$, and $P_{B}: \gS \rightarrow \Delta(\parents_{\gG})$ satisfy the \emph{detailed balance condition} for all transitions $s\rightarrow s'\in\gG$ such that $s'\neq \terminal$
    \begin{equation}
        F(s)P_{F}(s'\mid s) = F(s')P_{B}(s\mid s'),
    \end{equation}
    and satisfies the boundary conditions $F(x)P_{F}(\terminal\mid x) = R(x)$ for all $x\in\gX$, then the terminating state distribution associated with $P_{F}$ is proportional to the reward: $P_{F}^{\top}(x) \propto R(x)$.
\end{theorem}

\begin{proof}
    This is a direct consequence of \cref{thm:detailed-balance-condition} and \cref{prop:terminating-state-probability-from-flow}. Since $F$, $P_{F}$ and $P_{B}$ satisfy the detailed balance condition, then by \cref{thm:detailed-balance-condition} there exists a unique Markovian flow $F^{\star}$ whose state flow matches $F$, and whose forward transition probability $P^{\star}_{F}$ matches $P_{F}$. By definition of $P^{\star}_{F}$ in \cref{def:transition-probabilities-from-flow}, we know that for any terminating state $x\in\gX$, $F^{\star}(x\rightarrow \terminal) = F^{\star}(x)P^{\star}_{F}(\terminal \mid x) = F(x)P_{F}(\terminal\mid x)$. By \cref{prop:terminating-state-probability-from-flow}, the terminating state probability distribution associated with $P^{\star}_{F} = P_{F}$ is proportional to $F^{\star}(x\rightarrow \terminal) = F(x)P_{F}(\terminal\mid x)$. Using the boundary condition $F(x)P_{F}(\terminal\mid x) = R(x)$, we can conclude that $P_{F}^{\top}(x) \propto R(x)$.
\end{proof}

We can do the same exercise for the trajectory balance condition of \cref{thm:trajectory-balance-condition}. One advantage of the trajectory balance condition is that the flow at terminating edges (which appears in the boundary condition) is already part of its parametrization. We can therefore plug the reward function directly into the condition without the need for an additional explicit boundary condition, unlike for detailed balance and flow matching (although we saw in \cref{eq:flow-matching-boundary-condition} that the reward function can be integrated inside the flow matching condition). As a consequence, the trajectory balance condition is considered the most popular in the GFlowNet literature.

\begin{theorem}[Trajectory balance condition]
    \label{thm:trajectory-balance-proportional-reward}\index{Trajectory balance!Condition}
    Let $\gG = (\widebar{\gS}, \gA)$ be a pointed DAG, where $\gX\subseteq \gS$ is the set of terminating states of $\gG$, and let $R: \gX\rightarrow \sR_{+}$ be a reward function defined over the terminating states. If $Z > 0$, and the functions $P_{F}: \gS \rightarrow \Delta(\children_{\gG})$, and $P_{B}: \gS\rightarrow \Delta(\parents_{\gG})$ satisfy the \emph{trajectory balance condition} for all complete trajectories $\tau=(s_{0}, s_{1}, \ldots, s_{T}, \terminal) \in \gT$
    \begin{equation}
        Z\prod_{t=0}^{T}P_{F}(s_{t+1}\mid s_{t}) = R(s_{T})\prod_{t=1}^{T}P_{B}(s_{t-1}\mid s_{t}),
    \end{equation}
    with the convention $s_{T+1}=\terminal$, then the terminating state distribution associated with $P_{F}$ is proportional to the reward: $P_{F}^{\top}(x) \propto R(x)$.
\end{theorem}

\begin{proof}
    This is a direct consequence of \cref{thm:trajectory-balance-condition} and \cref{prop:terminating-state-probability-from-flow}. Since $Z$, $P_{F}$, and $P_{B}$ (and $F_{\terminal} \triangleq R$) satisfy the flow matching condition, then by \cref{thm:trajectory-balance-condition} there exists a unique Markovian flow $F^{\star}$ whose edge flow at the terminating transitions matches $R$, and whose forward transition probability $P^{\star}_{F}$ matches $P_{F}$. By \cref{prop:terminating-state-probability-from-flow}, the terminating state probability distribution associated with $P_{F}^{\star} = P_{F}$ is proportional to the reward: $P_{F}^{\top}(x) \propto R(x)$.
\end{proof}
We do not include a similar theorem based on the sub-trajectory balance condition of \cref{thm:sub-trajectory-balance-condition} though, because we saw that in general this was not a sufficient condition for a Markovian flow.

\subsection{Case study: Bayesian inference over decision trees}
\label{sec:case-study-application-bayesian-inference-decision-trees}
In this section, we will see an example of application of GFlowNets applied to Bayesian inference. Consider a dataset $\gD = \{(\vx_{1}, y_{1}), \ldots, (\vx_{N}, y_{N})\}$ of inputs/outputs, where each $\vx \in \{0, 1\}^{p}$ is a vector of $p$ binary features, and $y \in \{0, 1\}$ is also a binary label. A natural choice for tabular data like this is to use \emph{decision trees} in order to model the underlying system of interest. Instead of returning a single (weak) classifier though, most popular approaches use an ensemble of trees like a random forest \citep{breiman2001randomforest}, in order to increase the expressive power of the model. Alternatively, taking a Bayesian perspective from \cref{sec:bayesian-model-averaging}, we could consider \emph{all} the possible decision trees one could build over $\gD$ (\eg up to a certain maximum depth $d$), and reweight them using the posterior distribution over trees $P(T\mid \gD)$ when making our predictions.

Because the number of such decision trees is combinatorially large (in both $p$ and $d$), characterizing the whole posterior distribution can be very challenging, due to the intractable normalization constant (evidence)
\begin{equation}
    P(T\mid \gD) = \frac{P(\gD\mid T)P(T)}{P(\gD)} = \frac{P(\gD\mid T)P(T)}{\sum_{T'}P(\gD\mid T')P(T')}.
    \label{eq:bayes-rule-decision-trees}
\end{equation}
However, since decision trees (or at least their structure) are discrete and compositional objects, and the distribution is defined up to a normalization constant, we could instead model it using a generative flow network. Each state of this flow network would correspond to the structure of a decision tree with maximum depth $d$. The states are organized into a pointed DAG $\gG$, where the initial state $T_{0}$ corresponds to a ``tree'' with a single node (leaf) containing all the data in $\gD$, and where a transition $T\rightarrow T' \in \gG$ corresponds to \emph{expanding} a leaf of $T$ in order to obtain $T'$.

\begin{figure}[t]
    \centering
    \includegraphics[width=480pt]{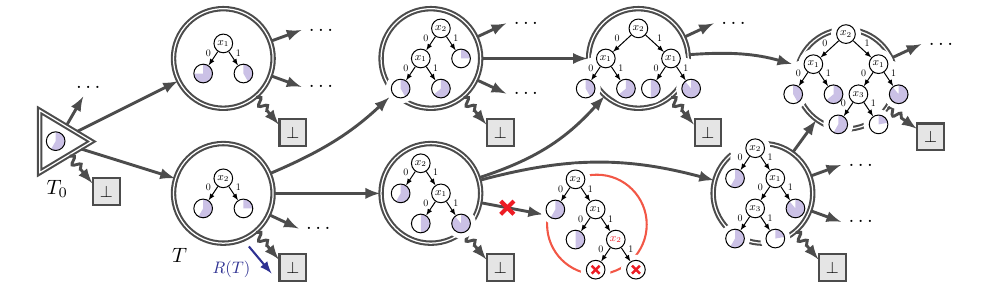}
    \caption[Bayesian inference of decision tree structures with a GFlowNet]{Bayesian inference of decision tree structures with a generative flow network. This is an illustration of the pointed DAG $\gG$, each state being the structure of a decision tree. The shading at the leaves indicates the label proportions. A transition $T \rightarrow T'$ corresponds to selecting a leaf and a variable and expanding the tree. The state in red is invalid since the same decision variable is used twice on the same path to a leaf node; it will not be included in the structure of $\gG$}
    \label{fig:gflownet-decision-trees}
\end{figure}

Concretely, transitioning to a new graph requires selecting which leaf to expand, and which variable $\{x_{1}, \ldots, x_{p}\}$ to use for the new decision node. But these transitions are also subject to additional constraints: (1) a leaf which is already at the maximum length cannot be further expanded, (2) the variable for the new decision node must be distinct from its ancestors in the tree (\ie we can't split on the same variable multiple times), and (3) we can't have ``trivial'' splits, where no example of $\gD$ ends up in one of the leaves, meaning that there is no unnecessary decision node. All of these rules can be easily verified, and they shape the structure of $\gG$; see \cref{fig:gflownet-decision-trees} for an illustration. Interestingly, all the states here are terminating ($\gX \equiv \gS$), since they all correspond to valid samples of the distribution \cref{eq:bayes-rule-decision-trees} (valid decision trees); we will see in \cref{sec:modified-detailed-balance} that this has some consequences in terms of the flow matching conditions that can be applied. Unlike in \cref{sec:probabilistic-inference-control-problem}, we are making full use of the DAG assumption for $\gG$ here, since there may be multiple ways of constructing the same decision tree, depending on the order in which decision nodes are added.

Given \cref{eq:bayes-rule-decision-trees} and the fact that generative flow networks are used to model distributions up to normalization, we can clearly choose $R(T) = P(\gD\mid T)P(T)$ as our reward function, which we assume can be computed analytically for any decision tree $T$. On the one hand, the prior $P(T)$ can be used for example to penalize more complex decision trees; \eg we can set $P(T) \propto \exp(-\beta|T|)$, where $|T|$ is the size of $T$, to mimic the \emph{Akaike information criterion} (AIC) frequently used in model selection \citep{akaike1974aic}. The marginal likelihood $P(\gD\mid T)$, on the other hand, can be decomposed into pieces following the partition of $\gD$ obtained with the decision tree
\begin{align}
    \log P(\gD\mid T) &= \log\left[\int_{\theta}P(\gD\mid T, \theta)P(\theta\mid T)d\theta\right] = \log \left[\int_{\theta}\left(\prod_{n=1}^{N}P(y_{n}\mid \vx_{n}, T, \theta)\right)P(\theta\mid T)d\theta\right]\nonumber\\
    &= \log \Bigg[\int_{\theta}\prod_{l\in T}\Bigg(P(\theta_{l})\prod_{n:\vx_{n}\in l}P(y_{n}\mid \theta_{l})\Bigg)d\theta\Bigg] \\
    &= \sum_{l \in T}\log\Bigg[\int_{\theta_{l}}\Bigg(\prod_{n: \vx_{n}\in l}P(y_{n}\mid \theta_{l})\Bigg)P(\theta_{l}) d\theta_{l}\Bigg],
    \label{eq:marginal-likelihood-decision-tree}
\end{align}
where the outer sum is over the leaves $l$ of the decision tree, the decisions at each leaf depending on some parameters $\theta_{l}$ that need to be integrated over, based on the subset of $\gD$ consistent with the leaf $l$. These integrals can be computed in closed form if we assume that the parameters $\theta_{l}$ have a Beta prior (\ie a Beta-Binomial model).

Since the normalization constant $P(\gD)$ is rarely tractable, GFlowNets are particularly appropriate for Bayesian inference in general, and especially in Bayesian model averaging\index{Bayesian inference!Bayesian model averaging} \citep{madigan1994occam} since these models are typically discrete objects. This formulation based on a pointed DAG over states allows us to have maximum flexibility, the extreme case being where $\gG$ is a rooted tree of depth 1, where the initial state (empty state) is connected to all possible outcomes of the posterior distribution (\eg the different models). However, flow networks become exponentially more effective when these objects exhibit some compositional structure, just like with decision trees, trading off the generation of an object in multiple steps for a much smaller branching factor (the components to add) in $\gG$. We will see in \cref{chap:dag-gflownet} and onward how GFlowNets can be used to model the posterior distribution over the DAG structure of a Bayesian network.

\section{Learning a generative flow network}
\label{sec:gflownet-learning-flow-network}
Up to this point, we assumed that the flow $F$, or any quantity parametrizing the generative flow network, were simply \emph{functions} that satisfied some conditions. Taking the example of the flow matching condition in \cref{sec:flow-matching-condition-boundary-constraint}, having a ``tabular representation'' of this flow function would then require specifying the value of $F(s\rightarrow s')$ for each possible transition $s \rightarrow s'\in\gG$, which can be prohibitively large in practice. Instead, we will often consider a parametric family of functions $F_{\phi}$ to \emph{amortize} the cost of finding a flow (as in \cref{sec:amortized-inference}), and find parameters \gls{phi} that approximately satisfy the conditions of \cref{thm:flow-matching-proportional-reward}. This is a \emph{variational approach}\index{Variational inference}, where inference over the unknown target distribution $P^{\star}$ is viewed as an optimization problem; we will come back to the links between GFlowNets \& variational inference in details in \cref{sec:gflownets-variational-inference}.

\subsection{Flow matching losses}
\label{sec:flow-matching-losses}
In order to find these optimal parameters that approximately satisfy the conditions given in \cref{sec:probabilistic-modeling-flow-networks}, we can convert any of those flow matching conditions into a (non-linear) least-square objective of the form
\begin{equation}
    \gL(\phi) = \frac{1}{2}\E_{\pi_{b}}\big[\Delta^{2}(\cdot; \phi)\big],
    \label{eq:least-square-objective}
\end{equation}
where $\pi_{b}$ is an arbitrary distribution over appropriate quantities (\ie states, transitions, or complete trajectories) with full support, meaning that it assigns a non-zero probability to all elements, and \gls{residual} are \emph{residuals}\index{Residual} that quantify the ``mismatch'' in the flow matching (and boundary) conditions. By analogy with flow matching conditions, we will call this a \emph{flow matching loss} (associated with $\Delta(\cdot;\phi)$). For example, taking inspiration from the formulation in \cref{eq:flow-matching-boundary-condition}, the residual associated with the flow matching condition of \cref{thm:flow-matching-proportional-reward} can be written as
\begin{equation}
    \Delta_{\mathrm{FM}}(s'; \phi) = \log \frac{\sum_{s\in\parents_{\gG}(s')}F_{\phi}(s\rightarrow s')}{\sum_{s''\in\underline{\children}_{\gG}(s')}F_{\phi}(s'\rightarrow s'') + R(s')}.
    \label{eq:flow-matching-loss}\index{Flow matching!Loss|textbf}
\end{equation}
In that case, $\pi_{b}$ is a distribution over states $s'\neq s_{0}$. The choice of working in logarithmic space is intuitive: since $F_{\phi}(s\rightarrow s')$ would ideally represent a flow, flows closer to the initial state will naturally have a significantly larger magnitude than those closer to the terminal state, since they would have to accumulate downstream flows (because of the flow matching conditions). Even for upstream and downstream flows at the level of a state $s'$, these changes in orders of magnitudes are better controlled numerically in logarithmic space. It can easily be shown that if the flow matching loss associated with $\Delta_{\mathrm{FM}}(s';\phi)$ is minimized, then the corresponding flow satisfies the conditions of \cref{thm:flow-matching-proportional-reward}, thanks to the full support of $\pi_{b}$, provided that the flow satisfying these conditions can be represented by $F_{\phi}$ (\ie the minimum flow matching loss is zero).

\begin{proposition}
    \label{prop:minimization-flow-matching-loss}
    Let $\gG = (\widebar{\gS}, \gA)$ be a pointed DAG, and let a family of functions $F_{\phi}: \gA \rightarrow \sR_{+}$ parametrized by $\phi$. Let $\gL_{\mathrm{FM}}(\phi)$ be the objective in \cref{eq:least-square-objective} using the flow matching residual $\Delta_{\mathrm{FM}}(s';\phi)$ defined in \cref{eq:flow-matching-loss}. There exists $\phi^{\star}$ such that $F_{\phi^{\star}}$ satisfies the flow matching \& boundary conditions of \cref{thm:flow-matching-proportional-reward} if and only if $\gL_{\mathrm{FM}}(\phi^{\star}) = 0$.
\end{proposition}
\begin{proof}
    Clearly if $F_{\phi^{\star}}$ satisfies the conditions of \cref{thm:flow-matching-proportional-reward}, then the loss will be zero since all the residuals themselves will be zeros. Conversely, if $\gL_{\mathrm{FM}}(\phi^{\star}) = 0$, we can write it more explicitly as
    \begin{equation}
        \gL_{\mathrm{FM}}(\phi^{\star}) = \frac{1}{2}\sum_{s'\neq s_{0}}\pi_{b}(s')\Delta_{\mathrm{FM}}^{2}(s';\phi^{\star}) = 0.
    \end{equation}
    Since we assumed that $\pi_{b}$ has full support, we know that for all $s'\neq s_{0}$, $\pi_{b}(s') > 0$. Moreover since $\Delta_{\mathrm{FM}}^{2}(s';\phi^{\star}) \geq 0$, this implies that the flow matching residuals are all zeros, and therefore \cref{eq:flow-matching-boundary-condition} is satisfied for all $s'\neq s_{0}$.
\end{proof}

The key property we used was that $\pi_{b}$ had full support. Indeed if there was a state for which $\pi_{b}(s) = 0$, then $\Delta_{\mathrm{FM}}(s;\phi)$ could be completely arbitrary while still minimizing $\gL_{\mathrm{FM}}$. We can also define flow matching losses based on other conditions, such as the detailed balance loss based on \cref{thm:detailed-balance-proportional-reward}, with the following residual defined over transitions $s\rightarrow s'$ such that $s'\neq \terminal$:
\begin{equation}
    \Delta_{\mathrm{DB}}(s\rightarrow s'; \phi) = \log \frac{F_{\phi}(s)P_{F}^{\phi}(s'\mid s)}{F_{\phi}(s')P_{B}^{\phi}(s\mid s')}.
    \label{eq:detailed-balance-loss}\index{Detailed balance!Loss|textbf}
\end{equation}
Unlike the flow matching loss above, this residual relies on a family of state flows $F_{\phi}: \gS \rightarrow \sR_{+}$, forward transitions probabilities $P_{F}^{\phi}: \gS \rightarrow \Delta(\children_{\gG})$, and backward transition probabilities $P_{B}^{\phi}: \gS \rightarrow \Delta(\parents_{\gG})$; we assume for clarity that $\phi$ encapsulates the parameters of all these functions. Another major difference with the flow matching residual in \cref{eq:flow-matching-loss} is that this covers the detailed balance condition, but not the boundary conditions which depend on the reward function. To address this, we need to define an additional residual for the boundary conditions at terminating transitions $x\rightarrow \terminal$, with $x\in\gX$:
\begin{equation}
    \Delta_{\mathrm{DB}}(x\rightarrow \terminal; \phi) = \log \frac{F_{\phi}(x)P_{F}^{\phi}(\terminal\mid x)}{R(x)}.
    \label{eq:reward-matching-loss}
\end{equation}
This type of residual is also called a \emph{reward matching loss}\index{Reward matching loss} \citep{bengio2023gflownetfoundations}. For the detailed balance loss, $\pi_{b}$ is a distribution with full support over all the transitions of $\gG$, including the terminating transitions. Similarly, we also have the trajectory balance loss based on \cref{thm:trajectory-balance-proportional-reward}, with the following residual defined over complete trajectories $\tau = (s_{0}, s_{1}, \ldots, s_{T}, \terminal)$:
\begin{equation}
    \Delta_{\mathrm{TB}}(\tau;\phi) = \log \frac{Z_{\phi}\prod_{t=0}^{T}P_{F}^{\phi}(s_{t+1}\mid s_{t})}{R(s_{T})\prod_{t=1}^{T}P_{B}^{\phi}(s_{t-1}\mid s_{t})}.
    \label{eq:trajectory-balance-loss}\index{Trajectory balance!Loss|textbf}
\end{equation}
In addition to a parametrization of the forward and backward transition probabilities, like in the detailed balance loss, the trajectory balance loss also uses an additional learned scalar $Z_{\phi} > 0$ to represent the total flow. For the trajectory balance loss, $\pi_{b}$ is a distribution with full support over complete trajectories. We present a simplified training loop to learn a GFlowNet based on the trajectory balance loss in \cref{alg:training-gflownet-tb-loss}. Note that even if working in logarithmic space for the detailed balance and the trajectory balance losses was inherited from the original flow matching loss \cref{eq:flow-matching-loss}, we will see in \cref{sec:equivalence-maxent-rl-gflownet} that it finds a natural interpretation when seeing these from the perspective of maximum entropy RL. Similar to \cref{prop:minimization-flow-matching-loss}, it can also be shown that minimizing the detailed balance loss satisfies the conditions of \cref{thm:detailed-balance-proportional-reward}, and minimizing the trajectory balance loss satisfies the conditions of \cref{thm:trajectory-balance-proportional-reward}, provided their respective quantities can be represented by the family parametrized by $\phi$ (\ie the minimum loss is zero).

Other losses that are not based on a least-square objective have also been introduced in the GFlowNet literature. For example, \citet{zhang2023gfnrobustscheduling} introduced a \emph{log-partition variance loss} \citep{richter2020vargrad} that minimizes the variance of an estimate of the log-partition function induced by the trajectory balance condition.

\begin{algorithm}[t]
    \caption{Training a generative flow network with the trajectory balance loss \cref{eq:trajectory-balance-loss}.}
    \label{alg:training-gflownet-tb-loss}
    \begin{algorithmic}[1]
        \Require A pointed DAG $\gG = (\widebar{\gS}, \gA)$, a reward function $R: \gX \rightarrow \sR^{+}$, a batch size $K$.
        \State Initialize the parameters $\phi$ of $P_{F}^{\phi}$, $P_{B}^{\phi}$ \& $Z_{\phi}$
        \Loop
            \LComment{Sample $K$ trajectories \& get the reward of their terminating states}
            \State $\gB \leftarrow \emptyset$
            \For{$k = 1, \dots, K$}
                \State Initialize the state: $s_{0}$
                \Repeat
                    \State Sample the next state: $s_{t+1} \sim \pi_{b}(\cdot \mid s_{t})$
                \Until{$s_{T+1} = \terminal$}
                \State $\gB \leftarrow \gB \cup \{(\tau, R(s_{T}))\}$ \Comment{$\tau = (s_{0}, s_{1}, \ldots, s_{T}, \terminal)$}
            \EndFor
            \State \vphantom{Update}
            \LComment{Update the parameters with stochastic gradient descent}
            \State Estimate the loss $\gL_{\mathrm{TB}}(\phi) = \frac{1}{2}\E_{\pi_{b}}\big[\Delta_{\mathrm{TB}}^{2}(\tau;\phi)\big]$ based on $\gB$ (Monte Carlo) \Comment{\cref{eq:trajectory-balance-loss}}
            \State Update the parameters: $\phi \leftarrow \phi - \beta \nabla_{\phi}\gL_{\mathrm{TB}}(\phi)$
        \EndLoop
        \State \Return the forward transition probability $P_{F}^{\phi}(s'\mid s)$
    \end{algorithmic}
\end{algorithm}

\subsection{Off-policy training}
\label{sec:off-policy-training}
The least-square objective in \cref{eq:least-square-objective} is an expected loss that depends on a distribution $\pi_{b}$ with full support. Although we used the same notation to denote distributions over various quantities depending on the type of flow matching loss (\eg $\pi_{b}$ is a distribution over states $s'$ for the flow matching loss \cref{eq:flow-matching-loss}, and it is a distribution over complete trajectories $\tau$ for the trajectory balance loss \cref{eq:trajectory-balance-loss}), they will all typically be derived from a distribution of the form $\pi_{b}(s_{t+1}\mid s_{t})$ consistent with $\gG$; for example to build a distribution $\pi_{b}(\tau)$ over complete trajectories thanks to \cref{lem:PF-distribution-suffix}. This is called a \emph{behavior policy}\index{Behavior policy|textbf} in the reinforcement learning literature \citep{sutton2018introrl}. The simplest choice for such a behavior policy is to use the current forward transition probability $\pi_{b}(s_{t+1}\mid s_{t}) = P_{F}^{\phi}(s_{t+1}\mid s_{t})$ (possibly derived from $F_{\phi}$ in the case of the flow matching loss, by normalizing the outgoing flows \cref{eq:forward-transition-probability-flow-matching}); this strategy is called \emph{on-policy}.

The behavior policy can be a priori completely arbitrary though, as long as it has full support. When $\pi_{b}$ is different from the target forward transition probability, this is called \emph{off-policy} learning. This approach is usually effective to encourage exploration, which is often essential to discover regions of the state space leading to yet unseen modes of the target distribution. A naive way to incorporate exploration is via the \emph{$\varepsilon$-sampling} strategy, where the next state is sampled from $P_{F}^{\phi}$ with probability $1 - \varepsilon$, and uniformly at random with probability $\varepsilon \in (0, 1]$. The behavior policy can therefore be written as the following mixture model:
\begin{equation}
    \pi_{b}(s_{t+1}\mid s_{t}) = (1 - \varepsilon)P_{F}^{\phi}(s_{t+1}\mid s_{t}) + \frac{\varepsilon}{|\children_{\gG}(s_{t})|}.
    \label{eq:epsilon-sampling-behavior-policy}
\end{equation}
An alternative strategy is to use a \emph{temperature} parameter $\alpha > 0$ to control how ``diffused'' the probabilities of reaching the next states are, based on the current $P_{F}^{\phi}$:
\begin{equation}
    \pi_{b}(s_{t+1}\mid s_{t}) \propto \big[P_{F}^{\phi}(s_{t+1}\mid s_{t})\big]^{1/\alpha}.
    \label{eq:tempering-behavior-policy}
\end{equation}
Although either $\varepsilon$ or $\alpha$ can be fixed hyperparameters, in practice it is customary to use a schedule to slowly transition from the uniform policy ($\varepsilon = 1$, or $\alpha \rightarrow +\infty$) to the forward transition probability $P_{F}^{\phi}$ ($\varepsilon \approx 0$, or $\alpha \approx 1$, in order to still keep some exploratory behavior) over the course of training. This is inherited once again from the reinforcement learning literature \citep{mnih2015dqn}.

There also exists more advanced strategies to define the behavior policy. \citet{rectorbrooks2023thompsongfn} introduced a strategy based on Thompson sampling \citep{thompson1933thompsonsampling} to improve the exploration in GFlowNets by maintaining an approximate posterior over policies. \citet{morozov2024mctsgfn} proposed to use Monte Carlo tree search \citep{xiao2019ments} to improve planning at both training and test time. QGFN \citep{lau2024qgfn} used a policy based on Q-values learned by standard reinforcement learning in order to encourage exploration in regions with high reward. Finally, an increasingly popular way to train generative flow networks off-policy is to use reweighted experience stored in a \emph{replay buffer} over the course of training, possibly acquired using an exploration-enhanced behavior policy, in order to compute the loss \citep{deleu2022daggflownet,vemgal2023gfnreplaybuffer}, as with \emph{prioritized experience replay} \citep{schaul2016prioritizedexperiencereplay}..

\paragraph{Differentiation through the behavior policy} We will use gradient based methods in order to optimize the flow matching losses we presented in the previous section (see line 14 of \cref{alg:training-gflownet-tb-loss}). We can show that the gradient of the objective in \cref{eq:least-square-objective} can also be written as an expectation over the behavior policy \citep{schulman2015gradientstochastic}
\begin{equation}
    \nabla_{\phi}\gL(\phi) = \frac{1}{2}\E_{\pi_{b}}\big[\nabla_{\phi}\Delta^{2}(\cdot;\phi) + \nabla_{\phi}\log \pi_{b}(\cdot)\Delta^{2}(\cdot;\phi)\big],
    \label{eq:least-square-gradient}
\end{equation}
where again the quantities the expectation is taken over depend on the specific flow matching loss. When the behavior policy $\pi_{b}$ is completely independent of $\phi$, then the second term of this expectation vanishes. However, if $\pi_{b}$ does depend on $\phi$, then this term should still appear; this is most obviously the case when we are on-policy, and the behavior policy $\pi_{b} \equiv P_{F}^{\phi}$ is the current forward transition probability. But in a subtle way, this would also be the case off-policy when we consider either $\varepsilon$-sampling in \cref{eq:epsilon-sampling-behavior-policy}, or tempering in \cref{eq:tempering-behavior-policy}, as they both depend on $\phi$ as well. In practice though, we will always ignore the second term of \cref{eq:least-square-gradient}, and estimate the gradient \emph{as if $\pi_{b}$ was completely independent of $\phi$}: $\nabla_{\phi}\gL(\phi) \approx \frac{1}{2}\E_{\pi_{b}}[\nabla_{\phi}\Delta^{2}(\cdot;\phi)]$. This does not hinder convergence, and it can be estimated with Monte Carlo based on samples from $\pi_{b}$.

\subsection{Convergence guarantees}
\label{sec:convergence-guarantees-gflownets}
The approximation of the flow matching conditions (and boundary conditions) inevitably leads to errors, the source of which is typically two-fold: (1) due to the choice of the variational family to parametrize the quantities of interest (\eg flows, transition probabilities), and (2) due to the finite nature of optimization. Unlike standard methods in variational inference that minimize some divergence between a variational approximation and the target distribution (typically the reverse KL-divergence; \citealp{jordan1999introductionvi}), GFlowNets instead minimize some losses that quantify the ``mismatch'' in the flow matching conditions which, upon being satisfied, ensure that the terminating state probability corresponds to the target distribution. By way of getting this mismatch as small as possible over the course of optimization, we hope that the model will indirectly get closer to the target in some way. However, does minimizing any of the flow matching losses presented in \cref{sec:flow-matching-losses} offer any guarantee that the corresponding terminating state probability distribution $P_{F}^{\top}$ approaches the target distribution $P^{\star}$?

\paragraph{Estimation of the partition function} Although convergence guarantees probably exist for other flow matching conditions, throughout this section we will consider the trajectory balance loss \cref{eq:trajectory-balance-loss} as a representative example, since it is a popular objective to train generative flow networks. One advantage of this loss is that it provides an approximation of the partition function with the learned scalar $Z_{\phi}$, in addition to an approximate sampler for $P^{\star}$. However, it has been shown empirically that $Z_{\phi}$ alone may yield an unreliable estimation of the partition function, which may be detrimental if it is used for model comparison via the Bayes factor \citep{morey2016bayesfactor}. The following proposition offers a more precise relation between the true partition function $Z$, its learned counterpart $Z_{\phi}$, and the residuals $\Delta_{\mathrm{TB}}(\tau;\phi)$.  %

\begin{proposition}[Estimation of the partition function]
    \label{prop:relation-log-partition-function}
    Let $\phi$ be the parameters of the forward transition probabilities $P_{F}^{\phi}$, the backward transition probabilities $P_{B}^{\phi}$, and the total flow $Z_{\phi}$ in the trajectory balance loss \cref{eq:trajectory-balance-loss}. The log-partition function $\log Z$ of the target distribution \cref{eq:target-distribution-gflownet-reward} is related to $\log Z_{\phi}$ via
    \begin{equation}
        \log Z = \log Z_{\phi} + \log \Big(\E_{\tau\sim P_{F}^{\phi}}\big[\exp(-\Delta_{\mathrm{TB}}(\tau;\phi))\big]\Big).
        \label{eq:relation-log-partition-function}
    \end{equation}
\end{proposition}

\begin{proof}
    Based on \cref{eq:trajectory-balance-loss}, for some complete trajectory $\tau = (s_{0}, s_{1}, \ldots, s_{T}, \terminal)$, we can find an expression for the product of the reward at the terminating state $s_{T}$ and the backward probability of $\tau$ as a function of the residual $\Delta_{\mathrm{TB}}(\tau;\phi)$
    \begin{equation}
        R(s_{T})P_{B}^{\phi}(\tau\mid s_{T}) = R(s_{T})\prod_{t=1}^{T}P_{B}^{\phi}(s_{t-1}\mid s_{t}) = Z_{\phi}\exp(-\Delta_{\mathrm{TB}}(\tau;\phi))\prod_{t=0}^{T}P_{F}^{\phi}(s_{t+1}\mid s_{t}),
    \end{equation}
    where we used the notation $P_{B}^{\phi}(\tau\mid s_{T}) = \prod_{t=1}^{T}P_{B}^{\phi}(s_{t-1}\mid s_{t})$ introduced in \cref{sec:backward-transition-probabilities}. Since $P_{B}^{\phi}(\cdot \mid x)$ is a distribution over the complete trajectories terminating at $x$, the reward $R(x)$ is given by
    \begin{equation}
        R(x) = \sum_{\tau: x\rightarrow \terminal\in\tau}R(x)P_{B}^{\phi}(\tau\mid x) = Z_{\phi}\sum_{\tau: x\rightarrow \terminal \in \tau}P_{F}^{\phi}(\tau)\exp(-\Delta_{\mathrm{TB}}(\tau;\phi))
    \end{equation}
    By definition of the partition function $Z$, and using the fact that $P_{F}^{\phi}$ induces a probability distribution over all the complete trajectories by \cref{lem:PF-distribution-suffix}
    {\allowdisplaybreaks%
    \begin{align}
        Z &= \sum_{x\in\gX}R(x)\\
        &= Z_{\phi}\sum_{x\in\gX}\sum_{\tau: x\rightarrow \terminal\in\tau}P_{F}^{\phi}(\tau)\exp(-\Delta_{\mathrm{TB}}(\tau;\phi))\\
        &= Z_{\phi}\sum_{\tau\in\gT}P_{F}^{\phi}(\tau)\exp(-\Delta_{\mathrm{TB}}(\tau;\phi))\\
        &= Z_{\phi}\E_{\tau\sim P_{F}^{\phi}}\big[\exp(-\Delta_{\mathrm{TB}}(\tau;\phi))\big].
    \end{align}}%
    We can conclude by taking the $\log$ of the equality above.
\end{proof}

This result is reminiscent of the estimation of the log-marginal likelihood via a variational lower-bound in PhyloGFN \citep{zhou2024phylogfn}, obtained by applying Jensen's inequality to \cref{eq:relation-log-partition-function}. Using \cref{lem:bound-log-expectation-exp}, this gives us an upper-bound this time on the error between the learned $\log Z_{\phi}$ and the true log-partition function $\log Z$ that is controlled by the maximal residual over all possible complete trajectories
\begin{equation}
    |\log Z_{\phi} - \log Z| \leq \max_{\tau\in\gT}|\Delta_{\mathrm{TB}}(\tau;\phi)|.
    \label{eq:bound-log-partition-function}
\end{equation}

\paragraph{Convergence of the terminating state distribution} While \cref{eq:bound-log-partition-function} gives us a guarantee on the convergence of $\log Z_{\phi}$ towards the target log-partition function as the trajectory balance loss gets minimized, we would like to also have stronger guarantees on the convergence of the terminating state probability $P_{F}^{\phi\top}(x)$ itself towards $P^{\star}(x)$. Using a similar argument, the following proposition shows that the difference between the log-terminating state probability and the target log-probability is directly controlled by the residual $\Delta_{\mathrm{TB}}(\tau;\phi)$ being minimized. This ensures that as this residual gets smaller, the approximation becomes more accurate at a particular $x$.
\begin{proposition}
    \label{prop:bound-difference-log-probs}
    Let $\phi$ be the parameters of the forward transition probabilities $P_{F}^{\phi}$, the backward transition probabilities $P_{B}^{\phi}$, and the total flow $Z_{\phi}$ in the trajectory balance loss \cref{eq:trajectory-balance-loss}. For any $x\in\gX$, the error between the log-terminating state probability associated with $P_{F}^{\phi}$ and the target log-probability is bounded by
    \begin{equation}
        |\log P_{F}^{\phi\top}(x) - \log P^{\star}(x)| \leq \max_{\tau\in\gT}|\Delta_{\mathrm{TB}}(\tau;\phi)| + \max_{\tau: s_{0}\rightsquigarrow x}|\Delta_{\mathrm{TB}}(\tau;\phi)|.
        \label{eq:bound-difference-log-probs}
    \end{equation}
\end{proposition}

\begin{proof}
    We can use \cref{eq:trajectory-balance-loss} to find an expression for $P_{F}^{\phi}(\tau)$ as a function of the residual $\Delta_{\mathrm{TB}}(\tau;\phi)$:
    \begin{equation}
        \prod_{t=0}^{T}P_{F}^{\phi}(s_{t+1}\mid s_{t}) = \exp(\Delta_{\mathrm{TB}}(\tau;\phi))\frac{R(s_{T})}{Z_{\phi}}\prod_{t=1}^{T}P_{B}^{\phi}(s_{t-1}\mid s_{t}).
        \label{eq:proof-bound-difference-log-probs-3}
    \end{equation}
    For any $x\in\gX$, using the notation $P_{B}^{\phi}(\tau\mid x) = \prod_{t=1}^{T}P_{B}^{\phi}(s_{t-1}\mid s_{t})$ introduced in \cref{sec:backward-transition-probabilities} (with $s_{T} = x$), which is a properly defined distribution over the complete trajectories terminating in $x$, and by \cref{def:terminating-state-probability} of the terminating state probability distribution associated with $P_{F}^{\phi}$, we get
    {\allowdisplaybreaks%
    \begin{align}
        P_{F}^{\phi\top}(x) &= \sum_{\tau: s_{0}\rightsquigarrow x}\prod_{t=0}^{T_{\tau}}P_{F}^{\phi}(s_{t+1}\mid s_{t})\\
        &= \frac{R(x)}{Z_{\phi}}\sum_{\tau: s_{0}\rightsquigarrow x}\exp(\Delta_{\mathrm{TB}}(\tau;\phi))P_{B}^{\phi}(\tau\mid x)\\
        &= \frac{R(x)}{Z}\frac{Z}{Z_{\phi}}\E_{\tau\sim P_{B}^{\phi}(\cdot\mid x)}\big[\exp(\Delta_{\mathrm{TB}}(\tau;\phi))\big]\\
        &= P^{\star}(x)\frac{Z}{Z_{\phi}}\E_{\tau\sim P_{B}^{\phi}(\cdot\mid x)}\big[\exp(\Delta_{\mathrm{TB}}(\tau;\phi))\big]\label{eq:proof-bound-difference-log-probs-1}
    \end{align}}%
    Although the normalization constant $Z$ of the target distribution is still unknown, we can fortunately write it as a function of the residual as well thanks to \cref{prop:relation-log-partition-function}. Combining it with \cref{eq:proof-bound-difference-log-probs-1}, we get an expression of the difference in log-probabilities as a function of $\Delta_{\mathrm{TB}}(\tau;\phi)$ only, where the expectations over complete trajectories are respectively taken wrt. $P_{F}^{\phi}$ and $P_{B}^{\phi}(\cdot \mid x)$
    \begin{equation}
        \log P_{F}^{\phi\top}(x) - \log P^{\star}(x) = \log \Big(\E_{\tau\sim P_{F}^{\phi}}\big[\exp(-\Delta_{\mathrm{TB}}(\tau;\phi)\big]\Big) + \log \Big(\E_{\tau\sim P_{B}^{\phi}(\cdot\mid x)}\big[\exp(\Delta_{\mathrm{TB}}(\tau;\phi))\big]\Big).
        \label{eq:proof-bound-difference-log-probs-4}
    \end{equation}
    Using the triangle inequality, we can conclude by applying \cref{lem:bound-log-expectation-exp} to both terms of the RHS
    {\allowdisplaybreaks%
    \begin{align}
        \left|\log \Big(\E_{\tau\sim P_{F}^{\phi}}\big[\exp(-\Delta_{\mathrm{TB}}(\tau;\phi))\big]\Big)\right| &\leq \max_{\tau\in\gT}|\Delta_{\mathrm{TB}}(\tau;\phi)|\\
        \left|\log \Big(\E_{\tau\sim P_{B}^{\phi}(\cdot\mid x)}\big[\exp(\Delta_{\mathrm{TB}}(\tau;\phi))\big]\Big)\right| &\leq \max_{\tau: s_{0}\rightsquigarrow x}|\Delta_{\mathrm{TB}}(\tau;\phi)|.
    \end{align}}%
\end{proof}
As an immediate consequence of this proposition, we can get a simpler (albeit looser, because it becomes independent of the terminating state $x$) bound of the form
\begin{equation}
    |\log P_{F}^{\phi\top}(x) - \log P^{\star}(x)| \leq 2\max_{\tau\in\gT}|\Delta_{\mathrm{TB}}(\tau;\phi)|.
    \label{eq:bound-difference-log-probs-loose}
\end{equation}
The term $\max_{\tau\in\gT}|\Delta_{\mathrm{TB}}(\tau;\phi)|$ in the bound of \cref{prop:bound-kl-divergence-terminating-state-prob}, independent of $x$, has an intuitive interpretation. We could allocate all the training capacity to learn parameters $\phi$ so that the trajectory balance conditions match almost perfectly for all the trajectories leading to a certain terminating state $x$ only (\eg using a behavior policy that focuses almost exclusively on those trajectories). In that case, we would have $\max_{\tau: s_{0}\rightsquigarrow x}|\Delta_{\mathrm{TB}}(\tau; \phi)| \approx 0$. However, that does not mean that we perfectly recovered $\log P^{\star}(x)$, because we have to take into account all the other terminating states as well in the computation of its normalization constant; this is precisely what the term $\max_{\tau\in\gT}|\Delta_{\mathrm{TB}}(\tau;\phi)|$ controls for.

Instead of considering what happens at the level of a single $x$, we can also get more global guarantees on the divergence between these two distributions, similar to what typical variational methods would minimize. The following proposition gives a bound on their KL-divergence that still depends exclusively on $\Delta_{\mathrm{TB}}(\tau;\phi)$.
\begin{proposition}
    \label{prop:bound-kl-divergence-terminating-state-prob}
    Let $\phi$ be the parameters of the forward transition probabilities $P_{F}^{\phi}$, the backward transition probabilities $P_{B}^{\phi}$, and the total flow $Z_{\phi}$ in the trajectory balance loss \cref{eq:trajectory-balance-loss}. The KL-divergence between the terminating state probability distribution associated with $P_{F}^{\phi}$ and the target distribution $P^{\star}$ is bounded by
    \begin{equation}
        \KL\big(P_{F}^{\phi\top}(x)\,\|\,P^{\star}(x)\big) \leq \E_{x\sim P_{F}^{\phi\top}}\Big[\max_{\tau: s_{0}\rightsquigarrow x}\Delta_{\mathrm{TB}}(\tau;\phi)\Big] - \min_{\tau\in\gT}\Delta_{\mathrm{TB}}(\tau;\phi).
        \label{eq:bound-kl-divergence-terminating-state-prob}
    \end{equation}
\end{proposition}

\begin{proof}
    This is a direct consequence of \cref{eq:proof-bound-difference-log-probs-4} in the proof of \cref{prop:bound-difference-log-probs} above:
    \begin{align}
        \KL&\big(P_{F}^{\phi\top}(x)\,\|\,P^{\star}(x)\big) = \E_{x\sim P_{F}^{\phi\top}}\big[\log P_{F}^{\phi\top}(x) - \log P^{\star}(x)\big]\\
        &= \log\Big(\E_{\tau\sim P_{F}^{\phi}}\big[\exp(-\Delta_{\mathrm{TB}}(\tau;\phi))\big]\Big) + \E_{x\sim P_{F}^{\phi\top}}\Big[\log \Big(\E_{\tau\sim P_{B}^{\phi}(\cdot\mid x)}\big[\exp(\Delta_{\mathrm{TB}}(\tau;\phi))\big]\Big)\Big]\\
        &\leq \Big[\max_{\tau\in\gT} -\Delta_{\mathrm{TB}}(\tau;\phi)\Big] + \E_{x\sim P_{F}^{\phi\top}}\Big[\max_{\tau: s_{0}\rightsquigarrow x}\Delta_{\mathrm{TB}}(\tau; \phi)\Big] \label{eq:proof-bound-kl-divergence-terminating-state-prob-1}\\
        &= \E_{x\sim P_{F}^{\phi\top}}\Big[\max_{\tau: s_{0}\rightsquigarrow x}\Delta_{\mathrm{TB}}(\tau; \phi)\Big] - \min_{\tau\in\gT}\Delta_{\mathrm{TB}}(\tau;\phi),
    \end{align}
    where the inequality in \cref{eq:proof-bound-kl-divergence-terminating-state-prob-1} is a consequence of \cref{eq:proof-bound-log-expectation-exp-1} in the proof of \cref{lem:bound-log-expectation-exp}.
\end{proof}

We will see an alternative bound on this KL-divergence in \cref{prop:data-processing-inequality-gflownets}. Because the bounds of \cref{prop:bound-difference-log-probs,prop:bound-kl-divergence-terminating-state-prob} both involve the maximum error being incurred for any complete trajectory, they are often impractical for precisely quantifying how close the model learned with the GFlowNet is to the target. Nevertheless, the objective of these results is to confirm the intuition that the approximation found by the GFlowNet gets better as the trajectory balance loss is minimized.

\section{GFlowNets \& variational inference}
\label{sec:gflownets-variational-inference}

We saw in the previous section that generative flow networks are closely related to variational inference, as they treat the problem of inference (via sampling) as an optimization problem (via the minimization of a flow matching loss in \cref{sec:flow-matching-losses}). We will show in this section that this connection is more than superficial, and we can actually represent a variational model as a GFlowNet and conversely \citep{malkin2023gfnhvi}. Due to the sequential nature of generation in a GFlowNet, the type of variational model will be slightly different from the ones we presented in \cref{sec:variational-inference}, and we will have to consider a \emph{hierarchy} of latent variables (corresponding to the intermediate states of a GFlowNet).

\subsection{Hierarchical variational inference}
\label{sec:hierarchical-variational-inference}\index{Variational inference!Hierarchical}
A \emph{hierarchical variational model}\glsadd{hvi} is a generalization of variational inference introduced in \cref{sec:variational-inference} that extends to a hierarchy of $T$ latent variables $z_{1}, \ldots, z_{T}$. In particular in the Markovian case, the generative process is given by the following Markov chain
\begin{equation}
    P_{\psi}(z_{1}, \ldots, z_{T}) = P^{\star}(z_{T})\prod_{t=1}^{T-1}P_{\psi}(z_{t}\mid z_{t+1})
    \label{eq:hvi-generative-model}
\end{equation}
This generative model depends on some parameters $\psi$, that may be learned, and its marginal $P^{\star}(z_{T})$ is a Gibbs distribution. Unlike a standard VAE though, where the objective is to find an approximation of the posterior \citep{kingma2013vae}, here the goal is to find an \emph{inference model} $Q_{\phi}(z_{1}, \ldots, z_{T})$ whose marginal approximates the (equally intractable) marginal $P_{\psi}(z_{T}) = P^{\star}(z_{T})$. The form of inference model we consider here is also a Markov chain, operating forward in time \citep{bachman2015datagensequential}
\begin{equation}
    Q_{\phi}(z_{1}, \ldots, z_{T}) = Q_{\phi}(z_{1})\prod_{t=1}^{T-1}Q_{\phi}(z_{t+1}\mid z_{t}).
    \label{eq:hvi-inference-model}
\end{equation}
Since our objective is to approximate the marginal distribution $P_{\psi}(z_{T})$ with $Q_{\phi}(z_{T})$, we can use the data processing inequality in \cref{prop:data-processing-inequality} in order to bound the KL divergence between these marginals by the KL divergence between their respective joint distributions over all latent variables
\begin{equation}
    \KL\big(Q_{\phi}(z_{T})\,\|\,P^{\star}(z_{T})\big) = \KL\big(Q_{\phi}(z_{T})\,\|\,P_{\psi}(z_{T})\big) \leq \KL\big(Q_{\phi}(z_{1:T})\,\|\,P_{\psi}(z_{1:T})\big).
    \label{eq:kl-marginal-kl-joint-gfn-hvi}
\end{equation}
A natural choice of objective function is therefore to use the RHS of the inequality above, in order to find parameters $(\phi, \psi)$ such that the marginal $Q_{\phi}(z_{T})$ is an accurate approximation of $P^{\star}(z_{T})$
\begin{equation}
    \gL_{\mathrm{HVI}}(\phi, \psi) = \KL\big(Q_{\phi}(z_{1:T})\,\|\,P_{\psi}(z_{1:T})\big).
\end{equation}
Interestingly, while this objective depends on the target distribution $P^{\star}(z_{T})$ via the generative model in \cref{eq:hvi-generative-model} which is unknown, its intractable partition function $Z$ disappears when computing the gradients $\nabla \gL_{\mathrm{HVI}}(\phi, \psi)$. Thus, it is possible to minimize this objective using gradient methods, based on samples from the inference model $Q_{\phi}(z_{1:T})$, despite knowing $P^{\star}(z_{T})$ only up to a normalization constant. This is reminiscent of the intractable evidence a priori appearing in \cref{eq:derivation-elbo-from-kl}.

\subsection{Construction of a GFlowNet from a hierarchical variational model}
\label{sec:gflownet-from-hvm}
Sampling a sequence $(z_{1}, \ldots, z_{T})$ from a hierarchical model is closely related to how complete trajectories are sampled in a generative flow network. To see this, consider a pointed DAG $\gG = (\widebar{\gS}, \gA)$ whose states are organized in ``layers'': the states at a distance $t$ of the initial state $s_{0}$ correspond to all the possible values the latent variable $z_{t}$ may take in the hierarchical model, and all the states at distance $t$ are connected to those at distance $t+1$ through an edge in $\gA$. All the states corresponding to the values the final latent $z_{T}$ may take are then connected to the terminal state $\terminal$ (\ie they are terminating states).

On top of this pointed DAG, we can also define forward and backward transition probabilities based on the generative and inference models in \cref{eq:hvi-generative-model} \& \cref{eq:hvi-inference-model}. On the one hand, the backward transition probabilities are given by the generative model
\begin{equation}
    P_{B}^{\psi}(z_{t}\mid z_{t+1}) = P_{\psi}(z_{t}\mid z_{t+1}),
\end{equation}
where the values $z_{t}$ \& $z_{t+1}$ must be interpreted as their corresponding states in $\gG$. Moreover, based on the structure of the pointed DAG, it is clear that $P_{B}^{\psi}(s_{0}\mid z_{1}) = 1$, no matter the state $z_{1}$ in the first layer. On the other hand, the forward transition probabilities are given by the inference model
\begin{align}
    P_{F}^{\phi}(z_{t+1}\mid z_{t}) &= Q_{\phi}(z_{t+1}\mid z_{t}) && \textrm{and} & P_{F}^{\phi}(z_{1}\mid s_{0}) = Q_{\phi}(z_{1}),
\end{align}
where again it is clear that $P_{F}^{\phi}(\terminal\mid z_{T}) = 1$ based on the structure of $\gG$. Rewriting the hierarchical model in terms of the forward and backward transition probabilities of this GFlowNet, we have for any complete trajectory $\tau = (s_{0}, s_{1}, \ldots, s_{T}, \terminal)$
\begin{align}
    P_{F}^{\phi}(\tau) &= \prod_{t=0}^{T}P_{F}^{\phi}(s_{t+1}\mid s_{t}) &&& P_{B}^{\psi}(\tau) &= P^{\star}(x_{\tau})P_{B}^{\psi}(\tau\mid x_{\tau}) = P^{\star}(s_{T})\prod_{t=1}^{T}P_{B}^{\psi}(s_{t-1}\mid s_{t}),
    \label{eq:data-processing-inequality-gflownets-2}
\end{align}
where $s_{T+1} = \terminal$. We use the notation $x_{\tau} \equiv s_{T}$ to denote the terminating state reached by $\tau$. Similar to how we transformed any hierarchical model into a GFlowNet, we showed that any GFlowNet can conversely be transformed into an equivalent hierarchical model as well \citep{malkin2023gfnhvi}. The following proposition is the equivalent of \cref{eq:kl-marginal-kl-joint-gfn-hvi}, this time working exclusively with quantities specific to GFlowNets and generalized to any $f$-divergence.

\begin{proposition}
    \label{prop:data-processing-inequality-gflownets}
    Let $f$ be a convex function. Let $P_{F}^{\phi}$ be a forward transition probability distribution parametrized by $\phi$, and $P_{B}^{\psi}$ be a  backward transition probability distribution parametrized by $\psi$. The following inequality holds:
    \begin{equation}
        D_{f}\big(P^{\star}(x)\,\|\,P_{F}^{\phi\top}(x)\big) \leq D_{f}\big(P_{B}^{\psi}(\tau)\,\|\,P_{F}^{\phi}(\tau)\big),
        \label{eq:data-processing-inequality-gflownets}
    \end{equation}
    where the RHS involves distributions over complete trajectories defined in \cref{eq:data-processing-inequality-gflownets-2}.
\end{proposition}

\begin{proof}
    This is a direct consequence of the data processing inequality in \cref{prop:data-processing-inequality}, bounding the $f$-divergence between marginal distributions (here, marginalized over trajectories leading to a terminating state) by the $f$-divergence between the joint distributions (here, the distributions over complete trajectories in \cref{eq:data-processing-inequality-gflownets-2}). Indeed, by \cref{def:terminating-state-probability}, $P_{F}^{\phi\top}(x)$ is the marginal of $P_{F}^{\phi}$ (as a distribution over complete trajectories) over the trajectories leading to $x\in\gX$. Similarly, the marginal of $P_{B}^{\psi}$ is
    \begin{equation}
        \sum_{\tau: s_{0}\rightsquigarrow x}P_{B}^{\psi}(\tau) = P^{\star}(x)\sum_{\tau: s_{0}\rightsquigarrow x}P_{B}^{\psi}(\tau\mid x) = P^{\star}(x),
    \end{equation}
    where we used \cref{lem:PB-distribution-prefix} to conclude
\end{proof}

It is important to reiterate that the distributions involved in the $f$-divergence on the RHS of \cref{eq:data-processing-inequality-gflownets} are distributions over complete trajectories, whereas those on the LHS are distributions over terminating states only. A particular case of this is using the convex function $f(x) = -\log x$, where we get a bound on the KL divergence between the terminating state distribution and the target $P^{\star}$
\begin{equation}
    \KL\big(P_{F}^{\phi\top}(x)\,\|\,P^{\star}(x)\big) \leq \KL\big(P_{F}^{\phi}(\tau)\,\|\,P_{B}^{\psi}(\tau)\big).
    \label{eq:bound-kl-divergence-terminating-state-prob-2}
\end{equation}
One way to interpret this inequality is that in order to approximate the target distribution, we must find a pair of distribution $(P_{F}^{\phi}, P_{B}^{\psi})$ that are as close as possible to one another (as distributions over complete trajectories). This notion of consistency between distributions over complete trajectories, either defined in terms of forward probabilities or backward probabilities, reminds us of the discussion in \cref{sec:trajectory-balance-condition} about the trajectory balance condition imposing an equality between those distributions.

\subsection{Analysis of the gradients}
\label{sec:analysis-gradients-gfn-hvi}
Recall that the trajectory balance loss is defined as $\gL_{\mathrm{TB}}(\phi, \psi, \omega) = \frac{1}{2}\E_{\pi_{b}}\big[\Delta_{\mathrm{TB}}^{2}(\tau; \psi, \psi, \omega)\big]$, where for a complete trajectory $\tau = (s_{0}, s_{1}, \ldots, s_{T}, \terminal)$,
\begin{equation}
    \Delta_{\mathrm{TB}}(\tau; \phi, \psi, \omega) = \log \frac{Z_{\omega}P_{F}^{\phi}(\tau)}{R(s_{T})P_{B}^{\psi}(\tau\mid s_{T})} = \log \frac{Z_{\omega}\prod_{t=0}^{T}P_{F}^{\phi}(s_{t+1}\mid s_{t})}{R(s_{T})\prod_{t=1}^{T}P_{B}^{\psi}(s_{t-1}\mid s_{t})}.
    \label{eq:residual-trajectory-balance-gfn-vi}
\end{equation}
Here we explicitly separate the parameters of the forward transition probabilities $\phi$, those of the backward transition probabilities $\psi$, and those of the approximation of the partition function $\omega$ to highlight the updates of each individual parameter. We saw in \cref{prop:bound-kl-divergence-terminating-state-prob} that this loss bounds the KL divergence between the corresponding terminating state distribution $P_{F}^{\phi\top}$ and the unknown target distribution $P^{\star}$. This is similar to the role played by the hierarchical variational inference objective $\gL_{\mathrm{HVI}}(\phi, \psi) = \KL\big(P_{F}^{\phi}(\tau)\,\|\,P_{B}^{\psi}(\tau)\big)$ introduced in the previous section. %

Is this connection between $\gL_{\mathrm{TB}}(\phi, \psi, \omega)$ and $\gL_{\mathrm{HVI}}(\phi, \psi)$ only superficial? It turns out that the hierarchical variational inference objective can be viewed as a surrogate loss for the trajectory balance loss. In other words, instead of updating the parameters of $P_{F}^{\phi}$ with gradient descent based on the trajectory balance loss, we could use the gradient of $\gL_{\mathrm{HVI}}$ (wrt. $\phi$).

\begin{proposition}
    \label{prop:gfn-vi-equality-gradients-tb}
    Let $P_{F}^{\phi}$ be a forward transition probability distribution parametrized by $\phi$, and $P_{B}^{\psi}$ be a  backward transition probability distribution parametrized by $\psi$. The gradient of the hierarchical variational inference objective is related to the gradient of the trajectory balance loss via
    \begin{equation}
        \nabla_{\phi}\KL\big(P_{F}^{\phi}(\tau)\,\|\,P_{B}^{\psi}(\tau)\big) = \frac{1}{2}\E_{\tau\sim P_{F}^{\phi}}\big[\nabla_{\phi}\Delta_{\mathrm{TB}}^{2}(\tau;\phi, \psi, \omega)\big].\label{eq:gfn-vi-equality-gradients-tb-1}
    \end{equation}
\end{proposition}

\begin{proof}
    First, recall that $P_{B}^{\psi}$ as a distribution over complete trajectories is defined for any complete trajectory $\tau = (s_{0}, s_{1}, \ldots, s_{T}, \terminal)$ by
    \begin{equation}
        P_{B}^{\psi}(\tau) = P^{\star}(s_{T})P_{B}^{\psi}(\tau\mid s_{T}) = \frac{R(s_{T})}{Z}P_{B}^{\psi}(\tau\mid s_{T}).
    \end{equation}
    Therefore, we can rewrite the difference in log-probabilities between $P_{F}^{\phi}(\tau)$ and $P_{B}^{\psi}(\tau)$ in terms of the residual of the trajectory balance condition \cref{eq:residual-trajectory-balance-gfn-vi}
    \begin{equation}
        \log \frac{P_{F}^{\phi}(\tau)}{P_{B}^{\psi}(\tau)} = \Delta_{\mathrm{TB}}(\tau; \phi, \psi, \omega) + \log Z - \log Z_{\omega}.
    \end{equation}
    To prove \cref{eq:gfn-vi-equality-gradients-tb-1}, we can write the KL-divergence as a function of the residual as well:
    \begin{equation}
        \KL\big(P_{F}^{\phi}(\tau)\,\|\,P_{B}^{\psi}(\tau)\big) = \E_{\tau\sim P_{F}^{\phi}}\left[\log \frac{P_{F}^{\phi}(\tau)}{P_{B}^{\psi}(\tau)}\right] = \E_{\tau \sim P_{F}^{\phi}}\big[\Delta_{\mathrm{TB}}(\tau)\big] + \log Z - \log Z_{\omega},
    \end{equation}
    where we compacted the notation for $\Delta_{\mathrm{TB}}(\tau)$, implicitly depending on $(\phi, \psi, \omega)$. Taking the gradient of this equation wrt. $\phi$, we get
    \begin{equation}
        \nabla_{\phi}\KL\big(P_{F}^{\phi}(\tau)\,\|\,P_{B}^{\psi}(\tau)\big) = \nabla_{\phi}\E_{\tau \sim P_{F}^{\phi}}\big[\Delta_{\mathrm{TB}}(\tau)\big] = \E_{\tau \sim P_{F}^{\phi}}\big[\nabla_{\phi}\log P_{F}^{\phi}(\tau)\Delta_{\mathrm{TB}}(\tau) + \nabla_{\phi}\Delta_{\mathrm{TB}}(\tau)\big].
        \label{eq:gfn-vi-equality-gradients-tb-proof-1}
    \end{equation}
    Moreover, the second term of \cref{eq:gfn-vi-equality-gradients-tb-proof-1} vanishes, since $P_{F}^{\phi}(\tau)$ is a properly defined distribution over complete trajectories, and
    \begin{align}
        \E_{\tau \sim P_{F}^{\phi}}\big[\nabla_{\phi}\Delta_{\mathrm{TB}}(\tau)\big] &= \E_{\tau \sim P_{F}^{\phi}}\big[\nabla_{\phi} \log P_{F}^{\phi}(\tau)\big]\\
        &= \sum_{\tau\in\gT}P_{F}^{\phi}(\tau)\nabla_{\phi}\log P_{F}^{\phi}(\tau) = \nabla_{\phi}\sum_{\tau\in\gT}P_{F}^{\phi}(\tau) = 0.\label{eq:gfn-vi-equality-gradients-tb-proof-2}
    \end{align}
    Based on the fact that $\nabla_{\phi}\Delta_{\mathrm{TB}}(\tau) = \nabla_{\phi}\log P_{F}^{\phi}(\tau)$, we can conclude
    \begin{equation}
        \nabla_{\phi}\KL\big(P_{F}^{\phi}(\tau)\,\|\,P_{B}^{\psi}(\tau)\big) = \E_{\tau \sim P_{F}^{\phi}}\big[\nabla_{\phi}\Delta_{\mathrm{TB}}(\tau)\Delta_{\mathrm{TB}}(\tau)\big] = \frac{1}{2}\E_{\tau \sim P_{F}^{\phi}}\big[\nabla_{\phi}\Delta_{\mathrm{TB}}^{2}(\tau)\big].
    \end{equation}
\end{proof}
It is important to observe that this equality holds only when the trajectory balance loss is \emph{on-policy}, meaning that we are using the same policy $\pi_{b} \equiv P_{F}^{\phi}$ for sampling trajectories and evaluating $\Delta_{\mathrm{TB}}(\tau)$. However since this is on-policy, the RHS of \cref{eq:gfn-vi-equality-gradients-tb-proof-1} does not correspond exactly to $\nabla_{\phi}\gL_{\mathrm{TB}}$. Indeed, there is an additional term appearing from differentiating through the sampling process
{\allowdisplaybreaks%
\begin{align}
    \nabla_{\phi}\gL_{\mathrm{TB}}(\phi, \psi, \omega) &= \nabla_{\phi}\left[\frac{1}{2}\E_{\tau \sim P_{F}^{\phi}}\big[\Delta_{\mathrm{TB}}^{2}(\tau; \phi, \psi, \omega)\big]\right]\\
    &= \frac{1}{2}\E_{P_{F}^{\phi}}\big[\nabla_{\phi}\log P_{F}^{\phi}(\tau)\Delta_{\mathrm{TB}}^{2}(\tau) + \nabla_{\phi}\Delta_{\mathrm{TB}}^{2}(\tau)\big]\\
    &= \nabla_{\phi}\gL_{\mathrm{HVI}}(\phi, \psi) + \frac{1}{2}\E_{P_{F}^{\phi}}\big[\nabla_{\phi}\log P_{F}^{\phi}(\tau)\Delta_{\mathrm{TB}}^{2}(\tau)\big].\label{eq:gradients-tb-hvi-pf-offset}
\end{align}}%
In practice, when considering the trajectory balance loss on-policy (or any other flow matching loss for that matter), it is standard to completely ignore the second term of \cref{eq:gradients-tb-hvi-pf-offset} as we saw in \cref{sec:off-policy-training}, effectively treating $\gL_{\mathrm{TB}}$ as if it were off-policy; see \cref{sec:jsp-gfn-behavior-policy} for example.

We similarly showed that the gradient of the trajectory balance loss, this time wrt. the parameters $\psi$ of the backward transition probabilities, is related to the gradient of the \emph{forward KL} objective \citep{malkin2023gfnhvi}
\begin{equation}
    \nabla_{\psi}\KL\big(P_{B}^{\psi}\,\|\,P_{F}^{\phi}\big) = \frac{1}{2}\E_{\tau\sim P_{B}^{\psi}}\big[\nabla_{\psi}\Delta_{\mathrm{TB}}^{2}(\tau;\phi, \psi, \omega)\big].\label{eq:gfn-vi-equality-gradients-tb-2}
\end{equation}
This result, along with \cref{prop:gfn-vi-equality-gradients-tb} may be extended to sub-trajectories \citep{malkin2023gfnhvi}. Unlike the gradient wrt.~$\phi$ though, the expectations appearing on both sides of this equation are taken over $P_{B}^{\psi}$ itself, which is intractable since it depends on $P^{\star}$ (see \cref{eq:data-processing-inequality-gflownets-2}). Although it is common practice to resort to importance sampling in those cases to get an unbiased estimator off-policy (\eg based on trajectories sampled with $P_{F}^{\phi}$; off-policy here wrt. $P_{B}^{\psi}$), this is impractical since $P_{B}^{\psi}(\tau)$ cannot even be evaluated (a necessary condition to apply importance sampling). The following proposition relates the gradient of the trajectory balance loss with the gradient of a combination of the reverse KL and a pseudo $f$-divergence.

\begin{proposition}
    \label{prop:on-policy-gradient-pB-gfn-hvi}
    Let $P_{F}^{\phi}$ be a forward transition probability distribution parametrized by $\phi$, and $P_{B}^{\psi}$ be a backward transition probability distribution parametrized by $\psi$. The gradient of the (on-policy) trajectory balance loss wrt.~$\psi$ is equal to
    \begin{equation}
        \nabla_{\psi}\left[\frac{1}{2}\E_{\tau \sim P_{F}^{\phi}}\big[\Delta_{\mathrm{TB}}^{2}(\tau; \phi, \psi, \omega)\big]\right] = \nabla_{\psi}\left[D_{0.5\log^{2}}\big(P_{B}^{\psi}\,\|\,P_{F}^{\phi}\big) + \log \frac{Z_{\omega}}{Z}\KL\big(P_{F}^{\phi}\,\|\,P_{B}^{\psi}\big)\right],
    \end{equation}
    where $D_{0.5\log^{2}}$ is the pseudo $f$-divergence, with the function $f(x) = 0.5\log^{2}(x)$.
\end{proposition}

\begin{proof}
    We will use the notation $c(\tau) = \log P_{F}^{\phi}(\tau) - \log P_{B}^{\psi}(\tau)$, which is implicitly a function of $\phi$ and $\psi$. Similarly, we use the notation $\Delta_{\mathrm{TB}}(\tau)$ to denote the residual of the trajectory balance condition, hiding its dependency on the parameters for clarity. It is clear that for any complete trajectory $\tau = (s_{0}, s_{1}, \ldots, s_{T}, \terminal)$, we have $\nabla_{\psi}\Delta_{\mathrm{TB}}(\tau) = - \nabla_{\psi}\log P_{B}^{\psi}(\tau \mid s_{T}) = -\nabla_{\psi}\log P_{B}^{\psi}(\tau) = \nabla_{\psi}c(\tau)$.
    {\allowdisplaybreaks%
    \begin{align}
        \nabla_{\psi}\left[\frac{1}{2}\E_{\tau \sim P_{F}^{\phi}}\big[\Delta_{\mathrm{TB}}^{2}(\tau)\big]\right] &= \E_{\tau \sim P_{F}^{\phi}}\big[\nabla_{\psi}\Delta_{\mathrm{TB}}(\tau)\Delta_{\mathrm{TB}}(\tau)\big] = \E_{\tau \sim P_{F}^{\phi}}\left[\nabla_{\psi}c(\tau) \log \frac{P_{F}^{\phi}(\tau)Z_{\omega}}{P_{B}^{\psi}(\tau)Z}\right]\\
        &= \nabla_{\psi}\left[\E_{\tau \sim P_{F}^{\phi}}\left[\frac{1}{2}c^{2}(\tau) + \log \frac{Z_{\omega}}{Z}c(\tau)\right]\right]\\
        &= \nabla_{\psi}\left[\frac{1}{2}\E_{\tau \sim P_{F}^{\phi}}\left[\left(\log \frac{P_{B}^{\psi}(\tau)}{P_{F}^{\phi}(\tau)}\right)^{2}\right] + \log \frac{Z_{\omega}}{Z}\E_{\tau\sim P_{F}^{\phi}}\left[\log \frac{P_{F}^{\phi}(\tau)}{P_{B}^{\psi}(\tau)}\right]\right]
    \end{align}}%
\end{proof}
This provides another surrogate loss for the (on-policy) trajectory balance loss, this time to update the parameters of $P_{B}^{\psi}$. However, this is equally impractical since the surrogate loss in question depends on the unknown partition function $Z$. \cref{tab:gfn-hvi-surrogate-losses} highlights the similarities between the surrogate losses considered for trajectory balance and optimizing the reverse KL objective directly, as is typically the case in the literature on variational inference. \citet{zhang2024directalignmentgfn} also used a similar connection between reverse KL and the detailed balance objective this time (\cref{thm:detailed-balance-proportional-reward}) to devise an on-policy algorithm inspired by \gls{ppo} \citep{schulman2017ppo}.

\begin{table}[t]
    \centering
    \caption[Summary of the surrogate losses for variational inference algorithms \& GFlowNets]{Summary of the surrogate losses for variational inference algorithms \& GFlowNets. This highlights that reverse KL \& (on-policy) trajectory balance optimize the same objective when updating the parameters $\phi$ of the inference model/forward transition probabilities.}
    \begin{tabular}{lccc}
    \toprule
     & \multicolumn{3}{c}{Surrogate loss}\\
     \cmidrule{2-4}
    Algorithm & $\phi \leftarrow \phi - \eta \nabla_{\phi}[-]$ && $\psi \leftarrow \psi - \eta \nabla_{\psi}[-]$ \\
    \midrule
    Reverse KL & $\KL\big(P_{F}^{\phi}\,\|\,P_{B}^{\psi}\big)$ && $\KL\big(P_{F}^{\phi}\,\|\,P_{B}^{\psi}\big)$\\
    Wake-sleep & $\KL\big(P_{B}^{\psi}\,\|\,P_{F}^{\phi}\big)$ && $\KL\big(P_{F}^{\phi}\,\|\,P_{B}^{\psi}\big)$\\
    \midrule 
    \multirow{2}{*}{(On-policy) Trajectory balance} & $\KL\big(P_{F}^{\phi}\,\|\,P_{B}^{\psi}\big)$ && \multirow{2}{*}{\cref{prop:on-policy-gradient-pB-gfn-hvi}} \\
    & \small \cref{prop:gfn-vi-equality-gradients-tb} && \\
    \bottomrule
    \end{tabular}
    \label{tab:gfn-hvi-surrogate-losses}
\end{table}

\subsection{Variance reduction of the gradient estimate}
\label{sec:variance-reduction-gradient-estimate}
To circumvent the complexity of studying the update of $P_{B}$ via the surrogate loss in \cref{prop:on-policy-gradient-pB-gfn-hvi}, we will consider it fixed in this section and focus on learning only $P_{F}^{\phi}$ (along with $Z_{\omega}$ in the case of trajectory balance). We saw that both the reverse KL and (on-policy) trajectory balance objectives correspond to the same surrogate loss when considering the update of $\phi$ (\cref{tab:gfn-hvi-surrogate-losses}). Recall that
\begin{equation}
    \gL_{\mathrm{HVI}}(\phi) = \KL\big(P_{F}^{\phi}(\tau)\,\|\,P_{B}(\tau)\big) = \E_{\tau \sim P_{F}^{\phi}}\left[\log \frac{ZP_{F}^{\phi}(\tau)}{R(x_{\tau})P_{B}(\tau\mid x_{\tau})}\right].
    \label{eq:hvi-loss}
\end{equation}
We will typically use gradient methods to update the parameters $\phi$ of the forward transition probabilities. In order to estimate the gradient of $\gL_{\mathrm{HVI}}$, we can take the \emph{score function estimator} \citep{mohamed2020mcestimationml}, also known as the REINFORCE estimator in the context of reinforcement learning \citep{williams1992reinforce}. This estimator is based on the following observation:
\begin{align}
    \nabla_{\phi}\gL_{\mathrm{HVI}}(\phi) &= \E_{\tau\sim P_{F}^{\phi}}\big[\nabla_{\phi}\log P_{F}^{\phi}(\tau) c(\tau;\phi)\big] && \textrm{where} & c(\tau;\phi) &= \log \frac{P_{F}^{\phi}(\tau)}{R(x_{\tau})P_{B}(\tau\mid x_{\tau})}.
\end{align}
The term $\E\big[\nabla_{\phi}c(\tau;\phi)\big] = 0$ that would have appeared in the expression of the gradient vanishes since $\E\big[\nabla_{\phi}\log P_{F}^{\phi}(\tau)\big] = 0$ (see also \cref{eq:gfn-vi-equality-gradients-tb-proof-2}). For similar reasons, it is standard to add a \emph{control variate} $b$ (also called a \emph{baseline} in reinforcement learning) independent of the trajectories $\tau$, in order to reduce the variance of the future estimator (although this is not guaranteed; \citealp{weaver2001optimalbaseline})
\begin{equation}
    \nabla_{\phi}\gL_{\mathrm{HVI}}(\phi) = \E_{\tau\sim P_{F}^{\phi}}\big[\nabla_{\phi}\log P_{F}^{\phi}(\tau) \big(c(\tau;\phi) - b\big)\big].
    \label{eq:gradient-hvi-baseline}
\end{equation}
While $b$ might be completely arbitrary, a typical choice is to use $b = \E\big[c(\tau;\phi)\big]$. We can estimate the gradient based on trajectories $\{\tau_{k}\}_{k=1}^{K}$ sampled iid. from the distribution $P_{F}^{\phi}(\tau)$
\begin{align}
    \nabla_{\phi}\gL_{\mathrm{HVI}}(\phi) &\approx \frac{1}{K}\sum_{k=1}^{K}\nabla_{\phi}\log P_{F}^{\phi}(\tau_{k})\big(c(\tau_{k};\phi) - b_{\mathrm{local}}\big) && \textrm{where} & b_{\mathrm{local}} &= \frac{1}{K}\sum_{k=1}^{K}c(\tau_{k};\phi),
    \label{eq:local-baseline-gfn-hvi}
\end{align}
and where the control variate is also estimated based on sample trajectories. The ``local'' aspect of this estimate is due to using only information from the current batch of trajectories $\{\tau_{k}\}_{k=1}^{K}$.

A more accurate approximation of $\E\big[c(\tau;\phi)\big]$ can be obtained by maintaining a running average of the values $c(\tau; \phi)$ across multiple batches, leading to a ``global'' baseline whose update is given by
\begin{equation}
    b_{\mathrm{global}} \leftarrow (1 - \eta) b_{\mathrm{global}} + \eta b_{\mathrm{local}}.
    \label{eq:global-baseline-gfn-hvi}
\end{equation}
Going back to the trajectory balance loss (on-policy), while we focused on the update of the parameters $\phi$ of the forward transition probabilities thus far, we can also study the gradient update of the parameters of $Z_{\omega}$. Suppose that we use the parametrization $Z_{\omega} = \exp(-\omega)$ (which is a standard practice; \citealp{malkin2022trajectorybalance}). The gradient of the trajectory balance objective is then given by
\begin{align}
    \nabla_{\omega}\gL_{\mathrm{TB}}(\phi, \omega) &= \nabla_{\omega}\left[\frac{1}{2}\E_{\tau\sim P_{F}^{\phi}}\Big[\big(\log Z_{\omega} + c(\tau;\phi)\big)^{2}\Big]\right]\\
    &= \frac{1}{2}\nabla_{\omega}\E_{\tau\sim P_{F}^{\phi}}\Big[\big(-\omega + c(\tau;\phi)\big)^{2}\Big] = \omega - \E_{\tau \sim P_{F}^{\phi}}\big[c(\tau;\phi)\big]
\end{align}
Using the notation $b_{\mathrm{local}}$ introduced in \cref{eq:local-baseline-gfn-hvi}, it is easy to see that the gradient wrt. the parameters $\omega$ can be approximated based on a batch of trajectories with $\nabla_{\omega}\gL_{\mathrm{TB}}(\phi, \omega) \approx \omega - b_{\mathrm{local}}$. With this approximation of the gradient, we can write the update rule of the parameters $\omega$ after one step of gradient descent with step size $\eta$ as
\begin{equation}
    \omega_{t+1} = \omega_{t} - \eta \nabla_{\omega}\gL_{\mathrm{TB}}(\phi, \omega_{t}) = \omega_{t} - \eta\big(\omega_{t} - b_{\mathrm{local}}\big) = (1 - \eta)\omega_{t} + \eta b_{\mathrm{local}},
\end{equation}
which corresponds exactly to the update rule of the global baseline in \cref{eq:global-baseline-gfn-hvi}. In other words, minimizing the trajectory balance loss (on-policy) coincides with minimizing the reverse KL loss with a control variate estimated with a moving average \cref{eq:global-baseline-gfn-hvi}. Since the gradient of $\gL_{\mathrm{HVI}}(\phi)$ is typically approximated using a local baseline \cref{eq:local-baseline-gfn-hvi} \citep{weber2015virl}, on-policy trajectory balance and reverse KL only differ in the way the control variate is being updated. We show this similarity between both methods empirically in \cref{sec:dag-gfn-effect-off-policy} on the problem of Bayesian structure learning, and more exhaustively in \citet{malkin2023gfnhvi}.

It is important to remember that this correspondence holds when we consider the trajectory balance loss \emph{on-policy}. One of the major strengths of the flow matching losses for GFlowNets is that they are also valid off-policy without any modification (see \cref{sec:off-policy-training}). This has the advantage of accelerating training since we can encourage exploration with a behavior policy $\pi_{b}$ decoupled from $P_{F}^{\phi}$ being learned. On the other hand, the reverse KL objective is inherently on-policy, and using trajectories sampled from another policy requires importance sampling in order to obtain an unbiased estimate of $\nabla_{\phi}\gL_{\mathrm{HVI}}(\phi)$
\begin{align}
    \nabla_{\phi}\gL_{\mathrm{HVI}}(\phi)&\approx \frac{1}{K}\sum_{k=1}^{K}w_{k}\nabla_{\phi}\log P_{F}^{\phi}(\widetilde{\tau}_{k})\big(c(\widetilde{\tau}_{k};\phi) - b_{\mathrm{local}}\big)&&\textrm{where}& w_{k} &= \frac{P_{F}^{\phi}(\widetilde{\tau}_{k})}{\pi_{b}(\widetilde{\tau}_{k})},
\end{align}
where $\{\widetilde{\tau}_{k}\}_{k=1}^{K}$ are trajectories sampled iid. from the behavior policy $\pi_{b}$. The variance of this estimator is usually high, despite the control variate, making the reverse KL less effective when training off-policy. The key observation here is that the trajectory balance objective (and its gradient) do \emph{not} require importance sampling, and therefore do not suffer from this high variance.

%% file: chapters/05_GFlowNets_MaxEnt_RL.tex
\chapter[GFlowNets \& Maximum Entropy Reinforcement Learning]{GFlowNets \& Maximum Entropy\\Reinforcement Learning}
\label{chap:gflownet-maxent-rl}

\begin{minipage}{\textwidth}
    \itshape
    This chapter contains material from the following paper:
    \begin{itemize}[noitemsep, topsep=1ex, itemsep=1ex, leftmargin=3em]
        \item \textbf{Tristan Deleu}, Padideh Nouri, Nikolay Malkin, Doina Precup, Yoshua Bengio (2024). \emph{Discrete Probabilistic Inference as Control in Multi-path Environments}. Conference on Uncertainty in Artificial Intelligence (UAI). \notecite{deleu2024gfnmaxentrl}
    \end{itemize}
    \vspace*{5em}
\end{minipage}

We saw in \cref{sec:probabilistic-inference-control-problem} that while being an appealing solution at first glance, maximum entropy reinforcement learning\index{Reinforcement learning!Maximum entropy reinforcement learning} (MaxEnt RL; \citealp{ziebart2010maxent,fox2016tamingnoiserl}) alone was not sufficient to sample from the Gibbs distribution \cref{eq:gibbs-distribution} without introducing some bias towards terminating states with multiple complete trajectories leading to them. This bias motivated the introduction of GFlowNets as a general solution for sampling as a sequential decision making problem, regardless of the structure of the pointed DAG $\gG$. Recall that given an energy function $\gE(x)$, our objective is to sample from the following Gibbs distribution
\begin{equation}
    P^{\star}(x) \propto \exp(-\gE(x) / \alpha),
    \label{eq:gibbs-distribution-alpha}
\end{equation}
where this distribution over the (discrete and compositional) sample space $\gX$ is slightly more general than \cref{eq:gibbs-distribution} by introducing a \emph{temperature} parameter $\alpha > 0$; we can get back to our original formulation by simply taking $\alpha=1$. In this chapter, we will show how we can incorporate a correction to MaxEnt RL in order to remove the bias, this time regardless of the structure of $\gG$.

\section[Unbiased sampling with maximum entropy reinforcement learning]{Unbiased sampling with maximum entropy RL}
\label{sec:unbiased-sampling-maxent-rl}
Once again, to be consistent with the reinforcement learning literature and similar to \cref{sec:probabilistic-inference-control-problem}, we will use the term ``policies'' and denote them by ``$\pi$'' when talking about quantities related to reinforcement learning, and keep ``forward transition probabilities'' and ``$P_{F}$'' for quantities related to GFlowNets. Recall that the objective of MaxEnt RL is to find an optimal stochastic policy $\pi_{\maxent}^{\star}$ that maximizes the expected return, augmented with the entropy of the policy $\gH(\pi(\cdot \mid s)) = -\sum_{s'\in\children_{\gG}(s)}\pi(s'\mid s)\log \pi(s'\mid s)$ along the trajectory:
\begin{equation}
    \pi^{\star}_{\maxent} = \argmax_{\pi}\E_{\tau\sim\pi}\Bigg[\sum_{t=0}^{T}r(s_{t}, s_{t+1}) + \alpha \gH(\pi(\cdot \mid s_{t}))\Bigg],
    \label{eq:maxent-rl-objective-2}
\end{equation}
with the convention $s_{T+1} = \terminal$. With this newly introduced temperature parameter $\alpha$ controlling the relative weight of the entropy regularization compared to the return, \citet{haarnoja2017sql} showed that the optimal policy maximizing \cref{eq:maxent-rl-objective-2} can be written for any $s\rightarrow s'\in\gG$ as
\begin{equation}
    \pi_{\maxent}^{\star}(s'\mid s) = \exp\left(\frac{1}{\alpha}\big(Q^{\star}_{\soft}(s, s') - V_{\soft}^{\star}(s)\big)\right),
    \label{eq:optimal-policy-maxentrl}
\end{equation}
where $Q_{\soft}^{\star}(s, s')$ is a state-action value function and $V_{\soft}^{\star}(s)$ is a state value function that satisfy the following soft Bellman optimality equations (adapted to our setting with a deterministic \& undiscounted MDP):
\begin{align}
    Q^{\star}_{\soft}(s, s') &= r(s, s') + V^{\star}_{\soft}(s')\label{eq:soft-bellman-optimality-equations}\\
    V^{\star}_{\soft}(s) &= \alpha \log \sum_{s''\in\children_{\gG}(s)}\exp\left(\frac{1}{\alpha}Q^{\star}_{\soft}(s, s'')\right).\label{eq:soft-bellman-optimality-equations-V}
\end{align}
Note that these equations slightly differ from \cref{eq:soft-bellman-optimality-equations-intro} \& \cref{eq:soft-bellman-optimality-equations-V-intro} because of the parameter $\alpha$.

\subsection{Bellman optimality \& flow matching on trees}
\label{sec:bellman-optimality-flow-matching-trees}
In this section only, we will assume that $\gG = (\widebar{\gS}, \gA)$ has a tree structure, meaning that every state $s\in\gS$ has a unique parent state (except of course the initial state $s_{0}$). Note that we omit the case $s=\terminal$, since by \cref{def:terminating-states} $\parents_{\gG}(\terminal) = \gX$. If the reward function is defined such that $\sum_{t=0}^{T}r(s_{t}, s_{t+1}) = -\gE(s_{T})$ (with $s_{T+1} = \terminal$), then we saw in \cref{sec:sampling-terminating-states-soft-mdp} that the optimal policy $\pi^{\star}_{\maxent}$ induces a distribution over complete trajectories $\tau = (s_{0}, s_{1}, \ldots, s_{T}, \terminal)$ such that
\begin{equation}
    \pi_{\maxent}^{\star}(\tau) \propto \exp(-\gE(s_{T})/\alpha),
    \label{eq:optimal-maxent-rl-propto-energy-alpha}
\end{equation}
once again slightly differing from \cref{prop:maxent-rl-distribution-propto-return} due to the parameter $\alpha$ (the proof of this proposition can be immediately adapted). In the case where $\gG$ is a tree, there is a unique complete trajectory leading to each terminating state $x\in\gX$, and therefore we can sample from the Gibbs distribution by following the optimal policy $\pi_{\maxent}^{\star}$. Let's consider the case of a sparse reward received only at the end of the trajectory, meaning that $r(s_{T}, \terminal) = -\gE(s_{T})$, and is zero otherwise. The following proposition shows that the flow matching condition we established in \cref{thm:flow-matching-proportional-reward} is exactly equivalent to the soft Bellman optimality equations. %

\begin{proposition}
    \label{prop:equivalence-flow-matching-bellman-equations}
    Let $\gM = (\gG, r)$ be a MDP such that $\gG$ is a tree over $\gS$, and the reward function is sparse, meaning that for any $x\in\gX$, $r(x, \terminal) = -\gE(x)$, and is zero otherwise (\ie for $s\rightarrow s' \in \gG$ such that $s'\neq \terminal$, $r(s, s') = 0$). The flow matching condition \& boundary conditions of \cref{thm:flow-matching-proportional-reward} are equivalent (in log-space) to the soft Bellman optimality equations \cref{eq:soft-bellman-optimality-equations} \& \cref{eq:soft-bellman-optimality-equations-V}, with the following correspondence for all $s\rightarrow s' \in \gG$:
    \begin{equation}
        Q_{\soft}^{\star}(s, s') = \alpha \log F(s \rightarrow s'),
        \label{eq:equivalence-flow-matching-bellman-equations-correspondence}
    \end{equation}
    where $F$ is the edge flow function satisfying the flow matching \& boundary conditions of \cref{thm:flow-matching-proportional-reward}.
\end{proposition}

\begin{proof}
    First, note that the boundary condition of \cref{thm:flow-matching-proportional-reward} reads as $F(x\rightarrow \terminal) = \exp(-\gE(x) / \alpha)$, when the reward function (in the sense of GFlowNets, which must not be confused with the reward $r$ of $\gM$) is $R(x) = \exp(-\gE(x) / \alpha)$. For some transition $s\rightarrow s' \in\gG$, we consider two cases:
    \begin{itemize}[leftmargin=*]
        \item If $s' = \terminal$, then we know that $r(s, \terminal) = -\gE(s)$ and that $\terminal$ has no children in $\gG$, meaning that $V^{\star}_{\soft}(\terminal) = 0$. Therefore, using the correspondence in \cref{eq:equivalence-flow-matching-bellman-equations-correspondence}, we can write \cref{eq:soft-bellman-optimality-equations} as
        \begin{equation}
            \alpha \log F(s\rightarrow \terminal) = Q_{\soft}^{\star}(s, \terminal) = r(s, \terminal) + V_{\soft}^{\star}(\terminal) = -\gE(s),
        \end{equation}
        which corresponds to the boundary condition in log-space.

        \item If $s'\neq\terminal$, then we know that $r(s, s') = 0$ (sparse reward). Using the correspondence in \cref{eq:equivalence-flow-matching-bellman-equations-correspondence}, we can write the soft Bellman optimality equations as
        \begin{align}
            \alpha \log F(s\rightarrow s') &= Q_{\soft}^{\star}(s, s') = r(s, s') + \alpha \log \sum_{s''\in\children_{\gG}(s')}\exp\left(\frac{1}{\alpha}Q_{\soft}^{\star}(s', s'')\right)\\
            &= \alpha \log \sum_{s''\in\children_{\gG}(s')}\exp\left(\frac{1}{\alpha}Q_{\soft}^{\star}(s', s'')\right) = \alpha \log \sum_{s''\in\children_{\gG}(s')}F(s'\rightarrow s'').\label{eq:equivalence-flow-matching-bellman-equations-proof-1}
        \end{align}
        Moreover, since $\gG$ has a tree structure, $s'$ has only a single parent which is $s$. Thus
        \begin{equation}
            \log \sum_{\tilde{s}\in\parents_{\gG}(s')}F(\tilde{s}\rightarrow s') = \log F(s\rightarrow s') = \log \sum_{s''\in\children_{\gG}(s')}F(s'\rightarrow s''),
        \end{equation}
        where we used \cref{eq:equivalence-flow-matching-bellman-equations-proof-1} in the second equality. This corresponds to the flow matching condition in log-space.
    \end{itemize}%
    \vspace*{-2em}
\end{proof}

An interesting point about this result is that the edge (log-)flow in a GFlowNet finds a natural interpretation as a state-value function in reinforcement learning; this intuitive correspondence will be confirmed throughout \cref{sec:equivalence-maxent-rl-gflownet}. In the general case though, recall that when $\gG$ does \emph{not} have a tree structure anymore, the terminating state distribution associated with the optimal policy $\pi_{\maxent}^{\star}$ is biased towards states having multiple paths leading to them:
\begin{equation}
    \pi_{\maxent}^{\star\top}(x) \propto n(x)\exp(-\gE(x)/\alpha),
    \label{eq:terminating-state-distribution-optimal-maxentrl-policy-recall}
\end{equation}
where $n(x)$ counts the number of paths going to a certain terminating state $x\in\gX$.

\subsection{Reward correction}
\label{sec:gflownet-maxentrl-reward-correction}
To correct the bias in \cref{eq:terminating-state-distribution-optimal-maxentrl-policy-recall} caused by multiple complete trajectories leading to the same terminating state, we could treat the terminating state distribution $\pi^{\star\top}_{\maxent}(x)$ in \cref{def:terminating-state-probability} not as a sum, but as an \emph{expectation} over trajectories leading to $x$, by reweighting $\pi^{\star}_{\maxent}(\tau)$ (which is constant for any complete trajectory terminating at $x$ \cref{eq:optimal-maxent-rl-propto-energy-alpha}) with a probability distribution over these trajectories. We saw in \cref{sec:backward-transition-probabilities} that a backward transition probability distribution induces a probability distribution $P_{B}(\tau\mid x)$ over complete trajectories terminating at $x$; this would be a natural choice for such a reweighting. The following theorem shows that if the reward function is defined using the energy function, as in \cref{eq:reward-soft-mdp}, but this time corrected by a fixed backward transition probability, then the terminating state distribution associated with the optimal policy $\pi^{\star}_{\maxent}$ matches the Gibbs distribution of interest.

\begin{theorem}
    \label{thm:reward-correction-maxentrl}
    Let $\gM = (\gG, r)$ be a MDP with a pointed DAG structure $\gG = (\widebar{\gS}, \gA)$, and let $P_{B}: \gS \rightarrow \Delta(\parents_{\gG})$ be an arbitrary backward transition probability consistent with $\gG$. For any $s\rightarrow s'\in \gA$, let $r(s, s')$ be the reward function of the MDP corrected with $P_{B}$, satisfying for any complete trajectory $\tau = (s_{0}, s_{1}, \ldots, s_{T}, \terminal)$:
    \begin{equation}
        \sum_{t=0}^{T}r(s_{t}, s_{t+1}) = -\gE(s_{T}) + \alpha \sum_{t=0}^{T-1}\log P_{B}(s_{t}\mid s_{t+1}),
        \label{eq:reward-correction-maxentrl}
    \end{equation}
    with the convention $s_{T+1}=\terminal$. Then the terminating state distribution associated with the optimal policy $\pi_{\maxent}^{\star}$ solution of \cref{eq:maxent-rl-objective-2} satisfies $\pi^{\star\top}_{\maxent}(x) \propto \exp(-\gE(x) / \alpha)$ for all $x\in\gX$.
\end{theorem}

\begin{proof}
    The proof of this theorem uses similar techniques as the proof of \cref{prop:maxent-rl-distribution-propto-return}. First, note that based on \cref{eq:optimal-policy-maxentrl} \& \cref{eq:soft-bellman-optimality-equations} we can write the optimal policy for any $s\rightarrow s' \in \gG$ as
    \begin{equation}
        \pi^{\star}_{\maxent}(s'\mid s) = \exp\left(\frac{1}{\alpha}\big(r(s, s') + V^{\star}_{\soft}(s') - V^{\star}_{\soft}(s)\big)\right).
    \end{equation}
    By \cref{def:terminating-state-probability}, we can write the terminating state probability distribution associated with $\pi_{\maxent}^{\star}$ for any terminating state $x\in\gX$ as
    {\allowdisplaybreaks%
    \begin{align}
        \pi_{\maxent}^{\star\top}(x) &= \sum_{\tau: s_{0}\rightsquigarrow x}\pi_{\maxent}^{\star}(\tau) = \sum_{\tau: s_{0}\rightsquigarrow x}\prod_{t=0}^{T_{\tau}}\pi_{\maxent}^{\star}(s_{t+1}\mid s_{t})\\
        &= \sum_{\tau: s_{0} \rightsquigarrow x}\exp\left(\frac{1}{\alpha}\sum_{t=0}^{T_{\tau}}\big(r(s_{t}, s_{t+1}) + V_{\soft}^{\star}(s_{t+1}) - V_{\soft}^{\star}(s_{t})\big)\right)\\
        &= \sum_{\tau: s_{0}\rightsquigarrow x}\exp\Bigg(\frac{1}{\alpha}\Bigg(\sum_{t=0}^{T_{\tau}}r(s_{t}, s_{t+1}) + \underbrace{V_{\soft}^{\star}(\terminal)}_{=\,0} - V_{\soft}^{\star}(s_{0})\Bigg)\Bigg)\label{eq:reward-correction-maxentrl-proof-1}\\
        &= \sum_{\tau: s_{0}\rightsquigarrow x}\exp\left(\frac{1}{\alpha}\left(-\gE(x) + \alpha \sum_{t=0}^{T_{\tau}-1}\log P_{B}(s_{t}\mid s_{t+1}) - V_{\soft}^{\star}(s_{0})\right)\right)\\
        &= \exp\left(\frac{1}{\alpha}(-\gE(x) - V_{\soft}^{\star}(s_{0}))\right)\underbrace{\sum_{\tau: s_{0}\rightsquigarrow x}\prod_{t=0}^{T_{\tau}-1}P_{B}(s_{t}\mid s_{t+1})}_{=\,1}\label{eq:reward-correction-maxentrl-proof-2}\\
        &= \frac{\exp(-\gE(x)/\alpha)}{\exp(V_{\soft}^{\star}(s_{0})/\alpha)}\propto \exp(-\gE(x)/\alpha),
    \end{align}}%
    where we used the telescoping of $V_{\soft}^{\star}$ in \cref{eq:reward-correction-maxentrl-proof-1}, and the fact that $P_{B}$ induces a probability distribution over the complete trajectories leading to $x$ in \cref{eq:reward-correction-maxentrl-proof-2} (see \cref{lem:PB-distribution-prefix}). Note that we used the convention $s_{T_{\tau}} = x$ and $s_{T_{\tau}+1} = \terminal$ throughout our derivation, where $T_{\tau}$ indicates the length of the complete trajectory $\tau$ (which may be different, even if they all terminate at the same $x$).
\end{proof}

In other words, \cref{thm:reward-correction-maxentrl} shows that by appropriately correcting the reward function of the MDP, we can emulate the behavior of a GFlowNet (\ie sampling from \cref{eq:gibbs-distribution-alpha}) with MaxEnt RL. This theorem generalizes the result of \citet{tiapkin2024gfnmaxentrl}, who considered a special case of reward correction which we will go back to in the next section. Unlike the reward function we considered in \cref{sec:probabilistic-inference-control-problem} where $\sum_{t=0}^{T}r(s_{t}, s_{t+1}) = -\gE(s_{T})$, the return now depends on the complete trajectory leading to $s_{T}$ via the second term in \cref{eq:reward-correction-maxentrl}, and not just the state $s_{T}$ it reaches at the end. Interestingly, the temperature parameter $\alpha$ introduced in the MaxEnt RL literature \citep{haarnoja2017sql} used to balance the effect of the entropy regularization in \cref{eq:maxent-rl-objective-2} finds a natural interpretation as the temperature of the Gibbs distribution \cref{eq:gibbs-distribution-alpha}. Note that the correction in \cref{eq:reward-correction-maxentrl} only involves the backward probability $P_{B}(\tau\mid x)$ of the whole complete trajectory $\tau$, making it applicable even with non-Markovian backward transition probabilities \citep{shen2023understandingtraininggfn,bengio2023gflownetfoundations} (\ie a generalization of \cref{sec:backward-transition-probabilities}).

This correction of the reward is fully compatible with our observation in \cref{sec:probabilistic-inference-control-problem} that sampling terminating states $\pi^{\star}_{\maxent}$ yields samples from \cref{eq:gibbs-distribution-alpha} when the soft MDP is a tree with the (uncorrected) reward in \cref{eq:reward-soft-mdp}, since in that case any state $s'\neq s_{0}$ has a unique parent $s$, and thus we necessarily have $P_{B}(s\mid s') = 1$, as also observed by \citet{tiapkin2024gfnmaxentrl}. Since $P_{B}$ can be arbitrary, a simple option is for example to set $P_{B}(s\mid s')$ as the uniform distribution over the parents of $s$. Another option is to set $P_{B}(s\mid s') = n(s) / n(s')$, where $n(s)$ is the number of partial trajectories from the initial state $s_{0}$ to $s$. \citet{mohammadpour2024maxentgfn} proved that this backward transition probability has the remarkable property of maximizing the flow entropy (\ie the entropy of $\pi_{\maxent}^{\star}(\tau)$ as a distribution over complete trajectories; \citealp{zhang2022unifyinggfn,shen2023understandingtraininggfn}); we will come back to this property in \cref{sec:modified-detailed-balance}.

\section{Equivalence between MaxEnt RL \& GFlowNet objectives}
\label{sec:equivalence-maxent-rl-gflownet}
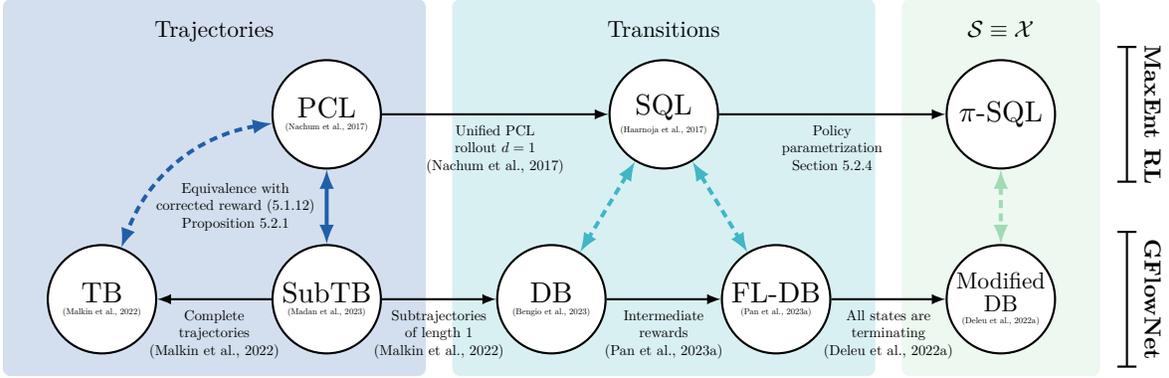
\begin{figure}[t]
\input{figures/chapter5/equivalences}
\caption[Equivalences between objectives in MaxEnt RL \& GFlowNets]{Equivalence between objectives in MaxEnt RL, with corrected rewards, and the objectives in GFlowNets. The objectives are classified based on whether they operate at the level of (complete) trajectories (left), transitions (middle), or if all the states are terminating (right). Further details about the form of the different residuals and the correspondences to transfer from one objective to another are available in \cref{fig:gfn-maxentrl-residuals-equivalences}.}
\label{fig:gfn-maxentrl-equivalences}
\end{figure}

\begin{figure}[hbtp]
    \vspace*{-4em}
    \centering
    \includegraphics{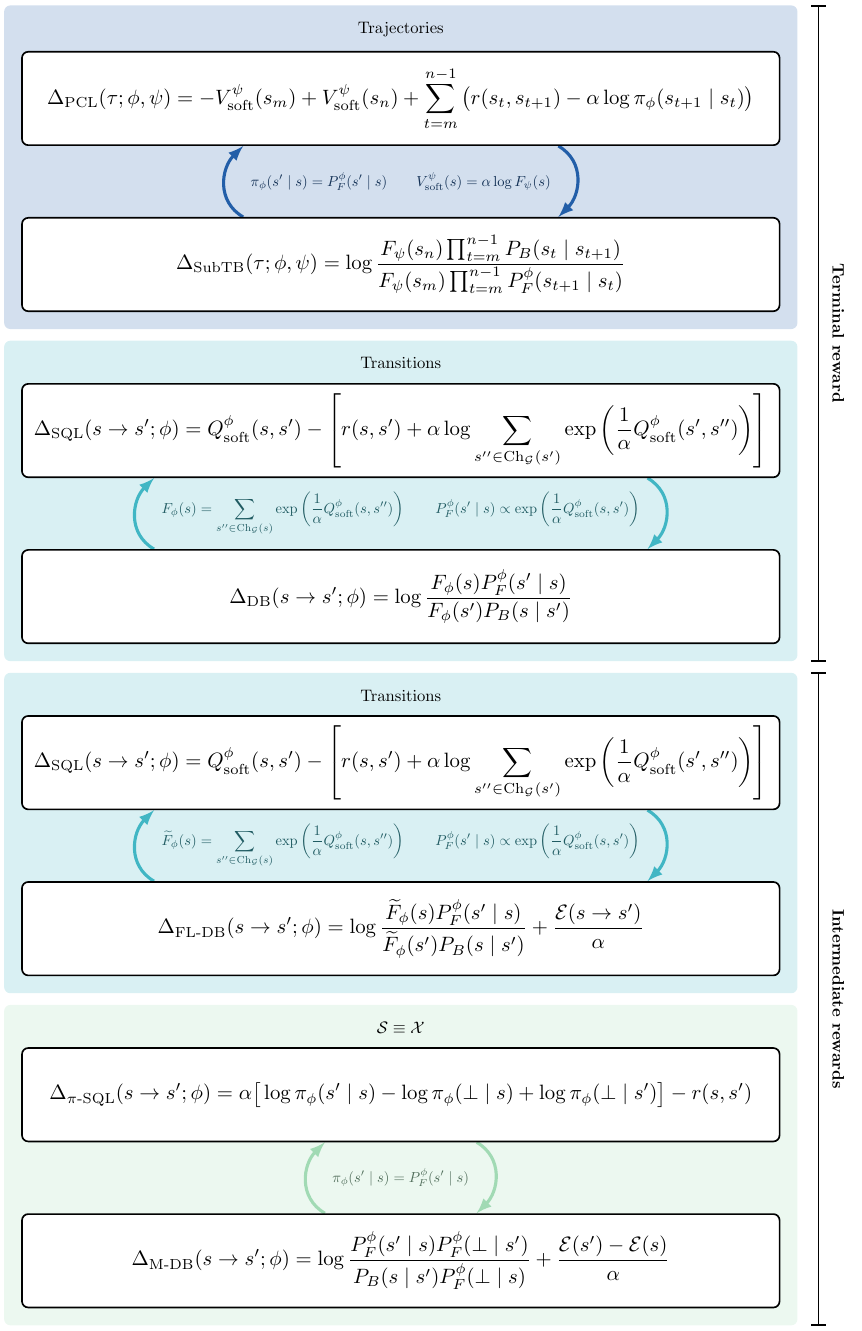}
    \caption[Summary of the equivalences between the residuals in MaxEnt RL \& GFlowNet]{Summary of the equivalences between the residuals in MaxEnt RL (top, in each box) and GFlowNet (bottom, in each box) objectives, using the classification of \cref{fig:gfn-maxentrl-equivalences}. All objectives can be written as $\gL(\cdot) = \frac{1}{2}\E_{\pi_{b}}[\Delta^{2}(\cdot)]$, where $\pi_{b}$ is a distribution over appropriate quantities (\emph{i.e.}, trajectories, or transitions). The \emph{terminal reward} setting corresponds to $r(s_{t}, s_{t+1}) = \alpha \log P_{B}(s_{t}\mid s_{t+1})$ \& $r(s_{T}, \terminal) = -\gE(s_{T})$ (\cref{sec:equivalence-pcl-subtb}), whereas the \emph{intermediate rewards} setting corresponds to $r(s_{t}, s_{t+1}) = -\gE(s_{t}\rightarrow s_{t+1}) + \alpha \log P_{B}(s_{t}\mid s_{t+1})$ \& $r(s_{T}, \terminal) = 0$ (\cref{sec:equivalence-sql-fldb}).}
    \label{fig:gfn-maxentrl-residuals-equivalences}
\end{figure}

\cref{thm:reward-correction-maxentrl} suggests that solving the MaxEnt RL problem in \cref{eq:maxent-rl-objective-2} with the corrected reward is comparable to finding a solution of a GFlowNet, as they both lead to a policy/forward transition probability whose terminating state distribution is \cref{eq:gibbs-distribution-alpha}. It turns out that there exist strong equivalences between specific algorithms solving these two problems, which we will detail in this section. Taking inspiration from \cref{sec:flow-matching-losses}, all the objectives we will consider here, including those for MaxEnt RL, will have the form $\gL(\phi) = \frac{1}{2}\E_{\pi_{b}}\big[\Delta^{2}(\cdot;\phi)\big]$ where $\pi_{b}$ is an arbitrary distribution over appropriate quantities (behavior policy), and $\Delta(\cdot;\phi)$ are residuals\index{Residual} that are algorithm-specific. A full summary of the connections between different MaxEnt RL and GFlowNet objectives is available in \cref{fig:gfn-maxentrl-equivalences}, with further details in \cref{fig:gfn-maxentrl-residuals-equivalences}. Throughout this section, we will assume that the backward transition probability $P_{B}$ used in \cref{eq:reward-correction-maxentrl} is fixed.

\paragraph{Is it the end for GFlowNets?} If GFlowNets and MaxEnt RL are two faces of the same coin, as we will see in this section, is all the research in GFlowNets going to be superseded by MaxEnt RL? After all, the latter is a well-established field, with a large community and applications well beyond the relatively narrow (in comparison) domain of probabilistic inference. Fortunately, the answer to this question is \emph{no}. The main reason being that this equivalence between these two fields holds \emph{under the assumption that $P_{B}$ is fixed}, even if it could be learned in GFlowNets \citep{malkin2022trajectorybalance}. This is necessary because the backward transition probability appears in the correction added to the reward $r(s, s')$, which in the standard formulation of reinforcement learning is required to be a known quantity (and not learned). Despite these theoretical considerations though, from a practical perspective it would be as easy to also learn $P_{B}$ within the framework of MaxEnt RL in this situation, without having to resort to inverse RL.

However, these connections invite some bridges to be made between the two fields, where a lot of existing tools from the reinforcement learning literature can be readily transferred to GFlowNets. We saw in \cref{sec:off-policy-training} some of them already, with the use of Thompson sampling to improve exploration \citep{rectorbrooks2023thompsongfn}, advanced planning methods with Monte Carlo tree search \citep{morozov2024mctsgfn}, or replay buffers \citep{deleu2022daggflownet}. We will also see in \cref{chap:dag-gflownet} that it may be advantageous to also use a \emph{target network} with objectives inspired by Q-Learning (\cref{sec:equivalence-pisql-mdb}), as popularized by the seminal work on Deep Q-Networks \citep{mnih2015dqn}. \citet{zhang2023distributionalgfn} introduced a distributional version of GFlowNets, taking inspiration from distributional RL \citep{bellemare2023distributionalrl}. This intersection is an exciting direction of research.

\subsection{Path consistency learning \& (sub-)trajectory balance}
\label{sec:equivalence-pcl-subtb}
Similar to how \cref{eq:reward-soft-mdp} covered the particular case of a sparse reward that is only received at the end of the trajectory (\ie $r(s_{T}, \terminal) = -\gE(s_{T})$, and zero everywhere else), in this section and \cref{sec:equivalence-sql-db} we will consider a reward function satisfying \cref{eq:reward-correction-maxentrl} where the energy function only appears at the end of the trajectory
\begin{align}
    r(s_{t}, s_{t+1}) &= \alpha \log P_{B}(s_{t}\mid s_{t+1}) && r(s_{T}, \terminal) = -\gE(s_{T}),
    \label{eq:maxentrl-correction-sparse}
\end{align}
as introduced in \citet{tiapkin2024gfnmaxentrl}; we can see that this reward function satisfies the condition \cref{eq:reward-correction-maxentrl} of \cref{thm:reward-correction-maxentrl}. With this particular choice of reward function, we will establish an equivalence between the \emph{Path Consistency Learning} algorithm (\gls{pcl}; \citealp{nachum2017pcl}) in MaxEnt RL and the \emph{sub-trajectory balance} loss (SubTB; \citealp{malkin2022trajectorybalance,madan2022subtb}) in the GFlowNet literature. On the one hand, the PCL objective encourages the consistency at the level of partial trajectories $\tau = (s_{m}, s_{m+1}, \ldots, s_{n})$ between a policy $\pi_{\phi}$ parametrized by $\phi$ and a state value function $V_{\soft}^{\psi}$ parametrized by $\psi$, whose residual can be written as
\begin{equation}
    \Delta_{\mathrm{PCL}}(\tau; \phi, \psi) = -V_{\soft}^{\psi}(s_{m}) + V_{\soft}^{\psi}(s_{n}) + \sum_{t=m}^{n-1}\big(r(s_{t}, s_{t+1}) - \alpha \log \pi_{\phi}(s_{t+1}\mid s_{t})\big),
    \label{eq:residual-pcl}
\end{equation}
where we enforce that $V_{\soft}^{\psi}(\terminal) = 0$. On the other hand, the SubTB objective also encourages some form of consistency at the level of partial trajectories, but this time between a forward transition probability $P_{F}^{\phi}$ parametrized by $\phi$, and a state flow function $F_{\psi}$ parametrized by $\psi$, where the residual is defined by
\begin{equation}
    \Delta_{\mathrm{SubTB}}(\tau; \phi, \psi) = \log \frac{F_{\psi}(s_{n})\prod_{t=m}^{n-1}P_{B}(s_{t}\mid s_{t+1})}{F_{\psi}(s_{m})\prod_{t=m}^{n-1}P_{F}^{\phi}(s_{t+1}\mid s_{t})},
    \label{eq:residual-subtb}\index{Sub-trajectory balance!Loss}
\end{equation}
if $s_{n} \neq \terminal$. It is important to recall that satisfying the sub-trajectory balance conditions for all partial trajectories of fixed length does not necessarily guarantee that the terminating state distribution $P_{F}^{\phi\top}$ matches \cref{eq:gibbs-distribution-alpha} (see \cref{sec:sub-trajectory-balance-condition}; although it does if we consider all partial trajectories of length $\leq L$ for any $L$). Nevertheless, this loss is interesting since it generalizes existing GFlowNet losses we introduced in \cref{sec:flow-matching-losses} and makes the presentation clearer. Similar to how the detailed balance loss in \cref{eq:detailed-balance-loss} required an additional reward matching loss\index{Reward matching loss} to account for the boundary conditions, if the end of the partial trajectory $s_{n} = \terminal$ is the terminal state, then the residual is
\begin{equation}
    \Delta_{\mathrm{SubTB}}(\tau; \phi, \psi) = \log \frac{\prod_{t=m}^{n-2}P_{B}(s_{t}\mid s_{t+1})}{F_{\psi}(s_{m})\prod_{t=m}^{n-1}P_{F}^{\phi}(s_{t+1}\mid s_{t})} - \frac{\gE(s_{n-1})}{\alpha}.
    \label{eq:residual-subtb-2}
\end{equation}
The following proposition establishes the equivalence between the two objectives induced by these residuals, up to a constant that only depends on the temperature $\alpha$, and provides a way to move from the policy/value function parametrization in MaxEnt RL to the forward transition probability/flow function parametrization in GFlowNets.

\begin{proposition}
    \label{prop:equivalence-pcl-subtb}
    The sub-trajectory balance objective (GFlowNet; \citealp{madan2022subtb}) is proportional to the Path Consistency Learning objective (MaxEnt RL; \citealp{nachum2017pcl}) on the MDP with the reward function defined in \cref{eq:maxentrl-correction-sparse}, in the sense that $\gL_{\mathrm{PCL}}(\phi, \psi) = \alpha^{2}\gL_{\mathrm{SubTB}}(\phi, \psi)$, with the following correspondence
    \begin{align}
        \pi_{\phi}(s'\mid s) &= P_{F}^{\phi}(s'\mid s) &&& V_{\soft}^{\psi}(s) &= \alpha \log F_{\psi}(s).
        \label{eq:equivalence-pcl-subtb-correspondence}
    \end{align}
\end{proposition}

\begin{proof}
    In order to show the equivalence between $\gL_{\mathrm{PCL}}$ and $\gL_{\mathrm{SubTB}}$, it is sufficient to only show the equivalence between their corresponding residuals on a partial trajectory $\tau = (s_{m}, s_{m+1}, \ldots, s_{n})$, by replacing the reward by \cref{eq:maxentrl-correction-sparse}. We will use the correspondence in \cref{eq:equivalence-pcl-subtb-correspondence} in order to move between the parameterizations of MaxEnt RL and GFlowNets. We consider two cases
    \begin{itemize}[leftmargin=*]
        \item If $s_{n} \neq \terminal$ is not the terminal state, then the residual for SubTB is given by \cref{eq:residual-subtb}. Substituting \cref{eq:maxentrl-correction-sparse} \& \cref{eq:equivalence-pcl-subtb-correspondence} into the residual $\Delta_{\mathrm{PCL}}$:
        {\allowdisplaybreaks%
        \begin{align}
            \Delta_{\mathrm{PCL}}&(\tau;\phi,\psi) = -V_{\soft}^{\psi}(s_{m}) + V_{\soft}^{\psi}(s_{n}) + \sum_{t=m}^{n-1}\big(r(s_{t}, s_{t+1}) - \alpha \log \pi_{\phi}(s_{t+1}\mid s_{t})\big)\\
            &= -\alpha \log F_{\psi}(s_{m}) + \alpha \log F_{\psi}(s_{n}) + \alpha \sum_{t=m}^{n-1}\big(\log P_{B}(s_{t}\mid s_{t+1}) - \log P_{F}^{\phi}(s_{t+1}\mid s_{t})\big)\\
            &= \alpha \log \frac{F_{\psi}(s_{n})\prod_{t=m}^{n-1}P_{B}(s_{t}\mid s_{t+1})}{F_{\psi}(s_{m})\prod_{t=m}^{n-1}P_{F}^{\phi}(s_{t+1}\mid s_{t})} = \alpha \Delta_{\mathrm{SubTB}}(\tau;\phi,\psi).
        \end{align}}%
        \item If $s_{n} = \terminal$, then the residual for SubTB is given by \cref{eq:residual-subtb-2}. Once again, substituting \cref{eq:maxentrl-correction-sparse} \& \cref{eq:equivalence-pcl-subtb-correspondence} into the residual $\Delta_{\mathrm{PCL}}$, and using the fact that $V_{\soft}^{\psi}(\terminal) = 0$, we have
        {\allowdisplaybreaks%
        \begin{align}
            \Delta_{\mathrm{PCL}}(\tau;\phi,\psi) &= -V_{\soft}^{\psi}(s_{m}) + V_{\soft}^{\psi}(\terminal) + \sum_{t=m}^{n-1}\big(r(s_{t}, s_{t+1}) - \alpha \log \pi_{\phi}(s_{t+1}\mid s_{t})\big)\\
            &= -\alpha \log F_{\psi}(s_{m}) - \big(\gE(s_{n-1}) + \alpha \log P_{F}^{\phi}(\terminal \mid s_{n-1})\big)\nonumber\\
            &\qquad + \alpha \sum_{t=m}^{n-2}\big(\log P_{B}(s_{t}\mid s_{t+1}) - \log P_{F}^{\phi}(s_{t+1}\mid s_{t})\big)\\
            &= \alpha\left[\log \frac{\prod_{t=m}^{n-2}P_{B}(s_{t}\mid s_{t+1})}{F_{\psi}(s_{m})\prod_{t=m}^{n-1}P_{F}^{\phi}(s_{t+1}\mid s_{t})} - \frac{\gE(s_{n-1})}{\alpha}\right] = \alpha \Delta_{\mathrm{SubTB}}(\tau;\phi,\psi).
        \end{align}}%
    \end{itemize}
    This concludes the proof, showing that $\gL_{\mathrm{PCL}}(\phi, \psi) = \alpha^{2}\gL_{\mathrm{SubTB}}(\phi, \psi)$.
\end{proof}

Similarities between PCL and SubTB have also been mentioned in prior work \citep{malkin2022trajectorybalance,jiralerspong2024eflownet,hu2024gfnllm,mohammadpour2024maxentgfn}, although without showing this strong equivalence. The equivalence between the value function in MaxEnt RL and the state flow function in GFlowNets was also found in \citet{tiapkin2024gfnmaxentrl}. When applied to complete trajectories, this proposition also shows an equivalence between PCL and the \emph{trajectory balance} loss\index{Trajectory balance!Loss} (TB; \citealp{malkin2022trajectorybalance}). Indeed, the trajectory balance loss in \cref{eq:trajectory-balance-loss} corresponds exactly to \cref{eq:residual-subtb-2} when $s_{m} = s_{0}$ is the initial state (using a scalar parametrization $Z$ instead of $F(s_{0})$).

\subsection{Soft Q-Learning \& detailed balance}
\label{sec:equivalence-sql-db}\index{Soft Q-Learning}
If we consider consistency over complete trajectories in PCL/TB as one end of a spectrum, on the other end of the spectrum we can consider objectives measuring the consistency at the level of transitions $s\rightarrow s'\in\gG$. We saw in \cref{sec:sampling-terminating-states-soft-mdp} that one of the simplest method for finding the optimal policy $\pi_{\maxent}^{\star}$ is the Soft Q-Learning algorithm (SQL; \citealp{haarnoja2017sql}), that operates on transitions via a state-action value function $Q(s, s')$. In the tabular case, we can interpret the procedure detailed in \cref{alg:soft-q-learning} as performing stochastic (semi-)gradient descent on a certain objective with the residual
{\allowdisplaybreaks%
\begin{align}
    \Delta_{\mathrm{SQL}}(s\rightarrow s'; \phi) &= Q_{\soft}^{\phi}(s, s') - \big(r(s, s') + V_{\soft}^{\phi}(s')\big) \label{eq:residual-sql}\\
    \mathrm{where}\quad V_{\soft}^{\phi}(s') &\triangleq \alpha \log \sum_{s''\in\children_{\gG}(s')}\exp\left(\frac{1}{\alpha}Q_{\soft}^{\phi}(s', s'')\right).
\end{align}}%
quantifying the mismatch in the soft Bellman optimality equations in \cref{eq:soft-bellman-optimality-equations} \& \cref{eq:soft-bellman-optimality-equations-V}. In the case where $s' = \terminal$ is the terminal state, then we have $V_{\soft}^{\phi}(s') = 0$. This residual depends only on a state-action value function $Q_{\soft}^{\phi}(s, s')$ parametrized by $\phi$.

Since SQL works at the level of transitions, it would be natural to study its equivalence with the \emph{detailed balance} loss (DB; \citealp{bengio2023gflownetfoundations}) in GFlowNets. The DB loss introduced in \cref{sec:flow-matching-losses}, whose residual (including for the reward matching loss, with the energy function here) is given by
\begin{align}
    \textrm{for $s\rightarrow s'\in\gG$,}&& \Delta_{\mathrm{DB}}(s\rightarrow s';\phi) &= \log \frac{F_{\phi}(s)P_{F}^{\phi}(s'\mid s)}{F_{\phi}(s')P_{B}(s\mid s')}\label{eq:residual-db}\\
    \textrm{for $x\in\gX$,}&&\Delta_{\mathrm{DB}}(x\rightarrow \terminal; \phi) &= \log \big(F_{\phi}(x)P_{F}^{\phi}(\terminal \mid x)\big) + \frac{\gE(x)}{\alpha},
    \label{eq:residual-db-2}\index{Detailed balance!Loss}
\end{align}
which corresponds exactly to the residuals of SubTB in \cref{eq:residual-subtb} \& \cref{eq:residual-subtb-2} (up to a change of sign) when the partial trajectory is of length 1 (\ie a transition). With this in mind, would it be possible to use the result of \cref{prop:equivalence-pcl-subtb} in order to establish a connection between SQL and DB?

There is a discrepancy between \cref{eq:residual-sql} and \cref{eq:residual-db} though: while SQL uses a single state-action value function $Q_{\soft}^{\phi}(s, s')$, DB depends on a state flow $F_{\psi}(s)$ and a forward transition probability $P_{F}^{\phi}(s'\mid s)$ which are in general parametrized by two separate functions. This separation also existed in the residual of PCL. Fortunately, \citet{nachum2017pcl} showed that SQL can be interpreted as a special case of PCL under what they called the \emph{Unified PCL} perspective, where the state value function $V_{\soft}^{\phi}(s)$ and the policy $\pi_{\phi}(s'\mid s)$ appearing in \cref{eq:residual-pcl} can be parametrized by a single state-action value function $Q_{\soft}^{\phi}(s, s')$ via
\begin{align}
    V_{\soft}^{\phi}(s) &= \alpha \log \sum_{s''\in\children_{\gG}(s)}\exp\left(\frac{1}{\alpha}Q_{\soft}^{\phi}(s, s'')\right) &&& \pi_{\phi}(s'\mid s) &\propto \exp\left(\frac{1}{\alpha}Q_{\soft}^{\phi}(s, s')\right).
    \label{eq:unified-pcl}
\end{align}
As a corollary of \cref{prop:equivalence-pcl-subtb} under this Unified PCL perspective, the following proposition shows that there exists a strong equivalence between DB and SQL, once again under our choice of reward function detailed in the previous section.

\begin{proposition}
    \label{prop:equivalence-sql-db}
    The detailed balance objective (GFlowNet; \citealp{bengio2023gflownetfoundations}) is proportional to the Soft Q-Learning objective (MaxEnt RL; \citealp{haarnoja2017sql}) on the MDP with the reward function defined in \cref{eq:maxentrl-correction-sparse}, in the sense that $\gL_{\mathrm{SQL}}(\phi) = \alpha^{2}\gL_{\mathrm{DB}}(\phi)$, with the following correspondence
    \begin{align}
        F_{\phi}(s) &= \sum_{s''\in\children_{\gG}(s)}\exp\left(\frac{1}{\alpha}Q_{\soft}^{\phi}(s, s'')\right) &&& P_{F}^{\phi}(s'\mid s) &\propto \exp\left(\frac{1}{\alpha}Q_{\soft}^{\phi}(s, s')\right).
        \label{eq:equivalence-sql-db-correspondence}
    \end{align}
\end{proposition}

\begin{proof}
    While this proposition can be seen as a corollary of \cref{prop:equivalence-pcl-subtb}, we prove it here as a standalone result for completeness. Let $s\rightarrow s'\in \gG$ be a transition. We consider two cases:
    \begin{itemize}[leftmargin=*]
        \item If $s' \neq \terminal$ is not the terminal state, then we know that $r(s, s') = \alpha \log P_{B}(s\mid s')$. Substituting it into the residual $\Delta_{\mathrm{SQL}}$, and using the correspondence in \cref{eq:equivalence-sql-db-correspondence}, we have:
        {\allowdisplaybreaks%
        \begin{align}
            \Delta_{\mathrm{SQL}}&(s\rightarrow s';\phi) = Q_{\soft}^{\phi}(s, s') - r(s, s') - \alpha \log \sum_{s''\in\children_{\gG}(s')}\exp\left(\frac{1}{\alpha}Q_{\soft}^{\phi}(s', s'')\right)\\
            &= Q_{\soft}^{\phi}(s, s') - \alpha \log P_{B}(s\mid s') - \alpha \log \sum_{s''\in\children_{\gG}(s')}\exp\left(\frac{1}{\alpha}Q_{\soft}^{\phi}(s', s'')\right)\\
            &= \alpha \log \left(\exp\left(\frac{1}{\alpha}Q_{\soft}^{\phi}(s, s')\right)\right) - \alpha \log P_{B}(s\mid s') - \alpha \log \sum_{s''\in\children_{\gG}(s')}\exp\left(\frac{1}{\alpha}Q_{\soft}^{\phi}(s', s'')\right)\nonumber\\
            &\qquad - \alpha \log \sum_{s''\in\children_{\gG}(s)}\exp\left(\frac{1}{\alpha}Q_{\soft}^{\phi}(s, s'')\right) + \alpha \log \sum_{s''\in\children_{\gG}(s)}\exp\left(\frac{1}{\alpha}Q_{\soft}^{\phi}(s, s'')\right)\\
            &= \alpha\left[\log P_{F}^{\phi}(s'\mid s) + \log F_{\phi}(s) - \log P_{B}(s\mid s') - \log F_{\phi}(s')\right] = \alpha \Delta_{\mathrm{DB}}(s \rightarrow s'; \phi)
        \end{align}}%

        \item If $s' = \terminal$ is the terminal state, then the reward is $r(s, \terminal) = -\gE(s)$. Substituting it again into the residual $\Delta_{\mathrm{SQL}}$, and using the correspondence in \cref{eq:equivalence-sql-db-correspondence}, we have:
        {\allowdisplaybreaks%
        \begin{align}
            \Delta_{\mathrm{SQL}}(s\rightarrow \terminal; \phi) &= Q_{\soft}^{\phi}(s, \terminal) - r(s, \terminal)\\
            &= \alpha \log \left(\exp\left(\frac{1}{\alpha}Q_{\soft}^{\phi}(s, \terminal)\right)\right) + \gE(s)\\
            &\qquad - \alpha \log \sum_{s''\in\children_{\gG}(s)}\exp\left(\frac{1}{\alpha}Q_{\soft}^{\phi}(s, s'')\right) + \alpha \log \sum_{s''\in\children_{\gG}(s)}\exp\left(\frac{1}{\alpha}Q_{\soft}^{\phi}(s, s'')\right)\nonumber\\
            &= \alpha\left[\log P_{F}^{\phi}(\terminal \mid s) + \log F_{\phi}(s) + \frac{\gE(s)}{\alpha}\right] = \alpha \Delta_{\mathrm{DB}}(s \rightarrow \terminal;\phi)
        \end{align}}%
    \end{itemize}
    This concludes the proof, showing that $\gL_{\mathrm{SQL}}(\phi) = \alpha^{2}\gL_{\mathrm{DB}}(\phi)$.
\end{proof}

Note that \citet{tiapkin2024gfnmaxentrl} established a similar connection between SQL and DB through a dueling architecture perspective \citep{wang2016dueling}, where the state-action value function $Q_{\soft}(s, s')$ is decomposed as (adapted to incorporate the parameter $\alpha$):
\begin{equation}
    \frac{1}{\alpha}Q_{\soft}^{\psi,\phi}(s, s') = V_{\soft}^{\psi}(s) + A_{\soft}^{\phi}(s, s') - \log \sum_{s''\in\children_{\gG}(s)}\exp\big(A_{\soft}^{\phi}(s, s'')\big),
    \label{eq:dueling-q-function}
\end{equation}
where $A_{\soft}^{\phi}(s, s')$ is an advantage function quantifying the added value of taking the transition $s\rightarrow s'$ compared to the ``baseline'' value $V_{\soft}^{\psi}(s)$ at state $s$. The additional log-sum-exp term in the decomposition above is to account for the identifiability between the advantage and the state-action value functions. The connection \citet{tiapkin2024gfnmaxentrl} established is then between the state value function $V_{\soft}^{\psi}(s)$ and the state flow function $F_{\psi}(s)$ on the one hand, and the advantage function $A_{\soft}^{\phi}(s, s')$ and the forward transition probability $P_{F}^{\phi}(s'\mid s)$ on the other hand, via
{\allowdisplaybreaks%
\begin{align}
    \log F_{\psi}(s) &= V_{\soft}^{\psi}(s) = \log \sum_{s''\in\children_{\gG}(s)}\exp\left(\frac{1}{\alpha}Q_{\soft}^{\psi,\phi}(s, s'')\right)\label{eq:equivalence-sql-dueling-correspondence-1}\\
    \log P_{F}^{\phi}(s'\mid s) &= A_{\soft}^{\phi}(s, s') - \log \sum_{s''\in\children_{\gG}(s)}\exp\big(A_{\soft}^{\phi}(s, s')\big).\label{eq:equivalence-sql-dueling-correspondence-2}
\end{align}}%
This differs from our result above in that \cref{prop:equivalence-sql-db} does not explicitly require a separate advantage function. However, we remark that both results are effectively equivalent to one another when the $V_{\soft}^{\phi}(s)$ and $A_{\soft}^{\phi}(s, s')$ share the same parametrization, since \cref{eq:equivalence-sql-dueling-correspondence-1} corresponds exactly to \cref{eq:equivalence-sql-db-correspondence}, and
\begin{equation}
    \log P_{F}^{\phi}(s'\mid s) = \frac{1}{\alpha}Q_{\soft}^{\phi}(s, s') - V_{\soft}^{\phi}(s) = \frac{1}{\alpha}Q_{\soft}^{\phi}(s, s') - \log \sum_{s''\in\children_{\gG}(s)}\exp\left(\frac{1}{\alpha}Q_{\soft}^{\phi}(s, s'')\right),
\end{equation}
where we plugged \cref{eq:dueling-q-function} into \cref{eq:equivalence-sql-dueling-correspondence-2} to also match the correspondence in \cref{prop:equivalence-sql-db}. This connection through a dueling architecture is closer to DB when the forward transition probability and the state flow function are parametrized separately. Empirically, we found that on some environments this strategy where the state value and the advantage functions (or equivalently the flow function and the forward transition probability in GFlowNets) are parametrized by two separate functions was more performant than having a single function parametrizing both \citep{deleu2024gfnmaxentrl}. However, a more extensive study is necessary in order to conclude on the effectiveness of both methods.

\subsection{Soft Q-Learning \& forward-looking detailed balance}
\label{sec:equivalence-sql-fldb}
\begin{figure}[t]
    \begin{adjustbox}{center}
        \includegraphics[width=480pt]{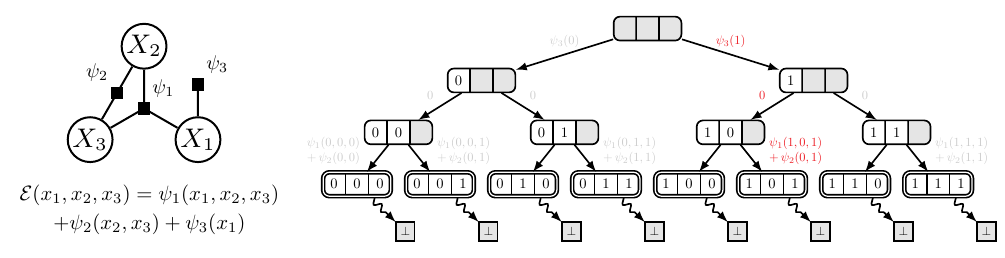}
    \end{adjustbox}
    \caption[Revisiting sampling in a discrete factor graph, with intermediate rewards]{Revisiting sampling in a discrete factor graph of \cref{sec:sampling-factor-graphs}, with intermediate rewards. Instead of computing the energy of the whole terminating state $\gE(x_{1}, x_{2}, x_{3})$ upon termination, there are some intermediate rewards along the way as soon as all the information necessary to evaluate the factors is available. The sum of rewards along a complete trajectory (in red) still equals $\gE(1, 0, 1)$.}
    \label{fig:treesample-fldb}
\end{figure}

\index{Soft Q-Learning}
Up until now, we have considered cases where the energy function was only received at the end of the complete trajectory. This was due to the fact that in general the energy function $\gE(x)$ is only defined for terminating states $x\in\gX$, and nowhere else (\ie it is undefined on intermediate states in $\gS\backslash\gX$). However there are situations where the energy function itself has a natural decomposition that would allow us to get intermediate signals. We illustrate this by revisiting the example detailed in \cref{sec:sampling-factor-graphs}, where we use MaxEnt RL (or GFlowNets) for inference in a discrete factor graph. To make this example more concrete, we consider the following energy function
\begin{equation}
    \gE(x_{1}, x_{2}, x_{3}) = \psi_{1}(x_{1}, x_{2}, x_{3}) + \psi_{2}(x_{2}, x_{3}) + \psi_{3}(x_{1}),
    \label{eq:energy-function-treesample-example-d3}
\end{equation}
where the functions (factors) $\psi_{m}$ are known and only depend on a subset of values, and each $x_{i}$ is a value in $\{0, 1\}$ (see \cref{fig:treesample-fldb}). The standard approach to evaluate this energy function would be to wait until all the random variables have been assigned a value in order to evaluate \cref{eq:energy-function-treesample-example-d3} all at once, as we did in \cref{sec:sampling-factor-graphs}.

Alternatively, we could evaluate the terms in \cref{eq:energy-function-treesample-example-d3} \emph{as soon as all the necessary information is available}, meaning as soon as all the variables appearing in a factor have been assigned a value (recall that the values are assigned one random variable at a time, following a fixed order; \citealp{buesing2020treesample}). This is illustrated in \cref{fig:treesample-fldb}. For example, at the first step of generation the value of $x_{1}$ is assigned a value in $\{0, 1\}$, meaning that we can readily evaluate the term $\psi_{3}(x_{1})$. Then at the second step of generation when the value of $x_{2}$ is assigned, no new term can be evaluated since all the factors in which $x_{2}$ appears also depend on the value of $x_{3}$ which has not been assigned yet. And finally once the value of $x_{3}$ is known at the final step of generation, we can evaluate the remaining two terms $\psi_{1}(x_{1}, x_{2}, x_{3}) + \psi_{2}(x_{2}, x_{3})$. This provides intermediate information along the way.

While in this example $\gG$ is a tree (defeating the purpose of the correction in \cref{sec:gflownet-maxentrl-reward-correction}), it is important to note that there is nothing preventing us from applying the same idea to an arbitrary pointed DAG $\gG$, as long as for any complete trajectory $\tau = (s_{0}, s_{1}, \ldots, s_{T}, \terminal)$, the energy function at the terminating state can be decomposed as
\begin{equation}
    \gE(s_{T}) = \sum_{t=0}^{T-1}\gE(s_{t} \rightarrow s_{t+1}),
    \label{eq:energy-decomposition-fl}
\end{equation}
where we overloaded the notation $\gE(s_{t} \rightarrow s_{t+1})$ to denote the intermediate energy received along $\tau$. In our example above, we have for instance $\gE\big((\cdot, \cdot, \cdot) \rightarrow (x_{1}, \cdot, \cdot)\big) = \psi_{3}(x_{1})$. It is not always possible to have such a decomposition though (\eg in molecule generation in \cref{sec:generation-small-organic-molecules}). However when it is available, \citet{pan2023flgfn} introduced a class of losses for GFlowNets called \emph{forward-looking} losses that takes advantage of this decomposition. In particular, the \emph{forward-looking detailed balance}\index{Detailed balance!Forward-looking} loss (\gls{fldb}) can be written with the following residual, for any transition $s\rightarrow s'\in\gG$ such that $s'\neq \terminal$:
\begin{equation}
    \Delta_{\fldb}(s\rightarrow s';\phi) = \log \frac{\widetilde{F}_{\phi}(s)P_{F}^{\phi}(s'\mid s)}{\widetilde{F}_{\phi}(s')P_{B}(s\mid s')} + \frac{\gE(s\rightarrow s')}{\alpha},
    \label{eq:residual-fldb}
\end{equation}
where $P_{F}^{\phi}(s'\mid s)$ is a forward transition probability, and $\widetilde{F}_{\phi}(s)$ is an \emph{offset} state flow function. This represents an offset to the (true) state flow function $F(s)$ in the sense that we make full use of the intermediate energy signal $\gE(s\rightarrow s')$ that we don't have to learn. The difference in state flows is related to $\widetilde{F}_{\phi}$ via
\begin{equation}
    \log F(s') - \log F(s) = \big(\log \widetilde{F}_{\phi}(s') - \log \widetilde{F}_{\phi}(s)\big) - \frac{\gE(s\rightarrow s')}{\alpha}.
    \label{eq:offset-fldb}
\end{equation}
Unlike the standard detailed balance loss, the forward looking detailed balance loss does not require an additional reward matching loss for the terminating transition, since it already leverages intermediate information with $\gE(s\rightarrow s')$. \citet{jang2024ledgfn} showed that this it is also possible to learn this decomposition of the energy function.

Under the decomposition of the energy function in \cref{eq:energy-decomposition-fl}, we can define the corrected reward as follows on a complete trajectory $\tau = (s_{0}, s_{1}, \ldots, s_{T}, \terminal)$, in order to satisfy the conditions of \cref{thm:reward-correction-maxentrl}:
\begin{align}
    r(s_{t}, s_{t+1}) &= -\gE(s_{t} \rightarrow s_{t+1}) + \alpha \log P_{B}(s_{t}\mid s_{t+1}) &&& r(s_{T}, \terminal) &= 0.
    \label{eq:maxentrl-correction-dense-fldb}
\end{align}
The same way we can consider the reward function \cref{eq:maxentrl-correction-sparse} we used in the previous two sections to be analogous to a ``sparse'' reward (where the energy is only obtained at the end, only the correction is added along the way), we reshaped the rewards in order to obtain ``dense'' rewards \citep{ng1999rewardshaping}, while still satisfying the correction in \cref{eq:reward-correction-maxentrl}. The following proposition, similar to \cref{prop:equivalence-sql-db}, establishes an equivalence between FL-DB and SQL under this new choice of reward function.
\begin{proposition}
    \label{prop:equivalence-sql-fldb}
    The forward-looking detailed balance objective (GFlowNet; \citealp{pan2023flgfn}) is proportional to the Soft Q-Learning objective (MaxEnt RL; \citealp{haarnoja2017sql}) on the MDP with the reward function defined in \cref{eq:maxentrl-correction-dense-fldb}, in the sense that $\gL_{\mathrm{SQL}}(\phi) = \alpha^{2}\gL_{\fldb}(\phi)$, with the following correspondence
    \begin{align}
        \widetilde{F}_{\phi}(s) &= \sum_{s''\in\children_{\gG}(s)}\exp\left(\frac{1}{\alpha}Q_{\soft}^{\phi}(s, s'')\right) &&& P_{F}^{\phi}(s'\mid s) &\propto \exp\left(\frac{1}{\alpha}Q_{\soft}^{\phi}(s, s')\right).
        \label{eq:equivalence-sql-fldb-correspondence}
    \end{align}
\end{proposition}

\begin{proof}
    The proof is similar to the proof of \cref{prop:equivalence-sql-db}. Let $s\rightarrow s'\in\gG$ be a transition such that $s'\neq \terminal$. We our choice of reward function in \cref{eq:maxentrl-correction-dense-fldb} we know that $r(s, s') = -\gE(s\rightarrow s') + \alpha \log P_{B}(s\mid s')$. Substituting it into the residual of SQL in \cref{eq:residual-sql}, we get
    \begin{align}
        \Delta_{\mathrm{SQL}}&(s\rightarrow s'; \phi) = Q_{\soft}^{\phi}(s, s') - \big(r(s, s') + V_{\soft}^{\phi}(s')\big)\\
        &= Q_{\soft}^{\phi}(s, s') - \alpha \log \sum_{s''\in\children_{\gG}(s')}\exp\left(\frac{1}{\alpha}Q_{\soft}^{\phi}(s', s'')\right) + \gE(s\rightarrow s') - \alpha \log P_{B}(s\mid s')\nonumber\\
        &\qquad + \alpha \log \sum_{s''\in\children_{\gG}(s)}\exp\left(\frac{1}{\alpha}Q_{\soft}^{\phi}(s, s'')\right) - \alpha \log \sum_{s''\in\children_{\gG}(s)}\exp\left(\frac{1}{\alpha}Q_{\soft}^{\phi}(s, s'')\right)\\
        &= \alpha\left[\log P_{F}^{\phi}(s'\mid s) - \log P_{B}(s\mid s') + \log \widetilde{F}_{\phi}(s) - \log \widetilde{F}_{\phi}(s') + \frac{\gE(s\rightarrow s')}{\alpha}\right]\\
        &= \alpha \Delta_{\fldb}(s\rightarrow s';\phi)
    \end{align}
    This concludes the proof, showing that $\gL_{\mathrm{SQL}}(\phi) = \alpha^{2}\gL_{\fldb}(\phi)$.
\end{proof}

Interestingly, the correspondence in \cref{eq:equivalence-sql-fldb-correspondence} between the (offset) state flow function and the forward transition probability on the one hand (GFlowNet) and the state-action value function on the other hand (MaxEnt RL) is exactly the same as the one in \cref{prop:equivalence-sql-db}.

\subsection{Policy parametrized SQL \& modified detailed balance}
\label{sec:equivalence-pisql-mdb}\index{Soft Q-Learning!Policy parametrized}
As a special case of the decomposition of the energy function we considered in the previous section, suppose that all the states of the states space are terminating $\gS \equiv \gX$. We saw an example of this situation in \cref{sec:case-study-application-bayesian-inference-decision-trees}, and we will come back to this setting in \cref{chap:dag-gflownet} in the context of GFlowNets over DAGs. In that case, it is clear that the energy function is defined for all the states $s\in\gS$, and it can trivially be decomposed along a trajectory $\tau = (s_{0}, s_{1}, \ldots, s_{T}, \terminal)$ as
\begin{equation}
    \gE(s_{T}) = \sum_{t=0}^{T-1}\big(\gE(s_{t+1}) - \gE(s_{t})\big),
    \label{eq:energy-decomposition-mdb}
\end{equation}
provided that $\gE(s_{0}) = 0$; this is not a limiting assumption, since we can always subtract a constant offset $\gE(s_{0})$ to the energy at all states without changing the underlying Gibbs distribution. We will see in \cref{sec:dag-gfn-modularity-computational-efficiency} that in some cases the difference in energies between two consecutive states can be computed efficiently. This difference of energies also appears in the acceptance probability of Metropolis-Hastings in \cref{sec:existing-approaches-sampling-ebm}. Using this intermediate signal $\gE(s_{t}\rightarrow s_{t+1}) = \gE(s_{t+1}) - \gE(s_{t})$ in the reward function \cref{eq:maxentrl-correction-dense-fldb}, we can rewrite it as
\begin{align}
    r(s_{t}, s_{t+1}) &= \gE(s_{t}) - \gE(s_{t+1}) + \alpha \log P_{B}(s_{t}\mid s_{t+1}) &&& r(s_{T}, \terminal) &= 0.
    \label{eq:maxentrl-correction-dense-mdb}
\end{align}
With this particular choice of rewards, and especially because no reward is received upon termination, we can express the objective of Soft Q-Learning as a function of a policy $\pi_{\phi}(s'\mid s)$ parametrized by $\phi$, instead of a state-value function $Q_{\soft}^{\phi}(s, s')$; we call this method \gls{pisql}.

\begin{proposition}[$\pi$-SQL]
    \label{prop:pi-sql}
    Let $\gM = (\gG, r)$ be a MDP such that all the states of $\gG = (\widebar{\gS}, \gA)$ are terminating (\ie $\gS \equiv \gX$), and the reward function satisfies $r(s, \terminal) = 0$ for all states $s$. Then the residual appearing in the objective of Soft Q-Learning (\ref{eq:residual-sql}; \citealp{haarnoja2017sql}) can be written as a function of a policy $\pi_{\phi}: \gS \rightarrow \Delta(\children_{\gG})$: for any transition $s\rightarrow s'\in \gA$ such that $s'\neq \terminal$
    \begin{equation}
        \Delta_{\pisql}(s\rightarrow s';\phi) = \alpha \big[\log \pi_{\phi}(s'\mid s) - \log \pi_{\phi}(\terminal \mid s) + \log \pi_{\phi}(\terminal \mid s')\big] - r(s, s').
        \label{eq:residual-pisql}
    \end{equation}
\end{proposition}

\begin{proof}
    Since we assume that $r(s, \terminal) = 0$, we can enforce the fact that $Q_{\soft}^{\phi}(s, \terminal) = 0$ in our parametrization of the state-action value function used in the residual of SQL; this grounding at the terminal transitions allows us to identify the state-action value function with a policy. If we define a policy $\pi_{\phi}$ as
    \begin{equation}
        \pi_{\phi}(s'\mid s) = \exp\left(\frac{1}{\alpha}\big(Q_{\soft}^{\phi}(s, s') - V_{\soft}^{\phi}(s)\big)\right),
    \end{equation}
    then we have in particular $\pi_{\phi}(\terminal \mid s) = \exp(-V_{\soft}^{\phi}(s)/\alpha)$ based on our observation above. Moreover, we can write the different value functions appearing in the residual of SQL \cref{eq:residual-sql} as a function of $\pi_{\phi}$ alone
    \begin{align}
        Q_{\soft}^{\phi}(s, s') - V_{\soft}^{\phi}(s') &= Q_{\soft}^{\phi}(s, s') - V_{\soft}^{\phi}(s) + V_{\soft}^{\phi}(s) - V_{\soft}^{\phi}(s')\\
        &= \alpha\big[\log \pi_{\phi}(s'\mid s) - \log \pi_{\phi}(\terminal \mid s) + \log \pi_{\phi}(\terminal \mid s')\big]
    \end{align}
    which concludes the proof, by substituting this in the residual of SQL.
\end{proof}

We will see in \cref{sec:modified-detailed-balance} that when all the states are terminating, it is also possible to modify the detailed balance loss to only depend on a forward transition probability $P_{F}^{\phi}(s'\mid s)$, without any state flow function \citep{deleu2022daggflownet}. With an energy function, the corresponding residual can be written, for $s\rightarrow s'\in\gG$ such that $s'\neq \terminal$, as
\begin{equation}
    \Delta_{\mdb}(s\rightarrow s';\phi) = \log \frac{P_{F}^{\phi}(s'\mid s)P_{F}^{\phi}(\terminal\mid s')}{P_{B}(s\mid s')P_{F}^{\phi}(\terminal\mid s)} + \frac{\gE(s') - \gE(s)}{\alpha}.
    \label{eq:residual-mdb-maxentrl}\index{Detailed balance!Modified detailed balance}\index{Modified detailed balance|see {Detailed balance}}
\end{equation}
It can be shown that this modified detailed balance objective is a special case of the forward-looking detailed balance objective introduced in \cref{sec:equivalence-sql-fldb} \citep{pan2023flgfn}. The following proposition establishes an equivalence between the modified detailed balance loss and $\pi$-SQL with the reward function \cref{eq:maxentrl-correction-dense-mdb}.
\begin{proposition}
    \label{prop:equivalence-pisql-mdb}
    The modified detailed balance objective (GFlowNet; \citealp{deleu2022daggflownet}) is proportional to the Soft Q-Learning objective with policy parametrization (MaxEnt RL; \cref{prop:pi-sql}) on the MDP with the reward function defined in \cref{eq:maxentrl-correction-dense-mdb}, in the sense that $\gL_{\pisql}(\phi) = \alpha^{2}\gL_{\mdb}(\phi)$, with the correspondence $\pi_{\phi}(s'\mid s) = P_{F}^{\phi}(s'\mid s)$.
\end{proposition}

\begin{proof}
    The proof is similar to the proof of \cref{prop:equivalence-sql-fldb}. Let $s\rightarrow s'\in\gG$ be a transition such that $s'\neq \terminal$. With our choice of reward function in \cref{eq:maxentrl-correction-dense-mdb} we know that $r(s, s') = \gE(s) - \gE(s') + \alpha \log P_{B}(s\mid s')$. Substituting it into the residual of $\pi$-SQL in \cref{eq:residual-pisql}, we get
    \begin{align}
        &\Delta_{\pisql}(s\rightarrow s';\phi) = \alpha\big[\log \pi_{\phi}(s'\mid s) - \log \pi_{\phi}(\terminal \mid s) + \log \pi_{\phi}(\terminal\mid s')\big] - r(s, s')\\
        &\qquad= \alpha\big[\log \pi_{\phi}(s'\mid s) - \log \pi_{\phi}(\terminal \mid s) + \log \pi_{\phi}(\terminal\mid s')\big] - \big[\gE(s) - \gE(s') + \alpha \log P_{B}(s\mid s')]\\
        &\qquad= \alpha\left[\log \pi_{\phi}(s'\mid s) + \log \pi_{\phi}(\terminal \mid s') - \log P_{B}(s\mid s') - \log \pi_{\phi}(\terminal \mid s) + \frac{\gE(s') - \gE(s)}{\alpha}\right]\\
        &\qquad= \alpha\left[\log \frac{P_{F}^{\phi}(s'\mid s)P_{F}^{\phi}(\terminal\mid s')}{P_{B}(s\mid s')P_{F}^{\phi}(\terminal \mid s)} + \frac{\gE(s') - \gE(s)}{\alpha}\right] = \alpha \Delta_{\mdb}(s\rightarrow s'; \phi)
    \end{align}
    which concludes the proof, showing that $\gL_{\pisql}(\phi) = \alpha^{2}\gL_{\mdb}(\phi)$.
\end{proof}

\section{Relative entropy regularization}
\label{sec:relative-entropy-regularization-maxent-rl}
Generalizing the notion of entropy-regularized reinforcement learning considered so far in this chapter, we can consider the problem of reinforcement learning regularized by the \emph{relative entropy} with an ``anchor'' policy $\mu$. The optimization problem then becomes
\begin{equation}
    \pi_{\mathrm{RelEnt}}^{\star} = \argmax_{\pi} \E_{\tau\sim \pi}\left[\sum_{t=0}^{T}r(s_{t}, s_{t+1}) - \alpha \mathrm{KL}\big(\pi(\cdot \mid s_{t})\,\|\,\mu(\cdot\mid s_{t})\big) \right].
    \label{eq:kl-regularized-rl}
\end{equation}
This type of relative entropy regularization has gained popularity recently with \emph{reinforcement learning from human feedback} (\gls{rlhf}; \citealp{ziegler2019rlhf,stiennon2020rlhfsummarize,bai2022assistantsanthropic,rafailov2023dpo}), a framework for aligning AI models. Similar to how the optimal policy in MaxEnt RL could be expressed in terms of soft value functions, it can be shown that the optimal policy maximizing \cref{eq:kl-regularized-rl} can be written, for any transition $s \rightarrow s' \in \gG$, as
\begin{equation}
    \pi^{\star}_{\mathrm{RelEnt}}(s'\mid s) = \mu(s'\mid s)\exp\left(\frac{1}{\alpha}\big(Q^{\star}_{\soft}(s, s') - V^{\star}_{\soft}(s)\big)\right),
    \label{eq:kl-regularized-optimal-policy}
\end{equation}
where $Q^{\star}_{\soft}(s, s')$ and $V^{\star}_{\soft}(s)$ satisfy some generalization of the soft Bellman optimality equations given in \cref{eq:soft-bellman-optimality-equations} \& \cref{eq:soft-bellman-optimality-equations-V}, taking the following form:
\begin{align}
    Q^{\star}_{\soft}(s, s') &= r(s, s') + V^{\star}_{\soft}(s')\label{eq:soft-bellman-optimality-kl}\\
    V^{\star}_{\soft}(s) &= \alpha \log \sum_{s'\in\children_{\gG}(s)}\mu(s'\mid s)\exp\left(\frac{1}{\alpha}Q^{\star}_{\soft}(s, s')\right).\label{eq:soft-bellman-optimality-kl-V}
\end{align}
Contrary to \cref{sec:gflownet-maxentrl-reward-correction}, we will consider in this section the case where the reward function of the MDP is \emph{uncorrected}, meaning that $\sum_{t=0}^{T}r(s_{t}, s_{t+1}) = -\gE(s_{T})$. Note that the case where $\mu$ is the uniform policy does not recover exactly the MaxEnt RL objective in \cref{eq:maxent-rl-objective-2}, despite the relative entropy to the uniform distribution being tightly related to the entropy; the reason is that $\mu(s'\mid s) = 1/|\children_{\gG}(s)|$ is not independent of the state $s$, and it would introduce an extra term depending on $\pi$ in \cref{eq:kl-regularized-rl}.

\subsection{Terminating state distribution associated with the optimal policy}
\label{sec:terminating-state-distribution-optimal-policy-kl}
The following proposition shows that the terminating state distribution associated with the optimal policy $\pi^{\star}_{\mathrm{RelEnt}}$ is related to the terminating state distribution associated with the anchor policy $\mu$.
\begin{proposition}
    \label{prop:terminating-state-optimal-policy-kl-anchor}
    Let $\gM = (\gG, r)$ be a MDP with a pointed DAG structure $\gG$, where the reward function satisfies, for any complete trajectory $\tau = (s_{0}, s_{1}, \ldots, s_{T}, \terminal)$, $\sum_{t=0}^{T}r(s_{t}, s_{t+1}) = -\gE(s_{T})$ (with the convention $s_{T+1} = \terminal$). Then the terminating state distribution associated with the optimal policy $\pi^{\star}_{\mathrm{RelEnt}}$ solution of \cref{eq:kl-regularized-rl} satisfies for all $x\in\gX$
    \begin{equation}
        \pi^{\star\top}_{\mathrm{RelEnt}}(x) \propto \mu^{\top}(x)\exp(-\gE(x)/\alpha).
    \end{equation}
\end{proposition}
\begin{proof}
    The proof is similar to the proof of \cref{thm:reward-correction-maxentrl}. We can again write the optimal policy $\pi^{\star}_{\mathrm{RelEnt}}$ in terms of the state value function $V^{\star}_{\soft}$ alone, using \cref{eq:soft-bellman-optimality-kl}:
    \begin{equation}
        \pi^{\star}_{\mathrm{RelEnt}}(s'\mid s) = \mu(s'\mid s)\exp\left(\frac{1}{\alpha}\big(r(s, s') + V^{\star}_{\soft}(s') - V^{\star}_{\soft}(s)\big)\right).
    \end{equation}
    By \cref{def:terminating-state-probability}, we can write the terminating state distribution associated with $\pi^{\star}_{\mathrm{RelEnt}}$ for any terminating state $x\in\gX$ as
    {\allowdisplaybreaks%
    \begin{align}
        \pi^{\star\top}_{\mathrm{RelEnt}}(x) &= \sum_{\tau: s_{0}\rightsquigarrow x}\prod_{t=0}^{T_{\tau}}\pi^{\star}_{\mathrm{RelEnt}}(s_{t+1}\mid s_{t})\\
        &= \sum_{\tau:s_{0}\rightsquigarrow x}\underbrace{\left[\prod_{t=0}^{T_{\tau}}\mu(s_{t+1}\mid s_{t})\right]}_{=\,\mu(\tau)}\exp\left(\frac{1}{\alpha}\sum_{t=0}^{T_{\tau}}\big(r(s_{t}, s_{t+1}) + V^{\star}_{\soft}(s_{t+1}) - V^{\star}_{\soft}(s_{t})\big)\right)\\
        &= \sum_{\tau:s_{0}\rightsquigarrow x}\mu(\tau)\exp\Bigg(\frac{1}{\alpha}\underbrace{\sum_{t=0}^{T_{\tau}}r(s_{t}, s_{t+1})}_{=\,-\mathcal{E}(x)} + \underbrace{\vphantom{\sum_{t=0}}V^{\star}_{\soft}(\terminal)}_{=\,0} - V^{\star}_{\soft}(s_{0})\Bigg)\\
        &= \frac{\exp(-\mathcal{E}(x)/\alpha)}{\exp(V^{\star}_{\soft}(s_{0})/\alpha)}\underbrace{\sum_{\tau:s_{0}\rightsquigarrow x}\mu(\tau)}_{=\,\mu^{\top}(x)} \propto \mu^{\top}(x)\exp(-\gE(x)/\alpha).
    \end{align}}%
\end{proof}
A similar result can also been found in various contexts in the literature on language model alignment in the case of autoregressive generation \citep{korbak2022rlwithklbayesian,mudgal2024controlleddecoding,rafailov2024fromrtoqstar}. Following our remark above noting that the case where the anchor policy $\mu$ is uniform does not correspond exactly to MaxEnt RL, it is also important to note that this proposition does not relate to our previous observation from \cref{sec:multi-path-environment-biased-sampling} that the terminating state distribution associated with $\pi^{\star}_{\maxent}$ is biased (towards terminating states having multiple trajectories leading to them) in the case where the reward function is uncorrected. Indeed, in general $\mu^{\top}(x) \not\propto n(x)$ if $\mu$ is the uniform policy, where $n(x)$ is the number of complete trajectories leading to $x\in\gX$.

\subsection{Consequences for Bayesian inference}
\label{sec:kl-regularized-consequence-bayesian-inference}
One potential application of this result with relative-entropy regularization is in Bayesian inference\index{Bayesian inference} \citep{korbak2022rlwithklbayesian}. Recall that by Bayes rule, we know that for some dataset of observations $\gD_{n}$ and a new observation $x_{n+1}$, we can write the posterior distribution of a model $\theta$ wrt. the new dataset $\gD_{n+1} = \gD_{n} \cup \{x_{n+1}\}$ (with the convention $\gD_{0} = \emptyset$) in terms of the posterior wrt.~$\gD_{n}$
\begin{equation}
    P(\theta \mid \gD_{n+1}) \propto P(\theta\mid \gD_{n})P(x_{n+1}\mid \theta),
    \label{eq:bayes-rule-update-datasets}
\end{equation}
provided the observations are mutually independent given $\theta$. If we model the posterior $P(\theta\mid \gD_{n})$ using a GFlowNet (or MaxEnt RL, using the techniques described in \cref{sec:unbiased-sampling-maxent-rl} \& \ref{sec:equivalence-maxent-rl-gflownet}), then this amounts to finding a policy $\pi_{n}$ such that its terminating state distribution matches the posterior: $\pi_{n}^{\top}(\theta) = P(\theta\mid \gD_{n})$. If we know the policy $\pi_{n}$, then \cref{prop:terminating-state-optimal-policy-kl-anchor} gives us a strategy to find a policy $\pi_{n+1}$ such that its terminating state distribution matches the posterior wrt. the updated dataset $\gD_{n+1}$: we simply have to find the optimal policy maximizing the following objective:
\begin{equation}
    \pi_{n+1} = \argmax_{\pi}\E_{\tau\sim \pi}\left[\sum_{t=0}^{T}r(s_{t}, s_{t+1}) - \mathrm{KL}\big(\pi(\cdot \mid s_{t})\,\|\,\pi_{n}(\cdot\mid s_{t})\big)\right],
    \label{eq:kl-reg-rl-bayesian-problem}
\end{equation}
where we set $\alpha = 1$, with $\pi_{n}$ serving as the anchor policy, and where the reward function is defined such that the corresponding return is equal to the log-likelihood of the new datapoint
\begin{equation}
    \sum_{t=0}^{T}r(s_{t}, s_{t+1}) = \log P(x_{n+1}\mid \theta),
\end{equation}
where $s_{T} = \theta$ is the terminating state, and $s_{T+1} = \terminal$. This guarantees that $\pi_{n+1}^{\top}(\theta) = P(\theta\mid \gD_{n+1})$. In order to find the solution of \cref{eq:kl-reg-rl-bayesian-problem}, we can use any algorithm similar to the ones presented in the previous section specifically adapted to the relative entropy regularization; for example, the recently introduced \emph{relative trajectory balance} objective \citep{venkatraman2024rtb} can be interpreted as Path Consistency Learning applied to the problem of KL-regularized reinforcement learning.

To find the initial policy $\pi_{0}$ whose terminating state distribution should match the prior $P(\theta)$, we can use any GFlowNet method described in \cref{chap:generative-flow-networks}, or their MaxEnt RL counterparts. In particular, if the prior distribution is uniform, we can simply train a GFlowNet with a constant reward $R(\theta) = 1$ for all models (again, $R$ here being a reward in the sense of GFlowNets \cref{eq:target-distribution-gflownet-reward}, not the reward $r$ in the reinforcement learning counterpart). This will be typically easier than training a GFlowNet to approximate the posterior $P(\theta\mid \gD)$ directly (with the reward $R(\theta) = P(\gD, \theta)$), since the prior is completely data-independent. We will come back to the practical challenges of training a GFlowNet for Bayesian inference in \cref{sec:limitations-gflownets-bayesian-inference}.

%% file: figures/chapter5/equivalences.tex
\begin{NoHyper}
\begin{adjustbox}{center}
\begin{tikzpicture}[x=10pt, y=10pt]
\useasboundingbox (0, 0) rectangle (48, 12);  
\node[] at (24, 6) {\begin{tikzpicture}[every node/.style={inner sep=0pt}, gfn_edge/.style={ultra thick, -latex}, alg_node/.style={thick, circle, draw, minimum size=41pt, fill=white, inner sep=2pt, align=center}, intra_alg_edge/.style={thick, -latex}, inter_alg_edge/.style={latex-latex, line width=1.5pt}, y=70pt, x=85pt]

\usetikzlibrary{positioning}
\pgfdeclarelayer{background}
\pgfsetlayers{background,main}

\definecolor{groupcolor1}{HTML}{225ea8}
\definecolor{groupcolor2}{HTML}{41b6c4}
\definecolor{groupcolor3}{HTML}{a1dab4}

\node[alg_node] (pcl) at (1, 1) {};
\node[] at (pcl.center) {\begin{tikzpicture}
    \node[scale=1] (title_pcl) {PCL};
    \node[below=2pt of title_pcl, scale=0.3] {\citep{nachum2017pcl}};
\end{tikzpicture}};

\node[alg_node] (subtb) at (1, 0) {};
\node[] at (subtb.center) {\begin{tikzpicture}
    \node[scale=1] (title_subtb) {SubTB};
    \node[below=2pt of title_subtb, scale=0.3] {\citep{madan2022subtb}};
\end{tikzpicture}};

\node[alg_node] (tb) at (0, 0) {};
\node[] at (tb.center) {\begin{tikzpicture}
    \node[scale=1] (title_tb) {TB};
    \node[below=2pt of title_tb, scale=0.3] {\citep{malkin2022trajectorybalance}};
\end{tikzpicture}};

\node[alg_node] (sql) at (2.5, 1) {};
\node[] at (sql.center) {\begin{tikzpicture}
    \node[scale=1] (title_sql) {SQL};
    \node[below=2pt of title_sql, scale=0.3] {\citep{haarnoja2017sql}};
\end{tikzpicture}};

\node[alg_node] (db) at (2, 0) {};
\node[] at (db.center) {\begin{tikzpicture}
    \node[scale=1] (title_db) {DB};
    \node[below=2pt of title_db, scale=0.3] {\citep{bengio2023gflownetfoundations}};
\end{tikzpicture}};

\node[alg_node] (fl_db) at (3, 0) {};
\node[] at (fl_db.center) {\begin{tikzpicture}
    \node[scale=1] (title_fl_db) {FL-DB};
    \node[below=2pt of title_fl_db, scale=0.3] {\citep{pan2023flgfn}};
\end{tikzpicture}};

\node[alg_node] (policy_sql) at (4, 1) {};
\node[] at (policy_sql.center) {\begin{tikzpicture}
    \node[scale=1] (title_policy_sql) {$\pi$-SQL};
\end{tikzpicture}};

\node[alg_node] (modified_db) at (4, 0) {};
\node[] at (modified_db.center) {\begin{tikzpicture}
    \node[scale=0.8, align=center] (title_modified_db) {Modified\\[-0.5ex]DB};
    \node[below=2pt of title_modified_db, scale=0.3] {\citep{deleu2022daggflownet}};
\end{tikzpicture}};

\draw[inter_alg_edge, draw=groupcolor1] (pcl) -- (subtb) node[left, midway, scale=0.5, align=center, anchor=east, inner sep=8pt] {Equivalence with\\corrected reward \cref{eq:reward-correction-maxentrl}\\\cref{prop:equivalence-pcl-subtb}};
\draw[inter_alg_edge, densely dashed, draw=groupcolor1] (pcl) to[bend right=30] (tb);
\draw[intra_alg_edge] (subtb) -- (tb) node[below, midway, scale=0.5, align=center, inner sep=8pt] {Complete\\trajectories\\\citep{malkin2022trajectorybalance}};
\draw[intra_alg_edge] (pcl) -- (sql) node[below, midway, scale=0.5, align=center, inner sep=8pt] {Unified PCL\\rollout $d=1$\\\citep{nachum2017pcl}};
\draw[intra_alg_edge] (subtb) -- (db) node[below, midway, scale=0.5, align=center, inner sep=8pt] {Subtrajectories\\of length 1\\\citep{malkin2022trajectorybalance}};
\draw[inter_alg_edge, densely dashed, draw=groupcolor2] (sql) -- (db);
\draw[inter_alg_edge, densely dashed, draw=groupcolor2] (sql) -- (fl_db);
\draw[intra_alg_edge] (sql) -- (policy_sql) node[below, midway, scale=0.5, align=center, inner sep=8pt] {Policy\\parametrization\\\cref{sec:equivalence-pisql-mdb}};
\draw[intra_alg_edge] (db) -- (fl_db) node[below, midway, scale=0.5, align=center, inner sep=8pt] {Intermediate\\rewards\\\citep{pan2023flgfn}};
\draw[intra_alg_edge] (fl_db) -- (modified_db) node[below, midway, scale=0.5, align=center, inner sep=8pt] {All states are\\terminating\\\citep{deleu2022daggflownet}};
\draw[inter_alg_edge, densely dashed, draw=groupcolor3] (policy_sql) -- (modified_db);

\node[fit=(pcl)(subtb)(tb)] (trajectories_group) {};
\node[fit=(sql)(fl_db)(db)] (transitions_group) {};
\node[fit=(policy_sql)(modified_db)] (all_terminating_group) {};

\node[anchor=south, scale=0.8, yshift=8pt] (trajectories_group_title) at (trajectories_group.{north}) {Trajectories};
\node[anchor=south, scale=0.8, yshift=8pt] (transitions_group_title) at (transitions_group.{north}) {\vphantom{j}Transitions};
\node[anchor=south, scale=0.8, yshift=8pt] (all_terminating_group_title) at (all_terminating_group.{north}) {\vphantom{j}$\gS \equiv \gX$};

\begin{pgfonlayer}{background}
\node[fit=(trajectories_group)(trajectories_group_title), fill=groupcolor1!20, inner sep=8pt, rounded corners=3pt, minimum width=160pt] {};
\node[fit=(transitions_group)(transitions_group_title), fill=groupcolor2!20, inner sep=8pt, rounded corners=3pt, minimum width=160pt] {};
\node[fit=(all_terminating_group)(all_terminating_group_title), fill=groupcolor3!20, inner sep=8pt, rounded corners=3pt, minimum width=75pt] (all_terminating_group_fill) {};
\end{pgfonlayer}

\coordinate (mid) at ($(pcl.south)!0.5!(subtb.north)$);
\coordinate[xshift=10pt] (mid_right) at (mid -| all_terminating_group_fill.east);
\draw[|-|, thick, shorten <=-5pt, shorten >=-5pt] (policy_sql.north -| mid_right) -- (policy_sql.south -| mid_right) node[midway, above, sloped, scale=0.8, font=\bfseries, inner sep=8pt] {MaxEnt RL};
\draw[|-|, thick, shorten <=-5pt, shorten >=-5pt] (modified_db.north -| mid_right) -- (modified_db.south -| mid_right) node[midway, above, sloped, scale=0.8, font=\bfseries, inner sep=8pt] {GFlowNet};

\end{tikzpicture}};
\end{tikzpicture}
\end{adjustbox}
\end{NoHyper}

%% file: chapters/06_GFlowNets_General_State_Spaces.tex
\chapter{GFlowNets over General State Spaces}
\label{chap:gflownets-general-state-spaces}

\begin{minipage}{\textwidth}
    \itshape
    This chapter contains material from the following papers:
    \begin{itemize}[noitemsep, topsep=1ex, itemsep=1ex, leftmargin=3em]
        \item Salem Lahlou, \textbf{Tristan Deleu}, Pablo Lemos, Dinghuai Zhang, Alexandra Volokhova, Alex Hern\'{a}ndez-Garc\'{i}a, L\'{e}na N\'{e}hale Ezzine, Yoshua Bengio, Nikolay Malkin (2023). \emph{A Theory of Continuous Generative Flow Networks}. International Conference on Machine Learning (ICML). \notecite{lahlou2023continuousgfn}
        \item \textbf{Tristan Deleu}, Yoshua Bengio (2023). \emph{Generative Flow Networks: a Markov Chain Perspective}. \notecite{deleu2023gfnmarkovchain}
    \end{itemize}
    \vspace*{5em}
\end{minipage}

So far, we have assumed that the elements of the sample space $\gX$ were discrete and had some compositional structure. This was mainly due to the fact that this case is relatively less explored in the context of variational inference, which has found more applications in continuous settings thanks to the reparametrization trick, as we discussed in \cref{sec:estimation-parameters}. To expand the scope of applications of generative flow networks, we extend in this chapter the results we have seen previously to more general state spaces; this includes, notably, continuous spaces but also mixtures of discrete (with some compositional aspect) and continuous spaces. The key takeaway of this chapter is that under some conditions, the flow matching losses that were introduced for discrete GFlowNets will still hold in the general case, possibly with minimal modifications. This will play a critical role in \cref{chap:jsp-gfn}, where we will be able to perform Bayesian inference over the structure and parameters of Bayesian networks under the unique framework of GFlowNets.

\section{Generative flow networks over discrete spaces}
\label{sec:gflownet-discrete-spaces}
Before going deeper into the extensions of generative flow networks to more general spaces, we will briefly revisit what we have seen so far about discrete GFlowNets from a different point of view. In this section, we will show that everything we have achieved with flow networks and pointed DAGs can be cast from the perspective of \emph{Markov chains}.

\subsection{Revisiting the flow matching condition}
\label{sec:general-revisiting-flow-matching}
We saw in \cref{def:transition-probabilities-from-flow} that the forward transition probability is defined in terms of a flow function by normalizing the outgoing edge flow. Conversely, the edge flow $F^{\star}(s\rightarrow s')$ can be viewed as the product of the state flow and the forward transition probability: $F^{\star}(s\rightarrow s') = F^{\star}(s)P^{\star}_{F}(s'\mid s)$; this is an observation we have already made for the boundary condition of detailed balance in \cref{sec:alternative-conditions}. To characterize a valid Markovian flow, we can rewrite the flow matching condition of \cref{thm:flow-matching-condition} not in terms of a conservation law between the incoming and outgoing edge flows, but using the decomposition above in conjunction with \cref{prop:identities-state-edge-flows}. The following proposition provides an alternative perspective on the flow matching condition, in terms of state flows and forward transition probabilities.

\begin{proposition}[Flow matching condition]
    \label{prop:flow-matching-condition-state-flow-pF}\index{Flow matching!Condition}
    Let $\gG = (\widebar{\gS}, \gA)$ be a pointed DAG. A function $F: \gS \rightarrow \sR_{+}$ defines the state flow, and $P_{F}: \gS \rightarrow \Delta(\children_{\gG})$ the forward transition probabilities of a unique Markovian flow $F^{\star}$ if and only if they satisfy the following condition for all $s\in\gS$ such that $s'\neq s_{0}$:
    \begin{equation}
        F(s') = \sum_{s\in\parents_{\gG}(s')}F(s)P_{F}(s'\mid s).
        \label{eq:flow-matching-condition-state-flow-pF}
    \end{equation}
    Moreover, under these conditions, the unique Markovian flow $F^{\star}$ whose state flow and forward transition probabilities match $F$ and $P_{F}$ respectively is defined, for any complete trajectory $\tau = (s_{0}, s_{1}, \ldots, s_{T}, \terminal)$, by
    \begin{equation}
        F^{\star}(\tau) = F(s_{0})\prod_{t=0}^{T}P_{F}(s_{t+1}\mid s_{t}),
        \label{eq:flow-matching-condition-state-flow-pF-markovian-flow}
    \end{equation}
    with the convention $s_{T+1} = \terminal$.
\end{proposition}

The proof of this proposition is almost identical to the proof of \cref{thm:flow-matching-condition}, and is deferred to \cref{app:alternative-conditions} for completeness. Up until now, we considered that the codomain of a forward transition probability $\Delta(\children_{\gG})$ was dependent on the current state $s$ (the children of $s$). Alternatively, we could simply write the flow matching condition in \cref{eq:flow-matching-condition-state-flow-pF} as %
\begin{equation}
    F(s') = \sum_{s\in\gS}F(s)P_{F}(s'\mid s),
\end{equation}
for all $s' \neq s_{0}$, where we use the convention $P_{F}(s'\mid s) = 0$ if $s\notin \parents_{\gG}(s')$. With the exception of the condition on $s'$ not being the initial state, the equation above looks oddly similar to the definition of $F$ being \emph{invariant} under the forward transition probability $P_{F}$. Recall that a vector $\mF \in \sR^{|\gS|}$ is said to be invariant under a stochastic matrix $\mP_{F}$ called the \emph{transition matrix} whose rows are $P_{F}(\cdot \mid s)$ if and only if $\mF = [\mP_{F}]^{\top}\mF$; note that we use the notation $[\mP_{F}]^{\top}$ to denote the transpose of $\mP_{F}$ to avoid any confusion with the terminating state distribution associated with $P_{F}$. In order to prepare for the generalization of GFlowNets to arbitrary state spaces, we will therefore make a detour first and view a (discrete) GFlowNet \emph{as if} $F$ was indeed the invariant measure of a certain transition matrix.

\subsection{The cyclic nature of flow networks}
\label{sec:cyclic-nature-flow-networks}
The way we sample from a Markovian flow network is to follow its forward transition probability until a terminating transition is taken (\cref{alg:sampling-terminating-state-probability}), and then restart again from the initial state in order to get a new (independent) sample. Instead of treating independent samples as being completely separate trajectories in a pointed DAG $\gG$, we can view this process as being a \emph{single} Markov chain that regenerates every time it reaches a special state. We will denote by $(X_{n})_{n\geq 0}$ the canonical (homogeneous) Markov chain with transition probability $P_{F}$ and starting at $s_{0}$:
\begin{equation}
    \sP_{s_{0}}(X_{n+1} = s'\mid X_{0} = s_{0}, X_{1} = s_{1}, \ldots, X_{n} = s) = \sP_{s_{0}}(X_{n+1} = s'\mid X_{n} = s) = P_{F}(s'\mid s),
\end{equation}
by the Markov property, and where $\sP_{s_{0}}(X_{0} = s) = \delta_{s_{0}}(s)$ is the Dirac measure at $s_{0}$. We use the notation ``$\sP_{s_{0}}$'' to denote the joint distribution of the chain (and later ``$\E_{s_{0}}$'' to denote expectations under this distribution). In the context of Markov chains, $P_{F}$ is also called a \emph{Markov kernel}.

\begin{figure}[t]
    \centering
    \begin{adjustbox}{center}
        \includegraphics[width=480pt]{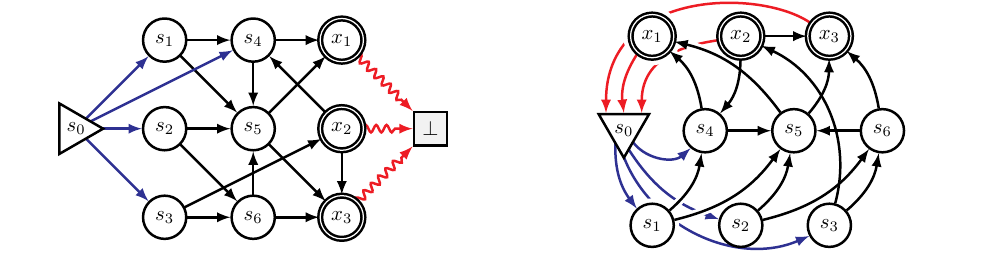}%
    \end{adjustbox}
    \begin{adjustbox}{center}%
        \begin{subfigure}[b]{240pt}%
            \caption{}%
            \label{fig:wrapped-around-gflownet-before}%
        \end{subfigure}%
        \begin{subfigure}[b]{240pt}%
            \caption{}%
            \label{fig:wrapped-around-gflownet-after}%
        \end{subfigure}%
    \end{adjustbox}
    \caption[Discrete GFlowNet as a recurrent Markov chain]{Discrete GFlowNet as a recurrent Markov chain. (a) The pointed DAG structure of a GFlowNet, with the flow going from the initial state $s_{0}$ to the terminal state $\terminal$. (b) The ``wrapped-around'' version of the same GFlowNet, where the initial state has been merged with the terminal state $s_{0}\equiv \terminal$. We can place a recurrent Markov chain over this state space that regenerates every time it hits $s_{0}$.}
    \label{fig:wrapped-around-gflownet}
\end{figure}

A flow network represented as a pointed DAG $\gG$ with a forward transition probability $P_{F}$ can be transformed into such a Markov chain by simply ``wrapping around'' the graph and merging the terminal state $\terminal$ with the initial state $s_{0}$, as depicted in \cref{fig:wrapped-around-gflownet}. It is clear that whenever this Markov chain goes through $s_{0} \equiv \terminal$, this would be effectively equivalent to starting a new complete trajectory in our standard formulation. Furthermore, $\gG$ was originally a DAG where all the complete trajectories are guaranteed to eventually end at the terminal state. This ensures that $s_{0}\in\gS$ is a \emph{recurrent state}, meaning that the Markov chain eventually comes back to $s_{0}$ with probability 1:
\begin{equation}
    \sP_{s_{0}}(\exists n \geq 1,\,X_{n} = s_{0}) = 1.
    \label{eq:recurrent-state-markov-chain}
\end{equation}
More than being simply recurrent, the Markov chain is also \emph{positive recurrent}\index{Recurrence!Positive recurrence}, meaning the expected time to return to $s_{0}$ is finite. Moreover, since we assumed in \cref{sec:elements-graph-theory} that $\gG$ was structured in such a way that all the states of $\gS$ are accessible from the initial state (in the sense that there exists a partial trajectory), and that the terminal state is also itself accessible from all the states, this Markov chain is also clearly \emph{irreducible}\index{Irreducibility}, meaning that all the states are accessible from any other state (possibly via $s_{0} \equiv \terminal$).

\subsection{Generative flow network as a Markov chain}
\label{sec:gflownet-as-markov-chain}
In the previous section, we showed how we could transform a flow network in order to create an equivalent irreducible and positive recurrent Markov chain. In this section, we will do the opposite: starting with a Markov chain with those properties, we will show that under appropriate boundary conditions we can sample objects proportionally to some reward function $R(x)$, just like we did with GFlowNets. We will assume here that the state space $\gS$ is discrete, but not necessarily finite anymore, and that we have a Markov chain with transition probability $P_{F}$ that is \emph{irreducible} and \emph{positive recurrent}. Unlike the case where $\gS$ is finite though, positive recurrence is not an immediate property when the chain is merely recurrent and $\gS$ is infinite \citep{deleu2023gfnmarkovchain}.

The fact that the Markov chain is recurrent guarantees that it will eventually return back to its initial state $s_{0}$ an infinite amount of times. We can define the \emph{return time} $\sigma_{s_{0}}$\glsadd{returntime} as being the (random) time the chain first goes back to the initial state:
\begin{equation}
    \sigma_{s_{0}} \triangleq \inf\{n \geq 1\mid X_{n} = s_{0}\}.
    \label{eq:return-time}
\end{equation}
Excursions of the Markov chain between two consecutive returns to $s_{0}$ are independent by the strong Markov property, since $\sigma_{s_{0}}$ is a \emph{stopping time} (\ie the decision whether the chain returns to $s_{0}$ for the first time does not depend on future information). In other words, the independence for two different trajectories through a flow network is preserved even with a single recurrent Markov chain running indefinitely. The return time $\sigma_{s_{0}}$ is guaranteed to be finite in expectation thanks to the positive recurrence of the chain (in particular, it is finite with probability $1$). Moreover, the irreducibility and positive recurrence of the chain ensure the existence of a measure $\lambda$ which is \emph{invariant} for $P_{F}$: $\lambda$ is a measure over $\gS$ that satisfies $\lambda P_{F} = \lambda$, in the sense that
\begin{equation}
    \lambda(s') = \sum_{s\in\gS}P_{F}(s'\mid s)\lambda(s),
    \label{eq:definition-invariant-measure-discrete}
\end{equation}
for all $s'\in\gS$. Note that $\lambda$ is not necessarily normalized. The following theorem shows the existence of a such an invariant measure, unique up to a multiplicative constant.
\begin{theorem}[Existence of an invariant measure; \citealp{douc2018markovchains}, Theorem 7.2.1]
    \label{thm:discrete-markov-chain-existance-invariant-measure}
    Let $(X_{n})_{n\geq 0}$ be an irreducible and positive recurrent Markov chain over a discrete state space $\gS$, starting at $s_{0}$ and with transition probability $P_{F}$. There exists a non-trivial invariant measure $\lambda$ for $P_{F}$ (\ie $\lambda P_{F} = \lambda$), unique up to a multiplicative positive constant, defined for all $s\in\gS$ by
    \begin{equation}
        \lambda(s) = \E_{s_{0}}\left[\sum_{n=0}^{\sigma_{s_{0}}-1}\mathds{1}(X_{n} = s)\right] = \sum_{n=0}^{\infty}\E_{s_{0}}\big[\mathds{1}(n < \sigma_{s_{0}})\mathds{1}(X_{n} = s)\big].
        \label{eq:discrete-markov-chain-existance-invariant-measure}
    \end{equation}
    In particular, $\lambda$ is the unique invariant measure for $P_{F}$ such that $\lambda(s_{0}) = 1$.
\end{theorem}

Intuitively, the way $\lambda(s)$ is constructed is by measuring, on average, how often the chain hits $s$ before going back to its initial state. This quantity is related to the \emph{occupancy measure} in the context of reinforcement learning \citep{szepesvari2020rltheory}. It may seem surprising that we could hit a state $s$ multiple times before returning to $s_{0}$ when we were working exclusively with pointed DAGs for GFlowNets (\ie we couldn't visit the same state multiple times since it would violate acyclicity). This will not invalidate our results further, since the most important property is the \emph{positive recurrence} of the chain, to guarantee that we can return to $s_{0}$ in finite time almost surely.

The same way we defined the terminating state distribution in \cref{def:terminating-state-probability} as being the marginal over trajectories terminating at a certain state, we can define it more generally for any positive recurrent Markov chain as being the distribution of states reached by the chain right before going back to $s_{0}$.

\begin{definition}[Terminating state distribution]
    \label{def:general-discrete-terminating-state-distribution}\index{Terminating state probability}
    Given an irreducible and positive recurrent Markov chain over $\gS$, with an initial state $s_{0}$ and a transition probability $P_{F}$, the \emph{terminating state probability distribution} is defined as the marginal distribution of the chain right before returning to its initial state: for all $x \in \gS$
    \begin{equation}
        P_{F}^{\top}(x) \triangleq \sP_{s_{0}}(X_{\sigma_{s_{0}}-1} = x) = \E_{s_{0}}\big[\mathds{1}(X_{\sigma_{s_{0}}-1} = x)\big].
    \end{equation}
\end{definition}

We can show that the terminating state distribution is again a properly defined probability distribution (\ie non-negative, and sums to 1); see \cref{prop:general-discrete-terminating-state-proper-distribution} for a result analogous to \cref{prop:terminating-state-proper-distribution} for flow networks, but this time working exclusively with Markov chains. The support of $P_{F}^{\top}$ is a subset $\gX \subseteq \gS$ corresponding to the states having non-zero probability of transitioning to $s_{0}$ (\ie the ``parents'' of $s_{0}$): $\gX = \{x\in\gS\mid P_{F}(s_{0}\mid x) > 0\}$. The terminating state distribution is actually related to invariant measures of $P_{F}$ in an interesting way.

\begin{proposition}
    \label{prop:terminating-state-distribution-invariant-measure}
    The terminating state distribution $P_{F}^{\top}$ is related to the invariant measure $\lambda$ defined in \cref{thm:discrete-markov-chain-existance-invariant-measure} by, $\forall x\in\gS$, $P_{F}^{\top}(x) = \lambda(x)P_{F}(s_{0}\mid x)$.
\end{proposition}
\begin{proof}
    For any $n \geq 0$ and any $x\in\gS$, we have
    {\allowdisplaybreaks%
    \begin{align}
        \E_{s_{0}}\big[\mathds{1}(n < \sigma_{s_{0}})\mathds{1}(X_{n} = x)\big]&P_{F}(s_{0}\mid x) = \E_{s_{0}}\big[\mathds{1}(n < \sigma_{s_{0}})\mathds{1}(X_{n} = x)P_{F}(s_{0}\mid X_{n})\big]\label{eq:terminating-state-distribution-invariant-measure-proof-1}\\
        &= \E_{s_{0}}\big[\mathds{1}(n < \sigma_{s_{0}})\mathds{1}(X_{n} = x)\E_{X_{n}}[\mathds{1}(X_{1} = s_{0})]\big]\label{eq:terminating-state-distribution-invariant-measure-proof-2}\\
        &= \E_{s_{0}}\big[\mathds{1}(n < \sigma_{s_{0}})\mathds{1}(X_{n} = x)\E_{s_{0}}[\mathds{1}(X_{n+1} = s_{0})\mid X_{0:n}]\big]\label{eq:terminating-state-distribution-invariant-measure-proof-3}\\
        &= \E_{s_{0}}\big[\mathds{1}(n < \sigma_{s_{0}})\mathds{1}(X_{n} = x)\mathds{1}(X_{n+1} = s_{0})\big]\label{eq:terminating-state-distribution-invariant-measure-proof-4}\\
        &= \E_{s_{0}}\big[\mathds{1}(\sigma_{s_{0}} = n+1)\mathds{1}(X_{n} = x)\big]\label{eq:terminating-state-distribution-invariant-measure-proof-5}.
    \end{align}}%
    In details, we used an equivalent definition of $P_{F}(s_{0}\mid X_{n})$ in \cref{eq:terminating-state-distribution-invariant-measure-proof-2}  as the expectation of the chain moving to $s_{0}$ after one step ($\mathds{1}(X_{1} = s_{0})$) starting at $X_{n}$, the Markov property in \cref{eq:terminating-state-distribution-invariant-measure-proof-3}, the law of total expectation in \cref{eq:terminating-state-distribution-invariant-measure-proof-4}, and finally the definition of the return time $\sigma_{s_{0}}$ \cref{eq:return-time} in \cref{eq:terminating-state-distribution-invariant-measure-proof-5}. Using the definition of the invariant measure $\lambda$, we then obtain the expected result:
    {\allowdisplaybreaks%
    \begin{align}
        \lambda(x)P_{F}(s_{0}\mid x) &= \sum_{n=0}^{\infty}\E_{s_{0}}\big[\mathds{1}(n < \sigma_{s_{0}})\mathds{1}(X_{n} = x)\big]P_{F}(s_{0}\mid x)\\
        &= \sum_{n=0}^{\infty}\E_{s_{0}}\big[\mathds{1}(\sigma_{s_{0}} = n+1)\mathds{1}(X_{n} = x)\big]\\
        &= \sum_{n=1}^{\infty}\E_{s_{0}}\big[\mathds{1}(\sigma_{s_{0}} = n)\mathds{1}(X_{n-1}=x)\big]\\
        &= \E_{s_{0}}\big[\mathds{1}(X_{\sigma_{s_{0}}-1}=x)\big] = P_{F}^{\top}(x).
    \end{align}}%
\end{proof}
It is important to observe that while it is related to an invariant measure, $P_{F}^{\top}(x)$ itself is \emph{not} invariant for $P_{F}$, despite being equal to $\lambda(x)$ up to a multiplicative factor (but dependent on $x$). This is the key difference with most of the literature on MCMC which constructs a Markov chain whose invariant distribution is the distribution of interest (see also \cref{sec:comparison-gflownets-mcmc}).

\paragraph{Boundary condition} Similar to the flow network perspective, our objective is not to find $P_{F}^{\top}$ given the Markov kernel $P_{F}$, but instead the inverse problem of finding a $P_{F}$ such that its corresponding terminating state distribution matches the target distribution (\ie the Gibbs distribution in \cref{eq:gibbs-distribution}). Just like we did in \cref{chap:generative-flow-networks}, we have to introduce \emph{boundary conditions} that the Markov kernel $P_{F}$ and one of its invariant measures $F$ must satisfy. In preparation for the more general case that we will see in \cref{sec:gflownet-general-state-spaces}, we will say that they satisfy the boundary condition for a finite measure $R$ over $\gX$ if for all $x\in\gX$
\begin{equation}
    F(x)P_{F}(s_{0}\mid x) = R(x).
    \label{eq:generalized-boundary-condition-discrete}
\end{equation}
Following the intuition that we ``wrapped-around'' the pointed DAG $\gG$ like in \cref{fig:wrapped-around-gflownet} with $s_{0} \equiv \terminal$, this condition above is identical to the one we saw initially for generative flow networks in \cref{eq:boundary-conditions}. If $P_{F}$ admits an invariant measure $F$ that satisfies the boundary condition above, then we obtain the counterpart of the fundamental theorem of GFlowNet in \cref{thm:flow-matching-proportional-reward}, but this time working exclusively with Markov chains.

\begin{theorem}
    \label{thm:markov-chain-discrete-terminating-state-propto-reward}
    Let $(X_{n})_{n\geq 0}$ be an irreducible and positive recurrent Markov chain over $\gS$, with initial state $s_{0}$ and transition probability $P_{F}$, that admits an invariant measure $F$ such that $\forall x \in \gX$, $F(x)P_{F}(s_{0}\mid x) = R(x)$, where $R$ is a finite measure on $\gX \subseteq \gS$. Then the terminating state probability distribution satisfies $\forall x\in\gX$, $P_{F}^{\top}(x) \propto R(x)$.
\end{theorem}

\begin{proof}
    Since $F$ is an invariant measure of $P_{F}$, by unicity of the invariant measure of $P_{F}$ up to a multiplicative constant (\cref{thm:discrete-markov-chain-existance-invariant-measure}), there exists a constant $\gamma > 0$ such that $F = \gamma \lambda$, where $\lambda$ is the invariant measure defined in \cref{eq:discrete-markov-chain-existance-invariant-measure}. Using \cref{prop:terminating-state-distribution-invariant-measure} and the boundary condition $F(x)P_{F}(s_{0}\mid x) = R(x)$, we get for all $x\in\gX$
    \begin{equation}
        P_{F}^{\top}(x) = \lambda(x)P_{F}(s_{0}\mid x) = \frac{1}{\gamma}F(x)P_{F}(s_{0}\mid x) = \frac{R(x)}{\gamma}.
    \end{equation}
    In fact, since $P_{F}^{\top}$ is a probability distribution (\cref{prop:general-discrete-terminating-state-proper-distribution}), the multiplicative constant happens to be $\gamma = \sum_{x\in\gX}R(x)$ (\ie the partition function).
\end{proof}

In practice though, showing that a measure $F$ is invariant proves to be at least as computationally expensive as our original problem of sampling from the Gibbs distribution. Indeed, verifying invariance involves in particular to check that \cref{eq:definition-invariant-measure-discrete} is satisfied for $s' = s_{0}$, where we would have to sum over all the elements of $\gX$; this is precisely what we wanted to avoid with the intractable partition function $Z$ of the Gibbs distribution in \cref{eq:gibbs-distribution}. In standard GFlowNets, it is common practice to only check this for any $s'\neq s_{0}$ (\cref{prop:flow-matching-condition-state-flow-pF}). We can show that these two perspectives are in fact equivalent in the discrete case, and checking \cref{eq:definition-invariant-measure-discrete} for all states except $s_{0}$ still implies invariance over the whole state space, including the initial state; see \cref{prop:markov-chain-flow-matching-invariant}.

\subsection{Comparison with Markov chain Monte Carlo methods}
\label{sec:comparison-gflownets-mcmc}\index{Markov chain Monte Carlo}
As the name implies, Markov chains play a central role in \emph{Markov chain Monte Carlo} methods (MCMC; see also \cref{sec:existing-approaches-sampling-ebm}). GFlowNets and MCMC methods have both been introduced to solve the same problem: sampling from a probability distribution that is defined up to a normalization constant, such as the Gibbs distribution. One of the main advantages of viewing GFlowNets from the perspective of Markov chains is that it places them under the same theoretical framework as MCMC, highlighting the similarities and differences between both methods. These differences are summarized in \cref{tab:comparison-gflownet-mcmc}.

\begin{table}[t]
    \centering
    \caption[Comparison between MCMC and GFlowNets, from the perspective of Markov chains.]{Comparison between MCMC methods and GFlowNets, from the perspective of Markov chains, defined on a discrete state space $\gS$ with a positive recurrent Markov kernel $P_{F}$ \& invariant measure/distribution $F$. In both cases, the objective is to approximate a target distribution $\propto R(x)$. The row ``GFlowNet'' corresponds to the Markov chain perspective introduced in \cref{sec:gflownet-as-markov-chain}.}
    \label{tab:comparison-gflownet-mcmc}
    \begin{tabular}{lcccc}
        \toprule
         & Sample & Conditions & Target distribution & \multirow{2}{*}{Sampling}\\
         & space & \footnotesize{(+ positive recurrence)} & $\propto R(x)$ & \\
        \midrule
        \multirow{2}{*}{MCMC} & \multirow{2}{*}{$\gS$} & Aperiodicity & $ = F(x)$ & Asymptotic \\
        & & \footnotesize{(convergence)} & \footnotesize{(invariant dist.)} & \footnotesize{(correlated)} \\[0.8em]
        \multirow{2}{*}{GFlowNet} & \multirow{2}{*}{$\gX \subseteq \gS$} & Boundary condition & $ = P_{F}^{\top}(x)$ & Finite time \\
        & & \footnotesize{(marginal)} & \footnotesize{(terminating state dist.)} & \footnotesize{(independent)} \\
        \bottomrule
    \end{tabular}
\end{table}

Recall that the goal of MCMC methods is to construct a Markov chain whose invariant distribution matches the target distribution defined up to a normalization constant. Samples from the distribution are then obtained by running the Markov chain until convergence to the invariant distribution, which typically requires to run the Markov chain for a long (burn-in) period. The convergence of iterates $\{P^{n}(\cdot\mid s_{0})\}_{n\geq 0}$ to the invariant distribution is guaranteed by the \emph{ergodicity} of the chain (\ie positive recurrent, \emph{and} aperiodic). In contrast, the Markov chain related to a GFlowNet as described in previous section is only required to be positive recurrent in order to guarantee the existence of an invariant measure $F$, but \emph{no guarantee on the convergence to $F$ is necessary}. Moreover, while the Markov kernel $P_{F}$ of MCMC methods needs to be carefully built to ensure that the invariant distribution matches the target distribution, the invariant measure of the Markov kernel in a GFlowNet may be arbitrary, as long as it matches the boundary condition \cref{eq:generalized-boundary-condition-discrete}; there may be multiple Markov kernels with different invariant distributions yielding the same terminating state distribution, as mentioned in \cref{sec:flow-matching-condition-boundary-constraint}. This could explain why the Markov kernels in MCMC methods are typically handcrafted (\eg Metropolis-Hastings in \cref{alg:metropolis-hastings-algorithm}), as opposed to GFlowNets where $P_{F}$ is learned (\eg with neural networks). For completeness, we note that some components of the kernel in MCMC may also be learned, but this would often be limited to the proposal distribution of Metropolis-Hastings for example \citep{wang2018metalearningmcmc}.

Probably the main difference between GFlowNets and MCMC is the relation between the invariant measure/distribution of the Markov chain and the target distribution. While MCMC requires the invariant distribution to match the target distribution, GFlowNets only require the marginal distribution of the chain (\ie the terminating state probability distribution; \cref{def:general-discrete-terminating-state-distribution}) to match the target distribution. The state space of the chain in MCMC therefore corresponds to the sample space of the target distribution; we saw in \cref{sec:sampling-ebm} that the chain moves exclusively between valid elements of the sample space. As a consequence, these chains are known to poorly mix, leading to slow convergence to the invariant distribution, especially in the presence of multiple modes \citep{robert2018acceleratingmcmc,roy2020convergencemcmc}. On the other hand, the Markov chain associated with a GFlowNet is constructed on an \emph{augmented} state space $\gS$, broader than the sample space $\gX \subseteq \gS$, allowing the chain to use these intermediate steps to move between modes more easily. It is worth noting that there exist some approaches such as Hamiltonian Monte Carlo methods (HMC; \citealp{neal1993inferencemcmc,mackay2003informationtheory}), where the target distribution is the marginal of the invariant distribution over an augmented space.

Finally, since MCMC methods rely on the convergence to the invariant distribution, samples of the target distribution are only guaranteed asymptotically. Moreover, consecutive samples are correlated by the Markov kernel $P_{F}$, and additional post-processing techniques are required to reduce the effect of the cross-correlation between samples. Conversely, samples of a GFlowNet are obtained in finite time thanks to the positive recurrence of the Markov chain, and are guaranteed to be independent from one another because of the strong Markov property.

\section{Generative flow networks over general state spaces}
\label{sec:gflownet-general-state-spaces}
One of the major obstacles to define a GFlowNet over continuous states spaces is that the notion of pointed DAG, which was essential throughout \cref{chap:flow-networks,chap:generative-flow-networks}, would not make sense anymore in a continuous state space. We will see in this section that approaching this from the angle of Markov chains, as we did in the previous section, will allow us to more naturally transition to more general spaces without relying so heavily on DAGs anymore. Before going into the details of this extension though, we need to first establish what we mean by ``general state space''. This is done via the concept of \emph{measurable space} from measure theory.

\begin{definition}[Measurable space]
    \label{def:measurable-space}\index{Measurable space}
    Let $\gS$ be a state space. The pair $(\gS, \Sigma)$ is called a \emph{measurable space} if $\Sigma$ is a $\sigma$-algebra associated to $\gS$, \ie $\Sigma$ is a set of sets of $\gS$ satisfying
    \begin{enumerate}
        \item $\gS \in \Sigma$;
        \item $\Sigma$ is closed under complementation: $\forall B \in \Sigma$, $\gS \backslash B \in \Sigma$;
        \item $\Sigma$ is closed under countable unions: if $\forall n, B_{n} \in \Sigma$, then $\cup_{n\geq 0}B_{n} \in \Sigma$.
    \end{enumerate}
\end{definition}

An example of measurable space is $\sR$ along with the \emph{Borel $\sigma$-algebra}, which is the $\sigma$-algebra generated by the open sets of $\sR$. Measurable spaces are critical in order to define notions of \emph{measures} over general state spaces, as well as the notion of \emph{measurable functions}, where the preimage of any measurable set (element of the $\sigma$-algebra) remains a measurable set in the domain of that function. In this thesis, we will always assume that the measures $\mu$ we consider are positive and $\sigma$-finite, meaning that $\gS$ is the countable union of measurable sets $(B_{n})_{n\geq 0}$ with $\mu(B_{n}) < \infty$.

In the discrete case, having a pointed DAG was convenient to structure how we could move through the state space. But what is more important is how $P_{F}$ was defined, which in itself captures this structure locally: if we are in a certain state, it encodes how can we move to the next state and with what probability. This can be naturally extended to measurable state spaces through the notion of \emph{Markov kernel}, a name we already used in \cref{sec:cyclic-nature-flow-networks}. A Markov kernel gives us a probability distribution that we can transition from while being in a state $s\in\gS$.

\begin{definition}[Markov kernel]
    \label{def:markov-kernel}\index{Kernel!Transition kernel}\index{Kernel!Markov kernel}\index{Markov kernel|see {Kernel}}\index{Transition kernel|see {Kernel}}
    Let $(\gS, \Sigma)$ be a measurable state space. A function $\kappa: \gS \times \Sigma \rightarrow [0, \infty)$ is called a positive $\sigma$-finite \emph{transition kernel} if
    \begin{enumerate}
        \item For any $B\in\Sigma$, the mapping $s\mapsto \kappa(s, B)$ is measurable, where the space $[0, \infty)$ is associated with the Borel $\sigma$-algebra $\gB([0, \infty))$;
        \item For any $s\in\gS$, the mapping $B \mapsto \kappa(s, B)$ is a positive $\sigma$-finite measure on $(\gS, \Sigma)$.
    \end{enumerate}
    Furthermore, if the mappings $\kappa(s, \cdot)$ are probability distributions (\ie $\kappa(s, \gS) = 1$), the transition kernel is called a \emph{Markov kernel}.
\end{definition}
Similar to the discrete case, we will denote by $(X_{n})_{n\geq 0}$ the canonical Markov chain \citep{petritis2015markovchainmeasurable} following the Markov kernel $P_{F}(s, B)$. We use the notation ``$P_{F}(s, B)$'' for a Markov kernel (instead of $\kappa(s, B)$) by analogy with the forward transition probabilities $P_{F}(s'\mid s)$ in the discrete case; this new notation with ``$s$'' (the state we are transitioning from) as the first argument highlights that we are no longer working exclusively in discrete space. We will use once again the notation $\sP_{s}$ to denote the joint distribution of the Markov chain starting at some state $s$. We leave the question of \emph{where} this Markov chain starts (the ``initial state'') for the next section.

Although we defined transition kernels over a single space $(\gS, \Sigma)$ (\ie we are transitioning into the same space), it is easy to extend this definition to transition kernels from one space to a different one \citep{petritis2015markovchainmeasurable}; see \cref{app:operations-transition-kernels} for details. This allows us to also define in \cref{app:operations-transition-kernels} different operations over transition kernels, such as the \emph{product kernel} $\kappa_{1} \otimes \kappa_{2}$, or the \emph{composition kernel} $\kappa_{1}\cdot\kappa_{2}$, which will be useful in what follows.

Finally, a property on which we have not focused our attention too much in discrete spaces is the notion of \emph{irreducibility} (\ie all the states are accessible from any other state); this was also a necessary property when we talked about pointed DAGs, as we discussed in \cref{sec:general-revisiting-flow-matching}. On a measurable state space, irreducibility encodes a notion of accessibility that can be formalized for sets that are not ``degenerate'', in the sense of a measure $\phi$. This is called \emph{$\phi$-irreducibility}.

\begin{definition}[$\phi$-irreducibility]
    \label{def:phi-irreducibility}\index{Irreducibility!phi-irreducibility@$\phi$-irreducibility}
    Let $\phi$ be a measure over a measurable space $(\gS, \Sigma)$. A Markov chain $(X_{n})_{n\geq 0}$ is said to be \emph{$\phi$-irreducible} if any set $B \in \Sigma$ such that $\phi(B) > 0$ is accessible, in the sense that for all $s\in\gS$
    \begin{equation}
        \sP_{s}(\eta_{B} < \infty) > 0,
    \end{equation}
    where $\eta_{B} = \inf \{n \geq 0\mid X_{n}\in B\}$\glsadd{hittingtime} is the hitting time of the Markov chain to $B$.
\end{definition}

In what follows, instead of being the Gibbs distribution \cref{eq:gibbs-distribution}, or some distribution defined by a reward function up to a normalization constant \cref{eq:target-distribution-gflownet-reward}, our objective will be to sample from a distribution $P^{\star}$ defined from a positive and finite measure $R$ over the measurable space $(\gX, \Sigma_{\gX})$ by
\begin{equation}
    P^{\star}(B) \propto R(B),
    \label{eq:general-target-distribution-gflownet-reward}
\end{equation}
for all $B \in \Sigma_{\gX}$, where $\Sigma_{\gX}$ is the trace $\sigma$-algebra of the subsets of $\gX \subseteq \gS$. This distribution is well-defined since we assumed that $R$ was finite. This section will follow loosely the structure of \cref{sec:gflownet-as-markov-chain}: (1) we will first ensure that there exists a notion of terminating state distribution, by taking inspiration from the ``wrapped-around'' construction of \cref{sec:cyclic-nature-flow-networks}; (2) then we will introduce an extension of positive recurrence for Markov chains on measurable state spaces to guarantee the existence of an invariant measure; and (3) finally, we will combine invariance with a boundary condition to find a Markov kernel whose corresponding terminating state distribution matches the target distribution $P^{\star}$.

\subsection{Creation of an atom via the splitting technique}
\label{sec:creation-atom-splitting-technique}
The key property preserved by wrapping around the state space at $s_{0} \equiv \terminal$ in \cref{sec:cyclic-nature-flow-networks} was that the Markov chain was effectively ``regenerating'' every time it was returning to the initial state $s_{0}$, thanks to the (strong) Markov property. In this context, $\{s_{0}\}$ is called an \emph{atom} of the Markov chain \citep{meyn1993markovchainsstability}, which informally corresponds to the chain ``forgetting'' about the past every time it goes through any state in the atom. We can generalize this notion to Markov chains over general state spaces.

\begin{definition}[Atom]
    \label{def:atom}\index{Atom}
    Let $(X_{n})_{n\geq 0}$ be a Markov chain over a measurable space $(\gS, \Sigma)$, with Markov kernel $P_{F}$. A set $A\in\Sigma$ is called an \emph{atom} if there exists a probability measure $\nu$ over $\gS$ such that $\forall s\in A$, and $\forall B\in\Sigma$,
    \begin{equation}
        P_{F}(s, B) = \nu(B).
        \label{eq:definition-atom}
    \end{equation}
\end{definition}

The notion of atom is evident for singletons in discrete spaces, but in general Markov chains may not contain any accessible atom. Although it was not interpreted this way, we can view the augmentation of the state space in \citet{lahlou2023continuousgfn} with a terminal state $\terminal \notin \gS$ as a way to create an accessible artificial atom at $\terminal$, if we were to also (informally) wrap around the state space at $s_{0} \equiv \terminal$ as in \cref{sec:cyclic-nature-flow-networks}. We will come back to this construction of an explicit atom in \cref{sec:practical-implementation-generalized-gflownets}.

Besides introducing an artificial atom though, which requires changes to the definition of the state space, we will show in this section how the \emph{split chain} construction \citep{nummelin1978splitting} can be used as an alternative way to create a pseudo-atom for a large class of Markov chains, without ever having to change the state space. Instead of introducing a new state, we will use a set $\gX\in\Sigma$ that satisfies a \emph{minorization condition}. The choice of the notation $\gX$ is not coincidental because this set will eventually be the sample space of the corresponding terminating state distribution that we will define in the next section (as well as the sample space of \cref{eq:general-target-distribution-gflownet-reward}).

\begin{definition}[Minorization condition; \citealp{nummelin1978splitting}]
    \label{def:minorization-condition}\index{Minorization condition}
    Let $(X_{n})_{n\geq 0}$ be a $\phi$-irreducible Markov chain over a measurable space $(\gS, \Sigma)$, with Markov kernel $P_{F}$. A set $\gX\in\Sigma$ such that $\phi(\gX) > 0$ is said to satisfy the \emph{minorization condition} if there exists a non-negative measurable function $\varepsilon$ such that $\varepsilon^{-1}((0, +\infty)) = \gX$ (\ie $\gX$ is the set on which $\varepsilon$ is positive), and a probability measure $\nu$ such that, $\forall s\in\gS$ and $\forall B\in\Sigma$,
    \begin{equation}
        P_{F}(s, B) \geq \varepsilon(s)\nu(B).
        \label{eq:definition-minorization-condition}
    \end{equation}
\end{definition}

Taking $B = \gS$ in the inequality above, we can see that $\varepsilon$ is necessarily bounded, with $\varepsilon(s) \in [0, 1]$; as such, we can interpret $\varepsilon(s)$ as a probability. This minorization condition is not particularly interesting when $s\notin\gX$, as it simply implies that $P_{F}(s, B) \geq 0$. When $x\in\gX$ though, this allows us to interpret a fixed Markov kernel $P_{F}$ satisfying the minorization condition as a mixture of two Markov kernels, whose mixture weights depend on $\varepsilon(x)$:
\begin{equation}
    P_{F}(x, B) = \big(1 - \varepsilon(x)\big)R_{\nu}(x, B) + \varepsilon(x)\nu(B),
    \label{eq:minorization-condition-mixture-markov-kernels}
\end{equation}
and where $R_{\nu}(x, B)$ is a ``remainder'' Markov kernel that can be defined as
\begin{equation}
    R_{\nu}(x, B) = \mathds{1}(\varepsilon(x) < 1) \frac{P_{F}(x, B) - \varepsilon(x)\nu(B)}{1 - \varepsilon(x)} + \mathds{1}(\varepsilon(x) = 1)\nu(B).
    \label{eq:remainder-markov-kernel}
\end{equation}
The exact form of the remainder kernel is not as important as the fact that it is a properly defined Markov kernel whose existence is guaranteed when $P_{F}$ satisfies the minorization condition. The important property to note is that the second kernel in the mixture \cref{eq:minorization-condition-mixture-markov-kernels}, $\nu(B)$, is completely independent of $x$---only the mixture weight $\varepsilon(x)$ depends on it. We can interpret \cref{eq:minorization-condition-mixture-markov-kernels} as follows: (1) we first select which of the two kernels to apply, with probability $\varepsilon(x)$, and upon the selection of the second kernel (2) we ``reset'' the Markov chain with $\nu(B)$. Using the terminology of GFlowNets used in previous chapters, and connecting with the Markov kernel $\widebar{P}_{F}$ of \citet{lahlou2023continuousgfn} detailed in \cref{sec:practical-implementation-generalized-gflownets} (we use a different notation to distinguish it from the Markov kernel $P_{F}$ considered here, to avoid confusion), we can view $\varepsilon(x) \approx \widebar{P}_{F}(x, \{\terminal\})$ the probability of terminating at $x\in\gX$.  %

\begin{figure}[t]
    \centering
    \begin{adjustbox}{center}
        \includegraphics[width=480pt]{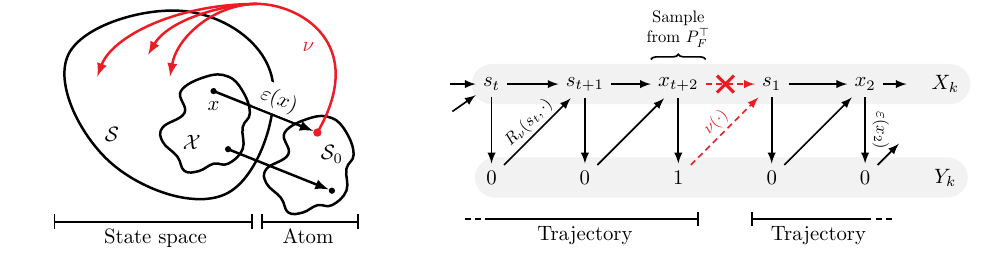}%
    \end{adjustbox}
    \begin{adjustbox}{center}%
        \begin{subfigure}[b]{200pt}%
            \caption{}%
            \label{fig:splitting-technique}%
        \end{subfigure}%
        \begin{subfigure}[b]{280pt}%
            \caption{}%
            \label{fig:splitting-technique-trajectories}%
        \end{subfigure}%
    \end{adjustbox}
    \caption[Construction of an atom via the splitting technique]{Construction of an atom via the splitting technique. (a) Illustration of the atom $\gS_{0} = \gX \times \{1\}$ created from $\gX$. Once the Markov chain reaches the atom $\gS_{0}$, it transitions to its next state with the kernel $\nu(\cot)$, independent of $x$. (b) The split chain $Z_{n} = (X_{n}, Y_{n})$, where $X_{n}$ are elements of $\gS$, and $Y_{n}$ is a binary value indicating which kernel to choose next: if $Y_{n} = 0$, the remainder kernel $R_{\nu}(X_{n}, \cdot)$ is picked, otherwise it is $\nu(\cdot)$ representing the ``end'' of a trajectory with a corresponding sample from the terminating state distribution $P_{F}^{\top}$ (\cref{def:generalized-terminating-state-distribution}).}
    \label{fig:splitting-technique-and-trajectories}
\end{figure}

This suggests the construction of a \emph{split chain} $(Z_{n})_{n\geq 0}$, where each element can be broken down into $Z_{n} = (X_{n}, Y_{n})$, with $X_{n}$ being a state in $\gS$, and $Y_{n}$ being a binary variable indicating which of the two kernels in the mixture \cref{eq:minorization-condition-mixture-markov-kernels} to select at the next step. This is a Markov chain over the product space $(\gS', \Sigma') = \big(\gS \times \{0, 1\}, \Sigma \otimes \sigma(\{0, 1\})\big)$, with Markov kernel $P_{F}^{\mathrm{split}} = Q_{\nu} \otimes b_{\varepsilon}$, where
\begin{align}
    Q_{\nu}((x, y), B) &= \mathds{1}(y=0)R_{\nu}(x, B) + \mathds{1}(y=1)\nu(B) && B\in\Sigma\label{eq:product-kernel-component-1}\\
    b_{\varepsilon}(x, C) &= \big(1 - \varepsilon(x)\big)\delta_{0}(C) + \varepsilon(x)\delta_{1}(C) && C \in \sigma(\{0, 1\}),\label{eq:product-kernel-component-2}
\end{align}
where $\delta_{y}$ is the Dirac measure at $y$, and $\sigma(\{0, 1\})$ is the discrete $\sigma$-algebra on $\{0, 1\}$. The notions of product $\sigma$-algebra $\Sigma \otimes \sigma(\{0, 1\})$ and product kernel $Q_{\nu}\otimes b_{\varepsilon}$ are defined in \cref{app:operations-transition-kernels}. This construction, illustrated in \cref{fig:splitting-technique-and-trajectories}, is similar to splitting the terminating states $x\in\gX$ in discrete GFlowNets into transitions $x \rightarrow x^{\top}$, were $x^{\top}$ has no children except the terminal state $\terminal$ \citep{malkin2022trajectorybalance}. It is easy to show that the set $\gX \times \{1\} \in \Sigma'$ is an atom of the split chain \citep{nummelin1978splitting}. We will call this atom $\gS_{0} \triangleq \gX \times \{1\}$, by analogy with the initial state in discrete flow networks, which plays a similar role (as an atom, although there exist subtle differences between $\gS_{0}$ and $\{s_{0}\}$; see the proof of \cref{thm:generalized-flow-matching-terminating-state} for details).  %

\subsection{Harris recurrence and invariant measures}
\label{sec:harris-recurrence-invariant-measures}
Similar to the discrete case, we will also assume
some form of recurrence for the Markov kernel in addition to ($\phi$-)irreducibility, in order to guarantee the existence of an invariant measure for $P_{F}$. Although other notions of recurrence exist for measurable state spaces, we will use a stronger notion called \emph{Harris recurrence}, which can be seen as a counterpart of positive recurrence in discrete space.

\begin{definition}[Harris recurrence]
    \label{def:harris-recurrence}\index{Recurrence!Harris recurrence}
    A $\phi$-irreducible Markov chain $(X_{n})_{n\geq 0}$ over a measurable state space $(\gS, \Sigma)$ is said to be \emph{Harris recurrent} if for all set $B\in\Sigma$ such that $\phi(B) > 0$, any sequence starting in $B$ eventually returns back to $B$ in finite time with probability 1:
    \begin{equation}
        \forall s\in B,\qquad \sP_{s}(\sigma_{B} < \infty) = 1,
    \end{equation}
    where $\sigma_{B} = \inf\{n \geq 1\mid X_{n} \in B\}$ is the return time of the Markov chain to $B$ (similar to \cref{eq:return-time}).
\end{definition}

The condition that the Markov chain returns to any accessible set with probability $1$ is reminiscent
of the ``finitely absorbing'' condition of \citet{lahlou2023continuousgfn} that requires all trajectories to be of bounded length as we will see in \cref{sec:separating-structure-transition-probabilities}. A measure $\lambda$ over $(\gS, \Sigma)$ is said to be \emph{invariant} for $P_{F}$ if for any bounded measurable function $f: \gS \rightarrow \sR$ we have
\begin{equation}
    \int_{\gS}f(s')\lambda(ds') = \iint_{\gS\times\gS}f(s')\lambda(ds)P_{F}(s, ds').
    \label{eq:invariant-measure-generalized}
\end{equation}
This equation transposes the notion of invariance in the same way \cref{eq:definition-invariant-measure-discrete} did in the discrete case as an equality between two measures; take for example $f(s) = \mathds{1}_{B}(s)$. We will often use the response of an arbitrary bounded measurable function $f$ to prove equality between measures, especially in \cref{sec:practical-implementation-generalized-gflownets}. The notation $\int f(s)\lambda(ds)$ must be understood as the integral of $f$ wrt.~the measure $\lambda$, and can often be interpreted as $\int f(s)g(s)ds$ with some appropriate function $g$ (called the \emph{density} of the measure $\lambda$, or its \emph{Radon-Nikodym derivative}); we will come back to densities in \cref{sec:generalized-flow-matching-losses}. The following theorem shows the existence of such an invariant measure for a Harris recurrent Markov chain, and is the counterpart of \cref{thm:discrete-markov-chain-existance-invariant-measure} for general state spaces.

\begin{theorem}[Existence of an invariant measure]
    \label{thm:harris-recurrent-mc-invariant-measure}
    Let $(X_{n})_{n \geq 0}$ be a Harris recurrent Markov chain over a measurable state space $(\gS, \Sigma)$, with Markov kernel $P_{F}$, such that $\gX$ satisfies the minorization condition in \cref{def:minorization-condition}. Then there exists a non-trivial invariant measure $\lambda$ for $P_{F}$ \cref{eq:invariant-measure-generalized}, unique up to a positive multiplicative constant, defined for all bounded measurable functions $f: \gS \rightarrow \sR$ by
    \begin{equation}
        \int_{\gS}f(s)\lambda(ds) = \E_{\gS_{0}}\Bigg[\sum_{n=1}^{\sigma_{\gS_{0}}}f(X_{n})\Bigg] = \sum_{n=1}^{\infty}\E_{\gS_{0}}\big[\mathds{1}(n \leq \sigma_{\gS_{0}})f(X_{n})\big].
        \label{eq:harris-recurrent-mc-invariant-measure}
    \end{equation}
    Moreover, $\lambda$ is the unique invariant measure such that $\int_{\gS}\varepsilon(s)\lambda(ds) = 1$, where $\varepsilon$ is the measurable function in the minorization condition.
\end{theorem}

\begin{proof}
    The existence and the form of the invariant measure $\lambda$ are given in \citep[][Theorem 3]{nummelin1978splitting}. It is therefore sufficient to prove that $\int_{\gS}\varepsilon(s)\lambda(ds) = 1$. We first show that for any $n \geq 1$, we have
    \begin{align}
        \E_{\gS_{0}}\big[\mathds{1}(n \leq \sigma_{\gS_{0}})\varepsilon(X_{n})\big] &= \E_{\gS_{0}}\big[\mathds{1}(n \leq \sigma_{\gS_{0}})\E_{Z_{n-1}}[\mathds{1}(Y_{1} = 1)\mid X_{1}]\big]\label{eq:harris-recurrent-mc-invariant-measure-proof-1}\\
        &= \E_{\gS_{0}}\big[\mathds{1}(n \leq \sigma_{\gS_{0}})\E_{\gS_{0}}[\mathds{1}(Y_{n} = 1)\mid Z_{0:n-1}, X_{n}]\big]\label{eq:harris-recurrent-mc-invariant-measure-proof-2}\\
        &= \E_{\gS_{0}}\big[\mathds{1}(n \leq \sigma_{\gS_{0}})\mathds{1}(Y_{n} = 1)\big]\label{eq:harris-recurrent-mc-invariant-measure-proof-3}\\
        &= \E_{\gS_{0}}\big[\mathds{1}(\sigma_{\gS_{0}} = n)\big],\label{eq:harris-recurrent-mc-invariant-measure-proof-4}
    \end{align}
    where we used an equivalent definition of $\varepsilon(X_{n})$ in \cref{eq:harris-recurrent-mc-invariant-measure-proof-1} as being the probability of setting the indicator variable $Y$ of the split chain to $1$ at the next step ($\mathds{1}(Y_{1} = 1)$) based on \cref{eq:product-kernel-component-2}, the Markov property of the chain $(Z_{n})_{n\geq 0}$ in \cref{eq:harris-recurrent-mc-invariant-measure-proof-2}, the law of total expectation in \cref{eq:harris-recurrent-mc-invariant-measure-proof-3}, and finally the definition of $\sigma_{\gS_{0}}$ in \cref{eq:harris-recurrent-mc-invariant-measure-proof-4} as being the return time to $\gS_{0}$, where we necessarily have $Y = 1$. Therefore, using the form of the invariant measure $\lambda$, and since $\varepsilon$ is a bounded measurable function (guaranteeing the existence of the integral), we have
    \begin{equation}
        \int_{\gS}\varepsilon(s)\lambda(ds) = \sum_{n=1}^{\infty}\E_{\gS_{0}}\big[\mathds{1}(n \leq \sigma_{\gS_{0}})\varepsilon(X_{n})\big] = \sum_{n=1}^{\infty}\E_{\gS_{0}}\big[\mathds{1}(\sigma_{\gS_{0}} = n)\big] = 1,
    \end{equation}
    where we were able to conclude since the split chain $(Z_{n})_{n\geq 0}$ is also Harris recurrent, meaning that it must return to $\gS_{0}$ with probability 1 (\cref{def:harris-recurrence}).
\end{proof}

Since $\gS_{0}$ is an atom of the split chain, the kernel $\nu$ in the mixture \cref{eq:minorization-condition-mixture-markov-kernels} is guaranteed to be selected to transition from any state of $\gS_{0}$; this justifies our notation $\E_{\gS_{0}}[\cdot]$ to denote $\E_{z}[\cdot]$ for any $z\in\gS_{0}$ (\ie the exact initial state $z$ has no impact on the expectation as long as that we start from a state of $\gS_{0}$).

\subsection{GFlowNets as recurrent Markov chains}
\label{sec:gflownet-recurrent-markov-chains}
Similar to \cref{sec:cyclic-nature-flow-networks}, we will define a GFlowNet in terms of a (Harris) recurrent Markov chain, this time over a general measurable state space $(\gS, \Sigma)$. Instead of wrapping around the GFlowNet to construct an atom $\{s_{0}\}$ though, we now use the atom $\gS_{0}$ constructed in \cref{sec:creation-atom-splitting-technique} via the splitting technique. Just like in \cref{def:general-discrete-terminating-state-distribution}, we can also define the \emph{terminating state distribution} over $\gS$ as the marginal distribution of the split chain returning to the atom $\gS_{0}$. This is guaranteed to exist for a Harris recurrent chain since it eventually returns to $\gS_{0}$ with probability $1$.

\begin{definition}[Generalized terminating state distribution]
    \label{def:generalized-terminating-state-distribution}\index{Terminating state probability!Generalized}
    Let $(X_{n})_{n\geq 0}$ be a Harris recurrent Markov chain over a measurable space $(\gS, \Sigma)$, with Markov kernel $P_{F}$, such that $\gX$ satisfies the minorization condition in \cref{def:minorization-condition}. The \emph{terminating state distribution} is defined as the marginal distribution of the split chain $(Z_{n})_{n\geq 0}$ defined in \cref{sec:creation-atom-splitting-technique} returning to the atom $\gS_{0} = \gX \times \{1\}$. For all $B \in \Sigma_{\gX}$:
    \begin{equation}
        P_{F}^{\top}(B) \triangleq \E_{\gS_{0}}\big[\mathds{1}_{B}(X_{\sigma_{\gS_{0}}})\big] = \E_{\gS_{0}}\big[\mathds{1}_{B\times \{1\}}(Z_{\sigma_{\gS_{0}}})\big],
        \label{eq:generalized-terminating-state-distribution}
    \end{equation}
    where $\Sigma_{\gX}$ is the trace $\sigma$-algebra induced by $\Sigma$ on $\gX$ (\ie $\sigma$-algebra of subsets of $\gX$).
\end{definition}

It is interesting to see that while the terminating state distribution in discrete spaces involved the state of the Markov chain at time $\sigma_{s_{0}} - 1$ (\cref{def:general-discrete-terminating-state-distribution}), the general case above does not have this offset by one. The reason is that in the split chain, we can treat $X_{n} \rightarrow Y_{n}$ as being a ``sub-transition'', which may be enough to make up for the extra step required in the discrete case. We can again show that the terminating state distribution is related to the invariant measure $\lambda$. The following proposition is the counterpart of \cref{prop:terminating-state-distribution-invariant-measure} for general state spaces.

\begin{proposition}
    \label{prop:generalized-terminating-state-distribution-invariant-measure}
    The terminating state distribution $P_{F}^{\top}$ is related to the invariant measure $\lambda$ defined in \cref{thm:harris-recurrent-mc-invariant-measure} by, $\forall B \in \Sigma_{\gX}$
    \begin{equation}
        P_{F}^{\top}(B) = \int_{B}\varepsilon(x)\lambda(dx),
    \end{equation}
    where $\varepsilon$ is the measurable function in \cref{def:minorization-condition}.
\end{proposition}

\begin{proof}
    The proof of this proposition follows the same pattern as the proof of \cref{prop:terminating-state-distribution-invariant-measure}. Following the same steps as \cref{eq:harris-recurrent-mc-invariant-measure-proof-1}--\cref{eq:harris-recurrent-mc-invariant-measure-proof-4} in the proof of \cref{thm:harris-recurrent-mc-invariant-measure}, we can show that for any $n \geq 1$, we have
    \begin{equation}
        \E_{\gS_{0}}\big[\mathds{1}(n \leq \sigma_{\gS_{0}})\mathds{1}_{B}(X_{n})\varepsilon(X_{n})\big] = \E_{\gS_{0}}\big[\mathds{1}(\sigma_{\gS_{0}} = n)\mathds{1}_{B}(X_{n})\big].\label{eq:generalized-terminating-state-distribution-invariant-measure-proof-4}
    \end{equation}
    Using the definition of the invariant measure $\lambda$ in \cref{thm:harris-recurrent-mc-invariant-measure} in addition to the equation above, and given that for any $B\in\Sigma_{\gX}$ the measurable function $\mathds{1}_{B}\varepsilon$ is bounded (guaranteeing the existence of the integral),
    {\allowdisplaybreaks%
    \begin{align}
        \int_{B}\varepsilon(x)\lambda(dx) &= \sum_{k=1}^{\infty}\E_{\gS_{0}}\big[\mathds{1}(n \leq \sigma_{\gS_{0}})\mathds{1}_{B}(X_{n})\varepsilon(X_{n})\big]\\
        &= \sum_{n=1}^{\infty}\E_{\gS_{0}}\big[\mathds{1}(\sigma_{\gS_{0}} = n)\mathds{1}_{B}(X_{n})\big]\label{eq:generalized-terminating-state-distribution-invariant-measure-proof-5}\\
        &= \E_{\gS_{0}}\big[\mathds{1}_{B}(X_{\sigma_{\gS_{0}}})\big] = P_{F}^{\top}(B).
    \end{align}}%
\end{proof}
The proposition above shows in particular that $P_{F}^{\top}$ is a properly defined distribution over $(\gX, \Sigma_{\gX})$, in the sense that $P_{F}^{\top}(\gX) = 1$, since $\lambda$ is the unique invariant measure of $P_{F}$ such that $\int_{\gS}\varepsilon(x)\lambda(dx) = 1$, and $\varepsilon$ is positive only on $\gX$.

Under the minorization condition of \cref{def:minorization-condition}, we saw that the Markov kernel $P_{F}$ can be interpreted as being a mixture of two separate kernels, whose mixture components depend on the function $\varepsilon$. This suggests a natural strategy to sample (continuously) from $P_{F}^{\top}$: (1) sample which mixture component to apply next based on the current state, and (2) sample from the corresponding Markov kernel, returning the state anytime the Markov chain is reset with $\nu$. This is illustrated in \cref{fig:splitting-technique-trajectories}, and detailed in \cref{alg:sampling-generalized-terminating-state-probability}.
\begin{algorithm}[t]
    \caption{Sampling from the terminating state distribution $P_{F}^{\top}$ (measurable space).}
    \label{alg:sampling-generalized-terminating-state-probability}
    \begin{algorithmic}[1]
        \Require A Markov kernel $P_{F}$ satisfying the conditions of \cref{def:generalized-terminating-state-distribution}, the measure $\nu$ and the function $\varepsilon$ in the minorization condition \cref{def:minorization-condition}.
        \Ensure Samples $x \sim P_{F}^{\top}(\cdot)$ of the terminating state distribution. %
        \State Initialization: $s_{0} \sim \nu(\cdot)$, $y_{0} \sim \mathrm{Bernoulli}(\varepsilon(s_{0}))$ %
        \Loop
            \If{$y_{t} = 0$}
                \State Sample the next state: $s_{t+1} \sim R_{\nu}(s_{t}, \cdot)$ \Comment{\cref{eq:remainder-markov-kernel}}
            \Else
                \State \textbf{yield} $x \equiv s_{t}$
                \State Sample the next state: $s_{t+1} \sim \nu(\cdot)$    
            \EndIf
            \State Sample the next mixture component: $y_{t+1} \sim \mathrm{Bernoulli}(\varepsilon(s_{t+1}))$
        \EndLoop
    \end{algorithmic}
\end{algorithm}

\paragraph{Boundary condition} In addition to the existence of some invariant measure $F$ thanks to Harris recurrence, and similar to discrete spaces in \cref{sec:gflownet-as-markov-chain}, we will require $F$ to also satisfy some boundary condition in order to obtain a terminating state distribution $P_{F}^{\top}$ that matches $P^{\star}$ in \cref{eq:general-target-distribution-gflownet-reward}. For some positive and finite measure $R$ over $(\gX, \Sigma_{\gX})$, where $\gX$ is the set that we defined in the minorization condition in \cref{def:minorization-condition}, we say that $F$ satisfies the boundary condition if for any bounded measurable function $f: \gX \rightarrow \sR$, we have
\begin{equation}
    \int_{\gX}f(x)R(dx) = \int_{\gX}f(x)\varepsilon(x)F(dx),
    \label{eq:generalized-boundary-condition}
\end{equation}
where $\varepsilon$ is the function defined in \cref{def:minorization-condition}; we saw that this function could be loosely interpreted as $\varepsilon(x) \approx \widebar{P}_{F}(x, \{\terminal\})$, making this boundary condition analogous to \cref{eq:generalized-boundary-condition-discrete} in the discrete case. We are finally ready to state the fundamental theorem of generative flow networks, extended to measurable state spaces.

\begin{theorem}
    \label{thm:markov-chain-generalized-terminating-state-propto-reward}
    Let $(X_{n})_{n \geq 0}$ be a Harris recurrent Markov chain over a measure state space $(\gS, \Sigma)$, with Markov kernel $P_{F}$, such that $\gX\in\Sigma$ satisfies the minorization condition in \cref{def:minorization-condition}. Moreover, assume that $P_{F}$ admits an invariant measure $F$ such that for any bounded measurable function $f: \gX \rightarrow \sR$,
    \begin{equation}
        \int_{\gX}f(x)R(dx) = \int_{\gX}f(x)\varepsilon(x)F(dx),
        \label{eq:markov-chain-generalized-terminating-state-propto-reward-boundary-condition}
    \end{equation}
    where $R$ is a positive and finite measure on $\gX$, and $\varepsilon$ is the measurable function in \cref{def:minorization-condition}. Then the terminating state distribution is proportional to the measure $R$: $\forall B \in \Sigma_{\gX}$, $P_{F}^{\top}(B) \propto R(B)$.
\end{theorem}

\begin{proof}
    The proof relies once again on the unicity of the invariant measure of a Harris recurrent Markov chain, up to a normalization constant. By \cref{thm:harris-recurrent-mc-invariant-measure}, there exists a constant $\gamma > 0$ such that $F = \gamma\lambda$. Using \cref{prop:generalized-terminating-state-distribution-invariant-measure}, we get for all $B\in\Sigma_{\gX}$
    \begin{equation}
        P_{F}^{\top}(B) = \int_{B}\varepsilon(x)\lambda(dx) = \frac{1}{\gamma}\int_{B}\varepsilon(x)F(dx) = \frac{R(B)}{\gamma} \propto R(B),
    \end{equation}
    where we used the boundary condition in \cref{eq:markov-chain-generalized-terminating-state-propto-reward-boundary-condition} with $f(x) = \mathds{1}_{B}(x)$.
\end{proof}

\section{Practical implementation of generalized GFlowNets}
\label{sec:practical-implementation-generalized-gflownets}
Although Markov chains provide an elegant framework to describe generative flow networks on general state spaces, in practice we need to find a way to \emph{learn} the Markov kernel $P_{F}$ and its corresponding invariant measure $F$ satisfying the boundary condition. Following how discrete GFlowNets are trained, we could transform these conditions (invariance \& boundary condition) into loss functions that can eventually be minimized. In this section, we will take even further inspiration from discrete GFlowNets and provide a construction of generalized GFlowNets as a special case of the framework described in \cref{sec:gflownet-general-state-spaces}, which offers some advantages for a practical implementation. Throughout this section, we will assume that $(\gS, \Sigma)$ is a measurable state space, and we identify a fixed initial state $s_{0} \in \gS$; moreover, we will also assume that $\{s_{0}\} \in \Sigma$.  %

\paragraph{Prior attempts to define continuous GFlowNets} Even if this section will culminate to the observation that the flow matching losses we defined in the discrete case in \cref{sec:flow-matching-losses} can be transferred almost directly with minimal modifications, the path to get there is not necessarily evident. Notably \citet{li2023cflownets} introduced \emph{CFlowNets}, which was meant as an initial attempt to extend discrete GFlowNets to continuous spaces. They did so by simply writing \emph{a} flow matching condition (distinct from \cref{thm:flow-matching-condition}) with integrals instead of sums. This was too closely inspired by the original formulation of \citet{bengio2021gflownet}, who worked with \emph{actions} instead of transitions (which is valid in discrete space), leading to the inaccurate ``continuous flow matching condition''
\begin{equation}
    \int_{s': s\rightarrow s'}F(s\rightarrow s')ds' = \int_{a}F(s\rightarrow T(s, a))da,
\end{equation}
where the second integral is taken over actions and $T(s, a)$ represents the state by taking the action $a$ in state $s$ (all quantities being loosely defined). This change of variable from transitions to actions is invalid in general: the integrand on the RHS is missing a Jacobian term $dT(s, a)/da$ which is not necessarily equal to $1$.

\subsection{Separating structure from transition probabilities}
\label{sec:separating-structure-transition-probabilities}
Apart from the nature of the state space, the key difference between discrete and generalized GFlowNets as described in the previous section is that, in the discrete case, the structure of the state space $\gG$ (as a pointed DAG) was distinct from the forward transition probabilities $P_{F}$ placed onto it. Conversely, generalized GFlowNets directly work at the level of the kernel $P_{F}$ without calling for an explicit underlying structure. This was motivated by the fact that it is impossible to define a notion of ``graph'' over a continuous state space for example, let alone ensuring that it is acyclic. Here, we will instead use another construction called \emph{measurable pointed graph}, whose structure will be defined in terms of transition kernels (\cref{def:markov-kernel}).

\begin{definition}[Measurable pointed graph]
    \label{def:measurable-pointed-graph}\index{Measurable pointed graph}
    Let $(\widebar{\gS}, \widebar{\Sigma})$ be a measureable space, where $\widebar{\gS} = \gS \cup \{\terminal\}$ is a state space augmented with a distinct state $\terminal \neq \gS$ (with $\widebar{\Sigma} = \sigma(\Sigma \cup \{\terminal\})$ being the $\sigma$-algebra induced by augmenting $\Sigma$ with $\terminal$), and let $s_{0}\in\gS$ be a fixed initial state. Let $\kappa_{F}$ \& $\kappa_{B}$ be two transition kernels and $\nu$ be a positive $\sigma$-finite measure on $\widebar{\gS}$. The tuple $\gG = ((\widebar{\gS}, \widebar{\Sigma}), \kappa_{F}, \kappa_{B}, \nu)$\glsadd{measurablepointedgraph} is called a \emph{measurable pointed graph} if it satisfies the following properties:
    \begin{enumerate}
        \item $\forall B \in \widebar{\Sigma}$, s.t. $\nu(B) > 0$, $\exists n \geq 0,\ \kappa_{F}^{n}(s_{0}, B) > 0$
        \item $\forall B \in \widebar{\Sigma}$, $\kappa_{F}(\terminal, B) = \mathds{1}(\terminal \in B)$
        \item $\forall B \in \widebar{\Sigma} \otimes \widebar{\Sigma}$, s.t. $(s_{0}, s_{0}) \notin B$ and $(\terminal, \terminal) \notin B$, $\nu \otimes \kappa_{F}(B) = \nu \otimes \kappa_{B}(B)$
        \item $\forall B \in \widebar{\Sigma},\ \kappa_{B}(s_{0}, B) = 0$
    \end{enumerate}
    The kernels \gls{forwardreferencekernel} and \gls{backwardreferencekernel}, and the measure $\nu$, are called the forward reference kernel, the backward reference kernel, and the reference measure of $\gG$ respectively.
\end{definition}

The reference kernels provide a sense of ``structure'' for the state space, where $\kappa_{F}$ encodes how to move forward in the state space, and $\kappa_{B}$ how to move backward. $\nu\otimes \kappa$ denotes the measure defined as the product of a measure $\nu$ with a transition kernel $\kappa$; see \cref{eq:product-measure-kernel} for the definition. In practice, if the structure of the measurable pointed graph is only defined by the forward reference kernel $\kappa_{F}$, then under mild assumptions we can find a measure $\nu$ and a kernel $\kappa_{B}$ satisfying the conditions of \cref{def:measurable-pointed-graph} \citep[][Definition 2]{cappe2005inferencehmm}. To draw a closer parallel with the notion of pointed DAGs in the discrete case, a measurable pointed graph is said to be \emph{finitely absorbing} if $\exists N > 0$, $\forall s \in \gS$
\begin{equation}
    \mathrm{supp}\big(\kappa_{F}^{N}(s, \cdot)\big) = \{\terminal\},
    \label{eq:finitely-absorbing}\index{Measurable pointed graph!Finitely absorbing}
\end{equation}
where $\mathrm{supp}(\cdot)$ represents the support of the measure, and where $\kappa_{F}^{N}$ is the composition kernel defined recursively in \cref{eq:recursive-composition-kernel} by applying the same kernel $\kappa_{F}$ $N$ times consecutively. This condition in \cref{eq:finitely-absorbing} means that any trajectory following $\kappa_{F}$ is guaranteed to eventually end up at $\terminal$ almost surely, in which case the minimal such integer $N$ is called the maximal trajectory length of $\gG$. Drawing further inspiration from the discrete case, each of the four properties in the definition above can be interpreted as follows: (1) encodes the fact that all the states are accessible from the initial state $s_{0}$, (2) represents the fact that $\terminal$ is a terminal state with no ``children'' other than itself, (3) encodes the fact that the kernels $\kappa_{F}$ and $\kappa_{B}$ are inverse of one another, relative to some measure $\nu$, and finally (4) represents the fact that the initial state $s_{0}$ has no ``parent'' (note that this does not adopt the ``wrapped-arround'' construction anymore).

\begin{example}[Pointed DAG as a measurable pointed graph]
    \label{ex:pointed-dag-measurable-pointed-graph}
    Finite state spaces with a pointed DAG structure are special cases of finitely absorbing measurable pointed graphs. Let $\gG = (\widebar{\gS}, \gA)$ be a pointed DAG, where $\widebar{\gS} = \gS \cup \{\terminal\}$ is a finite state space and $\gA \subseteq \gS \times \widebar{\gS}$ the set of directed edges, and let $s_{0} \in \gS$ be the initial state. The finite state space can be associated with the canonical $\sigma$-algebra $\widebar{\Sigma}$ corresponding to subsets of $\widebar{\gS}$ (\ie $\widebar{\Sigma} \equiv 2^{\widebar{\gS}}$ is the power set of $\widebar{\gS}$). For any state $s\in\gS$ and any subset $B \subseteq \widebar{\gS}$, the forward reference kernel $\kappa_{F}$ can be defined by the structure of $\gG$ as
    \begin{equation}
        \kappa_{F}(s, B) = \sum_{s'\in B}\mathds{1}(s' \in \children_{\gG}(s)) + \mathds{1}(s = \terminal)\mathds{1}(\terminal \in B).
    \end{equation}
    Similarly, for any state $s'\in\widebar{\gS}$ the backward reference kernel $\kappa_{B}$ can also be defined as
    \begin{equation}
        \kappa_{B}(s, B) = \sum_{s\in B}\mathds{1}(s\in\parents_{\gG}(s')),
    \end{equation}
    and the reference measure $\nu$ is simply the counting measure on $\widebar{\gS}$ (\ie the measure counting the number of elements in a subset of $\widebar{\gS}$). It can be shown that the kernels and measure defined this way satisfy all the properties of \cref{def:measurable-pointed-graph}. Moreover, \cref{eq:finitely-absorbing} is also satisfied thanks to the acyclicity of $\gG$.
\end{example}

Similar to how we defined forward and backward transition probability distributions over a pointed DAG in \cref{sec:forward-transition-probabilities,sec:backward-transition-probabilities}, we can also define a similar notion of Markov kernels \emph{compatible} with a measurable pointed graph $\gG$.
\begin{definition}[Compatible Markov kernels]
    \label{def:compatible-markov-kernels}\index{Kernel!Compatible kernels}
    Let $\gG = ((\widebar{\gS}, \widebar{\Sigma}), \kappa_{F}, \kappa_{B}, \nu)$ be a measurable pointed graph. A Markov kernel $\widebar{P}_{F}$ over $(\widebar{\gS}, \widebar{\Sigma})$ is called a \emph{forward Markov kernel} compatible with $\gG$ if for all $s\in\widebar{\gS}$, the distribution $P_{F}(s, \cdot)$ is absolutely continuous wrt.~the measure $\kappa_{F}(s, \cdot)$. Similarly, a Markov kernel $\widebar{P}_{B}$ is a \emph{backward Markov kernel} compatible with $\gG$ if for all $s\in\widebar{\gS}$, the distribution $P_{B}(s, \cdot)$ is absolutely continuous wrt.~the measure $\kappa_{B}(s, \cdot)$.
    \begin{align}
        \widebar{P}_{F}(s, \cdot) &\ll \kappa_{F}(s, \cdot) & \widebar{P}_{B}(s, \cdot) &\ll \kappa_{B}(s, \cdot)
    \end{align}
\end{definition}
We use the notation ``$\widebar{P}$'' to highlight the fact that these kernels are defined over $(\widebar{\gS}, \widebar{\Sigma})$ (unlike some of the kernels below). Although it was not initially interpreted as such by \citet{lahlou2023continuousgfn}, the new state $\terminal$ is introduced as an \emph{explicit} atom of a Markov chain that would be ``wrapped-around'' like we did in \cref{sec:cyclic-nature-flow-networks}. The following proposition defines the kernel of this Markov chain (over $(\gS, \Sigma)$ this time) precisely in terms of a forward Markov kernel compatible with a measurable pointed graph $\gG$. This new Markov kernel has the notable properties of being $\nu$-irreducible (\cref{def:phi-irreducibility}) and Harris recurrent (\cref{def:harris-recurrence}).
\begin{proposition}
    \label{prop:finitely-absorbing-measurable-pointed-graph-harris-recurrent}
    Let $\gG = ((\widebar{\gS}, \widebar{\Sigma}), \kappa_{F}, \kappa_{B}, \nu)$ be a finitely absorbing measurable pointed graph, and let $\widebar{P}_{F}$ be a forward Markov kernel compatible with $\gG$. The Markov kernel $P_{F}$ defined for any $s\in\gS$ and any $B\in\Sigma$ by
    \begin{equation}
        P_{F}(s, B) = \widebar{P}_{F}(s, B) + \widebar{P}_{F}(s, \{\terminal\})\delta_{s_{0}}(B),
    \end{equation}
    is $\nu$-irreducible and Harris recurrent over the measurable space $(\gS, \Sigma)$.
\end{proposition}

\begin{proof}
    $P_{F}$ is a well defined Markov kernel over $(\gS, \Sigma)$. Indeed, it is clearly a transition kernel since $\widebar{P}_{F}$ itself is a Markov kernel, and
    \begin{equation}
        P_{F}(s, \gS) = \widebar{P}_{F}(s, \gS) + \widebar{P}_{F}(s, \{\terminal\}) = \widebar{P}_{F}(s, \widebar{\gS}) = 1,
    \end{equation}
    since $s_{0}\in \gS$ and therefore $\delta_{s_{0}}(\gS) = 1$. Before showing Harris recurrence, we first show that $P_{F}$ is $\nu$-irreducible. Recall that by definition of finitely absorbing in \cref{eq:finitely-absorbing}, there exists a maximal trajectory length $N > 0$ of $\gG$ such that for any $s\in\gS$, $\mathrm{supp}\big(\kappa^{N}_{F}(s, \cdot)\big) = \{\terminal\}$. Since $\widebar{P}_{F}$ is compatible with $\gG$ (therefore $P_{F} \ll \kappa_{F}$), then we necessarily have that $\widebar{P}_{F}^{N}(s, \cdot) = \delta_{\terminal}(\cdot)$ as well. Moreover, we can show by induction that for any $m > 0$ and any set $A \in \Sigma$
    \begin{equation}
        P_{F}^{m}(s, A) \geq \widebar{P}_{F}^{m}(s, \{\terminal\})\delta_{s_{0}}(A).
    \end{equation}
    In particular, we have $P_{F}^{N}(s, A) \geq \delta_{s_{0}}(A)$ from what we saw above. Let $B\in\Sigma$ be a set such that $\nu(B) > 0$. By definition of a measurable pointed graph, in particular property (1) of \cref{def:measurable-pointed-graph}, there exists $n \geq 0$ such that $\kappa_{F}^{n}(s_{0}, B) > 0$. Again, since $\widebar{P}_{F}$ is compatible with $\gG$, we also have $\widebar{P}^{n}_{\!F}(s_{0}, B) > 0$. By the Chapman-Kolmogorov equation
    \begin{equation}
        P_{F}^{N+n}(s, B) = \int_{\gS}P_{F}^{N}(s, ds')P_{F}^{n}(s', B) \geq \int_{\gS}\delta_{s_{0}}(ds')P_{F}^{n}(s', B) = P_{F}^{n}(s_{0}, B) \geq \widebar{P}_{F}^{n}(s_{0}, B) > 0.
        \label{eq:finitely-absorbing-measurable-pointed-graph-harris-recurrent-proof-1}
    \end{equation}
    Since this is true for any state $s\in\gS$, and any set $B\in\Sigma$ such that $\nu(B) > 0$, we can conclude that $P_{F}$ is $\nu$-irreducible.
    
    Finally to show that $P_{F}$ is Harris recurrent, we can follow the same argument as above, but this time for any set $B \in \Sigma$ such that $\nu(B) > 0$, and any $s\in B$. The only additional requirement is to ensure that $N + n > 0$ (since the definition of Harris recurrence involves the \emph{return time} to $B$), which is clear since \cref{eq:finitely-absorbing-measurable-pointed-graph-harris-recurrent-proof-1} can be interpreted as hitting $\terminal\notin \gS$, and restarting at $s_{0}$ (therefore, there is at least a transition to $\terminal$ to account for).
\end{proof}
A consequence of this proposition is that the Markov kernel $P_{F}$ also admits an invariant measure, by \cref{thm:harris-recurrent-mc-invariant-measure}, \emph{regardless of the forward Markov kernel} compatible with $\gG$. It will play an important role in what follows. Interestingly, it is clear that $P_{F}$ also satisfies the minorization condition, since $P_{F}(s, B) \geq \widebar{P}_{F}(s, \{\terminal\})\delta_{s_{0}}(B)$; in that context, we have $\nu(B) = \delta_{s_{0}}(B)$ and $\varepsilon(s) = \widebar{P}_{F}(s, \{\terminal\})$ in \cref{def:minorization-condition}.

\subsection{Generalization of the flow matching condition}
\label{sec:generalized-flow-matching-condition}
We saw in the discrete case that the only difference between the flow matching condition and invariance was the inclusion of the initial state $s_{0}$ when we check those conditions. In measurable state space, we will have the same separation of the initial state. This \emph{generalized flow matching condition} will be similar to invariance in \cref{eq:invariant-measure-generalized}, except that we will not measure the response against any function $f$, but only those such that $f(s_{0}) = 0$, effectively ignoring the initial state.
\begin{definition}[Generalized flow matching condition]
    \label{def:generalized-flow-matching}\index{Flow matching!Generalized|textbf}
    Let $\gG = ((\widebar{\gS}, \widebar{\Sigma}), \kappa_{F}, \kappa_{B}, \nu)$ be a finitely absorbing measurable pointed graph. Let $\widebar{P}_{F}$ be a forward Markov kernel compatible with $\gG$, and $\widebar{F} \ll \nu$ be a $\sigma$-finite measure over $\widebar{\gS}$. The pair $(\widebar{F}, \widebar{P}_{F})$ is said to satisfy the \emph{flow matching condition} if for any bounded measurable function $f: \widebar{S} \rightarrow \sR$ such that $f(s_{0}) = 0$, it satisfies
    \begin{equation}
        \int_{\widebar{\gS}}f(s')\widebar{F}(ds') = \iint_{\gS\times\widebar{\gS}}f(s')\widebar{F}(ds)\widebar{P}_{F}(s, ds').
        \label{eq:generalized-flow-matching}
    \end{equation}
\end{definition}
Instead of being a function, the flow $\widebar{F}$ is now a measure over $\widebar{\gS}$. We saw in the previous section that we could construct a Markov kernel $P_{F}$, from $\widebar{P}_{F}$, that admits an invariant measure. The following proposition shows that this invariant measure is related to the flow when it satisfies the generalized flow matching condition; it happens to simply be the restriction of $\widebar{F}$ to the state space $\gS$ (\ie ignoring the terminal state $\terminal$).
\begin{proposition}
    \label{prop:generalized-flow-matching-invariant}
    Let $\gG = ((\widebar{\gS}, \widebar{\Sigma}), \kappa_{F}, \kappa_{B}, \nu)$ be a finitely absorbing measurable pointed graph. If a pair $(\widebar{F}, \widebar{P}_{F})$ satisfies the flow matching condition in \cref{def:generalized-flow-matching}, then the restriction of the measure $\widebar{F}$ over $\gS$ is invariant for the Markov kernel defined over the measurable space $(\gS, \Sigma)$ by
    \begin{equation}
        P_{F}(s, B) = \widebar{P}_{F}(s, B) + \widebar{P}_{F}(s, \{\terminal\})\delta_{s_{0}}(B).
    \end{equation}
\end{proposition}

\begin{proof}
    We use the notation $F$ to denote the restriction of $\widebar{F}$ (defined as a measure over $(\widebar{\gS}, \widebar{\Sigma})$) over $\gS$. For any bounded measurable function $g: \gS \rightarrow \sR$, there exists a unique bounded measurable function $f: \widebar{\gS} \rightarrow \sR$ such that $f(s_{0}) = 0$ that can be defined by
    \begin{align}
        f(s_{0}) &= 0 &&& f(\terminal) &= g(s_{0}) &&& \textrm{and $\forall s\neq s_{0}$,}\ \ f(s) &= g(s).
        \label{eq:generalized-flow-matching-invariant-proof-1}
    \end{align}
    This transformation is a bijection, meaning that we can go back and forth in a unique way between $g$ and $f$. Using such a function $g$, we have
    {\allowdisplaybreaks%
    \begin{align}
        \iint_{\gS\times\gS}&g(s')F(ds)P_{F}(s, ds')\nonumber\\
        &= \iint_{\gS\times\gS}g(s')\mathds{1}(s'\neq s_{0})F(ds)\widebar{P}_{F}(s, ds') + \int_{\gS}\underbrace{\left[\int_{\gS}g(s')\delta_{s_{0}}(ds')\right]}_{=\,g(s_{0})}F(ds)\widebar{P}_{F}(s, \{\terminal\})\label{eq:generalized-flow-matching-invariant-proof-2}\\
        &= \iint_{\gS\times\gS}f(s')\widebar{F}(ds)\widebar{P}_{F}(s, ds') + \int_{\gS}f(\terminal)\widebar{F}(ds)\widebar{P}_{F}(s, \{\terminal\})\label{eq:generalized-flow-matching-invariant-proof-3}\\
        &= \iint_{\gS\times\widebar{\gS}}f(s')\widebar{F}(ds)\widebar{P}_{F}(s, ds'),\\
        \intertext{where we used the definition of $P_{F}$ and the fact that $\widebar{P}_{F}(s, \{s_{0}\}) = 0$ since $\gG$ is finitely absorbing in \cref{eq:generalized-flow-matching-invariant-proof-2}, the bijection between $g$ \& $f$ and the fact that $\widebar{F}$ \& $F$ match on $\gS$ in \cref{eq:generalized-flow-matching-invariant-proof-3}. Continuing the derivation, we can apply the flow matching condition since $f$ is such that $f(s_{0}) = 0$:}
        &= \int_{\widebar{\gS}}f(s')\widebar{F}(ds')\\
        &= \int_{\gS}f(s')\widebar{F}(ds') + f(\terminal)\widebar{F}(\{\terminal\})\\
        &= \int_{\gS}g(s')\mathds{1}(s'\neq s_{0})F(ds') + g(s_{0})F(\{s_{0}\}) = \int_{\gS}g(s')F(ds'),\label{eq:generalized-flow-matching-invariant-proof-4}
    \end{align}}%
    where we used the bijection between $f$ and $g$, the fact that $\widebar{F}$ and $F$ match on $\gS$ and $\widebar{F}(\{\terminal\}) = \widebar{F}(\{s_{0}\}) = F(\{s_{0}\})$, and $f(s_{0}) = 0$ in \cref{eq:generalized-flow-matching-invariant-proof-4}. Since this holds for any bounded measurable function $g$, this shows that $F$ is invariant for $P_{F}$.
\end{proof}

\subsection{Terminating state distribution}
\label{sec:generalized-terminating-state-distribution}
Now that we have separated the structure of the state space through the finitely absorbing measurable pointed graph $\gG$ from the Markov kernel $\widebar{P}_{F}$, we can one last time define a notion of terminating state distribution. The domain $\gX$ of this distribution will be the ``parents'' of the terminal state $\terminal$ (\ie states from which $\terminal$ is directly accessible), just like in \cref{def:terminating-states} for pointed DAGs, as given by the forward transition kernel $\kappa_{F}$
\begin{equation}
    \gX = \{x\in\gS\mid \kappa_{F}(x, \{\terminal\}) > 0\}.
    \label{eq:generalized-terminating-state}
\end{equation}
This is fully compatible with the space $\gX$ in the minorization condition, which was the domain of distribution defined in \cref{def:generalized-terminating-state-distribution}, in the sense that $\varepsilon(x) = \widebar{P}_{F}(x, \{\terminal\})$ is zero if $\kappa_{F}(x, \{\terminal\}) = 0$ as well (by absolute continuity, because $\widebar{P}_{F}$ is compatible with $\gG$). We will use $(\widebar{X}_{n})_{n\geq 0}$ to denote the canonical Markov chain with kernel $\widebar{P}_{F}$ starting at $s_{0}$. Note that since $\gG$ is finitely absorbing \cref{eq:finitely-absorbing}, we necessarily have for all $n \geq N$, $\widebar{X}_{n} = \terminal$. Based on this chain, we can define the terminating state distribution associated with $\widebar{P}_{F}$ similarly to \cref{def:general-discrete-terminating-state-distribution}, as being for any $B\in\Sigma_{\gX}$
\begin{equation}
    \widebar{P}_{F}^{\top}(B) = \E_{s_{0}}\big[\mathds{1}_{B}(\widebar{X}_{\bar{\eta}_{\terminal} - 1})\big],
    \label{eq:generalized-terminating-state-distribution-practical}
\end{equation}
where $\bar{\eta}_{\terminal} = \inf\{n \geq 0 \mid \widebar{X}_{n} = \terminal\}$ is the \emph{hitting time} of the Markov chain at $\terminal$. Note that although the hitting time is typically defined as only being non-negative (as opposed to the \emph{return time} \cref{eq:return-time} being $\geq 1$), we necessarily have $\bar{\eta}_{\terminal} \geq 1$ since $s_{0} \in \gS$ cannot be equal to $\terminal \notin \gS$, making the terminating state distribution well defined. Once again, we can show that this is a properly defined probability distribution over $\gX$. To sample from $\widebar{P}_{F}^{\top}$, the strategy is exactly the same as in \cref{alg:sampling-terminating-state-probability} for pointed DAGs: starting at the initial state $s_{0}$, sample sequentially using the Markov kernel $\widebar{P}_{F}$ until we reach $\terminal$, and return the state obtained right before termination.

\paragraph{Boundary condition} Going back to our original problem of finding a Markov kernel whose terminating state distribution matches the target $P^{\star}$ in \cref{eq:general-target-distribution-gflownet-reward}, we still have to define a boundary condition, depending on the positive and finite measure $R$. Moreover, we will assume that $R \ll \nu$. Adapted to our setting here with a flow measure $\widebar{F}$ and a forward Markov kernel $\widebar{P}_{F}$ defined on top of a measurable pointed graph, the boundary condition in \cref{eq:generalized-boundary-condition} becomes
\begin{equation}
    \int_{\gX}f(x)R(dx) = \int_{\gX}f(x)\widebar{F}(dx)\widebar{P}_{F}(x, \{\terminal\}),
    \label{eq:practical-generalized-boundary-condition}
\end{equation}
for any bounded measurable function $f: \gX \rightarrow \sR$. Once again, this condition is reminiscent to \cref{eq:boundary-conditions} in the discrete case. Combining it with the flow matching condition, and after an excursion through the Markov chain perspective detailed in \cref{sec:gflownet-general-state-spaces}, we are now ready to state the direct extension of the fundamental theorem of GFlowNets (\cref{thm:flow-matching-proportional-reward}), this time for measurable spaces.

\begin{theorem}[Generalized flow matching condition]
    \label{thm:generalized-flow-matching-terminating-state}\index{Flow matching!Generalized}
    Let $\gG = ((\widebar{\gS}, \widebar{\Sigma}), \kappa_{F}, \kappa_{B}, \nu)$ be a finitely absorbing measurable pointed graph. Let $\widebar{P}_{F}$ be a forward Markov kernel compatible with $\gG$, $\widebar{F} \ll \nu$ be a $\sigma$-finite measure over $\widebar{\gS}$, and $R$ be a finite measure on $\gX \subseteq \gS$ such that $R\ll \nu$. If $(\widebar{F}, \widebar{P}_{F})$ satisfies the flow matching condition of \cref{def:generalized-flow-matching} and the following boundary condition for any bounded measurable function $f: \gX \rightarrow \sR$
    \begin{equation}
        \int_{\gX}f(x)R(dx) = \int_{\gX}f(x)\widebar{F}(dx)\widebar{P}_{F}(x, \{\terminal\}),
        \label{eq:generalized-boundary-condition-theorem}
    \end{equation}
    then the terminating state distribution associated with $\widebar{P}_{F}$ is proportional to the measure $R$: $\forall B\in\Sigma_{\gX}$, $\widebar{P}_{F}^{\top}(B) \propto R(B)$.
\end{theorem}
\begin{figure}[t]
    \centering
    \begin{adjustbox}{center}
    \includegraphics[width=480pt]{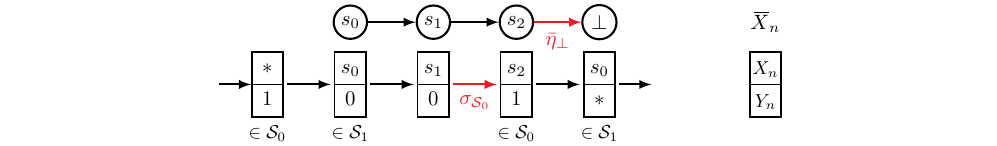}
    \end{adjustbox}
    \caption[Relation between the atom $\gS_{0}$ and $\gS_{1}$]{Relation between the atom $\gS_{0}$ for the split chain $Z_{n} = (X_{n}, Y_{n})$ and the set $\gS_{1}$. The value $\ast$ indicates that the value can be arbitrary. Returning to $\gS_{0}$ at random time $\sigma_{\gS_{0}}$ happens exactly one step before hitting the set $\gS_{1}$ (corresponding to hitting the terminating state for the Markov chain $\widebar{X}_{n}$).}
    \label{fig:return-time-proof}
\end{figure}
\begin{proof}
    This theorem is a direct consequence on \cref{thm:markov-chain-generalized-terminating-state-propto-reward}. If $(\widebar{F}, \widebar{P}_{F})$ satisfies the flow matching condition and, and since $\gG$ is a finitely absorbing measurable pointed graph, then by \cref{prop:finitely-absorbing-measurable-pointed-graph-harris-recurrent,prop:generalized-flow-matching-invariant} the Markov kernel over $(\gS, \Sigma)$ defined by
    \begin{equation}
        P_{F}(s, B) = \widebar{P}_{F}(s, B) + \widebar{P}_{F}(s, \{\terminal\})\delta_{s_{0}}(B)
        \label{eq:generalized-flow-matching-terminating-state-proof-4}
    \end{equation}
    is $\nu$-irreducible, Harris recurrent, and admits $F$, the restriction of $\widebar{F}$ on $\gS$, as an invariant measure. The set $\gX$ clearly satisfies the minorization condition of \cref{def:minorization-condition}, with the corresponding function $\varepsilon(s) = \widebar{P}_{F}(s, \{\terminal\})$: by \cref{eq:generalized-flow-matching-terminating-state-proof-4}, $P_{F}(s, B) \geq \varepsilon(s)\delta_{s_{0}}(B)$. Since $F$ and $\widebar{F}$ are equal on $\gX \subseteq \gS$, and with the form of $\varepsilon(s)$, the boundary condition \cref{eq:generalized-boundary-condition-theorem} is exactly equivalent to the one in \cref{eq:markov-chain-generalized-terminating-state-propto-reward-boundary-condition}. Therefore, by \cref{thm:markov-chain-generalized-terminating-state-propto-reward}, we get that for all $B\in\Sigma_{\gX}$, $P_{F}^{\top}(B) \propto R(B)$.

    To conclude, we only need to prove that $\widebar{P}_{F}^{\top}(B) = P_{F}^{\top}(B)$. First, note that in addition to the function $\varepsilon$, we can also identify the measure $\nu(B) = \delta_{s_{0}}(B)$ in the minorization condition. We introduce the set $\gS_{1} = \{s_{0}\} \times \{0, 1\}$, by analogy with $\gS_{0} = \gX \times \{1\}$ appearing in the definition of $P_{F}^{\top}$ \cref{eq:generalized-terminating-state-distribution}; we chose the index ``1'' because transitioning from any element of $\gS_{0}$ is guaranteed to yield the state $s_{0}$ as its first component (because $\nu = \delta_{s_{0}}$), and the second component of the split chain (called ``$Y$'' in \cref{sec:creation-atom-splitting-technique}) may be arbitrary.
    {\allowdisplaybreaks%
    \begin{align}
        P_{F}^{\top}(B) &= \E_{\gS_{0}}\big[\mathds{1}_{B}(X_{\sigma_{\gS_{0}}})\big] = \sum_{n=1}^{\infty}\E_{\gS_{0}}\big[\mathds{1}(n = \sigma_{\gS_{0}})\mathds{1}_{B}(X_{n})\big]\\
        &= \sum_{n=1}^{\infty}\E_{\gS_{0}}\big[\mathds{1}(n \leq \sigma_{\gS_{0}})\mathds{1}_{B}(X_{n})\widebar{P}_{F}(X_{n}, \{\terminal\})\big],\label{eq:generalized-flow-matching-terminating-state-proof-1}
        \intertext{where we used \cref{eq:generalized-terminating-state-distribution-invariant-measure-proof-4} and the definition of $\varepsilon(s) = \widebar{P}_{F}(s, \{\terminal\})$ in \cref{eq:generalized-flow-matching-terminating-state-proof-1}. Since we are guaranteed to transition to an element of $\gS_{1}$ from $\gS_{0}$ we get, using Markov property}
        &= \sum_{n=1}^{\infty}\E_{\gS_{1}}\big[\mathds{1}(n - 1 \leq \sigma_{\gS_{0}})\mathds{1}_{B}(X_{n-1})\widebar{P}_{F}(X_{n-1}, \{\terminal\})\big]\\
        &= \sum_{n=0}^{\infty}\E_{\gS_{1}}\big[\mathds{1}(n \leq \sigma_{\gS_{0}})\mathds{1}_{B}(X_{n})\widebar{P}_{F}(X_{n}, \{\terminal\})\big]\\
        &= \sum_{n=0}^{\infty}\E_{s_{0}}\big[\mathds{1}(n \leq \bar{\eta}_{\terminal} - 1)\mathds{1}_{B}(\widebar{X}_{n})\widebar{P}_{F}(\widebar{X}_{n}, \{\terminal\})\big],\label{eq:generalized-flow-matching-terminating-state-proof-2}
        \intertext{where we used the link between $\sigma_{\gS_{0}}$ and $\bar{\eta}_{\terminal}$ illustrated in \cref{fig:return-time-proof}, and the fact that the Markov chain $\widebar{X}_{n}$ behaves exactly as $X_{n}$ before hitting the terminal state to get \cref{eq:generalized-flow-matching-terminating-state-proof-2}. Using an argument similar to \cref{eq:generalized-terminating-state-distribution-invariant-measure-proof-4} used earlier in this proof, we can conclude that}
        &= \sum_{n=0}^{\infty}\E_{s_{0}}\big[\mathds{1}(n = \bar{\eta}_{\terminal} - 1)\mathds{1}_{B}(\widebar{X}_{n})\big] = \E_{s_{0}}\big[\mathds{1}_{B}(\widebar{X}_{\bar{\eta} - 1})\big] = \widebar{P}_{F}^{\top}(B).\label{eq:generalized-flow-matching-terminating-state-proof-3}
    \end{align}}%
    This concludes the proof.
\end{proof}
The same way we introduced flow matching losses in order to search for quantities that were satisfying the conditions of \cref{thm:flow-matching-proportional-reward} (or any flow matching condition for that matter), we will also introduce similar losses in \cref{sec:generalized-flow-matching-losses} to find a flow and Markov kernel $(\widebar{F}, \widebar{P}_{F})$ satisfying the conditions of this theorem.  %

\subsection{Generalized detailed balance \& trajectory balance conditions}
\label{sec:generalized-detailed-balance-trajectory-balance-conditions}
So far in this chapter, we have focused exclusively on the flow matching condition, and its relation to invariant measures, first from the perspective of Markov chains, then measurable pointed graphs (analogous to pointed DAGs). In this section, we will derive similar extensions of the other flow matching conditions we introduced in \cref{sec:flow-matching-conditions}. Starting with the \emph{detailed balance} condition, the following proposition shows in particular that the generalization of the detailed balance condition actually implies the flow matching condition of \cref{def:generalized-flow-matching}.

\begin{proposition}[Generalized detailed balance condition]
    \label{prop:generalized-detailed-balance}\index{Detailed balance!Generalized|textbf}
    Let $\gG = ((\widebar{\gS}, \widebar{\Sigma}), \kappa_{F}, \kappa_{B}, \nu)$ be a finitely absorbing measurable pointed graph. Let $\widebar{P}_{F}$ be a forward Markov kernel and $\widebar{P}_{B}$ be a backward Markov kernel compatible with $\gG$, and let $\widebar{F} \ll \nu$ be a $\sigma$-finite measure over $\widebar{\gS}$. The tuple $(\widebar{F}, \widebar{P}_{F}, \widebar{P}_{B})$ is said to satisfy the \emph{detailed balance condition} if for any bounded measurable function $f: \gS \times \widebar{\gS} \rightarrow \sR$ such that $f(s, s_{0}) = 0$ for any $s\in\gS$, it satisfies:
    \begin{equation}
        \iint_{\gS\times \widebar{\gS}}f(s, s')\widebar{F}(ds)\widebar{P}_{F}(s, ds') = \iint_{\gS\times \widebar{\gS}}f(s, s')\widebar{F}(ds')\widebar{P}_{B}(s', ds).
        \label{eq:generalized-detailed-balance}
    \end{equation}
    Moreover if $(\widebar{F}, \widebar{P}_{F}, \widebar{P}_{B})$ satisfy the detailed balance condition above, then $(\widebar{F}, \widebar{P}_{F})$ satisfy the flow matching condition of \cref{def:generalized-flow-matching}.
\end{proposition}
\begin{proof}
    Following the properties of \cref{def:measurable-pointed-graph}, especially (2) \& (3), we have $\kappa_{B}(s', \{\terminal\}) = 0$ for any $s'\in\gS$. Since $\widebar{P}_{B}$ is compatible with $\gG$, it is clear that we also have $\widebar{P}_{B}(s', \{\terminal\}) = 0$. For any bounded measurable function $f: \widebar{\gS} \rightarrow \sR$ such that $f(s_{0}) = 0$, we can define a function $g: \gS \times \widebar{\gS} \rightarrow \sR$ such that for all $(s, s') \in \gS \times \widebar{\gS}$, $g(s, s') = f(s')$. Note that $g$ satisfies $g(s, s_{0}) = 0$ for any $s\in\gS$. We can show that $(\widebar{F}, \widebar{P}_{F})$ satisfies the flow matching condition:
    {\allowdisplaybreaks%
    \begin{align}
        \iint_{\gS \times \widebar{\gS}}f(s')\widebar{F}(ds)\widebar{P}_{F}(s, ds') &= \iint_{\gS\times\widebar{\gS}}g(s, s')\widebar{F}(ds)\widebar{P}_{F}(s, ds')\\
        &= \iint_{\gS\times\widebar{\gS}}g(s, s')\widebar{F}(ds')\widebar{P}_{B}(s', ds)\label{eq:generalized-detailed-balance-proof-1}\\
        &= \int_{\widebar{\gS}}f(s')\widebar{F}(ds')\int_{\gS}\widebar{P}_{B}(s', ds) = \int_{\widebar{\gS}}f(s')\widebar{F}(ds'),\label{eq:generalized-detailed-balance-proof-2}
    \end{align}}%
    where we used the detailed balance condition in \cref{eq:generalized-detailed-balance-proof-1}, and the fact that $\widebar{P}_{B}$ is a Markov kernel and $\widebar{P}_{B}(s', \{\terminal\}) = 0$ in \cref{eq:generalized-detailed-balance-proof-2}.
\end{proof}

Since we reduced the generalized detailed balance condition to the generalized flow matching condition, it is clear by \cref{thm:generalized-flow-matching-terminating-state} that satisfying it along with the boundary condition \cref{eq:generalized-boundary-condition-theorem} implies that the corresponding terminating state distribution will also be proportional to the measure $R$. This result matches \cref{thm:detailed-balance-proportional-reward} in the case of discrete GFlowNets.

\begin{corollary}
    \label{cor:generalized-detailed-balance-propto-reward}
    If $(\widebar{F}, \widebar{P}_{F}, \widebar{P}_{B})$ satisfy the generalized detailed balance condition of \cref{prop:generalized-detailed-balance}, along with the boundary condition of \cref{eq:generalized-boundary-condition-theorem}, then the terminating state distribution associated with $\widebar{P}_{F}$ is proportional to the measure $R$: $\forall B\in\Sigma_{\gX}$, $\widebar{P}_{F}^{\top}(B) \propto R(B)$.
\end{corollary}

Finally, we can do the same exercise with the \emph{trajectory balance condition}. Since we saw that this is a conservation law at the level of complete trajectories, we have to extend the notion of (one step) Markov kernel $\widebar{P}_{F}$ (and $\widebar{P}_{B}$) to distributions over trajectories \citep{petritis2015markovchainmeasurable}; see \cref{app:operations-transition-kernels} for details. The following theorem generalizes \cref{thm:trajectory-balance-proportional-reward} to measurable spaces; this condition is more involved than those we have seen thus far.

\begin{theorem}[Generalized trajectory balance condition]
    \label{thm:generalized-trajectory-balance}\index{Trajectory balance!Generalized}
    Let $\gG = ((\widebar{S}, \widebar{\Sigma}), \kappa_{F}, \kappa_{B}, \nu)$ be a finitely absorbing measurable pointed graph. Let $\widebar{P}_{F}$ be a forward Markov kernel and $\widebar{P}_{B}$ be a backward Markov kernel compatible with $\gG$, and let $\widebar{Z} > 0$ be a positive scalar. The tuple $(\widebar{Z}, \widebar{P}_{F}, \widebar{P}_{B})$ is said to satisfy the \emph{trajectory balance condition} wrt. a positive and finite measure $R$ on $\gX \subseteq \gS$ (s.t.~$R\ll \nu$) if for any $T \geq 0$ and any bounded measurable function $f: \widebar{S}^{T+2} \rightarrow \sR$, it satisfies:
    \begin{equation}
    \begin{aligned}
        \int_{\widebar{\gS}^{T+1}}\widebar{Z}f(s_{0},& s_{1}, \ldots, s_{T}, s_{T+1})\mathds{1}(s_{T+1} = \terminal)\widebar{P}_{F}^{\otimes T+1}(s_{0}, ds_{1:T+1})\\
        &= \int_{\widebar{\gS}^{T+1}}R(ds_{T})f(s, s_{1}, \ldots, s_{T}, \terminal)\mathds{1}(s = s_{0})\widebar{P}_{B}^{\otimes T}(s_{T}, ds_{T-1:1}ds),
    \end{aligned}
    \label{eq:generalized-trajectory-balance}
    \end{equation}
    where $\widebar{P}_{F}^{\otimes T+1}$ \& $\widebar{P}_{B}^{\otimes T}$ are the product kernels defined recursively by \cref{eq:recursive-product-kernel}. Moreover, if $(\widebar{Z}, \widebar{P}_{F}, \widebar{P}_{B})$ satisfy the trajectory balance condition above, then the forward kernel $\widebar{P}_{F}$ and the measure $\widebar{F}$ defined for all $B\in\Sigma$ by
    \begin{equation}
        \widebar{F}(B) = \widebar{Z}\sum_{n=0}^{\infty}\widebar{P}_{F}^{n}(s_{0}, B),
        \label{eq:generalized-tb-invariant-measure}
    \end{equation}
    where $\widebar{P}_{F}^{n}$ is the composition kernel defined recursively by \cref{eq:recursive-composition-kernel}, satisfy the flow matching condition of \cref{def:generalized-flow-matching} as well as the boundary condition \cref{eq:generalized-boundary-condition-theorem}. Therefore, by \cref{thm:generalized-flow-matching-terminating-state}, the terminating state distribution associated with $\widebar{P}_{F}$ is proportional to $R$: $\widebar{P}_{F}^{\top}(B) \propto R(B)$ for any $B \in \Sigma_{\gX}$.
\end{theorem}

\begin{proof}
    Let $f: \widebar{\gS} \rightarrow \sR$ be a bounded measurable function such that $f(s_{0}) = 0$. We can first show that $(\widebar{F}, \widebar{P}_{F})$ with $\widebar{F}$ defined by \cref{eq:generalized-tb-invariant-measure} satisfy the flow matching condition.
    {\allowdisplaybreaks%
    \begin{align}
        \iint_{\gS\times \widebar{\gS}}f(s')F(ds)\widebar{P}_{F}(s, ds') &= \iint_{\gS\times \widebar{\gS}}f(s')Z\sum_{n=0}^{\infty}\widebar{P}_{F}^{n}(s_{0}, ds)\widebar{P}_{F}(s, ds')\\
        &= \int_{\widebar{\gS}}f(s')\widebar{Z}\sum_{n=0}^{\infty}\widebar{P}_{F}^{n+1}(s_{0}, ds')\label{eq:proof-generalized-trajectory-balance-1}\\
        &= \int_{\widebar{\gS}}f(s')\widebar{Z}\sum_{n=0}^{\infty}\widebar{P}_{F}^{n}(s_{0}, ds') = \int_{\widebar{\gS}}f(s')\widebar{F}(ds'),\label{eq:proof-generalized-trajectory-balance-2}
    \end{align}}%
    where we used the Chapman–Kolmogorov equation in \cref{eq:proof-generalized-trajectory-balance-1}, and the fact that $\widebar{P}_{F}^{0}(s_{0}, \cdot) = \delta_{s_{0}}(\cdot)$ and $f(s_{0}) = 0$ in \cref{eq:proof-generalized-trajectory-balance-2}. To show that they also satisfy the boundary condition, we can first observe that for any state $s\in\widebar{\gS}$
    \begin{equation}
        \sum_{n=0}^{\infty}\widebar{P}_{B}^{n}(s, \{s_{0}\}) = 1.
        \label{eq:proof-generalized-trajectory-balance-3}
    \end{equation}
    This is the generalization of \cref{lem:PB-distribution-prefix} establishing that a backward kernel induces a distribution over trajectories back to $s_{0}$; it follows from the finitely absorbing nature of $\gG$ and the properties of \cref{def:measurable-pointed-graph}. See \citep[][Proposition 5]{lahlou2023continuousgfn}.
    {\allowdisplaybreaks%
    \begin{align}
        \int_{\gX}f(s)\widebar{F}(ds)\widebar{P}_{F}(s, \{\terminal\}) &= \int_{\widebar{\gS}}\mathds{1}_{\gX}(s)f(s)Z\sum_{n=0}^{\infty}\widebar{P}_{F}^{n}(s_{0}, ds)\widebar{P}_{F}(s, \{\terminal\})\\
        &= \sum_{n=0}^{\infty}\int_{\widebar{\gS}^{n+1}}\mathds{1}_{\gX}(s_{n})f(s_{n})\widebar{Z}\mathds{1}(s_{n+1} = \terminal)\widebar{P}_{F}^{\otimes n+1}(s_{0}, ds_{1:n+1})\label{eq:proof-generalized-trajectory-balance-4}\\
        &= \sum_{n=0}^{\infty}\int_{\widebar{\gS}^{n+1}}R(ds_{n})\mathds{1}_{\gX}(s_{n})f(s_{n})\mathds{1}(s' = s_{0})\widebar{P}_{B}^{\otimes n}(s_{n}, ds_{n-1:1}ds')\label{eq:proof-generalized-trajectory-balance-5}\\
        &= \int_{\widebar{\gS}}\mathds{1}_{\gX}(s)f(s)R(ds)\sum_{n=0}^{\infty}\int_{\widebar{\gS}^{n}}\widebar{P}_{B}^{\otimes n}(s, ds_{n-1:1}ds')\mathds{1}(s'=s_{0})\\
        &= \int_{\widebar{\gS}}\mathds{1}_{\gX}(s)f(s)R(ds)\sum_{n=0}^{\infty}\widebar{P}_{B}^{n}(s, \{s_{0}\}) = \int_{\gX}f(s)R(ds),\label{eq:proof-generalized-trajectory-balance-6}
    \end{align}}%
    where we used the marginalization of $\widebar{P}_{F}^{\otimes n + 1}$ in \cref{eq:proof-generalized-trajectory-balance-4} (see \cref{prop:marginalization-product-kernel-composition-kernel}), the trajectory balance condition in \cref{eq:proof-generalized-trajectory-balance-5}, and \cref{eq:proof-generalized-trajectory-balance-3} in \cref{eq:proof-generalized-trajectory-balance-6}. This shows that $(\widebar{F}, \widebar{P}_{F})$ also satisfy the boundary condition. Finally, we can conclude that the terminating state distribution $\widebar{P}_{F}^{\top}$ is proportional to the measure $R$ by \cref{thm:generalized-flow-matching-terminating-state}.
\end{proof}

Interestingly, just like in the discrete case, the theorem above directly integrates the reward measure in the condition \cref{eq:generalized-trajectory-balance}. The functions $\mathds{1}(s_{T+1} = \terminal)$ and $\mathds{1}(s = s_{0})$ ensure that we only take into account ``complete trajectories''. Similar to the generalized detailed balance condition, we proved this theorem by reducing the trajectory balance condition to the flow matching condition (and boundary condition).

\subsection{Generalized flow matching losses}
\label{sec:generalized-flow-matching-losses}
Finally in this section, we will see that the flow matching losses we presented in \cref{sec:flow-matching-losses} transfer with minimal modifications to the general case. In particular, all of these losses will still take the form of a non-linear least square objective of the form
\begin{equation}
    \gL(\phi) = \frac{1}{2}\E_{\pi_{b}}\big[\Delta^{2}(\cdot; \phi)\big],
    \label{eq:generalized-least-square-objective}
\end{equation}
where $\Delta(\cdot; \phi)$ is a residual\index{Residual} which depends on the condition being applied, and $\pi_{b}$ is a behavior policy with full support. These losses are still \emph{off-policy}, and the considerations of \cref{sec:off-policy-training} still hold (we will see some other considerations more specifically for continuous spaces later in this section).

All of the conditions we have seen in \cref{sec:generalized-terminating-state-distribution,sec:generalized-detailed-balance-trajectory-balance-conditions} involve to some extent measures $\widebar{F}$ \& $R$ and/or Markov kernels compatible with a measurable pointed graph $\gG$. Since we assumed absolute continuity of all these quantities wrt.~the components of $\gG$, we can define them in terms of their \emph{density} (or Radon-Nikodym derivatives). For example, since we assumed that $\widebar{F} \ll \nu$ (\cref{def:generalized-flow-matching}) and $R \ll \nu$, there exist two functions $\widebar{f}: \widebar{\gS} \rightarrow \sR^{+}$ and $r: \gX \rightarrow \sR^{+}$ such that for all $B\in\Sigma$ and $B' \in \Sigma_{\gX}$,
\begin{align}
    \widebar{F}(B) &= \int_{B}\widebar{f}(s)\nu(ds)
    \label{eq:density-flow} && \textrm{and} & R(B') &= \int_{B'}r(x)\nu(dx).
\end{align}
Similarly, since the Markov kernels $\widebar{P}_{F}$ \& $\widebar{P}_{B}$ are compatible with $\gG$ (\cref{def:compatible-markov-kernels}), then there exist two functions $\widebar{p}_{F}: \gS \times \widebar{\gS} \rightarrow \sR^{+}$ and $\widebar{p}_{B}: \gS \times \gS \rightarrow \sR^{+}$ such that for all $B\in\widebar{\Sigma}$
\begin{align}
    \widebar{P}_{F}(B) &= \int_{B}\widebar{p}_{F}(s, s')\kappa_{F}(s, ds') && \textrm{and} & \widebar{P}_{B}(B) &= \int_{B}\widebar{p}_{B}(s', s)\kappa_{B}(s', ds).
    \label{eq:density-markov-kernel}
\end{align}
With the exception of $r$, which we will assumed to be fixed depending on our problem of interest (recall that we want to sample from a distribution proportional to $R$), all these densities may be parametrized more easily with some $\phi$; for example in a continuous space $\gS = \sR$, the density of the forward Markov kernel can be parametrized as the density of a Normal distribution with fixed variance and centered around a function of the previous state that would depend on $\phi$
\begin{equation}
    \widebar{p}_{F}^{\phi}(s, s') = \frac{1}{\sqrt{2\pi \sigma^{2}}}\exp\left(-\frac{(s' - \mu_{\phi}(s))^{2}}{2\sigma^{2}}\right).
    \label{eq:normal-distribution-pf-general}
\end{equation}
Taking inspiration from the flow matching condition in \cref{def:generalized-flow-matching} and \cref{thm:generalized-flow-matching-terminating-state}, we can define for any $s' \neq s_{0}$ the residual for the \emph{flow matching loss}
\begin{equation}
    \Delta_{\mathrm{FM}}(s'; \phi) = \log \frac{\int_{\gS}\widebar{f}_{\phi}(s)\widebar{p}^{\phi}_{F}(s, s')\kappa_{B}(s', ds)}{\widebar{f}_{\phi}(s')}.
    \label{eq:generalized-flow-matching-loss}\index{Flow matching!Loss}
\end{equation}
At first glance, this residual may seem quite different from the flow matching condition, because of the inclusion of $\kappa_{B}$. It is at least not as straightforward as the one we derived in \cref{eq:flow-matching-loss} based on \cref{eq:flow-matching-boundary-condition} in the discrete case. The following proposition shows that this formulation is actually correct, in the sense that if this residual is zero almost everywhere, then the generalized flow matching condition is satisfied.
\begin{proposition}
    \label{prop:generalized-flow-matching-densities}
    Let $\gG = ((\widebar{\gS}, \widebar{\Sigma}), \kappa_{F}, \kappa_{B}, \nu)$ be a finitely absorbing measurable pointed graph, and let $\widebar{f}: \widebar{\gS} \rightarrow \sR^{+}$ be the density of a flow measure $\widebar{F}$ and $\widebar{p}_{F}: \gS\times \widebar{\gS} \rightarrow \sR^{+}$ the density of a forward Markov kernel $\widebar{P}_{F}$ compatible with $\gG$. For $s'\neq s_{0}$, if
    \begin{equation}
        \widebar{f}(s') = \int_{\gS}\widebar{f}(s)\widebar{p}_{F}(s, s')\kappa_{B}(s', ds)
        \label{eq:generalized-flow-matching-densities}
    \end{equation}
    $\nu$-almost surely, then $(\widebar{F}, \widebar{P}_{F})$ satisfy the flow matching condition of \cref{def:generalized-flow-matching}.
\end{proposition}

\begin{proof}
    We can go back and forth between $\widebar{F}$ \& $\widebar{P}_{F}$ and their corresponding densities. For any bounded measurable function $f: \widebar{\gS} \rightarrow \sR^{+}$ such that $f(s_{0}) = 0$, we have
    {\allowdisplaybreaks%
    \begin{align}
        \int_{\widebar{\gS}}f(s')\widebar{F}(ds') &= \int_{\widebar{\gS}}f(s')\widebar{f}(s')\nu(ds')\\
        &= \int_{\widebar{\gS}}f(s')\int_{\gS}\widebar{f}(s)\widebar{p}_{F}(s, s')\kappa_{B}(s', ds)\nu(ds')\label{eq:generalized-flow-matching-densities-proof-1}\\
        &= \iint_{\gS\times \widebar{\gS}}f(s')\widebar{f}(s)\widebar{p}_{F}(s, s')\kappa_{F}(s, ds')\nu(ds) \label{eq:generalized-flow-matching-densities-proof-2}\\
        &= \iint_{\gS\times \widebar{\gS}}f(s')\widebar{F}(s)\widebar{P}_{F}(s, ds'),
    \end{align}}%
    where we used the condition \cref{eq:generalized-flow-matching-densities} in \cref{eq:generalized-flow-matching-densities-proof-1}, and property (3) of the definition of a measurable pointed graph (\cref{def:measurable-pointed-graph}) in \cref{eq:generalized-flow-matching-densities-proof-2}. We implicitly used the fact that $f(s_{0}) = 0$ because the condition \cref{eq:generalized-flow-matching-densities} is only valid $\nu$-almost surely for any $s'\neq s_{0}$. This concludes the proof.
\end{proof}
It is noteworthy that property (3) of \cref{def:measurable-pointed-graph} has never been used until this very point. Indeed, this condition was included in this definition specifically for working with densities; in fact, this is a property that would be necessary also to prove similar statements for other flow matching conditions. Regarding the flow matching condition, we also need to incorporate a residual for the boundary condition \cref{eq:generalized-boundary-condition-theorem}; for all $x\in\gX$
\begin{equation}
    \Delta_{\mathrm{FM}}(x; \phi) = \log \frac{\widebar{f}_{\phi}(x)\widebar{p}_{F}^{\phi}(x, \terminal)}{r(x)}.
    \label{eq:generalized-reward-matching-loss}
\end{equation}
This is also called a \emph{reward matching loss} \citep{bengio2023gflownetfoundations,lahlou2023continuousgfn}. In the case of these two residuals, the behavior policy $\pi_{b}$ must be a distribution over states. Similarly, the residual for the detailed balance condition in \cref{prop:generalized-detailed-balance}, leading to the \emph{generalized detailed balance loss}, is given for any $s, s'\in\gS$ by
\begin{equation}
    \Delta_{\mathrm{DB}}(s \rightarrow s'; \phi) = \log \frac{\widebar{f}_{\phi}(s)\widebar{p}_{F}^{\phi}(s, s')}{\widebar{f}_{\phi}(s')\widebar{p}_{B}^{\phi}(s', s)}.
    \label{eq:generalized-detailed-balance-loss}\index{Detailed balance!Loss}
\end{equation}
Just like the flow matching loss, this detailed balance loss must go along with a reward matching loss of the form \cref{eq:generalized-reward-matching-loss}, with $\Delta_{\mathrm{DB}}(x\rightarrow \terminal; \phi) = \Delta_{\mathrm{FM}}(x; \phi)$. The behavior policy $\pi_{b}$ is a distribution over transitions in $\gG$. Finally, the residual used in the \emph{generalized trajectory balance loss} is given for any $\tau = (s_{0}, s_{1}, \ldots, s_{T}, \terminal)$ by
\begin{equation}
    \Delta_{\mathrm{TB}}(\tau; \phi) = \log \frac{\widebar{Z}_{\phi}\prod_{t=0}^{T}\widebar{p}_{F}^{\phi}(s_{t}, s_{t+1})}{r(s_{T})\prod_{t=1}^{T}\widebar{p}_{B}^{\phi}(s_{t}, s_{t-1})}
    \label{eq:generalized-trajectory-balance-loss}\index{Trajectory balance!Loss}
\end{equation}
where we can also parametrize the scalar $\widebar{Z}_{\phi}$. The behavior policy $\pi_{b}$ in that case is a distribution over ``complete trajectories'' (\ie trajectories that start at $s_{0}$ and end in $\terminal$). It is worth noting that with the exception of the flow matching loss in \cref{eq:generalized-flow-matching-loss}, all the residuals presented here match almost identically their discrete counterparts in \cref{sec:flow-matching-losses}. The only major modification is to replace any probability mass functions by densities in the general case.

\paragraph{Off-policy training} We saw in \cref{sec:off-policy-training} some options such as $\varepsilon$-sampling in order to define a behavior policy $\pi_{b}$ that encourages exploration. In the case of a continuous state space though, this strategy often does not make sense any longer when $\bar{p}_{F}^{\phi}$ does not have a finite support (\eg when it is a Normal distribution). For example if the forward transition probability is a Normal distribution, an option to encourage exploration is to add an extra variance term (\ie replacing $\sigma^{2}$ by $\sigma^{2} + \varepsilon^{2}$ in \cref{eq:normal-distribution-pf-general}; \citealp{lahlou2023continuousgfn}).

%% file: chapters/07_DAG_GFlowNet.tex
\chapter[Bayesian Structure Learning with Generative Flow Networks]{Bayesian Structure Learning\\with Generative Flow Networks}
\label{chap:dag-gflownet}

\begin{minipage}{\textwidth}
    \itshape
    This chapter contains material from the following paper:
    \begin{itemize}[noitemsep, topsep=1ex, itemsep=1ex, leftmargin=3em]
        \item \textbf{Tristan Deleu}, Ant\'{o}nio G\'{o}is, Chris Emezue, Mansi Rankawat, Simon Lacoste-Julien, Stefan Bauer, Yoshua Bengio (2022). \emph{Bayesian Structure Learning with Generative Flow Networks}. Conference on Uncertainty in Artificial Intelligence (UAI). \notecite{deleu2022daggflownet}
    \end{itemize}
    \vspace*{5em}
\end{minipage}

In the first part of this thesis, we have presented a comprehensive view of the theory underlying generative flow networks as a framework to describe distributions over discrete and compositional objects. In the second part of this thesis, we will use these tools for the problem of \emph{structure learning} that we introduced in \cref{sec:structure-learning}, but this time from a \emph{Bayesian} perspective. GFlowNets appear as an ideal fit for this problem, since the DAG structure of a Bayesian network over which we will define a posterior distribution is a discrete object and has a natural decomposition as a collection of directed edges (\cref{fig:compositional-objects}).

\section{Bayesian structure learning}
\label{sec:bayesian-structure-learning}\index{Structure learning!Bayesian structure learning}
We saw in \cref{sec:structure-learning} that finding \emph{the best} DAG structure of a Bayesian network from data alone was particularly challenging for a number of reasons that we will briefly recall here. From a practical point of view, we saw in \cref{sec:score-based-methods} that treating structure learning as an optimization problem meant that we had to search for a single best scoring DAG in a space that was not only discrete (ruling out gradient based methods for optimization), but was most importantly combinatorially large (the number of DAGs with $d$ nodes being $2^{\Theta(d^{2})}$). Moreover, these methods also rely implicitly on the fact that we can ``trust'' the score being maximized, that is we have enough data to draw any conclusion. If the dataset $\gD$ is small though, we may be in situations where either (1) the DAG with the highest score may be an artifact of the finiteness of $\gD$, akin to ``overfitting'' in machine learning, or (2) many other (not necessarily Markov equivalent) DAGs might be slightly suboptimal due to the lack of evidence and wouldn't be considered if all we search for is a single graph.

Another major challenge of structure learning is the problem of \emph{identifiability}: even on the other end of the spectrum where we have a large amount of data, there are still inherent ambiguities remaining due to Markov equivalence (\cref{sec:markov-equivalence}). This means that any method finding a single best DAG has the risk of returning only an arbitrary element of the Markov equivalence class--- unless we work with the CPDAG representation \citep{spirtes2000causationpredictionsearch,chickering2002ges}, and even in that case making predictions would still require either (1) the selection of an arbitrary element of the Markov equivalence class (MEC), or (2) an aggregation of the predictions made for a possibly exponential number of elements in the MEC. This may not be problematic as long as we are fully aware that Bayesian networks encode purely statistical dependencies, but this becomes an issue as soon as we want to apply structure learning for causal purposes (\ie \emph{causal discovery}). Despite largely using the same tools as those of \cref{sec:structure-learning}, returning an arbitrary DAG from the Markov equivalence class now has the risk of leading to unsafe, or even confidently wrong, causal predictions.

We could instead consider \emph{all} the structures that are compatible with the data when making our predictions (\citealp{madigan1994occam}; see also \cref{sec:bayesian-model-averaging}). We take a Bayesian perspective of structure learning \citep{madigan1994enhancing}, where our objective is no longer to find a single DAG, but to characterize the whole posterior distribution $P(G\mid \gD)$ given a dataset of observations $\gD$. Recall from Bayes' rule that the posterior is given by
\begin{equation}
    P(G\mid \gD) = \frac{P(\gD\mid G)P(G)}{P(\gD)},
    \label{eq:marginal-posterior-dags}\index{Bayesian inference!Bayes' rule}
\end{equation}
where $P(\gD\mid G)$ is the marginal likelihood and $P(G)$ is the prior over graphs. We are using the \emph{marginal} likelihood here since the parameters $\theta$ of the Bayesian network have been marginalized over. Although computing $P(\gD\mid G)$ is in general intractable and may necessitate some approximations \citep{cheeseman1996cheesemanstutz,chickering1997approxmarginallikelihood}, we will assume in this chapter that we can compute the marginal likelihood analytically and efficiently \citep{cooper1992structurelearning}. For the prior $P(G)$, some options have been proposed in the literature that would encourage sparsity \citep{eggeling2019structureprior}, or are edge-dependent \citep{rittel2023priorbayesiancsl}.

Finding this posterior is at least as difficult as the original problem of structure learning. Indeed, now we don't simply have to search for one element in the vast space of DAGs, but we have to specify one value (probability) for each individual DAG, and all of this while maintaining the normalization of the distribution; this makes the constant $P(\gD)$ in \cref{eq:marginal-posterior-dags} intractable in general. That being said, having access to the posterior distribution gives us a more complete view of the problem: for example, maximizing the Bayesian score in \cref{eq:bayesian-score} reduces to \emph{maximum a posteriori} (MAP) estimation on \cref{eq:marginal-posterior-dags}.

In most cases though, sampling DAGs from this posterior distribution will be sufficient, rather than evaluating the posterior probability of some graph. This means that we have a distribution over discrete and compositional objects (directed acyclic graphs), known up to an intractable normalization constant ($P(\gD)$), that we want to sample from; this is a perfect case of application for GFlowNets, as we will see in this chapter. But before this, we will review some existing methods in the following sections based either on MCMC or variational inference.

\paragraph{Exact sampling} Even if the task seems daunting, sampling exactly from \cref{eq:marginal-posterior-dags} is sometimes possible in very specific cases using \emph{dynamic programming}. If the marginal likelihood has a certain structure, we can compute the marginal posterior probabilities of certain edges being present in time $O(d2^{d})$ \citep{koivisto2004exactbayesiansl,koivisto2006exact}, a significant improvement over the super-exponential number of DAGs, albeit still being exponential. Relaxing the structure to any modular marginal likelihood, \citet{tian2009exactposterior} showed that these edge marginals can be computed in $O(d3^{d})$. This in turns allows for exact sampling from the posterior distribution at relatively minimal cost \citep{he2016structurelearningexact,talvitie2020exactsampling}. Recently, \citet{harviainen2024fasterexactsampling} showed that the cost of exactly sampling from the posterior can be brought down to $O(2.829^{d})$ by bypassing completely the initial computation of the marginals. Dynamic programming can also be used in conjunction with MCMC to provide an informative proposal distribution allowing for larger moves in the space of DAGs \citep{eaton2007bayesian}.

\subsection{Structure Markov chain Monte Carlo}
\label{sec:structure-mcmc}
\begin{figure}[t]
    \centering
    \begin{adjustbox}{center}
    \includegraphics[width=480pt]{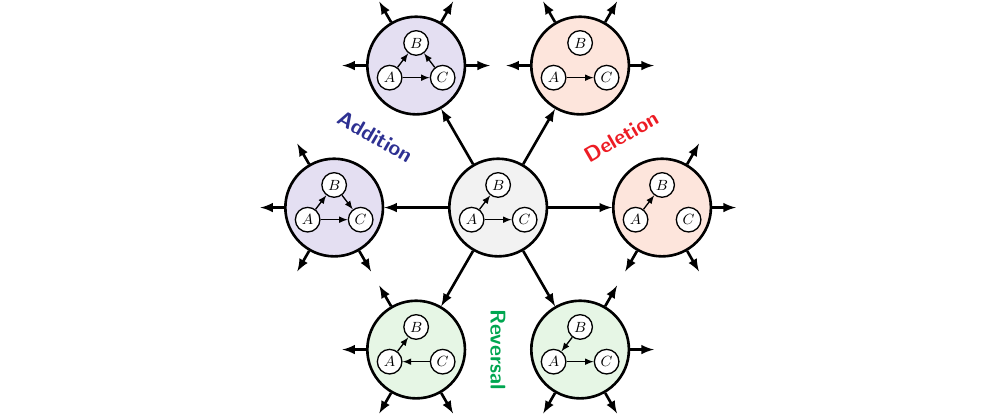}
    \end{adjustbox}
    \caption[Structure Markov chain Monte Carlo]{Illustration of the moves made by a proposal distribution in structure Markov chain Monte Carlo, where starting from a certain DAG, we can either add, subtract \citep{madigan1995structuremcmc}, or reverse \citep{grzegorczyk2008improving} a single edge.}
    \label{fig:structure-mcmc}
\end{figure}

\index{Markov chain Monte Carlo}
Markov chain Monte Carlo (MCMC) methods are in general natural choices for Bayesian inference, and it is no different for structure learning. The \emph{Structure MCMC} algorithm (MC\textsuperscript{3}; \citealp{madigan1995structuremcmc}) was one of the very first MCMC methods (and one of the first Bayesian structure learning methods in general) designed over the space of DAGs. It operates with a Metropolis-Hastings scheme (\cref{sec:existing-approaches-sampling-ebm}), where the proposal distribution applies mutations to the current graph by either adding or removing a single directed edge. While deleting an edge is always a legal move (in the sense that it also yields a valid DAG), adding an edge on the other hand requires a careful verification to ensure that it wouldn't introduce a cycle (\citealp{giudici2003improvingmcmc}; see also \cref{sec:efficient-verification-valid-actions}). This simple set of moves alone leads to very slow mixing though, consequently slowing down the convergence of the Markov chain to the posterior distribution. This set of moves was later enhanced with the reversal of a single edge \citep{grzegorczyk2008improving}, to encourage mixing between Markov equivalence classes \citep{giudici2003improvingmcmc}. \cref{fig:structure-mcmc} illustrates the legal moves allowed by these MCMC methods.

Instead of working in the space of DAGs directly, \citet{friedman2003ordermcmc} introduced \emph{Order MCMC} where the Markov chain moves in the space of node orders, further improving the mixing time but at the cost of introducing some bias \citep{ellis2008bias}. This prompted a number of derivatives operating on spaces different from the space of DAGs, such as the space of partial orders \citep{niinimaki2016partialordermcmc} or the space of node partitions \citep{kuipers2017partitionmcmc}. An alternative to improve convergence is to change the strategy to accept new moves, switching the standard Metropolis-Hastings update for Gibbs sampling \citep{mansinghka2006gibbs} or a birth-death process \citep{jennings2018birthdeathprocess}, or to change the set of legal moves \citep{castelo2003inclusiondrivenmcmc,kuipers2021efficient}. Recently, \citet{viinikka2020gadget} incorporated many of these advances into an efficient MCMC algorithm operating over partial orders called \emph{Gadget}.

\subsection{Variational inference}
\label{sec:variational-inference-bayesian-structure-learning}
Approximating the posterior distribution $P(G\mid \gD)$ with variational inference as we presented it in \cref{sec:variational-inference} faces a number of challenges. The first one is that variational inference is better adapted to problems with continuous variables, where we can leverage the reparametrization trick of \cref{sec:parameters-learning}; nevertheless, some works have successfully applied it to discrete problems \citep{maddison2017concretedistribution,jang2017gumbelsoftmax}. The second and major challenge though is that the acyclicity constraint is extremely difficult to encode in a variational framework. In other words, it is relatively easy to generate directed graphs with a variational distribution, but far more complex to guarantee that the support of this distribution is limited to directed \emph{acyclic} graphs. In this section, we will present two models representative of how the community has been approaching the problem of Bayesian structure learning from the perspective of variational inference. We will provide a more exhaustive overview in \cref{sec:joint-posterior-variational-inference}.

\paragraph{Continuous relaxation of the prior distribution} Recall that the objective of variational inference is to find a distribution $Q_{\phi}(G)$ that best approximates the intractable distribution (here $P(G\mid \gD)$) from a family of tractable distributions. \citet{lorch2021dibs} chose to model this as a latent variable model, with a latent variable $Z = (U, V)$, where $U, V \in \sR^{d\times k}$. The inference model is
\begin{align}
    Q(G\mid Z) &= \prod_{i\neq j}Q(G_{ij}\mid \vu_{i}, \vv_{j}) && \textrm{where} & Q(G_{ij} = 1\mid \vu_{i}, \vv_{j}) = \sigma\big(\vu_{i}^{\top}\vv_{j}\big),
\end{align}
and where $\sigma$ is the sigmoid function. This is a distribution over \emph{directed} graphs, since a priori $\sigma\big(\vu_{i}^{\top}\vv_{j}\big) \neq \sigma\big(\vu_{j}^{\top}\vv_{i}\big)$, but not over \emph{acyclic} graphs yet. To encode acyclicity, \citet{lorch2021dibs} took inspiration from the continuous relaxations mentioned in \cref{sec:structure-learning-continuous-relaxations}. More specifically, recall that there exist functions $h$ over directed graphs such that $h(G) = 0$ iff.~$G$ is a DAG; an example of such a function is $h(G) = \Tr\big(\exp(G)\big) - d$ \citep{zheng2018notears}. The acyclicity is then encouraged thanks to a prior over the latent variables define by the Gibbs distribution
\begin{equation}
    Q(Z) \propto \exp\big(\!-\!\lambda \E_{Q(G\mid Z)}\big[h(G)\big]\big),
\end{equation}
with $\lambda$ being a hyperparameter controlling the ``strength'' of this prior. In particular, as $\lambda \rightarrow \infty$, the support of this prior becomes restricted to DAGs, and in turn this makes the support of the overall distribution $Q(G) = \int_{Z}Q(G\mid Z)Q(Z)dZ$ also limited to DAGs. But for finite values of $\lambda$, there is unfortunately no guarantee that this distribution will be over DAGs only. Note that while it may seem like $Q(G)$ is parameter-free, the posterior distribution is actually approximated with an empirical distribution based on particles $\{Z^{(m)}\}_{m=1}^{M}$, which effectively constitute the variational parameters. This particular model is called \emph{DiBS} \citep{lorch2021dibs}, but the idea of having a continuous relaxation of acyclicity in the prior has been used more broadly \citep{annadani2021vcn}.

\paragraph{Variable ordering} A key observation is that under a certain ordering of the variables (a topological ordering), the adjacency matrix of a DAG is guaranteed to be strictly upper-triangular. In other words, any (weighted) adjacency matrix of a DAG can be written as $G = P^{\top}UP$, where $P$ is a permutation matrix and $U$ is strictly upper triangular. We can approximate $P(G\mid \gD)$ with a mean field approximation of the form $Q_{\phi}(G) = Q_{\phi}(P)Q_{\phi}(U)$, with two appropriate distributions. On the one hand, $Q_{\phi}(U)$ is a distribution over weighted upper triangular matrices, which is easy to model (\eg with a collection of Normal distributions). On the other hand, we can make use of continuous relaxations such as the Gumbel-Sinkhorn distribution \citep{mena2018gumbelsinkhorn} in order to model the distribution $Q_{\phi}(P)$ over permutation matrices. This particular method is called \emph{BCD Nets} \citep{cundy2021bcdnets}, and has been applied specifically to linear-Gaussian models where $G$ is a weighted adjacency matrix. This idea of decomposition with a permutation matrix has also been applied to a broader class of Bayesian networks \citep{charpentier2022differentiabledag,annadani2023bayesdag}.

\section{GFlowNet over directed acyclic graphs}
\label{sec:gflownet-over-dags}\index{Directed acyclic graph!DAG-GFlowNet}
As an alternative to MCMC and variational inference, we saw in \cref{chap:generative-flow-networks} that generative flow networks were particularly adapted (1) when the target distribution was known only up to an intractable normalization constant, and (2) when the objects to be generated were discrete and had a compositional structure, which is the case of DAGs. In this section we will present the foundations of how to model any distribution over DAGs with a GFlowNet, and defer the discussion about the specific application to Bayesian structure learning (\ie the choice of reward function) to \cref{sec:dag-gfn}.

\subsection{Structure of the GFlowNet}
\label{sec:structure-dag-gfn}
\begin{figure}[t]
    \centering
    \begin{adjustbox}{center}
    \includegraphics[width=480pt]{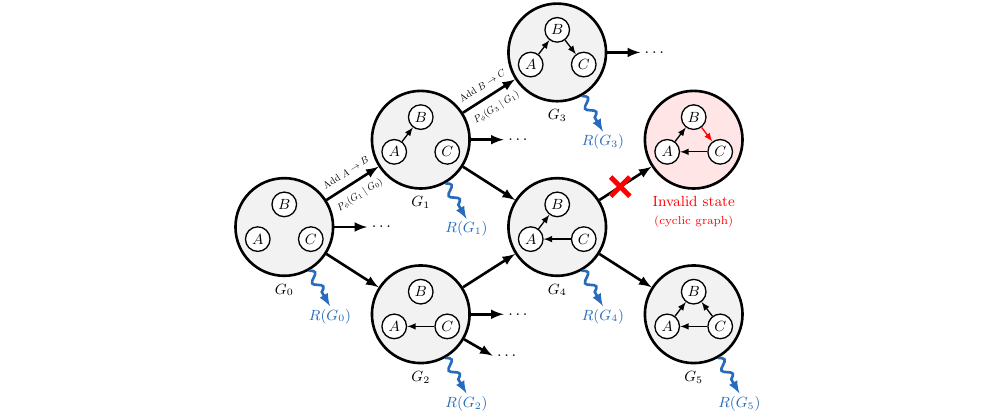}
    \end{adjustbox}
    \caption[Structure of a GFlowNet over DAGs]{Structure of a GFlowNet over DAGs. The states of the GFlowNet correspond to DAGs, with the initial state $G_{0}$ being the completely disconnected graph. Each state $G$ is terminating (\ie connected to the terminal state $\terminal$, represented by blue arrows for brevity) and associated with a reward $R(G)$. Transitioning from one state to another corresponds to adding an edge to the graph. The state in red is invalid since the graph includes a cycle.}
    \label{fig:dag-gflownet}
\end{figure}

\begin{figure}[hbtp]
    \centering
    \begin{adjustbox}{center}
    \includegraphics[width=520pt]{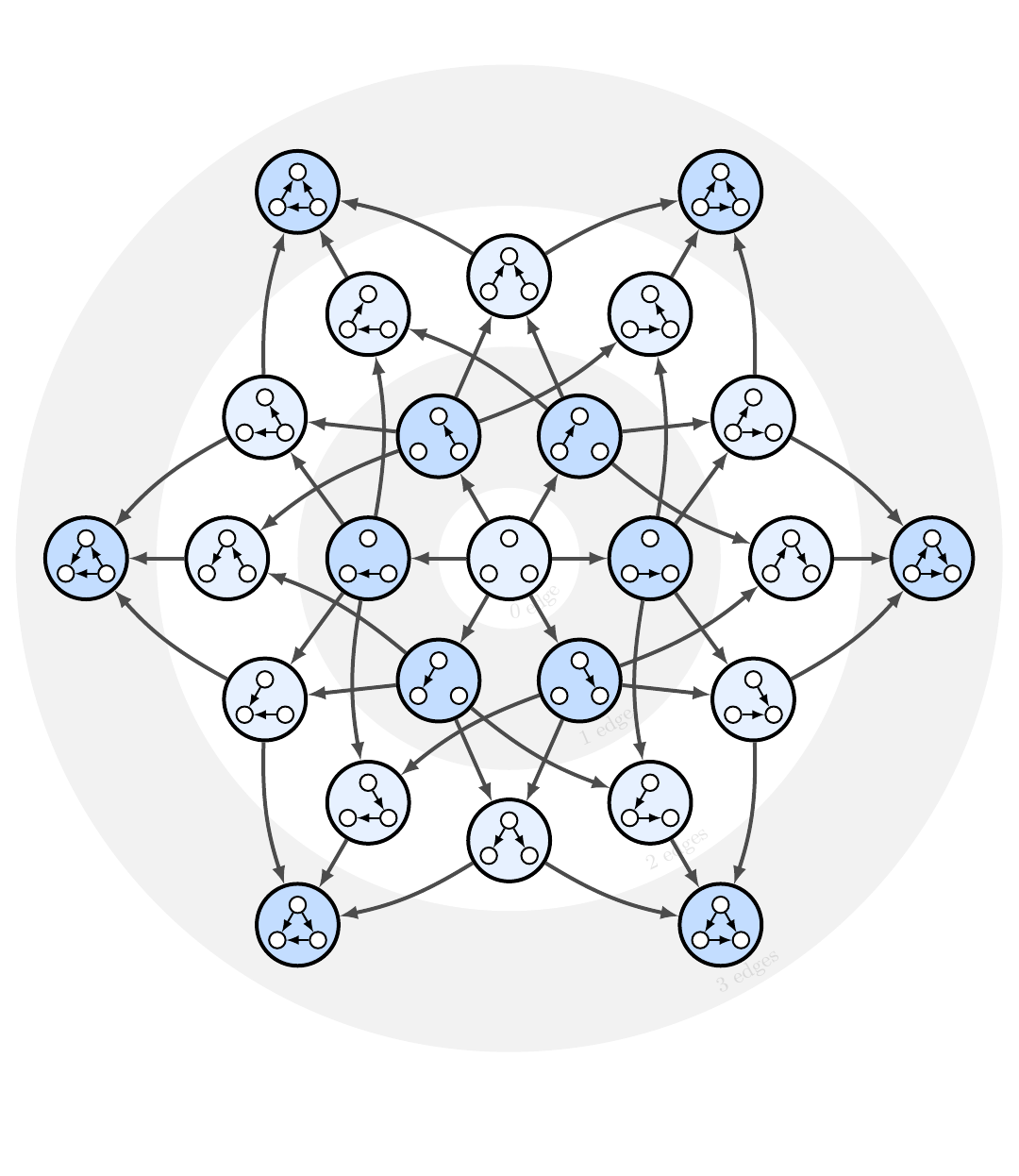}
    \end{adjustbox}
    \caption[Exhaustive structure of a GFlowNet over DAGs with $d=3$ nodes]{Exhaustive structure of a GFlowNet over DAGs with $d=3$ nodes. This pointed DAG contains $25$ states (one for each possible DAG) and $48$ edges. The initial state is the empty graph at the center of the figure.}
    \label{fig:structure-3nodes}
\end{figure}

We consider a GFlowNet $\gG$ where the states themselves are DAGs over $d$ (labeled) nodes; without loss of generality, we will assume that the vertices of all the DAGs are labeled with $\{1, \ldots, d\}$. Since the states of the GFlowNet are graphs, we will use the notation $G \in \gS$ to denote a state, in favor of $s$ as in \cref{chap:generative-flow-networks}; $\gS$ represents the space of all possible DAGs over $d$ nodes. A transition $G \rightarrow G' \in \gG$ in this GFlowNet corresponds to adding a single edge to $G$ in order to obtain the graph $G'$. In other words, the graphs are constructed one edge at a time, starting from the initial state $G_{0} \in \gS$ which is the fully disconnected graph over $d$ nodes. Since all the states of the GFlowNet are valid DAGs, they are all terminating states (\ie connected to the terminal state $\terminal$) with a corresponding reward $R(G)$; we will see in \cref{sec:modified-detailed-balance} that having $\gX \equiv \gS$ (to use the notations of \cref{sec:elements-graph-theory}) has consequences on the conditions one can apply here. \cref{fig:dag-gflownet} shows an illustration of the structure of such a GFlowNet, where the states are DAGs over $d=3$ nodes; see also \cref{fig:structure-3nodes} for the full structure of the GFlowNet over DAGs with $d=3$ nodes, with $|\gS| = 25$ states and $48$ transitions.

As mentioned in \cref{sec:flows-over-cyclic-graphs}, we take a \emph{constructive} approach with a GFlowNet, in the sense that we are only allowed to add edges, unlike the MCMC methods of \cref{sec:structure-mcmc} where the deletion and reversal of edges were also allowed. Only adding edges to the graphs guarantees that the structure of $\gG$ is itself a pointed DAG, since the number of edges in each state strictly increases as we transition. This application of GFlowNets to graphs also highlights the importance of the pointed DAG structure of the GFlowNet $\gG$ itself, as opposed to a tree structure as in \cref{sec:sampling-terminating-states-soft-mdp}, since there are multiple paths leading to the same state: for any graph $G$ with $K$ edges, there are $K!$ possible paths from the initial state $G_{0}$ leading to $G$, because the edges of $G$ may have been added in any order. This combinatorially large number of trajectories will constitute a challenge when \emph{evaluating} the terminating state probability of a certain state (\cref{def:terminating-state-probability}; not sampling from it though, which can always be done with \cref{alg:sampling-terminating-state-probability}); we will come back to this in \cref{sec:estimation-terminating-state-probability}.

\subsection{Efficient verification of valid actions}
\label{sec:efficient-verification-valid-actions}
To guarantee the integrity of the GFlowNet, we have to ensure that adding a new edge to some state $G$ also results in a valid DAG. This means that if we have a candidate edge we want to add, this edge (1) must not be already present in $G$, and (2) must not introduce a cycle. While the first condition is generally easy to verify, the second condition requires checking if the resulting graph contains any cycle; this can be done, for example, by finding a topological order of the vertices (if none can be found, then the action is invalid). Although each operation can be relatively inexpensive (typically linear in the size of $G$), the fact that this verification must be done for all $d(d-1)$ edges one may want to add makes this solution impractical to obtain all the valid actions from $G$, especially since we would need to do this verification frequently (\ie every time we transition in the GFlowNet).

Alternatively, it is also possible to filter out invalid actions using some \emph{binary mask} $\mM \in \{0, 1\}^{d \times d}$ that depends on the graph $G$, which can be updated efficiently since the graphs in the GFlowNet are constructed one edge at a time. \citet{giudici2003improvingmcmc} used a similar concept to efficiently obtain the legal moves their MCMC sampler may take. This mask depends not only on $G$ itself, but also on the transpose of its transitive closure.

\begin{definition}[Transpose of the transitive closure]
    The \emph{transpose of the transitive closure} of a DAG $G$, denoted by $\widebar{G}_{}^{\top}$, is the directed graph defined by
    \begin{equation}
        i \rightarrow j \in \widebar{G}_{}^{\top} \qquad \Leftrightarrow \qquad j \rightsquigarrow i \in G.
        \label{eq:transpose-transitive-closure}
    \end{equation}
    In other words, there is an edge $i \rightarrow j$ in $\widebar{G}_{}^{\top}$ if and only if there exists a path $j \rightsquigarrow i$ from $j$ to $i$ in $G$. We use the convention that $\widebar{G}_{}^{\top}[i, i] = 1$ for any node $i \in G$.
    \label{def:transpose-transitive-closure}
\end{definition}

In graph theory, the transitive closure of a DAG $G$ encodes in its edges the accessibility of any node from any other node through a path following the edges of $G$. Taking its transpose allows us to consider ``reverse paths'', which will be important in what follows; note that taking the transpose of a graph is exactly equivalent to transposing its adjacency matrix. It is easy to verify that, apart from the self-loops existing due the convention $\widebar{G}_{}^{\top}[i, i] = 1$, the transpose of the transitive closure of a DAG $G$ is also itself a DAG (\ie $\widebar{G}_{}^{\top}$ does not contain any cycle of length greater than 1). The following proposition shows how we can use both $G$ and $\widebar{G}_{}^{\top}$ in order to construct the binary mask $\mM$ to identify in constant time all the valid actions that can be taken from $G$.

\begin{proposition}[Mask]
    Let $G_{t}$ be a DAG and $\widebar{G}_{t}^{\top}$ the transpose of its transitive closure. Let $\mM_{t}$ be a $d \times d$ \emph{binary mask} associated with $G_{t}$, defined for any pair of nodes $(i, j)$ in $G_{t}$ as
    \begin{equation}
        \mM_{t}[i, j] \triangleq 1 - \big(G_{t}[i, j] \vee \widebar{G}_{t}^{\top}[i, j]\big).
        \label{eq:definition-mask-dag-gfn}
    \end{equation}
    Then adding an edge $u \rightarrow v$ to $G_{t}$ is a valid action for the GFlowNet over DAGs defined in \cref{sec:structure-dag-gfn} (in the sense that it is not already present in $G_{t}$, and it doesn't introduce a cycle) if and only if $\mM_{t}[u, v] = 1$.
    \label{prop:mask-valid-actions-dag-gfn}
\end{proposition}

\begin{proof}
    Adding an edge $u \rightarrow v$ to $G_{t}$ is a valid action in the GFlowNet over DAGs if and only if
    \begin{enumerate}
        \item the edge $u \rightarrow v$ is not already present in $G_{t}$, meaning that $G_{t}[u, v] = 0$;
        \item and the edge $u \rightarrow v$ does not introduce a cycle, meaning that there is no path $v \rightsquigarrow u$ from $v$ to $u$ in $G_{t}$ (otherwise $u \rightarrow v$ would close the cycle). In terms of the transpose of the transitive closure of $G_{t}$, this means that $\widebar{G}_{t}^{\top}[u, v] = 0$.
    \end{enumerate}
    Using \cref{eq:definition-mask-dag-gfn}, both conditions above can be summarized as $\mM_{t}[u, v] = 1$.
\end{proof}

Although this allows the verification of all $d(d-1)$ possible edges at once in constant time, constructing the transpose of the transitive closure $\widebar{G}_{}^{\top}$ still remains an operation equally as expensive as the naive solution described above (\eg the transitive closure can be constructed in $O(d^{3})$ using the Floyd-Warshall algorithm). Fortunately, we can leverage the fact that graphs in the GFlowNet of \cref{sec:structure-dag-gfn} are constructed one edge at a time to efficiently update the transpose of the transitive closure every time a new edge is added.

\begin{proposition}[Update of the mask components]
    Let $G_{t}$ be a DAG and $\widebar{G}_{t}^{\top}$ the transpose of its transitive closure. Suppose that $G_{t+1}$ is the result of adding the edge $u \rightarrow v$ to $G_{t}$. Then for any pair of nodes $(i, j)$ in the graph, we have
    \begin{align}
        \widebar{G}_{t+1}^{\top}[i, j] = \widebar{G}_{t}^{\top}[i, j] \vee \big(\widebar{G}_{t}^{\top}[u, j] \wedge \widebar{G}_{t}^{\top}[i, v]\big).
        \label{eq:update-mask-transpose-transitive-closure}
    \end{align}
    Moreover, we also have $G_{t+1}[i, j] = G_{t}[i, j] \vee \big(\mathds{1}(i=u)\mathds{1}(j=v)\big)$.
    \label{prop:update-mask-transpose-transitive-closure}
\end{proposition}

\begin{proof}
    The update of $G_{t+1}$ is immediate. By definition of $\widebar{G}_{t+1}^{\top}$, for some pair of nodes $(i, j)$, we have that $\widebar{G}_{t+1}^{\top}[i, j] = 1$ if and only if there exists a path $j \rightsquigarrow i$ from $j$ to $i$ in $G_{t+1}$. Since $G_{t+1}$ is the result of adding a single edge $u \rightarrow v$ to $G_{t}$, there are only two ways a path $j \rightsquigarrow i$ could exist in $G_{t+1}$:
    \begin{enumerate}
        \item either this path already existed in $G_{t}$, in which case we have $\widebar{G}_{t}^{\top}[i, j] = 1$;
        \item or this path goes through $u \rightarrow v$, meaning there exist two paths $j \rightsquigarrow u$ and $v \rightsquigarrow i$ in $G_{t}$. In that case, we have $\widebar{G}_{t}^{\top}[u, j] = 1$ and $\widebar{G}_{t}^{\top}[i, v] = 1$ respectively. This also covers the cases where either $j \equiv u$ or $i \equiv v$, since by convention $\widebar{G}_{t}^{\top}[u, u] = \widebar{G}_{t}^{\top}[v, v] = 1$.
    \end{enumerate}
    This concludes the proof, since \cref{eq:update-mask-transpose-transitive-closure} corresponds to rewriting the conditions above in logical form.
\end{proof}

\begin{figure}[t]
    \centering
    \begin{adjustbox}{center}
    \includegraphics[width=480pt]{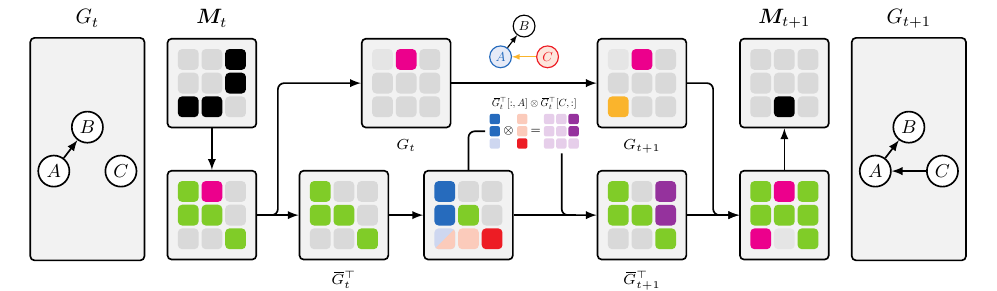}
    \end{adjustbox}
    \caption[Online update of the mask $\mM$]{Online update of the mask $\mM$. The mask $\mM_{t}$ associated with $G_{t}$ represents (in black) the edges that can be added to $G_{t}$ to obtain a valid DAG. $\mM_{t}$ is decomposed in two parts (\cref{prop:mask-valid-actions-dag-gfn}): (1) the adjacency matrix of $G_{t}$ (top) and (2) the transpose of its transitive closure $\widebar{G}^{\top}_{t}$ (bottom). To update the mask and obtain $\mM_{t+1}$ associated with $G_{t+1}$, the result of adding the edge $C \rightarrow A$ to $G_{t}$, each component must be updated separately (\cref{prop:update-mask-transpose-transitive-closure}), and then recombined.}
    \label{fig:dag-gfn-update-mask}
\end{figure}

The decomposition of the mask $\mM$ as well as the update of its two components are illustrated in \cref{fig:dag-gfn-update-mask}. The second term in \cref{eq:update-mask-transpose-transitive-closure} can be viewed in matrix form as a (binary) outer product $\widebar{G}_{t}^{\top}[:, v] \otimes \widebar{G}_{t}^{\top}[u, :]$. Overall, the update of the mask is dominated by the update of $\widebar{G}_{}^{\top}$, which is done in $O(d^{2})$; however, since these are all binary operations, this can be implemented very efficiently in practice. \cref{prop:update-mask-transpose-transitive-closure} suggests that we can keep track of $\widebar{G}_{t}^{\top}$ along with the graph $G_{t}$ itself, and update both every time we make a transition in the GFlowNet; the mask $\mM_{t}$ is then built ``lazily'' from both components any time it is necessary (\eg to filter out invalid actions in the forward transition probability distribution, see \cref{sec:parametrization-forward-transition-probabilities}). The transpose of the transitive closure is initialized as $\widebar{G}_{0}^{\top} = \mI_{d}$, due to the convention adopted in \cref{def:transpose-transitive-closure}. The pseudo-code for sampling from the GFlowNet over DAGs, and its interplay with the mask, is available in \cref{alg:dag-gflownet-interaction}.

\begin{algorithm}[t]
    \caption{Sampling a DAG from the GFlowNet defined in \cref{sec:gflownet-over-dags}.}
    \label{alg:dag-gflownet-interaction}
    \begin{algorithmic}[1]
        \Require A forward transition probability $P_{F}(\cdot\mid G; \mM)$ depending on a mask $\mM$; see \cref{sec:dag-gfn-neural-network-parametrization}.
        \Ensure A sample DAG $G \sim P_{F}^{\top}(G)$ from the terminating state distribution.
        \State Initialization of the state: $s_{0} = (G_{0}, \mI_{d})$, where $G_{0}$ is the empty DAG over $d$ nodes
        \Repeat
            \State Construct the mask $\mM_{t}$ based on $s_{t}$ \Comment{\cref{prop:mask-valid-actions-dag-gfn}}
            \State Sample the next graph: $G_{t+1} \sim P_{F}(\cdot\mid G_{t}; \mM_{t})$
            \If{$G_{t+1} \neq \terminal$}
            \State Update the state $s_{t+1} = (G_{t+1}, \widebar{G}_{t+1}^{\top})$ \Comment{\cref{prop:update-mask-transpose-transitive-closure}}
            \EndIf
        \Until{$G_{T+1} = \terminal$}
        \State \Return $G_{T}$
    \end{algorithmic}
\end{algorithm}

\subsection{Additional structural constraints}
\label{sec:additional-structural-constraints}
In addition to the acyclicity constraint, necessary for the structure of a Bayesian network, we can also encode various structural constraints about the graphs to be generated by the GFlowNet. For example, we may want to generate sparse structures in order to enable efficient inference in those Bayesian networks. This could be done by biasing the reward function $R(G)$ towards graphs that have the desired properties, by introducing a prior term that encourages some properties in a \emph{soft} way. Alternatively, we can also encode these constraints in the structure of $\gG$ itself, to have explicit guarantees about the support of the distribution over DAGs. For example, sparsity may be enforced by restricting the number of parents each node may have to be at most $d_{\max}$, which is a common assumption in structure learning \citep{koller2009pgm}. In that case, we can encode this by limiting the support of the distribution to be DAGs whose nodes have a maximum of $d_{\max}$ parents; this can be achieved by removing from $\gG$ any transition to a graph that would violate this condition. In practice, we can reuse the binary mask of \cref{prop:mask-valid-actions-dag-gfn} for that purpose, and filter out entire columns that have already exhausted their maximum number of parents:  %
\begin{equation}
    \widetilde{\mM}_{t}[i, j] = \mathds{1}\left(\sum_{k=1}^{d}G_{t}[k, j] < d_{\max}\right)\mM_{t}[i, j].
    \label{eq:binary-mask-maximum-parents}
\end{equation}

In the context of learning regulatory networks, further properties may also be enforced on the structure of the Bayesian network based on expert knowledge. For example, \citet{peer2006minreg} used prior biological studies to motivate having a limited number of \emph{regulator genes}, \ie nodes in the graph with an out-degree greater than zero, from a set of candidate regulators $\gC$. Similar to \cref{eq:binary-mask-maximum-parents} for limiting the number of parents of each node, these properties can be easily incorporated in the generation process of the GFlowNet by adapting the binary mask $\mM$ based on these structural constraints.

\subsection{Modified detailed balance condition}
\label{sec:modified-detailed-balance}

Although we could use any flow-matching condition from \cref{sec:flow-matching-condition-boundary-constraint,sec:alternative-conditions} to guarantee that the GFlowNet over DAGs induces a terminating state distribution $P_{F}^{\top}$ proportional to the reward, we can exploit the special structure of the GFlowNet to derive more adapted conditions. In particular, since all the states of the GFlowNet are terminating, we can modify the detailed balance condition of \cref{thm:detailed-balance-proportional-reward} to only depend on the transitions probabilities $P_{F}$ and $P_{B}$, and not explicitly on a state flow function anymore.

\begin{proposition}[Modified detailed balance]
    \label{prop:modified-detailed-balance}\index{Detailed balance!Modified detailed balance|textbf}
    Let $\gG$ be a GFlowNet over a state space $\gS$, with a reward function $R$, a forward transition probability $P_{F}$, and a backward transition probability $P_{B}$. If all the states of the GFlowNet are terminating (\ie $\gX \equiv \gS$), then for any transition $s \rightarrow s' \in \gG$ with $s'\neq \terminal$, the detailed balance condition in \cref{thm:detailed-balance-proportional-reward} (in conjunction with the boundary condition) can be written as
    \begin{equation}
        R(s')P_{B}(s\mid s')P_{F}(\terminal \mid s) = R(s)P_{F}(s'\mid s)P_{F}(\terminal \mid s').
        \label{eq:modified-detailed-balance}
    \end{equation}
\end{proposition}

\begin{proof}
    Recall from \cref{thm:detailed-balance-proportional-reward} that for any transition $s \rightarrow s' \in \gG$ such that $s' \neq \terminal$, the detailed balance condition is given by
    \begin{equation}
        F(s)P_{F}(s'\mid s) = F(s')P_{B}(s\mid s'),
        \label{eq:db-condition-proof-modified-db}
    \end{equation}
    where $F(s)$ is a state flow function, in addition to the boundary condition $F(s \rightarrow \terminal) = R(s)$. If all the states are terminating, then we can write the state flow $F(s)$ in terms of the reward function and the forward transition probability as
    \begin{equation}
        P_{F}(\terminal \mid s) = \frac{F(s \rightarrow \bot)}{\sum_{s'\in \children_{\gG}(s)}F(s \rightarrow s')} = \frac{R(s)}{F(s)} \qquad \Leftrightarrow \qquad F(s) = \frac{R(s)}{P_{F}(\terminal \mid s)}.
        \label{eq:modified-detailed-balance-proof-1}
    \end{equation}
    Replacing the state flow function $F$ in \cref{eq:db-condition-proof-modified-db}, we get the expected condition \cref{eq:modified-detailed-balance}.
\end{proof}

Note that \cref{prop:modified-detailed-balance} is expressed using the general notation ``$s$'' for states, since this result can be generally applied to any GFlowNet where $\gX \equiv \gS$ (the GFlowNet defined in \cref{sec:structure-dag-gfn} being one example). This modified version of the detailed balance condition offers a better credit assignment, while remaining local at the level of individual transitions $s \rightarrow s'$, by directly making use of the reward signal coming from both $R(s)$ and $R(s')$. It can be interpreted as an instance of forward-looking GFlowNet (FL-GFN; \citealp{pan2023flgfn}), which is a class of GFlowNets we mentioned in \cref{sec:equivalence-sql-fldb} when establishing equivalences with Soft Q-Learning, where local credit is obtained via a decomposition of the reward function; we will come back to this decomposition in the context of structure learning of Bayesian networks in \cref{sec:dag-gfn-modularity-computational-efficiency}. Moreover, because this is only a reformulation of the detailed balance condition, we can rely on \cref{thm:detailed-balance-proportional-reward} and guarantee that if \cref{eq:modified-detailed-balance} is satisfied for all transitions in the GFlowNet, then the terminating state probability distribution associated with $P_{F}$ is proportional to $R(s)$.

\begin{corollary}
    \label{cor:modified-db-proportional-reward}
    Suppose that $\gX \equiv \gS$. If the modified detailed balance condition of \cref{prop:modified-detailed-balance} is satisfied for all the transitions $s\rightarrow s'\in \gG$ s.t.~$s'\neq \terminal$, then the terminating state distribution associated with the corresponding $P_{F}$ is proportional to $R$: $P_{F}^{\top}(s) \propto R(s)$.
\end{corollary}

\subsection{Fixed backward transition probability}
\label{sec:fixed-backward-transition-probability}\index{Transition probability!Backward transition probabilities}
Because the system of equations in \cref{eq:modified-detailed-balance} may admit many solutions (\ie forward and backward transition probabilities), we can set the backward transition probability $P_{B}$ to some fixed and arbitrary distribution. Throughout this chapter, we will set $P_{B}$ as the uniform distribution over the parent states: for a transition $G \rightarrow G'$ in the GFlowNet defined in \cref{sec:structure-dag-gfn}, where $G'$ is a DAG with $K$ edges, we have
\begin{equation}
    P_{B}(G\mid G') = \frac{1}{K},
    \label{eq:backward-transition-probability-uniform}
\end{equation}
since $G'$ admits exactly $K$ parents, each one being the result of removing a single edge from $G'$. Therefore, the only unknown quantity in \cref{eq:modified-detailed-balance} becomes the forward transition probability distribution $P_{F}$, whose parametrization will depend on the binary mask $\mM$ introduced in \cref{prop:mask-valid-actions-dag-gfn}; see \cref{sec:parametrization-forward-transition-probabilities} for details. The following proposition shows that once the backward transition probability is fixed, then the modified detailed balance condition admits a unique solution $P_{F}$.

\begin{proposition}
    Let $\gG = (\widebar{\gS}, \gA)$ be the GFlowNet over DAGs defined in \cref{sec:structure-dag-gfn} with a reward function $R$. If the backward transition probability $P_{B}$ is a fixed distribution over parent states (\eg as in \cref{eq:backward-transition-probability-uniform}), then there exists a unique (positive) forward transition probability distribution $P_{F}$ that satisfies the modified detailed balance condition in \cref{prop:modified-detailed-balance} for all $G\rightarrow G' \in \gG$ such that $G' \neq \terminal$.
    \label{prop:unique-PF-modified-detailed-balance}
\end{proposition}

\begin{proof}
    The proof is based on a reparametrization of the (non-linear) system of equations in $P_{F}$ in \cref{eq:modified-detailed-balance} with linear constraints (coming from the fact that $\sum_{G'\in\children_{\gG}(G)}P_{F}(G'\mid G) = 1$) into an unconstrained system of linear equations; \cref{eq:modified-detailed-balance} is non-linear in $P_{F}$ due to the product $P_{F}(s'\mid s)P_{F}(\terminal\mid s')$ appearing on the RHS. For any transition $G\rightarrow G'$ such that $G'\neq \terminal$, we define
    \begin{equation}
        x(G'\mid G) \triangleq \frac{P_{F}(G'\mid G)}{P_{F}(\terminal \mid G)}.
        \label{eq:proof-reparametrization-linear-system}
    \end{equation}
    This is well defined since we are looking for a positive solution, and therefore $P_{F}(\terminal \mid G) > 0$. We can note that since $P_{F}$ is a properly normalized probability distribution, we have
    \begin{equation}
        \sum_{G'\in \underline{\children}_{\gG}(G)}x(G'\mid G) = \frac{1}{P_{F}(\terminal\mid G)}\sum_{G'\in\underline{\children}_{\gG}(G)}P_{F}(G'\mid G) = \frac{1 - P_{F}(\terminal\mid G)}{P_{F}(\terminal\mid G)} = \frac{1}{P_{F}(\terminal\mid G)} - 1,
        \label{eq:proof-unique-solution-normalization}
    \end{equation}
    where we used the notation $\underline{\children}_{\gG}(G) = \children_{\gG}(G) \backslash \{\terminal\}$ to denote the children of $G$, except the terminal state $\terminal$ (recall that all the states of the GFlowNet are terminating, and therefore $\terminal$ is a child of any state $G$). Using $x(G'\mid G)$ and \cref{eq:proof-unique-solution-normalization}, we can rewrite the modified detailed balance equation in \cref{eq:modified-detailed-balance} for any transition $G\rightarrow G'$ such that $G'\neq \terminal$ as
    \begin{align}
        & \frac{R(G')P_{B}(G\mid G')}{P_{F}(\terminal\mid G')} = R(G)\frac{P_{F}(G'\mid G)}{P_{F}(\terminal \mid G)}\\
        \Leftrightarrow\qquad & R(G)x(G'\mid G) - R(G')P_{B}(G\mid G')\sum_{G''\in \underline{\children}_{\gG}(G')}x(G''\mid G') = R(G')P_{B}(G\mid G').\label{eq:proof-unique-solution-linear-system}
    \end{align}
    This is a system of linear equations in $x(G'\mid G)$, with $p \triangleq |\gA| - |\gS|$ equations (one for each transition $G\rightarrow G'$ such that $G'\neq \terminal$), and as many unknowns. If we organize the unknowns into a vector $\vx \in \sR^{p}$ according to a topological order of the DAG $\gG$ (in the sense that $x(G_{1}'\mid G_{1}) \prec x(G_{2}'\mid G_{2})$ in $\vx$ if and only if $G_{1}$ is an ancestor of $G_{2}$ in $\gG$), we can rewrite \cref{eq:proof-unique-solution-linear-system} as $\mA\vx = \vb$ (with $\vx > 0$), where $\mA$ is an appropriately defined $p \times p$ matrix. In fact, due to the topological order used in $\vx$, the matrix $\mA$ is upper-triangular, its diagonal elements are the rewards $R(G) > 0$ and its off-diagonal elements are $-R(G')P_{B}(G\mid G')$. Therefore the matrix $\mA$ is invertible, and the system $\mA\vx = \vb$ admits a unique solution. Using variable elimination starting at the ``leaves'' of $\gG$ (\ie states with no children other than $\terminal$) and moving all the way to the initial state $G_{0}$ in reverse topological order, we can see that the solution $\vx$ is positive---the vector $\vb$ itself being positive as well.

    Since there is a one-to-one mapping between $x(G'\mid G)$ and the forward transition probability via the transformation
    \begin{align}
        P_{F}(\terminal\mid G) &= \left[\sum_{G'\in\underline{\children}_{\gG}(G)}x(G'\mid G) + 1\right]^{-1} && P_{F}(G'\mid G) = P_{F}(\terminal\mid G)x(G'\mid G),
    \end{align}
    based on \cref{eq:proof-unique-solution-normalization}, we can conclude that the modified detailed balance condition admits a unique positive forward transition probability $P_{F}$ when $P_{B}$ is a fixed distribution over parent states.
\end{proof}

The proof of \cref{prop:unique-PF-modified-detailed-balance} does not rely on the exact form of the backward transition probability (only on the fact that it is fixed), and we could choose any distribution over the parent states for $P_{B}$. In some cases, it has even been shown empirically that learning $P_{B}$ along with $P_{F}$, thus leading to an over-parametrization of the problem, may be advantageous \citep{malkin2022trajectorybalance}. Our choice of \cref{eq:backward-transition-probability-uniform} is motivated by the fact that we don't want to encourage specific trajectories $\tau$ while generating a DAG $G$ one edge at a time. This can be interpreted as finding the corresponding $P_{F}$ with maximum conditional entropy.

\begin{proposition}
    \label{prop:uniform-maximizes-conditional-entropy-trajectories}
    Let $\gG = (\widebar{\gS}, \gA)$ be the GFlowNet over DAGs defined in \cref{sec:structure-dag-gfn} with a reward function $R$. The backward transition probability distribution $P_{B}$ defined in \cref{eq:backward-transition-probability-uniform} (\ie uniform over the parent states) maximizes the conditional entropy over trajectories
    \begin{equation}
        \gH(\tau\mid G) = -\sum_{G\in\gS}\sum_{\tau: G_{0}\rightsquigarrow G}P_{F}(\tau)\log \frac{ZP_{F}(\tau)}{R(G)},
        \label{eq:conditional-entropy-trajectories}
    \end{equation}
    where $Z = \sum_{G\in\gS}R(G)$ is the partition function, and $P_{F}$ is the (unique) forward transition probability distribution associated with $P_{B}$ satisfying the modified detailed balance condition of \cref{prop:modified-detailed-balance}.
\end{proposition}

\begin{proof}
    Let $\tau_{G}$ be a trajectory $G_{0} \rightsquigarrow G$ from the initial state $G_{0}$ to some DAG $G$ with $K$ edges; we use the subscript ``$G$'' to emphasize the dependency of the trajectory on the end state. Let us first recall the definition of the conditional entropy over trajectories
    \begin{equation}
        \gH(\tau \mid G) \triangleq -\sum_{G\in \gS}\sum_{\tau_{G}:G_{0}\rightsquigarrow G}P_{F}(\tau_{G})\log \frac{P_{F}(\tau_{G})}{P_{F}^{\top}(G)},
        \label{eq:proof-conditional-entropy-max-definition}
    \end{equation}
    where $P_{F}^{\top}(G)$ is the terminating state probability distribution, which is the marginal of $P_{F}(\tau) = \prod_{t=0}^{T}P_{F}(G_{t+1}\mid G_{t})$ for $\tau = (G_{0}, G_{1}, \ldots, G, \terminal)$ with the conventions $G_{T} = G$ and $G_{T+1} = \terminal$ (see also \cref{sec:forward-transition-probabilities}). Note that $\tau$ and $G$ in the notation ``$\gH(\tau\mid G)$'' represent random variables, and not specific assignments. Since we consider $P_{F}$ and $P_{B}$ such that they satisfy the modified detailed balance condition of \cref{prop:modified-detailed-balance}, we know that $P_{F}^{\top}(G) = R(G)/Z$ and that the following trajectory balance condition is satisfied: $ZP_{F}(\tau_{G}) = R(G)P_{B}(\tau_{G}\mid G)$ (\cref{thm:trajectory-balance-proportional-reward}). Therefore,
    \begin{align}
        \gH(\tau\mid G) &= -\sum_{G\in\gS}\sum_{\tau_{G}:G_{0}\rightsquigarrow G}P_{F}(\tau_{G})\log \frac{ZP_{F}(\tau_{G})}{R(G)}\\
        &= -\sum_{G\in\gS}\frac{R(G)}{Z}\sum_{\tau_{G}: G_{0}\rightsquigarrow G}P_{B}(\tau_{G}\mid G)\log P_{B}(\tau_{G}\mid G)
        \label{eq:proof-maximization-problem-condtional-entropy}
    \end{align}
    As a function of $P_{B}(\tau_{G}\mid G)$, it is well-known that the distribution maximizing \cref{eq:proof-maximization-problem-condtional-entropy} is the uniform distribution (over the trajectories leading to $G$):
    \begin{equation}
        P_{B}(\tau_{G}\mid G) = \frac{1}{K!},
        \label{eq:proof-uniform-distribution-trajectories}
    \end{equation}
    where $K!$ is the total number of trajectories leading to $G$ (see \cref{sec:structure-dag-gfn}). Let $G'$ be the result of adding one edge to $G$, and $\tau_{G'}$ being a trajectory from $G_{0}$ to $G'$ that can be decomposed into a trajectory $\tau_{G}$ from $G_{0}$ to $G$, followed by the transition $G\rightarrow G'$. By definition of the backward transition probability of a trajectory and using \cref{eq:proof-uniform-distribution-trajectories}, we get
    \begin{equation}
        P_{B}(\tau_{G'} \mid G') = P_{B}(G\mid G')P_{B}(\tau_{G}\mid G) \qquad \Rightarrow \qquad P_{B}(G\mid G') = \frac{1}{K+1},
    \end{equation}
    which corresponds to $P_{B}$ being the uniform distribution over parent states \cref{eq:backward-transition-probability-uniform}, since $G'$ has $K+1$ edges.
\end{proof}
\citet{zhang2022ebgfn} also showed that, in certain cases further identified by \citet{mohammadpour2024maxentgfn}, the uniform backward transition probability $P_{B}$ maximizes the expected total entropy of the unique corresponding forward transition probability distribution along complete trajectories, which they defined as a measure of entropy of the overall Markovian flow.

\section{Parametrization of the forward transition probabilities}
\label{sec:parametrization-forward-transition-probabilities}
In practice, the forward transition probability distribution will be parametrized by a function, whose parameters will be learned through interactions with the GFlowNet. In this section, we will denote by $\phi$ the parameters of this function (\eg they may be parameters of a neural network), and use the notation $P_{\phi}$ to denote the learned components of the forward transition probabilities, instead of $P_{F}^{\phi}$, for clarity and to emphasize their parametrization in $\phi$. Recall from the previous section that the backward transition probabilities $P_{B}$ are set to the uniform distribution over parent states \cref{eq:backward-transition-probability-uniform}, and only $P_{\phi}$ is learned.

\subsection{Learning objective}
\label{sec:dag-gfn-learning-objective}
Similar to how any flow matching condition could be transformed into a learning objective in \cref{sec:flow-matching-losses}, we can also turn the modified detailed balance condition of \cref{prop:modified-detailed-balance} into a loss function that can be minimized in order to fit the parameters of the neural networks $\phi$ parametrizing the forward transition probabilities, by treating it as a non-linear least square problem of the form $\gL(\phi) = \frac{1}{2}\E_{\pi_{b}}[\Delta^{2}(G\rightarrow G'; \phi)]$, where the residual is defined as
\begin{equation}
    \Delta(G\rightarrow G';\phi) = \log \frac{R(G')P_{B}(G\mid G')P_{\phi}(\terminal \mid G)}{R(G)P_{\phi}(G'\mid G)P_{\phi}(\terminal \mid G')},
    \label{eq:dag-gfn-learning-objective}\index{Detailed balance!Modified detailed balance}
\end{equation}
where $\pi_{b}$ is an arbitrary distribution over transitions $G\rightarrow G' \in \gG$ (such that $G'\neq \terminal$) with full support. This objective was also introduced in \cref{sec:equivalence-pisql-mdb}, where we showed its equivalence with a policy-parametrized version of the Soft Q-Learning algorithm \citep{haarnoja2017sql}. In practice, we often use the Huber loss \citep{huber1992huberloss} instead of the squared loss in $\gL(\phi)$ for stability, following the recommendations from the deep reinforcement learning literature \citep{mnih2015dqn}.

\subsection{Improved exploration with off-policy learning}
\label{sec:dag-gfn-tools-reinforcement-learning}
As we showed in \cref{sec:off-policy-training}, training a GFlowNet may be done completely \emph{off-policy}, meaning that the (behavior) policy\index{Behavior policy} $\pi_{b}$ appearing in the loss may be arbitrary as long as it has full-support (\ie there is no transition s.t.~$\pi_{b}(G\rightarrow G') = 0$). This is essential to improve exploration in the vast space of DAGs. We use the simple \emph{$\varepsilon$-sampling} strategy described in \cref{sec:off-policy-training} to collect trajectories through interactions with $\gG$, which consists in using the current policy $P_{\phi}$ most of the time and sampling uniformly at random with probability $\varepsilon$. At the beginning of training, we set $\varepsilon = 1$ in order to enable a fully exploratory behavior, and then we gradually decrease it over the course of training to a baseline value $\varepsilon_{\min} > 0$ to use more often the current $P_{\phi}$ that would become more reliable (while still leaving an option to explore). This is a strategy inspired from the literature in reinforcement learning, and in particular the pioneering work on Deep Q-Networks (\gls{dqn}; \citealp{mnih2015dqn}); this is not surprising since we saw in \cref{chap:gflownet-maxent-rl} that GFlowNets share close ties with (maximum entropy) reinforcement learning.

These collected transitions $G \rightarrow G'$ are not directly used to evaluate the loss though, as is standard for the trajectory balance loss in \cref{alg:training-gflownet-tb-loss}, but they are fed into a \emph{replay buffer} \citep{lin1992experiencereplay,mnih2015dqn,schaul2016prioritizedexperiencereplay}. This type of buffer stores (up to a certain capacity) all the experience acquired through the interaction with $\gG$ throughout training, which is then ``replayed'' in the sense that a batch of transitions is sampled from it in order to compute a Monte Carlo estimate of the modified detailed balance loss. This adds yet another layer of ``off-policiness'' to our behavior policy $\pi_{b}$, which is effectively the combination of $\varepsilon$-sampling (to acquire data) \& sampling from the replay buffer (to create a batch to update $\phi$). Our work applying GFlowNet to Bayesian structure learning spearheaded the use of replay buffers in the GFlowNet literature \citep{deleu2022daggflownet}, which has now gained popularity in this community \citep{vemgal2023gfnreplaybuffer,zhang2023graphcogfn}.

\subsection{Hierarchical representation of the transition probabilities}
\label{sec:hierarchical-representation-dag-gfn}\index{Transition probability!Forward transition probabilities}
At every step of the generation of a DAG with the GFlowNet described in \cref{sec:gflownet-over-dags}, we have to make a decision: either (1) we continue adding an edge to the current graph $G$ and transition to a new graph $G'$ with probability $P_{\phi}(G'\mid G)$, or (2) we decide to stop adding edges with probability $P_{\phi}(\terminal \mid G)$ and return $G$ as a sample of our distribution. Although we could treat each decision equally by defining a \emph{flat} distribution $P_{\phi}(\cdot \mid G)$, the description above suggests a \emph{hierarchical} definition of the forward transition probabilities, where we first decide whether we want to continue adding an edge or not with probability $P_{\phi}(\stopaction \mid G)$, and upon choosing to continue, we can select which edge to add with probability $P_{\phi}(G'\mid G, \neg\stopaction)$.

\begin{definition}[Hierarchical representation]
    \label{def:hierarchical-representation-dag-gfn}
    Let $G \in \gG$ be a state of the GFlowNet. The forward transition probability distribution $P_{\phi}$ has a \emph{hierarchical representation} if
    \begin{align}
        P_{\phi}(\terminal \mid G) &= P_{\phi}(\stopaction\mid G)\label{eq:hierarchical-representation-dag-gfn-1}\\
        P_{\phi}(G'\mid G) &= \big(1 - P_{\phi}(\stopaction\mid G)\big)P_{\phi}(G'\mid G, \neg\stopaction),\label{eq:hierarchical-representation-dag-gfn-2}
    \end{align}
    where $P_{\phi}(\stopaction\mid G)$ parametrizes the probability to stop adding new edges to $G$, and $P_{\phi}(G'\mid G, \neg\stopaction)$ parametrizes a distribution over the next non-terminal states $G'$.
\end{definition}

Because the GFlowNet is structured in such a way that a transition is the result of adding a single edge to $G$, we can model $P_{\phi}(G'\mid G, \neg\stopaction)$ not as being a distribution over the next states, but as a probability distribution over the $d(d-1)$ possible \emph{edges} one could add to $G$. We can use the binary mask $\mM$ from \cref{prop:mask-valid-actions-dag-gfn} in order to filter out actions (edges) that would not lead to a valid DAG $G'$ (\eg if this edge were to introduce a cycle), and set $P_{\phi}(G'\mid G, \neg\stopaction) = 0$ for any invalid action, as well as normalize it accordingly.

Moreover, if there is no valid action from $G$ other than terminating (\eg if $G$ already contains $d(d-1)/2$ edges, which is the maximum number of edges a DAG over $d$ nodes can have), we can force termination by setting $P_{\phi}(\stopaction \mid G) = 1$. This last constraint is enforced as a post-processing step on top of the (raw) values returned by the neural networks parametrizing $P_{\phi}(\stopaction\mid G)$, based on whether the mask $\mM \equiv \vzero$ is identically zero.  %

\subsection{Neural network parametrization}
\label{sec:dag-gfn-neural-network-parametrization}
Given that we want to represent the forward transition probability distribution in a hierarchical way, we need to parametrize both $P_{\phi}(\stopaction\mid G)$ and $P_{\phi}(G'\mid G, \neg\stopaction)$. We choose to parametrize these two components using neural networks taking the current state $G$ as an input, to benefit from their capacity to generalize to states that were not encountered during training. In practice, instead of defining two distinct neural networks, we use a single neural network with a common backbone encoding $G$ and two separate heads to take advantage of parameter sharing. We experimented with two families of neural network architectures, depending on how the graph $G$ is represented: (linear) Transformers \citep{katharopoulos2020lineartransformers}, and Graph Neural Networks \citep{battaglia2018graphnetwork}. There will be no comparative analysis of both architectures in \cref{sec:dag-gfn-comparison-exact-posterior} and onwards though, since they correspond to two different generations of network architectures in our experiments, the latter having superseded the former for its simplicity and (unreported) equal performance on a non-exhaustive set of cases, and will be used in \cref{chap:jsp-gfn}.

\begin{figure}[t]
    \centering
    \begin{adjustbox}{center}
    \includegraphics[width=520pt]{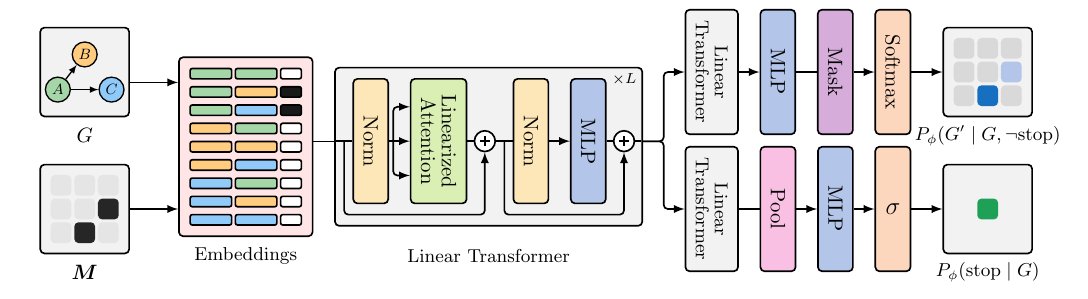}
    \end{adjustbox}
    \caption[Linear Transformer architecture for the forward transition probabilities $P_{\phi}(\cdot \mid G)$]{Linear Transformer architecture for the forward transition probabilities $P_{\phi}(\cdot \mid G)$. The input graph $G$ is encoded as a set of $d^{2}$ possible edges (including self-loops); each directed edge is embedded using the embeddings of its source and target, with an additional vector indicating whether the edge is present in $G$. The embeddings are fed into a Linear Transformer \citep{katharopoulos2020lineartransformers}, with two separate output heads. The first head (above) gives the probability to add an edge $P_{\phi}(i\rightarrow j\mid G, \neg\stopaction)$, using the mask $\mM$ to filter out invalid actions; here, the only valid actions are either adding $B \rightarrow C$ or $C\rightarrow B$. The second head (below) gives the probability to terminate the trajectory $P_{\phi}(\stopaction\mid G)$.}
    \label{fig:dag-gfn-linear-transformer}
\end{figure}

\paragraph{Linear Transformer} In \citep{deleu2022daggflownet,nishikawa2023vbg}, we represented a graph as a collection of $d^{2}$ edges, each associated with a binary flag indicating whether the edge was present in $G$ (see \cref{fig:dag-gfn-linear-transformer}). Our choice of neural network architecture was then motivated by two factors: we wanted an architecture (1) that is \emph{permutation equivariant}, meaning that any permutation of the edges of $G$ in the input would be reflected in the output of the neural network for $P_{\phi}(G'\mid G, \neg\stopaction)$ (we saw in \cref{sec:hierarchical-representation-dag-gfn} that this can be interpreted as a distribution over edges to add), and (2) whose parameters $\phi$ do not scale \emph{too much} with the number of nodes $d$. A natural option fitting these requirements would be a Transformer architecture \citep{vaswani2017transformers}; however, because the input of the neural network is of size $d^{2}$ already (\ie one entry for each possible pair of nodes), the self-attention layers of a Transformer would scale in $O(d^{4})$, severely limiting our ability to use a GFlowNet to model a distribution over larger DAGs.

We opted for a Linear Transformer \citep{katharopoulos2020lineartransformers} instead, which relies on a linearized attention mechanism, and has the advantage to not suffer from this quadratic scaling in the input size. \cref{fig:dag-gfn-linear-transformer} shows an illustration of the neural network architecture used in \citep{deleu2022daggflownet}. On the one hand, the distribution $P_{\phi}(i \rightarrow j \mid G, \neg\stopaction)$ over the next edges we can add to $G$ (more precisely, $P_{\phi}(G \cup \{i \rightarrow j\}\mid G, \neg\stopaction)$) is
\begin{equation}
    P_{\phi}(i\rightarrow j\mid G, \neg\stopaction) \propto \mM[i, j]\exp(\vx_{ij}),
\end{equation}
where $\vx$ are the logits returned by one head of the Linear Transformer, and where $\mM$ is the mask associated with $G$; as mentioned in \cref{sec:hierarchical-representation-dag-gfn}, this degenerates to all probabilities being equal to $0$ if the GFlowNet does not admit any valid action from $G$ ($\mM \equiv \vzero$). On the other hand, the probability of terminating $P_{\phi}(\stopaction\mid G) = f_{\phi}(\vg)$ is simply a neural network taking as input an embedding $\vg$ of $G$ (\eg pooling the values returned by the second head of the Linear Transformer), with a sigmoid output.

\paragraph{Graph Neural Network} In subsequent works, starting with \citet{malkin2023gfnhvi}, we chose to replace the linear Transformer with an architecture based on Graph Neural Networks (GNN). This family of neural networks fits all the requirements mentioned above, but with the advantage of being more adapted to graph inputs, and overall simpler to implement \citep{jraph2020github,fey2019pytorchgeometric}. We use a combination of graph networks \citep{battaglia2018graphnetwork} and self-attention blocks \citep{vaswani2017transformers} to encode information about the graph $G$, which
appears in the conditioning of all the quantities of interest. This common backbone returns a graph
embedding $\vg$ of $G$, as well as 2 embeddings $\vu_{i}$ \& $\vv_{i}$ for each node $X_{i}$ in $G$
\begin{equation}
    \vg, \{\vu_{i}, \vv_{i}\}_{i=1}^{d} = \mathrm{SelfAttention}_{\phi}\big(\mathrm{GraphNet}_{\phi}(G)\big).
    \label{eq:dag-gfn-gnn-backbone}
\end{equation}
Same as above, we can parametrize the probability of selecting the ``$\stopaction$'' action using $\vg$ with $P_{\phi}(\stopaction\mid G) = f_{\phi}(\vg)$,
where $f_{\phi}$ is a neural network with a sigmoid output. Inspired by \citet{lorch2021dibs}, we chose to parametrize the remaining distribution of continuing to add an edge as
\begin{equation}
    P_{\phi}(i\rightarrow j\mid G, \neg \stopaction) \propto \mM[i, j]\exp\big(\vu_{i}^{\top}\vv_{j}\big).
    \label{eq:dag-gfn-gnn-parametrization-continue}
\end{equation}
We will see in subsequent chapters that this neural network architecture is flexible in that it allows us to parametrize additional quantities, necessary for example to sample the parameters of the Bayesian network along with its structure $G$ (\cref{sec:jsp-gfn-parametrization-forward-trasition-probabilities}).

\section{DAG-GFlowNet: Bayesian structure learning with GFlowNets}
\label{sec:dag-gfn}\index{Structure learning!Bayesian structure learning}\index{Directed acyclic graph!DAG-GFlowNet|textbf}
So far, the description in \cref{sec:gflownet-over-dags} covered the design of a GFlowNet over DAGs, and was completely independent of the application and the choice of reward function $R(G)$. In this section, we will use such a GFlowNet in order to model the posterior distribution $P(G\mid \gD)$ in \cref{eq:marginal-posterior-dags} over Bayesian networks, thus providing a novel method for Bayesian structure learning (we saw in \cref{sec:bayesian-structure-learning} alternative algorithms). Since a GFlowNet that satisfies the modified detailed balance condition of \cref{prop:modified-detailed-balance} induces a distribution $P_{F}^{\top}(G) \propto R(G)$ (\cref{cor:modified-db-proportional-reward}), we will naturally define its reward function as the joint probability between a DAG $G$ and the dataset $\gD$:
\begin{equation}
    R(G) \triangleq P(\gD\mid G)P(G),
    \label{eq:dag-gfn-reward}
\end{equation}
where $P(G)$ is a prior over DAGs \citep{eggeling2019structureprior}, and $P(\gD\mid G)$ is the marginal likelihood; even though the marginal likelihood is in general intractable to compute, we will consider in this chapter only cases where we can compute it analytically in an efficient way. Note that the reward depends implicitly on the fixed dataset $\gD$. By Bayes' rule \cref{eq:bayes-rule}, a GFlowNet over DAGs with this choice of reward function will indeed model the posterior distribution $P(G\mid \gD) \propto R(G)$. We call our method \emph{DAG-GFlowNet} \citep{deleu2022daggflownet}.

\subsection{Modularity \& computational efficiency}
\label{sec:dag-gfn-modularity-computational-efficiency}
Following prior works on Bayesian structure learning, we assume that both the prior over parameters $P(\theta\mid G)$ of the Bayesian network (required to compute the marginal likelihood) and the prior over structures $P(G)$ are \emph{modular} \citep{heckerman1995bde,chickering1995learning}, meaning that they can be written as a product of terms involving each $X_{j}$ and its parents only. As a consequence, the reward $R(G)$ is also itself modular, and its logarithm can be written as a sum of local scores that only depend on individual variables and their parents in $G$:
\begin{equation}
    \log R(G) = \sum_{j=1}^{d}\mathrm{LocalScore}\big(X_{j}\mid \parents_{G}(X_{j})\big).
    \label{eq:dag-gfn-modular-reward}
\end{equation}
With our choice of reward, $\log R(G)$ corresponds to the Bayesian score we introduced in \cref{sec:score-based-methods}. Examples of modular scores for the log-marginal likelihood $\log P(\gD \mid G)$ include the BDe score \citep{heckerman1995bde} and the BGe score \citep{geiger1994bge,kuipers2014bgeaddendum}; see also \cref{app:bayesian-scores} for details about these scores. We can observe that the objective in \cref{eq:dag-gfn-learning-objective} used to fit the parameters $\phi$ of the neural network for $P_{\phi}$ only involves the difference in log-rewards $\log R(G') - \log R(G)$ between two consecutive states, where $G'$ is the result of adding some edge $X_{i} \rightarrow X_{j}$ to the DAG $G$. Using the assumption of modularity, we can compute this difference efficiently, as the terms in \cref{eq:dag-gfn-modular-reward} remain unchanged for any $j'\neq j$: %
\begin{align}
    \delta(G, G') &\triangleq \log R(G') - \log R(G)\label{eq:delta-score}\\
    &= \mathrm{LocalScore}\big(X_{j}\mid \parents_{G}(X_{j})\cup \{X_{i}\}\big) - \mathrm{LocalScore}\big(X_{j}\mid \parents_{G}(X_{j})\big).
\end{align}
This difference in local scores is sometimes called the \emph{delta score}, or the \emph{incremental value} \citep{friedman2003ordermcmc}, and it has been employed in the literature to improve the efficiency of search algorithms \citep{chickering2002ges,koller2009pgm}.

\subsection{Target network}
\label{sec:target-network}
Continuing with our inspiration from reinforcement learning, we can make use of other tools that have proven to be effective in that field. Since the modified detailed balance loss we use here is equivalent to (a policy-parametrized version of) the Soft Q-Learning algorithm \citep{haarnoja2017sql}, a ``soft'' version of the classic Q-Learning algorithm, it is all the more natural to again borrow tools from Deep Q-Learning \citep{van2018deadlytriad}. In particular, we used a \emph{target network} in our evaluation of $P_{\widebar{\phi}}(\terminal \mid G')$ in \cref{eq:dag-gfn-learning-objective}, where the parameters $\widebar{\phi}$ are only updated periodically at a lower frequency. The residual then becomes
\begin{equation}
    \Delta(G\rightarrow G';\phi) = \log \frac{R(G')P_{B}(G\mid G')P_{\phi}(\terminal \mid G)}{R(G)P_{\phi}(G'\mid G)P_{\textcolor{Red}{\widebar{\phi}}}(\terminal \mid G')}.
    \label{eq:dag-gfn-learning-objective-target-network}\index{Detailed balance!Modified detailed balance}
\end{equation}
The use of a target network has been popularized in reinforcement learning \citep{mnih2015dqn,mnih2016a3c,lillicrap2016ddpg}, and later in self-supervised learning \citep{grill2020byol,he2020moco,caron2021dino}, to stabilize training.

But beyond the simple analogy with reinforcement learning, we found that having a target network was actually necessary to avoid degeneracy during training. This would happen because of the seemingly lack of ``grounding'' of the condition in \cref{prop:modified-detailed-balance}: everything seem to evolve completely freely. To see this better, recall that the modified detailed balance condition is
\begin{equation}
    R(G')P_{B}(G\mid G')P_{\phi}(\terminal \mid G) = R(G)P_{\phi}(G'\mid G)P_{\phi}(\terminal\mid G'),
    \label{eq:modified-db-condition-target-network}
\end{equation}
for a transition $G\rightarrow G'$, where the only learnable quantity is $P_{\phi}$. At first glance, a simple way to satisfy this equation regardless of the transition or the reward function would be to set $P_{\phi}(\terminal\mid G) \approx 0$ for all the DAGs $G$, since the probability of terminating appears on both sides of \cref{eq:modified-db-condition-target-network}, making this equation trivial. Of course since $P_{\phi}(\terminal\mid G)$ is typically parametrized by a neural network with a sigmoid output (\cref{sec:dag-gfn-neural-network-parametrization}), it would never be truly zero; but it could be so small that it would be effectively as if it were. This would yield a degenerate forward transition probability $P_{\phi}$ that would learn to always add edges to the graph, and to never stop. It would be forced to stop only when it reaches the end of the state space, where the graph is ``complete'' (\ie it has the maximum number of edges $d(d-1)/2$), there is no edge that can be added without introducing a cycle, and termination is the only option. This would be regardless of the reward function, and clearly not a good way to sample approximately from the posterior distribution $P(G\mid \gD)$.

This last point is in fact critical: we \emph{can't} set $P_{\phi}(\terminal \mid G) \approx 0$ everywhere, since at least for all the complete graphs $G$, we are forced to terminate and we must have $P_{\phi}(\terminal \mid G) = 1$. That's why this observation does not contradict our \cref{cor:modified-db-proportional-reward}, because it's not entirely true: there exist some graphs that ``ground'' the value of $P_{\phi}(\terminal \mid G)$. But in practice, transitions $G \rightarrow G'$ where $G'$ is complete are few and far between in our data collection to see the real effect of grounding. Thus, without a target network, what we observe in practice is $P_{\phi}(\terminal\mid G)$ \& $P_{\phi}(\terminal\mid G')$ both going towards $0$ together. This kind of degeneracy is also a known issue in reinforcement learning. Having a target network for $P_{\widebar{\phi}}(\terminal \mid G')$ allows us to artificially ground that value for some iterations (this one can't go to $0$ along with $P_{\phi}(\terminal\mid G)$) and mitigates this behavior.

\subsection{Limitations of GFlowNets for Bayesian inference}
\label{sec:limitations-gflownets-bayesian-inference}
Bayesian inference is particularly appealing when the dataset $\gD$ is small to moderately large. Indeed, these are the cases that would benefit the most from an estimation of the epistemic uncertainty over the hypotheses of interest (here, the structure of a Bayesian network). When the dataset becomes very large though, the posterior distribution becomes more and more peaky, and in the limit approaches a distribution whose support is the Markov equivalence class of the ``true'' structure $G^{\star}$ (following the result of \cref{thm:structure-identifiability}, provided that the support of the prior covers it). As a consequence, the delta-score in \cref{eq:delta-score}, which is required to compute the modified detailed balance loss, can take a wide range of values: adding an edge to a graph can drastically increase or decrease its score (\ie the log-reward $\log R(G)$). In turn, the neural network parametrizing $P_{\phi}(G'\mid G)$ needs to compensate for these large fluctuations, making it harder to train.

Unfortunately, some of the standard techniques used in machine learning to tackle this issue, such as the normalization of the inputs, cannot be applied here. Normalizing the delta-scores would be equivalent to modifying the rewards $R(G)$ and $R(G')$ themselves, and it would inevitably change the distribution being approximated. Indeed, if were were to normalize the delta-score by a constant $\alpha$ in the residual in \cref{eq:dag-gfn-learning-objective-target-network}, we would get
\begin{equation}
    \frac{\log R(G') - \log R(G)}{\alpha} + \log \frac{P_{B}(G\mid G')P_{\phi}(\terminal\mid G)}{P_{\phi}(G'\mid G)P_{\widebar{\phi}}(\terminal\mid G')} = \log \frac{R(G')^{1/\alpha}P_{B}(G\mid G')P_{\phi}(\terminal\mid G)}{R(G)^{1/\alpha}P_{\phi}(G'\mid G)P_{\widebar{\phi}}(\terminal\mid G')}.
\end{equation}
In other words, instead of approximating the posterior $P(G\mid \gD)$, we would approximate a distribution $\propto P(G\mid \gD)^{1/\alpha}$. Solutions may include:
\begin{enumerate}
    \item Using a scheduled temperature, slowly decreasing the temperature from a large value of $\alpha$ (simpler to model with a GFlowNet) to $\alpha = 1$ over the course of training, inspired by \emph{simulated annealing} \citep{kirkpatrick1983simulatedannealing}. We could even have the GFlowNet being conditioned on the temperature (\citealp{kim2024logitgfn}; see also \cref{sec:conditional-flow-networks});
    \item Reparametrizing the forward transition probability distribution $P_{\phi}(G'\mid G)$ to better handle large fluctuations of delta-scores. For example in a different context, \citet{metz2019learnedoptimizers} have a network output 2 quantities $o_{1}$ and $o_{2}$ that are then combined into ``$\exp(o_{1})o_{2}$''. This is inspired by the exponent/mantissa decomposition of the floating point representation;
    \item Rescaling the outputs of the neural network (logits) with the size of the dataset $|\gD|$. We have found that this simple solution works well in some cases, but at the cost of worsening the stability of optimization.
\end{enumerate}
Note that none of this is specific to DAG-GFlowNet alone, and these considerations hold for modeling posterior distributions with a GFlowNet in general when the dataset is large. Another promising direction is to take an incremental approach as described in \cref{sec:kl-regularized-consequence-bayesian-inference}. This would allow us to have a good enough starting point every time new data gets incorporated, to warm start training and refine the parameters $\phi$ of the GFlowNet more and more.

\subsection{Limitations of DAG-GFlowNet}
\label{sec:limitations-dag-gfn}
In addition to the general limitations of applying GFlowNets to the problem of Bayesian inference mentioned in the previous section, there are also limitation specific to DAG-GFlowNet itself. In particular, we have made the strong assumption in this chapter that the marginal likelihood $P(\gD\mid G)$ (necessary to compute the reward, or the delta-score) could be computed in an efficient way. This limits the scope of application of DAG-GFlowNet to cases where the following marginalization
\begin{equation}
    P(\gD\mid G) = \int_{\theta}P(\gD\mid G, \theta)P(\theta\mid G)d\theta,
\end{equation}
can be done analytically, which is only possible for a handful of models and carefully chosen priors $P(\theta\mid G)$. In this chapter, we will only consider linear-Gaussian models as introduced in \cref{ex:linear-gaussian-model} with a Normal-Wishart prior over parameters leading to the \emph{Bayesian Gaussian equivalent} score (BGe; \citealp{geiger1994bge,kuipers2014bgeaddendum}), and Bayesian networks with categorical random variables with a Dirichlet prior over parameters leading to the \emph{Bayesian Dirichlet equivalent} score (BDe; \citealp{heckerman1995bde}). The ``equivalent'' in both names means that two Markov equivalent structures will have the same marginal likelihood under these scores. We note that marginalization is also possible analytically in cases where the conditional probability distributions are parametrized by Gaussian Processes \citep{vonkugelgen2019gpstructure}, which may encode non-linear relations. Instead, we will see in the next chapter how to extend the ideas of DAG-GFlowNet to more general models, including non-linear ones, where the marginalization above will not be necessary anymore.

Finally, another limitation is that we assume that the model has no hidden variables, and we fully observe the whole system. This is a standard assumption in most of the literature on structure learning, with some notable exceptions \citep{friedman1998structuralem,boyen1999semdbn,murphy2002dbn}.

\section{Empirical comparison with the exact posterior}
\label{sec:dag-gfn-comparison-exact-posterior}
Even if generative flow networks were designed specifically to sample from distributions where the partition function was too expensive to compute, it can sometimes be useful to apply them to smaller problems where the target distribution is fully known, in order to test the accuracy of the approximation returned by the GFlowNet. In the context of Bayesian structure learning, this means testing it on scenarios where the number of random variables is small enough, so that the number of possible DAGs is not prohibitively large. In those settings, we can compute the posterior distribution $P(G\mid \gD)$ with an exhaustive enumeration of all the DAGs $G$. This is feasible for $d=5$ variables, where we have a total of $29,281$ DAGs.

\paragraph{Experimental setup} We generated a synthetic dataset of observations $\gD$ using a randomly constructed linear-Gaussian Bayesian network, as we described in \cref{ex:linear-gaussian-model}. We follow the experimental setup of \citet{lorch2021dibs}, which we describe below.
\begin{enumerate}
    \item We first sampled the DAG structure $G^{\star}$ of the Bayesian network over $d=5$ random variables using an Erd\H{o}s-R\'{e}nyi model \citep{erdos1960ergraphs}. The probability of creating an edge between two nodes was scaled in such a way that there is on average $d$ edges; this is a setting commonly referred to as ``ER1'' in the structure learning literature \citep{lachapelle2020grandag};
    \item Once the structure is known, we sample the parameters $\theta^{\star}_{ji}\sim \gN(0, 1)$ of the linear-Gaussian model \cref{eq:linear-gaussian-model} randomly from a unit Normal distribution. The noise variables are given by $\varepsilon_{i} \sim \gN(0, 0.01)$. This specifies completely the Bayesian network;
    \item Once the full Bayesian network $(G^{\star}, \theta^{\star})$ is known, we generate $N = 100$ observations from it using ancestral sampling, as described in \cref{sec:sampling-probabilistic-inference}, to form the dataset $\gD$.
\end{enumerate}
After this process, the Bayesian network $(G^{\star}, \theta^{\star})$ used to generate data is completely discarded, and we only keep the resulting dataset $\gD$, required to compute the reward $R(G)$. In this section, we will compute the exact posterior distribution $P(G\mid \gD)$ by computing the reward of all the possible DAGs over $d=5$ nodes, and normalizing their values to sum to $1$. We used the \emph{BGe score} \citep{geiger1994bge} in order to compute the reward, with the default hyperparameters suggested by \citet{kuipers2014bgeaddendum}. We also used a uniform prior $P(G)$ over DAGs.

\subsection{Evaluation of the terminating state probability with dynamic programming}
\label{sec:evaluation-terminating-state-probability-dynamic-programming}
We introduced generative flow networks only as sampling machines. While we saw in \cref{sec:sampling-ebm} that this is sufficient for most applications, it might be sometimes necessary to also compute the probability $P_{F}^{\top}(x)$ for some terminating state $x\in\gX$, once we have learned $P_{F}$. This is the case for example if we want to compare the approximation $P_{F}^{\top}(x)$ with the target distribution $\propto R(x)$ (in cases where we can compute it). In this section, we place ourselves in the general case, where the we have a GFlowNet over a pointed DAG $\gG = (\widebar{\gS}, \gA)$ (where all the states are not necessarily all terminating, unlike the application in this chapter), with a forward transition probability $P_{F}$; this $P_{F}$ may have been learned with any flow matching loss, and not necessarily the modified detailed balance loss we consider in this chapter.

\begin{algorithm}[t]
    \caption{Evaluation of $P_{F}^{\top}(x)$ with dynamic programming.}
    \label{alg:terminating-state-probability-dynamic-programming}
    \begin{algorithmic}[1]
        \Require A pointed DAG $\gG = (\widebar{\gS}, \gA)$, a forward transition probability $P_{F}$.
        \Ensure The probability $P_{F}^{\top}(x)$ for all terminating states $x\in\gX$.
        \State Find a topological ordering $\sigma$ of $\gG$ \Comment{Excluding $s_{0}$ \& $\terminal$}
        \State Initialize $F(s_{0}) \leftarrow 1$
        \ForAll{$s' \in \sigma$}
            \State $F(s') \leftarrow \sum_{s\in\parents_{\gG}(s)}F(s)P_{F}(s'\mid s)$
        \EndFor
        \State \Return $P_{F}^{\top}(x) \equiv F(x)P_{F}(\terminal \mid x)$
    \end{algorithmic}
\end{algorithm}

In general, computing the terminating state distribution $P_{F}^{\top}(x)$ requires summing over all the trajectories leading to $x$ (\cref{def:terminating-state-probability}), which may be prohibitively large and involves redundant components across different terminating states. Here, we will instead look for a solution that can compute the whole distribution $P_{F}^{\top}$ for all the terminating states at once, based on \emph{dynamic programming}. The key observation is to use \cref{prop:terminating-state-probability-from-flow}: if we could construct a flow whose forward transition probability is $P_{F}$, then we could simply read $P_{F}^{\top}$ from the flow at the terminating edges. To build the flow, the idea is to ``push'' a single unit of flow at the initial state $s_{0}$ down $\gG$, following $P_{F}$. The algorithm is given in \cref{alg:terminating-state-probability-dynamic-programming}, and we can show that it is correct.

\begin{proposition}
    \label{eq:terminating-state-probability-dynamic-programming-valid}
    \cref{alg:terminating-state-probability-dynamic-programming} is correct, and returns the probability $P_{F}^{\top}(x)$ for all $x\in\gX$.
\end{proposition}

\begin{proof}
    The flow defined by \cref{alg:terminating-state-probability-dynamic-programming} is a Markovian flow, defined uniquely by the edge flow $F(s\rightarrow s') = F(s)P_{F}(s'\mid s)$ (it follows the flow matching condition of \cref{prop:flow-matching-condition-state-flow-pF}). From this, it is clear that $P_{F}$ is the forward transition probability associated with this flow (\cref{def:transition-probabilities-from-flow}). Moreover, its total flow is $Z = F(s_{0}) = 1$ (\cref{prop:initial-flow-total-flow}), thanks to the initialization of the algorithm. We can conclude with \cref{prop:terminating-state-probability-from-flow} that $P_{F}^{\top}(x) = F(x\rightarrow \terminal) = F(x)P_{F}(\terminal\mid x)$.
\end{proof}

This algorithm allows us to evaluate the whole terminating state distribution in $O(|\gA|)$, making it applicable only on small scale problems, such as the GFlowNet presented in this chapter, for a small enough number of variables $d$.

\subsection{Comparison with the exact posterior}
\label{sec:dag-gfn-small-graphs-comparison}
To evaluate the quality of the posterior approximation returned by DAG-GFlowNet, we test it on various structural \emph{features}. A feature is a marginal probability on quantities of interest for Bayesian networks. For example, an \emph{edge feature} \citep{buntine1991refinementbn,cooper1992structurelearning} corresponds to the marginal probability of an edge $X_{i} \rightarrow X_{j}$ belonging to the graph:
\begin{equation}
    P(X_{i}\rightarrow X_{j}) = \E_{G\sim P}\big[\mathds{1}(X_{i} \rightarrow X_{j} \in G)\big],
    \label{eq:edge-feature}
\end{equation}
where the distribution $P$ over DAGs may be either the true posterior distribution $P(G\mid \gD)$ (target probability), or the terminating state distribution of the GFlowNet $P_{\phi}^{\top}(G)$ (estimate). We can also compare these two distributions in terms of their \emph{path feature}, the marginal probability of a directed path $X_{i} \rightsquigarrow X_{j}$ existing from $X_{i}$ to $X_{j}$, or their \emph{Markov feature}, the marginal probability of $X_{i}$ being in the Markov blanket of $X_{j}$ \citep{friedman2003ordermcmc}.

\begin{figure}[t]
    \centering
    \begin{adjustbox}{center}%
    \includegraphics[width=470pt]{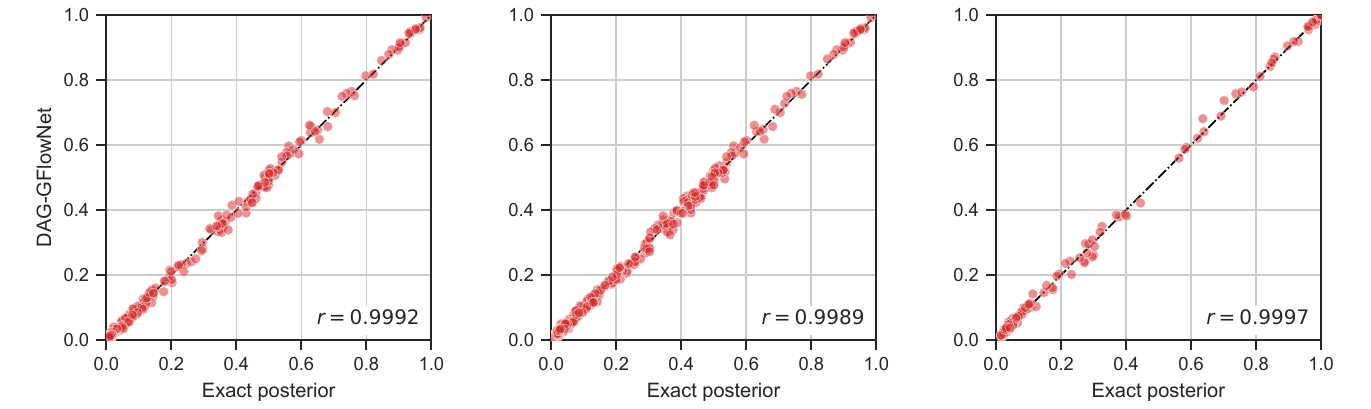}%
    \end{adjustbox}
    \begin{adjustbox}{center}%
        \begin{subfigure}[b]{160pt}%
            \caption{Edge features}%
            \label{fig:dag-gfn-small-graphs-edge}%
        \end{subfigure}%
        \begin{subfigure}[b]{155pt}%
            \caption{Path features}%
            \label{fig:dag-gfn-small-graphs-path}%
        \end{subfigure}%
        \begin{subfigure}[b]{155pt}%
            \caption{Markov blanket features}%
            \label{fig:dag-gfn-small-graphs-markov}%
        \end{subfigure}%
    \end{adjustbox}
    \caption[Comparison between the exact posterior and the approximation from DAG-GFlowNet]{Comparison between the exact posterior distribution and the approximation from DAG-GFlowNet, for different structural features: (a) edge features $X_{i} \rightarrow X_{j}$, (b) path features $X_{i} \rightsquigarrow X_{j}$, and (c) Markov blanket features $X_{i} \sim_{M} X_{j}$. Each point corresponds to a feature (\ie marginal probability) computed for a specific pair $(X_{i}, X_{j})$ in a graph over $d = 5$ nodes, either based on the exact posterior $P(G\mid \gD)$ (x-axis), or the approximation found with DAG-GFlowNet (y-axis; see \cref{sec:evaluation-terminating-state-probability-dynamic-programming}). Points on the diagonal line $y = x$ correspond to an exact agreement between the exact posterior and the approximation \emph{on that pair of variables}. The Pearson correlation coefficient $r$ is included at the bottom-right corner of each plot.}
    \label{fig:dag-gfn-small-graphs}
\end{figure}

In \cref{fig:dag-gfn-small-graphs}, we compare the probabilities of these features for both the exact posterior and the distribution induced by DAG-GFlowNet, where we repeated the experiment described above with 20 different datasets coming from different Bayesian networks $(G^{\star}, \theta^{\star})$. We observe that the probabilities of all structural features estimated by the GFlowNet are strongly correlated with the exact marginal probabilities. This shows that DAG-GFlowNet is capable of learning a very accurate approximation of the posterior distribution over graphs $P(G\mid \gD)$.

\subsection{Hierarchical variational inference \& the advantage of off-policy learning}
\label{sec:dag-gfn-effect-off-policy}\index{Variational inference!Hierarchical}
In \cref{sec:gflownets-variational-inference}, we saw that GFlowNets are closely related to hierarchical variational models, especially when they are trained on-policy. In turn, this highlighted a major advantage of GFlowNets over these other variational methods: they can directly be trained off-policy, without having to resort to importance sampling. In this section, we will show this advantage empirically on DAG-GFlowNet. In addition to training the GFlowNet for Bayesian structure learning with the modified detailed balance loss, we also trained it using the trajectory balance loss in \cref{eq:trajectory-balance-loss}, as well as with the reverse KL objective introduced in \cref{eq:hvi-loss}, for comparison. For the latter two losses, we considered two settings: either on-policy, using the current $P_{F}$ to acquire data, and off-policy using the behavior policy presented in \cref{sec:off-policy-training}; note that the reverse KL objective is trained with the estimate presented in \cref{sec:variance-reduction-gradient-estimate} with a local control variate (called ``$b_{\mathrm{local}}$'' in that section), and it necessitates importance sampling to be trained off-policy.

\begin{table}[t]
    \centering
    \caption[Comparison of the Jensen-Shannon divergence for Bayesian structure learning]{Comparison of the Jensen-Shannon divergence for Bayesian structure learning, showing the advantage of off-policy TB over on-policy TB, and on-policy or off-policy HVI. Lower is better.}
    \begin{tabular}{@{}lccc}
    \toprule
    &\multicolumn{3}{c}{Number of nodes}\\\cmidrule{2-4}
    Objective & 3 & 4 & 5 \\
    \midrule
    Modified detailed balance &  $5.32 \pm 4.15 \times 10^{-6}$ &  $2.05 \pm 0.70 \times 10^{-5}$ &  $\mathbf{4.65 \pm 1.08 \times 10^{-4}}$ \\
    Off-policy trajectory balance &  $\mathbf{3.70 \pm 2.51 \times 10^{-7}}$ &  $\mathbf{9.35 \pm 2.99 \times 10^{-6}}$ &  $5.44 \pm 2.47 \times 10^{-4}$ \\
    On-policy trajectory balance &               $0.022 \pm 0.007$ &               $0.123 \pm 0.028$ &               $0.277 \pm 0.040$ \\
    On-policy reverse KL (HVI) &               $0.022 \pm 0.007$ &               $0.125 \pm 0.027$ &               $0.306 \pm 0.042$ \\
    Off-policy reverse KL (HVI) &               $0.014 \pm 0.008$ &               $0.605 \pm 0.019$ &               $0.656 \pm 0.009$ \\
    \bottomrule
    \end{tabular}
    \label{tab:dag-gfn-hvi-jsd}
\end{table}

We compare the performance of all these methods based on the \emph{Jensen Shannon divergence} (\gls{jsd}) between the distribution induced by the GFlowNet $P_{\phi}^{\top}(G)$ and the true posterior distribution $P(G\mid \gD)$. Recall that the JSD is given by
\begin{align}
    \mathrm{JSD}\big(P_{\phi}^{\top}(G), P(G\mid \gD)\big) &= \frac{1}{2}\left[\KL\big(P_{\phi}^{\top}(G)\,\|\,M(G)\big) + \KL\big(P(G\mid \gD)\,\|\,M(G)\big)\right] \label{eq:jensen-shannon-divergence}\\ \textrm{where}\quad M(G) &= \frac{P_{\phi}^{\top}(G) + P(G\mid \gD)}{2}.
\end{align}
We show in \cref{tab:dag-gfn-hvi-jsd} the performance of all these methods on Bayesian networks with $d\leq 5$ random variables, to ensure that the true posterior $P(G\mid \gD)$ can still be computed analytically. We observe that on the easiest setting (graphs over $d = 3$ nodes), all methods accurately approximate the posterior distribution. But as we increase the complexity of the problem, we observe that the accuracy of the approximation with off-policy reverse KL degrades significantly, while the ones found with the off-policy GFlowNet objectives (modified detailed balance \& trajectory balance) remain very accurate. We also note that the performance of the on-policy algorithms degrades too, but not as rapidly; furthermore they achieve the same JSD, confirming our observation in \cref{sec:analysis-gradients-gfn-hvi} that trajectory balance and reverse KL are closely connected when trained on-policy.

\begin{figure}[t]
    \centering
    \begin{adjustbox}{center}%
    \includegraphics[width=470pt]{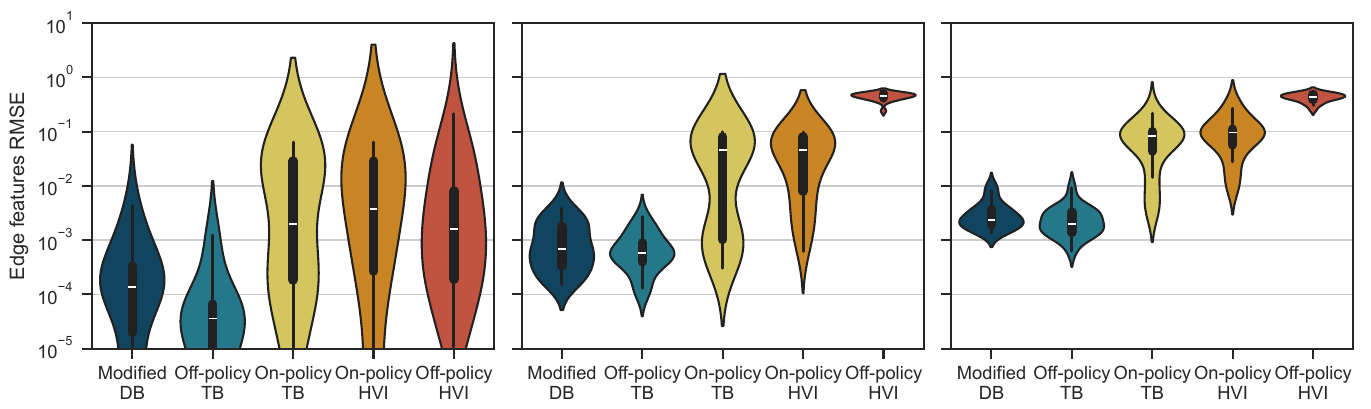}%
    \end{adjustbox}
    \begin{adjustbox}{center}%
        \begin{subfigure}[b]{174pt}%
            \caption{$d = 3$ variables}%
            \label{fig:gfn-hvi-edge-marginals-3}%
        \end{subfigure}%
        \begin{subfigure}[b]{134pt}%
            \caption{$d = 4$ variables}%
            \label{fig:gfn-hvi-edge-marginals-4}%
        \end{subfigure}\hfill%
        \begin{subfigure}[b]{138pt}%
            \caption{$d = 5$ variables}%
            \label{fig:gfn-hvi-edge-marginals-5}%
        \end{subfigure}%
    \end{adjustbox}
    \caption[Comparison of the edge features computed with GFlowNet objectives and Reverse KL]{Comparison of the edge features computed either with GFlowNet objectives, or the Reverse KL objective in \cref{eq:hvi-loss}. Performance is reported as the root mean square error (RMSE) between the edge features computed with the exact posterior $P(G\mid \gD)$ and the approximation (\cref{sec:evaluation-terminating-state-probability-dynamic-programming}); lower is better. Each method is run on $20$ different datasets.}
    \label{fig:gfn-hvi-edge-marginals}
\end{figure}

This trend gets confirmed when we look at structural features, as we did in the previous section. In \cref{fig:gfn-hvi-edge-marginals}, we show the root mean-square error (\gls{rmse}) of the edge features (in other words, this is the RMSE of the linear fit in \cref{fig:dag-gfn-small-graphs-edge} on the diagonal line). This highlights even more the similar behavior of the trajectory balance and the reverse KL losses on-policy.

\section{Evaluation on simulated data}
\label{sec:dag-gfn-evaluation-large-graphs}
In this section, we will consider a larger setting to fully leverage the advantage of GFlowNets to approximate distributions whose partition function is intractable. We take a similar experimental setup as the one described in the previous section (linear-Gaussian model, where data is obtained from randomly generated Bayesian networks), except that the Bayesian networks will now be over $d=20$ random variables. To go along with this increase in the number of nodes, we will still generate DAGs $G^{\star}$ based on an Erd\H{o}s-R\'{e}nyi model, but this time with $2d$ edges on average (ER2), to match the experimental settings studied in prior work \citep{lorch2021dibs,cundy2021bcdnets}. The reward is still computed using the BGe score, and any experiment is conducted on $20$ different randomly generated datasets $\gD$ from $20$ different Bayesian networks $(G^{\star}, \theta^{\star})$.

\subsection{Estimation of the terminating state probability}
\label{sec:estimation-terminating-state-probability}
While we saw a method to compute the terminating state probability $P_{F}^{\top}(x)$ using dynamic programming in \cref{sec:evaluation-terminating-state-probability-dynamic-programming}, this algorithm is impractical once the pointed DAG $\gG$ of the GFlowNet is too large. This is the case for example in DAG-GFlowNet as soon as $d \geq 7$. For lack of a more efficient way to compute $P_{F}^{\top}(x)$ for a single terminating state $x$, we will show in this section how to approximate it. We will take the example of DAG-GFlowNet for concreteness here. Recall that the terminating state probability $P_{F}^{\top}(G)$ is given by
\begin{equation}
    P_{F}^{\top}(G) = \sum_{\tau: G_{0}\rightsquigarrow G}P_{F}(\tau).
    \label{eq:terminating-state-distribution-definition-estimation-dag-gfn}
\end{equation}
If $G$ has $K$ edges, then this means that there are $K!$ trajectories from $G_{0}$ to $G$ (\ie adding the edges in any order), making this sum intractable in general. We can leverage the fact that the backward transition probability $P_{B}$ induces a distribution $P_{B}(\tau\mid G)$ over the trajectories leading to $G$ (\cref{lem:PB-distribution-prefix}) to write it as
\begin{equation}
    P_{F}^{\top}(G) = \sum_{\tau: G_{0}\rightsquigarrow G}P_{F}(\tau) = \sum_{\tau: G_{0}\rightsquigarrow G}\frac{P_{B}(\tau\mid G)}{P_{B}(\tau\mid G)}P_{F}(\tau) = K!\cdot \E_{\tau\sim P_{B}(\tau\mid G)}\big[P_{F}(\tau)\big],
\end{equation}
where $P_{B}(\tau\mid G) = 1/K!$ regardless of the trajectory $\tau$, since we assumed that the backward transition probability is fixed to the uniform distribution here (\cref{sec:fixed-backward-transition-probability}). This suggests a way to get an unbiased estimate of the terminating state probability: we can build a Monte Carlo estimate of this expectation based on sample trajectories $\{\tau^{(m)}\}_{m=1}^{M}$ from $P_{B}(\tau \mid G)$
\begin{equation}
    P_{F}^{\top}(G) \approx \frac{K!}{M}\sum_{m=1}^{M}P_{F}\big(\tau^{(m)}\big).
    \label{eq:mc-estimate-terminating-state-probability-dag-gfn}
\end{equation}
We can get those trajectories by removing one edge at time uniformly at random, starting at $G$.

While \cref{eq:mc-estimate-terminating-state-probability-dag-gfn} provides an unbiased estimate of $P_{F}^{\top}(G)$, in practice the variance of this estimate will be large due to the combinatorially large space of trajectories leading to $G$, and therefore due to the wide range of values $P_{F}(\tau)$ may take. In order to reduce the variance, we can first identify some trajectories that would contribute the most to the sum in \cref{eq:terminating-state-distribution-definition-estimation-dag-gfn} and complement them with some randomly sampled trajectories as in \cref{eq:mc-estimate-terminating-state-probability-dag-gfn}. In other words, if we had access to a subset $\gT_{\mathrm{top}}$ of $B$ trajectories $\tau: G_{0}\rightsquigarrow G$ that have a large $P_{F}(\tau)$, then
\begin{equation}
    P_{F}^{\top}(G) = \sum_{\tau\in\gT_{\mathrm{top}}}P_{F}(\tau) + \sum_{\tau \notin \gT_{\mathrm{top}}}P_{F}(\tau) \approx \sum_{\tau\in\gT_{\mathrm{top}}}P_{F}(\tau) + \frac{K! - B}{M}\sum_{m=1}^{M}P_{F}\big(\tau^{(m)}\big),
    \label{eq:beam-search-estimate-terminating-state-probability-dag-gfn}
\end{equation}
where the sample trajectories $\{\tau^{(m)}\}_{m=1}^{M}$ can be obtained using rejection sampling, with a uniform proposal as above. The estimate in \cref{eq:beam-search-estimate-terminating-state-probability-dag-gfn} is still unbiased, but with a lower variance. We can use \emph{beam-search} to find the ``top-scoring'' trajectories in $\gT_{\mathrm{top}}$, with a beam-size $B$, as described in \cref{alg:beam-search}. More precisely, we need to run beam-search, starting at $G_{0}$, in such a way that the trajectories are guaranteed to end at $G$. We can achieve this by constraining the set of actions one can take at each step of expansion to move from a graph $G_{t}$ to $G_{t+1} = G_{t} \cup \{e\}$ (using this notation to denote that $G_{t+1}$ is the result of adding
the edge $e$ to $G_{t}$), with the following score:
\begin{equation}
    \widetilde{P}_{F}(G_{t+1}\mid G_{t}) = \mathds{1}(e\in G)P_{F}(G_{t+1}\mid G_{t}).
\end{equation}
In other words, we only keep transitions corresponding to adding edges that we know are in $G$. Note that even though $\widetilde{P}_{F}$ is not a properly defined probability distribution (it does not sum to $1$), we can still use this as a scoring function to run beam-search. Since beam-search is only a heuristic procedure, we are not guaranteed to get the top-$B$ trajectories with highest $P_{F}(\tau)$ necessarily; in practice though, it is an effective method to obtain trajectories with ``high enough'' scores.

Although we presented these estimates of the terminating state probabilities in the case of DAG-GFlowNet, note that this is a valid strategy whenever the objects to be generated are collections of parts that can be added in any order (\eg graphs as a collection of edges). This is not true for any arbitrary GFlowNet though. A notable example that does \emph{not} satisfy this requirement is the molecule generation environment described in \cref{sec:generation-small-organic-molecules}.

\subsection{Medium-scale linear-Gaussian models}
\label{sec:medium-scale-linear-gaussian-models}
In this section, we will compare the performance of DAG-GFlowNet against representative methods for Bayesian structure learning. This includes two methods based on MCMC (\cref{sec:structure-mcmc}), MC\textsuperscript{3} \citep{madigan1995structuremcmc} and Gadget \citep{viinikka2020gadget}, and two methods based on variational inference (\cref{sec:variational-inference-bayesian-structure-learning}), DiBS \citep{lorch2021dibs} and BCD Nets \citep{cundy2021bcdnets}. In addition to these 4 methods, we also consider two methods based on bootstrapping the dataset $\gD$ (\ie resampling $N$ observations with replacement; \citealp{friedman1999bootstrap}), each set being used to learn a single DAG with either the constraint-based method PC \citep{spirtes2000causationpredictionsearch} or the score-based method GES \citep{chickering2002ges}.

\begin{figure}[t]
    \centering
    \begin{adjustbox}{center}%
    \includegraphics[width=520.714285714pt]{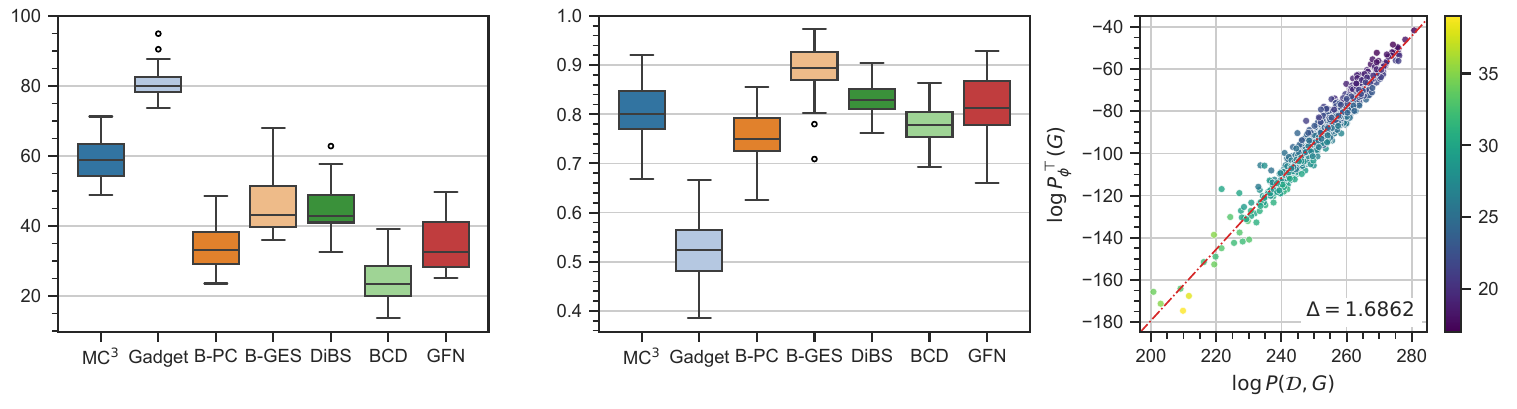}%
    \end{adjustbox}
    \begin{adjustbox}{center}%
        \begin{subfigure}[b]{189.99034749pt}%
            \caption{Expected SHD}%
            \label{fig:dag-gfn-lingauss20-eshd}%
        \end{subfigure}%
        \begin{subfigure}[b]{189.99034749pt}%
            \caption{AUROC}%
            \label{fig:dag-gfn-lingauss20-auroc}%
        \end{subfigure}%
        \begin{subfigure}[b]{140.733590734pt}%
            \caption{Linear-Gaussian}%
            \label{fig:dag-gfn-lingauss20-diag}%
        \end{subfigure}%
    \end{adjustbox}
    \caption[Bayesian structure learning of linear-Gaussian Bayesian networks with $d=20$ nodes]{Bayesian structure learning of linear-Gaussian Bayesian networks with $d=20$ nodes. Results for (a) $\E$-SHD (lower is better) \& (b) AUROC (higher is better) are aggregated over 25 randomly generated datasets $\gD$. Labels: B-PC =
    Bootstrap-PC, B-GES = Bootstrap-GES, BCD = BCD Nets, GFN = DAG-GFlowNet. (c) Comparison between the terminating state log-probability $\log P_{\phi}^{\top}(G)$ (estimated, see \cref{sec:estimation-terminating-state-probability}) and the log-reward $\log R(G) = \log P(\gD, G)$. Each of the $1,000$ points represents a graph sampled with DAG-GFlowNet. The color of each point indicates the number of edges in the corresponding graph. The slope $\Delta$ of the linear fit obtained with RANSAC \citep{fischler1981ransac} is shown at the bottom-right corner.}
    \label{fig:dag-gfn-lingauss20}
\end{figure}

Since we know the ground-truth graph $G^{\star}$ that was used to generate $\gD$, we evaluate the performance of each algorithm based on structural metrics comparing to $G^{\star}$. Following the literature in Bayesian structure learning \citep{lorch2021dibs,cundy2021bcdnets}, we consider the \emph{expected structural hamming distance} (\gls{eshd}) between samples of a posterior approximation $\widehat{P}(G\mid \gD)$ and $G^{\star}$
\begin{equation}
    \E\textrm{-}\mathrm{SHD} \triangleq \E_{\widehat{P}(G\mid \gD)}\big[\mathrm{SHD}(G, G^{\star})\big],
    \label{eq:expected-shd}
\end{equation}
where $\mathrm{SHD}(G, G^{\star})$ is the structural hamming distance measuring the edit distance (\ie adding, or removing edges) between the two DAGs. We also compute the \emph{area under the receiver operating characteristic curve} (\gls{auroc}) of the edges of $G^{\star}$ against the predictions made from the edge marginals (of the posterior approximation) under varying thresholds \citep{ellis2008bias}. We observe in \cref{fig:dag-gfn-lingauss20-eshd,fig:dag-gfn-lingauss20-auroc} that DAG-GFlowNet is competitive against all other methods, achieving a good trade-off between both metrics, behaving closest to Bootstrap PC.

But going past a comparison with other Bayesian structure learning methods, we would like to assess that DAG-GFlowNet is \emph{intrinsically} an accurate approximation of $P(G\mid \gD)$. Since the state space is now too large ($\approx 10^{72}$ DAGs over $d=20$ nodes), we can't perform the same evaluation as in \cref{sec:dag-gfn-small-graphs-comparison} anymore. Instead, we can compare the terminating state log-probability of DAG-GFlowNet $\log P_{\phi}^{\top}(G)$ with the log reward $\log R(G)$; indeed, since the terminating state probability of the GFlowNet approximated the posterior, we should ideally get by Bayes' rule
\begin{equation}
    \log P_{\phi}^{\top}(G) \approx \log P(G\mid \gD) = \log P(\gD, G) - \log P(\gD) = \log R(G) - \log P(\gD).
    \label{eq:comparison-terminating-state-reward-correlation}
\end{equation}
This means that if $\log P_{\phi}^{\top}(G)$ is an accurate approximation of $\log P(G\mid \gD)$, it should correlate with $\log R(G)$ with a slope of $1$ \citep{bengio2021gflownet}, $\log P(\gD)$ being simply a constant offset. Even though the terminating state probability can't be computed either, we saw in in the previous section how it can be estimated with beam-search \& Monte Carlo. We show in \cref{fig:dag-gfn-lingauss20-diag} that there is indeed a strong linear correlation between $\log P_{\phi}^{\top}(G)$ \& $\log R(G)$ computed on $1,000$ samples from DAG-GFlowNet, with a slope $\Delta$ relatively close to $1$. This suggests that it is again an accurate approximation of the posterior on this medium-size problem, \emph{at least around the modes it captures}.

\subsection{Criticism of the evaluation metrics}
\label{sec:criticism-evaluation-metrics}
In (Bayesian) structure learning, the evaluation of the performance is often made relative to the ``ground-truth'' structure $G^{\star}$ used to generate the dataset $\gD$. This explains largely why this community has focused a lot of its attention on evaluation with synthetic data, to have a fully controlled setting. However in real-world applications, and especially in scientific discovery where we hope that these tools will be used to find new scientific theories, there is no such thing as a \emph{known} ground truth structure, precisely because that is what we want to infer from data. Thus, any metric relying on such a ground-truth, which the \gls{shd} and the AUROC we used in the previous section are examples of, are not applicable \emph{by design} beyond synthetic (or pseudo-real) data.

Another drawback of a structural metric like the SHD that measures differences with $G^{\star}$ is that it tends to favor sparse graphs with very few edges \citep{lorch2022avici}. In the limit case, a trivial algorithm that would always predict the empty graph would fare reasonably well since its SHD would be exactly the number of edges in $G^{\star}$ (edits from the empty graph are just edge additions). For example in the experiment of \cref{fig:dag-gfn-lingauss20-eshd}, this would mean having a $\E$-SHD of about $40$, since the ground-truth graphs have on average $2d$ edges. This would artificially make some methods look better than others, without being of any use in practice.

But beyond these limitations, which are true for the evaluation in structure learning in general, there is a fundamental issue with using a metric like the (expected) SHD for Bayesian structure learning. The reason being that the objective of the Bayesian perspective is \emph{not} to recover $G^{\star}$, but to provide a faithful approximation of the posterior distribution $P(G\mid \gD)$. Suppose that we have an algorithm that would return a Dirac $\delta(G = G^{\star})$; we would always sample $G^{\star}$. Then it would achieve a remarkable $\E$-SHD of $0$. But this would be a terrible approximation of the posterior, first because it has no quantification of epistemic uncertainty, second because at best we can only recover the Markov equivalence class of $G^{\star}$ based on our discussion in \cref{sec:faithfulness-identifiability}, and third because the posterior distribution $P(G\mid \gD)$ that we wish to approximate is a fundamentally ``finite-sample'' quantity. This last point means in particular that $G^{\star}$ may not even be a mode of $P(G\mid \gD)$, due to lack of data. This is true for the AUROC as well.

Probably the most reliable evaluation is on a downstream task. For example, if we have access to a separate set of test data $\gD'$ representative of the system being modeled \citep{eaton2007bayesian}, then we can test the predictive performance of the algorithm based on Bayesian model averaging as described in \cref{sec:bayesian-model-averaging}. Some caution needs to be taken when we consider the problem of (marginal) Bayesian structure learning though: while new data is independent of $\gD$ given the full specification of the Bayesian network (\ie $\gD'\independent \gD \mid G, \theta$), it is not the case anymore when the parameters have been marginalized over: $\gD'\not\independent\gD\mid G$, meaning that $P(\gD'\mid G, \gD) \neq P(\gD'\mid G)$, which was necessary for Bayesian model averaging (see also \cref{fig:meta-bayesian-network}).

Finally, even if it is a more robust measure of the accuracy of the posterior approximation, comparing $\log P_{\phi}^{\top}(G)$ to $\log R(G)$ (as we did in \cref{fig:dag-gfn-lingauss20-diag}) suffers from being biased as we are only comparing those two quantities \emph{on a mode of $P_{\phi}^{\top}$}. Indeed, we use graphs sampled with the GFlowNet for evaluation, and those will tend to be in the mode(s) of the terminating state distribution. This does not necessarily capture the whole complexity of the true posterior $P(G\mid \gD)$, which may not be reflected in the distribution induced by the GFlowNet. Moreover, since $P_{\phi}^{\top}(G)$ can only be estimated (\cref{sec:estimation-terminating-state-probability}), this estimate becomes less reliable  as the size of the problem increases (\ie the number of random variables $d$ increases), because its variance will inevitably increase.

\section{Application to protein signaling networks}
\label{sec:dag-gfn-application-protein-signaling-networks}
Finally to complement our experimental results on simulated data, we also evaluated DAG-GFlowNet on real-world flow cytometry data \citep{sachs2005causal}. The data consists of continuous measurements of $d = 11$ phosphoproteins in individual T-cells. Out of all the measurements, we first selected the $N = 853$ observations corresponding to the first experimental condition of \citet{sachs2005causal} (control) as our dataset $\gD$; we will see in \cref{sec:dag-gfn-sachs-experimental-data} how to also integrate the additional data from other experimental regimes. This is considered a standard dataset in the structure learning literature, and following prior work we used the DAG inferred by the authors of that study as our reference graph (``ground-truth''), containing $d = 11$ nodes and $17$ edges. However, it should be noted that this ``consensus graph'' may not represent a realistic and complete description of the system being modeled here \citep{mooij2020jci}, biological systems often containing feedback processes that can't be modeled as an acyclic graph; we will come back to this limitation in \cref{sec:jsp-gfn-biological-structures-real-data}. We standardized the data, and used again the BGe score to compute $R(G)$.

\subsection{Inference from observational data}
\label{sec:dag-gfn-sachs-inference-observation-data}
\begin{table}[t]
    \centering
    \caption[Learning protein signaling networks with DAG-GFlowNet]{Learning protein signaling networks from flow cytometry data \citep{sachs2005causal}. All results include a 95\% confidence interval estimated with bootstrap resampling. Label: $\E$-\#~Edges = average number of edges in the sampled graphs.}
    \label{tab:dag-gfn-sachs-observational}
\begin{tabular}{lccc}
    \toprule
     & $\mathbb{E}$-\#~Edges & $\mathbb{E}$-SHD  & AUROC \\
    \midrule
    MC\textsuperscript{3} & $10.96 \pm 0.09$ & $22.66 \pm 0.11$ & $0.508$ \\
    Gadget & $10.59 \pm 0.09$ & $21.77 \pm 0.10$ & $0.479$ \\
    Bootstrap GES & $11.11 \pm 0.09$ & $23.07 \pm 0.11$ & $\mathbf{0.548}$ \\
    Bootstrap PC & $\phantom{1}7.83 \pm 0.04$ & $20.65 \pm 0.06$ & $0.520$ \\
    DiBS & $12.62 \pm 0.16$ & $23.32 \pm 0.14$ & $0.518$ \\
    BCD Nets & $\phantom{1}4.14 \pm 0.09$ & $\mathbf{18.14 \pm 0.09}$ & $0.510$ \\
    \midrule
    DAG-GFlowNet & $11.25 \pm 0.09$ & $22.88 \pm 0.10$ & $0.541$ \\
    \bottomrule
\end{tabular}
\end{table}
We compare the performance of DAG-GFlowNet on this protein signaling inference problem against the same baseline methods as the ones we considered in \cref{sec:medium-scale-linear-gaussian-models}. In \cref{tab:dag-gfn-sachs-observational}, we once again report the expected SHD \& AUROC (based on the ``ground-truth'' graph), despite the comments we made in the previous section. In addition to those metrics, we also include the average number of edges present in the graphs to get a more comprehensive view; all the metrics are computed based on $1,000$ samples from each posterior approximation.

We observe that while BCD Nets and Bootstrap PC have a smaller $\E$-SHD, suggesting that the distribution is concentrated closer to the consensus graph than other methods, in reality they tend to be more conservative and sample graphs with fewer edges. Overall, DAG-GFlowNet offers a good trade-off between performance (as measured by the $\E$-SHD and the AUROC), and getting a distribution that assigns higher probability to DAGs with more edges, on par with Bootstrap GES. Interestingly, we also observed that $1.50\%$ of the graphs samples with DiBS contained a cycle, validating empirically what we saw in \cref{sec:variational-inference-bayesian-structure-learning} (\ie methods based on continuous relaxations of the acyclicity constraint have a risk to sample cyclic graphs).

\begin{figure}[t]
    \centering
    \begin{adjustbox}{center}
    \includegraphics[width=470pt]{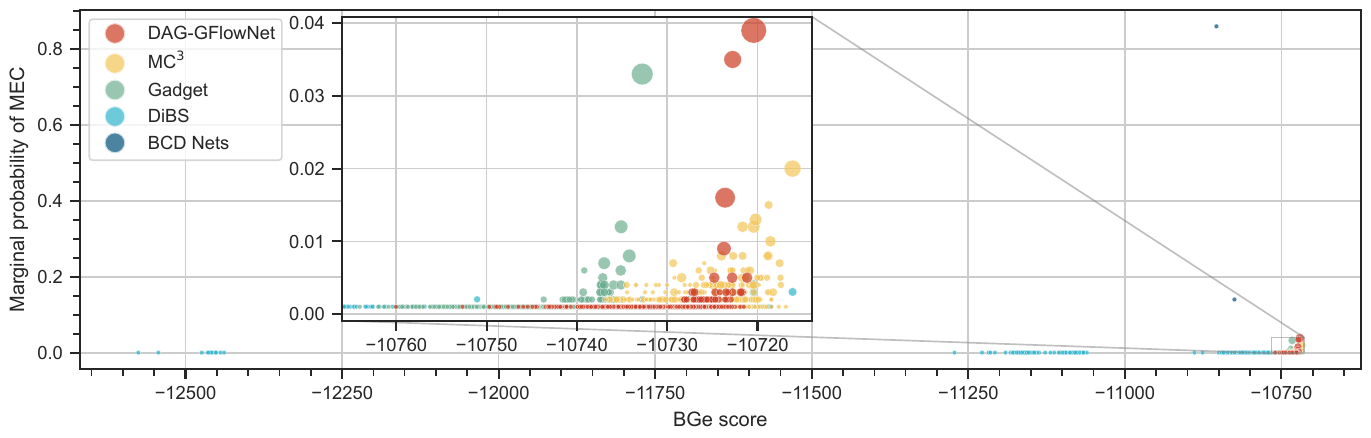}
    \end{adjustbox}
    \caption[Coverage of the posterior approximations learned on flow cytometry data]{Coverage of the posterior approximations learned on flow cytometry data. Each point corresponds to a sampled Markov equivalence class, and its size represents the number of unique DAGs sampled from the posterior approximation (with $1,000$ samples) in the MEC. The x-axis is the BGe score of the MEC, and the y-axis is the marginal probability of the MEC (\ie the normalized count of DAGs falling in that MEC). Note the multiple points in light blue close to the $y = 0$ line (DiBS), and the 2 points in dark blue around $x = -10,850$ (BCD Nets).}
    \label{fig:dag-gfn-sachs}
\end{figure}

Beyond these metrics though, and in line with our comments in \cref{sec:criticism-evaluation-metrics}, we would like to test if the advantages of Bayesian structure learning are also reflected in the distribution induced by DAG-GFlowNet. In particular, we want to study (1) if this distribution covers multiple high-scoring DAGs, instead of being peaked at a single most likely graph, and (2) if DAG-GFlowNet can sample a variety of DAGs from the same Markov equivalence class (MEC; see \cref{sec:markov-equivalence}), showing the inherent uncertainty over equivalent graphs ($P(G\mid \gD)$ itself assigning the same probability to equivalent DAGs since BGe is an equivalent score, and the prior is uniform). In \cref{fig:dag-gfn-sachs}, we visualize the MECs of the graphs sampled with DAG-GFlowNet, along with other methods based on MCMC and variational inference. Each dot represents a MEC, and its size represents the number of unique DAGs in that MEC out of $1,000$ samples. The x-axis is the BGe score of the MEC (all equivalent DAGs sharing the same score), and the y-axis represents the marginal probability of the DAGs in that MEC (\ie the proportion of DAGs falling into the equivalence class).

We first observe that DAG-GFlowNet does not collapse to a single most-likely graph, and covers multiple MECs. Moreover, it is also capable of sampling different equivalent DAGs (larger dots), showing again that the distribution does not collapse to a single representative of the MECs with higher marginal probability. Its behavior is largely similar to methods based on MCMC (MC\textsuperscript{3} \& Gadget), and its maximal MEC reaches a high score, albeit lower than MC\textsuperscript{3} and DiBS; as a point of reference, the score of the best MEC obtained with GES \citep{chickering2002ges} is $-10,716.12$.

Moreover, despite comparing favorably in terms of $\E$-SHD \& AUROC in \cref{tab:dag-gfn-sachs-observational}, we can assess more precisely the quality of the posterior approximation given by BCD Nets \& DiBS in \cref{fig:dag-gfn-sachs}. The first observation is that out of $1,000$ samples from BCD Nets, the graphs belong to one of only two MECs (with a BGe score around $-10,850$ and a higher marginal probability). Furthermore, as shown by the size of each point, those MECs happen to only contain \emph{a single} unique DAG. Overall, this means that BCD Nets only returned $2$ unique DAGs, showing a clear lack of diversity. On the other hand, DiBS sampled a significant number of low-scoring DAGs (\ie having a very low probability under $P(G\mid \gD)$), with BGe scores as low as $-12,550$, which is not something any other method suffers from. We can also see that DiBS tends to return many graphs that belong to unique MECs (many unique points with a low marginal probability), but rarely multiple graphs from the same MEC. This shows that while DiBS benefits from a high diversity in terms of MECs (including ones covering low-scoring DAGs), it suffers from a lack of diversity within each MEC, which would be expected from a faithful approximation of the posterior distribution.

\subsection{Integration of experimental data}
\label{sec:dag-gfn-sachs-experimental-data}
An advantage of working with biological data is that we can acquire \emph{experimental data}. In addition to the observational data we used in the previous section, \citet{sachs2005causal} also provided flow cytometry data under different experimental conditions, where the T-cells were perturbed with some reagents; this effectively corresponds to \emph{interventional data} \citep{pearl2009causality}, which we saw in \cref{sec:causal-identifiability} is one possible ingredient for better causal identification. Although a molecular intervention may be imperfect and affect multiple proteins (\citealp{eaton2007belief}; \ie a molecule is used to ``knock out'' a protein, but this molecule may affect other proteins as well), we assume here that these interventions are perfect and the intervention targets are known. This time, we used a discretized dataset of $N = 5,400$ samples from $9$ experimental conditions, of which $6$ are interventions. Since the data is now discrete, we used a modification of the BDe score \citep{heckerman1995bde} to handle this mixture of observational and interventional data \citep{cooper1999interventional}. This is done by effectively ignoring the terms corresponding to the intervened node in \cref{eq:bde-score} (see also \cref{sec:interventions}).

\begin{table}[t]
    \centering
    \caption[Combining discrete interventional and observational flow cytometry data]{Combining discrete interventional and observational flow cytometry data. ${}^{\star}$Results reported in \citep{eaton2007belief}. All results include a 95\% confidence interval estimated with bootstrap resampling.}
    \label{tab:dag-gfn-sachs-interventional}
\begin{tabular}{lccc}
    \toprule
     & $\mathbb{E}$-\#~Edges & $\mathbb{E}$-SHD  & AUROC \\
    \midrule
    Exact posterior${}^{\star}$ & --- & --- & $\mathbf{0.816}$ \\
    MC\textsuperscript{3} & $25.97 \pm 0.01$ & $\mathbf{25.08 \pm 0.02}$ & $0.665$ \\
    \midrule
    DAG-GFlowNet & $30.66 \pm 0.04$ & $27.77 \pm 0.03$ & $0.700$ \\
    \bottomrule
\end{tabular}
\end{table}

In \cref{tab:dag-gfn-sachs-interventional}, we compare with the results reported by \citet{eaton2007belief}, where they computed the AUROC of the exact posterior $P(G\mid \gD)$ itself using dynamic programming (at the expense of computing only the edge marginals, and not the whole distribution over DAGs), therefore working as an upper bound for what a posterior approximation can achieve. We observe that MC\textsuperscript{3} predicts sparser graphs with a higher $\E$-SHD than DAG-GFlowNet, but with a lower AUROC. Note that this experimental setup is different from prior work which used continuous data instead \citep{brouillard2020differentiable,faria2022differentiable}, and was designed specifically to showcase the ability of DAG-GFlowNet to also work with discrete (and interventional) data.

%% file: chapters/08_JSP_GFN.tex
\chapter{Joint Inference of Structure and Parameters}
\label{chap:jsp-gfn}

\begin{minipage}{\textwidth}
    \itshape
    This chapter contains material from the following papers:
    \begin{itemize}[noitemsep, topsep=1ex, itemsep=1ex, leftmargin=3em]
        \item Mizu Nishikawa-Toomey, \textbf{Tristan Deleu}, Jithendaraa Subramanian, Yoshua Bengio, Laurent Charlin (2023). \emph{Bayesian learning of Causal Structure and Mechanisms with GFlowNets and Variational Bayes}. Graphs and More Complex Structures for Learning and Reasoning Workshop (AAAI). \notecite{nishikawa2023vbg}
        \item \textbf{Tristan Deleu}, Mizu Nishikawa-Toomey, Jithendaraa Subramanian, Nikolay Malkin, Laurent Charlin, Yoshua Bengio (2023). \emph{Joint Bayesian Inference of Graphical Structure and Parameters with a Single Generative Flow Network}. Advances in Neural Information Processing Systems (NeurIPS). \notecite{deleu2023jspgfn}
    \end{itemize}
    \vspace*{5em}
\end{minipage}

One of the major drawbacks of working with the (marginal) posterior distribution $P(G\mid \gD)$ over the structures alone is that it limits the scope of applications. Indeed, this assumes that the marginal likelihood $P(\gD\mid G)$ can be computed efficiently (oftentimes analytically), which is only possible for a handful of models that we studied in the previous chapter. This notably rules out many Bayesian networks whose conditional probability distributions are defined with non-linear functions (\eg neural networks), which is sometimes essential to model the intricacies of real-world systems (\eg in systems biology). In this chapter, we will build upon the foundations of \cref{chap:dag-gflownet} and present two different ways to model the posterior over entire Bayesian networks (\ie their structure \emph{and} parameters) for complete flexibility with a GFlowNet: in \cref{sec:vbg-variational-bayes-approach} we will introduce a direct extension of DAG-GFlowNet in conjunction with Variational Bayes, whereas in \cref{sec:joint-inference-single-gflownet} and onwards we will present a solution with \emph{a single} GFlowNet that can be trained end-to-end.

\section{Posterior distribution over graphs \& parameters}
\label{sec:posterior-distribution-graphs-parameters}\index{Structure learning!Bayesian structure learning}
\cref{chap:dag-gflownet} was entirely dedicated to modeling the \emph{marginal} posterior distribution $P(G\mid \gD)$; we call it ``marginal'' to highlight the fact that the parameters of the conditional distributions $\theta$ have been marginalized out. In this chapter, we will instead explicitly model the \emph{joint} posterior distribution $P(G, \theta\mid \gD)$ over both the structure $G$ and the parameters $\theta$ of a Bayesian network. Again with Bayes' rule, this joint posterior distribution is given by
\begin{equation}
    P(G, \theta\mid \gD) = \frac{P(\gD\mid G, \theta)P(\theta\mid G)P(G)}{P(\gD)},
    \label{eq:joint-posterior-dags-parameters}\index{Bayesian inference!Bayes' rule}
\end{equation}
where $P(\gD\mid G, \theta)$ is now the likelihood function (as opposed to the \emph{marginal} likelihood before), and $P(\theta\mid G)$ is a prior over parameters that we will consider as known and fixed (just like we did with the prior over graphs $P(G)$). The fact that we don't have to marginalize over $\theta$ anymore and we can use the likelihood directly makes it applicable to any Bayesian network: we simply have to evaluate $P(\gD\mid G, \theta)$ with \cref{eq:bayesian-network}. Modeling this joint posterior inherits many of the challenges that we mentioned in the previous chapter, namely that $G$ is still a member of a discrete and combinatorially large space of DAGs, making \cref{eq:joint-posterior-dags-parameters} intractable.

But it also comes with its own set of challenges. The first one being that the sample space is now a mixed space with a discrete component $G$ and a continuous one $\theta$. The second challenge is that the space of parameters \emph{depends on the structure of the Bayesian network}. To see this, consider two Bayesian networks over two binary random variables $A$ \& $B$, with structures $G_{1}: A \phantom{\rightarrow} B$ and $G_{2}: A \rightarrow B$. If we list the minimal set of parameters necessary to define both Bayesian networks
\begin{align}
    G_{1}&& P(A = 1) &= \theta_{1} & P(B = 1) &= \theta_{2}\\
    G_{2}&& P(A = 1) &= \theta_{1} & P(B = 1\mid A = 0) &= \theta_{2} & P(B = 1\mid A = 1) = \theta_{3}
\end{align}
all other conditional probabilities can be derived from those. We can see that $G_{1}$ has 2 parameters, whereas $G_{2}$ has 3. The space $\Theta_{G}$ which the parameters $\theta$ associated with a structure $G$ belong to not only changes depending on the DAG, but its dimensionality changes as well. This makes the sample space $\gX$ of $P(G, \theta\mid \gD)$ highly complex, and we will give a formal definition in \cref{sec:jsp-gfn-structure-gflownet}. This explains why joint Bayesian inference has received relatively little attention compared to its marginal counterpart.

\subsection{Reversible-jump MCMC}
\label{sec:rjmcmc}\index{Markov chain Monte Carlo}
We saw in \cref{sec:structure-mcmc} how we can build Markov chain over the space of DAGs with local moves $G \rightarrow G'$ corresponding to adding, deleting, or reversing a single edge. How can we expect building a similar Markov chain on the joint sample space, moving $(G, \theta) \rightarrow (G', \theta')$, when $\theta\in\Theta_{G}$ and $\theta'\in\Theta_{G'}$ may not even be elements of the same space, let alone with the same dimensionality? This is something which is impossible within the standard framework of MCMC. A naive solution could be to run independent chains \citep{chib1995jointmcmcindependentchains}, one for each DAG $G$, and to query the appropriate chain after any move in the space of DAGs. This is clearly not appropriate for Bayesian structure learning, as the number of possible structures is extremely large.

A more flexible and principled option is to allow \emph{trans-dimensional} moves, to move from any $\Theta_{G}$ to any other $\Theta_{G'}$ and back. \emph{Reversible-jump MCMC} methods (RJ-MCMC; \citealp{green1995rjmcmc}) is an extension of the Metropolis-Hastings scheme introduced in \cref{sec:existing-approaches-sampling-ebm}, but where the proposal is designed in such a way that moves between spaces of different dimensions are allowed, while preserving detailed balance (necessary to guarantee that the Markov chain will eventually converge to the target distribution, here $P(G, \theta\mid \gD)$). This has been used for Bayesian model selection more generally \citep{hastie2012modelselectionrjmcmc}, as well as Bayesian structure learning \citep{fronk2002rjmcmcdags}.

As a simple alternative that doesn't require the integration of complex trans-dimensional moves, we will consider in our experiments two extensions of structure MCMC (MC\textsuperscript{3}, described in \cref{sec:structure-mcmc}) studied by \citet{lorch2021dibs}. The first one is using a Metropolis-Hastings scheme to jointly sample structures and parameters together, where the parameters belong to a shared sample space and are appropriately ``masked out'' depending on the adjacency matrix of $G$; this is called \emph{Metropolis-Hastings MC\textsuperscript{3}} (M-MC\textsuperscript{3}). The second is a \emph{Metropolis-within-Gibbs} method (G-MC\textsuperscript{3}) that alternates between proposing updates to the graph and parameters \citep{godsill2001mcmcmodeluncertainty,castelletti2024jointposteriorrjmcmc}.

\subsection{Variational inference}
\label{sec:joint-posterior-variational-inference}
In this section, we give a more exhaustive comparison between various methods for Bayesian structure learning based on variational inference, for both joint inference and marginal inference, to complement \cref{sec:variational-inference-bayesian-structure-learning}. We give a comparison of various algorithms over multiple criteria in \cref{tab:comparison-variational-inference-gflownet}, where we also included methods based on GFlowNets since they are effectively variational methods as we saw in \cref{sec:gflownets-variational-inference} \citep{malkin2023gfnhvi,zimmermann2022vigfn}.

\begin{table}[t]
    \centering
    \caption[Comparison of variational inference \& GFlowNets for Bayesian structure learning] {Comparison of different methods based on variational inference and GFlowNets for Bayesian structure learning. See the text for a detailed description of each category. ${}^{\star}$Methods based on GFlowNets that are presented in this thesis.}
    \begin{adjustbox}{center}
    \begin{tabular}{clccccccc}
        \toprule
         && Joint & Non & DAG & Discrete & Max. & \multirow{2}{*}{Sampler} & Mini \\[-0.1em]
         && $G$ \& $\theta$ & Linear & Support & Obs. & Parents & & Batch \\
        \midrule
        \multirow{8}{*}{\rotatebox[origin=c]{90}{Variational inference}} & VCN \citep{annadani2021vcn} & \xcomp & \xcomp & \xcomp & \qcomp & \xcomp & \ccomp & \xcomp \\
        & BCD Nets \citep{cundy2021bcdnets} & \ccomp & \xcomp & \ccomp & \xcomp & \xcomp & \ccomp & \xcomp \\
        & DiBS \citep{lorch2021dibs} & \ccomp & \ccomp & \xcomp & \ccomp & \xcomp & \xcomp & \ccomp \\
        & \textsc{Trust} \citep{wang2022trust} & \xcomp & \xcomp & \ccomp & \qcomp & \xcomp & \qcomp & \xcomp \\
        & VI-DP-DAG \citep{charpentier2022differentiabledag} & \xcomp & \ccomp & \ccomp & \xcomp & \xcomp & \ccomp & \xcomp \\
        & AVICI \citep{lorch2022avici} & \xcomp & \ccomp & \xcomp & \qcomp & \xcomp & \ccomp & \xcomp \\
        & DPM-DAG \citep{rittel2023priorbayesiancsl} & \xcomp & \ccomp & \ccomp & \xcomp & \xcomp & \ccomp & \xcomp \\
        & BayesDAG \citep{annadani2023bayesdag} & \ccomp & \ccomp & \ccomp & \xcomp & \xcomp & \ccomp & \ccomp \\
        \midrule
        \multirow{4}{*}{\rotatebox[origin=c]{90}{GFlowNet}} & DAG-GFlowNet${}^{\star}$ \citep{deleu2022daggflownet} & \xcomp & \xcomp & \ccomp & \ccomp & \ccomp & \ccomp & \xcomp\\
        & DynGFN \citep{atanackovic2023dyngfn} & \qcomp & \xcomp & \qcomp & \qcomp & \ccomp & \ccomp & \ccomp\\
        & VBG${}^{\star}$ \citep{nishikawa2023vbg} & \ccomp & \qcomp & \ccomp & \qcomp & \ccomp & \ccomp & \xcomp\\
        & JSP-GFN${}^{\star}$ \citep{deleu2023jspgfn} & \ccomp & \ccomp & \ccomp & \ccomp & \ccomp & \ccomp & \ccomp\\
        \bottomrule
    \end{tabular}
    \end{adjustbox}
    \label{tab:comparison-variational-inference-gflownet}
\end{table}

\paragraph{Joint $\bm{G}$ \& $\bm{\theta}$} This category, of main interest in the context of this chapter, indicates whether the model can approximate the joint posterior $P(G, \theta\mid \gD)$, or if they are limited to approximating the marginal posterior $P(G\mid \gD)$. Both methods we presented in \cref{sec:variational-inference-bayesian-structure-learning} can also be adapted to model the joint posterior. In the case of BCD Nets \citep{cundy2021bcdnets}, or another method based on the ordering of the random variables called BayesDAG \citep{annadani2023bayesdag}, the model readily parametrizes a weighted adjacency matrix, whose non-zero elements encode the structure of the DAG, but their exact values also encode the parameters of a linear-Gaussian model. But this is not necessarily the case of other methods falling into this same family as well (\eg VI-DP-DAG, or DPM-DAG). In the case of DiBS \citep{lorch2021dibs}, which uses a continuous relaxation of acyclicity as a soft prior, the posterior approximation can incorporate the parameters into the joint distribution $Q(Z, \theta)$ together with the latent variables $Z$ used to determine the graphs downstream. On the other hand, another method also based on a continuous relaxation of acyclicity called VCN \citep{annadani2021vcn} doesn't include the parameters of the Bayesian network.

For methods based on GFlowNets, we saw in the previous chapter that DAG-GFlowNet was only capable of modeling the marginal posterior. We will see in this chapter two alternatives that do model the joint posterior: VBG \citep{nishikawa2023vbg} that we will introduce in \cref{sec:vbg-variational-bayes-approach}, and JSP-GFN \citep{deleu2023jspgfn} that we will introduce in the rest of this chapter. In the case of DynGFN \citep{atanackovic2023dyngfn}, which is an adaptation of DAG-GFlowNet to learn the structure of \emph{dynamic Bayesian networks} \citep{friedman1998structurelearningdbn,murphy2002dbn}, the parameters are estimated with maximum likelihood instead of having a completely Bayesian treatment.

\paragraph{Non-linear} This category indicates whether the model can be applied to Bayesian networks whose conditional probability distributions are parametrized by non-linear functions (\eg neural networks). While most methods approximating $P(G\mid \gD)$ may be applied to cases where the conditional probabilities are parametrized by Gaussian Processes (allowing for an explicit marginalization; \citealp{vonkugelgen2019gpstructure}), we only consider here methods that handle any non-linearity by design. It is natural to find models approximating $P(G, \theta\mid \gD)$ also being capable of working with non-linear conditional probability distributions, with the notable exception of BCD Nets which has an explicit ``linear-Gaussian'' assumption. VI-DP-DAG \citep{charpentier2022differentiabledag}, and DPM-DAG \citep{rittel2023priorbayesiancsl} which is based upon it, use maximum likelihood in order to estimate a point estimate of the parameters of a non-linear function. Finally even though VBG can be applied to non-linear cases, in practice \citet{nishikawa2023vbg} only studied linear-Gaussian models.

\paragraph{DAG support} This category indicates whether the posterior approximation is guaranteed to have support over the space of DAGs only. VCN and DiBS only encourage acyclicity via a prior term, meaning that those methods may return graphs containing cycles. AVICI \citep{lorch2022avici} uses a similar prior term in some cases, but in general this framework does not seek to enforce acyclicity by design (\eg to model feedback loop in certain domains, such as biological systems; \citealp{sethuraman2023nodagsflow}). \textsc{Trust} \citep{wang2022trust}, just like all other methods based on a variable ordering, guarantees acyclicity via a distribution over the variable orders that can be learned using sum-product networks \citep{poon2011sumproductnetwork}. On the other hand, methods based on GFlowNets are guaranteed to have a support over the space of DAGs by design; DynGFN considers dynamic Bayesian networks (without instantaneous edges), which are necessarily acyclic.

\paragraph{Discrete observations} This category indicates whether the posterior approximation may be applied to Bayesian networks with discrete random variables. Although a number of models make the explicit assumption to have a Bayesian network over continuous random variables (sometimes even requiring the variables to be normally distributed), some methods may in principle be applied to discrete Bayesian networks. The authors of VCN mention that their approach is also applicable to discrete observations. While there is no experiment in the discrete setting in \citep{lorch2021dibs}, this extension can be found in the official code release for DiBS; this also transfers to \textsc{Trust}, which may use DiBS as its underlying routine for structure learning.

\paragraph{Maximum parents} This category indicates whether a maximum number of parents can be specified for each variable. Although this is a very common constraint in the structure learning literature to improve efficiency \citep{koller2009pgm}, none of the variational methods for Bayesian structure learning allow for such a (hard) constraint. Some methods may introduce a sparsity-inducing prior \citep{annadani2021vcn,cundy2021bcdnets,lorch2021dibs,annadani2023bayesdag}, or use post-processing of the sampled DAGs \citep{charpentier2022differentiabledag} to reduce the number of edges. Conversely, we saw in \cref{sec:additional-structural-constraints} that this type of additional structural constraints can be naturally added to any GFlowNet over DAGs.

\paragraph{Sampler} The category indicates whether one can freely sample from the model once it is fully trained. While this seems like a basic requirement for Bayesian inference, a notable exception is DiBS which uses a particle-based approach \citep{liu2016svgd} to approximate the posterior (be it marginal, or joint), and therefore the number of particles is fixed ahead of time; once fully trained, it is impossible to sample new Bayesian networks $(G, \theta)$ from this model anymore (distinct from the particles themselves).

\paragraph{Mini-batch} This category indicates whether the model can be updated with mini-batches of observations from $\gD$, instead of the the full dataset. With only a few exceptions, most methods for Bayesian structure learning necessitate the whole dataset $\gD$ to learn the posterior approximation. DiBS uses mini-batch updates for their experiments on protein signaling networks, where the number of observations is large ($N = 7,466$). We will see in \cref{sec:jsp-gfn-minibatch-training} that JSP-GFN can naturally leverage mini-batch training, while DAG-GFlowNet couldn't (nor can VBG for that matter).

\section{Variational Bayes approach}
\label{sec:vbg-variational-bayes-approach}\index{Variational Bayes!Variational Bayes DAG-GFlowNet}
Prior to our work extending generative flow networks to more general state spaces \citep{lahlou2023continuousgfn} that we presented in \cref{chap:gflownets-general-state-spaces}, the only way to model the joint posterior $P(G, \theta\mid \gD)$ with a GFlowNet (to leverage the tools developed in the previous chapter) was to treat the discrete component $G$ and the continuous one $\theta$ separately. We will see in this section how we can use \emph{Variational Bayes}, that was introduced in \cref{sec:variational-bayes}, in conjunction with GFlowNets in order to approximate the joint posterior distribution. Since it takes inspiration from DAG-GFlowNet, we call this method \emph{Variational Bayes-DAG-GFlowNet} (VBG; \citealp{nishikawa2023vbg}).

\subsection{Evidence lower bound}
\label{sec:vbg-elbo}
To treat the graphs and parameters separately, we will approximate the joint posterior distribution $P(G, \theta \mid \gD)$ as the product of two distribution: a distribution $Q_{\phi}(G)$ over DAGs, parametrized by $\phi$, and a distribution $Q_{\psi}(\theta\mid G)$, parametrized by $\psi$
\begin{equation}
    P(G, \theta\mid \gD) \approx Q_{\phi}(G)Q_{\psi}(\theta\mid G).
\end{equation}
Our objective will be to find the best approximation of the posterior (\ie the best parameters $(\phi^{\star}, \psi^{\star})$). We saw in \cref{sec:variational-bayes} that Variational Bayes approaches this problem by coordinate ascent on an \emph{evidence lower bound} (ELBO) of the log-evidence $\log P(\gD)$. The following proposition gives an explicit form for this ELBO.

\begin{proposition}[ELBO]
    \label{prop:vbg-elbo}
    Let $Q_{\phi}(G)$ \& $Q_{\psi}(\theta\mid G)$ be two distributions such that their product is an approximation of the joint posterior distribution $P(G, \theta\mid \gD) \approx Q_{\phi}(G)Q_{\psi}(\theta\mid G)$. Then the \emph{evidence lower bound} $\gJ(\phi, \psi)$ for this problem (\ie $\log P(\gD) \geq \gJ(\phi, \psi)$) can be written as
    \begin{equation}
        \gJ(\phi, \psi) = \E_{G\sim Q_{\phi}}\left[\E_{\theta\sim Q_{\psi}}\big[\log P(\gD\mid G, \theta)\big] - \KL\big(Q_{\psi}(\theta\mid G)\,\|\,P(\theta\mid G)\big)\right] - \KL\big(Q_{\phi}(G)\,\|\,P(G)\big).
        \label{eq:vbg-elbo}
    \end{equation}
\end{proposition}

\begin{proof}
    We start with writing the joint distribution between the data, the graph and parameters:
    \begin{equation}
        P(\gD, \theta, G) = P(\gD\mid G, \theta)P(\theta\mid G)P(G).
        \label{eq:vbg-elbo-proof-1}
    \end{equation}
    The derivation of the ELBO follows the standard steps in variational inference (as in \cref{sec:inference-as-optimization}), except both the joint distribution $P(\gD, \theta, G)$ and the posterior approximation are more structured.
    {\allowdisplaybreaks%
    \begin{align}
        \log &P(\gD) = \log\left(\sum_{G}\int_{\theta}P(\gD, \theta, G)d\theta\right)\\
        &= \log \left(\sum_{G}\left[\int_{\theta}\frac{P(\gD, \theta, G)}{Q_{\phi}(G)Q_{\psi}(\theta\mid G)}Q_{\psi}(\theta\mid G)d\theta\right]Q_{\phi}(G)\right)\\
        &\geq \sum_{G}\left[\int_{\theta}\log \frac{P(\gD, \theta, G)}{Q_{\phi}(G)Q_{\psi}(\theta\mid G)}Q_{\psi}(\theta\mid G)d\theta\right]Q_{\phi}(G)\label{eq:vbg-elbo-proof-2}\\
        &= \sum_{G}\left[\int_{\theta}\log \frac{P(\gD\mid G, \theta)P(\theta\mid G)P(G)}{Q_{\phi}(G)Q_{\psi}(\theta\mid G)}Q_{\psi}(\theta\mid G)d\theta\right]Q_{\phi}(G)\label{eq:vbg-elbo-proof-3}\\
        &= \sum_{G}\left[\int_{\theta} \log P(\gD\mid G, \theta)Q_{\psi}(\theta\mid G)d\theta + \int_{\theta}\log \frac{P(\theta\mid G)}{Q_{\psi}(\theta\mid G)}Q_{\psi}(\theta\mid G)d\theta + \log \frac{P(G)}{Q_{\phi}(G)}\right]Q_{\phi}(G)\nonumber\\
        &= \E_{G\sim Q_{\phi}}\left[\E_{\theta\sim Q_{\psi}}\big[\log P(\gD\mid G, \theta)\big] - \KL\big(Q_{\psi}(\theta\mid G)\,\|\,P(\theta\mid G)\big)\right] - \KL\big(Q_{\phi}(G)\,\|\,P(G)\big),
    \end{align}}%
    where we used Jensen's inequality in \cref{eq:vbg-elbo-proof-2}, and the decomposition of the joint distribution \cref{eq:vbg-elbo-proof-1} in \cref{eq:vbg-elbo-proof-3}. This concludes the proof.
\end{proof}
Although we didn't give any detail on how these distributions are parametrized exactly, we will see in the next section that $Q_{\phi}(G)$ in particular will be parametrized with a GFlowNet. On the other hand, the distribution $Q_{\psi}(\theta\mid G)$ will be problem specific, depending on the form of the conditional distributions appearing in the Bayesian networks we are modeling.

\subsection{Modeling the distribution over DAGs with a GFlowNet}
\label{sec:vbg-distribution-dags-gflownet}
Since $Q_{\phi}(G)$ is a distribution over DAGs, it is natural to model it as a GFlowNet, similar to the one we presented in \cref{sec:gflownet-over-dags}. While this informs us about how the GFlowNet is structured (\cref{sec:structure-dag-gfn}) and the loss being used (\cref{sec:modified-detailed-balance}), we still need to specify its reward function. Recall that a GFlowNet models a target distribution which is proportional to that reward function. The question then becomes: what is this target distribution here? Given that Variational Bayes operates by coordinate ascent, the target distribution for a fixed $Q_{\psi}(\theta\mid G)$ is the one maximizing the ELBO $\gJ(\phi, \psi)$ we saw in the previous section.
\begin{proposition}
    \label{prop:vbg-maximization-graphs}
    For a fixed distribution $Q_{\psi}(\theta\mid G)$, the distribution $Q_{\phi}^{\star}(G)$ maximizing the ELBO $\gJ(\phi, \psi)$ in \cref{prop:vbg-elbo} satisfies for any DAG $G$
    \begin{equation}
        \log Q_{\phi}^{\star}(G) = \E_{\theta\sim Q_{\psi}}\big[\log P(\gD\mid G, \theta)\big] - \KL\big(Q_{\psi}(\theta\mid G)\,\|\,P(\theta\mid G)\big) + \log P(G) + C,
    \end{equation}
    where $C$ is a constant independent of $G$.
\end{proposition}

\begin{proof}
    There is a finite number of DAGs (albeit possibly extremely large) one can create with $d$ nodes. We call them $\{G_{m}\}_{m=1}^{M}$. Maximizing the ELBO wrt.~the distribution $Q_{\phi}(G)$ is equivalent to solving the following constrained optimization problem
    \begin{align}
        \max_{\{Q_{\phi}(G_{m})\}_{m=1}^{M}} & \gJ(Q_{\phi}, Q_{\psi})\\[-0.5em]
        \textrm{s.t.}\ & \sum_{m=1}^{M}Q_{\phi}(G_{m}) = 1,
    \end{align}
    where we make the dependence of the ELBO on the distributions $Q_{\phi}$ \& $Q_{\psi}$ more explicit. The constraint only guarantees that $Q_{\phi}(G)$ is a valid distribution over DAGs; there should be an additional non-negative constraint, which we ignore here for simplicity. To find solutions of this optimization problem, we an look at critical points of the Lagrangian, defined as
    {\allowdisplaybreaks%
    \begin{align}
        L&\big(\{Q_{\phi}(G_{m})\}_{m=1}^{M}, \lambda) = \gJ(Q_{\phi}, Q_{\psi}) + \lambda\left(\sum_{m=1}^{M}Q_{\phi}(G_{m}) - 1\right)\\
        &= \sum_{m=1}^{M}\Bigg[\E_{\theta\sim Q_{\psi}}\big[\log P(\gD\mid G_{m}, \theta)\big] - \KL\big(Q_{\psi}(\theta\mid G_{m})\,\|\,P(\theta\mid G_{m})\big) + \log \frac{P(G_{m})}{Q_{\phi}(G_{m})} + \lambda\Bigg]Q_{\phi}(G_{m}) - \lambda.
    \end{align}}%
    Taking the derivative of the Lagrangian with respect to a particular $Q_{\phi}(G_{m})$, and equating it to zero to find the critical point, we get
    {\allowdisplaybreaks%
    \begin{align}
        \frac{\partial L}{\partial Q_{\phi}(G_{m})} &= \E_{\theta\sim Q_{\psi}}\big[\log P(\gD\mid G_{m}, \theta)\big] - \KL\big(Q_{\psi}(\theta\mid G_{m})\,\|\,P(\theta\mid G_{m})\big)\nonumber\\&\qquad \qquad + \log P(G_{m}) - \log Q_{m}^{\star}(G_{m}) - 1 + \lambda = 0\\
        \Leftrightarrow \qquad \log Q_{\phi}^{\star}(G_{m}) &= \E_{\theta\sim Q_{\psi}}\big[\log P(\gD\mid G_{m}, \theta)\big] - \KL\big(Q_{\psi}(\theta\mid G_{m})\,\|\,P(\theta\mid G_{m})\big) + \log P(G_{m}) + C,
    \end{align}}%
    where $C = \lambda - 1$ is a constant independent of $G_{m}$.
\end{proof}
The proposition above shows that the optimal $Q_{\phi}^{\star}(G)$ for a fixed $Q_{\psi}(\theta\mid G)$ is a distribution over DAGs defined up to a normalization constant. This is perfect application for generative flow networks, as we saw in \cref{chap:dag-gflownet}. Unlike DAG-GFlowNet though, where the reward of the GFlowNet was fixed to $R(G) = P(\gD\mid G)P(G)$ \citep{deleu2022daggflownet}, the reward here is defined by
\begin{equation}
    \log \widetilde{R}(G) = \E_{\theta\sim Q_{\psi}}\big[\log P(\gD\mid G, \theta)\big] - \KL\big(Q_{\psi}(\theta\mid G)\,\|\,P(\theta\mid G)\big) + \log P(G).
    \label{eq:reward-gflownet-vbg}
\end{equation}
The key difference with DAG-GFlowNet is that the reward $\widetilde{R}(G)$ now depends on $Q_{\psi}(\theta\mid G)$ that may evolve (the reward function remains fixed while updating the GFlowNet though). In a different context, this relates to EB-GFN \citep{zhang2022ebgfn} and GFN-EM \citep{hu2023gfnem} where an iterative procedure is used to update both a GFlowNet and the reward function, instead of the latter being fixed. Interestingly, if the distribution $Q_{\psi}(\theta\mid G) = P(\theta\mid G, \gD)$ matches the posterior distribution over parameters, given a DAG $G$, then the reward above simply reduces to the reward we used in DAG-GFlowNet
\begin{align}
    \log \widetilde{R}(G) &= \E_{\theta\sim P(\theta\mid G, \gD)}\big[\log P(\gD\mid G, \theta)\big] - \KL\big(P(\theta\mid G, \gD)\,\|\,P(\theta\mid G)\big) + \log P(G)\\
    &= \int_{\theta}\log \frac{P(\gD\mid G, \theta)P(\theta\mid G)}{P(\theta\mid G, \gD)}P(\theta\mid G, \gD)d\theta + \log P(G)\\
    &= \int_{\theta}\log P(\gD\mid G)P(\theta\mid G, \gD)d\theta + \log P(G)\\
    &= \log P(\gD\mid G) + \log P(G) = \log R(G).
\end{align}

Moreover, if we make the additional assumption that the distribution $Q_{\psi}(\theta)$ also decomposes along the parameters $\theta_{i}$ of the conditional distributions of the Bayesian network
\begin{equation}
    Q_{\psi}(\theta) = \prod_{i=1}^{d}Q_{\psi}(\theta_{i}),
\end{equation}
then we can show that the delta-score $\log \widetilde{R}(G') - \log \widetilde{R}(G)$ appearing in the modified detailed balance loss for a transition $G \rightarrow G'$ in the GFlowNet (\cref{sec:dag-gfn-modularity-computational-efficiency}) can also be easily computed without having to evaluate the whole log-reward of either graphs \citep{nishikawa2023vbg}.

Following the algorithm in \cref{sec:variational-bayes}, we can find the posterior approximation by interlacing the following steps: (1) for a fixed $Q_{\psi}(\theta\mid G)$, fit the parameters $\phi$ of a GFlowNet over DAGs as described in \cref{sec:gflownet-over-dags}, with a reward function $\widetilde{R}(G)$ defined in \cref{eq:reward-gflownet-vbg}, and (2) for a fixed $Q_{\phi}(G)$, maximize the ELBO in \cref{prop:vbg-elbo} wrt.~$\psi$; see also \cref{alg:vbg-training}.

\begin{algorithm}[t]
    \caption{Training of Variational Bayes-DAG-GFlowNet (VBG).}
    \label{alg:vbg-training}
    \begin{algorithmic}[1]
        \Require A dataset $\gD$, and a model $P(\gD, \theta, G) = P(\gD\mid G, \theta)P(\theta\mid G)P(G)$.
        \Ensure An approximation of the joint posterior $P(G, \theta\mid \gD) \approx Q_{\phi}(G)Q_{\psi}(\theta\mid G)$.
        \State Initialization of the parameters $(\phi, \psi)$
        \Loop
            \Repeat
                \State Update the parameters $\phi$ of the GFlowNet (\cref{sec:gflownet-over-dags}) with reward $\widetilde{R}(G)$ \Comment{\cref{eq:reward-gflownet-vbg}}
            \Until{convergence (\eg $\gL_{\mdb}(\phi_{t+1}) - \gL_{\mdb}(\phi_{t}) < \varepsilon$)}
            \State Sample graphs from $Q_{\phi}(G)$ modeled by the GFlowNet
            \State Update $\psi$ to maximize $\gJ(\phi, \psi)$ \Comment{\cref{prop:vbg-elbo}; gradient updates, or analytical}
        \EndLoop
        \State \Return $Q_{\phi}(G)Q_{\psi}(\theta\mid G)$
    \end{algorithmic}
\end{algorithm}

\subsection{Update of the posterior approximation over parameters}
\label{sec:vbg-update-posterior-approximation-parameters}
The last ingredient for Variational Bayes is to determine how to parametrize $Q_{\psi}(\theta\mid G)$, and how to update its parameters $\psi$. Since $\theta$ represents the parameters of a Bayesian network (continuous quantity), the posterior approximation will typically be represented as a Normal distribution, following \citet{kingma2013vae,rezende2014dlgm}
\begin{equation}
    Q_{\psi}(\theta_{i}\mid G) = \gN\big(\theta_{i}\mid \mu_{\psi}^{(i)}(G), \Sigma_{\psi}^{(i)}(G)\big),
\end{equation}
where $\mu_{\psi}^{(i)}(G)$ and $\Sigma_{\psi}^{(i)}(G)$ are functions that depend on the graph $G$. To maximize the ELBO wrt.~$\psi$, we can then apply gradient ascent. Since $Q_{\psi}(\theta\mid G)$ appears as the distribution wrt.~which we take an expectation in \cref{eq:vbg-elbo}, we can use the reparametrization trick (\cref{sec:estimation-parameters}) to estimate the gradient of $\gJ(\phi, \psi)$. In some special cases, such as linear-Gaussian Bayesian networks, we can show that this maximization of $\gJ(\phi, \psi)$ wrt.~$\phi$ can be done analytically \citep{nishikawa2023vbg}, and does not require gradient updates.

\section{JSP-GFN: Joint inference with a single GFlowNet}
\label{sec:joint-inference-single-gflownet}\index{Structure learning!Bayesian structure learning}\index{JSP-GFN}
Instead of taking directly DAG-GFlowNet and working around it to incorporate a distribution over parameters with variational Bayes as we did in VBG in the previous section, we can fully leverage the results of \cref{chap:gflownets-general-state-spaces} and define \emph{a single} GFlowNet over a mixed discrete/continuous space capable of doing joint inference over the structure $G$ and the parameters $\theta$. We call this model \emph{JSP-GFN}, for Joint Structure and Parameters Bayesian inference with a GFlowNet.

\subsection{Structure of the GFlowNet}
\label{sec:jsp-gfn-structure-gflownet}
\begin{figure}[t]
    \centering
    \begin{adjustbox}{center}
        \includegraphics[width=480pt]{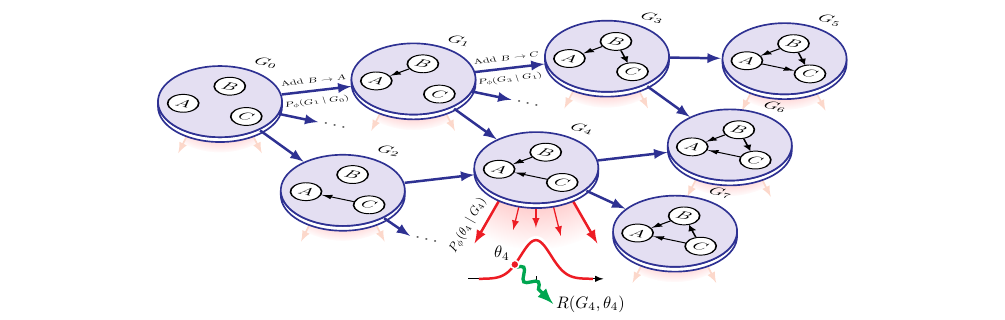}
    \end{adjustbox}
    \caption[Structure of JSP-GFN]{Structure of JSP-GFN to approximate the joint posterior distribution $P(G, \theta\mid \gD)$. A graph $G$ and the parameters $\theta$ are constructed as follows: starting from the empty graph $G_{0}$, (1) the graph $G$ is first generated one edge at a time (blue), as in \citep{deleu2022daggflownet}. Then once we select the ``$\stopaction$'' action indicating that we stop adding edges to the graph, (2) we generate the parameters $\theta$ (red), conditioned on the graph $G$.  Finally given $G$ and $\theta$, (3) we receive a reward $R(G, \theta)$ (green).}
    \label{fig:jsp-gfn}
\end{figure}

Unlike in \cref{chap:dag-gflownet}, where we model a distribution only over DAGs, here we need to define a GFlowNet whose terminating states are pairs $(G, \theta)$, where $G$ is a DAG and $\theta$ is a set of (continuous-valued) parameters whose dimension may depend on $G$. These terminating states are obtained through two phases, shown in \cref{fig:jsp-gfn}: (1) the DAG $G$ is first constructed one edge at a time, following our description in \cref{sec:structure-dag-gfn} \citep{deleu2022daggflownet}, and then (2) the corresponding parameters $\theta$ are generated, conditioned on $G$. We denote by $(G, \ast)$ states where the DAG $G$ has no parameters associated to it yet (\ie states in blue in \cref{fig:jsp-gfn}); they are intermediate states during the first phase of the GFlowNet, and do not correspond to valid samples. Using our notations of \cref{sec:elements-graph-theory}, $(G, \theta) \in \gX$, whereas $(G, \ast) \in \gS\backslash \gX$.

Starting at the empty graph $(G_{0}, \ast)$, the DAG is constructed one edge at a time during the first phase, following the forward transition probabilities $P_{F}(G'\mid G)$; the same considerations as in \cref{sec:efficient-verification-valid-actions} also hold here, namely that the edges added must not be already present nor introduce a cycle. This first phase ends when a special ``$\stopaction$'' action is selected with $P_{F}$, indicating that we stop adding edges to the graph and we are entering the second phase. Then during the second phase, we generate $\theta$ conditioned on $G$, following the forward transition probabilities $P_{F}(\theta\mid G)$. For a fixed $G$, all the terminating states $(G, \theta)$ can be seen as forming an (infinitely-wide) tree rooted at $(G, \ast)$.

Note that since the states of the GFlowNet are really \emph{pairs} of objects, $P_{F}(\theta\mid G)$ (resp.~$P_{F}(G'\mid G)$) is an abuse of notations, and actually represents $P_{F}\big((G, \theta)\mid (G, \ast)\big)$ (resp.~$P_{F}\big((G', \ast)\mid (G, \ast)\big)$). In particular, it is important to see that $P_{F}(\theta\mid G)$ is \emph{not} a distribution over $\theta$, in the sense that it is not normalized to $1$, since
\begin{equation}
    \int_{\theta}P_{F}(\theta\mid G)d\theta + \sum_{G'\in\underline{\children}_{\gG}(G)}P_{F}(G'\mid G) = 1,
\end{equation}
where $\underline{\children}_{\gG}(G)$ corresponds to the children states of $G$ during the first phase only (\ie children DAGs $G'$ being the result of adding a single edge to $G$).

\paragraph{Reward function} Since our objective is to approximate the joint posterior distribution $P(G, \theta\mid \gD) \propto P(\gD, \theta, G)$, it is natural to define the reward function of a terminating state $(G, \theta)$ as
\begin{equation}
    R(G, \theta) = P(\gD\mid G, \theta)P(\theta\mid G)P(G),
    \label{eq:reward-function-jsp-gfn}
\end{equation}
where the likelihood model $P(\gD\mid G, \theta)$ can be completely arbitrary now (\eg with conditional distributions parametrized by neural networks), $P(\theta\mid G)$ is the prior over parameters, and $P(G)$ is the prior over graphs.

\paragraph{Measurable pointed graph perspective} While the description of the GFlowNet's structure above is perfectly intuitive, we saw in \cref{sec:practical-implementation-generalized-gflownets} that some care must be taken when it comes to defining GFlowNets over general state spaces. This includes in particular mixed spaces with discrete and continuous components, as is the case here. Recall that for some DAG structure $G$, the space of parameters associated with $G$ is denoted by $\Theta_{G}$. In particular, we saw in \cref{sec:rjmcmc} that the dimensionality of $\Theta_{G}$ may depend on $G$. For example in a linear-Gaussian model, the space of parameters is $\Theta_{G} \simeq \sR^{K}$, where $K$ is the number of edges in $G$.

With this, we can formally define the state space $\gS$ and the space of terminating states $\gX$ as being the finite union over all the DAGs over $d$ nodes of product spaces
\begin{align}
    \gS &= \bigcup_{G\in\mathrm{DAG}}\{G\}\times \widebar{\Theta}_{G} &&\textrm{and} & \gX &= \bigcup_{G\in\mathrm{DAG}}\{G\} \times \Theta_{G},
\end{align}
where $\widebar{\Theta}_{G} = \Theta_{G}\cup \{\ast\}$ is the space of parameters, augmented with the symbol $\ast$ indicating that we are still in the first phase.

We now have to define the structure of the GFlowNet, in terms of a forward and backward reference kernels. For a DAG $G$, the measure $\kappa_{F}\big((G, \ast), \cdot\big)$ is the sum of a discrete measure (to transition to another intermediate state $(G', \ast)$ during the first phase) and a continuous measure (to transition to a terminating state $(G, \theta)$ in the second phase). We can write this measure as
\begin{equation}
    \kappa_{F}\big((G, \ast), \cdot\big) = \sum_{G'\in\underline{\children}_{\gG}(G)}\delta_{(G', \ast)}(\cdot) + \big(\delta_{G} \otimes \lambda_{\Theta_{G}}\big)(\cdot),
\end{equation}
where $\delta$ is the Dirac measure, and $\lambda_{\Theta_{G}}$ is the Lebesgue measure over $\Theta_{G}$. Moreover, we also have $\kappa_{F}\big((G, \theta), \cdot\big) = \delta_{\terminal}(\cdot)$ for all the terminating states $(G, \theta) \in \gX$; in other words, there is no transition from a terminating state other than to the terminal state $\terminal$. This clearly shows in particular that the measurable pointed graph will be finitely absorbing \cref{eq:finitely-absorbing}, with a maximum length $d(d-1)/2 + 1$. Unlike RJ-MCMC, this approach does not require trans-dimensional steps from $\Theta_{G}$ to another $\Theta_{G'}$ because once we select a graph $G$ during the first phase, we are committing to that space $\Theta_{G}$; to get another sample, we need to start again the process from scratch. On the other hand, the backward reference kernel is always discrete, regardless of the state
\begin{align}
    \kappa_{B}\big((G, \theta), \cdot\big) &= \delta_{(G, \ast)}(\cdot) && \textrm{and} & \kappa_{B}\big((G', \ast), \cdot\big) &= \sum_{G\in\parents_{\gG}(G)}\delta_{(G, \ast)}(\cdot).
\end{align}
We can observe that $(G, \theta)$ only has one ``parent'', which will be useful in the next section to simplify the flow matching condition we will use here. Finally, even though the reward and ``forward transition probabilities'' should really be measures ($R(G, d\theta)$ and $P_{F}(d\theta\mid G)$), we will only work with their Radon-Nikodym derivatives throughout this chapter, without mentioning it again.

\subsection{Sub-trajectory balance condition}
\label{sec:jsp-gfn-sub-trajectory-balance-condition}
To obtain a generative process that samples pairs $(G, \theta)$ proportionally to the reward, the GFlowNet needs to satisfy some flow matching condition, such as the ones described in \cref{sec:flow-matching-conditions}. Unlike DAG-GFlowNet though, we can't use the modified detailed balance of \cref{prop:modified-detailed-balance} directly as not all the states are terminating here. But we can take inspiration from it, and consider a condition between two terminating states $(G, \theta)$ \& $(G', \theta)$ instead, where $G'$ is the result of adding an edge to $G$. We can achieve this with the \emph{sub-trajectory balance condition} we presented in \cref{sec:sub-trajectory-balance-condition} \citep{madan2022subtb,malkin2022trajectorybalance}, and in particular with its generalization to undirected paths going ``backward then forward'' in \cref{eq:generalized-sub-trajectory-balance-condition}. For an undirected path of length 3 of the form $(G, \theta) \leftarrow (G, \ast) \rightarrow (G', \ast) \rightarrow (G', \theta')$ (illustrated in red in \cref{fig:sub-trajectory-balance}), we can rewrite \cref{eq:generalized-sub-trajectory-balance-condition} as
\begin{equation}
    F(G, \theta)P_{B}(G\mid \theta)P_{F}(G'\mid G)P_{F}(\theta'\mid G') = F(G', \theta')P_{B}(G'\mid \theta')P_{B}(G\mid G')P_{F}(\theta\mid G),
    \label{eq:sub-tb-3-raw-jsp-gfn}\index{Sub-trajectory balance!Condition}
\end{equation}
where we used a similar abuse of notation $P_{B}(G\mid \theta)$ to denote $P_{B}\big((G, \ast)\mid (G, \theta)\big)$, and $F(G, \theta)$ is a state flow function. In fact, since the terminating state $(G, \theta)\in\gX$ has only a single parent state $(G, \ast)$, we necessarily have $P_{B}(G\mid \theta) = 1$ (and similarly for $(G', \theta')$). Furthermore, we can use the same observation we made in \cref{eq:modified-detailed-balance-proof-1} in the proof of \cref{prop:modified-detailed-balance} to write the state flow $F(G, \theta)$ at a terminating state as a function of its reward:
\begin{equation}
    F(G, \theta) = \frac{R(G, \theta)}{P_{F}\big(\terminal \mid (G, \theta)\big)}.
    \label{eq:sub-tb-3-state-flow-reward}
\end{equation}
We can further simplify this equation by observing that in the GFlowNet described in the previous section, $\terminal$ is the \emph{only} child of the terminating state $(G, \theta)$ and therefore we necessarily have $P_{F}\big(\terminal\mid (G, \theta)\big) = 1$ (hence $F(G, \theta) = R(G, \theta)$). With all these simplifications, the sub-trajectory balance condition \cref{eq:sub-tb-3-raw-jsp-gfn} can be written as
\begin{equation}
    R(G', \theta')P_{B}(G\mid G')P_{F}(\theta\mid G) = R(G, \theta)P_{F}(G'\mid G)P_{F}(\theta'\mid G').
    \label{eq:sub-trajectory-balance-condition-jsp-gfn}
\end{equation}
This particular sub-trajectory balance condition happens to be very similar to the modified detailed balance we used in DAG-GFlowNet (\cref{prop:modified-detailed-balance}), where the probability of terminating $P_{F}(\terminal \mid G)$ has been replaced by $P_{F}(\theta\mid G)$, in addition to the reward now depending on both $G$ \& $\theta$. This is not a coincidence, as we will see later in \cref{sec:jsp-gfn-dag-gfn-special-case}.

The downside of using the sub-trajectory balance condition (instead of the detailed balance condition, as a basis for \cref{prop:modified-detailed-balance}) is that we saw in \cref{sec:sub-trajectory-balance-condition} that it was not sufficient to define a valid Markovian flow. Despite having no guarantee in general that satisfying the sub-trajectory balance conditions would yield a flow (let alone a distribution proportional to the reward), the following theorem shows that this GFlowNet does induce a distribution $\propto R(G, \theta)$ if the sub-trajectory balance condition in \cref{eq:sub-trajectory-balance-condition-jsp-gfn} is satisfied for all undirected paths $(G, \theta) \leftarrow (G, \ast) \rightarrow (G', \ast) \rightarrow (G', \theta')$.

\begin{theorem}[Sub-trajectory balance condition]
    \label{thm:sub-trajectory-balance-condition-jsp-gfn}\index{Sub-trajectory balance!Condition}
    Let $\gG$ be the GFlowNet described in \cref{sec:jsp-gfn-structure-gflownet}, with a reward function $R$, a forward transition probability $P_{F}$, and a backward transition probability $P_{B}$. If the following \emph{sub-trajectory balance condition}
    \begin{equation}
        R(G', \theta')P_{B}(G\mid G')P_{F}(\theta\mid G) = R(G, \theta)P_{F}(G'\mid G)P_{F}(\theta'\mid G')
        \label{eq:sub-trajectory-balance-condition-jsp-gfn-theorem}
    \end{equation}
    is satisfied for all undirected paths of length 3 between any $(G, \theta)$ and $(G', \theta')$ of the form $(G, \theta) \leftarrow (G, \ast) \rightarrow (G', \ast) \rightarrow (G', \theta')$, then the terminating state distribution is proportional to the reward: $P_{F}^{\top}(G, \theta) \propto R(G, \theta)$.
\end{theorem}

\begin{proof}
    Let $G \neq G_{0}$ be a fixed DAG distinct from the empty graph, and $\theta\in\Theta_{G}$ a set of corresponding parameters. Let $\tau = (G_{0}, \ldots, G_{T-1}, G)$ be an arbitrary trajectory from $G_{0}$ to $G$ during the first phase described in \cref{sec:jsp-gfn-structure-gflownet}, where we use the convention $G_{T} \equiv G$. For any $t < T$, if $\theta_{t} \in \Theta_{G_{t}}$ is a fixed set of parameters associated with $G_{t}$, then the sub-trajectory balance condition \cref{eq:sub-trajectory-balance-condition-jsp-gfn-theorem} can be written for any timestep as
    \begin{equation}
        R(G_{t+1}, \theta_{t+1})P_{B}(G_{t}\mid G_{t+1})P_{F}(\theta_{t}\mid G_{t}) = R(G_{t}, \theta_{t})P_{F}(G_{t+1}\mid G_{t})P_{F}(\theta_{t+1}\mid G_{t+1}),
    \end{equation}
    again using the convention $\theta_{T}\equiv \theta$. Taking the product of the ratio between $P_{F}$ and $P_{B}$ over the trajectory $\tau$, we get by telescoping
    \begin{equation}
        \prod_{t=0}^{T-1}\frac{P_{F}(G_{t+1}\mid G_{t})}{P_{B}(G_{t}\mid G_{t+1})} = \prod_{t=0}^{T-1}\frac{P_{F}(\theta_{t}\mid G_{t})R(G_{t+1}, \theta_{t+1})}{R(G_{t}, \theta_{t})P_{F}(\theta_{t+1}\mid G_{t+1})} = \frac{P_{F}(\theta_{0}\mid G_{0})R(G, \theta)}{R(G_{0}, \theta_{0})P_{F}(\theta\mid G)}.
    \end{equation}
    By definition of the terminating state distribution:
    {\allowdisplaybreaks%
    \begin{align}
        P_{F}^{\top}(G, \theta) &= P_{F}(\theta\mid G)\sum_{\tau: G_{0}\rightsquigarrow G}\prod_{t=0}^{T-1}P_{F}(G_{t+1}\mid G_{t})\\
        &= \frac{P_{F}(\theta_{0}\mid G_{0})}{R(G_{0}, \theta_{0})}R(G, \theta)\sum_{\tau: G_{0}\rightsquigarrow G}\prod_{t=0}^{T-1}P_{B}(G_{t}\mid G_{t+1})\\
        &= \frac{P_{F}(\theta_{0}\mid G_{0})}{R(G_{0}, \theta_{0})}R(G, \theta),\label{eq:sub-trajectory-balance-condition-jsp-gfn-proof-1}
    \end{align}}%
    where we used \cref{lem:PB-distribution-prefix} in \cref{eq:sub-trajectory-balance-condition-jsp-gfn-proof-1}. Moreover, we can prove that $P_{F}(\theta_{0}\mid G_{0}) / R(G_{0}, \theta_{0})$ is independent of $\theta_{0}$ (albeit depending on $G_{0}$, which is the fixed initial state) with \cref{lem:subtb-3-constant-function}. Therefore, this shows that $P_{F}^{\top}(G, \theta) \propto R(G, \theta)$.
\end{proof}
This is a remarkable result, and was only possible thanks to the particular structure of the GFlowNet we consider here. Similar to \cref{sec:fixed-backward-transition-probability}, we will use a fixed backward transition probability $P_{B}(G\mid G')$ as being uniform over the parents of $G'$ for convenience.

\subsection{Parametrization of the forward transition probabilities}
\label{sec:jsp-gfn-parametrization-forward-trasition-probabilities}\index{Transition probability!Forward transition probabilities}
Since $P_{B}$ is fixed, the only quantity that we will learn is once again the forward transition probability $P_{F}$, similar to \cref{sec:parametrization-forward-transition-probabilities}. We will therefore call it $P_{\phi}$ to match the conventions of the previous chapter, where $\phi$ are the parameters of the forward transition probability. In the case of JSP-GFN though, we need to parametrize both $P_{\phi}(G'\mid G)$ the probability to transition to a new graph $G'$ by adding an edge to $G$ (first phase), and $P_{\phi}(\theta\mid G)$ the probability of generating the parameters given $G$ (second phase). We will largely follow the hierarchical representation of \cref{sec:hierarchical-representation-dag-gfn}, where we will parametrize 3 quantities: (1) the probability of terminating the first phase $P_{\phi}(\stopaction\mid G)$, (2) the probability of transitioning to the next graph given that we continue the first phase $P_{\phi}(G'\mid G, \neg\stopaction)$, and (3) the distribution over parameters given that we are in the second phase $P_{\phi}(\theta\mid G, \stopaction)$. The forward transition probabilities are then defined as a combination of all these quantities:
\begin{align}
    P_{\phi}(G'\mid G) &= \big(1 - P_{\phi}(\stopaction \mid G)\big)P_{\phi}(G'\mid G, \neg\stopaction)\label{eq:jsp-gfn-hierarchical-representation-1}\\
    P_{\phi}(\theta\mid G) &= P_{\phi}(\stopaction\mid G)P_{\phi}(\theta\mid G, \stopaction).\label{eq:jsp-gfn-hierarchical-representation-2}
\end{align}
We already saw in \cref{sec:dag-gfn-neural-network-parametrization} how to parametrize both $P_{\phi}(\stopaction\mid G)$ \& $P_{\phi}(G'\mid G, \neg\stopaction)$. To take the example of a graph neural network parametrization (which is the one we use in practice; \citealp{deleu2023jspgfn}), recall that we have a shared backbone from which we extract an embedding $\vg$ of the graph $G$, along with 3 embeddings $\vu_{i}$, $\vv_{i}$ \& $\vw_{i}$ for each node $X_{i}$ in $G$
\begin{equation}
    \vg, \{\vu_{i}, \vv_{i}, \vw_{i}\}_{i=1}^{d} = \mathrm{SelfAttention}_{\phi}\big(\mathrm{GraphNet}_{\phi}(G)\big).
\end{equation}
The embedding $\vg$ was already used to define $P_{\phi}(\stopaction\mid G)$, and the embeddings $\vu_{i}$ \& $\vv_{i}$ to define $P_{\phi}(G'\mid G, \neg\stopaction)$. Finally, we will use the embeddings $\vw_{i}$ to define the distribution $P_{\phi}(\theta\mid G, \stopaction)$; it is important to note that unlike $P_{\phi}(\theta\mid G)$, this is a well-defined probability distribution over $\theta$, and it will effectively be an approximation of the marginal posterior $P(\theta\mid G, \gD)$. Under the same assumption of modularity of the prior $P(\theta\mid G)$ that was already made in the previous chapter (and standard in structure learning), we can show that the posterior $P(\theta\mid G, \gD)$ also decomposes along the variables $X_{i}$. In turn, this motivates our modularization of $P_{\phi}(\theta\mid G, \stopaction)$ along the variables
\begin{equation}
    P_{\phi}(\theta\mid G, \stopaction) = \prod_{i=1}^{d}P_{\phi}(\theta_{i}\mid G, \stopaction).
\end{equation}
Finally, these distributions over $\theta_{i}$ of the conditional probability distributions of $X_{i}$ are parametrized as a multivariate Normal distribution with diagonal covariance (unless specified otherwise), following the literature on Bayesian neural networks \citep{graves2011bnn,blundell2015bayesbybackprop}
\begin{equation}
    P_{\phi}(\theta_{i}\mid G, \stopaction) = \gN\big(\theta_{i}\mid \mu_{\phi}(\vw_{i}), \Sigma_{\phi}(\vw_{i})\big),
    \label{eq:jsp-gfn-posterior-approximation-theta}
\end{equation}
where $\mu_{\phi}$ and $\Sigma_{\phi}$ are two neural networks, with appropriate non-linearities to guarantee that $\Sigma_{\phi}(\vw_{i})$ is a well-defined diagonal covariance matrix. In addition to providing an approximation of the marginal posterior $P(\theta\mid G, \gD)$ via this learned distribution, JSP-GFN also provides an approximation of the marginal posterior $P(G\mid \gD)$, by only following the first phase of the generation process (to generate $G$) until the ``$\stopaction$'' action is selected, and not continuing into the generation of the parameters.

\subsection{DAG-GFlowNet as a special case}
\label{sec:jsp-gfn-dag-gfn-special-case}\index{Directed acyclic graph!DAG-GFlowNet}
Despite the similarities between the modified detailed balance condition and the sub-trajectory balance condition we saw in \cref{sec:jsp-gfn-sub-trajectory-balance-condition}, this new GFlowNet seems quite distinct from DAG-GFlowNet at first glance. However, suppose in this section only that we are in a scenario where we \emph{can} compute the marginal likelihood $P(\gD\mid G)$ (recall that we introduced VBG \& JSP-GFN specifically because we \emph{couldn't} compute it). In that case, we also know the marginal posterior $P(\theta\mid G, \gD)$ as a byproduct, since by Bayes' rule
\begin{equation}
    P(\theta\mid G, \gD) = \frac{P(\gD\mid G, \theta)P(\theta\mid G)}{P(\gD\mid G)} = \frac{R(G, \theta)}{P(\gD\mid G)P(G)}.\label{eq:jsp-gfn-fixed-posterior}
\end{equation}
If we know $P(\theta\mid G, \gD)$, this means in particular that we don't have to learn an approximation of it anymore, and the probability of generating $\theta$ during the second phase now becomes
\begin{equation}
    P_{\phi}(\theta\mid G) = P_{\phi}(\stopaction\mid G)P(\theta\mid G, \gD),
    \label{eq:jsp-gfn-hierarchical-representation-fixed-posterior}
\end{equation}
based on our discussion in the previous section, and the hierarchical representation in \cref{eq:jsp-gfn-hierarchical-representation-2}. If we write the sub-trajectory balance condition from \cref{thm:sub-trajectory-balance-condition-jsp-gfn} for some undirected path $(G, \theta) \leftarrow (G, \ast) \rightarrow (G', \ast) \rightarrow (G', \theta')$ more precisely based on the equation above
{\allowdisplaybreaks%
\begin{align}
    && R(G', \theta')P_{B}(G\mid G')P_{\phi}(\theta\mid G) &= R(G, \theta)P_{\phi}(G'\mid G)P_{\phi}(\theta\mid G)\\
    \Leftrightarrow&& R(G', \theta')P_{B}(G\mid G')P_{\phi}(\stopaction\mid G)P(\theta\mid G, \gD) &= R(G, \theta)P_{\phi}(G'\mid G)P_{\phi}(\stopaction\mid G')P(\theta'\mid G', \gD)\nonumber\\
    \Leftrightarrow&& \widetilde{R}(G')P_{B}(G\mid G')P_{\phi}(\stopaction\mid G) &= \widetilde{R}(G)P_{\phi}(G'\mid G)P_{\phi}(\stopaction\mid G'),\label{eq:jsp-gfn-fixed-posterior-dag-gfn-condition}
\end{align}}%
where we used \cref{eq:jsp-gfn-fixed-posterior} in \cref{eq:jsp-gfn-fixed-posterior-dag-gfn-condition}, and the notation $\widetilde{R}(G) = P(\gD\mid G)P(G)$ to denote the reward of DAG-GFlowNet. We observe that in that case, the sub-trajectory balance condition is in fact \emph{exactly} equivalent to the modified detailed balance condition of \cref{prop:modified-detailed-balance}, based on the parametrization \cref{eq:hierarchical-representation-dag-gfn-1} for $P_{\phi}(\terminal\mid G) \equiv P_{\phi}(\stopaction\mid G)$ in DAG-GFlowNet. In that sense, we can treat JSP-GFN as a direct generalization of DAG-GFlowNet, the latter being recovered with the parametrization \cref{eq:jsp-gfn-hierarchical-representation-fixed-posterior}.

\subsection{Limitations of JSP-GFN}
\label{sec:limitations-jsp-gfn}
In light of what we saw in the previous section, JSP-GFN inherits from many of the limitations that DAG-GFlowNet suffers from, namely that the system is fully observed (\cref{sec:limitations-dag-gfn}) and the difficulties when $\gD$ becomes large (\cref{sec:limitations-gflownets-bayesian-inference}). But there are also challenges specific to JSP-GFN itself. In particular, we made the choice in \cref{sec:jsp-gfn-parametrization-forward-trasition-probabilities} to parametrize the posterior approximation over parameters $P_{\phi}(\theta\mid G, \stopaction)$ as a Normal distribution (with diagonal covariance, except for \emph{JSP-GFN (full)} in \cref{sec:jsp-gfn-joint-posterior-small-graphs}). This choice was made purely for simplicity, but it limits its expressivity to unimodal distributions only; this is an assumption which is commonly made with Bayesian neural networks \citep{hinton1993bnn}. However in general, the posterior distribution $P(\theta\mid G, \gD)$ may be \emph{highly multimodal}, especially when the model is non-linear. To see this, consider a Bayesian network whose conditional probability distributions are parametrized with 2-layer MLPs (as in \cref{sec:jsp-gfn-gaussian-bayesian-networks-simulated-data}). The weights and biases of both layers can be transformed (through an equivariant transformation) in such a way that the hidden units may be permuted, while preserving the values of the outputs. In other words, there are many sets of parameters $\theta$ leading to the same likelihood $P(\gD\mid G, \theta)$, and under mild assumptions on the priors they would have the same posterior probability $P(\theta\mid G, \gD)$.

To address this issue of unimodality, we can use a more expressive posterior approximation $P_{\phi}(\theta\mid G, \stopaction)$, such as ones parametrized with a normalizing flow \citep{rezende2015nf}; this is a drop-in replacement in JSP-GFN since their likelihood can be evaluated. We could even imagine combining it with other generative models, such as diffusion \citep{song2019sorebasedmodels} or a flow-matching models \citep{lipman2023flowmatching}, although parametrizing a score function/flow in those cases might be challenging since this would require the definition of a vector field over the space of parameters $\Theta_{G}$, and would probably necessitate something akin to a hypernetwork \citep{ha2017hypernetworks}. Alternatively, we could also consider multiple steps of a continuous GFlowNet (\cref{chap:gflownets-general-state-spaces}) instead of a single one to generate $\theta$.

\section{Learning objective of JSP-GFN}
\label{sec:jsp-gfn-learning-objective}
Inspired by how we turned flow matching conditions into objectives that can then be optimized using gradient methods in order to approximately satisfy them, we could do the same with the sub-trajectory balance condition of \cref{thm:sub-trajectory-balance-condition-jsp-gfn} here. We saw this type of \emph{sub-trajectory balance loss} already in \cref{sec:equivalence-pcl-subtb} when we established the equivalence between this loss and Path Consistency Learning (PCL; \citealp{nachum2017pcl}) in MaxEnt RL. This would again take the form of a non-linear least-square objective $\widetilde{\gL}(\phi) = \frac{1}{2}\E_{\pi_{b}}\big[\widetilde{\Delta}^{2}(G, \theta, G', \theta'; \phi)\big]$, where $\widetilde{\Delta}(G, \theta, G', \theta'; \phi)$ is a residual derived from \cref{eq:sub-trajectory-balance-condition-jsp-gfn-theorem}
\begin{equation}
    \widetilde{\Delta}(G, \theta, G', \theta'; \phi) = \frac{R(G', \theta')P_{B}(G\mid G')P_{\phi}(\theta\mid G)}{R(G, \theta)P_{\phi}(G'\mid G)P_{\phi}(\theta'\mid G')}.
    \label{eq:sub-tb-jsp-gfn-residual-raw}\index{Sub-trajectory balance!Loss}
\end{equation}
The behavior policy $\pi_{b}$ is a distribution over $(G, \theta, G', \theta')$ with full support, whose details will be given in the next section.

\begin{figure}[t]
    \centering
    \begin{adjustbox}{center}
        \includegraphics[width=500pt]{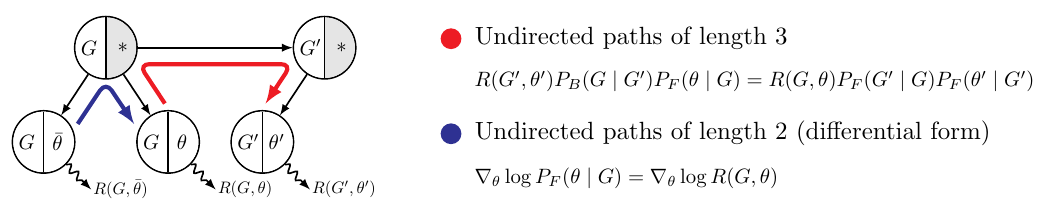}
    \end{adjustbox}
    \caption[Sub-trajectory balance conditions for undirected paths of length $3$ \& $2$]{Sub-trajectory balance conditions for undirected paths of length $3$ (red) \& $2$ (blue) considered here, with an illustration of these paths on an excerpt of the pointed DAG of \cref{fig:jsp-gfn}.}
    \label{fig:sub-trajectory-balance}
\end{figure}

But we can go one step further. The sub-trajectory balance condition we work with here only considers undirected paths of length 3 of the form $(G, \theta) \leftarrow (G, \ast) \rightarrow (G', \ast) \rightarrow (G', \theta')$. However, there are other types of undirected paths that we might be able to also make use of. Consider for example an undirected path of length 2 this time, of the form $(G, \theta) \leftarrow (G, \ast) \rightarrow (G, \bar{\theta})$ with $\theta \neq \bar{\theta}$ (see \cref{fig:sub-trajectory-balance}); instead of transitioning to another graph, we directly go down the second phase again from $(G, \ast)$. In that case, we can show that the corresponding sub-trajectory balance condition can be written in simple form as
\begin{equation}
    R(G, \theta)P_{\phi}(\bar{\theta}\mid G) = R(G, \bar{\theta})P_{\phi}(\theta\mid G).
    \label{eq:sub-tb-2-jsp-gfn}
\end{equation}
Equivalently, we can say that the function $f_{G}(\theta) = \log R(G, \theta) - \log P_{\phi}(\theta\mid G)$ is constant (\ie independent of $\theta$, but with a constant that depends on $G$). Since this function is clearly differentiable (wrt.~$\theta$), another way to write this condition is $\nabla_{\theta}f_{G}(\theta) = 0$. This gives us a \emph{differential formulation} of the sub-trajectory balance condition \cref{eq:sub-tb-2-jsp-gfn} over undirected paths of length 2:
\begin{equation}
    \nabla_{\theta}R(G, \theta) = \nabla_{\theta}P_{\phi}(\theta\mid G).
    \label{eq:sub-tb-2-jsp-gfn-differential}
\end{equation}
Interestingly, this formulation is equivalent to the notion of \emph{score matching} \citep{hyvarinen2005scorematching} to model unnormalized distributions, since $R(G, \theta)$ here corresponds to the unnormalized (marginal) posterior distribution $P(\theta\mid G, \gD)$, where the normalization is over $\theta$.

Another interesting point is that by \cref{lem:subtb-3-constant-function}, we show that \cref{eq:sub-tb-2-jsp-gfn-differential} is effectively redundant when the sub-trajectory condition of \cref{thm:sub-trajectory-balance-condition-jsp-gfn} is satisfied for all undirected paths of length 3. Although satisfying them exactly for \emph{all} undirected paths of length 3 is unlikely in practice, they will eventually be satisfied approximately over the course of training. Another way to encourage this is to ``embed'' the condition \cref{eq:sub-tb-2-jsp-gfn-differential} into our loss function. Going back to the residual \cref{eq:sub-tb-jsp-gfn-residual-raw}, using the notation $\widetilde{\Delta}(\phi)$ for clarity, since $\theta$ \& $\theta'$ may depend on $\phi$ via the reparametrization trick (we will see in \cref{sec:jsp-gfn-behavior-policy} that they are sampled on-policy), we have
\begin{align}
    \frac{1}{2}\frac{d}{d\phi}\widetilde{\Delta}^{2}(\phi) = \widetilde{\Delta}(\phi)\frac{d}{d\phi}\Big[\log R(G', \theta') &+ \log P_{\phi}(\theta\mid G)\nonumber\\ &- \log R(G, \theta) - \log P_{\phi}(G'\mid G) - \log P_{\phi}(\theta\mid G)\Big].
\end{align}
Isolating one term in particular, by the law of total derivatives
{\allowdisplaybreaks%
\begin{align}
    \frac{d}{d\phi}\Big[\log P_{\phi}(\theta\mid G) - \log R(G, \theta)\Big] &= \underbrace{\left[\frac{\partial}{\partial \theta}\log P_{\phi}(\theta\mid G) - \frac{\partial}{\partial \theta}\log R(G, \theta)\right]}_{=\,0}\frac{d\theta}{d\phi} + \frac{\partial}{\partial \phi}\log P_{\phi}(\theta\mid G)\nonumber\\
    &= \frac{\partial}{\partial \phi}\log P_{\phi}(\theta\mid G),\label{eq:jsp-gfn-objective-proof-1}
\end{align}}%
and similarly for the other term in $(G', \theta')$. We integrated the differential formulation of the sub-trajectory balance condition \cref{eq:sub-tb-2-jsp-gfn-differential} over undirected paths of length 2 directly into the computation of the gradient. This derivative of the objective then becomes
\begin{equation}
    \frac{1}{2}\frac{d}{d\phi}\widetilde{\Delta}^{2}(\phi) = \widetilde{\Delta}(\phi)\frac{\partial}{\partial \phi}\Big[\log P_{\phi}(\theta\mid G) - \log P_{\phi}(\theta'\mid G') - \log P_{\phi}(G'\mid G)\Big].
    \label{eq:jsp-gfn-objective-proof-2}
\end{equation}
An alternative way to obtain the same derivative in \cref{eq:jsp-gfn-objective-proof-1} is to take $d\theta/d\phi = 0$ instead, meaning that we would not differentiate through $\theta$ (and $\theta'$). Using the stop-gradient operator $\bot(\cdot)$ (to not confuse with the terminal state!), this means that we can define the new sub-trajectory balance objective as $\gL(\phi) = \frac{1}{2}\E_{\pi_{b}}\big[\Delta^{2}(G, \theta, G', \theta'; \phi)\big]$, where the residual now is
\begin{equation}
    \Delta(G, \theta, G', \theta'; \phi) = \log\frac{R\big(G', \textcolor{Red}{\bot(}\theta'\textcolor{Red}{)}\big)P_{B}\big(G\mid G'\big)P_{\phi}\big(\textcolor{Red}{\bot(}\theta\textcolor{Red}{)}\mid G\big)}{R\big(G, \textcolor{Red}{\bot(}\theta\textcolor{Red}{)}\big)P_{\phi}\big(G'\mid G\big)P_{\phi}\big(\textcolor{Red}{\bot(}\theta'\textcolor{Red}{)}\mid G'\big)},
    \label{eq:jsp-gfn-sub-tb-residual}\index{Sub-trajectory balance!Loss}
\end{equation}
whose gradient matches exactly \cref{eq:jsp-gfn-objective-proof-2}. While optimizing this objective alone leads to eventually satisfying the sub-trajectory balance conditions over undirected paths of length 2, it may be advantageous to explicitly encourage the behavior, especially in cases where $d$ is larger and/or for non-linear models. We can incorporate some penalty to the loss function, such as
\begin{align}
    \widebar{\gL}(\phi) = \gL(\phi) + \frac{\lambda}{2}\E_{\pi_{b}}\Big[\|\nabla_{\theta}\log P_{\phi}(\theta\mid G) &- \nabla_{\theta}\log R(G, \theta)\|^{2}\nonumber\\
    & + \|\nabla_{\theta'}\log P_{\phi}(\theta'\mid G') - \nabla_{\theta'}\log R(G', \theta')\|^{2}\Big].
\end{align}
Finally, the same considerations about the use of a target network that we highlighted in \cref{sec:target-network} also apply here when we evaluate $P_{\phi}\big(\bot(\theta')\mid G'\big)$.

\subsection{Behavior policy}
\label{sec:jsp-gfn-behavior-policy}\index{Behavior policy}
In DAG-GFlowNet, we used a combination of $\varepsilon$-sampling with a replay buffer in order to encourage exploration during training (\cref{sec:off-policy-training}). If we were to do the same thing here, we would be confronted with two problems. First, the notion of $\varepsilon$-sampling is not as well-defined for continuous quantities such as the parameters $\theta$ as it is with discrete ones; this can be addressed by adding an extra variance term to \cref{eq:jsp-gfn-posterior-approximation-theta} \citep{malkin2023gfnhvi}. But the second issue, more practical this time, is that storing experience in the replay buffer here would mean storing tuples $(G, \theta, G', \theta')$, which could be extremely large since $\theta$ represents all the parameters of the Bayesian network.

Instead, we will adopt a ``lazy'' strategy where the behavior policy $\pi_{b}$ is a combination of off-policy and on-policy. Taking inspiration from DAG-GFlowNet \citep{deleu2022daggflownet}, we will still store transitions $G \rightarrow G'$ in the replay buffer, following \cref{sec:off-policy-training}, whereas their corresponding parameters $\theta$ \& $\theta'$ are sampled on-policy using our current $P_{\phi}(\theta\mid G, \stopaction)$. Therefore, a key difference with \citet{deleu2022daggflownet} is that the reward $R(G, \theta)$ is computed ``lazily'' when the loss is evaluated (\ie only once $\theta$ and $\theta'$ are known), as opposed to being computed during the interaction with the state space and stored in the replay buffer alongside the transitions. \cref{alg:jsp-gfn-training} gives a description of the training algorithm of JSP-GFN.

\begin{algorithm}[t]
    \caption{JSP-GFN training}
    \label{alg:jsp-gfn-training}
    \begin{algorithmic}[1]
        \State Initialize the trajectory at $G_{0}$ the empty graph
        \Loop
            \LComment{Interactions with the environment}
            \For{$T$ steps}
                \State Sample whether the trajectory continues or not: $a \sim P_{\phi}(\stopaction\mid G_{t})$
                \If{$a = 1$}
                    \State Reset the trajectory: $G_{t+1} = G_{0}$ is the empty graph
                \Else
                    \State Sample a new graph $G_{t+1} \sim P_{\phi}(G_{t+1}\mid G_{t}, \neg\stopaction)$\Comment{see also \cref{alg:dag-gflownet-interaction}}
                    \State Store the transition $G_{t} \rightarrow G_{t+1}$ in the replay buffer
                \EndIf
            \EndFor
            \State \vphantom{A blank line}
            \LComment{Update of the parameters $\phi$}
            \State Sample a mini-batch $\gB$ of transitions $G \rightarrow G'$ from the replay buffer
            \State Sample $\theta \sim P_{\phi}(\theta\mid G, \stopaction)$ (and similarly for $\theta'$) for each transition in $\gB$
            \State Evaluate the rewards $R(G, \theta)$ and $R(G', \theta')$
            \State Evaluate the loss $\gL(\phi)$ based on $\gB$\Comment{\cref{eq:jsp-gfn-sub-tb-residual}}
            \State $\phi \leftarrow \phi - \beta \nabla_{\phi}\gL(\phi)$\Comment{Update the parameters with gradient methods}
        \EndLoop
        \State \Return The forward transition probability $P_{\phi}(\cdot \mid G)$
    \end{algorithmic}
\end{algorithm}

\subsection{Mini-batch training}
\label{sec:jsp-gfn-minibatch-training}
\begin{figure}[t]
    \centering
    \begin{adjustbox}{center}
        \includegraphics[width=480pt]{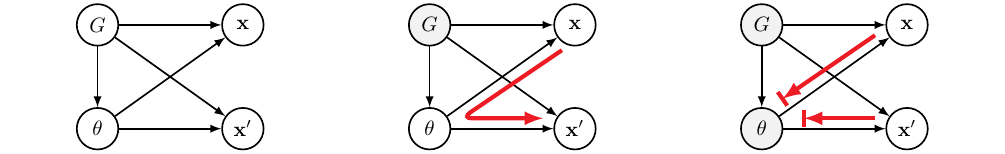}%
    \end{adjustbox}
    \begin{adjustbox}{center}%
        \begin{subfigure}[b]{160pt}%
            \caption{}%
            \label{fig:meta-bayesian-network-1}%
        \end{subfigure}%
        \begin{subfigure}[b]{160pt}%
            \caption{}%
            \label{fig:meta-bayesian-network-2}%
        \end{subfigure}%
        \begin{subfigure}[b]{160pt}%
            \caption{}%
            \label{fig:meta-bayesian-network-3}%
        \end{subfigure}%
    \end{adjustbox}
    \caption[Conditional dependencies between the graph, parameters, and data]{Conditional dependencies between the DAG $G$, the parameters $\theta$, and datapoints $\vx$ \& $\vx' \in \gD$. (a) The conditional dependencies represented as a (higher level) Bayesian network. (b) The effect of conditioning only on $G$; this leaves an open path between $\vx$ and $\vx'$ through $\theta$, making them conditionally \emph{dependent}. (c) The effect of conditioning on both $G$ \& $\theta$ this time, closing all paths from $\vx$ to $\vx'$ and thus making them conditionally \emph{independent}. Conditional independencies are read via d-separation presented in \cref{def:d-separation}.}
    \label{fig:meta-bayesian-network}
\end{figure}

So far when we talked about Bayesian structure learning in general, we always assumed that inference was done over the whole dataset $\gD$ at once. As the size of the dataset grows, computing the reward becomes more and more expensive since it is linear in the size of $\gD$, regardless of our discussion in \cref{sec:limitations-gflownets-bayesian-inference}. Taking inspiration from how deep learning models are trained, it would be far more convenient if we could also train our GFlowNet not based on the whole dataset, but on mini-batches of data. This proves to impossible though for DAG-GFlowNet (and more generally to approximate the marginal posterior $P(G\mid \gD)$). The reason being that if we have 2 datapoints $\vx$ \& $\vx'$ from $\gD$, then they are \emph{not} independent given $G$ alone: $\vx \not\independent\vx' \mid G$; there exists an open path through $\theta$ in the (higher-level) Bayesian network shown in \cref{fig:meta-bayesian-network-2} describing the distribution $P(\gD, \theta, G)$.

If we model the joint posterior $P(G, \theta\mid \gD)$ though, there is no such open path anymore, and we do have $\vx \independent \vx'\mid G, \theta$. Therefore beyond the capacity to model arbitrary (\eg non-linear) likelihood models, another advantage of approximating the joint posterior is that we can train the GFlowNet using mini-batches of observations. Concretely, for a mini-batch $\gB = \{\vx^{(m)}\}_{m=1}^{M}$ of $M$ observations sampled uniformly at random from the dataset $\gD$, we can define
\begin{equation}
    \log \widehat{R}_{\gB}(G, \theta) = \log P(\theta\mid G) + \log P(G) + \frac{N}{M}\sum_{m=1}^{M}\log P\big(\vx^{(m)}\mid G, \theta\big),
    \label{eq:jsp-gfn-estimate-log-reward}
\end{equation}
which is an unbiased estimate of the log-reward. We can likewise define a (stochastic) loss function $\widehat{\gL}_{\gB}(\phi)$ by simply replacing the reward by this estimate in the residual \cref{eq:jsp-gfn-sub-tb-residual}. The following proposition shows that minimizing this estimated loss wrt.~the parameters $\phi$ also minimizes the original loss function that uses the whole dataset $\gD$.
\begin{proposition}
    \label{prop:jsp-gfn-mini-batch-minimization}
    Suppose that $\gB$ is a mini-batch of observations sampled uniformly at random from the dataset $\gD$, and let $\widehat{\gL}_{\gB}(\phi)$ be the loss where the reward has been replaced by the estimate $\widehat{R}_{\gB}(G, \theta)$ in \cref{eq:jsp-gfn-estimate-log-reward} in the residual \cref{eq:jsp-gfn-sub-tb-residual}. Then we have $\gL(\phi) \leq \E_{\gB}\big[\widehat{\gL}_{\gB}(\phi)\big]$.
\end{proposition}
\begin{proof}
    We will first show that $\log \widehat{R}_{\gB}(G, \theta)$ is an unbiased estimate of the log-reward $\log R(G, \theta)$ under a uniform distribution of the mini-batches $\gB$. By conditional independence of the observations $\vx^{(n)}$ given $G$ and $\theta$, we have
    \begin{equation}
        \log P(\gD\mid G, \theta) = \sum_{n=1}^{N}\log P\big(\vx^{(n)}\mid G, \theta\big) = N\E_{\vx}\big[\log P(\vx\mid G, \theta)\big],
    \end{equation}
    where the expectation is taken over the uniform distribution over $\gD = \{\vx^{(n)}\}_{n=1}^{N}$. It is important to note that the independence of the observations is critical here to get this decomposition. Similarly, we have
    \begin{equation}
        \E_{\gB}\Bigg[\sum_{\vx^{(m)}\in\gB}\log P\big(\vx^{(m)}\mid G, \theta\big)\Bigg] = M\E_{\vx}\big[\log P(\vx\mid G, \theta)\big].
    \end{equation}
    Combining these two equations together, we can show that the estimate of the log-reward is indeed unbiased:
    {\allowdisplaybreaks%
    \begin{align}
        \E_{\gB}\big[\log \widehat{R}_{\gB}(G, \theta)\big] &= \frac{N}{M}\E_{\gB}\Bigg[\sum_{\vx^{(m)}\in\gB}\log P\big(\vx^{(m)}\mid G, \theta\big)\Bigg] + \log P(\theta\mid G) + \log P(G)\\
        &= \log P(\gD\mid G, \theta) + \log P(\theta\mid G) + \log P(G) = \log R(G, \theta).
    \end{align}}%
    Recall that the stochastic loss is defined as $\widehat{\gL}_{\gB}(\phi) = \frac{1}{2}\E_{\pi_{b}}\big[\widehat{\Delta}^{2}_{\gB}(G, \theta, G', \theta'; \phi)\big]$, where the residual is
    \begin{equation}
        \widehat{\Delta}_{\gB}(G, \theta, G', \theta'; \phi) = \log\frac{\widehat{R}_{\gB}\big(G', \bot(\theta')\big)P_{B}\big(G\mid G'\big)P_{\phi}\big(\bot(\theta)\mid G\big)}{\widehat{R}_{\gB}(G, \bot(\theta)\big)P_{\phi}\big(G'\mid G\big)P_{\phi}\big(\bot(\theta')\mid G'\big)}.
    \end{equation}
    Taking the expectation of the stochastic loss wrt.~a random batch $\gB$, we get
    \begin{equation}
        \E_{\gB}\big[\widehat{\gL}_{\gB}(\phi)\big] = \frac{1}{2}\E_{\pi_{b}}\Big[\E_{\gB}\big[\widehat{\Delta}^{2}_{\gB}(\phi)\big]\Big] \geq \frac{1}{2}\E_{\pi_{b}}\Big[\E_{\gB}\big[\widehat{\Delta}_{\gB}(\phi)\big]^{2}\Big] = \frac{1}{2}\E_{\pi_{b}}\big[\Delta^{2}(\phi)\big] = \gL(\phi),
    \end{equation}
    where we used the convexity of the squared function and Jensen's inequality, and the unbiasedness of $\log \widehat{R}_{\gB}(G, \theta)$ (as well as $\log \widehat{R}_{\gB}(G', \theta')$).
\end{proof}
Note that we only used the convexity of the squared loss in the proof above to conclude, not its exact form. This is interesting because as we mentioned in \cref{sec:dag-gfn-learning-objective}, we use the Huber loss in practice for stability, which is also convex. In the case of the squared loss, we can actually show a stronger result in terms of unbiasedness of the gradient estimator.
\begin{proposition}
    \label{prop:jsp-gfn-mini-batch-gradient-unbiased}
    The mini-batch gradient estimator is unbiased: $\nabla_{\phi}\gL(\phi) = \E_{\gB}\big[\nabla_{\phi}\widehat{\gL}_{\gB}(\phi)\big]$. Therefore, the local and global minima of the expected mini-batch loss coincide with those of the full-batch loss.
\end{proposition}
\begin{proof}
    We can first observe that $\nabla_{\phi}\Delta(\phi) = \nabla_{\phi}\widehat{\Delta}_{\gB}(\phi)$, since only the terms corresponding to the reward differ between $\Delta(\phi)$ \& $\widehat{\Delta}_{\gB}(\phi)$, and they do not depend on $\phi$ (in particular thanks to the stop-gradient operator). Therefore
    \begin{equation*}
        \frac{1}{2}\E_{\gB}\big[\nabla_{\phi}\widehat{\Delta}_{\gB}^{2}(\phi)\big] = \E_{\gB}\big[\widehat{\Delta}_{\gB}(\phi)\nabla_{\phi}\widehat{\Delta}_{\gB}(\phi)\big] = \E_{\gB}\big[\widehat{\Delta}_{\gB}(\phi)\big]\nabla_{\phi}\Delta(\phi) = \Delta(\phi)\nabla_{\phi}\Delta(\phi) = \frac{1}{2}\nabla_{\phi}\Delta^{2}(\phi).
    \end{equation*}
    Taking the expectation over the behavior policy $\E_{\pi_{b}}[\cdot]$ of both sides of the equation above concludes the proof.
\end{proof}
This proposition implies that the expected mini-batch loss and the full-batch loss only differ by a constant factor \citep{deleu2023jspgfn}. Note that mini-batch training alone does not address the limitations mentioned in \cref{sec:limitations-gflownets-bayesian-inference} for Bayesian inference when the dataset is large, since the estimate of the log-reward \cref{eq:jsp-gfn-estimate-log-reward} still scales with the total number of observations $N$. %

\section{Experimental results}
\label{sec:jsp-gfn-experimental-results}
To validate the effectiveness of both VBG and JSP-GFN to model the joint posterior over DAGs and the parameters of the conditional probability distributions, we will revisit the experiments we performed in \cref{chap:dag-gflownet}. We will first test these methods on simulated data, both on small graphs to allow comparison with $P(G, \theta\mid \gD)$ directly and on medium scale, and then we will study two real-world biological applications.

For comparison with existing algorithms, we use a similar set of baseline methods as the one we already used throughout \cref{chap:dag-gflownet}: 2 methods based on MCMC \citep{giudici2003improvingmcmc}, 2 based on bootstrapping \citep{friedman1999bootstrap}, and 2 based on variational inference. The ones based on variational inference, DiBS \citep{lorch2021dibs} \& BCD Nets \citep{cundy2021bcdnets}, are already approximations of the joint posterior as we saw in \cref{sec:joint-posterior-variational-inference}, although BCD Nets is only applicable on linear-Gaussian models. However the other 4 methods are by default approximating the \emph{marginal} posterior $P(G\mid \gD)$, and they need to be adapted for joint inference. In the case of bootstrapping, still using either GES \citep{chickering2002ges} or the PC algorithm \citep{spirtes2000causationpredictionsearch}, we approximate $P(\theta\mid G, \gD)$ by a Dirac distribution placed at the maximum likelihood estimate $\widehat{\theta}_{\mathrm{MLE}}$ for each sampled graphs \citep{agrawal2019abcd}. For MCMC, we enhance structure MCMC (MC\textsuperscript{3}) described in \cref{sec:structure-mcmc} with the ability to also sample parameters $\theta$ using two options \citep{lorch2021dibs}: either (1) the proposal distribution sample jointly $(G', \theta')$ for the Metropolis-Hastings step (\emph{MH-MC\textsuperscript{3}}), or (2) new parameters $\theta'$ are only proposed once we have already proposed to move to a new $G'$ with structure MCMC (\emph{Gibbs-MC\textsuperscript{3}}).

\subsection{Joint posterior over small graphs}
\label{sec:jsp-gfn-joint-posterior-small-graphs}
Following \cref{sec:dag-gfn-small-graphs-comparison}, we first evaluate the quality of the posterior approximation for both VBG (\cref{sec:vbg-variational-bayes-approach}) and JSP-GFN (\cref{sec:joint-inference-single-gflownet}) on small simulated problems, in order to allow direct comparison with the exact posterior $P(G, \theta\mid \gD)$. Computing this distribution can only be done exactly in limited cases, where (1) $P(\theta\mid G, \gD)$ can be computed analytically, and (2) for a small enough number of variables $d$ such that all the DAGs can be enumerated to compute $P(G\mid \gD)$. Therefore, we use the same experimental setup described in \cref{sec:dag-gfn-comparison-exact-posterior} with graphs over $d=5$ nodes, keeping a linear-Gaussian assumption to compute the posterior over parameters analytically. We once again generated $20$ different datasets of $N=100$ observations from randomly generated linear-Gaussian Bayesian networks $(G^{\star}, \theta^{\star})$. The likelihood function $P(\gD\mid G, \theta)$ is described in \cref{eq:linear-gaussian-model}, and we place a unit Normal distribution as the prior over parameters $P(\theta_{ij}\mid G) = \gN(0, 1)$; this differs from the BGe score we used in \cref{sec:dag-gfn-comparison-exact-posterior} in that $\sigma^{2}$ is treated as a hyperparameter here, instead of being a parameter of the Bayesian network, and therefore the resulting log-reward is not \emph{score-equivalent} anymore (\ie placing the same reward for Markov equivalent DAGs; \citealp{koller2009pgm}). We use a uniform prior over graphs $P(G)$.

\begin{figure}[t]
    \centering
    \begin{adjustbox}{center}
        \includegraphics[width=470pt]{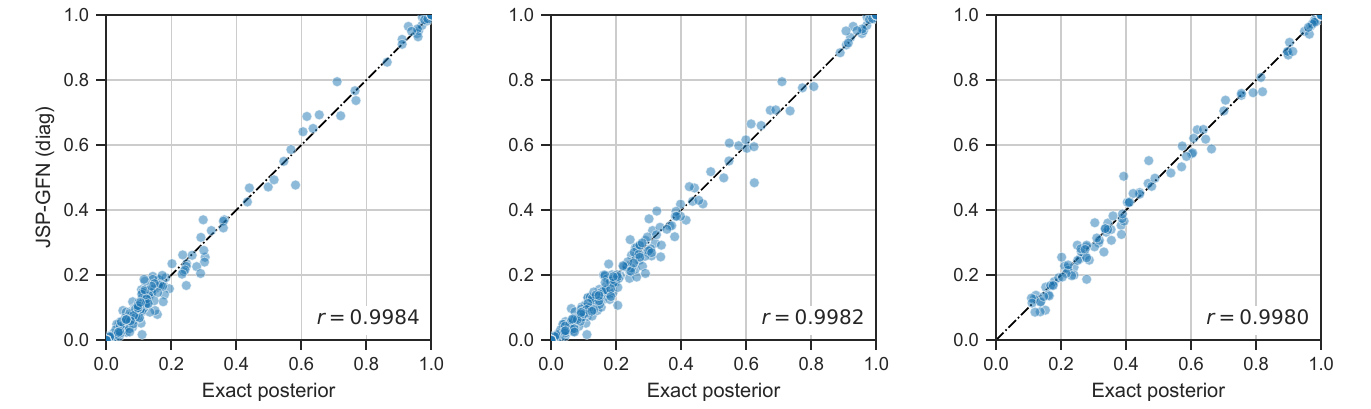}%
    \end{adjustbox}
    \begin{adjustbox}{center}%
        \begin{subfigure}[b]{160pt}%
            \caption{Edge features}%
            \label{fig:jsp-gfn-small-graphs-1}%
        \end{subfigure}%
        \begin{subfigure}[b]{160pt}%
            \caption{Path features}%
            \label{fig:jsp-gfn-small-graphs-2}%
        \end{subfigure}%
        \begin{subfigure}[b]{160pt}%
            \caption{Markov blanket features}%
            \label{fig:jsp-gfn-small-graphs-3}%
        \end{subfigure}%
    \end{adjustbox}
    \begin{adjustbox}{center}
        \includegraphics[width=470pt]{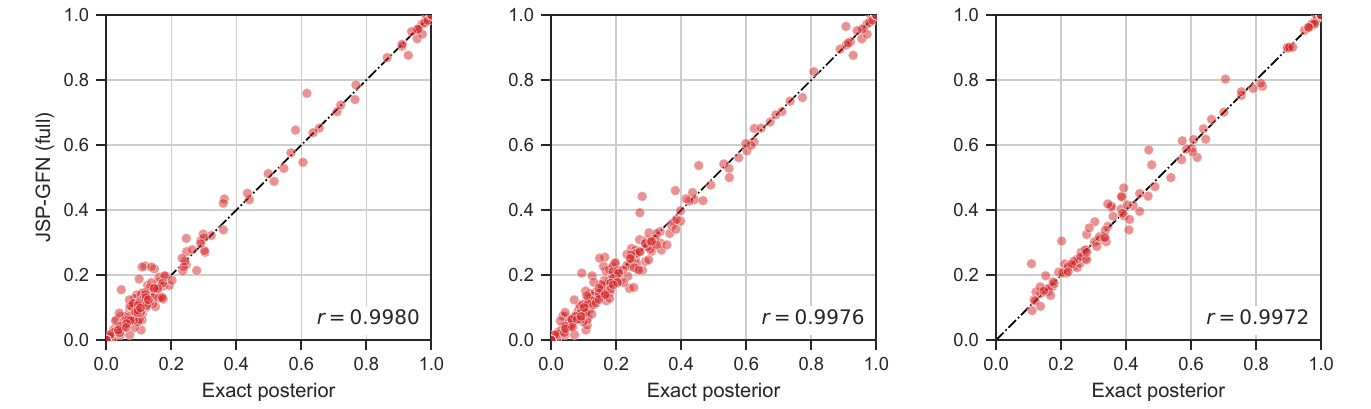}%
    \end{adjustbox}
    \begin{adjustbox}{center}%
        \begin{subfigure}[b]{160pt}%
            \caption{Edges features}%
            \label{fig:jsp-gfn-small-graphs-4}%
        \end{subfigure}%
        \begin{subfigure}[b]{160pt}%
            \caption{Path features}%
            \label{fig:jsp-gfn-small-graphs-5}%
        \end{subfigure}%
        \begin{subfigure}[b]{160pt}%
            \caption{Markov blanket features}%
            \label{fig:jsp-gfn-small-graphs-6}%
        \end{subfigure}%
    \end{adjustbox}
    \caption[Comparison between the exact posterior and the approximation from JSP-GFN]{Comparison between the exact posterior distribution and the approximation from JSP-GFN, for different structural features. (a-c) the posterior approximation over parameters $P(\theta\mid G, \stopaction)$ is parametrized as a Normal distribution with diagonal covariance (JSP-GFN (diag)). (d-f) We use a Normal distribution with full covariance matrix (JSP-GFN (full)). Each point corresponds to a feature (\ie marginal probability) computed for a specific pair $(X_{i}, X_{j})$ in a graph over $d = 5$ nodes, either based on the exact posterior $P(G\mid \gD)$ (x-axis), or the approximation found with JSP-GFN (y-axis; see \cref{sec:evaluation-terminating-state-probability-dynamic-programming}). The Pearson correlation coefficient $r$ is included at the bottom-right corner of each plot.}
    \label{fig:jsp-gfn-small-graphs}
\end{figure}

The quality of the joint posterior approximation is evaluated separately for $G$ and $\theta$. For the DAGs, we compare the approximation and the exact posterior on different marginals of interest, also called \emph{features} (\citealp{friedman2003ordermcmc}; see also \cref{sec:dag-gfn-small-graphs-comparison}). We show a comprehensive evaluation of JSP-GFN on the 3 categories of features in \cref{fig:jsp-gfn-small-graphs}, with two versions that only differ in the way $P_{\phi}(\theta\mid G, \stopaction)$ is parametrized: both use a Normal distribution, but ``\emph{JSP-GFN (diag)}'' uses a diagonal covariance matrix $\Sigma_{\phi}$ in \cref{eq:jsp-gfn-posterior-approximation-theta}, whereas ``\emph{JSP-GFN (full)}'' uses a full covariance matrix. We consider both parametrizations since it is known that in the case of a linear-Gaussian model the posterior $P(\theta\mid G, \gD)$, which is approximated by $P_{\phi}(\theta\mid G, \stopaction)$, is a Normal distribution with a full covariance matrix in general \citep{murphy2023pml2book,nishikawa2023vbg}, ``JSP-GFN (full)'' hence serving as a possibly more faithful approximation. Similar to DAG-GFlowNet in \cref{fig:dag-gfn-small-graphs}, we observe that both versions of JSP-GFN are capable of accurately approximating the marginal posterior $P(G\mid \gD)$ based at least on these structural features.

\begin{table}[t]
    \centering
    \caption[Quantitative evaluation of different methods for joint inference on small graphs]{Quantitative evaluation of different methods for joint posterior approximation on small graphs with $d=5$ nodes, both in terms of edge features and cross-entropy of between the approximation and the exact posterior over parameters $P(\theta\mid G, \gD)$. All values correspond to the mean and 95\% confidence interval across $20$ experiments.}
    \label{tab:jsp-gfn-small-graphs}
    \begin{tabular}{lccc}
        \toprule
         & \multicolumn{2}{c}{Edge features} & \multirow{2}{*}{$\mathbb{E}_{G, \theta}\big[\!-\!\log P(\theta\mid G, \mathcal{D})\big]$} \\
        \cmidrule(lr){2-3}
         & RMSE & Pearson's $r$ & \\
        \midrule
        MH-MC\textsuperscript{3} & $0.357 \pm 0.022$ & $0.067 \pm 0.143$ & $\phantom{-}5.39 \pm 1.41 \times 10^{2}$ \\
        Gibbs-MC\textsuperscript{3} & $0.357 \pm 0.022$ & $0.028 \pm 0.127$ & $\phantom{-}9.02 \pm 1.54 \times 10^{5}$ \\
        Bootstrap-GES & $0.263 \pm 0.070$ & $0.635 \pm 0.180$ & $\phantom{-}1.56 \pm 0.97 \times 10^{2}$ \\  %
        Bootstrap-PC & $0.305 \pm 0.057$ & $0.570 \pm 0.138$ & $\phantom{-}1.57 \pm 0.87 \times 10^{2}$ \\  %
        DiBS & $0.312 \pm 0.038$ & $0.737 \pm 0.071$ & $\phantom{-}9.49 \pm 7.34 \times 10^{3}$ \\
        BCD Nets & $0.215 \pm 0.055$ & $0.819 \pm 0.097$ & $\phantom{-}7.04 \pm 3.21 \times 10^{1}$ \\
        \midrule
        VBG & $0.237 \pm 0.037$ & $0.816 \pm 0.064$ & $\phantom{-}1.24 \pm 0.49 \times 10^{2}$ \\
        JSP-GFN (diag) & $\mathbf{0.018 \pm 0.005}$ & $\mathbf{0.998 \pm 0.001}$ & $\mathbf{-4.91 \pm 0.51 \times 10^{0}}$ \\
        JSP-GFN (full) & $\mathbf{0.019	\pm 0.007}$ & $\mathbf{0.998 \pm 0.001}$ & $\mathbf{-5.00 \pm 0.52 \times 10^{0}}$ \\
        \bottomrule
    \end{tabular}
\end{table}

But to get a complete view of how accurate this approximation is, we also evaluate the performance of the posterior approximation over $\theta$. This time we not only compare JSP-GFN, but also VBG and all other baseline methods described in the previous section. In addition to quantitative results of edge features (purely structural quantities as we saw above), \cref{tab:jsp-gfn-small-graphs} also shows the cross-entropy between the posterior approximations (over which the expectation is taken) and the exact posterior $P(\theta\mid G, \gD)$. This measures how likely sample parameters from the approximations are under the exact posterior (lower is better). We observe that out of all methods, JSP-GFN is capable of sampling parameters $\theta$ that are significantly more probable under $P(\theta\mid G, \gD)$. On the other hand, VBG fares relatively well, being on par with bootstrapping methods (only surpassed by BCD Nets).

\subsection{Gaussian Bayesian networks from simulated data}
\label{sec:jsp-gfn-gaussian-bayesian-networks-simulated-data}
To evaluate whether our observations hold on medium-size graphs, we also compare the performance of VBG \& JSP-GFN on data simulated from larger Gaussian Bayesian networks with $d=20$ variables, following the experimental setup described in \cref{sec:dag-gfn-evaluation-large-graphs}. In addition to linear Gaussian models though, this time we also experimented with models whose conditional distributions are non-linear functions parametrized by neural networks. Following \citet{lorch2021dibs}, we parametrized the conditional probability distributions of each variable with a 2-layer MLP, for a total of $|\theta| = 2,220$ parameters. For both experimental settings, we generated $20$ datasets of $N=100$ observations from randomly generated Bayesian networks $(G^{\star}, \theta^{\star})$. We use the same priors over parameters $P(\theta\mid G)$ and over graphs $P(G)$ as the ones described in the previous section.

\begin{figure}[t]
    \centering
    \begin{adjustbox}{center}
        \includegraphics[width=505.714285714pt]{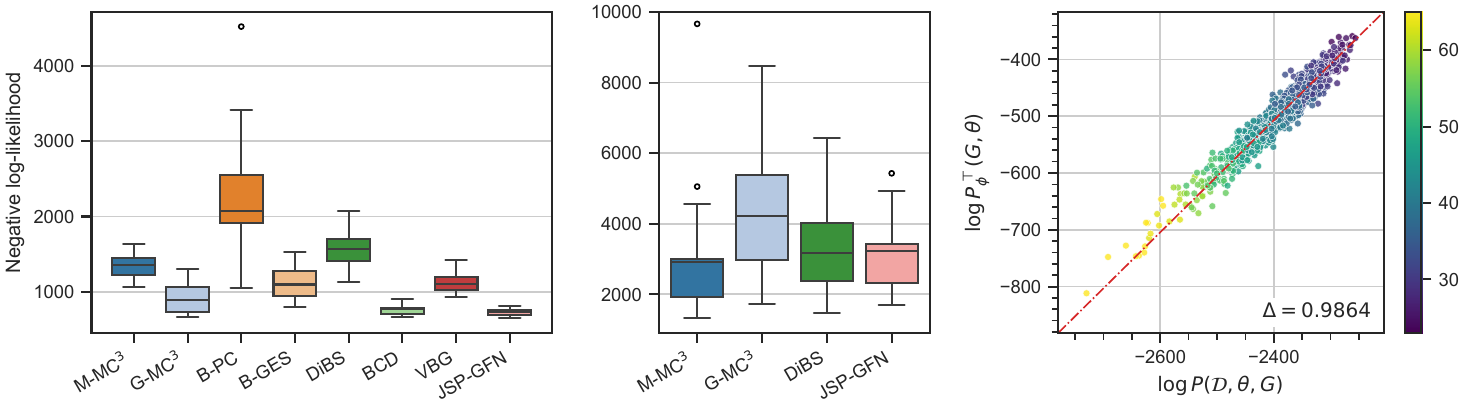}%
    \end{adjustbox}
    \begin{adjustbox}{center}%
        \begin{subfigure}[b]{160pt}%
            \caption{Linear Gaussian}%
            \label{fig:jsp-gfn-gaussian-20-linear}%
        \end{subfigure}%
        \begin{subfigure}[b]{160pt}%
            \caption{Non-linear Gaussian}%
            \label{fig:jsp-gfn-gaussian-20-nonlinear}%
        \end{subfigure}%
        \begin{subfigure}[b]{160pt}%
            \caption{Non-linear Gaussian}%
            \label{fig:jsp-gfn-gaussian-20-scatter}%
        \end{subfigure}%
    \end{adjustbox}
    \caption[Evaluation of JSP-GFN on Gaussian Bayesian networks with $d=20$ variables]{Evaluation of JSP-GFN on Gaussian Bayesian networks with $d=20$ variables. (a-b) Comparison of the negative log-likelihood on $N'=100$ held-out observations for different Bayesian structure learning methods, aggregated across 20 experiments on different datasets $\gD$. (c) Linear correlation between the log-reward (x-axis) and the terminating state log-probability (y-axis, estimated with \cref{sec:estimation-terminating-state-probability}) for $1,000$ samples $(G, \theta)$ from JSP-GFN. The color of each point indicates the number of edges in the corresponding graph. The slope $\Delta$ of the linear fit obtained with RANSAC \citep{fischler1981ransac} is shown at the bottom-right corner.}
    \label{fig:jsp-gfn-gaussian-20}
\end{figure}

In light of the discussion of \cref{sec:criticism-evaluation-metrics}, we opted for evaluating all methods based on the negative log-likelihood on held-out data \citep{eaton2007bayesian} (\ie $N' = 100$ new observations from the same $(G^{\star}, \theta^{\star})$), as opposed to structural measures with the ground-truth graphs $G^{\star}$ such as the $\E$-SHD or the AUROC. This provides a more representative perspective of the eventual performance on downstream tasks. In \cref{fig:jsp-gfn-gaussian-20-linear} we show the negative log-likelihood for linear-Gaussian models, and observe that JSP-GFN achieves the best performance out of all other methods, with VBG being a close contender. In \cref{fig:jsp-gfn-gaussian-20-nonlinear} we show the negative log-likelihood, this time on non-linear models, and therefore ruling out the methods based on bootstrapping and BCD Nets, as well as VBG since it was primarily introduced for linear-Gaussian models \citep{nishikawa2023vbg}, despite our discussion in \cref{sec:vbg-update-posterior-approximation-parameters}. We can observe that JSP-GFN achieves this time a performance similar to DiBS and competitive with MH-MC\textsuperscript{3}.

To complement these metrics, and in line with \cref{fig:dag-gfn-lingauss20-diag} for DAG-GFlowNet, we also assess the quality of the posterior approximation for JSP-GFN by comparing the terminating state log-probability and the log-reward of sampled $(G, \theta)$. Recall that ideally, the terminating state probability distribution approximates the joint posterior, meaning that
\begin{equation}
    \log P_{\phi}^{\top}(G, \theta) \approx \log P(G, \theta\mid \gD) = \log P(\gD, \theta, G) - \log P(\gD),
\end{equation}
where $\log P_{\phi}^{\top}(G, \theta)$ is once again estimated using the technique described in \cref{sec:estimation-terminating-state-probability}, and $P(\gD)$ is a constant offset. In \cref{fig:jsp-gfn-gaussian-20-scatter}, we can see that there is indeed a strong linear correlation across multiple samples $(G, \theta)$ from JSP-GFN, with a slope $\Delta$ close to $1$, suggesting again that it accurately approximates the joint posterior distribution at least around the modes it captures.

\subsection{Learning biological structures from real data}
\label{sec:jsp-gfn-biological-structures-real-data}
We finally evaluate JSP-GFN on real-world biological data for two separate tasks: the discovery of protein signaling networks from flow cytometry data \citep{sachs2005causal} following \cref{sec:dag-gfn-application-protein-signaling-networks}, and the discovery of small gene regulatory networks from gene expression data.

\paragraph{Protein signaling network} We use the data described in \cref{sec:dag-gfn-sachs-experimental-data}, consisting of a discretized version of flow cytometry data with both observational and interventional data \citep{sachs2005causal,eaton2007bayesian}. In preparation for the evaluation based on negative log-likelihood, we split the data into a dataset $\gD$ of $N = 4,200$ measurements of the activity of $d=11$ phosphoproteins from $7$ experimental settings (including control), leaving out $N'=1,200$ measurements from $2$ separate experimental settings; in other words, we will evaluate the performance of JSP-GFN on completely unseen interventions (via the \emph{interventional} negative log-likelihood). To account for the complexity of biological systems, we model it as a non-linear discrete Bayesian network, where the conditional probability distributions are parametrized using a 2-layer MLP with $16$ hidden units, \ie $P\big(X_{i} \mid \parents_{G}(X_{i}); \theta_{i}\big) = \mathrm{Categorical}(\pi_{i})$, with
\begin{equation}
    \pi_{i} = \mathrm{MLP}(\mM_{i}\mX; \theta_{i}),
\end{equation}
where $\mX$ encodes the inputs as one-hot vectors, $\mM_{i}$ is an appropriate mask depending on the structure of $G$, and the MLP has a softmax activation function for the output layer. In total, the model has $|\theta| = 6,545$ parameters. The priors over parameters $P(\theta\mid G)$ and over graphs $P(G)$ are the same as the ones described in \cref{sec:jsp-gfn-joint-posterior-small-graphs}. The purpose of this experiment is to show that JSP-GFN can work with a flexible model on discrete data.

\begin{figure}[t]
    \centering
    \begin{adjustbox}{center}
        \includegraphics[width=395pt]{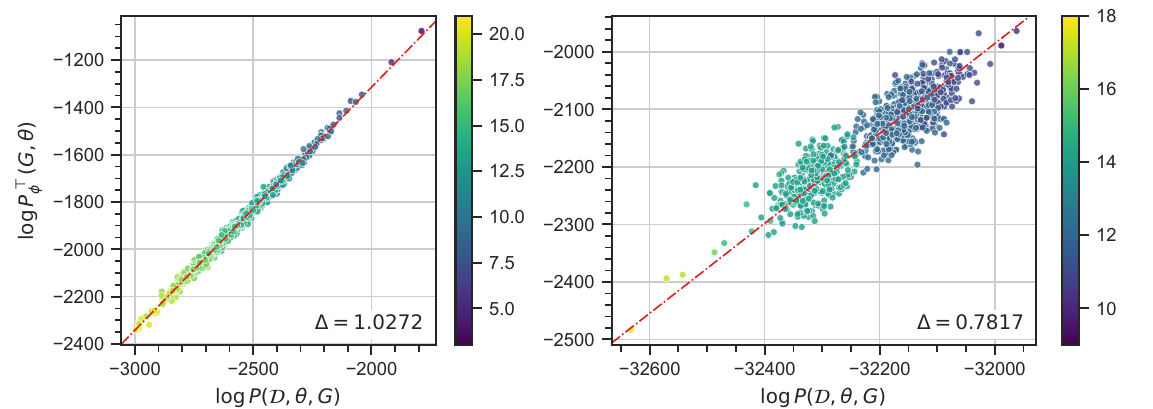}%
    \end{adjustbox}
    \begin{adjustbox}{center}%
        \begin{subfigure}[b]{170pt}%
            \caption{Sub-sampled dataset ($N=100$)}%
            \label{fig:jsp-gfn-sachs-subsample}%
        \end{subfigure}%
        \begin{subfigure}[b]{220pt}%
            \caption{Full dataset ($N = 4,200$)}%
            \label{fig:jsp-gfn-sachs-full}%
        \end{subfigure}%
    \end{adjustbox}
    \caption[Performance of JSP-GFN on the inference of protein signaling networks]{Performance of JSP-GFN on the inference of protein signaling networks with flow cytometry data \citep{sachs2005causal}. (a) Linear correlation between the log-reward (x-axis) and terminating state log-probability (y-axis) on a model trained with a subsampled dataset. Each of the $1,000$ points represents a $(G, \theta)$ sampled from JSP-GFN. (b) Same graph with JSP-GFN trained on the full dataset. The color of each point indicates the number of edges in the corresponding graph. The slope $\Delta$ of the linear fit obtained with RANSAC \citep{fischler1981ransac} is shown at the bottom-right corner.}
    \label{fig:jsp-gfn-sachs}
\end{figure}

To measure the quality of the posterior approximation, we again compare in \cref{fig:jsp-gfn-sachs} the terminating state log-probability with the log-reward, similar to the previous section. We observe there is a linear correlation between these two quantities. However unlike \cref{fig:jsp-gfn-gaussian-20-scatter}, the slope $\Delta = 0.7817$ in \cref{fig:jsp-gfn-sachs-full} is not as close to $1$, suggesting that JSP-GFN under-estimates the probability of $(G, \theta)$. We also observe that that the samples are ``clustered'' together, which can be explained by the fact that the posterior approximation is concentrated at only a few graphs, since the size of the dataset $\gD$ is larger. To confirm this observation, we show in \cref{fig:jsp-gfn-sachs-subsample} a similar plot but this time on a dataset of $N = 100$ datapoints randomly sampled from $\gD$, matching the experimental setup we considered in prior sections. We observe a much closer linear fit this time, with a slope $\Delta$ close to $1$. We also report in \cref{tab:jsp-gfn-biological-data} the performance of JSP-GFN as the negative log-likelihood on held-out interventional data (with new unseen interventions).

\begin{table}[t]
    \centering
    \caption[Comparison of JSP-GFN with MCMC methods on real-world biological data]{Comparison of JSP-GFN with MCMC methods on real-world biological data, in terms of the negative (interventional) log-likelihood on held-out data (and held-out interventions).}
    \label{tab:jsp-gfn-biological-data}
    \begin{tabular}{lcc}
        \toprule
        & Flow cytometry & Gene expression \\
        & $d = 11, N = 4,200$ & $d = 61, N = 2,628$ \\
        \midrule
        MH-MC\textsuperscript{3} & $\mathbf{2.578 \times 10^{4}}$ & $1.073 \times 10^{6}$\\
        Gibbs-MC\textsuperscript{3} & $3.490 \times 10^{5}$ & $4.882 \times 10^{6}$\\
        \midrule
        JSP-GFN & $2.679 \times 10^{4}$ & $\mathbf{2.651 \times 10^{5}}$\\
        \bottomrule
    \end{tabular}
\end{table}

\paragraph{Gene regulatory networks} Taking inspiration from prior works using Bayesian networks to model gene regulatory networks \citep{friedman2000expressiondata,friedman2004cellular,peer2006minreg,kaderali2008inferringgeneregulatorynets,lopez2022dcdfg}, we also tested JSP-GFN on a similar inference task. Following \citet{sethuraman2023nodagsflow}, we used a subset of $N=2,628$ datapoints corresponding to expression data from $d=61$ genes \citep{dixit2016perturbseq} under various experimental settings (both experimental \& control). The data is pre-processed following \citet{lopez2022dcdfg,sethuraman2023nodagsflow}. Gene expression data is composed of either non-zero continuous data (when a gene is expressed) or \emph{exactly} zero values (when the gene is inhibited). Therefore, we model this system as a non-linear Bayesian network, where the conditional probability distributions are \emph{zero-inflated Normal} distributions
\begin{equation}
    P\big(X_{i}\mid \parents_{G}(X_{i});\theta_{i}\big) = \alpha_{i}\delta_{0}(X_{i}) + (1 - \alpha_{i})\gN(X_{i};\mu_{i}, \sigma_{i}^{2}),
\end{equation}
where $\mu_{i}$ is the result of a 2-layer MLP with $16$ hidden units similar to the one described above for protein signaling networks. The parameters of the conditional probability distributions contain the parameters of the MLPs, as well as the mixture parameters $\alpha_{i}$ and the variances of the observation noise $\sigma_{i}^{2}$, for a total of $|\theta| = 61,671$ parameters.

The purpose of this experiment is again to show the complete flexibility of the modeling for the Bayesian network, with non-standard distributions, but also to test JSP-GFN on a larger system of $d=61$ variables. In \cref{tab:jsp-gfn-biological-data} we report the negative log-likelihood of held-out data, with unseen interventions. We observe that JSP-GFN achieves a lower interventional negative log-likelihood than both methods based on MCMC.

\paragraph{Biological plausibility of the acyclicity assumption} One of the strengths of JSP-GFN, and DAG-GFlowNet before it, is the capacity to obtain a distribution over the DAG structure of a Bayesian network (and its parameters). The acyclicity assumption is necessary to properly define the likelihood model in \cref{eq:bayesian-network}. Although the flow cytometry dataset of \citet{sachs2005causal} is a well established dataset in the structure learning literature, and there is a growing interest in using those techniques for inferring gene regulatory networks, it is important to note that biological systems often exhibit some feedback processes that cannot be captured by acyclic graphs \citep{mooij2020jci}. Therefore, the graphs found by JSP-GFN (or any structure learning algorithm) must be carefully interpreted. As a general framework though, GFlowNets can be adapted to ignore the acyclic nature of the graphs sampled by ignoring parts of the mask $\mM$ defined in \cref{sec:efficient-verification-valid-actions}. Alternatively, we can view the generation of a cyclic graph by unrolling it \citep{murphy2002dbn}, and use a GFlowNet similar to the one defined by \citet{atanackovic2023dyngfn} but with a full Bayesian treatment.

\subsection{Scaling to larger graphs}
\label{sec:jsp-gfn-scaling-larger-graphs}
Throughout this chapter, and more generally the second part of this thesis, we carefully evaluated the performance of DAG-GFlowNet, VBG and JSP-GFN on a range of experimental settings, from small-scale problems where we could compare against the exact posterior, to medium and larger scale problems, up to Bayesian networks over $d=61$ variables in the previous section. This is the order of magnitude the community of (Bayesian) structure learning has been mostly focused on. A question one might ask is: what does it take to scale these models to larger problems? For example, gene regulatory networks at the human genome level would require systems over about $20,000$ variables, far beyond the scale we considered in this thesis.

The theoretical framework we devised in \cref{chap:dag-gflownet,chap:jsp-gfn} does not limit the scale at which they can be applied a priori. But of course the challenge now becomes practical, as larger networks will inevitably require more computational resources and larger neural networks to model the forward transition probability $P_{\phi}$. The advantage of GFlowNets in general though is that they acquire their own data via interactions with the pointed DAG $\gG$ (just like online reinforcement learning), meaning that they are in principle not bound by the amount of ``training data'' they have access to. However the quality of this data is of utmost importance, and this requires an extensive \emph{exploration} of the state space to possibly capture multiple modes of the posterior distribution. This is largely an open problem, not only for GFlowNets but also in the reinforcement learning literature.

Finally another challenge of scaling to larger graphs is the problem of \emph{evaluation}: how can we make sure that even if we can train JSP-GFN on larger graphs, the posterior approximation we find is faithful? Considering the limitations of structural metrics mentioned in \cref{sec:criticism-evaluation-metrics}, throughout this chapter we saw three evaluation methods: (1) the comparison with the exact posterior (\cref{fig:jsp-gfn-small-graphs}), (2) the comparison between $\log P_{\phi}^{\top}(G, \theta)$ and $\log R(G, \theta)$ (\eg \cref{fig:jsp-gfn-gaussian-20-scatter}), and (3) the evaluation of the (interventional) negative log-likelihood on held-out data (\eg \cref{fig:jsp-gfn-gaussian-20-linear}). The first evaluation is clearly limited to small graphs and specific conditional distributions in order to compute $P(G, \theta\mid \gD)$ analytically. The second one requires the estimation of the terminating state log-probability as described in \cref{sec:estimation-terminating-state-probability}, which is reliable only up to a certain scale (the limiting factor here being the number of edges $K$ more than the number of nodes $d$, since we are estimating a sum with $K!$ terms). The third one is the most universally applicable, but it only serves as a comparative measure with other methods; it is not a measure of the intrinsic quality of JSP-GFN \emph{by itself} as an approximation of $P(G, \theta\mid \gD)$. But even as a comparison, does a good predictive power (even on new interventions) necessarily reflect something about the posterior distribution over the \emph{structure} of the Bayesian network? This question of robust evaluation in Bayesian structure learning is again an open problem.

%% file: chapters/09_Conclusion.tex
\chapter*{Conclusion}
\label{chap:conclusion}
\addcontentsline{toc}{chapter}{Conclusion}
In this thesis, we presented a comprehensive view of the theory of \emph{generative flow networks}, with a particular accent on their applications to \emph{Bayesian structure learning}. GFlowNets offer a new set of tools to model distributions over discrete and compositional objects, which has been a relatively under-represented area of (modern) generative modeling. We saw that they originated from addressing the shortcomings of using maximum entropy reinforcement learning to sample from these distributions. We also showed throughout the first part how they eventually unify many domains of machine learning and statistics, such as variational inference (\cref{sec:gflownets-variational-inference}), reinforcement learning (\cref{chap:gflownet-maxent-rl}), and Markov chains (\cref{sec:gflownet-general-state-spaces}). But despite their apparent generality, it should be emphasized that the ambition of GFlowNets is \emph{not} to supersede any other class of generative models, but to complement them and play to their (GFlowNets') strengths. This is precisely why modeling the posterior distribution over the full characterization of a Bayesian network, in particular with its directed acyclic graph structure, arises as a natural fit for GFlowNets.

However, modeling a distribution over models is not an end in itself. Going back to the work of George E. P. Box, he argued that scientific investigation is an iterative process interlacing \emph{deduction} (\ie using hypotheses to make predictions) and \emph{induction} (\ie inferring hypotheses from new observations) \citep{box1976sciencestats}. Taking a Bayesian perspective for model selection fits this vision by design, since the posterior can be used for predictions and subsequently be updated with new actively acquired evidence to refine our model of the real-world \citep{buntine1991refinementbn}; this update may be over the whole system \citep{tong2001activelearningstructure,murphy2001activelearning,scherrer2021learningcausalmodelactive,tigas2022interventionswhere,toth2022abci}, or only part of it \citep{agrawal2019abcd}. This interplay between scientific discovery and active learning has been a major driving force in the GFlowNet literature \citep{jain2022gfnbioseq,jain2023gfnscientific,jain2023mogfn,hernandez2023multifidelitygfn}.

More generally, GFlowNets have come a long way since their origins as a model to generate diverse small molecules \citep{bengio2021gflownet}, and our work laying their theoretical foundations as a generative model \citep{bengio2023gflownetfoundations}. They have found applications in a wide spectrum of domains, including combinatorial optimization \citep{zhang2023gfnrobustscheduling,zhang2023graphcogfn,kim2024antcolonygfn}, symbolic regression and neural architecture search \citep{li2023gfnsr,chen2024orderpreservinggfn}, reasoning in large language models \citep{hu2024gfnllm,song2024latentlogictreegfn}, AI safety \citep{lee2024redteaminggfn}, probabilistic inference \citep{falet2024deltaai}, astrophysics \citep{bunao2024gfncovariantloopquantum}, genetics \citep{zhou2024phylogfn}, materials science \citep{milaai4science2023crystalgfn,nguyen2023hierarchicalgfncrystal,cipcigan2024matgfn}, and computational chemistry for drug design \citep{volokhova2024conformationgeneration,cretu2024synflownet}.

The future is bright for GFlowNets, but some outstanding challenges still remain. Chief among them is the problem of efficient \emph{exploration} of the state space. We saw in \cref{sec:off-policy-training} that GFlowNets are inherently off-policy methods, allowing for more flexibility in the way data is acquired, and providing clear benefits empirically (\cref{sec:dag-gfn-effect-off-policy}). But with massive state spaces, such as the ones we experience in structure learning and scientific discovery in general \citep{fink2005moleculenumber}, naive strategies such as $\varepsilon$-sampling will most likely not be sufficient to effectively explore all the modes of the target distribution. Some methods borrowed from the reinforcement learning literature have been proposed to address this difficult question \citep{pan2023gafn,rectorbrooks2023thompsongfn,morozov2024mctsgfn}, but further inspiration from reinforcement learning (where the question of exploration is predominant) is certainly necessary.

One characteristic of GFlowNets, once again inherited from their close ties with reinforcement learning, is that they require \emph{simulation} through the state space $\gG$: data is acquired by rolling out a behavior policy $\pi_{b}$ (possibly storing the experience in a replay buffer) always starting from the initial state; there are notable exceptions though \citep{zhang2022ebgfn,kim2024lsgfn}. This is in stark contrast with other modern generative models such as diffusion models \citep{song2019sorebasedmodels} or flow-matching models \citep{lipman2023flowmatching,tong2024conditionalflowmatching} which tend to go towards \emph{simulation-free} objectives where a noisy version of the data is obtained directly without going through an iterative process; this is arguably what makes them so successful. While the situation remains quite different for GFlowNets (we never have to backpropagate through the simulation), we could leverage the fact that unlike most RL environments $\gG$ is known (at least locally). This means that we could get training data starting from \emph{any state} without having to systematically go back to $s_{0}$, which could again help for exploration \citep{ecoffet2021goexplore}.

The synergies between GFlowNets and those other modern generative models should also be further exploited. These have already been the subject of some works using GFlowNets-like objectives to train diffusion models for example \citep{zhang2024dgfs,zhang2024directalignmentgfn,sendera2024diffusiongfn,venkatraman2024rtb}. An exciting direction of research is to use a GFlowNet not as a way to \emph{tune} these models, but rather to \emph{guide} an a priori completely unstructured generative model like a diffusion model (\eg for de novo molecule generation) with the highly structured and often discrete GFlowNet on top of it (\eg to enforce synthesizability at low cost). The use of these powerful generative models can also have major benefits in conjunction with a GFlowNet like JSP-GFN (\cref{chap:jsp-gfn}) to capture the multi-modeality of the target posterior distribution.

Finally, specifically for the application of GFlowNets to Bayesian structure learning, the next frontier is to learn from low-level observations (\eg images), as opposed to directly observing the values of the random variables. In this setting, the Bayesian network would be defined over \emph{latent variables}, the values of which now would have to be inferred (based on low-level observations) along with the structure and the parameters of the model, the latter two being covered in \cref{chap:jsp-gfn}. This is largely an unexplored area, apart from the seminal work of \citet{friedman1998structuralem} in (non-Bayesian) structure learning with latent variables \citep{murphy2002dbn}. Although some of our preliminary findings extending DAG-GFlowNet to latent variable models can be found in \citep{manta2024latentdaggfn}, substantial additional work is necessary to make it broadly applicable.

%% file: appendix/01_Generative_Flow_Networks.tex
\chapter{Generative Flow Networks}
\label{app:generative-flow-networks}

\section{Generative flow networks}
\label{app:gflownets}

\subsection{Alternative conditions}
\label{app:alternative-conditions}
In this section, we will prove an alternative formulation of the flow matching condition stated in \cref{prop:flow-matching-condition-state-flow-pF}. We recall this proposition here and prove it; the proof follows the same pattern as the one of \cref{thm:flow-matching-proportional-reward}.

\begin{proposition611}[Flow matching condition]
    Let $\gG = (\widebar{\gS}, \gA)$ be a pointed DAG. A function $F: \gS \rightarrow \sR_{+}$ defines the state flow, and $P_{F}: \gS \rightarrow \Delta(\children_{\gG})$ the forward transition probabilities of a unique Markovian flow $F^{\star}$ if and only if they satisfy the following condition for all $s\in\gS$ such that $s'\neq s_{0}$:
    \begin{equation}
        F(s') = \sum_{s\in\parents_{\gG}(s')}F(s)P_{F}(s'\mid s).
    \end{equation}
    Moreover, under these conditions, the unique Markovian flow $F^{\star}$ whose state flow and forward transition probabilities match $F$ and $P_{F}$ respectively is defined, for any complete trajectory $\tau = (s_{0}, s_{1}, \ldots, s_{T}, \terminal)$, by
    \begin{equation}
        F^{\star}(\tau) = F(s_{0})\prod_{t=0}^{T}P_{F}(s_{t+1}\mid s_{t}),
    \end{equation}
    with the convention $s_{T+1} = \terminal$.
\end{proposition611}

\begin{proof}
    $\Rightarrow$: If $F^{\star}$ is a Markovian flow, then by \cref{prop:identities-state-edge-flows} its edge and state flows satisfy the following identity for all $s'\in\gS$ such that $s'\neq s_{0}$:
    \begin{equation}
        F^{\star}(s') = \sum_{s\in\parents_{\gG}(s')}F^{\star}(s\rightarrow s').
    \end{equation}
    Moreover, by definition of the forward transition probability induced by a flow (\cref{def:transition-probabilities-from-flow}), we know that $F^{\star}(s\rightarrow s') = F^{\star}(s)P_{F}^{\star}(s'\mid s)$ for any $s\rightarrow s'\in\gG$. Therefore, this proves the necessity of the condition for the state flow and the forward transition probabilities of a Markovian flow.

    $\Leftarrow$: Let $F$ and $P_{F}$ be functions satisfying the condition in \cref{eq:flow-matching-condition-state-flow-pF} for all state $s'\in\gS$ such that $s'\neq s_{0}$, and let $F^{\star}$ be the trajectory flow defined in \cref{eq:flow-matching-condition-state-flow-pF-markovian-flow}. Since $F(s_{0})$ is constant and $P_{F}$ is a forward transition probability consistent with $\gG$, the flow $F^{\star}$ is Markovian by \cref{prop:characterization-markovian-flow}. Moreover, this also shows that $P_{F}$ matches the forward transition probabilities of $F^{\star}$ (again, by \cref{prop:characterization-markovian-flow}). To show the sufficiency of the condition in \cref{eq:flow-matching-condition-state-flow-pF}, we then need to prove that (1) $F$ matches the state flow of $F^{\star}$, and that (2) the Markovian flow $F^{\star}$ is unique.

    \begin{enumerate}[leftmargin=*]
        \item We can prove that the function $F$ matches the state flow of $F^{\star}$ by strong induction. For any state $s\in\gS$, let $d_{s}$ denote the maximum length of a partial trajectory from the initial state $s_{0}$ to $s$.

        \emph{Base case:} The initial state is the unique state such that $d_{s_{0}} = 0$. By \cref{prop:characterization-markovian-flow} and the form of the Markovian flow in \cref{eq:flow-matching-condition-state-flow-pF-markovian-flow}, we know that the total flow of $F^{\star}$ is given by $Z^{\star} = F(s_{0})$. Furthermore, by \cref{prop:initial-flow-total-flow}, we know that the total flow is equal to the state flow at the initial state: $F^{\star} = F^{\star}(s_{0})$. Therefore, this shows that $F^{\star}(s_{0}) = F(s_{0})$.

        \emph{Induction step:} Suppose that $F^{\star}(s) = F(s)$ for all states $s$ such that $d_{s} \leq d$, for some $d > 0$. Since $\gG$ is a pointed DAG, there exists a state $s'\in\gG$ such that $d_{s'} = d + 1$; it is clear that $s' \neq s_{0}$. By \cref{prop:identities-state-edge-flows} \& \cref{def:transition-probabilities-from-flow}, we then have
        \begin{align}
            F^{\star}(s') &= \sum_{s\in\parents_{\gG}(s')}F^{\star}(s \rightarrow s') = \sum_{s\in\parents_{\gG}(s')}F^{\star}(s)P_{F}^{\star}(s'\mid s)\\
            &= \sum_{s\in\parents_{\gG}(s')}F(s)P_{F}(s'\mid s) = F(s'),\label{eq:flow-matching-condition-state-flow-pF-proof1}
        \end{align}
        where we used the fact that $P_{F}^{\star}(s'\mid s) = P_{F}(s'\mid s)$ and the induction assumption in \cref{eq:flow-matching-condition-state-flow-pF-proof1}, as well as the condition in \cref{eq:flow-matching-condition-state-flow-pF}. This shows that the function $F$ matches the state flow of $F^{\star}$.

        \item The proof of unicity of the Markovian flow is identical to the proof of \cref{thm:detailed-balance-condition}. Suppose that there exists another Markovian flow $F'^{\star}$ such that $F$ matches its state flow and $P_{F}$ matches its forward transition probabilities. We denote by $P_{F}'^{\star}$ the forward transition probabilities associated with $F'^{\star}$. By definition, we then have for any transition $s\rightarrow s'\in \gG$ (where $s'$ may be equal to the terminal state $\terminal$)
        \begin{equation}
            F'^{\star}(s\rightarrow s') = F'^{\star}(s)P_{F}'^{\star}(s'\mid s) = F(s)P_{F}(s'\mid s) = F^{\star}(s)P_{F}^{\star}(s'\mid s) = F^{\star}(s\rightarrow s'),
        \end{equation}
        where we used the fact that $F^{\star}$ and $F'^{\star}$ have both their state flows and forward transition probabilities equal to $F$ and $P_{F}$ respectively. Since this is valid for all transitions $s\rightarrow s'\in \gG$, by definition of equivalent flows we have $F'^{\star} \sim F^{\star}$. By \cref{prop:equivalent-markovian-flows}, this means that the two flows are equal, showing the unicity of $F^{\star}$.
    \end{enumerate}
\end{proof}

\subsection{Convergence guarantees}
\label{app:convergence-guarantees}
In this section, we will prove a general upper-bound on the log-expectation-exp based on the maximum element. This lemma is useful for proving multiple results for the convergence guarantees of GFlowNets (\eg \cref{prop:bound-difference-log-probs,prop:bound-kl-divergence-terminating-state-prob}).
\begin{lemma}[Bound on log-expectation-exp]
    \label{lem:bound-log-expectation-exp}
    Let $\vp = (p_{1}, \ldots, p_{n})$ be a vector of probabilities (\ie $p_{i} \geq 0$ and $\sum_{i} p_{i} = 1$), and $\vx = (x_{1}, \ldots, x_{n})$ be an arbitrary vector. Then
    \begin{equation}
        \left|\log \sum_{i=1}^{n}p_{i}\exp(x_{i})\right| \leq \max_{i}|x_{i}|.
    \end{equation}
\end{lemma}

\begin{proof}
    For any vector $\vy = (y_{1}, \ldots, y_{n})$, and since $p_{i} \geq 0$, we have the following inequalities
    \begin{equation}
        \sum_{i=1}^{n}p_{i}\Big(\min_{j} y_{j}\Big) \leq \sum_{i=1}^{n}p_{i}y_{i}\leq \sum_{i=1}^{n}p_{i}\Big(\max_{j}y_{j}\Big).
    \end{equation}
    Since $\sum_{i}p_{i} = 1$, both sides of these inequalities can be further simplified, only involving the minimum and maximum of $\vy$. We can apply these inequalities to $y_{i} = \exp(x_{i})$, and observe that $\min_{j}\exp(x_{j}) = \exp(\min_{j} x_{j})$ (and similarly for the $\max$), because the exponential is a monotonically increasing function
    \begin{equation}
        \exp\Big(\min_{j}x_{j}\Big) \leq \sum_{i=1}^{n}p_{i}\exp(x_{i}) \leq \exp\Big(\max_{j} x_{j}\Big).
    \end{equation}
    Taking the logarithm of the inequalities above, and again using the fact that $\log$ is monotonically increasing
    \begin{equation}
        \min_{j} x_{j} \leq \log \sum_{i=1}^{n}p_{i}\exp(x_{i}) \leq \max_{j}x_{j}.
        \label{eq:proof-bound-log-expectation-exp-1}
    \end{equation}
    Another way to write \cref{eq:proof-bound-log-expectation-exp-1} is
    \begin{align}
        \log \sum_{i=1}^{n}p_{i}\exp(x_{i}) &\leq \max_{i}x_{i} \leq \max_{i}|x_{i}|\\
        -\log \sum_{i=1}^{n}p_{i}\exp(x_{i}) &\leq -\min_{i}x_{i} \leq \max_{i}|x_{i}|,
    \end{align}
    which concludes the proof.
\end{proof}

\subsection{GFlowNets \& variational inference}
\label{app:gflownets-variational-inference}
In this section, we recall a well-known result called the \emph{data processing inequality}, that relates the $f$-divergence between joint distributions to the $f$-divergence between their marginals. This result, taken from \citep{zhang2019variationalfdivergence} and recalled for completeness, plays an important role in proving \cref{prop:data-processing-inequality-gflownets}, which establishes a bound on the $f$-divergence between the terminating state probability distribution of a GFlowNet and the target Gibbs distribution, in terms of the $f$-divergence between distributions over complete trajectories \citep{malkin2022trajectorybalance}.

\begin{proposition}[Data processing inequality]
    \label{prop:data-processing-inequality}\index{f-divergence@$f$-divergence}
    Let $f$ be a convex function. Let $P(x, z)$ and $Q(x, z)$ be two distributions. The $f$-divergence of the marginals is bounded by the $f$-divergence of the joint distributions:
    \begin{equation}
        D_{f}\big(P(x)\,\|\,Q(x)\big) \leq D_{f}\big(P(x, z)\,\|\,Q(x, z)\big).
        \label{eq:data-processing-inequality}
    \end{equation}
\end{proposition}

\begin{proof}
    The proof is taken from \citep{zhang2019variationalfdivergence}, and recalled here for completeness. Using the general decomposition $Q(x, z) = Q(x)Q(z\mid x)$ we have
    \begin{align}
        D_{f}\big(P(x, z)\,\|\,Q(x, z)\big) &= \iint Q(x, z) f\left(\frac{P(x, z)}{Q(x, z)}\right) dxdz\\
        &= \int Q(x) \int Q(z\mid x) f\left(\frac{P(x, z)}{Q(x, z)}\right)dz dx\\
        &\geq \int Q(x) f\left(\int q(z\mid x)\frac{P(x, z)}{Q(z\mid x)Q(x)}dz\right)dx\label{eq:data-processing-inequality-proof-1}\\
        &= \int Q(x) f\left(\frac{P(x)}{Q(x)}\right)dx = D_{f}\big(P(x)\,\|\,Q(x)\big),
    \end{align}
    where we used Jensen's inequality in \cref{eq:data-processing-inequality-proof-1}.
\end{proof}

\section{GFlowNets over General State Spaces}
\label{app:gflownets-general-state-spaces}

\subsection{Generative flow network as a Markov chain}
\label{app:gflownet-as-markov-chain}
The following proposition shows that the terminating state probability distribution defined purely over Markov chains in \cref{def:general-discrete-terminating-state-distribution} (in the discrete case) is a properly defined distribution over $\gS$. This result is analoguous as \cref{prop:terminating-state-proper-distribution} in the case of discrete GFlowNets.
\begin{proposition}
    \label{prop:general-discrete-terminating-state-proper-distribution}
    The terminating state distribution $P_{F}^{\top}$ is a well-defined probability distribution over $\gS$.
\end{proposition}
\begin{proof}
    It is clear that for all $x\in\gX$, $P_{F}^{\top}(x) \geq 0$. Moreover, using \cref{prop:terminating-state-distribution-invariant-measure}, we know that $P_{F}^{\top}(x) = \lambda(x)P_{F}(s_{0}\mid x)$ where $\lambda$ is the unique invariant measure of $P_{F}$ such that $\lambda(s_{0}) = 1$ (\cref{thm:discrete-markov-chain-existance-invariant-measure}). Therefore
    \begin{equation}
        \sum_{x\in\gS}P_{F}^{\top}(x) = \sum_{x\in\gS}\lambda(x)P_{F}(s_{0}\mid x) = \lambda(s_{0}) = 1,
    \end{equation}
    where we used the invariance of $\lambda$ in the second equality.
\end{proof}

\subsection{From invariance back to flow matching}
\label{sec:invariance-back-flow-matching}
The Markov chain perspective we devised in \cref{sec:gflownet-as-markov-chain} is elegant, but in practice the objective is to find a transition probability $P_{F}$ and a measure $F$, which may be parametrized by neural networks, that satisfy the conditions of \cref{thm:markov-chain-discrete-terminating-state-propto-reward}: $F$ must satisfy the boundary conditions and must be invariant for $P_{F}$. For this, we can take inspiration from the loss functions introduced in \cref{sec:flow-matching-losses}. The summation in \cref{eq:definition-invariant-measure-discrete}, necessary for checking the invariance of $F$, is typically inexpensive for most states if the state space is well structured (\ie if the objects have a clear compositional nature, as described in \cref{sec:compositional-objects}), since we only have to sum over the ``parents'' of $s'$ (all the states with non-zero probability of transitioning to $s'$).

However, checking the invariance of $F$ at the initial state $s_{0}$ proves to be as difficult as computing the partition function itself, since the parent set of $s_{0}$ is the whole sample space $\gX$. Fortunately, the following proposition shows that we only have to check that $F$ satisfies \cref{eq:definition-invariant-measure-discrete} for any state $s' \neq s_{0}$.

\begin{proposition}
    \label{prop:markov-chain-flow-matching-invariant}
    Let $(X_{n})_{n\geq 0}$ be an irreducible and positive recurrent Markov chain over $\gS$, with initial state $s_{0}$ and transition probability $P_{F}$. A measure $F$ is invariant for $P_{F}$ if and only if for all $s' \neq s_{0}$, we have
    \begin{equation}
        F(s') = \sum_{s\in\gS}F(s)P_{F}(s'\mid s).
        \label{eq:markov-chain-flow-matching-invariant}
    \end{equation}
\end{proposition}

\begin{proof}
    The necessity of this proposition is immediate. For sufficiency, we can first show (by induction) that if \cref{eq:markov-chain-flow-matching-invariant} is satisfied for all $s'\neq s_{0}$, then
    \begin{equation}
        F(s) = F(s_{0})\sum_{n=0}^{\infty}\E_{s_{0}}\big[\mathds{1}(n < \sigma_{s_{0}})\mathds{1}(X_{n} = s)\big].
        \label{eq:markov-chain-flow-matching-invariant-proof-1}
    \end{equation}
    Since the Markov chain is irreducible and positive recurrent, then by \cref{thm:discrete-markov-chain-existance-invariant-measure} it admits an invariant measure. The only non-trivial statement one must show is that if \cref{eq:markov-chain-flow-matching-invariant} is satisfied for any $s'\neq s_{0}$, then it must also hold for $s' = s_{0}$. Since the Markov chain is positive recurrent, it must eventually return to $s_{0}$ in finite time. In other words
    \begin{equation}
        \sum_{n=1}^{\infty}\E_{s_{0}}\big[\mathds{1}(\sigma_{s_{0}} = n)\big] = 1.
        \label{eq:markov-chain-flow-matching-invariant-proof-2}
    \end{equation}
    Furthermore, it is clear that at any point in time $n \geq 0$, $X_{n}$ must be in one of the states of $\gS$, meaning that $\sum_{s\in\gS}\mathds{1}(X_{n} = s) = 1$. Therefore
    \begin{align}
        \sum_{s\in\gS}F(s)P_{F}(s_{0}\mid s) &= \sum_{s\in\gS}\sum_{n=0}^{\infty}F(s_{0})\E_{s_{0}}\big[\mathds{1}(n < \sigma_{s_{0}})\mathds{1}(X_{n} = s)\big]P_{F}(s_{0}\mid s)\label{eq:markov-chain-flow-matching-invariant-proof-3}\\
        &= F(s_{0})\sum_{s\in\gS}\sum_{n=1}^{\infty}\E_{s_{0}}\big[\mathds{1}(\sigma_{s_{0}} = n)\mathds{1}(X_{n-1} = s)\big]\label{eq:markov-chain-flow-matching-invariant-proof-4}\\
        &= F(s_{0})\sum_{n=1}^{\infty}\E_{s_{0}}\Bigg[\mathds{1}(\sigma_{s_{0}} = n)\underbrace{\sum_{s\in\gS}\mathds{1}(X_{n-1} = s)}_{=\,1}\Bigg]\\
        &= F(s_{0})\sum_{n=1}^{\infty}\E_{s_{0}}\big[\mathds{1}(\sigma_{s_{0}} = n)\big] = F(s_{0}),\label{eq:markov-chain-flow-matching-invariant-proof-5}
    \end{align}
    where we substituted $F(s)$ by \cref{eq:markov-chain-flow-matching-invariant-proof-1} in the first step \cref{eq:markov-chain-flow-matching-invariant-proof-3}, we used \cref{eq:terminating-state-distribution-invariant-measure-proof-5} from the proof of \cref{prop:terminating-state-distribution-invariant-measure} in the second step \cref{eq:markov-chain-flow-matching-invariant-proof-4}, and finally \cref{eq:markov-chain-flow-matching-invariant-proof-2} to conclude in \cref{eq:markov-chain-flow-matching-invariant-proof-5}. This concludes the proof, showing that invariance also holds at $s_{0}$.
\end{proof}
This results mirrors exactly the practical implementation of flow networks as a pointed DAG (\cref{prop:flow-matching-condition-state-flow-pF}), where the flow matching condition is never checked at the terminal state $\terminal$ since it is an abstract state outside of the state space $\gS$.%

\subsection{Operations on transitions kernels}
\label{app:operations-transition-kernels}
We saw in \cref{def:markov-kernel} that a transition kernel $\kappa$ was defined with a single measurable state space $(\gS, \Sigma)$. This effectively means that we are transitioning from an element $s\in\gS$ to another element of $\gS$ with measure $\kappa(s, \cdot)$. But we could easily extend this definition to different spaces. In this section, we will use the notations $(\gS, \Sigma_{S})$, $(\gT, \Sigma_{T})$ and $(\gU, \Sigma_{U})$ to denote three measurable spaces (some may be equal); note that in this context $\gT$ denotes a state space, and is not the set of complete trajectories in \cref{def:complete-trajectories}. We first revisit the definition of a transition kernel in this general setting

\begin{definition}[Markov kernel]
    \label{def:generalized-markov-kernel}\index{Kernel!Transition kernel}\index{Kernel!Markov kernel}
    Let $(\gS, \Sigma_{S})$ \& $(\gT, \Sigma_{T})$ be two measurable state spaces. A function $\kappa: \gS \times \Sigma_{T} \rightarrow [0, \infty)$ is called a positive $\sigma$-finite \emph{transition kernel} if
    \begin{enumerate}
        \item For any $B\in\Sigma_{T}$, the mapping $s\mapsto \kappa(s, B)$ is measurable;
        \item For any $s\in\gS$, the mapping $B \mapsto \kappa(s, B)$ is a positive $\sigma$-finite measure on $(\gT, \Sigma_{T})$.
    \end{enumerate}
    Furthermore, if the mappings $\kappa(s, \cdot)$ are probability distributions (\ie $\kappa(s, \gT) = 1$), the transition kernel is called a \emph{Markov kernel}.
\end{definition}

With this broader definition of a transition kernel, we are now ready to define two important operations over kernels: the product and the composition. These operations are used throughout \cref{chap:gflownets-general-state-spaces}, and especially for the practical aspects in \cref{sec:practical-implementation-generalized-gflownets}.

\paragraph{Product kernel} Before defining the product between two kernels, we first need to understand what it means to have a product of measurable spaces. We will say that the product space $\gS \times \gT$ is associated with the \emph{product $\sigma$-algebra} $\Sigma_{S} \otimes \Sigma_{T}$, corresponding to the $\sigma$-algebra of product sets
\begin{equation}
    \gls{productsigmaalgebra} \triangleq \sigma\big(\{B_{S} \times B_{T}\mid B_{S}\in\Sigma_{S}, B_{T}\in\Sigma_{T}\}\big).
    \label{eq:product-sigma-algebra}
\end{equation}
\begin{definition}[Product kernel]
    \label{def:product-kernel}\index{Kernel!Product kernel}
    Let $\kappa_{1}$ be a transition kernel from $\gS$ to $\gT$ (\ie $\kappa_{1}: \gS \times \Sigma_{T} \rightarrow [0, \infty)$), and $\kappa_{2}$ a transition kernel from $\gS \times \gT$ to $\gU$ (\ie $\kappa_{2}: (\gS \times \gT) \times \Sigma_{U} \rightarrow [0, \infty)$). We define the \emph{product kernel} \gls{productkernel} from $\gS$ to $\gT \times \gU$ as
    \begin{equation}
        \big[\kappa_{1}\otimes \kappa_{2}\big](s, B) = \int_{\gT}\kappa_{1}(s, dt)\int_{\gU}\kappa_{2}\big((s, t), du\big)\mathds{1}_{B}(t, u),
        \label{eq:product-kernel}
    \end{equation}
    for $s \in \gS$ and $B \in \Sigma_{T} \otimes \Sigma_{U}$.
\end{definition}
The definition of a product kernel was useful when we defined the Markov kernel of the split chain in \cref{sec:creation-atom-splitting-technique}. In particular, if both $\kappa_{1}$ \& $\kappa_{2}$ are Markov kernel, then it is easy to show that the product $\kappa_{1}\otimes \kappa_{2}$ is also a Markov kernel (over $\gT\times \gU$). This definition naturally extends to the case where $\kappa_{2}$ is a transition kernel from $\gT$ to $\gU$ (since we could define $\bar{\kappa}_{2}((s, t), B) = \kappa_{2}(t, B)$ for all $s\in\gS$). This is especially interesting to define recursive products of the same Markov kernel. Let $P_{F}$ be a Markov kernel over $(\gS, \Sigma_{S})$ (into itself, the setting studied in \cref{chap:gflownets-general-state-spaces}). For $T > 0$, we define the product kernel \gls{productkernelrec} from $\gS$ into $\gS^{T}$ recursively by
\begin{equation}
    P_{F}^{\otimes T+1}(s, B) \triangleq \big[P_{F} \otimes P_{F}^{\otimes T}\big](s, B) = \int_{\gS^{T+1}}P_{F}(s, ds')P_{F}^{\otimes T}(s', ds_{1:T})\mathds{1}_{B}(s', s_{1}, \ldots, s_{T}),
    \label{eq:recursive-product-kernel}
\end{equation}
where $s\in\gS$ and $B \in \Sigma_{S}^{\otimes T}$ (an immediate generalization of \cref{eq:product-sigma-algebra}), and where $P_{F}^{\otimes 0}(s, B) = \delta_{s}(B)$ is the Dirac measure at $s$. This product Markov kernel is used in the definition of the generalized trajectory balance condition in \cref{thm:generalized-trajectory-balance}. Finally, we can also define the product of a measure $\nu$ over $\gS$ with a transition kernel $\kappa$ from $\gS$ to $\gT$ as 
\begin{equation}
    \big[\nu \otimes \kappa\big](B) = \int_{\gS}\nu(ds)\int_{\gT}\kappa(s, dt)\mathds{1}_{B}(t),
    \label{eq:product-measure-kernel}
\end{equation}
where $B \in \Sigma_{T}$; this is a consequence of \cref{def:product-kernel}, since a measure $\nu$ can be seen as a (trivial) transition kernel that ignores where it transitions from ($\widebar{\nu}(s, B_{S}) = \nu(B_{S})$, for $s\in\gS$ and $B_{S}\in\Sigma_{S}$). This is necessary in our definition of a measurable pointed graph in \cref{def:measurable-pointed-graph} to guarantee that the reference kernels are ``inverse'' of one another.

\paragraph{Composition kernel} In addition to the product kernel, we can also define another operation between two kernels called the \emph{composition kernel}.
\begin{definition}[Composition kernel]
    \label{def:composition-kernel}\index{Kernel!Composition kernel}
    Let $\kappa_{1}$ be a transition kernel from $\gS$ to $\gT$ (\ie $\kappa_{1}: \gS \times \Sigma_{T} \rightarrow [0, \infty)$), and $\kappa_{2}$ a transition kernel from $\gS\times \gT$ to $\gU$ (\ie $\kappa_{2}: (\gS \times \gT)\times \Sigma_{U} \rightarrow [0, \infty)$). We define the \emph{composition kernel} \gls{compositionkernel} from $\gS$ to $\gU$ as
    \begin{equation}
        \big[\kappa_{1}\cdot \kappa_{2}\big](s, B) = \int_{\gT}\kappa_{1}(s, dt)\int_{\gU}\kappa_{2}\big((s, t), du\big)\mathds{1}_{B}(u),
        \label{eq:composition-kernel}
    \end{equation}
    for $s\in\gS$ and $B\in \Sigma_{U}$.
\end{definition}
Similar to the product kernel, it is easy to show that if both $\kappa_{1}$ \& $\kappa_{2}$ are Markov kernels, then $\kappa_{1}\cdot \kappa_{2}$ is also a Markov kernel. Aligned with our discussion for product kernels above, it is also easy to define the composition kernel if $\kappa_{2}$ is a transition kernel from $\gT$ to $\gU$. This allows us in particular to also have a recursive definition of a transition kernel (similar to \cref{eq:recursive-product-kernel}). Let $\kappa$ be a transition kernel over $\gS$ (into itself). For $T > 0$, we define the composition kernel \gls{compositionkernelrec} from $\gS$ into itself recursively by
\begin{equation}
    \kappa^{T+1}(s, B) \triangleq \big[\kappa \cdot \kappa^{T}\big](s, B) = \int_{\gS}\kappa(s, ds')\int_{\gS}\kappa^{T}(s', ds'')\mathds{1}(s''\in B),
    \label{eq:recursive-composition-kernel}
\end{equation}
where $s\in \gS$ and $B\in\Sigma_{S}$, and where $\kappa^{0}(s, B) = \delta_{s}(B)$ is the Dirac measure at $s$. This definition is useful throughout \cref{sec:practical-implementation-generalized-gflownets}, and in particular when we introduce the notion of \emph{finitely absorbing} measurable pointed graph \cref{eq:finitely-absorbing}, and to define the invariant measure of $\widebar{P}_{F}$ in the generalized trajectory balance condition in \cref{thm:generalized-trajectory-balance}.

\paragraph{Marginalization} The product kernel in \cref{eq:recursive-product-kernel} and the composition kernel \cref{eq:recursive-composition-kernel} are related to one another by \emph{marginalization}. We will use this marginalization property in the proof of the generalized trajectory balance condition \cref{thm:generalized-trajectory-balance}.
\begin{proposition}
    \label{prop:marginalization-product-kernel-composition-kernel}
    Let $P_{F}$ be a Markov kernel over a measurable space $(\gS, \Sigma_{S})$ (into itself), and let $T > 0$. The marginalization of the product kernel $P_{F}^{\otimes T+1}$ corresponds to the composition kernel: for $s\in \gS$ and $B \in \Sigma_{S}$,
    \begin{equation}
        P_{F}^{T}(s, B) = \int_{\gS}\mathds{1}_{B}(s'')\int_{\gS^{T-1}}P_{F}^{\otimes T}(s, ds'ds''),
        \label{eq:marginalization-product-kernel-composition-kernel}
    \end{equation}
    where the outer integral is over $s''$ and the inner one over $s'$.
\end{proposition}
\begin{proof}
    For $B \in \Sigma_{S}$, we will consider the product set $\gS^{T-1} \times B$, which is an element of $\Sigma_{S}^{\otimes T}$. Another way to write \cref{eq:marginalization-product-kernel-composition-kernel} is therefore as $P_{F}^{T}(s, B) = P_{F}^{\otimes T}(s, \gS^{T-1}\times B)$. We will prove this statement by induction.

    \emph{Base case:} If $T = 1$, then we clearly have $P_{F}^{\otimes 1}(s, B) = P_{F}^{1}(s, B) = P_{F}(s, B)$. This satisfies the statement above, since $\gS^{0} = \emptyset$.

    \emph{Induction step:} Suppose that $P_{F}^{T}(s, B) = P_{F}^{\otimes T}(s, \gS^{T-1}\times B)$ for some $T$. Then using the recursions \cref{eq:recursive-product-kernel} \& \cref{eq:recursive-composition-kernel}, we get
    \begin{align}
        P_{F}^{\otimes T+1}(s, \gS^{T}\times B) &= \int_{\gS^{T+1}}P_{F}(s, ds')P_{F}^{\otimes T}(s', ds_{1:T})\mathds{1}_{\gS^{T}\times B}(s', s_{1}, \ldots, s_{T})\\
        &= \int_{\gS}P_{F}(s, ds')\int_{\gS^{T}}P_{F}^{\otimes T}(s', ds_{1:T})\mathds{1}_{\gS^{T-1}\times B}(s_{1}, \ldots, s_{T})\\
        &= \int_{\gS}P_{F}(s, ds')P_{F}^{\otimes T}(s', \gS^{T-1}\times B)\\
        &= \int_{\gS}P_{F}(s, ds')P_{F}^{T}(s', B)\\
        &= \int_{\gS}P_{F}(s, ds')\int_{\gS}P_{F}^{T}(s', ds'')\mathds{1}_{B}(s'')\\
        &= P_{F}^{T+1}(s, B).
    \end{align}
    This concludes the proof.
\end{proof}

%% file: appendix/02_Bayesian_Structure_Learning.tex
\chapter{Bayesian Structure Learning}
\label{app:bayesian-structure-learning}

\section{Bayesian Structure Learning with Generative Flow Networks}
\label{app:dag-gflownet}

\subsection{Bayesian scores}
\label{app:bayesian-scores}
In \cref{chap:dag-gflownet}, we considered two types of Bayesian networks: either (1) linear-Gaussian models, or (2) discrete Bayesian networks with a Dirichlet prior over parameters. These two settings give rise to two different ways to compute the marginal likelihood $P(\gD\mid G)$, respectively the \emph{BGe score} and the \emph{BDe score} that we will define here. Both of these have the remarkable property of giving the same value for two Markov equivalent graphs \citep{koller2009pgm}. Recall that in general, the marginal likelihood is given by
\begin{equation}
    P(\gD\mid G) = \int_{\theta}P(\gD\mid \theta, G)P(\theta\mid G)d\theta,
\end{equation}
which is intractable, unless we put some careful assumptions on the likelihood model and the prior.

\paragraph{BGe score} The \emph{Bayesian Gaussian equivalent score}\glsadd{bgescore} \citep{geiger1994bge,kuipers2014bgeaddendum} can be used to compute the marginal likelihood of a linear-Gaussian model, as described in \cref{ex:linear-gaussian-model}. We place a Normal prior over the linear parameters, and a Wishart prior over the variance parameters. If the data is $\gD = \{\vx^{(n)}\}_{n=1}^{N}$, the (log-)marginal likelihood is given by
\begin{equation}
    \log P(\gD\mid G) = \sum_{i=1}^{d} C + \log \frac{\Gamma\big(\frac{N + \alpha_{w} - d + l + 1}{2}\big)}{\Gamma\big(\frac{\alpha_{w} - d + l + 1}{2}\big)} + lt + \frac{1}{2}\log \frac{|\mR_{\pi_{i}}|^{N + \alpha_{w} - d + l}}{|\mR_{i}|^{N + \alpha_{w} - d + l + 1}},
    \label{eq:bge-score}
\end{equation}
where $\Gamma$ is the Gamma function, $C$ is a constant term independent of $i$, $l$ is the number of parents of $X_{i}$ in $G$, and $\mR$ is the matrix defined by
\begin{equation}
    \mR = t\mI + \sum_{n=1}^{N}(\vx^{(n)} - \bar{\vx})^{\top}(\vx^{(n)} - \bar{\vx}) +  \frac{N\alpha_{\mu}}{N + \alpha_{\mu}}\bar{\vx}^{\top}\widebar{\vx},
\end{equation}
where $\bar{\vx}$ is the average observation in $\gD$, and $t = \alpha_{\mu}(\alpha_{w} - d - 1)/(\alpha_{\mu} + 1)$ is a constant scalar. The matrix $\mR_{\pi_{i}}$ corresponds to the sub-matrix $\mR$ where only the rows and columns of $\parents_{G}(X_{i})$ are selected; similarly, $\mR_{i}$ is the sub-graph where the rows and columns of $\parents_{G}(X_{i})$ and $X_{i}$ are selected. The strength of the prior is defined by the hyperparameters $\alpha_{\mu} > 0$ and $\alpha_{w} > d + 1$.

\paragraph{BDe score} The \emph{Bayesian Dirichlet equivalent score}\glsadd{bdescore} \citep{heckerman1995bde} can be used to compute the marginal likelihood of a Bayesian network over discrete random variables (with Categorical distributions), whose parameters organized as conditional probability \emph{tables} have a Dirichlet prior. We will assume for simplicity that all the random variables $X_{1}, \ldots, X_{d}$ may take values in $\{1, \ldots, K\}$. In that case, the (log-)marginal likelihood is given by
\begin{equation}
    \log P(\gD\mid G) = \sum_{i=1}^{d}\sum_{\vu_{i}\in\Pi^{G}_{i}}\log \frac{\Gamma\big(N'[\vu_{i}]\big)}{\Gamma\big(N'[\vu_{i}] + N[\vu_{i}]\big)}\sum_{k=1}^{K}\log \frac{\Gamma\big(N'[k, \vu_{i}] + N[k, \vu_{i}]\big)}{\Gamma\big(N'[k, \vu_{i}]\big)},
    \label{eq:bde-score}
\end{equation}
where $\Pi^{G}_{i} = \mathrm{Val}(\parents_{G}(X_{i})) \simeq K^{|\parents_{G}(X_{i})|}$ is the set of all the possible values the parents of $X_{i}$ in $G$ may take, and $\Gamma$ is the Gamma function. $N[k, \vu_{i}]$ represents the number of observations in $\gD$ where $X_{i} = k$ \& $\parents_{G}(X_{i}) = \vu_{i}$, and $N[\vu_{i}] = \sum_{k=1}^{K}N[k, \vu_{i}]$. The strength of the prior in the BDe score is controlled by a scalar hyperparameter $N'$ called the \emph{equivalent sample size}, from which we define $N'[k, \vu_{i}] = N'\cdot P'\big(X_{i} = k, \parents_{G}(X_{i}) = \vu_{i}\big)$ where $P'$ is a joint distribution over $X_{1}, \ldots X_{d}$ that can be defined as a Bayesian network, and is often chosen to be the uniform distribution \citep{buntine1991bdeu}; similarly, we have $N'[\vu_{i}] = \sum_{k=1}^{K}N'[k, \vu_{i}]$.

\subsection{Estimation of the terminating state probability}
\label{app:estimation-terminating-state-probability}
In \cref{sec:estimation-terminating-state-probability}, we saw a method to approximate the terminating state probability of a specific graph $G$ using a combination of beam-search to extract high-scoring complete trajectories, and Monte-Carlo. We provide in \cref{alg:beam-search} a pseudo-code of the beam-search procedure for completeness.
\begin{algorithm}[t]
    \caption{Beam-search algorithm for finding high-probability trajectories $G_{0} \rightsquigarrow G$.}
    \label{alg:beam-search}
    \begin{algorithmic}[1]
        \Require A DAG $G$ with $K$ edges, the forward transition probability $P_{F}$ of the GFlowNet, the beam width $B$.
        \Ensure $B$ complete trajectories $\{\tau_{m}\}_{m=1}^{B}$ with high-probability $P_{F}(\tau_{m})$.
        \State Initialization of the trajectories: $G_{0}^{(m)} \leftarrow G_{0}$ \Comment{$\forall m \in \{1, \ldots, B\}$}
        \State Initial log-probabilities: $\ell_{0}^{(m)} \leftarrow 0$ \Comment{$\forall m \in \{1, \ldots, B\}$}
        \For{$k=0$ \textbf{to} $K-1$}
            \For{$m=1$ \textbf{to} $B$}
                \ForAll{edge $e \in G$}
                    \State $s(m, e) \leftarrow \ell_{k}^{(m)} + \log P_{F}(G^{(m)}_{k} \cup \{e\}\mid G^{(m)}_{k})$
                \EndFor
                \Statex \vspace*{-0.8em}
            \EndFor
            \For{$m=1$ \textbf{to} $B$}
                \State Find $(b, e)$ such that $s(b, e)$ is the $m$-th largest score
                \State Save the partial trajectory $b$: $\widetilde{G}^{(m)}_{0:k} \leftarrow G^{(b)}_{0:k}$
                \State Add the new graph to the trajectory $m$: $G^{(m)}_{k+1} \leftarrow G^{(b)}_{k} \cup \{e\}$
                \State Update the log-probabilities: $\ell_{k+1}^{(m)} \leftarrow s(b, e)$
            \EndFor
            \State Re-assign the partial trajectories: $G^{(m)}_{0:k} \leftarrow \widetilde{G}^{(m)}_{0:k}$ \Comment{$\forall m \in \{1, \ldots, B\}$}
        \EndFor
        \State \Return trajectories $\{\tau_{m} = G^{(m)}_{0:K}\}_{m=1}^{B}$
    \end{algorithmic}
\end{algorithm}

\section{Joint Inference of Structure and Parameters}
\label{app:jsp-gfn}

\subsection{Sub-trajectory balance condition}
The following lemma shows that if the sub-trajectory balance condition in \cref{eq:sub-trajectory-balance-condition-jsp-gfn} is satisfied for all undirected paths of length 3 in JSP-GFN, then the different between the log-reward and the forward transition log-probability is independent of the parameters $\theta$. This is useful to show that satisfying these sub-trajectory balance condition alone is sufficient to have JSP-GFN sample proportionally to the reward (unlike the general case in \cref{sec:sub-trajectory-balance-condition}).

\begin{lemma}
    \label{lem:subtb-3-constant-function}
    Let $\gG$ be the GFlowNet described in \cref{sec:jsp-gfn-structure-gflownet}, with a reward function $R$, and a forward transition probability $P_{F}$. For two DAGs $G$ \& $G'$, if the sub-trajectory balance condition \cref{eq:sub-trajectory-balance-condition-jsp-gfn} is satisfied for all undirected paths of the form $(G, \theta) \leftarrow (G, \ast) \rightarrow (G', \ast) \rightarrow (G', \theta')$ (for any $\theta$ \& $\theta'$), then the following function
    \begin{equation}
        f_{G}(\theta) \triangleq \log R(G, \theta) - \log P_{F}(\theta\mid G)
        \label{eq:subtb-3-constant-function}
    \end{equation}
    is constant (\ie independent of $\theta$, but a constant depending on $G$).
\end{lemma}

\begin{proof}
    Let $\theta' \in \Theta_{G'}$ be a fixed set of parameters associated with $G'$, and $\theta$ \& $\bar{\theta} \in \Theta_{G}$ two arbitrary sets of parameters associated with $G$. We can write the sub-trajectory balance condition for the undirected paths between $(G, \theta)$ \& $(G', \theta')$ and between $(G, \bar{\theta})$ \& $(G', \theta')$:
    \begin{align}
        R(G', \theta')P_{B}(G\mid G')P_{F}(\theta\mid G) &= R(G, \theta)P_{F}(G'\mid G)P_{F}(\theta'\mid G')\\
        R(G', \theta')P_{B}(G\mid G')P_{F}(\bar{\theta}\mid G) &= R(G, \bar{\theta})P_{F}(G'\mid G)P_{F}(\theta'\mid G')
    \end{align}
    Using these two equations, we have
    \begin{equation}
        \frac{R(G, \theta)}{P_{F}(\theta\mid G)} = \frac{P_{F}(G'\mid G)P_{F}(\theta'\mid G')}{R(G', \theta')P_{B}(G\mid G')} = \frac{R(G, \bar{\theta})}{P_{F}(\bar{\theta}\mid G)}.
    \end{equation}
    Taking the $\log$ of both sides of this equation, we can conclude that $f_{G}(\theta)$ in \cref{eq:subtb-3-constant-function} is constant since this is true for any arbitrary $\theta$ \& $\bar{\theta}$.
\end{proof}